%% file: report.tex
\documentclass[10pt,letterpaper]{article}

\usepackage{booktabs}
\usepackage{amsmath}
\usepackage{amssymb}
\usepackage{amsthm}
\usepackage{array}
\usepackage{multirow}
\usepackage{tabularx}
\usepackage{graphicx}
\usepackage{fvextra}
\usepackage{placeins}
\usepackage{float}
\usepackage{multicol}
\usepackage{mmla_technical_report}
\usepackage{orcidlink}
\graphicspath{{./figures/}}

\newcommand{\afrt}{\textsc{AFRT}}
\newcommand{\mmla}{\textsc{MMLA}}
\newcommand{\pdsa}{\textsc{PDSA}}
\newcommand{\pdsaj}{\textsc{PDSA-J}}
\newcolumntype{Y}{>{\raggedright\arraybackslash}X}
\theoremstyle{definition}
\newtheorem{reportdefinition}{Definition}[section]
\input{theory_family/integrated_preamble}

\newcommand{\mmlaboundary}{%
  \par\medskip\noindent
  At the September 13, 2026 evidence cutoff, MMLA has conditional formal results and protocol-specific component evidence, including restricted latent generation, update, and restoration. Generalized continuous-event latent readout, full authoritative-row conformance, a strict within-problem policy-only effect, predictive admission, and end-to-end PDSA benefit remain unvalidated.
  \par\medskip}
\let\originalthebibliography\thebibliography
\renewcommand{\thebibliography}[1]{%
  \originalthebibliography{#1}%
  \setlength{\itemsep}{-1pt}%
  \setlength{\parsep}{0pt}%
}

\title{MMLA: Memory-Mediated Learning Architecture\\for Predictive Dual-State Adaptation}
\author{%
\textbf{Junyi Zou}\textsuperscript{1,*\,\orcidlink{0009-0009-1367-7428}}\quad\textbullet\quad
\textbf{Avrova Donz}\textsuperscript{1,2,*\,\orcidlink{0009-0009-0100-0719}}\\[-0.06em]
{\fontsize{7.3pt}{8.2pt}\selectfont \textsuperscript{1}MMLA-org\quad
\textsuperscript{2}Communication University of China (CUC), No. 1 Dingfuzhuang East Street, Chaoyang District, Beijing 100024, China}\\[-0.10em]
{\small \textsuperscript{*}Equal contribution}\\[-0.10em]
{\scriptsize Email:\enspace
\href{mailto:lior.j.zou@gmail.com}{\nolinkurl{lior.j.zou@gmail.com}}
\enspace$\cdot$\enspace
\href{mailto:avrovadonz@icloud.com}{\nolinkurl{avrovadonz@icloud.com}}}}
\date{September 14, 2026}

\hypersetup{
  pdftitle={MMLA: Memory-Mediated Learning Architecture for Predictive Dual-State Adaptation},
  pdfauthor={Junyi Zou and Avrova Donz},
  pdfsubject={Complete technical report on predictive dual-state adaptation: conditional theory, component evidence, latent readout, and unresolved system claims},
  pdfkeywords={memory-mediated learning architecture, predictive dual-state adaptation, reasoning-time training, authoritative memory, predictive admission},
  pdfcreator={MMLA Technical Report}
}

\begin{document}
\makemmlatitle{
Memory-Mediated Learning Architecture (MMLA) separates slow base parameters theta, a bounded numerical policy carrier Phi, and a bounded authoritative memory M. Predictive Dual-State Adaptation (PDSA) lets feedback update Phi while one problem remains active and lets a trusted lifecycle atomically commit one typed row or exact NULL. Later reasoning may read both states, but their writers, resets, rollback domains, and ledgers remain distinct. Realized futures supervise values only during training; deployment is causal and future-blind.

\medskip\noindent We give conditional theory and falsifiable contracts for reasoning-time updates, completed-segment consolidation, predictive admission, authoritative memory, and dual-state attribution. Assumptions, counterexamples, capacity and cost ledgers, recovery duties, and identifying experiments are explicit; these are not implementation guarantees.

\medskip\noindent Controlled studies show exact lifecycle execution on 300/300 held-out records for each of three seeds, calibrated retrieval gains over frozen-hidden dense and BM25 baselines, and exact typed anchor-filler transport on 240/240 held-out records per seed. These validate restricted components, not natural-language memory management or complete PDSA.

\medskip\noindent Post-training studies retain positive and negative evidence. Later protocols obtain restricted readout progress, but a nine-trajectory comparison finds that matched latent readout, a full-width bridge, and bridge plus frozen text-teacher alignment all fail continuous-event qualification across both task families. A separately adapted text reference and restoration checks pass. At the September 13, 2026 evidence cutoff, no strict policy-only reasoning-time-training effect, predictive-admission oracle margin, learned future-blind admission policy, or policy-by-memory factorial advantage is established.
}

\begin{center}
\begin{minipage}{0.92\textwidth}
\centering
\includegraphics[width=\linewidth]{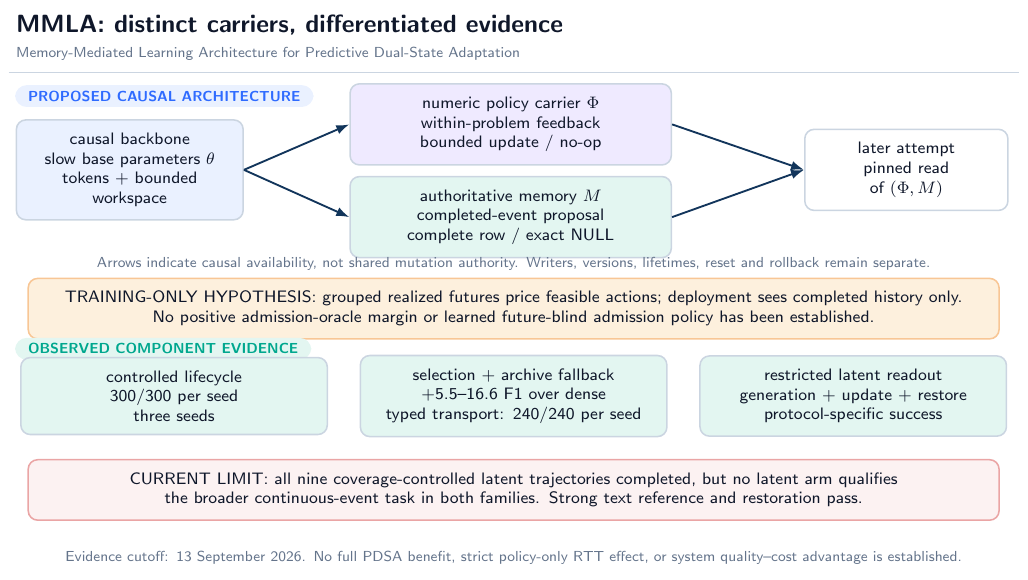}
\captionsetup{hypcap=false}
\captionof{figure}{\textbf{Proposal, evidence, and present boundary in one view.} The causal path is a design contract; future-valued supervision is training-only. Controlled component successes and restricted readout progress coexist with unresolved generalized latent readout and an unvalidated complete PDSA loop.}
\label{fig:overview}
\end{minipage}
\end{center}

\mmlanote{This archival revision is dated September 14, 2026 and freezes scientific evidence at September 13, 2026, 10:11:45 UTC. The revision date, evidence cutoff, and eventual arXiv announcement date are distinct. Earlier GitHub and arXiv PDFs remain historical artifacts (Appendix~\ref{app:version-method}); the accompanying manifest states exactly which reconstruction materials are included and which scientific artifacts are unavailable.}

\input{theory_family/integrated_overview}

\clearpage
% Keep the printed contents page as a high-level map: parts and sections only.
% Subsections remain available through the PDF bookmarks and the section
% headings themselves.  Two columns keep this 104-section report navigable
% without delaying the Introduction.
\setcounter{tocdepth}{1}
\hypersetup{bookmarksdepth=2}
\begingroup
\small
\makeatletter
\renewcommand{\@pnumwidth}{2.6em}
\renewcommand{\@tocrmarg}{3.3em}
% Article's default 1.5em number box is too narrow once section numbers reach
% three digits.  Widen it locally for the printed contents page only.
\renewcommand*\l@section[2]{%
  \ifnum \c@tocdepth >\z@
    \addpenalty\@secpenalty
    \addvspace{1.0em \@plus\p@}%
    \setlength\@tempdima{2.35em}%
    \begingroup
      \parindent \z@ \rightskip \@pnumwidth
      \parfillskip -\@pnumwidth
      \leavevmode \bfseries
      \advance\leftskip\@tempdima
      \hskip -\leftskip
      #1\nobreak\hfil
      \nobreak\hb@xt@\@pnumwidth{\hss #2%
                                 \kern-\p@\kern\p@}\par
    \endgroup
  \fi}
\makeatother
\begin{multicols}{2}
\tableofcontents
\end{multicols}
\endgroup
\clearpage

\section{Introduction}

We retain the project name \mmla{} while making its scope explicit: \mmla{} means \emph{Memory-Mediated Learning Architecture}. Earlier versions used \emph{Memory-Managed Long-Context Attention}; that phrase now names a historical formulation rather than the architecture's current definition. The architecture is mixer-agnostic and is organized around \emph{Predictive Dual-State Adaptation} (\pdsa{}).

Efficient long-context modeling has progressed from the Transformer \cite{transformer2017} through linear attention, structured state spaces, recurrent and convolutional hybrids, and sparse attention \cite{lineartransformer2020,s4_2021,retnet2023,rwkv2023,mamba2023,hyena2023,gdn2024,bigbird2020,nsa2025,deepseekv322025}. These methods reduce the per-token state size or the number of attended positions. They do not, by themselves, answer a different question: when should an observation change what the model will carry forward?

A million-token prompt can still contain mutually inconsistent versions of the same fact, preserve an obsolete value, or force the model to repeatedly process material whose relevance was impossible to know when it arrived. Conversely, a small persistent state can be useful if it supports reliable revision, deletion, abstention, and later reuse. The central object of this report is therefore not an unlimited context window. It is an architecture in which two bounded state carriers mediate learning at different time scales: a numerical policy carrier for within-problem adaptation and an authoritative memory carrier for controlled persistence.

Predictive writing poses a difficult decision beyond reliable reading. Our current experiments also show that latent readout itself can remain the practical bottleneck. Immediate salience is a poor proxy for future utility: an ordinary sentence may become important only after later events, while a vivid sentence may never be used again. Training data, however, contains the future that inference lacks. This suggests a division of labor. Real subsequent trajectories can price alternative memory actions; a deployable controller must predict their \emph{conditional expected} cost from completed history alone. Predictive information and information-bottleneck principles supply the target: compress the past while retaining information about the future \cite{tishby2000ib,bialek2001predictive}. The expectation is essential. A policy fitted to the action preferred by one realized continuation need not minimize risk over the future distribution faced at deployment.

The memory plane uses \emph{predictive memory consolidation}. The model first predicts causally. After a local event, sentence, or chunk has fully arrived, a small bidirectional module may reinterpret that completed segment without changing any output already emitted. It eventizes the segment into a bounded ordered list. For each event, a target-conditioned neural proposal and trusted assembler construct a complete candidate row; a controller chooses one resident target to overwrite or chooses NULL. The policy plane may separately update a declared numerical carrier from authenticated feedback before the problem ends. \pdsa{} composes these planes through distinct state interfaces rather than a single hidden state. Insert, update, consolidating update, eviction, deletion, protection, and stale rejection are lifecycle interpretations of the memory operator. The observation horizon and the mixture of local attention, recurrent state, memory read, and policy-adaptation modules are deployment choices rather than architectural constants.

Earlier versions of this report treated lifecycle, fallback, and relational compilation as the endpoint. We now treat them as evidence-bearing components of a broader memory architecture. Controlled lifecycle text exercises write--manage--read--fallback--generate behavior. Held-out multi-hop QA measures the point at which a query-independent writer should abstain. A third study isolates role binding when content and position are unreliable. The new formulation connects these pieces through future-valued consolidation while preserving every earlier result and failure.

\textbf{Contributions.}
\begin{enumerate}
  \item We define MMLA as Memory-Mediated Learning Architecture and \pdsa{} as a typed composition of a short-lived numerical policy carrier and a persistent authoritative memory carrier. Joint reads are allowed; ownership, reset, rollback, and evidence are not conflated (\S\ref{sec:pdsa-identity}).
  \item We specify five state layers---causal token state, completed-segment workspace, policy-adaptation carrier, authoritative memory, and slower base parameters---with distinct writers, lifetimes, and reset domains. The only authoritative memory transition is one complete-row overwrite or NULL per ordered event (\S\ref{sec:predictive-memory}).
  \item We define state-change selection as conditional action-value estimation. Grouped continuations with identical observed history provide counterfactual branch risks; a causal value predictor sees only completed history and chooses the lowest-risk feasible action. We state the no-leakage interface, information-theoretic objective, stability limits, and system boundary explicitly.
  \item We separate policy-effect identification, representation qualification, and predictive overwrite learning. Historical frozen-checkpoint interfaces fail even on fitted support. Subsequent post-training establishes restricted candidate and generated-answer following under different model and protocol lineages, while generalized continuous-event readout remains below qualification (\S\ref{sec:posttraining-evidence}--\ref{sec:coverage-readout}). None of these studies identifies a strict policy-only RTT effect or conditional expected-risk admission benefit.
  \item We retain component evidence for bounded lifecycle execution (\S\ref{sec:tracka}), calibrated selection and fallback (\S\ref{sec:trackb}--\ref{sec:boundary}), and typed relational binding (\S\ref{sec:afrt}); and report all nine completed trajectories of the coverage-controlled bridge/teacher comparison, including the registered negative outcome, uncertainty, denominator, and cost boundaries.
\end{enumerate}

The empirical evidence supports the reported component results; the relative benefits of dual-state adaptation and predictive admission remain to be established. The first two studies use frozen language models, and the relational study trains a compact synthetic compiler. A complete claim requires an independently identified policy-only effect, a semantically usable current-state interface, a positive same-checkpoint expected-risk intervention, and then the complete policy-by-memory factorial; compute-matched pretraining is conditional on those results.

\begin{figure}[H]
\centering
\includegraphics[width=\linewidth]{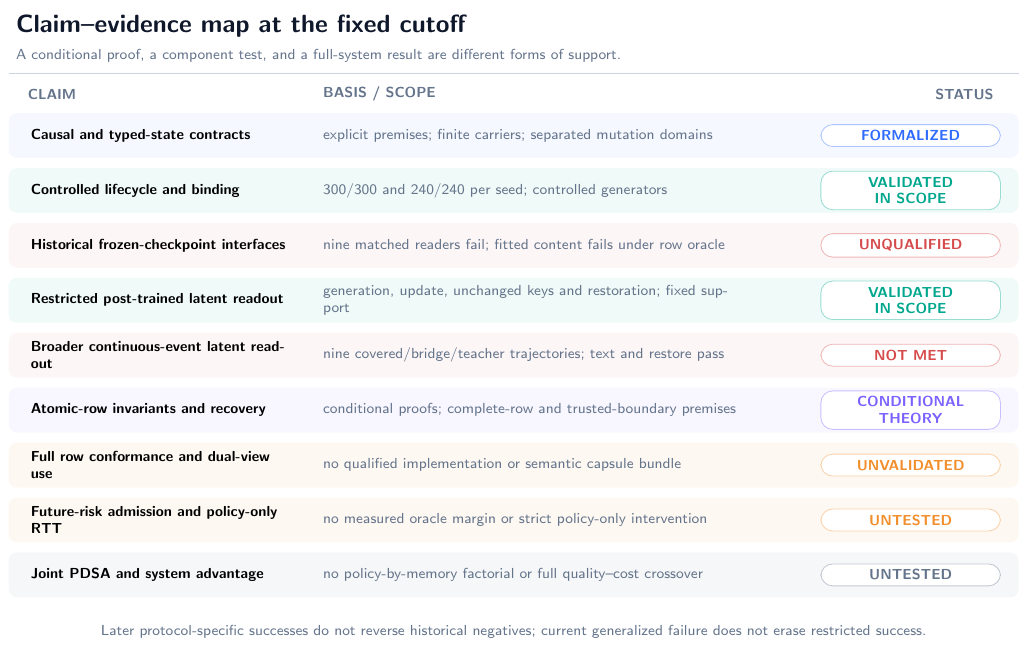}
\caption{\textbf{Claim--evidence matrix.} Status belongs to the strongest justified statement, not to the architecture as a whole. ``Implemented'' denotes a mechanical ancestor; ``validated'' requires evidence admitted under the frozen protocol.}
\label{fig:evidence-map}
\end{figure}

\section{MMLA Identity: Predictive Dual-State Adaptation}
\label{sec:pdsa-identity}

\subsection{Architecture definition}

\begin{reportdefinition}[Memory-Mediated Learning Architecture]
An \mmla{} system is a causal learner in which later behavior can be mediated by at
least two explicitly typed state carriers: a bounded numerical policy-adaptation
carrier $\Phi_t\in\mathcal P$ and a bounded authoritative memory carrier
$M_t\in\mathcal M$. The carriers have separately declared writers, readers,
privileges, lifetimes, version lines, reset operations, rollback domains, and cost
ledgers. A compatible pinned read may expose both carriers to one attempt, but does
not merge their mutation authority.
\end{reportdefinition}

The word \emph{mediated} is deliberate. The architecture does not require every
behavior change to be stored as a human-readable row, and it does not treat every
numerical update as a factual memory. It requires the path from observation to later
behavior to identify which carrier changed, which later computation read it, and
which receipt supports that claim. The historical ``Memory-Managed Long-Context
Attention'' formulation is one possible mixer realization inside this broader
architecture, not the definition of the architecture as a whole.

\begin{reportdefinition}[Predictive Dual-State Adaptation]
\pdsa{} is an MMLA execution regime in which authorized transitions of a bounded
policy carrier $\Phi$ and a bounded authoritative memory carrier $M$ are selected or
evaluated according to their conditional consequences for later behavior, while every
deployed decision is measurable with respect to the causal information available before
that transition. The two carriers retain separate writers, transition algebras,
versions, ledgers, lifetimes, reset operations, rollback domains, and failure semantics.
A compatibility-verified coordination manifest may bind their versions for a later
read, but it grants neither carrier-write authority nor cross-carrier rollback.
\end{reportdefinition}

\begin{reportdefinition}[Optional joint predictive-control profile]
The optional \pdsaj{} profile evaluates or selects a paired action for a completed event
$e$:
\begin{equation}
d_e=(u_e,a_e)\in
(\mathcal U_e\cup\{\mathsf{no\mbox{-}op}_{\Phi}\})
\times
(\mathcal A_e\cup\{\mathsf{no\mbox{-}op}_{M}\}),
\label{eq:pdsa-joint-action}
\end{equation}
where $u_e$ is a bounded policy-carrier action and $a_e$ is a typed authoritative
memory action. The expected consequence of $d_e$ is supervised from training-only
future outcomes, while the deployed selector receives only the causal information
available at the event boundary. Each carrier validates, commits, rejects, resets,
and rolls back through its own domain. A later coordination manifest may bind the two
carrier-local receipts, but it does not grant joint write or rollback authority.
\end{reportdefinition}

The memory no-op is the exact $\mathrm{NULL}$ action developed throughout this
report. The policy no-op is equally important: it certifies that feedback was
observed but no declared numerical carrier changed. Neither no-op can be replaced by
a small nonzero gate unless the implementation proves that no corresponding state bit
or optimizer bit changed.

\begin{table}[H]
\centering
\small
\setlength{\tabcolsep}{4pt}
\begin{tabularx}{\linewidth}{p{0.18\linewidth}YY}
\toprule
Property & Policy-adaptation carrier $\Phi$ & Authoritative memory carrier $M$ \\
\midrule
Representation & Numerical adapter, TTT layer, fast weights, or another fixed-budget declared object & Typed complete rows with canonical and neural views \\
Primary role & Change how later attempts reason within a still-active problem & Decide what observed content may persist with authority \\
Writer & Feedback-bound update rule & Neural proposer plus trusted assembler and transaction controller \\
No-op & Exact update rejection / unchanged carrier receipt & Exact NULL / rejection receipt \\
Lifetime & Usually attempt-, episode-, or session-bounded & Declared row lifetime; may cross episodes within authorization scope \\
Reset / rollback & Exact carrier and optimizer reset to named epoch or snapshot & Versioned row rollback with fresh epoch and ledger parent \\
Evidence question & Did feedback cause a later attempt to read a changed policy? & Did an admitted row improve later behavior without harmful persistence? \\
\bottomrule
\end{tabularx}
\caption{The two PDSA carriers can affect one later attempt but are not aliases.}
\label{tab:pdsa-carriers}
\end{table}

\subsection{Causal schedule and predictive objective}

\paragraph{Historical notation and the object actually tested.}
The earlier slow-parameter/fast-memory description used $(\theta,M)$ to contrast
long-lived learned weights with editable resident state. The present contract
refines that description: $\theta$ denotes the slow base parameters,
$\Phi$ a separately declared bounded numerical policy carrier, and $M$ the
independent authoritative memory. A numerical adapter may implement $\Phi$ only
when its writer, lifetime, feedback boundary, later-read witness, and reset domain
satisfy the policy contract. Training a readout adapter offline and freezing it
for evaluation is not, by itself, a within-question $\Phi$ update. Likewise,
contextual encoder states or temporary native-embedding diagnostics are not
additional authoritative carriers. Historical experiments retain their original
objects and endpoints; the notation refinement does not retroactively turn a
memory-only result into a joint $(\Phi,M)$ intervention.

\paragraph{Evidence cutoff.}
Empirical statements in this report refer to work completed by September 13,
2026, 10:11:45 UTC. The full manifest binds this fixed evidence slice separately
from the later document build. Subsequent work cannot silently enter a table
described as current here.

For one unresolved problem, the logical schedule is:
\begin{enumerate}
  \item an attempt reads a pinned pair $(v_t^{\Phi},v_t^M)$ and generates causally;
  \item feedback and completed local observations receive immutable event identities;
  \item the policy path accepts one bounded update or exact no-op and publishes a
  policy receipt;
  \item the memory path eventizes only completed observations, constructs a complete
  target-conditioned row, and accepts one transaction or exact NULL;
  \item a later attempt reads a fresh compatible pair; and
  \item offline evaluators attribute policy, memory, and interaction effects through
  interventions that preserve the two receipt lines.
\end{enumerate}

The deployed action must be measurable with respect to the causal information set
$\mathcal I_e$. Let $F_e$ be grouped future continuations available to an offline
teacher only. For the optional joint-control profile, a complete paired-action risk is
\begin{equation}
\mathcal R_e(d)=
\mathbb E\!\left[
L_{\rm task}(d;F_e)+\lambda_{\rm harm}L_{\rm harm}(d;F_e)
\mid\mathcal I_e\right]
+\langle\boldsymbol\lambda,\mathcal C_e(d)\rangle,
\label{eq:pdsa-joint-risk}
\end{equation}
where $\mathcal C_e$ charges policy backward and optimizer work, reset verification,
candidate construction, complete-row assembly, validation, action scoring,
counterfactual replay, commit/NULL, archive and index traffic, reads, latency, memory,
and energy. A factorized selector may use
\[
q_\theta(u,a\mid\mathcal I_e)
=q_\theta^{\Phi}(u\mid\mathcal I_e)
q_\theta^{M}(a\mid u,\mathcal I_e),
\]
but factorization does not eliminate paired-action cost or the need to estimate an
interaction. Base PDSA does not require every event to be produced by one joint
controller: independently authorized policy and memory transitions remain conforming
when their coordination manifest is verified at read time.

At deployment all $F_e$ features, teacher states, branch labels, and post-boundary
evaluations are amputated. A future-blind claim therefore requires a materialized
feature/data-flow audit, not only the absence of an obvious future-token argument in
one equation.

\subsection{Identification and the experiment ladder}

The first policy experiment disables authoritative memory and compares active-update
real feedback, active-update sham feedback, update-then-reset real feedback,
update-then-reset sham feedback, and context-only real feedback with dummy matched
update compute. Its contrast isolates a declared within-problem policy effect only if
the update carrier, reset domain, read witness, randomization, and experimental unit
are fixed in advance.

The memory path first requires a same-checkpoint semantic interface. Only then may
same-prefix futures establish an expected-risk oracle gap over current-risk,
write-all, no-write, and heuristic policies. Only after that gap exists may a
future-blind selector be trained. The integrated experiment is a policy-by-memory
factorial with cells $(0,0),(1,0),(0,1),(1,1)$ and interaction
\begin{equation}
\tau_{\Phi M}=Y_{11}-Y_{10}-Y_{01}+Y_{00}.
\label{eq:pdsa-interaction}
\end{equation}
We use a loss convention throughout this report: lower loss is better, so a negative
main effect is favorable. Two favorable isolated main effects do not imply
complementarity; on this scale, complementarity additionally requires
$\tau_{\Phi M}<0$ on the registered loss endpoint.

The current evidence has not reached this factorial. Historical frozen-checkpoint
interfaces fail; later post-training establishes restricted readout but not the
broader continuous-event qualification. The strict policy-only RTT intervention
remains unexecuted; no oracle admission margin has been measured; and no learned joint selector exists.
These are separate empirical gates of one architecture.

\section{Memory Plane: Predictive Memory Consolidation}
\label{sec:predictive-memory}

\subsection{Five state layers and distinct time scales}

Let $x_1,x_2,\ldots$ be a token stream. The report distinguishes five state layers:
token-level causal state $h_t$, completed-segment workspace $B_n$, a bounded
policy-adaptation carrier $\Phi_n$, bounded authoritative memory $M_n$, and slower
base parameters $\theta$. The first four can be read by a deployment attempt only
through declared interfaces; their writers, lifetimes, and reset domains remain
distinct. The authoritative resident state is
\begin{equation}
M_n=(r_{n,1},\ldots,r_{n,S})\in\mathcal R_B^S,
\qquad
\mathcal R_B=\mathcal I_{B_i}\times\mathcal C_{L_c}\times\mathcal V_d\times\mathcal K_{d_k}\times\mathcal U_{B_m},
\label{eq:bounded-row-state}
\end{equation}
where $\mathcal I_{B_i}$ is a fixed-width identity field, $\mathcal C_{L_c}$ is a padded canonical encoding over one fixed alphabet $\Sigma$ with at most $L_c$ symbols, $\mathcal V_d$ and $\mathcal K_{d_k}$ are fixed-width neural payload and routing-key spaces, and $\mathcal U_{B_m}$ is a fixed-width metadata record of at most $B_m$ bits. Concretely,
\[
B_c=\left\lceil\log_2(L_c+1)\right\rceil+
L_c\left\lceil\log_2(|\Sigma|+1)\right\rceil,\qquad
B_v=d\,p_v,\qquad B_k=d_k\,p_k,
\]
for fixed payload/key precisions $p_v,p_k$; padding and the stored length make the canonical serialization unique. Aliases have a fixed per-row count and length within $B_c+B_m$; provenance and rollback fields contain bounded identifiers or digests, never an append-only history. The resident table is therefore bounded by
\begin{equation}
\operatorname{bits}(M_n)\le S\bigl(B_i+B_c+B_v+B_k+B_m\bigr)=:B_{\mathrm{resident}},
\label{eq:resident-bit-bound}
\end{equation}
independently of stream length. A full provenance document, superseded canonical value,
rollback snapshot, retrieval corpus, or audit log belongs to a separately charged
external ledger $A_n$, not to $M_n$. If that ledger grows, only resident state---not
the total system---is bounded.

\begin{table}[H]
\centering
\small
\setlength{\tabcolsep}{4pt}
\begin{tabularx}{\linewidth}{@{}p{0.18\linewidth}p{0.28\linewidth}Y Y@{}}
\toprule
Layer & State object & Primary writer & Lifetime / read boundary \\
\midrule
Token state & $h_t$ and local causal activations & causal backbone & one token/attempt; never reads an unobserved future \\
Segment workspace & completed segment $B_n$ and event summary $z_n$ & bounded consolidator after closure & one boundary; may be read only after the segment closes \\
Policy carrier & numerical adapter and optimizer state $\Phi_n$ & feedback-bound policy updater & declared problem/episode scope; separate reset and rollback \\
Authoritative memory & complete typed rows $M_n$ & trusted assembler/committer & declared retention scope; later pinned reads only \\
Slow parameters & base weights $\theta$ & offline or pretraining optimizer & slower training lifecycle; not an implicit memory write \\
\bottomrule
\end{tabularx}
\caption{Five state layers in the architecture. A joint read can expose compatible layers, but no layer inherits another layer's write privilege or lifetime.}
\label{tab:state-scale-map}
\end{table}

\begin{figure}[t]
\centering
\includegraphics[width=\linewidth]{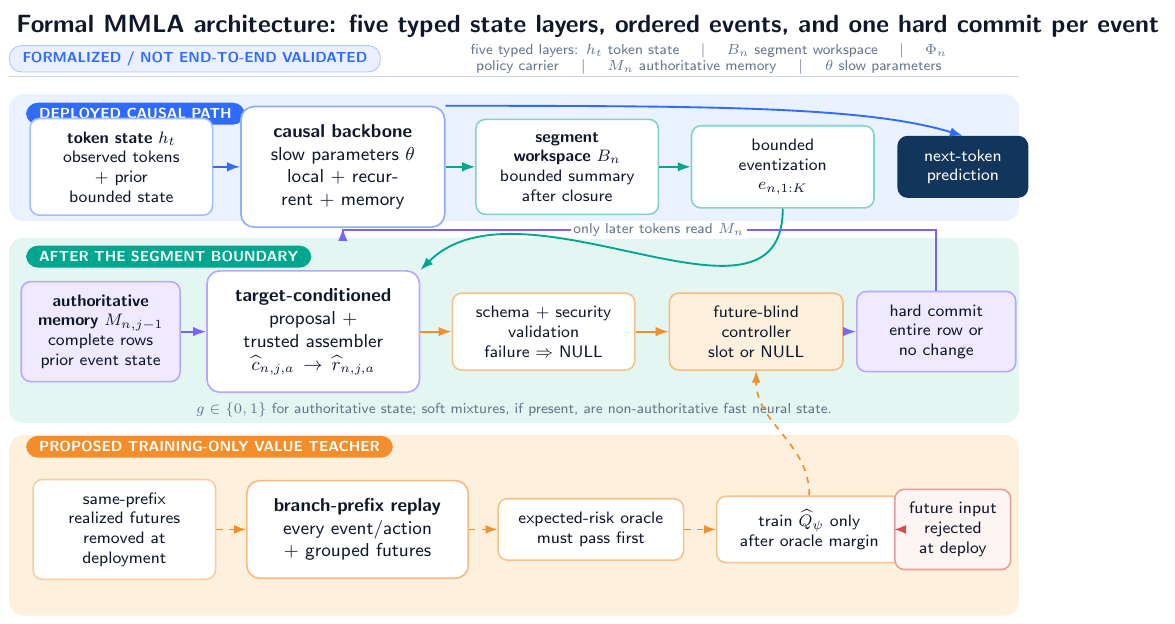}
\caption{\textbf{Predictive memory consolidation.} The language model predicts causally and reads only prior persistent state. After a local segment is complete, a bounded bidirectional consolidator may reinterpret that observed segment and emit ordered events. At training time, several realized continuations with the same prefix estimate the conditional risk of every registered event-level action. A causal value model regresses those risks. At deployment the continuation branches are absent; one complete row is overwritten or NULL leaves state unchanged per event, and only later tokens read the resulting segment state.}
\label{fig:predictive-consolidation}
\end{figure}

A block may expose local attention, a recurrent or linear state, and bounded memory read. Their ratio is not fixed. A budget-aware mixture is
\begin{equation}
\begin{aligned}
y_t&=\sum_{j\in\{\mathrm{local},\mathrm{linear},\mathrm{memory}\}}
w_{t,j}y_{t,j},\\
w_t&=\operatorname{softmax} r_\theta(h_t,M_{n(t)-1},b_t),
\end{aligned}
\label{eq:dynamic-mixer}
\end{equation}
where $b_t$ describes the available compute budget. An implementation may evaluate only the selected branches or place memory blocks sparsely. The invariant is bounded, editable state, not a particular layer ratio.

For a reproducible first end-to-end falsification, we nevertheless fix a proposed \emph{mixed-backbone reference profile}: 12 mixer blocks in four repetitions of [local attention, recurrent/linear state, memory read], a 2,048-token local window, 256-token completed segments, $S=256$ resident rows, at most four ordered events per segment, and at most one hard commit per event. The archive is disabled during semantic-interface qualification. Neural computation uses BF16 with FP32 normalization and controller reductions. This is a reference test configuration, not an empirically optimized product or a universal mixer ratio.

\subsection{Local bidirectional consolidation without future leakage}

Let the $n$th completed segment occupy $(b_{n-1},b_n]$. Token prediction remains autoregressive,
\begin{equation}
p_\theta(x_{1:T})=\prod_{t=1}^{T}
p_\theta(x_t\mid x_{<t},M_{n(t)-1}).
\label{eq:causal-factorization}
\end{equation}
Only after $x_{b_n}$ has been processed does the consolidator form
\begin{equation}
\begin{aligned}
z_n&=C_\phi\!\left(h_{b_{n-1}+1:b_n},M_{n-1}\right),\\
M_n&\not\rightarrow p_\theta(x_t\mid x_{<t})\quad(t\le b_n).
\end{aligned}
\label{eq:local-consolidation}
\end{equation}
Thus $C_\phi$ may use bidirectional attention \emph{inside an already observed segment}; its output can affect only later tokens. This is local reinterpretation, not inference-time access to an unobserved future. The segment may be a fixed chunk in the first experiment and a learned stopping interval later. A maximum horizon is a safety and cost bound, not a claim that consolidation should always wait a fixed number of turns.
Equivalently, the deployed consolidator must satisfy
\begin{equation}
z_n\perp X_{>b_n}\mid(X_{\le b_n},M_{n-1}),
\label{eq:deploy-future-independence}
\end{equation}
while training-only losses may use $X_{>b_n}$ as a target. This distinction is structural: the forward API contains no future-bearing argument, and the updated state is unavailable to logits at or before the boundary. ``Local bidirectionality'' therefore means retrospective encoding of observations already received, not bidirectional generation and not a hidden full-sequence look-ahead.

The consolidator next eventizes the bounded segment summary into an ordered list,
\begin{equation}
E_\eta(z_n)=(e_{n,1},\ldots,e_{n,K_n}),\qquad K_n\le K_{\max}.
\label{eq:eventization}
\end{equation}
A segment containing several facts is decomposed before one-row actions are considered. Bounded multi-row atomic transactions are deferred and excluded from the single-overwrite contract.

Within a completed segment, let $M_{n,0}=M_{n-1}$ and process the ordered events sequentially. The state after event $j$ is $M_{n,j}$, and only the completed segment state $M_n=M_{n,K_n}$ becomes available to later tokens.

\subsection{One atomic overwrite per event}

For event $e_{n,j}$ and target $a\in\{1,\ldots,S\}$, a neural constructor proposes semantic content,
\begin{equation}
\widehat c_{n,j,a}=G_\phi(z_n,r_{n,j-1,a},M_{n,j-1},e_{n,j},a).
\label{eq:row-constructor}
\end{equation}
The proposal contains bounded canonical value and aliases, neural payload, typed routing key, uncertainty, and proposed operation or tombstone intent. A trusted assembler then forms the complete row,
\begin{equation}
\widehat r_{n,j,a}=\operatorname{AssembleTrusted}(\widehat c_{n,j,a},\text{tenant},\text{ACL},\text{old protection},\text{next version},\text{provenance},\text{rollback lineage}).
\label{eq:trusted-assembler}
\end{equation}
Tenant, ACL/security scope, monotone version, protection inheritance, source receipt, rollback lineage, and deletion verification are system-owned fields, not free neural outputs. Their resident representation is fixed-width: receipts and rollback lineage are bounded identifiers into $A_n$, not embedded histories. Aliases live inside the bounded capsule or in a deterministic derived index; they never form an independently mutable authoritative map. Schema, size, security, and version checks run before action selection. An oversized or invalid candidate fails to NULL.
The version counter is an unsigned $B_u$-bit member of the metadata budget. A write is feasible only when $u_{\mathrm{old}}<2^{B_u}-1$ and the trusted assembler sets $u_{\mathrm{new}}=u_{\mathrm{old}}+1$; exhaustion rejects the write (or requires an explicit external re-key/migration protocol) and never wraps, saturates, or reuses a prior $(\iota,u)$ pair.

We propose a dual-view contract for future evaluation: canonical content makes a row auditable and reproducible, while a synchronized neural payload supports differentiable routing and readout. The two views are not merely co-located.  For row identity $\iota$ and version $u$, let $c$ be the canonical view and $(v,k)$ the neural payload and routing key.  A deterministic canonicalizer $\operatorname{Can}$ and a versioned encoder $E_{\omega,u}$ define the consistency receipt
\begin{equation}
\rho(r)=\Bigl(\iota,u,H(\operatorname{Can}(c)),H(E_{\omega,u}(\operatorname{Can}(c))),H(v\Vert k),H(\omega)\Bigr).
\label{eq:dual-view-receipt}
\end{equation}
A candidate is committable only if its identity/version/security fields agree and the fixed-width serialized neural bytes satisfy
\begin{equation}
\operatorname{Ser}_{v,k}\!\left(E_{\omega,u}(\operatorname{Can}(c))\right)
=\operatorname{Ser}_{v,k}(v,k).
\label{eq:dual-view-byte-equality}
\end{equation}
The hashes in $\rho$ are audit accelerators and tamper-evident receipts, not a substitute for the byte-equality check; a digest collision therefore cannot make unequal resident views valid. Canonical reconstruction and fresh-query semantic gates must also pass under the frozen versioned reader, and the receipt is included in the same atomic commit. The fixed-width $H(\omega)$ identifies an external immutable encoder artifact; encoder weights are not copied into the row. Any later neural re-encoding is a new row version, never an in-place drift. On load, recovery, export, and before use as teacher state, the system rechecks byte equality and $\rho$; failure quarantines the row and makes its action infeasible rather than choosing which view to trust. A deterministic derived index is rebuilt only from validated rows and is never authoritative. This closes the formal consistency invariant; whether a learnable encoder can satisfy its semantic gates remains unverified. The three negative semantic-interface studies motivate this contract; they do not validate it.

\begin{figure}[H]
\centering
\includegraphics[width=\linewidth]{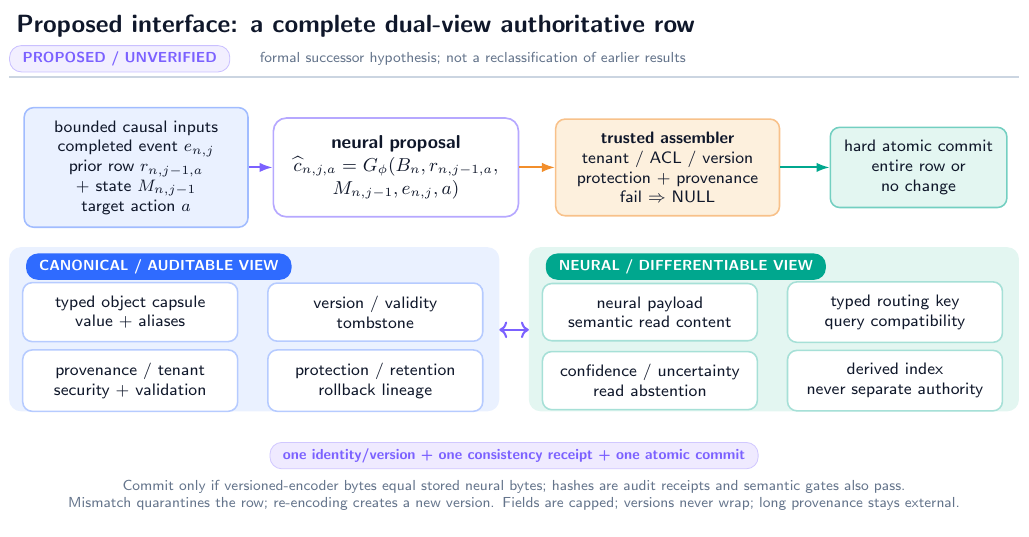}
\caption{\textbf{Proposed complete-row contract.} Canonical and neural views share one identity, validation boundary, version, and security scope. This proposed interface is explicitly unverified.}
\label{fig:complete-row}
\end{figure}

For each event $j$, define the feasible set only after every bounded candidate has been assembled and validated,
\begin{equation}
\mathcal A_{n,j}=\{\varnothing\}\cup
\left\{a\in\{1,\ldots,S\}:\widehat r_{n,j,a}\in\mathcal R_B,\ \operatorname{Valid}(\widehat r_{n,j,a})=1\right\}.
\label{eq:feasible-candidate-set}
\end{equation}
The controller chooses $a_{n,j}\in\mathcal A_{n,j}$ and a hard authoritative gate $g_{n,j}\in\{0,1\}$, with $g_{n,j}=0$ exactly when $a_{n,j}=\varnothing$. The state transition is
\begin{equation}
M_{n,j}=
\begin{cases}
M_{n,j-1}, & g_{n,j}=0\quad(\text{NULL}),\\
\operatorname{Replace}(M_{n,j-1},a_{n,j},\widehat r_{n,j,a_{n,j}}), & g_{n,j}=1,
\end{cases}
\qquad M_n=M_{n,K_n}.
\label{eq:atomic-overwrite}
\end{equation}
Insert, update, consolidating update, eviction, and deletion are lifecycle interpretations of replacing one complete row. Merging two authoritative resident rows requires two ordered events or a future bounded transaction, because changing one target cannot silently invalidate another row. A tombstone is itself a committed, versioned row; NULL changes nothing. Protection removes forbidden actions. A soft mixture, if useful, belongs only to non-authoritative fast neural state and never to the versioned fact table.

\begin{figure}[H]
\centering
\includegraphics[width=\linewidth]{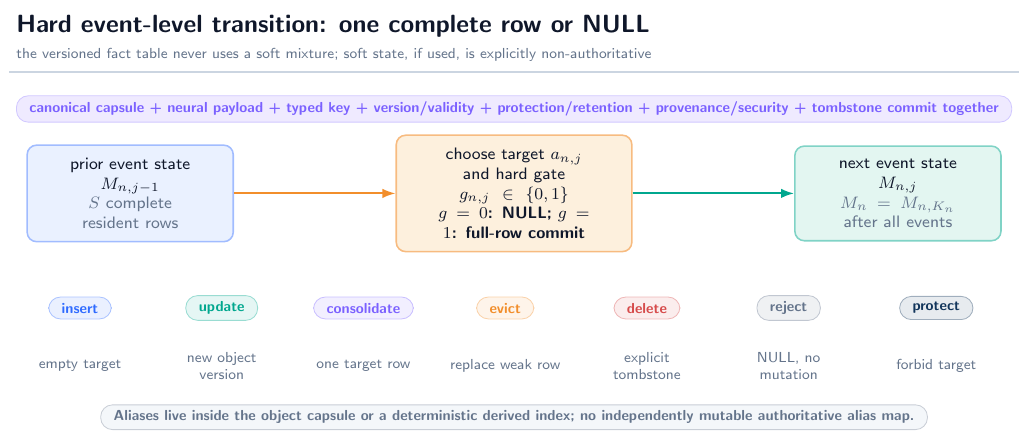}
\caption{\textbf{One physical transition, several lifecycle meanings.} The selected target, complete replacement row, and NULL option define the authoritative operation.}
\label{fig:atomic-meanings}
\end{figure}

If neural payloads satisfy $\lVert v(r)\rVert_2\le B$, a committed payload change obeys
\begin{equation}
\lVert v(\widehat r_{n,j,a_{n,j}})-v(r_{n,j-1,a_{n,j}})\rVert_2\le2B.
\label{eq:overwrite-drift}
\end{equation}
This is a one-step payload bound, not a convergence theorem. A poor policy can still oscillate or erase useful state. Permanent recovery of every historical version is impossible under fixed resident capacity unless versions are summarized or moved to a separately charged archive.

\begin{figure}[H]
\centering
\includegraphics[width=\linewidth]{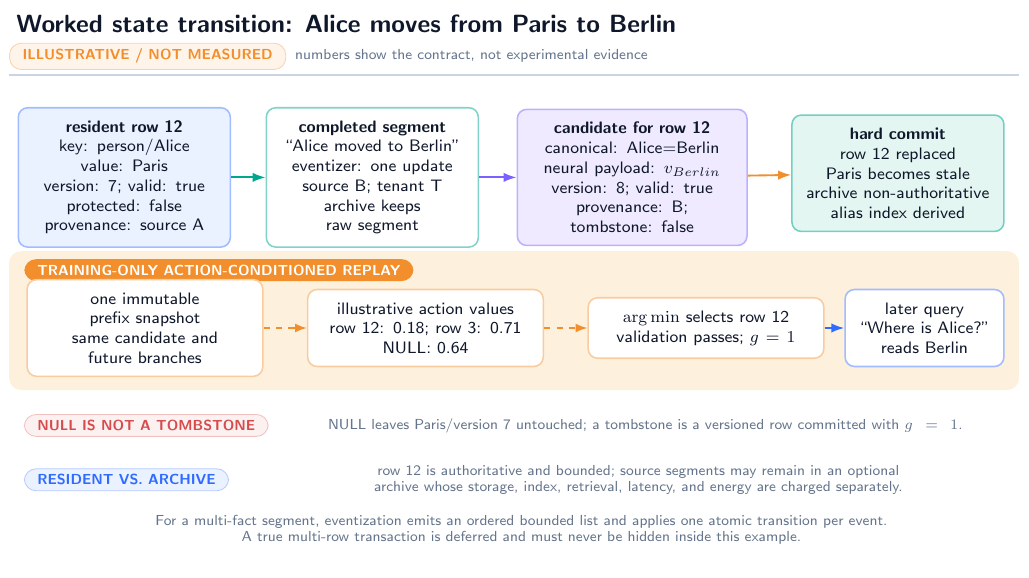}
\caption{\textbf{Worked state transition.} Alice's authoritative row advances from Paris/version 7 to Berlin/version 8. Numerical action values are illustrative, not measured. NULL preserves version 7; a tombstone would be a committed version 8 row. The replay teacher replays each target-conditioned action from the same snapshot; the archive remains non-authoritative and separately charged.}
\label{fig:alice-transition}
\end{figure}

\paragraph{System and recovery contract.}
The action mask enforces protection and tenant scope before values are compared. A failed schema, signature, provenance, monotone-version, or security check returns NULL. A committed row and its audit record publish atomically. Recovery restores a prior versioned snapshot and invalidates deterministic derived indexes. Verified deletion writes a tombstone and applies the declared archive-retention policy; it is never conflated with NULL. These are formal requirements, not end-to-end validated mechanisms.

\subsection{Multi-event, multi-future risk teacher and causal action values}

Let the bounded deployment summary for event $j$ be $s_{n,j}=S_\theta(h_{b_n-w+1:b_n},z_n,M_{n,j-1},e_{n,j})$. The notation $x_{\le b_n}$ describes only the causal information set: deployment cannot reread the full prefix unless archive retrieval and its cost are explicit. A segment is one finite-horizon decision episode with events $j=1,\ldots,K_n$. For any event-action sequence $\mathbf a_{1:j}=(a_1,\ldots,a_j)$, define the prefix state recursively by
\begin{equation}
M_{n,0}^{()}=M_{n-1},\qquad
M_{n,j}^{(\mathbf a_{1:j})}=\operatorname{Commit}\!\left(M_{n,j-1}^{(\mathbf a_{1:j-1})},a_j,\widehat r_{n,j,a_j}^{(\mathbf a_{1:j-1})}\right).
\label{eq:multi-event-prefix-state}
\end{equation}
The event summary, candidate, feasible mask, and later event state are recomputed from that branch's validated prefix; mixing a candidate from one prefix with another prefix is invalid. The future continuation $F$ remains random under $P(F\mid s_{n,1})$.

The exact sequence teacher prices a policy or complete action sequence $\mathbf a=(a_1,\ldots,a_{K_n})$ from one immutable pre-segment snapshot. A deterministic sequence is a special future-blind policy and is useful for exhaustive enumeration:
\begin{equation}
\begin{aligned}
J_H^{\mathrm{seq}}(\mathbf a;s_{n,1},F)={}&
\sum_{j=1}^{K_n}\lambda_{\mathrm{write}}\mathbf 1[a_j\ne\varnothing]
+\lambda_{\mathrm{int}}D_{\mathrm{collateral}}(M_{n,K_n}^{(\mathbf a)})\\
&+\sum_{k=1}^{H}\gamma^{k-1}\ell\!\left(F_k;M_{n,K_n}^{(\mathbf a)}\right).
\end{aligned}
\label{eq:sequence-hindsight-risk}
\end{equation}
For a feasible action $a$ at event $j$ after branch prefix $\mathbf a_{<j}$, let $\Pi_{j+1:K_n}^{\mathrm{fb}}$ be the preregistered class of deterministic or stochastic continuation policies whose action at event $q$ is measurable only with respect to the completed-segment event history and branch state $s_{n,q}$, never the identity or tokens of sampled $F$. Its teacher value is the future-blind cost-to-go
\begin{equation}
Q^{\mathrm T}_{n,j}(a\mid s_{n,j}^{(\mathbf a_{<j})})
=\min_{\pi\in\Pi_{j+1:K_n}^{\mathrm{fb}}}
\mathbb E_{F,\,\mathbf a_{j+1:K_n}\sim\pi}\!\left[
J_H^{\mathrm{seq}}((\mathbf a_{<j},a,\mathbf a_{j+1:K_n});s_{n,1},F)
\right].
\label{eq:multi-event-teacher-value}
\end{equation}
The order of minimization and expectation is essential: moving the minimum inside $\mathbb E_F$ would allow a clairvoyant continuation action chosen after seeing the sampled future. Exact dynamic programming minimizes the grouped expected return at every branch prefix; an approximation uses one frozen future-blind rollout policy declared before held-out futures are opened. Thus later events are either explicitly optimized or explicitly governed; they are never silently ignored. When $K_n=1$, this reduces to the single-event risk below. For the first matched causal intervention, the registered protocol deliberately sets $K_n=1$ and forbids later writes inside the scoring horizon, so that every branch isolates exactly one action:
\begin{equation}
\begin{aligned}
J_H(a;s_{n,j},F)={}&\sum_{k=1}^{H}\gamma^{k-1}
\ell\!\left(F_k;M_{n,j}^{(a)}\right)\\
&+\lambda_{\mathrm{int}}D_{\mathrm{collateral}}(a)
+\lambda_{\mathrm{write}}\mathbf{1}[a\neq\varnothing],
\end{aligned}
\label{eq:hindsight-risk}
\end{equation}
where $M_{n,j}^{(a)}$ is the event-level state after action $a$. The frozen reader must replay the continuation under that post-action state; adding memory only after a future hidden trace has already been cached is a fixed-trace diagnostic, not an action-conditioned counterfactual. The first matched experiment forbids later writes inside this horizon, so that $J_H$ identifies the consequence of exactly one intervention. The collateral term measures damage to facts that were not modified, leakage of obsolete values, or destruction of a still-useful slot.

The deployable single-event objective is the conditional action value
\begin{equation}
\overline J_H(a\mid s_{n,j})
=\mathbb E_{F\sim P(\cdot\mid s_{n,j})}
\left[J_H(a;s_{n,j},F)\right].
\label{eq:conditional-risk}
\end{equation}
For either the sequence return in Eq.~\ref{eq:sequence-hindsight-risk} or its registered $K_n=1$ restriction, the formal benchmark approximates the future expectation with $K_F$ physically grouped continuations sharing the identical prefix and pre-decision memory snapshot,
\begin{equation}
\widetilde J_H(a\mid s_{n,j})
=\sum_{k=1}^{K_F}w_k J_H(a;s_{n,j},F_k),
\qquad \sum_k w_k=1,
\label{eq:empirical-risk}
\end{equation}
with fixed preregistered weights $w_k$ that are normalized within a group and never inferred from the preferred action. Group IDs, not continuation branches, are split across fit, calibration, and held-out data. For $K_n>1$, the same grouped futures price every branch prefix in the exact dynamic program or the declared frozen rollout policy. A future-blind value model $\widehat Q_\psi(a\mid s_{n,j})$ regresses the detached cost-to-go target for \emph{every} feasible action, for example with a Huber loss,
\begin{equation}
\mathcal L_Q=\sum_{a\in\mathcal A_{n,j}}
\operatorname{Huber}\!\left(
\widehat Q_\psi(a\mid s_{n,j})-
\operatorname{stopgrad}\widetilde Q^{\mathrm T}_{n,j}(a\mid s_{n,j})
\right).
\label{eq:action-value-regression}
\end{equation}
At deployment it processes events in order, choosing $\arg\min_a\widehat Q_\psi(a\mid s_{n,j})$, committing the chosen row, recomputing $s_{n,j+1}$ from the resulting state, and optionally requiring a calibrated margin over NULL. It receives no continuation, future hidden state, answer, or gold action. In the first causal intervention, $K_n=1$ and one payload with $g_{n,1}=1$ may deliberately be shared across slot actions to isolate \emph{where or whether to write}. That controlled simplification is not the complete target-conditioned architecture of Equations~\ref{eq:row-constructor}--\ref{eq:trusted-assembler} and cannot establish multi-event credit assignment.

This distinction repairs a failure of the earlier formulation. In general,
\begin{equation}
\mathbb E_F\!\left[\operatorname{softmax}(-J_F/\tau)\right]
\neq
\operatorname{softmax}\!\left(-\mathbb E_F[J_F]/\tau\right).
\label{eq:expectation-policy-mismatch}
\end{equation}
Consequently, KL imitation of a Gibbs policy built from one realized future does not by itself optimize conditional expected risk. The earlier code remains useful for interface qualification, but it is not the decisive predictive-consolidation estimator. A learned latent rollout may later approximate $P(F\mid s_{n,j})$, as in world-model and latent-imagination systems \cite{worldmodels2018,dreamer2020}; it is not the first source of supervision, because a misspecified predictor can reward memories that merely make its own imagined future self-consistent.

\subsection{Motivation only: predictive information under finite capacity}

The persistent state is a lossy code of the observed past. We want it to approach a sufficient statistic for the future,
\begin{equation}
P(X_{>b_n}\mid X_{\le b_n})
\approx P(X_{>b_n}\mid M_n),
\end{equation}
rather than reconstruct every previous token. A predictive information-bottleneck objective is
\begin{equation}
\max_{q(M\mid X_{\le b_n})}
I(M;X_{>b_n})-
\beta I(M;X_{\le b_n})-
\lambda\,\mathbb E[N_{\mathrm{overwrite}}].
\label{eq:predictive-bottleneck}
\end{equation}
This is the information-bottleneck principle with future observations as the relevant variable \cite{tishby2000ib,bialek2001predictive}. If the deployed slots use a finite representation of $b$ bits each, then
\begin{equation}
\begin{aligned}
I(M;X_{\le b_n})&\le H(M)\le Sb,\\
I(M;X_{>b_n})&\le I(X_{\le b_n};X_{>b_n}),
\end{aligned}
\label{eq:finite-information}
\end{equation}
where the second inequality follows from data processing. The architecture cannot create predictive information or evade finite storage; it can only learn which predictive information to preserve. A variational implementation may replace the mutual-information penalty with a KL term to a bounded prior \cite{alemi2016vib}.

For the finite action set, let $E(a)=\overline J_H(a\mid s_{n,j})$ and $Z=\sum_a\exp[-E(a)/\tau]$. Any entropy-regularized action distribution $q$ obeys the exact identity
\begin{equation}
\begin{aligned}
\mathcal F[q]
&=\mathbb E_q[E(a)]-\tau H(q)\\
&=\tau D_{\mathrm{KL}}(q\Vert q^*)-\tau\log Z,\\
q^*(a)&=\frac{e^{-E(a)/\tau}}{Z}.
\end{aligned}
\label{eq:variational-free-energy}
\end{equation}
Equation~\ref{eq:variational-free-energy} is a regularized decision identity for one \emph{already averaged} conditional-risk vector. It does not justify averaging Gibbs policies formed separately for sampled futures; Eq.~\ref{eq:expectation-policy-mismatch} shows why. The first intervention trains action values directly. Here $\tau$ is an optional optimization temperature, not the physical temperature of a GPU or a brain.

For completeness, logical erasure also has a thermodynamic accounting boundary.
For erasing $b_{\mathrm{erase}}$ independent, initially unbiased bits without
exploitable side information in an ideal isothermal process at physical
temperature $T$, Landauer's lower bound is
\begin{equation}
Q_{\min}\ge b_{\mathrm{erase}}k_BT\ln2.
\label{eq:landauer}
\end{equation}
This standard result \cite{landauer1961} is not a memory-architecture contribution
or evidence of energy efficiency. Correlated information, physical encoding,
and the complete computational cycle require their own accounting. Present
hardware costs are far above this ideal limit; fewer logical updates need not
reduce measured wall-plug energy. The optimization temperature $\tau$ above
is not this physical temperature.

\subsection{Bounded resident state and the full system cost}

Let $T$ be total stream length, $W$ the local window, $S$ the resident slots, $K_r\ll S$ the active slots read per token, $K_e$ the maximum events per segment, and $C$ the minimum tokens per consolidation. Let $R_q(S,d)$ be the routing cost for one token, $C_{\mathrm{cons}}(W,d)$ the completed-segment encoding/eventization cost, $C_G(d,L_c)$ the cost of constructing one target-conditioned semantic proposal, $C_A(B_m)$ the trusted assembly/dual-view validation cost per candidate, $C_Q(S,d)$ the action-scoring cost per event after candidates exist, and $C_{\mathrm{commit}}(B_{\mathrm{row}})$ the bounded row-commit cost. If the complete architecture materializes all $S$ target candidates before choosing an action, the deployment context-mixing budget is
\begin{equation}
\begin{aligned}
C_{\mathrm{deploy}}(T)={}&O(TWd)+O\!\left(TR_q(S,d)\right)+O(TK_r d)\\
&+O\!\left(\frac{T}{C}\left[C_{\mathrm{cons}}(W,d)
+K_e\left(S(C_G+C_A)+C_Q+C_{\mathrm{commit}}\right)\right]\right),\\
\operatorname{bits}(M_n)\le{}&B_{\mathrm{resident}}=S(B_i+B_c+B_v+B_k+B_m).
\end{aligned}
\label{eq:context-compute}
\end{equation}
Here $C_G$ and $C_A$ are arguments of their displayed shapes; the shorthand suppresses them only inside the bracket. A shared precomputation may reduce $S C_G$ to $C_G^{\mathrm{shared}}+S C_G^{\mathrm{target}}$, and a preregistered target shortlist of size $L\ll S$ may replace $S$ only if shortlist construction, misses, and excluded actions are charged and reported. Neither optimization may omit complete-candidate construction from accounting. An exact read scan has $R_q(S,d)=\Theta(Sd)$. An approximate index can reduce query work only by adding explicit build, update, storage, and miss costs; those costs do not disappear from the system boundary.

Training cost is larger because the teacher branches actions and futures. For exact multi-event dynamic programming, let $N_{\mathrm{node}}\le\sum_{j=0}^{K_e-1}(S+1)^j$ be the number of reachable branch prefixes before the next action after feasibility pruning. With $K_F$ grouped futures and replay cost $C_{\mathrm{replay}}(H)$ per complete sequence leaf, let $N_{\mathrm{leaf}}\le(S+1)^{K_e}$ be the corresponding number of complete feasible action sequences. The additional teacher budget is
\begin{equation}
C_{\mathrm{teacher}}=
O\!\left(\frac{T}{C}\left[
N_{\mathrm{node}}\left(S(C_G+C_A)+C_Q+N_{\mathrm{child}}C_{\mathrm{commit}}\right)
+N_{\mathrm{leaf}}K_F C_{\mathrm{replay}}(H)
\right]\right).
\label{eq:teacher-compute}
\end{equation}
where $N_{\mathrm{child}}\le S$ is the maximum number of feasible non-NULL children materialized from a prefix (NULL reuses the parent state). The registered first intervention has $K_e=1$, so $N_{\mathrm{node}}=1$, $N_{\mathrm{leaf}}=S+1$, and $N_{\mathrm{child}}=S$ before feasibility pruning: all $S$ complete candidates are built, all $S$ non-NULL successor rows are materialized, and all $S+1$ actions, including NULL, are replayed across every grouped future. If an implementation uses immutable structural sharing, it may replace $S C_{\mathrm{commit}}$ by a separately measured branch-state materialization cost, but it may not charge only the ultimately selected deployment commit. Any multi-event successor must report pruning or rollout-policy approximation separately rather than hiding the exponential branch factor.

Sparse fallback over an archive $A_T$ makes the \emph{resident} state bounded but not the total retained information: archive and audit-ledger size, index size, retrieval latency, and retrieval energy must be reported separately. Under the stated fixed-budget assumptions, the operation count scales linearly with stream length. Practical throughput remains implementation-dependent: unfused small kernels may incur substantial launch and memory-traffic overhead despite favorable asymptotics. This is not a claim of constant total memory or small constant factors. BF16 is the reference training path, FP8 is appropriate for qualified matrix multiplications, and normalization, accumulated risk, and discrete action values retain the FP32 islands required for stability. The design introduces no FP64 requirement. FP4 is a later inference or post-training option whose overwrite stability must be tested separately.

\subsection{Historical registered intervention}

An unexecuted historical preregistration template proposed freezing one completed small-model checkpoint and its reader, then training only parameter-matched consolidation/controller modules on new, physically separated data. Its four deployable arms and one oracle would share the state size, candidate payload, gate, action candidates, overwrite budget, future-blind deployment interface, optimization exposure, and final checkpoint:
\begin{center}
\small
\setlength{\tabcolsep}{3.5pt}
\begin{tabular}{p{0.25\linewidth}p{0.27\linewidth}p{0.34\linewidth}}
\toprule
Arm & Segment representation & Write supervision \\
\midrule
No-op memory & causal & always NULL \\
Immediate writer & causal final state & current-only risk \\
Local consolidation & bidirectional completed segment & current-only risk \\
Expected-risk consolidation & same local consolidator & grouped multi-future action-value regression \\
Expected-risk oracle & same local consolidator & empirical conditional risk at evaluation; non-deployable ceiling \\
\bottomrule
\end{tabular}
\end{center}
In that template, the immediate-to-local difference would isolate post-segment reinterpretation, and the local-to-expected-risk difference would isolate a continuation distribution rather than current-only risk. Every benchmark group would contain an identical observed prefix with four to eight continuations, including ambiguous prefixes for which NULL is optimal in expectation and cases whose value appears only later. No later write would be allowed during the scoring horizon. The oracle would first test whether the data and action set contain useful conditional future value; if it did not beat current-risk writing, scale could not rescue that intervention. Student evaluation would report future loss, action regret, strict overwrite success, stale-value leakage, preservation of untouched facts, overwrite rate, latency, resident and archive memory, routing/index cost, kernel count, and energy. This ancestor protocol was not executed and is not represented as the current next experiment.

\subsection{Evidence-gated training allocation}

The proposed memory capability is primarily a \emph{mid-training or post-training} intervention, not a reason to begin with an expensive from-scratch model. Let $\theta_0$ be one frozen backbone, $\omega$ a current-state interface, and $\psi$ the future-blind action-value model. The evidence order is
\begin{equation}
\begin{aligned}
\hat\omega&=\arg\min_\omega \mathcal L_{\mathrm{interface}}(\omega;\theta_0),
&&\theta=\theta_0,\\
\hat\psi&=\arg\min_\psi \mathcal L_Q(\psi;\theta_0,\hat\omega),
&&\theta=\theta_0.
\end{aligned}
\label{eq:staged-training}
\end{equation}
The first line asks whether the resident state can be read and reconstructed at all; the second asks whether a deployable controller can exploit that interface to reduce future risk. Only if both fixed-checkpoint stages beat matched controls without collateral damage does joint training become scientifically justified:
\begin{equation}
\min_{\theta,\omega,\psi}
\mathcal L_{\mathrm{LM}}(\theta)
+\lambda_{\mathrm{mem}}\mathcal L_{\mathrm{interface}}
+\lambda_Q\mathcal L_Q.
\label{eq:conditional-joint-training}
\end{equation}
Equation~\ref{eq:conditional-joint-training} is a conditional scale study, not a completed result. This sequence keeps the comparatively cheap representation and policy hypotheses falsifiable before allocating a full pretraining budget.

\paragraph{Current evidence boundary.}
An earlier one-realization CPU loop remains a useful interface demonstration, but its cached future and realization-specific Gibbs target do not estimate Eq.~\ref{eq:conditional-risk}. Three three-seed current-state qualifications failed their preregistered semantic conditions. The first used an external semantic adapter; the second asked whether a raw hidden representation was directly usable. The third, the \emph{frozen-checkpoint semantic-interface study} (hereafter \emph{frozen-interface qualification}), compared a dense row, a structured $4\times192$ span, and the checkpoint-native reader under matched data, training, and final-only evaluation. All nine jobs completed 2,048 updates, materialized predictions before evaluator truth, and were evaluated unconditionally. Candidate exactness remained 0--45/23,040, query exactness 10--1,536/9,216, and record-macro Brier 0.998676--0.999093. Old-value suppression and physical-map invariance passed, but every semantic qualification failed.

The \emph{fitted-support fault-localization study} (hereafter \emph{fit-support localization}) reuses those immutable final checkpoints without optimizer updates and evaluates the complete registered fit population before opening the historical fresh summaries. Across nine jobs, direct candidate reconstruction is $40/73{,}728$, direct present readout and learned-content lattice exactness are both $0/73{,}728$, and a routing oracle still leaves learned content at $0/73{,}728$. Oracle content with learned routing reaches $18{,}544/73{,}728$, exactly the learned routing count, while the fully oracle-specified path is the required mechanical ceiling at $73{,}728/73{,}728$. Historical fresh results remain near zero on content-bearing probes. The preregistered tree therefore takes its fit/support-failure branch: the evidence cannot distinguish representation geometry, decoder, and optimization within that branch, but it does rule out fresh composition as the primary failure. No arm comparison is admissible, the same pure latent-row interface is not extended, and expected-risk overwrite remains unopened.

The analogy to complementary learning systems is useful but limited: fast editable state and slow parameter learning play different computational roles \cite{mcclelland1995cls}. We do not claim fixed slots, atomic overwrite, or conditional-risk supervision as a biological model. More generally, a system that can maintain revisable world state under a fixed budget could be useful for persistent agents, but planning, goals, exploration, causal modeling, safety, and slow continual learning remain separate problems.

\section{Implemented Evidence Path}
\label{sec:implemented-path}
\label{sec:method}

\subsection{Components}

\begin{figure}[t]
\centering
\includegraphics[width=\linewidth]{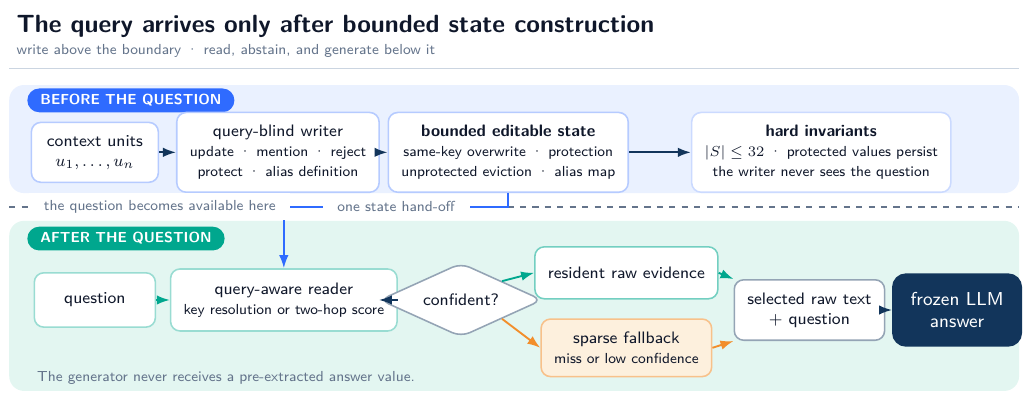}
\caption{The implemented path. State construction finishes before the question crosses the dashed boundary. At query time, the reader either uses resident evidence or abstains to budgeted sparse retrieval. The frozen language model receives the selected raw text and the question, never a pre-extracted answer value.}
\label{fig:schematic}
\end{figure}

Let a request context be segmented into candidate units $u_1,\dots,u_n$ (event lines in the lifecycle study; whole passages in the multi-hop QA study). The path executes in the following order.

\textbf{Writer (query-independent).} A classifier scores each unit from unit-local features only; its interface does not accept the final question. A unit test enforces this restriction, and identifiers, values, and digits are masked in the controlled lifecycle features. In that study the writer labels each line as \texttt{update}, \texttt{mention}, \texttt{reject}, \texttt{lock}, or \texttt{alias-def}. In the multi-hop QA study it predicts passage salience.

\textbf{Controlled-language lifecycle executor.} Units stream in document order into a memory of at most $32$ slots, keyed by entity in the lifecycle study and ranked by salience in the QA study. \texttt{update} writes; \texttt{lock} writes and protects, making later writes to that key void; \texttt{reject} and \texttt{mention} never write; alias definitions populate a separate alias map. When a new key arrives at capacity, the least-recently-updated unprotected slot is evicted. This executor tests lifecycle semantics under controlled cues. Because its alias map is independently mutable, it is not itself an implementation of the single-row atomic state contract in Eq.~\ref{eq:atomic-overwrite}.

\textbf{Reader (query-aware).} At question time the reader either resolves the requested key, directly or through the alias map, or scores resident passages with a learned two-hop reranker. The reranker combines dense-retrieval-style features \cite{dpr2020}, BM25 \cite{bm251995}, lexical overlap, and bridge features. Its dense feature is a frozen-language-model hidden state, not a separately trained DPR encoder.

\textbf{Calibrated sparse fallback.} A confidence signal --- the writer probability of a slot's source line in the lifecycle study, or the mean of the two highest resident-passage scores in the QA study --- is compared with a threshold chosen from calibration data only. A miss or a score below that threshold sends the request to budget-matched sparse retrieval over the full context. The threshold must be calibrated at the intended context length. When it was calibrated on short contexts, where 32 slots contain every passage, fallback never fired and extended-length evidence coverage fell from $0.88$ to $0.31$ (\S\ref{sec:negatives}).

\textbf{Generation.} A frozen instruction-tuned model receives only the \emph{raw selected evidence} (verbatim context lines/passages --- asserted at prompt-build time, never a pre-extracted answer value) plus the question, and generates the answer greedily.

The state transition and the fallback decision can be written compactly. Let
$S_t=\{(k_i,v_i,p_i,\rho_i)\}_{i=1}^{m_t}$ contain each resident key, value,
protection bit, and retention score. For a writer action $a_t$, the lifecycle
operator obeys
\begin{equation}
\begin{aligned}
S_{t+1}&=\mathcal{L}(S_t,u_t,a_t),\\
|S_{t+1}|&\leq M,\\
p_i=1&\Rightarrow (k_i,v_i)\text{ is protected}.
\end{aligned}
\label{eq:lifecycle}
\end{equation}
After the question $q$ becomes available, the evidence passed to generation is
\begin{equation}
E(q,S_n,x)=
\begin{cases}
R(q,S_n), & \mathrm{hit},\ c(q,S_n)\geq\tau,\\
\operatorname{SparseRetrieve}(q,x;B), & \text{otherwise},
\end{cases}
\label{eq:fallback}
\end{equation}
where $B$ is the fixed evidence budget and $\tau$ is chosen on calibration data. Equations~\ref{eq:lifecycle}--\ref{eq:fallback} expose the two guarantees that are easy to conflate: the lifecycle constrains how resident state may change, while fallback handles evidence that was absent from that state.

\textbf{Two instantiations, disclosed.} The controlled versioning study exercises typed events, same-key overwrite, protection with void writes, LRU eviction, and alias resolution. The multi-hop QA study contains static encyclopedia passages, so its memory reduces to a \emph{bounded salience cache}: the 32 passages with the highest query-independent writer scores. No overwrite or protection event fires in that setting. Its writer is query-free at inference, although its training labels come from question-linked supporting facts in the training split. These differences limit what the real-text result can say about lifecycle semantics.

\subsection{How the completed evidence maps to the proposal}
\label{sec:trainable-architecture}

The implemented path predates the conditional-risk teacher and does not instantiate the entire model in Figure~\ref{fig:predictive-consolidation}. Its role is to test semantics that a learned overwrite must eventually preserve. The controlled versioning study exercises same-key update, protection, stale rejection, eviction, and NULL-like non-writes under a controlled grammar. The multi-hop QA study measures the writer's information boundary and the value of abstention. The relational study tests a candidate-construction side path before lifecycle execution. None compares immediate writing with local or expected-risk consolidation.

A separate native scaffold uses pre-norm RMSNorm, rotary embeddings, grouped-query local attention, SwiGLU, and intermittent memory-managed mixers \cite{rmsnorm2019,roformer2021,gqa2023,mistral7b2023,swiglu2020}. It has completed a 50{,}003{,}968-token real-data pilot. Predictive consolidation and conditional-risk supervision were not part of that training, so the run is evidence of trainability and checkpoint availability rather than evidence for the new mechanism. The staged studies in Eq.~\ref{eq:staged-training} reuse its final checkpoint precisely to avoid attributing backbone or data-order differences to memory-interface learning.

The first CPU implementation enforces useful interface invariants: reads depend only on prior persistent state; local consolidation sees only a completed segment; deployment methods reject future-bearing arguments; and a hard update changes no more than one payload row. It does not replay future tokens under each action, estimate risk from multiple continuations with the same prefix, or include all lifecycle metadata in the resident row. It is therefore a mechanical ancestor, not positive evidence for the learned architecture.

\begin{table}[H]
\centering
\scriptsize
\setlength{\tabcolsep}{3.6pt}
\begin{tabular}{llrrrrl}
\toprule
Interface & Seed & Candidate exact & Query exact & Mean Brier & Preservation & Physical map \\
\midrule
External semantic adapter & 2026082101 & $358/6656$ & $244/2048$ & $0.6490$ & $13/1536$ & pass \\
 & 2026082102 & $314/6656$ & $441/2048$ & $0.6263$ & $3/1536$ & pass \\
 & 2026082103 & $406/6656$ & $258/2048$ & $0.6342$ & $20/1536$ & pass \\
Raw representation probe & 2026083101 & $12/2048$ & $343/131072$ & $1.4201$ & --- & pass \\
 & 2026083102 & $43/2048$ & $263/131072$ & $1.4230$ & --- & pass \\
 & 2026083103 & $36/2048$ & $513/131072$ & $1.4023$ & --- & pass \\
\bottomrule
\end{tabular}
\caption{Registered frozen-checkpoint interface negatives. ``Physical map'' denotes the preregistered invariance checks, not semantic success. The external adapter also passed old-value suppression but failed reconstruction, query, calibration, and unrelated-fact preservation. The raw probe failed its first five semantic conditions and passed only two mapping conditions on every seed. These experiments localize an interface problem; they do not reject the native memory mechanism or predictive consolidation.}
\label{tab:interface-negatives}
\end{table}

\begin{table}[H]
\centering
\scriptsize
\setlength{\tabcolsep}{3.4pt}
\begin{tabular}{llrrrrrl}
\toprule
Seed & Reader & Candidate exact & Query exact & Missing $\rightarrow$ absent & Mean Brier & Map & Qualified \\
\midrule
2026091101 & dense row & $0/23040$ & $1536/9216$ & $1536/1536$ & $0.998981$ & pass & no \\
 & structured span & $15/23040$ & $1536/9216$ & $1536/1536$ & $0.998676$ & pass & no \\
 & native reader & $30/23040$ & $645/9216$ & $640/1536$ & $0.998992$ & pass & no \\
\addlinespace
2026091102 & structured span & $21/23040$ & $10/9216$ & $0/1536$ & $0.999035$ & pass & no \\
 & native reader & $45/23040$ & $10/9216$ & $0/1536$ & $0.999046$ & pass & no \\
 & dense row & $0/23040$ & $1536/9216$ & $1536/1536$ & $0.998815$ & pass & no \\
\addlinespace
2026091103 & native reader & $30/23040$ & $10/9216$ & $0/1536$ & $0.999093$ & pass & no \\
 & dense row & $0/23040$ & $1536/9216$ & $1536/1536$ & $0.998818$ & pass & no \\
 & structured span & $0/23040$ & $1536/9216$ & $1536/1536$ & $0.998977$ & pass & no \\
\bottomrule
\end{tabular}
\caption{Frozen-interface qualification from one frozen checkpoint. Predictions were materialized before evaluator truth; all 55,296 record rows were reduced. Mean Brier is near the 1,025-class uniform reference $1024/1025=0.999024$. Every job passes old-value suppression and the registered 64-row mapping checks, but none passes the semantic conditions. Because no interface qualifies at all three seeds, no arm comparison, preferred-reader claim, or bootstrap comparison is defined.}
\label{tab:semantic-interface-qualification}
\end{table}

The frozen-interface qualification moves one level closer to the model than Table~\ref{tab:interface-negatives}: the dense reader reads a row, the structured reader preserves a structured $4\times192$ span, and the native reader calls the checkpoint-native memory path. Table~\ref{tab:semantic-interface-qualification} shows that the result is not a narrow failure of one external codec. The nine jobs preserve the registered structural invariants yet remain close to chance on semantic output. The comparison therefore rules out these three frozen interfaces under their matched budgets; it does not show that the frozen backbone lacks potentially useful information, nor does it distinguish an optimization bottleneck from an unsuitable compositional output geometry.

\begin{table}[H]
\centering
\scriptsize
\setlength{\tabcolsep}{4.1pt}
\begin{tabularx}{\linewidth}{lrlY}
\toprule
Fit-support probe & Exact & Population & Diagnostic reading \\
\midrule
Direct candidate reconstruction & $40/73{,}728$ & registered fit records & fit support is not reconstructed \\
Learned route / learned content (L/L) & $0/73{,}728$ & grouped fit probes & ordinary interface fails on fit \\
Oracle route / learned content (O/L) & $0/73{,}728$ & grouped fit probes & target-row isolation does not repair content \\
Learned route / oracle content (L/O) & $18{,}544/73{,}728$ & grouped fit probes & equals learned row-routing count \\
Oracle route / oracle content (O/O) & $73{,}728/73{,}728$ & grouped fit probes & mechanical ceiling, not learned success \\
Historical candidate exactness & $141/207{,}360$ & frozen-interface fresh records & fresh semantics also fail \\
Historical known/latest query exactness & $7/13{,}824$ each & frozen-interface fresh queries & content-bearing queries remain near zero \\
Historical map / old suppression & $82{,}944/82{,}944$; $13{,}824/13{,}824$ & frozen-interface checks & structural invariants still pass \\
\bottomrule
\end{tabularx}
\caption{Accepted fit-support localization. The nine fixed frozen-interface final checkpoints are reused without retraining, checkpoint selection, threshold changes, or arm preference. O/L supplies the registered target-row oracle while retaining learned content; L/O supplies oracle content while retaining learned routing. The complete O/O path is a mechanical ceiling. Because learned content fails even under target-row isolation on the fit population, the preregistered terminal is fit/support failure rather than fresh-composition-only failure.}
\label{tab:fitted-support-localization}
\end{table}

The fit-support localization is diagnostic, not a second qualification or a retrospective reader contest. It motivated alternative interface hypotheses, including canonical token spans, typed tuples, and dual-view rows, without validating any of them. It does not distinguish representation geometry, decoder design, and optimization as the cause of fitted-support failure, or establish that the frozen backbone lacks useful memory information. The later post-training experiments in Section~\ref{sec:posttraining-evidence} use separately declared model and protocol lineages.

\subsection{Preregistration and test discipline}

Before generating confirmatory data, we fixed the comparisons, the numerical conditions that would count as support, and the circumstances that would stop the run. Frozen holdouts were then read once; the multi-hop QA holdouts were read once for each of the two reader models specified in advance. The lifecycle benchmark used for the reported result was required to defeat the strongest declared lexical rule by construction before its held-out split was opened. Section~\ref{sec:evaluation-contracts} states the evaluation contracts and failure-preserving interpretation used in the final analysis.

\section{Controlled Lifecycle Study}
\label{sec:tracka}

\textbf{Data.} 1{,}200/300/300 train/dev/test records (split-disjoint sentence frames, shared cue vocabulary --- a disclosed controlled-language boundary), six balanced scenarios: \texttt{base-overwrite}, \texttt{alias-query} (the query names the key only through an alias defined once in context), \texttt{stale-remention} (post-final-update audit lines repeat old values), \texttt{invalid-write} (rejected change requests propose wrong values), \texttt{protected-slot} (a lock freezes the value; later updates are void), and \texttt{slot-pressure} (48 distinct written keys exceed the 32-slot bound). Generator tests prove that ``take the last line mentioning the queried key'' is wrong by construction on the four middle scenarios --- and returns exactly the designed wrong value --- while remaining correct on the other two.

\begin{table}[t]
\centering
\footnotesize
\setlength{\tabcolsep}{4.5pt}
\begin{tabular}{lccc}
\toprule
Method & Held-out exact match (3 seeds) & Purpose \\
\midrule
Lexical first-key & $0.087$ & floor control \\
Lexical last-key & $0.333$ & \textbf{strongest non-learned baseline} \\
Naive all-writes lifecycle & $0.333$ & no event typing \\
Writer allowed to see the question & $0.333$ & leakage control \\
Learned writer; lifecycle disabled & $0.490/0.490/0.493$ & component removal \\
Lifecycle active; fallback disabled & $0.990$ (all seeds) & component removal \\
\textbf{Complete controlled lifecycle path} & $\mathbf{1.000}$ (all seeds) & proposed path \\
Lifecycle with oracle event labels & $1.000$ & diagnostic ceiling \\
\bottomrule
\end{tabular}
\caption{Controlled-language lifecycle results on 300 held-out records, read once. Before the split was opened, we required improvement over the strongest hand-written rule on every seed with a positive paired-bootstrap 95\% confidence interval, substantially fewer lifecycle errors than the learned writer without lifecycle control, at least 0.90 accuracy on protected values, recovery of most capacity misses, and no more than 32 resident slots. The complete path met each condition: a $+66.7$-point margin with 95\% CI $[+61.3,+72.0]$, zero lifecycle errors versus $0.51$, protected-value accuracy $1.0$, recovery of all $3/3$ target evictions, and peak occupancy $32$. These numbers validate controlled execution, not learned natural-language memory management.}
\label{tab:tracka}
\end{table}

\begin{figure}[t]
\centering
\includegraphics[width=0.94\linewidth]{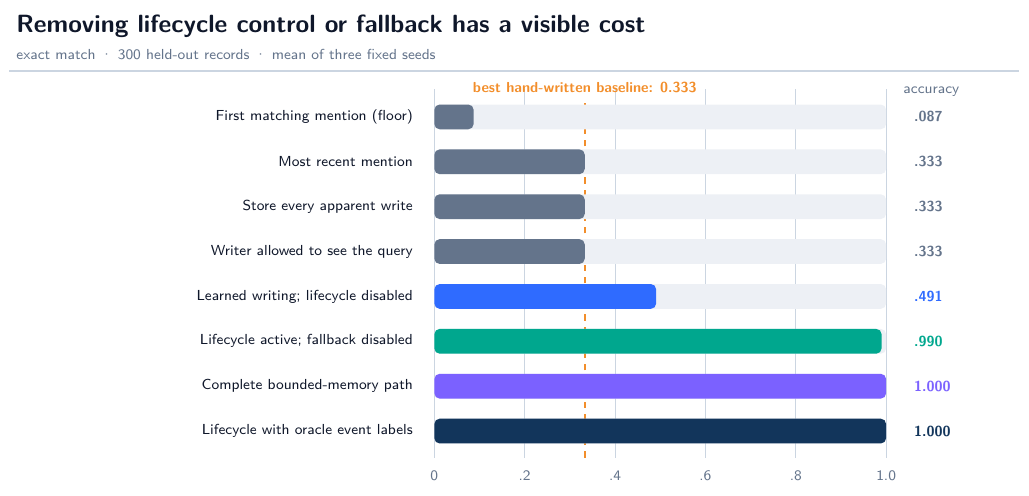}
\caption{Held-out exact match in the controlled lifecycle study, averaged over three fixed seeds. The dashed line is the strongest hand-written rule. Under the registered implementation and training recipe, the complete typed lifecycle path outperforms the specified lexical baselines and component-removal controls; disabling lifecycle execution roughly halves accuracy, while disabling fallback loses only the capacity-pressure cases.}
\label{fig:tracka}
\end{figure}

\begin{figure}[!t]
\centering
\includegraphics[width=0.94\linewidth]{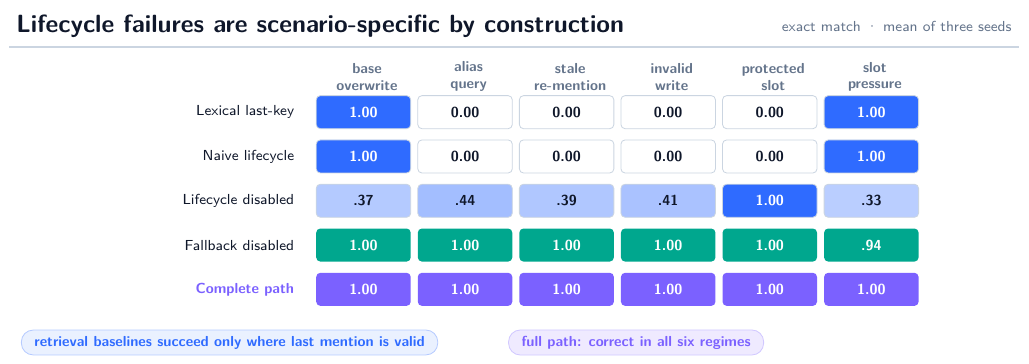}
\caption{Held-out exact match by lifecycle scenario, averaged over three fixed seeds. Lexical and untyped rules solve only the cases in which the most recent mention is valid. Without lifecycle execution, the learned writer cannot arbitrate versions; without fallback, only capacity-pressure evictions are lost. The complete path is correct in all six scenarios.}
\label{fig:tracka-scen}
\end{figure}

Table~\ref{tab:tracka} and Figures~\ref{fig:tracka}--\ref{fig:tracka-scen} make the component costs visible. Removing event typing or allowing the writer to depend on the question returns accuracy to the 0.333 lexical level. Keeping the learned writer but skipping lifecycle execution yields about 0.49. Keeping lifecycle execution but removing fallback loses three capacity-pressure cases, reducing exact match from 1.000 to 0.990. The writer also transfers across disjoint sentence frames (development macro-F1 $1.0$; shuffled-label control $0.069$).

Fallback has a narrow role in this controlled grammar. All three calibrated thresholds reached the minimum of the search grid because the writer was both confident and correct, and keys and values were recoverable by the controlled parser. The low-confidence branch therefore never fired; fallback was used only for the three target slots evicted under capacity pressure, and recovered all three. The multi-hop QA study supplies the complementary setting, with fallback rates of $0.39$--$0.72$.

\textbf{Generation.} This secondary generation comparison uses the fixed writer seed 202607094, not three independent generation runs. Given selected raw lines rather than pre-extracted values, the complete path answers $300/300$ with an average prompt length of $146$ tokens. The same frozen Llama-3.1-8B \cite{llama32024}, when given the entire log, answers $172/300$ with $729$ tokens. Its errors concentrate on protection and alias resolution. That distribution is consistent with explicit lifecycle handling being useful, but the comparison does not separately identify the contributions of lifecycle execution, evidence selection, and prompt length.

\section{Held-Out Multi-Hop QA at Two Lengths}
\label{sec:trackb}

\textbf{Data hygiene.} An early selector had already consumed labels for 20 examples from the local LongBench HotpotQA slice, so that pool is retained only for diagnostics. The confirmatory questions instead come from the official HotpotQA training split \cite{hotpotqa2018}. After removing exact and near-duplicate questions against all 500 local LongBench examples, we froze 1{,}200 training, 300 calibration, and 300 held-out records. Their median context length is $933$ words, far below the roughly 7.9k words used by LongBench \cite{longbench2023}. We therefore created a second held-out condition with a median of $8{,}221$ words by adding answer-excluded distractor passages drawn from the training-record pool. The questions and gold passages remain held out, but the distractor text is train-derived; this is a test of held-out questions among familiar distractor text, not of wholly unseen documents. Before developing the selector, we fixed evidence budgets at $25\%$ of context words for natural records and $10\%$ for extended records. Frozen Llama-3.1-8B and Qwen2.5-14B models serve as readers \cite{llama32024,qwen252024}.

For selected passages $E$ and full context $x$, the intended evidence budget is
\begin{equation}
\begin{aligned}
b(E;x)&=\frac{\sum_{p\in E}|p|_{\mathrm{word}}}
{|x|_{\mathrm{word}}}\leq\beta,\\
\beta&=0.25\ \text{(natural)},\quad
0.10\ \text{(extended)}.
\end{aligned}
\label{eq:evidence-budget}
\end{equation}
The first Llama execution used whole-prompt tokens rather than Eq.~\ref{eq:evidence-budget} and is retained as failed under that frozen definition; the Qwen follow-up enforces this equation record by record. Section~\ref{sec:evaluation-contracts} gives the full accounting.

\begin{table}[t]
\centering
\footnotesize
\setlength{\tabcolsep}{4.5pt}
\begin{tabular}{llcccc}
\toprule
Reader model & Method & \multicolumn{2}{c}{Natural (25\% tier)} & \multicolumn{2}{c}{Extended 8.2k words (10\% tier)} \\
 & & F1 & tokens & F1 & tokens \\
\midrule
\multirow{6}{*}{Llama-3.1-8B}
 & Full context & $0.714$ & 1477 & $0.666$ & 12024 \\
 & Dense (budget-matched) & $0.516$ & 432 & $0.621$ & 1300 \\
 & BM25 (budget-matched) & $0.601$ & 437 & $0.636$ & 1293 \\
 & Reader only (no state bound) & $0.659$--$0.663$ & 434 & $0.695$--$0.706$ & 1310 \\
 & \textbf{Resident cache + archive fallback} & $\mathbf{0.659}$--$\mathbf{0.663}$ & 434 & $\mathbf{0.677}$--$\mathbf{0.684}$ & 1325 \\
 & Oracle evidence (ceiling) & $0.717$ & 438 & $0.786$ & 1301 \\
\midrule
\multirow{6}{*}{Qwen2.5-14B}
 & Full context & $0.805$ & 1521 & $0.715$ & 12646 \\
 & Dense (budget-matched) & $0.594$ & 425 & $0.753$ & 1338 \\
 & BM25 (budget-matched) & $0.701$ & 430 & $0.771$ & 1330 \\
 & Reader only (no state bound) & $0.758$--$0.760$ & 428 & $0.827$--$0.828$ & 1351 \\
 & \textbf{Resident cache + archive fallback} & $\mathbf{0.755}$--$\mathbf{0.760}$ & 428 & $\mathbf{0.813}$--$\mathbf{0.830}$ & 1368 \\
 & Oracle evidence (ceiling) & $0.804$ & 432 & $0.854$ & 1340 \\
\bottomrule
\end{tabular}
\caption{Held-out multi-hop QA results, with 300 questions in each length condition. Ranges show three fixed selector seeds; token counts are rounded whole-prompt averages over records and, for learned methods, all three seeds. The selector --- frozen-Llama hidden-state features, a learned two-hop reranker, a query-independent writer, and calibrated thresholds --- was developed with Llama and then applied to Qwen without re-tuning.}
\label{tab:trackb}
\end{table}

\begin{figure}[t]
\centering
\includegraphics[width=\linewidth]{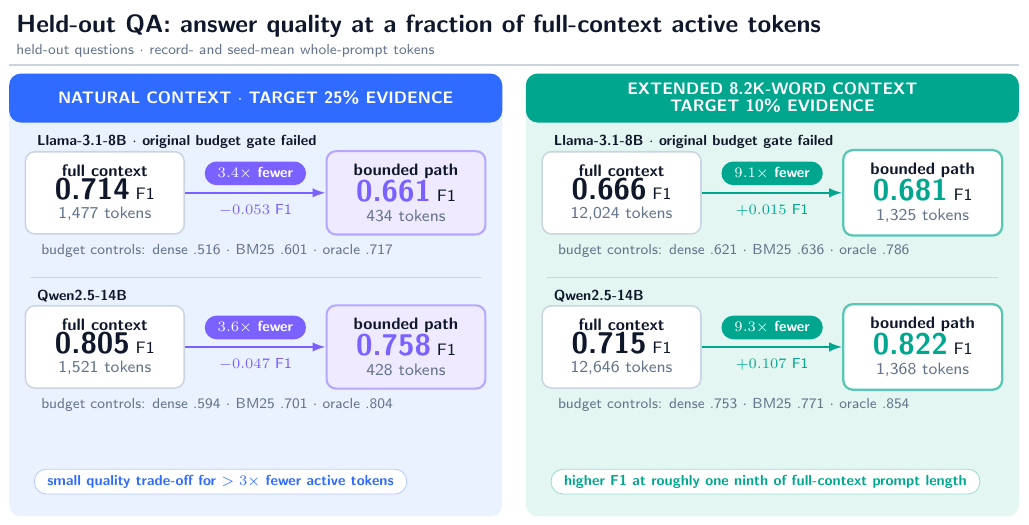}
\caption{Answer quality and active-prompt cost on the two held-out question sets. Each row compares the resident-cache plus archive-fallback path with the same reader given the full context; budget-matched dense, BM25, and oracle selectors are reported alongside. At natural length, selection trades a small amount of F1 for $3.4$--$3.6\times$ fewer prompt tokens. At 8.2k words, it improves F1 for both readers while using roughly one ninth as many prompt tokens. These counts exclude archive storage, indexing, retrieval latency, and retrieval energy.}
\label{fig:trackb}
\end{figure}

The comparison with budget-matched dense retrieval was fixed in advance. Across all selector seeds, the learned path improves F1 by $14.3$--$16.6$ points at natural length and $5.5$--$7.6$ points at extended length. The lower end of every paired-bootstrap 95\% confidence interval remains positive: at least $+8.9$ and $+1.45$ for Llama, and $+10.6$ and $+2.0$ for Qwen. That dense feature is a weak frozen-hidden baseline (train-only recall $0.14$), so the more informative margin is against BM25: $4.0$--$6.2$ F1. A trained modern retriever, answer-masked and no-evidence controls, and end-to-end archive cost remain necessary before making a general retrieval claim.

The long condition changes the usual cost--quality trade-off. Reading all 8.2k words lowers F1 for both readers ($0.714\!\to\!0.666$ for Llama and $0.805\!\to\!0.715$ for Qwen), consistent with previously reported lost-in-the-middle behavior \cite{lostmiddle2024}. Selecting at most $10\%$ of the evidence words instead reaches $102$--$103\%$ of full-context F1 for Llama and $114$--$116\%$ for Qwen. No part of the selector was re-tuned for Qwen, yet its margins are larger.

The reader-only rows clarify where the gain comes from. Removing the 32-slot bound does not reduce quality; on the extended set it is slightly better than the cache-plus-fallback path for Llama and roughly tied for Qwen. The learned selector creates the quality margin, while the resident cache preserves most of it. Calibrated fallback fires on $39\%$ of natural and $72\%$ of extended records and searches the retained full context, so total memory is not bounded by 32 slots. Because these passages never overwrite one another, this study supports selection and abstention with bounded hot state, not bounded total memory or the overwrite and protection claims established by the lifecycle study.

\section{Why Fallback Is Necessary When Relevance Is Unknown}
\label{sec:boundary}

\begin{figure}[!t]
\centering
\includegraphics[width=\linewidth]{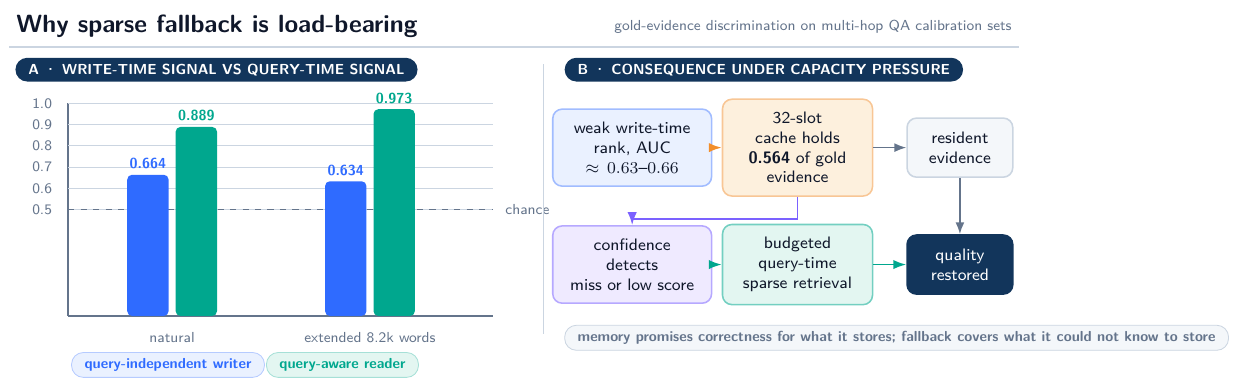}
\caption{Gold-evidence discrimination on calibration questions. Passage-only features provide little information about which evidence a future question will need. Once the question is available, the same selection problem becomes much easier. A bounded writer therefore needs an explicit way to abstain and retrieve at query time.}
\label{fig:boundary}
\end{figure}

Figure~\ref{fig:boundary} measures the asymmetry directly. A query-independent salience probe over passage-only features reaches AUC $0.664$ on natural records and $0.634$ on extended records, only modestly above chance. The query-aware reranker reaches $0.889$ and $0.973$, although part of the extended-set signal comes from the construction procedure. Under capacity pressure, the 32-slot cache consequently contains only $0.564$ of the gold evidence. Without fallback, every missing gold passage is unrecoverable; with a threshold calibrated at the deployment length, most of the selector's quality is retained. A bounded memory can enforce correct lifecycle behavior for resident information. It cannot promise to store an ordinary fact whose future relevance was unavailable when the write decision was made.

\section{Relational Binding with Factorized Anchor--Filler Transport}
\label{sec:afrt}

The lifecycle study relies on a deliberately controlled event parser. We therefore ask a narrower trainability question: once event boundaries and semantic role anchors must be predicted, what inductive bias lets a differentiable compiler bind each role to the correct filler under unseen compositions? Pointer networks motivate content-conditioned selection \cite{pointernet2015}, while relational models motivate explicit structural bias \cite{relationnet2017,relationalbias2018}. Our first group-disjoint compiler fit every training example but obtained $0/70$ held-out successes for role binding and whole-record execution. Optimization was sufficient; the content-matching readout did not compose. This failure motivated \afrt{} as a structured side path, not as a replacement for all self-attention.

\begin{figure}[t]
\centering
\includegraphics[width=\linewidth]{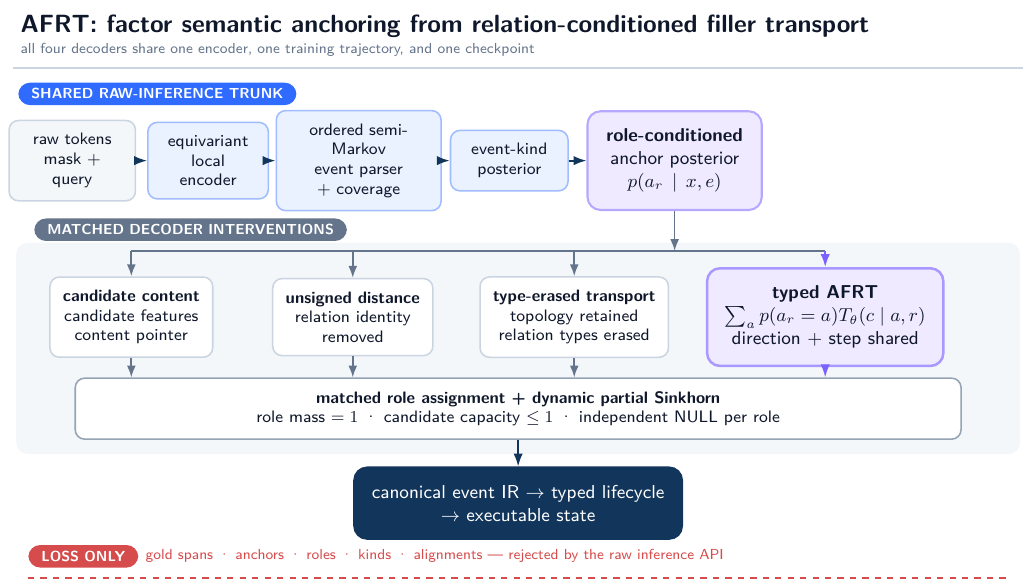}
\caption{\textbf{Relational-binding architecture and matched controls.} The raw encoder, semi-Markov event parser, event-kind head, role-anchor posterior, canonical intermediate representation, and typed lifecycle are shared. The four decoders use candidate content, unsigned distance, type-erased topology, or typed \afrt{} transport. Direction and path-step parameters are shared in the typed decoder. Gold spans, anchors, roles, kinds, and alignments enter the losses only and are absent from raw inference.}
\label{fig:afrt}
\end{figure}

For event $e$, role $r$, and candidate filler $c$, \afrt{} factorizes the pointer as
\begin{equation}
p(f_r=c\mid x,e)=\sum_a p(a_r=a\mid x,e)\,T_{\theta}(c\mid a,r,x,e),
\label{eq:afrt}
\end{equation}
where $p(a_r=a\mid x,e)$ is a predicted semantic-anchor posterior and $T_{\theta}$ transports its mass along registered relation paths. Transport depends on relation type, while the two directions and repeated path steps share parameters. Candidate logits are masked by predicted event coverage. They then enter a dynamic partial Sinkhorn assignment $P$. For the real candidate set $C$ and the private NULL candidate $\varnothing_r$ of each role,
\begin{equation}
\begin{aligned}
\sum_{c\in C}P_{rc}+P_{r\varnothing_r}&=1,\\
\sum_r P_{rc}&\leq1\quad(c\in C),\\
P_{r\varnothing_{r'}}&=0\quad(r\neq r').
\end{aligned}
\label{eq:partial-sinkhorn}
\end{equation}
Thus every role places one unit of mass, real fillers cannot be reused, and each role may independently remain unfilled. The typed decoder receives no filler-content feature, absolute position, signed-direction parameter, template or wrapper identifier, or gold span, anchor, role, kind, or alignment.

\textbf{Matched controls.} Each seed trains one model and publishes one final checkpoint. Four jointly trained decoders are read from it: a candidate-content decoder, an unsigned-distance decoder, a type-erased transport decoder, and the typed \afrt{} decoder. They share the raw encoder, event partition, coverage, kind and anchor heads, initialization, data order, optimizer, token exposure, and checkpoint. Their paired differences do not mix decoder structure with independently selected checkpoints. They do, however, permit co-adaptation through the jointly trained shared encoder, and none is a type-aware non-factorized control. The present comparison therefore identifies a need for typed topology relative to the registered ablations; it does not isolate factorization as the unique cause.

\textbf{Richer topology and fixed repetitions.} Each of three fresh data seeds contains 512 fitting records and 240 physically disjoint held-out records. The generator varies path length from 2--8, relation composition, distractor bridges, event count from 2--6, and optional or NULL roles. Six stress subsets contain 40 records each. Every model is trained for 32 complete epochs and 512 optimizer steps at effective batch 32. We fixed the three data/model seed pairs before training and did not use early stopping, choose among checkpoints, replace a failed seed, or tune a threshold after seeing a held-out result.

For decoder $d$ and seed $s$, the paired whole-record gain over a matched control $b$ is
\begin{equation}
\Delta_{d-b}^{(s)}=\sum_{i=1}^{240}
\left(\mathbf{1}[\hat y_{i,d}=y_i]-\mathbf{1}[\hat y_{i,b}=y_i]\right).
\label{eq:paired-delta}
\end{equation}
This quantity is positive only when the proposed decoder solves more of the same records than its control.

\begin{table}[t]
\centering
\small
\setlength{\tabcolsep}{4.2pt}
\begin{tabular}{lrrrr}
\toprule
Predefined check & Seed 0 & Seed 1 & Seed 2 & Required result \\
\midrule
Each of 11 typed-decoder output checks & 480/480 & 480/480 & 480/480 & $\geq432/480$ \\
Cross-view consistency & 240/240 & 240/240 & 240/240 & $\geq216/240$ \\
Whole-record exactness & 240/240 & 240/240 & 240/240 & $\geq216/240$ \\
Each of six stress subsets & 40/40 & 40/40 & 40/40 & $\geq34/40$ \\
Gain over candidate-content decoder & +240 & +240 & +240 & $\geq+24$ \\
Gain over unsigned-distance decoder & +240 & +240 & +240 & $\geq+18$ \\
Gain over type-erased transport & +240 & +240 & +240 & $\geq+12$ \\
All conditions met for this seed? & yes & yes & yes & yes \\
\bottomrule
\end{tabular}
\caption{Fresh three-seed confirmation on physically disjoint fitting and held-out records. The eleven output checks cover boundary, active anchor, present anchor, NULL anchor, role, kind, canonical intermediate representation, execution trace, state, resident selection, and payload exactness. Every matched control has zero whole-record successes on every seed. These are controlled synthetic results, not natural-language, long-context, or full-language-model results.}
\label{tab:afrt}
\end{table}

\begin{figure}[t]
\centering
\includegraphics[width=\linewidth]{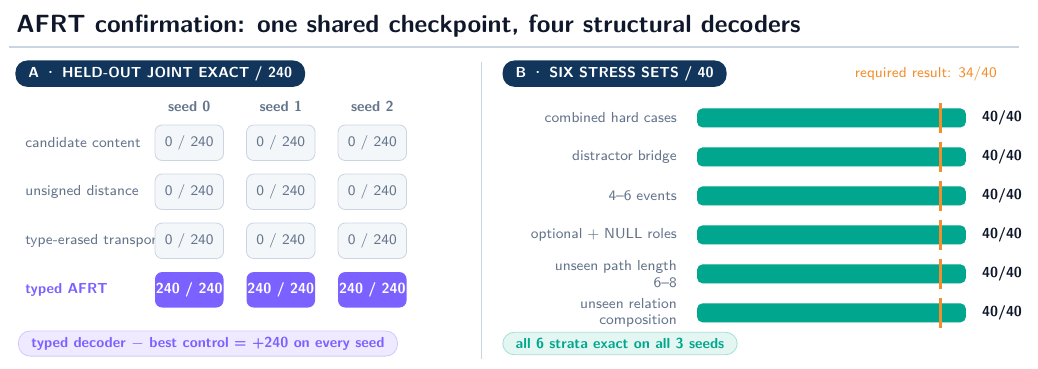}
\caption{Matched-control result for relational binding. Every decoder is read from the same checkpoint for a given seed. Typed \afrt{} is exact on all 240 held-out records for each of three fresh seeds; candidate-content, unsigned-distance, and type-erased transport decoders have no whole-record successes. The right panel separates the six predefined stress subsets.}
\label{fig:afrt-results}
\end{figure}

\textbf{Result.} We counted the hypothesis as supported only if every one of the three fixed seed pairs met every condition in Table~\ref{tab:afrt}; no failed seed could be replaced. All three did. Typed \afrt{} obtains 240/240 whole-record exactness and cross-view consistency on each seed, and all six stress subsets are 40/40. Each registered control has no whole-record successes, yielding $\Delta=+240$ against every control on every seed. The result supports a narrow causal statement: in this controlled topology and joint-training setup, typed relation-conditioned transport supplies a reproducible binding signal absent from content, unsigned-distance, and type-erased alternatives. A decisive factorization claim still requires a parameter-matched, type-aware non-factorized decoder and independently trained same-initialization/equal-compute repetitions. The result does not establish natural-language relation discovery, arbitrary graph reasoning, long-context scaling, or superiority to external sequence architectures.

\textbf{Evaluation contract.} The initial group-disjoint compiler and the first three-seed richer-topology study are not counted as demonstrations. The reported confirmation uses fresh data and model seeds. Payload exactness includes the separately stored lifecycle version; anchor equality is evaluated only on active events; and aggregate validation requires exact metric-key-set equality, so dictionary insertion order cannot affect acceptance. Missing or extra metrics fail closed. These definitions were fixed before training and before any confirmatory metric was observed.

\textbf{Originality boundary.} Ordered semi-Markov segmentation \cite{sarawagi2004semimarkov}, Sinkhorn assignment \cite{cuturi2013sinkhorn}, canonical intermediate representations, typed lifecycle execution, pointer-style selection \cite{pointernet2015}, and generic relational inductive bias \cite{relationalbias2018} are inherited tools. The proposed mechanism is the anchor-posterior/filler-pointer factorization in Eq.~\ref{eq:afrt}, filler-content-invariant typed transport with direction and step sharing, the partial assignment constraints in Eq.~\ref{eq:partial-sinkhorn}, and the same-checkpoint structural intervention. We do not attribute novelty to the standard components that make the mechanism executable.

\section{Evaluation Contracts and Failure-Preserving Analysis}
\label{sec:evaluation-contracts}

The reported analyses apply three rules. First, a benchmark must exercise the claimed mechanism rather than admit a trivial lexical solution. The controlled lifecycle generator therefore makes the last mention deliberately wrong in the alias, stale-remention, invalid-write, and protected-slot scenarios, while retaining capacity pressure and fallback cases.

Second, a failed registered constraint remains a failure even when quality metrics are favorable. The Llama multi-hop run used whole-prompt tokens divided by context tokens, thereby counting the fixed question and instruction tokens against the evidence budget. It exceeded its allowances: $29.4\%$ rather than $26.25\%$ at natural length and $11.0\%$ rather than $10.5\%$ at extended length. The separate evidence-word calculation is reported descriptively but does not convert that run into a budget success. The Qwen study uses the intended evidence-word definition with a hard per-record cap and is evaluated as a distinct follow-up.

Third, a corrected evaluator requires fresh confirmatory data when the correction changes the scientific predicate. The relational confirmation therefore uses fresh data and model seeds, includes lifecycle version in payload exactness, excludes inactive padded events from anchor equality, and requires all three fixed seed pairs to meet every predefined condition. Earlier compiler studies remain negative or non-confirmatory and are not rescored into positive evidence.

\section{Supporting Controlled Evidence (Corrected)}
\label{sec:controlled}

The integrated path builds on a longer sequence of controlled studies. A logic audit found nine issues, including a fixed target position, two evaluation shortcuts, an entity-identity split overlap, and invalid pooled confidence intervals over repeated use of the same examples. We corrected and reran the affected studies and report repetitions separately.

\begin{itemize}
  \item \textbf{2M-token synthetic stress (collision-free rerun):} explicit memory + sparse fallback scores $49/50$ at $2{,}097{,}152$ tokens with $\leq 132$ active chunks; fixed-state and sparse-only proxies each fail complementary scenario classes.
  \item \textbf{Trainability:} a 2.74M-parameter causal event-token backbone learns routing signals (199, 196, and 200 of 200 across three seeds with lite write supervision).
  \item \textbf{Frozen-hidden bridge:} all 18 predefined model--seed comparisons across six model families reach at least $57/60$ on entity-disjoint splits, showing that published-model hidden states contain lifecycle signal when key identity is supplied.
  \item \textbf{Controlled grounding and generation:} oracle-free five-digit lifecycle at $90/90$ per model/seed (Llama/Qwen); causal writer + hard lifecycle + generation at $90/90$; trained soft-token value compression is a documented negative (unstable $85$--$89/90$ across regularizers), with lossless span replay strictly better ($90/90/88$).
  \item \textbf{Open-domain selector precursor (IDF multi-hop):} a non-learned IDF multi-hop selector recovers structured variable-tracking chains ($+58.0$ points, CI $[49.8, 65.8]$) but fails open-domain HotpotQA ($-7.1$ F1, CI $[-14.6, 0.25]$), motivating the learned two-hop selector.
  \item \textbf{Differentiable compiler sequence:} the first compiler fits but does not generalize role binding; a single-seed same-checkpoint study isolates an $80/80$ typed-transport result against $0/80$ for content matching; the first three-seed richer-topology study remains failed; two audits identify evaluator defects; and a fresh three-seed confirmation tests the corrected question (Table~\ref{tab:afrt}).
\end{itemize}

\section{Negative Results}
\label{sec:negatives}

All artifacts are retained. Raw frozen-language-model hidden states are weak retrievers (layer-probe recall $0.14$; dense-only F1 $0.47$ at natural length) and become useful only as features in a learned reranker. An IDF-weighted bridge feature and two anchors did not improve that reranker. Presenting evidence in document order rather than score order cost six F1 points at extended length. Calibrating fallback at the wrong context length reduced evidence coverage from $0.88$ to $0.31$. Trained soft-token value compression never reached the predefined accuracy requirement and was replaced by lossless span replay.

Two earlier results are explicitly superseded or retained as failures. A first scaffold reported 0.02 accuracy without lifecycle control, but its tie-break deliberately preferred stale answers; Table~\ref{tab:tracka} uses a neutral tie-break and obtains 0.49. The first group-disjoint compiler fits its compact training distribution yet has no held-out role-binding successes. The first three-seed richer-topology study also remains failed even though later audits show that part of its apparent payload and anchor failure came from the evaluator. We answer the corrected question with fresh data and seeds rather than rewriting either verdict.

\input{studies/posttraining_evidence}
\input{studies/coverage_controlled_readout}

\section{Related Work}

\textbf{Efficient sequence backbones.} Linear attention, structured state spaces, recurrent hybrids, and long convolutions compress history into a recurrent or convolutional state \cite{lineartransformer2020,performer2020,s4_2021,retnet2023,rwkv2023,mamba2023,mamba22024,hyena2023,gdn2024,gdn22026}. Sparse attention limits the set of positions read by each query \cite{bigbird2020,nsa2025,msa2026,deepseekv322025}. Both directions primarily redesign token mixing. We study the semantics imposed on a bounded state when it must support overwrite, protection, eviction, and abstention.

\textbf{Kernel substrate.} FlashKDA is a high-performance CUTLASS implementation of Kimi Delta Attention with token-parallel and head-parallel recurrent updates \cite{flashkda2026}. Its recurrent state has shape $[B,H,V,K]$; current source documentation reports BF16 or FP32 state, FP32 update accumulation in the optimized path, automatic dispatch through Flash Linear Attention, and H20 kernel speedups over earlier KDA/GDN implementations. Those are upstream kernel measurements, not MMLA measurements. FlashKDA does not define semantic row authority, hindsight supervision, overwrite versus NULL, or lifecycle evaluation. It is nevertheless a plausible complementary substrate for a recurrent/linear MMLA branch if the state geometry and precision contract match. We have not integrated or benchmarked that combination.

\textbf{Addressable and external memory.} Memory Networks and Neural Turing Machines introduced differentiable addressable stores \cite{memorynetworks2014,ntm2014}. Recurrent-memory and virtual-context systems carry state across segments or calls \cite{rmt2022,memgpt2023}. Other Transformer memory systems use recurrence, compression, nearest-neighbor retrieval, retrieved text, or learned memory layers \cite{txl2019,compressive2019,memorizing2022,longmem2023,memorylayers2024,rag2020,retro2021,knnlm2019,titans2025,infini2024}. H2O, Scissorhands, SnapKV, and Landmark Attention retain or index selected KV-cache entries \cite{h2o2023,scissorhands2023,snapkv2024,landmark2023}. Sparse memory over static documents (MSA \cite{msa2026}) is a relevant retrieval comparator, not the sole nearest system. Our controlled experiments exercise same-key versioning, protected state, explicit eviction rules, and calibrated abstention when future relevance is unavailable at write time. The closer learning-and-memory comparisons in \S\ref{sec:nearest-neighbors} delimit which of these operations and learning signals already have direct precedents.

\begin{table}[H]
\centering
\small
\setlength{\tabcolsep}{4.2pt}
\renewcommand{\arraystretch}{1.10}
\begin{tabularx}{\linewidth}{@{}p{0.16\linewidth}YY@{}}
\toprule
Dimension & MMLA & FlashKDA \cite{flashkda2026} \\
\midrule
Problem & Future-use-aware bounded persistence & Fast delta-attention recurrence \\
Persistent state & Proposed $S$ authoritative rows plus a separate policy carrier & Recurrent matrix $[B,H,V,K]$ \\
Operation & Proposed complete-row overwrite or NULL per event & Numerical recurrent update, not semantic overwrite \\
Training & Component and latent-readout training tested; grouped-future admission untested & Kernel implementation; no memory-policy objective \\
Inference update & Proposed causal admission model; tested readout uses an external rule writer and oracle selection & Fused KDA state transition \\
Evidence / limit & Controlled and restricted readout successes; generalized latent loop unqualified; no learned predictive-overwrite result & Upstream H20 kernel results only; no MMLA integration or lifecycle semantics \\
\bottomrule
\end{tabularx}
\caption{Architecture--kernel comparison of two complementary objects, not a leaderboard. ``Operation'' distinguishes numerical state evolution from semantically authoritative memory mutation. Proposed MMLA system behavior is not a tested kernel integration.}
\label{tab:systems-comparison}
\end{table}

\textbf{Predictive state and consolidation.} Predictive state representations describe latent state through predictions of future observations rather than a reconstruction of history \cite{littman2001psr}. Predictive information formalizes the mutual information shared by past and future \cite{bialek2001predictive}, and the information bottleneck asks for a compact code that preserves a designated relevant variable \cite{tishby2000ib,alemi2016vib}. Complementary learning systems separate fast episodic storage from slower integration into distributed parameters \cite{mcclelland1995cls}. World models and latent-imagination agents learn compressed dynamics for prediction and control \cite{worldmodels2018,dreamer2020}. Our construction combines these motivations in a narrower language-model operation: grouped actual futures estimate the conditional action value of one bounded overwrite, then a causal value model predicts those values for deployment. We do not claim novelty for predictive state, value estimation, or consolidation in isolation.

\textbf{Retrieval and relational binding.} The QA study combines BM25 \cite{bm251995} with dense-retrieval-style features \cite{dpr2020}, evaluates on HotpotQA \cite{hotpotqa2018} at LongBench-scale lengths \cite{longbench2023}, and interprets the full-context comparison in light of lost-in-the-middle behavior \cite{lostmiddle2024}. The relational study draws on pointer selection \cite{pointernet2015}, relational modules \cite{relationnet2017,relationalbias2018}, ordered semi-Markov inference \cite{sarawagi2004semimarkov}, and Sinkhorn normalization \cite{cuturi2013sinkhorn}. We claim neither the inherited components nor a general graph reasoner; the tested mechanism is their anchor-posterior/typed-transport factorization under same-checkpoint content, distance, and type-erasure controls. RULER supplies complementary long-context stress tests \cite{ruler2024}; Llama and Qwen are frozen readers \cite{llama32024,qwen252024}, served with vLLM \cite{vllm2023}.

\input{studies/nearest_neighbors}

\section{Reproducibility Scope}

The controlled lifecycle construction uses split seeds 202607091--93 and writer seeds 202607094--96; its secondary generation comparison uses writer seed 202607094. The HotpotQA construction excludes exact and high-overlap question duplicates against the diagnostic pool; extended contexts use seeds 202607098/99. Selector training uses seeds 202607102--104, with bootstrap seeds 20260709/20260710. The relational confirmation uses data seeds 2026080411--13 and model seeds 2026080421--23. Splits, thresholds, support conditions, and final-only evaluation rules are specified in the corresponding method and protocol sections. The later readout studies retain separate models, seeds, parent weights, data strata, and evaluation units (\S\ref{sec:posttraining-evidence}--\ref{sec:coverage-readout}); they are not a pooled benchmark. The v4 reconstruction package includes the full document source, bibliography, figure assets, build instructions, and a neutral aggregate count table. Raw training datasets, weights, runtime logs, private review records, and executable scientific packages are not included. The historical paper repository is \url{https://github.com/MMLA-org/mmla-memory}; Appendix~\ref{app:version-method} distinguishes the earlier public artifacts from this complete revision and documents availability limits.

\section{\texorpdfstring{Discussion: Learning Beyond\\Context Length}{Discussion: Learning Beyond Context Length}}

Ultra-long context remains a stringent stress test, but it is not the final objective. A capable adaptive system should improve the appropriate state as evidence arrives rather than keep replaying an ever-growing transcript or indiscriminately changing all weights. The distinction matters for long-running assistants, agents that revisit a changing world, and any model expected to learn during use. In the PDSA view, policy-state quality is measured by isolated within-problem improvement, exact reset, and bounded drift; authoritative-memory quality is measured by revision, preservation, deletion, reuse, and future consequence. Neither is measured by the largest accepted token count.

The architecture also suggests a testable scaling hypothesis. Let $G^*(c)$ be the future-risk reduction available to an oracle memory policy at model capability $c$, and let $\epsilon_Q(c)=\sup_{s,a}|\widehat Q_c(a\mid s)-\overline J_c(a\mid s)|$ be the action-value error. The greedy decision bound in Appendix~\ref{app:stability} gives
\begin{equation}
G_{\mathrm{value}}(c)\ge G^*(c)-2\epsilon_Q(c).
\label{eq:capability-scaling}
\end{equation}
If stronger models both extract more value from correct memories and estimate conditional action values more accurately, $G^*(c)$ may rise while $\epsilon_Q(c)$ falls. Memory could then become more useful as the model becomes more capable, rather than merely compensating for a weak backbone. This is an empirical scaling hypothesis.

Persistent, revisable state may be useful for long-running adaptive systems, but the present report evaluates only the specified memory and policy mechanisms.

\section{Limitations and Next Steps}

\begin{figure}[H]
\centering
\includegraphics[width=\linewidth]{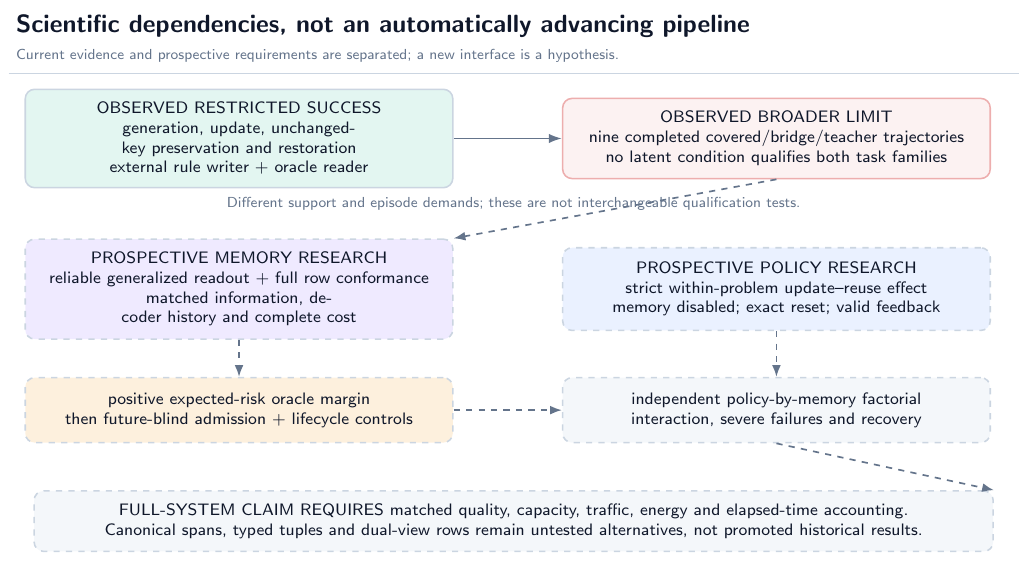}
\caption{\textbf{Scientific dependency roadmap at the fixed cutoff.} Restricted readout evidence does not imply generalized readout, learned admission, a policy-only effect, or a joint PDSA advantage. Dashed branches are prospective research questions, not newly executed studies.}
\label{fig:vision-roadmap}
\end{figure}

The lifecycle and multi-hop QA experiments use fixed language models; only the writer, reranker, and thresholds are trained. The versioning language is controlled and shares cue words across otherwise disjoint sentence frames, so it does not establish open-language event parsing. The QA passages are static and exercise only the bounded cache and fallback, not same-key overwrite. Its natural-length budget may also require truncating a passage in very short contexts, and passage-level selection leaves sentence-level budgeting unexplored.

The relational experiment closes one binding failure in a compact synthetic compiler. Its perfect result is tied to the registered generator, seen relation primitives, short sequences, and a jointly trained shared encoder. It is not evidence that \afrt{} improves a full language model, that factorization is uniquely responsible, or that the model discovers arbitrary relations in natural text. The historical interface studies in Tables~\ref{tab:interface-negatives} and~\ref{tab:semantic-interface-qualification} show that physically stable rows can remain semantically unusable. The fitted-support fault-localization study (Table~\ref{tab:fitted-support-localization}) finds failure already on fitted examples despite target-row isolation. These outcomes remain valid for those checkpoints and protocols. Later post-training studies establish narrower successes under different interfaces and curricula; they neither rerun nor reverse the historical tests. In the latest completed comparison, all three latent conditions fail generalized continuous-event qualification despite strong text reference and exact restoration. Teacher and student decoder histories differ, so that gap cannot be attributed solely to representation space. The predictive architecture adds further open questions: whether conditional future utility is predictable from the past, whether completed-segment bidirectionality helps beyond equal-compute causal encoding, whether learned payload construction can be separated from admission, and whether branch-generation cost is amortized by deployment savings.

Persistent writable memory also creates a security surface that the present experiments do not test. Relevant failures include poisoning a candidate so that it survives consolidation, exhausting slots through malicious writes or locks, cross-user leakage through shared resident state or archives, alias or routing collisions, unauthorized deletion, and rollback that resurrects stale values. A deployable system needs provenance, tenant isolation, write-rate limits, protected-capacity policy, auditable tombstones, deletion verification, and recovery semantics. These are system requirements, not consequences of the information-bottleneck objective.

The next scientific choice should be informed by the completed comparison, not by treating its negative result as proof that any one new interface must work. Covered supervision, added full-width mapping capacity, and this fixed output-alignment objective have now been compared together; none qualifies the broader task. A text/latent decoder-history-matched comparison, token-preserving alternatives, canonical spans, typed tuples, and a dual-view row remain hypotheses with different costs and information access. Any future comparison needs a separately frozen intervention, matched training support and compute accounting, all subjects, and independent validation. No proposed successor retroactively promotes the older interface or changes its threshold.

Separately, the policy plane lacks a memory-disabled within-problem RTT intervention with an exact carrier and backbone. A favorable result there would identify a different mechanism, not repair a latent readout failure. Predictive admission additionally requires a registered expected-risk oracle margin over supported slot actions and NULL from matched snapshots before a future-blind value model is scientifically justified. Such a controller must report action regret, NULL behavior, uncertainty, stale leakage, collateral damage, and write rate. A full PDSA benefit claim then requires a policy-by-memory factorial, complete lifecycle and recovery checks, and a cost comparison including backward/optimizer/reset work, archive storage, indexing, retrieval, branch replay, kernel count, memory traffic, latency, and energy. These are future research requirements; this report preparation executes none of them.

\section{Conclusion}

MMLA now names a Memory-Mediated Learning Architecture rather than one long-context attention layout. Its core mechanism, Predictive Dual-State Adaptation, asks two related but nonidentical questions: should feedback change the numerical policy used by a later attempt, and should a completed observation receive authority to persist? A causal model reads a pinned pair; the policy carrier may accept a bounded update or exact no-op; the memory carrier may reinterpret a completed segment and commit one complete row or NULL; training-only futures may price the joint consequence; and deployment remains future-blind. Finite typed carriers make bounded execution possible. They do not guarantee learnability, efficiency, safety, or intelligence.

The completed evidence establishes useful components. Explicit lifecycle execution resolves overwrite, protection, and eviction in controlled language; calibrated archive fallback handles evidence whose relevance was unknowable at write time; typed transport repairs one controlled relational-binding failure relative to registered ablations. Historical frozen-checkpoint semantic failures remain failures, including a fitted-support bottleneck under a routing oracle. Later post-training separates candidate ranking from free generation and establishes restricted generation, update, and restoration with content controls. The latest coverage-controlled study completes nine formal trajectories but does not qualify generalized continuous-event latent readout; bridge and teacher alignment provide no registered material improvement, while text reference and restoration pass. This is evidence about the tested curricula and interfaces, not a universal impossibility result.

The unresolved full-system claims are explicit: broadly reliable semantic readout with authoritative-row conformance, a strict policy-only RTT effect, a positive expected-risk admission oracle and learned future-blind policy, and an independently identified policy-by-memory advantage with complete resource and recovery accounting. Neither a proposed canonical/dual-view representation nor a conditional theorem supplies those observations. MMLA/PDSA is therefore a formal architecture with differentiated component evidence and falsifiable remaining claims, not a completed demonstration of adaptive superiority.

\input{theory_family/modules/index}

\mmlaboundary

\section*{AI-Assisted Tooling Disclosure}

AI-assisted tools were used for code assistance, language editing, evidence
cross-checks, figure preparation, and build checks. The author list and
contribution statement are preserved from the complete integrated manuscript.
This archival revision does not itself record final publication authorization;
the authors retain responsibility for the claims and release decision.

\section*{Author Contributions}

Junyi Zou and Avrova Donz contributed equally to conceptualization,
methodology, formal analysis, experiments, visualization, and manuscript
preparation.

\begingroup
\small
\bibliographystyle{plain}
\bibliography{references,theory_family/references_formal,near_neighbors}
\endgroup

\clearpage
\appendix
\small
\titlespacing*{\section}{0pt}{10pt plus 2pt minus 1pt}{4pt}

\section{Experiment Roadmap}
\label{app:phases}

\begin{table}[H]
\centering
\footnotesize
\begin{tabular}{p{0.30\linewidth}p{0.62\linewidth}}
\toprule
Experiment & Core question \\
\midrule
Five-task slot-memory prototype & Can explicit slots solve five synthetic memory tasks?  \\
Token-to-chunk neural interface & Do neural key/value/write signals survive tokenization and chunking?  \\
128K matched architecture comparison & How does the hybrid compare to Delta/GDN/KDA, DSA/NSA, and MSA proxies?  \\
Trainable causal event encoder & Can a causal GRU encoder replace hand-crafted event vectors?  \\
Two-million-token collision-free stress & Does the hybrid hold at two million tokens without key collisions?  \\
Generated-language boundary check & Does generated natural language preserve the synthetic boundaries?  \\
Local long-context benchmark replication & Do Needle/RULER-style local probes replicate the hybrid pattern?  \\
Official benchmark pipeline check & Can the pipeline ingest LongBench/RULER and produce valid predictions?  \\
Frozen-model lexical diagnostics & What do real Llama/Qwen models reveal about simple lexical selectors?  \\
Small trainable memory backbone & Can a 2.74M-parameter model jointly train fast state, slots, and sparse fallback?  \\
Six-family frozen-representation study & Can frozen hidden states from six model families support a memory lifecycle when key identity is supplied?  \\
Learned grounding without oracle keys & Can learned parsers replace oracle key IDs under controlled syntax?  \\
Controlled generation from selected memory & Can selected memory drive frozen-LLM generation, causal lifecycle, soft injection, and learned value-span selection?  \\
First open-domain selector & Does a non-learned IDF multi-hop selector transfer from structured chains to real HotpotQA text? (No: $-7.1$ F1.)  \\
Controlled lifecycle path & Do query-independent writing, lifecycle execution, and fallback each contribute when lexical retrieval is wrong by construction? (\S\ref{sec:tracka})  \\
Held-out multi-hop QA & Does bounded selection beat budget-matched retrieval at two context lengths and with two frozen reader models? (\S\ref{sec:trackb})  \\
Relational compiler & Can typed anchor-to-filler transport repair content-binding failure across fresh seeds, richer paths, distractors, multiple events, and NULL roles when all decoders share a checkpoint? (\S\ref{sec:afrt}) \\
Single-realization consolidation qualification & Can local consolidation, a detached one-future target, a future-blind controller, and one payload-row overwrite execute against the same completed language-model checkpoint? (Yes mechanically; insufficient for conditional-risk or full-state claims.) \\
External semantic adapter (external-semantic-adapter study) & Can a compact external current-state codec recover, query, and preserve the frozen memory state? (No under its registered three-seed contract; mapping invariance and old-value suppression pass.) \\
Raw representation probe (raw-representation study) & Is the frozen hidden representation directly decodable before choosing a denser or more native interface? (No under its registered three-seed contract; downstream interface arms remain unopened.) \\
Direct interface comparison (frozen-checkpoint semantic-interface study) & Which of dense row, structured span, or true checkpoint-native readout yields a usable current-state interface under one matched frozen checkpoint? (Complete: all nine fixed jobs failed semantic qualification while structural mapping passed; no interface or arm comparison qualified.) \\
Fit-versus-generalization diagnostic (fitted-support fault-localization study) & Does the frozen-checkpoint semantic-interface study failure already appear on fitted examples, emerge only on fresh composition, or disappear under routing or content oracles? (Complete: fit/support fails; routing oracle plus learned content is $0/73{,}728$, so the same latent-row interface stops.) \\
Matched explicit-row qualification & Can canonical token spans, typed tuples, or a dual-view row pass a new matched semantic contract? (Proposed, not executed; not a substitute for the separate post-training evidence.) \\
Same-checkpoint expected-risk intervention & Do identical-prefix grouped futures and action-conditioned replay support useful future-blind admission under equal-compute controls? (Unexecuted; no positive oracle margin.) \\
Post-trained readout progression & Which budgets, objectives, prompts and interfaces improve candidate following, free generation and update/restoration? (Completed restricted successes and broader failures; Section~\ref{sec:posttraining-evidence}.) \\
Coverage-controlled latent readout & Do matched layout, a full-width bridge or frozen text-teacher alignment qualify broader continuous-event readout? (Nine trajectories complete; no latent arm qualifies both families; Section~\ref{sec:coverage-readout}.) \\
Matched from-scratch pretraining & If the intervention succeeds, does predictive consolidation improve quality and cost across model scales and long-running natural streams at equal token and hardware budgets? \\
\bottomrule
\end{tabular}
\caption{Experiment roadmap in execution order. Mechanical qualification, semantic-interface identification, conditional-risk overwrite learning, and from-scratch scale studies are deliberately separated. Scale remains conditional on both a usable interface and a deployable gain in the corrected intervention.}
\label{tab:phases}
\end{table}

\section{Extended Related Work}
\label{app:related}

Linear attention and efficient sequence models reduce attention cost by replacing quadratic softmax attention with recurrent, kernelized, state-space, or long-convolution states \cite{lineartransformer2020,performer2020,s4_2021,retnet2023,rwkv2023,mamba2023,mamba22024,hyena2023}. Delta-style models and recent hybrid architectures improve the expressiveness and update dynamics of fixed states \cite{gdn2024,gdn22026,kimi2025}. VLA frames linear attention as stable associative memory \cite{vla2026}. Our focus differs: we do not only stabilize the state; we add explicit memory lifecycle management outside the fast state.

Long-term and neural memory systems augment sequence models with differentiable stores, recurrence, memory layers, or retrieved memories \cite{memorynetworks2014,ntm2014,txl2019,compressive2019,rmt2022,memorizing2022,longmem2023,memorylayers2024,titans2025,infini2024,memgpt2023}. RAG-style and nearest-neighbor methods retrieve external text or hidden states \cite{rag2020,retro2021,knnlm2019}; KV-cache methods retain high-value tokens or landmarks during inference \cite{h2o2023,scissorhands2023,snapkv2024,landmark2023}. These works motivate memory augmentation; our studied unit is request-local editable memory with explicit overwrite, protection, eviction metadata, and calibrated abstention executed in the evaluation path.

Sparse attention systems reduce the attended set through fixed or learned patterns, from BigBird's structured sparsity to NSA's hierarchical selection, DeepSeek's learned sparse indexer, and MSA's document/chunk memory retrieval \cite{bigbird2020,nsa2025,deepseekv322025,msa2026}. RACE replaces softmax similarity with random projections and soft locality-sensitive hashing \cite{race2025}. Sparse retrieval complements explicit lifecycle management: it is useful when relevance is unavailable at write time, but does not by itself define versioning semantics. RULER and LongBench pressure-test these distinctions \cite{ruler2024,longbench2023}; HotpotQA supplies multi-hop supervision \cite{hotpotqa2018}.

\section{Controlled Lifecycle Protocol Details}
\label{app:tracka}

\textbf{Scenarios.} Each record has a target key with $2$--$4$ versioned updates, distractor-key updates, and filler prose; sentence frames are disjoint across train/dev/test while cue phrases (update, mention, reject, lock, and alias markers) are shared --- the disclosed controlled-language boundary. Scenario semantics: \texttt{alias-query} defines the alias once (``\emph{X is an alias for Y}'') and the question names only the alias, which never appears in update lines; \texttt{stale-remention} appends audit lines (``\emph{an earlier revision listed Y as v}'') after the final update; \texttt{invalid-write} appends rejected change requests with fresh values; \texttt{protected-slot} freezes the key at version $k$ (``\emph{locked at $v_k$; subsequent writes are void}'') followed by later void updates; \texttt{slot-pressure} writes 48 distinct keys against the 32-slot bound. Generator unit tests assert that the lexical last-key baseline returns exactly the designed wrong value (the stale, rejected, or void value, or nothing for aliases) on the four breaking scenarios.

\textbf{Writer.} Multinomial logistic regression over hashed bag-of-words features of the line with keys, values, and digit runs masked ($256$ dimensions; class-weighted SGD). The feature function accepts only the line text; a unit test asserts two lines differing only in keys/values/step numbers produce identical features. Dev macro-F1 is $1.0$ across split-disjoint frames; a shuffled-label control collapses to $0.069$.

\textbf{What counted as support.} Before generating the data, we required the complete path to beat the strongest budget-matched hand-written rule by at least three exact-match points on every seed, with the lower end of a paired-bootstrap 95\% confidence interval above zero. Its stale, rejected, and void-answer rate also had to be at most half that of the learned writer without lifecycle execution. Accuracy on protected values had to reach 0.90. Fallback had to activate on at least 80\% of memory misses and recover at least 90\% of the capacity-pressure accuracy achieved by the unbounded lexical rule. Finally, occupancy could never exceed 32 and the writer interface had to remain query-independent.

Before the single held-out read, training and calibration checks also required adequate headroom, failure under shuffled labels, parser precision and recall of at least 0.99, successful one-batch overfitting, no split overlap, and evidence that every component was actually exercised. The runner was tested end to end on calibration data, committed, and then executed once. A completion marker makes a second execution refuse.

\section{Held-Out Multi-Hop QA Protocol Details}
\label{app:trackb}

\textbf{Contamination control.} The history ledger records every LongBench record ever touched by earlier phases and freezes the local LongBench HotpotQA pool as diagnostic-only (20 early label-tuned record IDs are unrecoverable). Holdouts derive from the official HotpotQA training parquet: 90{,}447 rows scanned; 13 in-split duplicates, 0 exact and 2 near-duplicate (token-Jaccard $\geq 0.8$) question overlaps against all 500 local LongBench questions excluded; 1{,}200/300/300 train/dev/holdout frozen by seeded hash rank.

\textbf{Length-matched extension.} Each holdout/dev record is extended to $\geq 8{,}000$ words by inserting whole distractor passages drawn from \emph{other} train records (pool of 11{,}713 deduplicated passages), excluding any passage whose normalized text contains the record's normalized answer (token-boundary match) or whose title collides with an original passage; original passages, including gold, are shuffled to seeded random positions. Original passages and supporting sentences are verified verbatim-present per record.

\textbf{Selector stack.} Dense features are frozen Llama-3.1-8B layer-32 hidden states (attention-masked mean pooling, L2-normalized; the layer was chosen by a train-only retrieval probe --- raw hidden states are weak retrievers, recall $0.14$, and act only as features). The reader is a logistic reranker over (dense cosine, BM25, lexical overlap, title-in-query, position, length) with a second hop round anchored on the top-1 pick (max-cosine and title-link bridge features), trained on train gold-passage labels, three seeds. The query-independent writer scores passages from position/length/capitalization/digit/title statistics only and fills 32 slots.

\textbf{Calibration in the deployment length regime.} The fallback threshold (confidence = mean of top-2 reader scores over memory) is chosen on a 200-record \emph{train-derived extended} slice: calibrating at natural length, where 32 slots hold all $\sim$10 passages, yields a threshold at which fallback never fires and extended coverage collapses from $0.88$ to $0.31$ --- a recorded negative. The slice's natural-length labels were seen by the reader during training. This overlap breaks calibration independence; its net effect on the selected threshold, fallback frequency, quality, and cost was not separately identified.

\textbf{Evidence budgets and per-seed results.} The budgets were fixed from calibration measurements logged before the selector existed. At $25\%$ of an 8.2k-word context, all selectors nearly saturate evidence recall (BM25 $0.970$); at $10\%$, BM25 reaches $0.867$ and the learned two-hop selector $0.956$. A 10\% budget is too small at natural length, where even oracle selection falls to F1 $0.60$, so that condition uses 25\%.

For Llama, the three seed-wise gains over dense retrieval are $+14.64/+14.44/+14.31$ F1 at natural length, with confidence-interval lower ends $+9.38/+8.99/+8.94$, and $+5.98/+6.27/+5.52$ at extended length, with lower ends $+1.66/+1.96/+1.45$. Qwen gains $+16.55/+16.48/+16.04$ and $+7.10/+7.64/+5.99$, with lower ends at least $+10.6$ and $+2.00$. Relative to reading the entire context, Llama retains $92.8/92.5/92.4\%$ of F1 at natural length and reaches $102.3/102.8/101.7\%$ at extended length. Qwen reaches $94.3/94.3/93.7\%$ and $115.2/116.0/113.7\%$.

\textbf{The two budget readings.} Under the frozen whole-prompt-token definition, the Llama run uses $29.4\%$ against a $26.25\%$ allowance at natural length and $11.0\%$ against $10.5\%$ at extended length, so it fails that requirement. Counting selected evidence words instead, the extended per-record maximum is exactly $10.00\%$. Natural records average $22.57\%$, but the original packer's first-passage exception reaches $61.36\%$ on one record. The forward-looking Qwen follow-up uses a hard per-record cap and truncates the top passage only if no whole passage fits; both length conditions then remain at or below their stated evidence budgets.

\section{Relational-Binding Protocol Details}
\label{app:afrt}

\textbf{Data geometry.} The fresh confirmation uses data seeds 2026080411--13 and model seeds 2026080421--23, all disjoint from the failed earlier three-seed study. Each seed contains 512 \texttt{train\_fit} and 240 \texttt{fresh\_train\_heldout} records with unique record and group namespaces across fitting data, held-out data, and seeds. The held-out records are balanced into six 40-record subsets: combined hard cases, distractor bridges, 4--6 events, optional or NULL roles, unseen path lengths 6--8, and unseen relation compositions. Relation primitives are shared across fitting and held-out data. The question is whether seen primitives compose in new ways, not whether the model discovers unseen relation semantics.

\textbf{Shared model and four decoders.} A translation-equivariant local encoder feeds an exact-count ordered semi-Markov event parser \cite{sarawagi2004semimarkov}, predicted event coverage, an event-kind head, and a role-conditioned anchor posterior. Candidate-content, unsigned-distance, type-erased transport, and typed \afrt{} decoders are trained jointly and read from one model-only checkpoint per seed. Dynamic partial Sinkhorn assignment inherits differentiable matrix scaling \cite{cuturi2013sinkhorn}; we do not claim Sinkhorn itself as novel. Raw inference accepts tokens, masks, and queries only. Gold spans, anchors, roles, kinds, and cross-view alignments are loss-only inputs and are rejected by the raw API.

\textbf{Training and what counted as support.} Training consists of 32 complete fitting epochs, 512 AdamW steps \cite{adamw2019}, and effective batch 32, without a scheduler, dropout, early stopping, or checkpoint choice. For each seed, every one of eleven typed-decoder output checks had to reach $432/480$; cross-view consistency and whole-record exactness each had to reach $216/240$; and every 40-record stress subset had to contain at least 34 exact records. The paired whole-record gains over the candidate-content, unsigned-distance, and type-erased decoders had to reach $+24$, $+18$, and $+12$, respectively. We would regard the overall hypothesis as supported only if all three fixed seeds met every condition, with no seed replacement.

\textbf{Evaluator definitions and accepted result.} Payload exactness compares both the resident payload and the separately stored lifecycle version. Anchor exactness is restricted to active events and reports present and NULL roles separately; inactive padding is excluded. The four decoder conditions are candidate content, unsigned distance, type-erased transport, and typed transport. The aggregate is complete, all three fixed seeds meet every condition, and both Dev/Test touched flags are false.

\section{Evaluation Validity Checks}
\label{app:audit}

The final controlled analyses enforce randomized target positions, query-independent writing, bounded evidence access, entity-disjoint representation splits, per-seed reporting, and collision-free filler/target construction. Evaluators treat full-context access and generator-template matching as non-memory controls rather than bounded-memory methods. The two-million-token collision-free stress test obtains $49/50$ under these definitions.

For relational binding, payload exactness includes lifecycle version, active-anchor equality excludes inactive padding, and aggregate validation uses exact metric-key sets. The confirmatory study uses fresh fitting and held-out namespaces and fresh data/model seeds. Earlier non-confirmatory compiler results remain negative and are not promoted by the final confirmation.

\section{Public Reproducibility Pointer}
\label{app:repro}

The manuscript and currently public paper artifacts are available from \url{https://github.com/MMLA-org/mmla-memory}. Availability of code, protocols, data, checkpoints, and execution logs is specified explicitly in the accompanying artifact manifest. The scientific split, seed, metric, threshold, and stop-rule definitions needed to interpret the reported numbers are stated in this report.

\section{Interface Qualification and Conditional Expected-Risk Protocol}
\label{app:predictive-protocol}

\textbf{Historical checkpoint binding.} The ancestor implementation used one frozen model-only checkpoint from the completed real-data pilot. All 295 tensor keys loaded with no missing or unexpected key, and every backbone parameter remained frozen. The project's ``160M'' label describes the hidden-stack configuration; the serialized model contains 352{,}347{,}121 parameters including the large vocabulary interfaces. This distinction prevents an architecture-size label from being mistaken for an exact parameter count.

\textbf{Mechanical ancestor and its boundary.} A parameter-free memory reader is exactly identical to the frozen language-model head when persistent state is empty. The local consolidator sees only a completed 64-token segment. The ancestor implementation evaluates each registered slot and a private NULL action from one immutable snapshot, detaches a target, rejects future-bearing deployment arguments, and changes one payload row or none. Thirty-five focused CPU tests and a one-record finite-gradient probe qualify software plumbing only. The code uses one realized continuation, converts its risks into a Gibbs policy, and applies memory after a cached future representation. It therefore neither estimates Eq.~\ref{eq:conditional-risk} nor identifies the effect of an action on future hidden-state evolution.

\textbf{External-adapter and raw-representation terminal evidence.} The external-semantic-adapter study trained the adapter at three fixed seeds and evaluated 393{,}216 raw rows. The resulting reductions appear in Table~\ref{tab:interface-negatives}; utility and predictive-admission experiments remained unopened. The raw-representation study trained a probe at three new fixed seeds and recorded the gate vector $(0,0,0,0,0,1,1)$ for every seed: all semantic gates failed, while both mapping gates passed. These terminal outcomes are retained rather than repaired or rerun.

\textbf{Frozen-interface qualification.} The frozen-checkpoint semantic-interface study compared dense row, structured $4\times192$ span, and checkpoint-native readout with active parameter counts $1{,}541{,}377/1{,}496{,}577/1{,}460{,}737$. It fixed three seeds, one Latin order, 2{,}048 updates per job, and final-only checkpoints. Predictions were materialized before evaluator truth, and all 55,296 qualification rows were reduced under the same registered gates. Every job was unqualified: candidate exactness was 0--45/23,040, query exactness 10--1,536/9,216, and mean record-macro Brier 0.998676--0.999093. Old-value suppression and mapping invariance passed. The study closed without selecting an interface or defining a bootstrap arm comparison.

\textbf{Fit-support localization terminal.} The fitted-support fault-localization study reused the nine fixed frozen-interface final checkpoints without training updates, checkpoint selection, seed replacement, threshold changes, or arm preference. It materialized target-free probability vectors before evaluator truth and evaluated 59,392 diagnostic rows per job, 534,528 in total. Historical fresh summaries were used only after the diagnostic results were fixed.

The exact reductions are reported in Table~\ref{tab:fitted-support-localization}. Learned content obtains $0/73{,}728$ exact with learned routing and $0/73{,}728$ with the target-row oracle. Oracle content plus learned routing obtains $18{,}544/73{,}728$, exactly the row-routing count; the fully oracle-specified path is $73{,}728/73{,}728$ by construction. Direct candidate reconstruction is $40/73{,}728$. Historical fresh content-bearing probes remain near zero, while map consistency is $82{,}944/82{,}944$ and old-value suppression is $13{,}824/13{,}824$. This selects the preregistered fit/support-failure branch. It does not identify which of representation geometry, decoder design, and optimization is causal, cannot rescue the frozen-interface qualification, and cannot select an interface retrospectively.

\textbf{Historical conditional expected-risk successor.} This unexecuted stage remained closed by the fit-support terminal. Its template required a new interface to qualify at every fixed seed before each future experiment group could contain one identical observed prefix and four to eight continuations with frozen weights. Group identifiers, not individual branches, would be physically separated across fitting, calibration, and fresh held-out namespaces. The generator would remove direct lexical cues to the correct write and include strict replacement, stale evidence, preservation of unrelated facts, deletion, delayed relevance, multi-hop consequences, ambiguous prefixes, and groups for which NULL is optimal in expectation. From one immutable snapshot, every action would be scored by replaying every continuation through the frozen reader under the corresponding post-action memory. No further write would be permitted within the horizon. These rules are preserved as a historical preregistration reference, not as an executed result or the current next experiment.

The template specified that the no-op, immediate/current-risk, local/current-risk, and local/expected-risk arms would share the same frozen checkpoint, slot count, registered actions, candidate payload, fixed gate $g=1$, write budget, train exposure, optimizer family, and evaluation budget. The local/current-risk to local/expected-risk contrast would change the supervision target only. All trainable arms would be parameter matched, all feasible actions would receive value targets, and a non-deployable empirical expected-risk oracle would be reported only as a ceiling. Payload learning, gate learning, recurrent rollout policy, index learning, and multi-write planning were outside that historical successor.

\textbf{Falsification before scale.} The direction stops if the empirical expected-risk oracle has negligible margin over current-risk writing, if the future-blind value model cannot recover a useful action ordering across fresh seeds, if lower future loss is bought by more stale leakage or collateral damage, or if gains disappear under parameter/compute matching. No checkpoint selection, failed-seed replacement, future-dependent deployment input, or post-held-out horizon change is allowed. A successful intervention authorizes a short joint-training study; it does not directly authorize a larger-scale or broader system claim.

\section{Frozen-Hidden-Bridge Protocol Details}
\label{app:hiddenbridge}

The six-family frozen-hidden bridge is a controlled representation study, not a learned or open-domain parser. Each generated sample supplies event sentences, a query sentence, integer \texttt{key\_ids}, integer \texttt{value\_ids}, a target event index, and optional write labels. The frozen backbone mean-pools last-layer hidden states over each complete event and query sentence. Separately, the protocol forms the canonical string \texttt{entity N} from each supplied key identifier, encodes it with the same backbone, and mean-pools that string. Event and query representations concatenate full-sentence vectors with the canonical-key vectors. No runtime regular expression, named-entity recognizer, exact-string span matcher, or value-span extractor is used; same-key replacement and branch arbitration use exact integer equality. The study therefore asks whether frozen hidden representations support a lifecycle when key identity is supplied. It does not establish entity discovery, alias resolution, or coreference, which the learned QA selector addresses only at passage level.

\section{Mathematical Properties and Stability Boundaries}
\label{app:stability}

\textbf{Causal boundary.} At boundary $b_n$, $z_n$ is a deterministic or stochastic function of $(X_{\le b_n},M_{n-1})$. The updated state $M_n$ is used only for positions greater than $b_n$. Therefore the conditional distribution for any already emitted token is unchanged by consolidation. Bidirectionality within $(b_{n-1},b_n]$ does not introduce a path from $X_{>b_n}$ to $p(X_t\mid X_{<t})$ for $t\le b_n$. The implementation enforces this property structurally rather than relying on a loss penalty.

\textbf{Atomicity.} Let $\Delta_{n,j}^{v}$ be the difference between matrices formed by stacking the neural payload views of $M_{n,j}$ and $M_{n,j-1}$. From Eq.~\ref{eq:atomic-overwrite}, $\operatorname{rank}(\Delta_{n,j}^{v})\le1$ and $\lVert\Delta_{n,j}^{v}\rVert_{0,\mathrm{row}}\le1$ for every event. Across the full segment, $\lVert V(M_n)-V(M_{n-1})\rVert_{0,\mathrm{row}}\le K_n$. The authoritative-state definition is stronger: every mutable canonical, neural, routing, version, security, tombstone, and rollback field of the selected row commits together. Any single event trace that modifies two resident rows, or silently mutates an external authoritative alias/index state, is outside the contract. Protection can only remove actions from $\mathcal A_{n,j}$; it cannot add a hidden second write.

\textbf{Resident boundedness.} Equation~\ref{eq:resident-bit-bound} is a bit-level contract, not merely a fixed row count. Every resident field has a fixed maximum width and there are exactly $S$ cells, so $\operatorname{bits}(M_n)\le B_{\mathrm{resident}}$ for every $n$. Monotone versions use bounded integers with an explicit fail-closed exhaustion policy: the last representable version cannot be incremented, and counters never wrap, saturate, or reuse identity/version pairs. Canonical content, aliases, and metadata that exceed their caps likewise make the candidate infeasible. Unbounded provenance, rollback history, archives, indexes, and audit events are external objects referenced by fixed-width digests and charged to $A_n$. Thus resident state is truly bounded while total retained system information may grow.

\textbf{Dual-view consistency.} A row is valid only when Eq.~\ref{eq:dual-view-byte-equality} holds and Eq.~\ref{eq:dual-view-receipt} binds one identity and version to both canonical and neural bytes under one frozen encoder version. Hashes are receipts rather than the equality predicate. The receipt is committed atomically with the row, and byte equality plus the receipt are rechecked at load, recovery, export, teacher replay, and index rebuild. A mismatch cannot be repaired by selecting one view: the row is quarantined, removed from the feasible read/write set, and requires a new validated version. This provides mechanical byte consistency and a fail-closed semantic qualification boundary; it does not prove that the learned neural view is semantically adequate.

\textbf{Multi-event teacher semantics.} Equations~\ref{eq:multi-event-prefix-state}--\ref{eq:multi-event-teacher-value} define each event value on its own branch prefix and include later event decisions in the cost-to-go. Exact dynamic programming, a frozen continuation policy, and the registered $K_n=1$ intervention are distinct protocols and must be labeled as such. The policy minimization occurs outside the future expectation; a per-future argmin would be a clairvoyant teacher unavailable to deployment. Reusing the pre-segment candidates at later events, averaging event-local argmins, or scoring each event while silently holding later writes undefined is not a valid teacher. Grouped future weights are fixed before outcomes are opened, and group identity is the data-splitting unit.

\textbf{Complete-candidate accounting.} Equations~\ref{eq:context-compute} and~\ref{eq:teacher-compute} charge semantic proposal, trusted assembly, dual-view validation, action scoring, commit, branch-prefix expansion, and future replay. Any target shortlist or shared constructor must publish its own construction, miss, and excluded-action costs. Consequently, the simpler $O(TSd)$ read expression is not an end-to-end MMLA cost bound.

\textbf{Entropy-regularized optimum for a fixed expected-risk vector.} Expanding Eq.~\ref{eq:variational-free-energy} gives
\[
\mathcal F[q]+\tau\log Z
=\tau\sum_a q(a)\log\frac{q(a)}{q^*(a)}\ge0.
\]
Equality holds exactly at $q=q^*$. The Gibbs distribution is therefore the unique regularized minimizer when $E(a)=\overline J_H(a\mid s)$ is fixed, every registered action has finite risk, and $\tau>0$. The identity neither estimates $\overline J_H$ nor permits the expectation to move through softmax. The future expected-risk successor estimates the conditional values first and uses direct value regression.

\textbf{Greedy action-value regret.} Let $a^*=\arg\min_a\overline J_H(a\mid s)$ and $\hat a=\arg\min_a\widehat Q(a\mid s)$. If $\max_a|\widehat Q(a\mid s)-\overline J_H(a\mid s)|\le\epsilon$, then
\[
\overline J_H(\hat a\mid s)-\overline J_H(a^*\mid s)
\le 2\epsilon.
\]
Indeed, $\overline J_H(\hat a)\le\widehat Q(\hat a)+\epsilon\le\widehat Q(a^*)+\epsilon\le\overline J_H(a^*)+2\epsilon$. This bound motivates held-out value error and action regret. It does not guarantee that a finite grouped benchmark estimates the true deployment distribution; that requires representativeness and fresh groups.

\textbf{Long-horizon boundary.} Authoritative rows do not obey a soft global-convergence claim: a newer fact causes a discrete hard commit, and routing, thresholded writes, and eviction form a switched process. A persistent false-write rate can therefore bias or oscillate even though each row and update is bounded. The predictive teacher is intended to reduce this error; only a long-horizon overwrite benchmark can establish repeated-write stability.

\input{studies/version_and_corrections}

% The detailed indexes are intentionally placed after the technical content so
% that the front matter remains a short orientation path.  Explicit bookmarks
% preserve direct navigation in PDF readers without adding entries to the
% high-level printed contents page.
\clearpage
\phantomsection
\pdfbookmark[1]{List of Figures}{list-of-figures}
\listoffigures

\clearpage
\phantomsection
\pdfbookmark[1]{List of Tables}{list-of-tables}
\listoftables

\end{document}

%% file: theory_family/integrated_preamble.tex
\newif\ifRtfIncludeRTT
\newif\ifRtfIncludeAMR
\newif\ifRtfIncludePMA
\newif\ifRtfIncludeDualState
\newif\ifRtfIncludeCSBC
\RtfIncludeRTTtrue
\RtfIncludeAMRtrue
\RtfIncludePMAtrue
\RtfIncludeDualStatetrue
\RtfIncludeCSBCtrue

\ifRtfIncludeRTT
  \input{theory_family/modules/rtt/preamble}
\fi
\ifRtfIncludeAMR
  \input{theory_family/modules/amr/preamble}
\fi
\ifRtfIncludePMA
  \input{theory_family/modules/pma/preamble}
\fi
\ifRtfIncludeDualState
  \input{theory_family/modules/dual_state/preamble}
\fi
\ifRtfIncludeCSBC
  \input{theory_family/modules/csbc/preamble}
\fi

\newcommand{\RtfBeginModule}[1]{%
  \begingroup
  \edef\RtfModulePrefix{tf:#1:}%
  \let\RtfBaseLabel\label
  \let\RtfBaseRef\ref
  \let\RtfBasePageRef\pageref
  \renewcommand{\label}[1]{\RtfBaseLabel{\RtfModulePrefix##1}}%
  \renewcommand{\ref}[1]{\RtfBaseRef{\RtfModulePrefix##1}}%
  \renewcommand{\pageref}[1]{\RtfBasePageRef{\RtfModulePrefix##1}}%
}

%% file: theory_family/integrated_overview.tex
\section*{Integrated Report Overview}
\addcontentsline{toc}{section}{Integrated Report Overview}

This report develops one Memory-Mediated Learning Architecture organized around
Predictive Dual-State Adaptation (PDSA). It separates a bounded numerical policy
carrier from a bounded authoritative-memory carrier, states their causal and
operational contracts, and distinguishes validated component evidence from the
unresolved end-to-end system claim. The integrated formal parts cover
Reasoning-Time Training, Atomic Memory Rows, Predictive Memory Admission,
dual-state composition, and completed-segment retrospective consolidation.

\begin{table}[H]
\centering
\small
\setlength{\tabcolsep}{4pt}
\begin{tabularx}{\linewidth}{p{0.27\linewidth}Yp{0.22\linewidth}}
\toprule
Layer & Role in the integrated architecture & Current scientific status \\
\midrule
Component evidence & Controlled lifecycle execution, bounded selection with fallback, and typed relational binding & Validated within the reported controlled settings \\
Post-trained latent readout & Candidate following, unconstrained generation, rebinding, and restoration under separately specified model/data lineages & Restricted success; generalized continuous-event qualification not met by any of nine latest trajectories \\
PDSA identity & Policy and authoritative-memory carriers with separate authority, lifetime, reset, rollback, and cost semantics & Formalized; end-to-end benefit unvalidated \\
Reasoning-Time Training & Within-problem policy-update witness and causal identification requirements & Policy-only intervention unexecuted \\
Atomic Memory Rows & Complete typed reads and writes, trusted assembly, exact NULL, version lineage, dual-view consistency, and recovery obligations & Formal contract formalized; conditional results proved; implementation and semantic qualification incomplete \\
Predictive Memory Admission & Grouped futures, expected-risk oracle, uncertainty, and future-blind action selection & No measured oracle margin or learned admission result \\
Completed-segment consolidation & Post-closure inspection, ordered event recursion, authoritative publication, and later exposure & Formal contract; end-to-end experiment unexecuted \\
Dual-state composition & Independently controlled policy and memory factors with an interaction estimand & Factorial unexecuted \\
\bottomrule
\end{tabularx}
\caption{Scientific scope and evidence status of the integrated report. Formal
results state conditional consequences; empirical status is determined by the
corresponding experiments and qualification gates.}
\label{tab:integrated-report-overview}
\end{table}

%% file: studies/posttraining_evidence.tex
\section{Post-Training Evidence: Binding, Generation, and Generalization}
\label{sec:posttraining-evidence}

The studies in this section use pretrained MiniCPM5 models and small learned
interfaces. They are a separate model lineage from the compact compiler and
frozen-checkpoint semantic-interface studies above. Their restricted successes
do not alter the latter studies' negative outcomes, and the latter failures
do not imply that every later interface is unlearnable. This section reports
completed work through September 13, 2026. The subsequent
coverage-controlled study is detailed in Section~\ref{sec:coverage-readout}.

\subsection{Models, carrier identity, and measurement units}
\label{sec:posttraining-units}

The locked models are \texttt{openbmb/MiniCPM5-1B-Base}, revision
\nolinkurl{74f1b424630b2ede03e19b3e317ed77c4f014524};
\texttt{openbmb/MiniCPM5-1B-SFT}, revision
\nolinkurl{a60b37f1fc409c54e1e337b0723aaac6f92dfec0}; and
\texttt{openbmb/MiniCPM5-1B}, revision
\nolinkurl{87179e5c1f455ef22e6223592d2d61351b525bfc}.
The last is a distinct checkpoint, not an alias for either of the first two.
Its upstream training history is not identified causally by its repository
name. Later generation/readout studies fix the SFT model rather than treating
each new intervention as another three-model comparison.

\paragraph{Training an interface is not within-problem policy learning.}
The early post-training interface freezes the backbone and contextual row
encoder, pools valid token states by a masked mean, and uses rank-64
input/value residual adapters with fixed gate 0.5. Oracle training updates
four input/value matrices and freezes query matrices. Top-two training has a
different active query boundary. These offline adapter updates are not a
demonstration that $\Phi$ changes and is reused while a single evaluation
problem remains unresolved. Evaluation freezes the learned interface while
external rule-based writes change $M$. This is a component readout study, not
an empirical identification of both PDSA carriers.

\paragraph{Binding response and candidate following.}
Let $L(A,a)$ be mean answer-token NLL under memory state $A$ and candidate
answer $a$. States $A$ and $B$ keep the same question, key set, and value set
but exchange bindings so that the correct answer changes from $a$ to $b$.
The response contrast is
\begin{equation}
d=\tfrac12\{L(A,b)-L(A,a)+L(B,a)-L(B,b)\}.
\label{eq:posttraining-d}
\end{equation}
The strict candidate-following score $F$ requires the correct answer to have
strictly lower teacher-forced NLL than both other candidates in \emph{each}
state; ties fail. Positive $d$ need not imply $F=1$ on a pair, and neither
quantity is free-generation accuracy.

\paragraph{Generated updating and restoration.}
The later score
\begin{equation}
G=\frac1N\sum_i c_{i,A}c_{i,B}c_{i,A2}
\label{eq:posttraining-G}
\end{equation}
requires the first generated word to match exactly at the original state,
after changing two bindings, and after restoring the original state. Each
version is actually persisted and read. Generation is greedy over the full
vocabulary, with at most 16 new tokens and a fixed first-word parser. It is
not restricted to candidate answers and receives no gold answer suffix or
forced correct prefix. Two additional queries check an unchanged key.
$A2$ is a restoration check on the same group, not an independent sample.
Later continuous-event experiments instead require all eight prescribed
queries in a 32-event episode to succeed, yielding $G_{\rm loop}$.

\paragraph{Different meanings of ``new.''}
Training fit, a new binding on a trained key, a new key using trained words,
and a new key using task-untrained words are distinct strata. A word excluded
from adapter training may still occur in unknown pretraining or SFT corpora.
Familiar/unfamiliar conditions sometimes deliberately share physical groups
and keys to form a paired intervention. Repeated development on related
generators can induce distribution-level adaptation even when exact keys
do not overlap. We neither estimate a numerical contamination probability
nor equate exact-key isolation with universal absence of leakage.

\subsection{Short-budget initialization studies and finite-budget validation}

The short-budget matched-oracle study, completed in September 2026, trains
on 1,024 paired examples with a 64-update small-set phase followed by only
64 diverse updates. A paired top-two versus oracle training comparison is
evaluated through the same oracle reader. The oracle-trained mean $F$ is
19.01\% for Base, 11.46\% for SFT, and 10.16\% for MiniCPM5-1B. Each uses
three subjects and 256 evaluation pairs; all fail the registered 50\%
following requirement. The corresponding mean NLL-response uplifts are
0.408658, 0.238458, and 0.201687 nats per target token. Their recorded
within-model sign-flip $p$ values are all
$1.5258556\times10^{-5}$, but response significance does not replace the
failed reliability criterion. The comparison does not establish a general
cross-model ranking: the two additional checkpoints deliberately repeat the
original protocol and input seeds, rather than supplying independent new
test distributions.

Earlier fixed-answer fitting and rebinding diagnostics in that lineage
showed low correct and mismatched NLL together with persistent old-answer
preference. They are consistent with a weak binding interface or a
task-answer shortcut; they do not uniquely establish parameter memorization
or corpus leakage. Evaluation-only oracle selection did not repair the
short-budget interface. Training and evaluation route matching helped, but
still did not qualify. Limited coadaptation branches that were stopped before
execution remain \emph{not run}, not additional failed model experiments.

The next finite-budget study holds SFT, oracle selection, pooling, rank,
gate, prompt, and optimizer fixed. It compares the original batch-level
correct/mismatched objective with a per-state three-candidate ranking
objective. Each of six trajectories has predetermined snapshots after one
and eight diverse passes over the same 1,024 pairs: total updates 128 and
576, including the common 64-update small-set phase. These are paired
points on one optimizer trajectory, not independent runs or eight times
as many distinct data.

\begin{table}[tbp]
\centering\small
\caption{Finite-budget candidate following on new keys, September 7, 2026.
All entries are $F$, not generated $G$. Each count is out of 256. Rows in
the first block use seeds 20261011--20261013; the independent-data second
block uses 20261111--20261113. ``Original'' is the batch-level objective;
``per-state'' changes both ranking granularity and negative coverage.}
\label{tab:posttraining-budget}
\begin{tabularx}{\linewidth}{@{}Ylll@{}}
\toprule
Model / objective / diverse passes & All subject counts & Mean $F$ & Sample SD \\
\midrule
SFT / original / one & 24, 15, 22 & 7.94\% & 1.85 pp \\
SFT / original / eight & 223, 233, 246 & 91.41\% & 4.50 pp \\
SFT / per-state / one & 16, 14, 15 & 5.86\% & 0.39 pp \\
SFT / per-state / eight & 239, 246, 241 & 94.53\% & 1.41 pp \\
\midrule
Base / original / one & 27, 30, 32 & 11.59\% & 0.98 pp \\
Base / original / eight & 231, 227, 225 & 88.93\% & 1.19 pp \\
SFT / original / one & 25, 31, 33 & 11.59\% & 1.63 pp \\
SFT / original / eight & 250, 248, 235 & 95.44\% & 3.18 pp \\
\bottomrule
\end{tabularx}
\end{table}

For SFT, the original-objective budget gain is 83.4635 percentage points
(pointwise 99\% interval [79.2969, 87.3698]; Holm-adjusted
$p=0.0000305171$). The per-state objective adds 3.1250 points at the high
budget (interval [0.2604, 5.9896]; adjusted $p=0.00387567$), below the
registered 10-point materiality criterion. Thus the new ranking objective
is not necessary for the observed high $F$, but its effect is not proved
to be exactly zero. A fresh paired Base/SFT budget study with new data and
new adapters obtains gains of 77.3438 and 83.8542 points respectively;
both adjusted $p$ values are 0.0000305171. Every high-budget subject passes
the candidate-following requirement. These are direct counterexamples to
the claim that this Base checkpoint cannot learn any usable latent binding
under this interface. They do not establish free generation or unlimited
benefit from further training.

\subsection{Candidate success does not imply free generation}

On a frozen-checkpoint generation diagnostic using the six high-budget
subjects, the three Base $G$ counts are 25, 17, and 16 out of 64; the SFT
counts are 43, 28, and 21. Mean $G$ is 30.21\% and 47.92\%, although
same-input candidate $F$ is 91.15\% and 90.10\%. The frozen text controls
are also weak: 27/64 for Base and 36/64 for SFT. Nine conditional training
ablations and the final evaluation were therefore not run. Correct-row
requests did not hit the length cap, so longer maximum generation is not
a supported explanation for these failures.

A first-token margin intervention subsequently trains six new Base
trajectories on matched two-row swaps, retaining the original full-vocabulary
CE. Original-objective $G$ counts are 52, 54, and 126 out of 256; with the
fixed margin they are 83, 106, and 79 (seeds 20261311--20261313).
The mean gain is 4.6875 points, recorded $p=0.0145414$, with pointwise
99\% interval [$-0.6510$, 10.0260] points. It fails the prespecified
10-point and $p<0.01$ rule and decreases the third paired subject.
Low training loss, high candidate following, first-token competition,
and exact generated words are different observations; none can silently
replace another's endpoint.

The zero-training prompt calibration that followed compared plain and native
SFT prompt configurations using correct single-row text. The initial attempt
was incomplete because of a tokenizer-return representation incompatibility;
only 320 calibration requests had completed. The corrected diagnostic
completed 26,240 requests with zero optimizer updates. Plain-text calibration
won 42/64 while the registered native configuration won 0/64, selecting plain
before development evaluation. The single frozen SFT text reference then
achieved only 138/256. This is a negative result for those prompt
configurations, not evidence that native chat templates are generally poor.
It motivates strengthening the text reference before attributing the
latent/text gap entirely to representation.

\subsection{A strong text reference and restricted prefix readout}

In the text-generation adaptation study (September 8), three SFT subjects
train only the last decoder block and final normalization, using ordinary
answer-token CE and the same plain prompt. Each completes 576 updates.
All three obtain 256/256 new-key $G$, with complete fit and rebinding
diagnostics; the same-batch unadapted reference obtains 143/256. The
recorded paired 95\% improvement interval is [37.8906, 50.3906] points.
These separately adapted text subjects become a positive reference, not
an equal-parameter or equal-training-history competitor for a latent student.

Adding last-block/norm coadaptation to the old pooled/global-additive latent
interface does not fix it under a matched six-trajectory, 576-update study.
The original pooled $G$ counts are 139, 145, and 136; coadapted counts
are 133, 126, and 132 out of 256. The latter-minus-former mean is
$-3.7760$ points, with recorded 95\% interval [$-8.5938$, 1.0417].
Training fit and rebinding are also unreliable. The encoder remains the
original frozen encoder; decoder coadaptation must not quietly change what
the row representation means.

The next interface uses the selected row's tokenwise contextual states to
form a continuous prefix anchored by a fixed nonzero BOS vector. The first
unanchored attempt is retained as numerically invalid: one pooled trajectory
completed 576 updates, but the first prefix backward was nonfinite and no
main evaluation ran. The fresh anchored study has all six trajectories
complete. With seeds 20261721--20261723, pooled $G$ is 144, 92, and
165 out of 256, whereas anchored-prefix $G$ is 256/256 for every subject.
The prefix intervention changes representation, position, and initialization
together; its improvement cannot be attributed uniquely to removing pooling.

An independent-input confirmation freezes these subjects on new keys and
bindings: the prefix remains 256/256 for all three, while pooled readout
obtains 150, 95, and 147. It introduces no new training seeds and no new
answer-word distribution. A matched prefix ablation then trains three
tokenwise prefixes and three mean-broadcast prefixes at the same physical
width and positions. Tokenwise $G$ is 256, 256, 256; mean-broadcast is
249, 251, 254 out of 256 (seeds 20261811--20261813). The 1.8229-point
gap, recorded interval [0.7813, 2.9948], is below the 10-point materiality
criterion, and both qualify. This does not establish equivalence, but it
does show that tokenwise residual preservation has not been established as
necessary for that familiar 40-word task. Mean broadcast is not a one-token
prefix and need not erase all order information from contextual states.

\subsection{Word transfer, fitting budget, and the fixed content path}

Frozen evaluation on new answer words exposes a sharp boundary. The
tokenwise prefix scores 256/256 on familiar words for all three subjects
but 0/256 on task-untrained words. Mean-broadcast scores 253, 253, and
254 on familiar words and also zero on untrained words. Separately
adapted text scores 244, 245, and 245 on those untrained words. The
paired construction shares keys/layout and changes word values; it is
neither a new training run nor evidence about unknown pretraining exposure.

Expanding training vocabulary from 40 to 256 at a fixed 576 updates still
produces zero new-word $G$ and weak fit for the expanded vocabulary.
Correct-target exposure per word falls as vocabulary coverage grows. A
subsequent 256-word fitting-budget study therefore retains fixed snapshots
after 576 and 4,160 updates on the same trajectory. It establishes excellent
fit, rebinding, and new-key/old-word performance without reliable new-word
transfer (Table~\ref{tab:posttraining-fit-budget}). Increasing exposure
solves a support-fitting problem here, not the entire generalization problem.

\begin{table}[tbp]
\centering\small
\caption{Within-vocabulary fitting-budget study, September 9. Seeds
20262111--20262113; each triplet lists all subjects. Short and long endpoints
are paired within the same optimizer trajectory. $G$ is strict three-state
generation; fitting observations are not held-out data.}
\label{tab:posttraining-fit-budget}
\begin{tabularx}{\linewidth}{@{}Ylll@{}}
\toprule
Stratum & Groups each & Short counts & Long counts \\
\midrule
Complete training fit & 1,024 & 538, 458, 269 & 1024, 1024, 1024 \\
Training keys, new bindings & 256 & 111, 77, 44 & 255, 256, 253 \\
New keys, trained words & 256 & 116, 84, 52 & 254, 253, 254 \\
New keys, untrained words & 256 & 0, 0, 0 & 0, 2, 0 \\
\bottomrule
\end{tabularx}
\end{table}

The matched content-path study changes one interface term while matching
data, updates, and active parameters. For normalized frozen contextual
token state $n=\operatorname{LN}(h)$ and learned value residual $V(n)$,
\begin{align}
\text{learned prefix}&=\text{BOS}+0.5V(n),\\
\text{content-preserving prefix}&=\text{BOS}+0.5n+0.5V(n).
\label{eq:posttraining-content-path}
\end{align}
The extra term is contextual content, not original token embeddings, a gold
suffix, or a text bypass. Both interfaces retain the four-matrix adapter
(393,216 active parameters). Six discarded engineering pilots precede six
formal trajectories at 4,160 updates each. At the fixed endpoint the learned
prefix has zero new-word successes; content-preserving counts are 213, 229,
and 234 out of 256 (seeds 20262211--20262213). The mean gain is
88.02 points, with the recorded descriptive 95\% group interval
[84.51, 91.28]. Both arms fit all 1,024 training groups and generalize to
new bindings using trained words. Thus this is a large matched improvement
after support fitting, not a subtraction from an old run.

The full reliability requirement still fails: only the third new-word
subject exceeds 231/256, and unchanged-key success is also below its
requirement for two subjects. Removing the learned value residual while
retaining the fixed content and trained input adapter yields zero whole-group
successes. This is a frozen-path dependence result, not proof that every
retrained architecture requires both paths. All predeclared failed subjects
remain in the mean and in the retained evidence.

\subsection{Independent inputs and restricted continuation success}

A zero-training independent-input confirmation gives the same fixed content
subjects 229, 248, and 234 new-word successes out of 256, while the
learned-prefix reference remains zero. The mean is 92.58\%, but the first
subject still fails 231/256. Learned-only and fixed-only ablations are both
zero for those frozen weights. This confirms that a substantial path effect
persists on new inputs; it does not convert the all-subject qualification
into a mean-only rule.

A new paired continuation study starts from those parent adapters, with
fresh optimizers and data, and compares original CE/ranking with later-answer
token weight two. Each of six branches receives 576 updates. Original-objective
new-word $G$ counts are 243, 251, and 242; token-weighted counts are
254, 248, and 240 (all /256; seeds 20262411--20262413). Both arms pass
all registered subject, unchanged-key, content-control, and restoration
requirements. The weighted-minus-original gain is only 0.78125 points,
with recorded 95\% interval [$-0.6510$, 2.2135], below the registered
two-point materiality requirement. The simpler original objective is thus a
qualified restricted baseline; this result is not a claim that token
weighting is universally useless.

This is genuine but limited read--update--generate--restore evidence with
four rows, oracle key selection, an external rule writer, fixed controlled
language and finite word sets. It does not train an autonomous router,
semantic writer, eviction policy, or predictive future-utility controller.
The next stress test changes the event and word distribution, so its lower
score is not a time-series decline of the same measured object.

\subsection{Continuous events: reliable text, incomplete latent closure}

With the continuation subjects frozen, a 32-event, eight-query protocol
scores whole episodes rather than three-state groups. Normal latent readout
achieves binding counts 219, 190, 212 and preference counts 230, 194, 214
out of 256. The text reference is perfect, all registered negative-control
whole-episode counts are zero, and restore checks pass. Normal-minus-no-write
has mean 81.97 points (95\% interval [79.43, 84.37]), yet the absolute
and unchanged-key requirements fail. A detectable update effect is not
reliable episode completion. An existing-output audit finds errors concentrated
in particular word forms; it does not run another model evaluation.

The event vocabulary/budget experiment compares 256 versus 1,024 training
words and paired 576 versus 4,160 update endpoints. On binding, the
four trained cells' mean $G_{\rm loop}$ is 22.79\%, 28.26\%, 44.01\%,
and 52.86\%; on preference it is 12.11\%, 17.06\%, 32.16\%, and
38.80\%, in that order. The prespecified original-vocabulary budget
effect is 5.2083 points, and the high-budget coverage effect is 23.1771
points. Their recorded Bonferroni 97.5\% intervals are [2.5391, 7.8776]
and [19.3359, 27.0833]. Both are material improvements, but all absolute
qualifications fail and the text reference itself is now insufficient.
This prevents a clean attribution of all remaining error to latent readout.

A text-strengthening and latent-coadaptation study fixes a COPY prompt using
text-only calibration, then trains nine trajectories: text-tail adaptation,
content-preserving adapters, and adapters plus decoder-tail coadaptation.
The first builds the strong text parents used in the later matched-layout
and teacher-alignment studies. The main matrix is
Table~\ref{tab:posttraining-event-reference}; conditions are evaluated on
the same physical episodes but have different decoder adaptation histories.

\begin{table}[tbp]
\centering\small
\caption{Text-strengthening and latent-coadaptation study, September 11.
Seeds 20263311--20263313; counts are whole eight-query episodes out of 256.
All three trained conditions receive 1,088 updates, but active parameters and
compute differ. No best-seed selection is used.}
\label{tab:posttraining-event-reference}
\begin{tabularx}{\linewidth}{@{}Ylll@{}}
\toprule
Condition & Binding counts & Preference counts & Mean by family \\
\midrule
Adapted text & 240, 239, 239 & 248, 252, 248 & 93.49\%, 97.40\% \\
Content-preserving latent & 127, 141, 136 & 186, 189, 176 & 52.60\%, 71.74\% \\
Latent plus tail coadaptation & 121, 117, 125 & 143, 159, 180 & 47.27\%, 62.76\% \\
\bottomrule
\end{tabularx}
\end{table}

The text condition qualifies; both latent conditions do not. The matched
coadaptation-minus-latent contrast is $-7.1615$ points (recorded 95\%
interval [$-10.1563$, $-4.1016$]). It is a negative result for that
coadaptation intervention, not a theorem that a decoder must always remain
frozen. Training fit, new bindings on trained keys, and new keys with
trained words are strong; new-word readout remains the salient limitation.

\subsection{Matched presentation and the incomplete bridge precursor}

The first native-embedding layout diagnostic completes its frozen comparison
but fails the prerequisite for bridge training. No bridge optimizer update
occurs. A fresh matched-presentation diagnostic then uses the \emph{adapted
text decoder}, not the original frozen student decoder. Native embeddings
with joint tokenization and ordering reproduce text behavior, and a
physically padded matched layout remains qualified. These are engineering
and input-form controls, not an implementation of latent memory.

For binding, text, joint embeddings, and joint padded embeddings all score
63, 63, and 62 out of 64; separately tokenized prefix embeddings score
52, 52, and 54. For preference, text/joint embeddings score 63, 61, 61,
joint padded embeddings 62, 62, 62, and separate-prefix embeddings
52, 52, 52. Zero-content is zero throughout. All required matched-layout
unchanged-key and restore checks pass. Identical text/joint-embedding
success vectors and first-logit parity establish equivalence in the tested
API path; text versus padded layout is \emph{not} bitwise identical.
Nor does a nonsignificant test establish universal equivalence.

The ensuing latent/bridge precursor remains \emph{incomplete}. Of six planned
formal trajectories, the first baseline completes 1,088 updates and the
first bridge branch completes only its 64-update small-set phase. The
bridge passes 16/16 changed-key checks but only 15/16 unchanged-key checks;
the same wrong word appears consistently in both unchanged states. No main
event matrix is run. The missed control word was not in the tiny direct
positive-answer set, but the baseline succeeded on that word, so uncovered
supervision is not a sufficient causal diagnosis. This is not a completed
negative bridge-versus-baseline experiment. The next study explicitly covers
both changed and unchanged queries in all arms and completes its full matrix
rather than changing the old precursor's status.

\subsection{Evidence limits and current implications}

All retrospective comparisons above preserve protocol changes, failures,
incomplete runs, and unexecuted branches. Recorded intervals are copied
from the original calculations; this report performs no resampling,
generation, training, or opening of a final holdout. Sample standard
deviations describe the three recorded subjects, not uncertainty over all
possible training seeds. Group-resampled intervals retain conditions and
subjects together and are conditional on the fixed word sets. Early
sign-flip tests further depend on their stated symmetry or exchangeability
assumptions; they are not automatically exact tests of any weak mean null.

The strongest defensible story is neither ``the base cannot learn'' nor
``the complete architecture is validated.'' Finite-budget candidate following,
familiar-word prefix generation, new-word content-path gains, and a restricted
continuation success are supported at their own scopes. Broader continuous
events remain unqualified. Native embeddings are a diagnostic input, not a
required PDSA component. Proposed canonical and dual-view memory contracts
remain prospective; they do not retroactively qualify a failed pure latent
interface or the complete predictive-admission system.

%% file: studies/coverage_controlled_readout.tex
\section{Coverage-Controlled Latent Readout with a Bridge and Text Teacher}
\label{sec:coverage-readout}

\subsection{Question, lineage, and completed scope}

The September 13, 2026 study asks whether a fixed contextual memory interface
becomes reliable on new task words when supervision covers both changed and
unchanged queries, and whether a larger mapping or frozen text-teacher outputs
add a useful improvement. It is a completed three-condition, three-seed study,
not an unexecuted proposal and not a continuation of an interrupted invocation.
All nine formal trajectories reached their fixed endpoints before development
evaluation. None of the three latent conditions met the registered qualification;
the strong text reference and the restoration checks did meet their requirements.

The backbone is MiniCPM5-1B-SFT, with the model revision fixed to
\texttt{a60b37f1fc409c54e1e337b0723aaac6f92dfec0}. The three optimization seeds
are 20263711, 20263712, and 20263713. They inherit the corresponding contextual
readout adapters from the September 11 text/readout co-adaptation study, but not
its optimizer state. Their text teachers inherit that study's separately
text-adapted decoder tails. Thus the student decoder and the strong text decoder
have different adaptation histories. A text--latent difference here is not a
pure intervention on representation space with an otherwise identical decoder.
All comparisons below concern these fixed lineages, task words, and generators.

The task retains four memory slots, externally specified rule writes, exact
key-based oracle row selection, completed events, and the fixed copying prompt.
There is no trained router, learned writer, future-utility admission controller,
within-problem policy optimizer, or from-scratch backbone in this experiment.
Offline adapter training does not constitute the feedback-bound policy-state
transition required by the RTT theory.

\subsection{Matched presentation and the three interventions}

The instruction, complete current key--value row, and question are jointly
tokenized. Offset mappings locate the memory span without splitting a token
across an instruction/memory/question boundary. The latent decoder replaces
that span and preserves the valid-token positions when padding the physical
memory width to 64. The student does not receive the original memory token
embeddings as a parallel content channel: their token IDs are replaced before
the decoder embedding lookup. The selected tokens instead enter an independent
frozen encoder. The same student forward path is used for training and generation.

Let $h$ be the tokenwise layer-normalized contextual states of that encoder,
$V$ the trained rank-64 value residual, and $e_{\rm anchor}$ the fixed anchor.
The memory-span input is $e_{\rm anchor}+r$, with
\begin{align}
 r_{\rm A}&=0.5h+0.5V(h),\\
 r_{\rm B}=r_{\rm C}&=0.5h+0.5V(h)+Wh.
 \label{eq:covered-readout-residual}
\end{align}
The full-width linear bridge $W$ has no bias and is initialized to zero. Input
and value adapters are active; the query adapter, original student decoder,
and independent encoder remain frozen. The input adapter also operates on the
non-memory part of the decoder input. Native embeddings occur only in a separate
frozen text-equivalence diagnostic, not as a proposed mandatory PDSA component.

\begin{table}[H]
\centering\small
\begin{tabularx}{\linewidth}{lYrp{0.30\linewidth}}
\toprule
Condition & Intervention & Active parameters & Supervision \\
\midrule
A & Matched contextual readout & 393,216 & Covered answer CE and changed-only hinge \\
B & A plus zero-initialized full-width bridge & 2,752,512 & Identical to A \\
C & Same bridge and capacity as B & 2,752,512 & B plus frozen text-teacher token KL \\
\bottomrule
\end{tabularx}
\caption{Coverage-controlled latent readout conditions. A, B, and C denote the
explicit interventions in this table, not different models or data splits.
Each condition has three paired optimization seeds.}
\label{tab:covered-conditions}
\end{table}

B minus A tests the addition of the bridge under the common covered objective.
C minus B tests the addition of this fixed output-alignment objective at the same
student capacity. All three conditions change supervision coverage relative to
the earlier interrupted study. There is no uncovered condition here, so the
causal effect of coverage itself is not identified by a cross-study subtraction.

\subsection{Supervised roles and the exact optimized objective}

Each training pair contains two states that exchange the bindings of a changed
key while preserving an independently queried control key. The changed query in
both states and the unchanged query in both states contribute positive answer
cross-entropy. An unchanged answer is identical across the two states and is
never treated as the other state's contradictory negative. Restoration of an
identical state is checked separately rather than counted as additional
independent answer supervision.

For one logical update, let $N_c$ and $N_u$ be the positive answer-token counts
for changed and unchanged roles. Let $S_c^+$, $S_u^+$, and $S_c^-$ be the sums of
answer-token NLL for changed-correct, unchanged-correct, and changed-mismatched
conditioning. The mismatched condition retains the target answer and exchanges
its memory conditioning; its token count remains $N_c$. Define
\begin{align}
 L_{\rm all}^+ &= (S_c^++S_u^+)/(N_c+N_u),\\
 g_c &= S_c^-/N_c-S_c^+/N_c,\\
 \mathcal L_{\rm covered} &=L_{\rm all}^+
             +0.1\max(0,0.05-g_c).
 \label{eq:covered-objective}
\end{align}
This is a logical-batch, token-weighted objective. Coverage means positive
supervision of all four query/state roles; it does not mean a new full-candidate
ranking loss, a per-example hinge, or equal weighting of unequal-length answers.

For C only, the frozen teacher supplies a full-vocabulary distribution at every
positive answer position, including the first position and later word-form
positions. With the same observed memory, question, and teacher-forced answer
prefix, the registered objective is
\begin{equation}
 \mathcal L_{\rm C}=\mathcal L_{\rm covered}
  +\frac{0.1\tau^2}{N_c+N_u}
   \sum_{j\in\text{positive answer positions}}
       D_{\rm KL}\!\left(p_T^\tau(\cdot\mid m,q,y_{<j})\,\Vert\,
                           p_C^\tau(\cdot\mid M,q,y_{<j})\right),
 \qquad \tau=1.
 \label{eq:covered-teacher-objective}
\end{equation}
Teacher logits are detached. Hard-label CE remains present, teacher errors do not
filter the training data, and neither teacher outputs nor answer suffixes enter
student generation. C uses no top-$k$ teacher truncation, candidate-only
distribution, native-embedding reconstruction target, or development-label
supervision. The teacher uses the actual current row, not a row manufactured
from an evaluator's gold field.

The implementation evaluates the two changed conditions without gradients to
determine the logical-batch hinge branch. It then uses three independent graphs
for changed-correct, unchanged-correct, and changed-mismatched sequences. For
$a=\mathbf1[g_c<0.05]$, their summed-CE coefficients are respectively
\[
 (N_c+N_u)^{-1}+0.1a/N_c,\qquad
 (N_c+N_u)^{-1},\qquad -0.1a/N_c.
\]
The derivative at the hinge kink is zero. Three backward calls accumulate these
contributions before one clipped AdamW update. Recomputed changed-condition
losses must match the no-gradient pass. Logged loss reconstructs the actual
CE plus hinge plus KL, not the piecewise-linear gradient surrogate with its
constant omitted. This distinction matters when interpreting telemetry.

\subsection{Data strata, schedule, and leakage boundary}

The formal training set contains 8,192 pairs and 1,024 new task words. The short
phase repeats the first 16 pairs for 64 updates. Two full main-phase passes add
1,024 updates, for a fixed endpoint of 1,088 updates per trajectory. A logical
update comprises 16 pairs, 64 positive sequences, and 32 changed-key mismatched
sequences. The three graphs use 32 sequences each; including the hinge prepass,
there are 160 student target-forward sequence presentations per update.

All nine short phases finish before any long phase, and each trajectory retains
its own live AdamW moments, step counter, and RNG across those phases. Formal
training follows a separate discarded engineering exercise of three 64-update
trajectories. Neither engineering weights nor a best development checkpoint
initialize a formal trajectory. All formal endpoints are fixed and reported.

AdamW uses learning rate $2\times10^{-4}$ for both adapters and bridge,
$(\beta_1,\beta_2)=(0.9,0.999)$, $\epsilon=10^{-8}$, weight decay $0.01$, and
gradient-norm clipping at 1, without a learning-rate schedule. Active parameters
are FP32, neural forward computation uses BF16 autocast, and CE/KL reductions use
FP32. Dropout and gradient checkpointing are disabled. Answer targets preserve
the specified leading space and trailing line feed; training is not restricted
to the first token.

Separate fixed diagnostic sets contain 64 fitted pairs, 64 new bindings with
training keys, and 64 new keys with training words. These are three different
forms of support checking, not a pooled unseen-word evaluation. The layout
diagnostic and the main development set each contain 64 event groups per family,
with 80 words per family/domain. Those four word sets and the engineering domain
are separate from formal training. The allocation uses the registered two- and
three-token answer strata without selecting words by model performance.

Historical isolation covers 55 explicitly bound manifests and the identifiable
task-training vocabulary of the text teachers. This does not establish zero
overlap with unknown pretraining or SFT corpora, unbound historical manifests,
near duplicates, or all semantically related word forms. Repeated development
on a common word source and generator also creates distribution-adaptation
risk. New keys or directories alone do not establish independence. The old final
holdout is not used, and this development result is not an independent final test.

\subsection{Strict continuous-event endpoint and fixed decision rule}

An evaluation group contains four slots and 32 completed events. The eight
predeclared queries occur at event/slot positions
$(4,0),(8,0),(12,2),(16,1),(16,1\text{ after reload}),(20,3),(26,0),(32,1)$.
The repeated event-16 query checks restoration of the same state. Generation is
greedy over the full model vocabulary, limited to 16 new tokens, and scored by
the original first-word parser. It is not candidate-constrained decoding.
For group $i$, let $c_{i,j}$ indicate exact first-word correctness on query $j$.
The primary endpoint is
\begin{equation}
 G_{\rm loop}=\frac{1}{N}\sum_{i=1}^N\prod_{j=1}^{8}c_{i,j}.
 \label{eq:covered-loop-score}
\end{equation}
The queries are dependent. This empirical conjunction is neither the product of
eight marginal accuracies nor the earlier teacher-forced candidate-following
score. The two unchanged-key queries and actual same-state restoration are also
reported separately.

Qualification requires each of three subjects in each of two families to reach
at least $58/64$ on both $G_{\rm loop}$ and unchanged-key correctness. Each
registered negative control must have at most $6/64$ strict successes, and the
positive-minus-negative difference must be at least 0.5. All restoration
disagreements must be zero, and all three text references must satisfy their own
reliability and control requirements. No cross-seed average substitutes for
these per-subject conditions. Short-phase learning accuracy is diagnostic in
this protocol; integrity, nonfinite gradients, or failed restoration still
invalidate the relevant execution. This changes no earlier study's verdict.

\subsection{All subjects and the two primary comparisons}

\begin{table}[H]
\centering\small
\setlength{\tabcolsep}{3.8pt}
\begin{tabularx}{\linewidth}{llp{0.21\linewidth}rp{0.22\linewidth}}
\toprule
Family & Condition & Strict successes /64 & $G_{\rm loop}$ mean $\pm$ SD (\%) & Unchanged successes /64 \\
\midrule
Binding & A: contextual & 42, 43, 41 & $65.63\pm1.56$ & 59, 56, 56 \\
 & B: bridge & 31, 41, 47 & $61.98\pm12.63$ & 53, 54, 58 \\
 & C: bridge + teacher & 24, 41, 36 & $52.60\pm13.65$ & 50, 54, 54 \\
 & Strong text & 64, 64, 64 & $100.00\pm0.00$ & 64, 64, 64 \\
\midrule
Preference & A: contextual & 38, 45, 45 & $66.67\pm6.31$ & 54, 59, 57 \\
 & B: bridge & 42, 41, 42 & $65.10\pm0.90$ & 58, 57, 57 \\
 & C: bridge + teacher & 38, 43, 40 & $63.02\pm3.93$ & 55, 57, 58 \\
 & Strong text & 64, 64, 64 & $100.00\pm0.00$ & 64, 64, 64 \\
\bottomrule
\end{tabularx}
\caption{Completed coverage-controlled readout study, September 13, 2026.
Within each latent row, subjects are ordered 20263711/12/13; the text row contains
the three corresponding frozen text-adapted subjects 20263311/12/13, not three
copies of one model. SD is the sample standard deviation across subjects, in
percentage points, not a standard error. All denominators are 64 physical groups.}
\label{tab:covered-results}
\end{table}

All three latent conditions fail qualification. No latent subject reaches the
strict-success threshold in either family. Several unchanged-key counts also
fall below threshold. The no-write, wrong-binding, and content-zero strict
success counts are all zero; the text no-memory controls are also zero. Zero
negative scores alone do not imply that every content-dependence requirement
passes: binding B and C fail that requirement for at least one subject. All
registered restoration checks pass. This is evidence of nontrivial but
insufficient content-dependent readout, not an absence of all memory response.

The two registered primary contrasts average the two families equally and retain
the pairing of all eight queries and three fixed subjects within each episode.
Their stored intervals come from 65,536 within-family episode bootstrap draws
with seed 20263718. Each is a 97.5\% percentile interval with a Bonferroni
allocation for the two comparisons (familywise $\alpha=0.05$). They are conditional
on these words and subjects, not confidence intervals for every training seed or
word population. The report transcribes the existing intervals; it does not
re-bootstrap or select a new analysis.

\begin{table}[H]
\centering\small
\begin{tabularx}{\linewidth}{Yrrp{0.22\linewidth}}
\toprule
Registered contrast & Mean (pp) & Stored 97.5\% interval (pp) & Three paired subject differences (pp) \\
\midrule
Bridge minus contextual readout (B$-$A) & $-2.60$ & $[-9.64,4.17]$ & $-5.47,-4.69,2.34$ \\
Teacher alignment minus bridge (C$-$B) & $-5.73$ & $[-11.72,0.00]$ & $-8.59,1.56,-10.16$ \\
\bottomrule
\end{tabularx}
\caption{Primary paired comparisons of the completed coverage-controlled study.
The second upper endpoint is a stored floating-point value within numerical
roundoff of zero; the display does not imply a positive improvement. Material
improvement required at least 10 pp, a positive corrected lower bound, and
positive differences for all three paired subjects. Neither contrast qualifies.}
\label{tab:covered-contrasts}
\end{table}

The outcome does not support adding the bridge or this fixed KL supervision for
the registered improvement target. A negative mean is not proof that all
possible bridges or distillation objectives are harmful: each contrast has a
subject with a positive difference, and no alternative temperature, weight,
decoder, or optimization schedule was tested here. There is no qualified
condition to choose under the prespecified simplicity order A then B then C.

\subsection{Support diagnostics, teacher behavior, and measured cost}

All nine fixed fitted-pair diagnostics and all nine new-key/training-word
diagnostics score $64/64$. Eight of the nine training-key/new-binding diagnostics
score $64/64$; bridge-plus-teacher seed 20263711 scores $63/64$. These narrow
support results do not erase the new-word continuous-event failure. They also
do not reproduce the earlier small-model fitted-support failure: the models,
representations, supervision, and evaluation units differ.

For each of the three formal teachers, the cached training supervision contains
32,768 positive sequences, 81,920 target positions, and the full 130,560-entry
vocabulary. Logits are stored in their original BF16 dtype and converted to FP32
for KL computation. Teacher target-token NLL is respectively 0.01148, 0.01121,
and 0.01170; token entropy is 0.04094, 0.04023, and 0.04140; mean correct-token
rank is 1.00531, 1.00518, and 1.00548. These are teacher-forced training-domain
statistics, not free-generation accuracy or proof of teacher infallibility.
No training example is discarded because the teacher is wrong.

\begin{table}[H]
\centering\small
\begin{tabularx}{\linewidth}{YrY}
\toprule
Measured quantity & Value & Scope \\
\midrule
Formal optimizer updates & 9,792 & Nine trajectories, $64+1,024$ each \\
Discarded engineering updates & 192 & Three separate 64-update exercises \\
Total backward calls & 29,952 & Three per logical update, including engineering \\
Generation requests & 62,064 & Layout, support, engineering, and main evaluation \\
Main event blocks & 84 & 43,008 requests; all completed \\
Layout event blocks & 18 & 9,216 requests; separate from main evaluation \\
Formal teacher-cache bytes & 64,187,203,584 & Sum of three recorded caches; not total runtime storage \\
Formal teacher-cache time (s) & 196.565 & Sum of three cache-generation times \\
Worker elapsed time (s) & 8,346.204 & Includes initialization, CPU work, caching, training, and evaluation \\
\bottomrule
\end{tabularx}
\caption{Observed execution cost, not budget estimates. The worker used one
RTX 4090 with the declared expanded memory capacity. Worker wall time is not
pure GPU training time, GPU utilization, or total project time. B/C have more
parameters than A and C incurs teacher computation and I/O; equal update counts
do not make these equal-FLOPs or equal-wall-time comparisons.}
\label{tab:covered-cost}
\end{table}

\subsection{What is and is not explained}

The completed matrix resolves an earlier missing comparison: the matched-layout
bridge and teacher conditions have now been trained and evaluated in all fixed
subjects. It does not repair the earlier interrupted study retroactively. The
new matrix also distinguishes a strong text task reference from inadequate
latent readout on the same new-word continuous-event endpoint. Its support
diagnostics favor a remaining transfer/reliability limitation over a claim that
the student cannot fit any binding at all.

The data do not uniquely identify pooling, normalization, decoder geometry,
teacher sharpness, optimization, or representation information loss as the
cause. In particular, the teacher has a text-adapted decoder tail, whereas the
student retains the original frozen SFT decoder. High teacher confidence is a
measured diagnostic, not a causal demonstration that the teacher distribution
was either sufficient or inappropriate. Rule writes and oracle reads remove
important learned decisions from this test. Reliable autonomous admission,
within-problem policy improvement, independent carrier resets under learning,
global capacity behavior, and joint PDSA benefit remain separate unverified
claims. A canonical or dual-view interface remains a proposed future
intervention, not a way to rename these latent conditions as successful.

%% file: studies/nearest_neighbors.tex
\subsection{Closest memory-learning systems: mechanisms rather than names}
\label{sec:nearest-neighbors}

The comparison below uses version-specific original papers available by the
evidence cutoff. It is not a benchmark ranking: their models, tasks, training
data, evaluators, costs, and success units differ from ours. In particular,
memory-operation accuracy, downstream question answering, game return, and
discovery of one best solution cannot be numerically compared with our
three-state or eight-query exact-generation criteria. The public bibliographic
records identify the consulted versions. A mechanism described as a design is
not silently upgraded to a demonstrated property of every implementation.
``Not established'' below means that the consulted evidence does not establish
the particular claim, not that the other system lacks the capability.
This comparison concerns mechanisms rather than matched benchmark performance.
The end-to-end benefits of the complete PDSA contract remain to be tested.

\paragraph{Dual adaptation has direct precedents.}
MCTR, \emph{Adapting Like Humans: A Metacognitive Agent with Test-time
Reasoning} (version~1), maintains natural-language task knowledge and a
trajectory buffer while adapting an action policy at test time
\cite{li2025adaptinglikehumansmetacognitive}. Its meta-reasoning module issues
\texttt{add}, \texttt{delete}, or \texttt{keep}; a scheduler regulates when the
knowledge is updated. Its action module conditions on that knowledge and
periodically applies a GRPO-style update using majority-vote agreement over
actions for past states. The original evaluation concerns seen and unseen Atari
games, with component ablations and adaptation dynamics. Thus neither an
explicit memory plus a changing policy, nor a test-time interaction between
them, is new in isolation. Our distinction is a more restrictive typed
contract for separately bounded numerical policy and authoritative memory,
with separate intervention and restoration obligations.

TTT-Discover, \emph{Learning to Discover at Test Time} (version~2), is an
especially important within-problem precedent
\cite{yuksekgonul2026learningdiscovertesttime}. Its Algorithm~1 retains a buffer
of attempts, evaluates a candidate, updates the policy, and reuses the result
while solving the same discovery problem. It combines an entropic objective
favoring high-reward outcomes with PUCT-inspired state reuse. The objective is
to discover one best solution, not to learn a policy with uniformly reliable
performance after deployment. Its reported mathematics and code-optimization
experiments therefore preclude a broad claim that online numerical adaptation
within a problem was introduced here. They also do not establish our specific
bounded-authority, future-blind admission, or independent policy/memory reset
requirements.

TTRL (version~3) samples multiple responses to an unlabeled test prompt,
estimates an answer by majority vote, and assigns agreement rewards for RL
\cite{tf:dual:zuo2025ttrl}. This is a concrete test-time
parameter-learning method, not merely longer inference. However, a benchmark
adaptation protocol and a strict single unresolved attempt need not share the
same initialization, lifetime, or information boundary. Our RTT classification
must be applied to a recorded execution schedule, not inferred from the phrase
``test time.'' A majority-derived reward can also be confidently wrong; it is
not an automatically correct predictive-memory teacher.

\paragraph{Downstream utility and memory credit are already learned.}
UMA, \emph{Learning to Remember: End-to-End Training of Memory Agents for
Long-Context Reasoning} (version~2), constructs query-agnostic memory by
processing context chunks and branches each final state into multiple QA
sessions \cite{zhang2026learningrememberendtoendtraining}. Its state contains
a core summary, a structured key--value bank, current focus, and an interaction
history. The tool interface supports writes, exact retrieval, listing, and
raw-context retrieval. Task-Stratified GRPO assigns each memory session its
own tool reward plus the mean downstream QA outcome for that memory
trajectory; memory and question-specific QA advantages are normalized in
different groups. Ledger-QA tests accumulated state tracking, in addition to
other retrieval and test-time-learning benchmarks. Accordingly, grouped
future usefulness and upstream credit from later QA cannot be claimed as
generic inventions of predictive admission here.

There is nevertheless a precise distinction. Our proposed predictive-admission
target compares admissible event actions, including NULL, on the same causal
prefix, with later decisions represented by a specified continuation policy
and with the policy minimization outside the future expectation. UMA's
trajectory-level mean QA signal is not, merely by having several QA branches,
the same event-local counterfactual estimand. Conversely, UMA reports an
end-to-end learned memory agent; our predictive-admission oracle margin and
its deployed approximation remain unvalidated.

Mem-T (version~2) explicitly addresses dense credit assignment
\cite{yue2026memtdensifyingrewardslonghorizon}. Its formation and evolution
policies build working, factual, experiential, and raw memories; evolution
includes ADD, UPDATE, DELETE, and IGNORE. Multi-turn retrieval is trained using
memory-operation trees, node-wise reward backpropagation, and intra-/inter-tree
advantages. Crucially, its construction policy uses hindsight credit from
terminal retrieval paths and ground-truth evidence where available, selects
the top half of valid actions within each operation category, and learns from
these selected demonstrations. It is inaccurate to describe the whole method
as a single undifferentiated terminal-reward GRPO update. Its training-only
hindsight supervision is legitimate when separated from execution inputs;
the corresponding separation in our design must be audited rather than
asserted by analogy.

Memory-R1 (version~5) trains a manager to choose ADD, UPDATE, DELETE, or NOOP,
and trains an answer agent to select useful retrieved memories
\cite{yan2026memoryr1enhancinglargelanguage}. The manager's reward comes from
the downstream answer after the proposed memory operation, using a frozen
answer agent during that update. The paper compares PPO/GRPO variants and
reports memory-management and answer-agent ablations. This is direct prior
work for outcome-trained CRUD and abstention. Its NOOP does not establish our
exact byte-accounted NULL transition, transaction lifetime, or adversarial
restore contract.

Mem-$\alpha$ (version~1) trains memory construction while retrieval and
generation remain fixed \cite{wang2025memalphalearningmemoryconstruction}.
It combines final-memory QA correctness and compression with per-action tool
validity and content-quality rewards, using core, semantic, and episodic
memory. Correctness and compression rewards are shared across the sequence;
tool/content components vary by action. Thus it already makes subsequent
answer quality a training signal for what is stored. The distinction here is
not ``future reward versus no future reward,'' but whether a specified causal,
event-level action comparison and complete cost ledger have been established.

AgeMem (version~3) jointly treats long-term storage and short-term context
control as tool actions \cite{yu2026agenticmemorylearningunified}. Its three
stages expose information, reset the short-term context while retaining
long-term memory, introduce distractors, and then require a memory-dependent
answer. The terminal group-normalized advantage is broadcast to preceding
actions in its step-wise GRPO formulation. This reset provides a concrete
information-isolation intervention. It differs from independently rolling back a numerical policy carrier
and an authoritative memory at specified points within one attempt. Those
stronger operations still require evidence in our own system.

\paragraph{Admission scores and non-parametric value learning.}
MemRL (version~2) keeps the LLM frozen and stores intent--experience--utility
triplets \cite{zhang2026memrlselfevolvingagentsruntime}. It first recalls a
similarity-filtered pool, then combines normalized similarity and learned
utility to select memories. For injected items its practical update moves the
stored utility toward observed reward; new summarized experiences are added
to memory. An empty candidate pool falls back to the frozen LLM. These are
runtime learning operations even without a model-weight update. Whether a
numerical utility belongs to $\Phi$ or $M$ in our terminology depends on its
actual storage, function, budget, and independent interventions; renaming a
memory field does not create a new policy channel. The paper's stability
analysis assumes conditions such as a frozen inference policy and stationarity;
it should not be quoted as an unconditional guarantee under arbitrary drift.

A-MAC (version~1) learns an interpretable admission score from utility,
confidence, novelty, recency, and type prior
\cite{zhang2026adaptivememoryadmissioncontrol}. A threshold admits or rejects
a candidate, with conflict handling for replacements/merges. Its utility
feature uses an LLM assessment of likely future usefulness; the score weights
and threshold are selected against labeled admission decisions. This is
predictive admission in an ordinary sense. It is not
the same measured quantity as a replay-based, continuation-aware downstream
loss difference. The latter remains a proposed target in this report rather
than a demonstrated general improvement over feature-based admission.

\paragraph{Governance and operation languages.}
MemOS (version~4) explicitly distinguishes plaintext, activation, and parameter
memory, with MemCubes carrying payloads, identities, governance metadata, and
usage information \cite{li2025memosmemoryosai}. Its version chains,
scheduling, access control, and cross-type transformations are relevant prior
art for lifecycle management and consolidation. The separation of state types
and independently stored memory therefore precedes this report. Its system
description and examples do not establish our particular finite-bit carrier
accounting or policy-by-memory causal effect.

Text2Mem is \emph{Text2Mem: A Unified Memory Operation Language for Memory
Operating System}, in Findings of ACL 2026, pages 2105--2119
\cite{wang-etal-2026-text2mem}. It specifies twelve verbs and a typed operation
schema, processed by a validator--parser--adapter pipeline. Its SQL reference
backend tests execution of governed transitions, including locking, expiry,
and scope constraints. It explicitly leaves large-scale cross-backend
comparisons for future work. A schema, CRUD operations, and a validator are
established tools. Our authoritative-memory particularization concerns bounded
numerical payload and control state, causal assembly, identity, and
publication/restoration obligations.

\begin{table}[tbp]
\centering\small
\caption{State and learning distinctions in the closest systems. These are
method descriptions, not a common benchmark. ``Policy'' denotes each paper's
own learnable decision mechanism, not a retroactive assignment to our $\Phi$.}
\label{tab:neighbor-state}
\begin{tabularx}{\linewidth}{@{}p{0.13\linewidth}Y Y@{}}
\toprule
System & State and update timing & Credit / admission distinction \\
\midrule
MCTR & Explicit knowledge and trajectory memory; periodic test-time policy updates & Majority-action agreement; add/delete/keep \\
UMA & Core/bank maintenance over chunks; QA after construction & Mean branched QA utility plus tool reward; task-stratified advantages \\
Mem-T & Hierarchical construction, evolution, and multi-turn retrieval & Retrieval-tree rewards and hindsight-selected construction actions; IGNORE \\
Memory-R1 & Learned manager and answer agent over an external bank & Downstream answer correctness; explicit NOOP \\
Mem-$\alpha$ & Learned construction; frozen retriever and generator & Final QA/compression plus operation-level validity/content rewards \\
AgeMem & Long-term store and short-term context, staged trajectories & Terminal advantage broadcast across stages; memory/context penalties \\
MemRL & Frozen LLM; updated utility in episodic triplets & Observed-reward utility update; empty-pool fallback \\
A-MAC & Candidate memory score at admission & Labeled admission selection; prospective utility feature and threshold \\
MemOS & Plaintext, activation, parameter payloads plus governance & Scheduling/transformation framework; no single universal learning rule \\
Text2Mem & Typed operations on an explicit store & Validated state transitions; not itself a utility-learning algorithm \\
TTRL & Numerical model parameters during test adaptation & Estimated majority-label reward \\
TTT-Discover & Numerical policy plus reused solution buffer within one problem & Entropic discovery objective and PUCT-inspired reuse \\
\bottomrule
\end{tabularx}
\end{table}

\subsection{What the comparison does and does not establish}
\label{sec:novelty-boundary}
The common ingredients are well precedented: external memory, stateful
retrieval, policy adaptation, CRUD, abstention, downstream rewards, value
learning, reset controls, and consolidation. The mathematical tools in the five
theory parts---conditional mutual information, finite-channel capacity,
noninterference arguments, union bounds, factorial contrasts, and
dynamic-programming recursions---are likewise standard tools. Their
contribution here is the explicit specialization to a typed, bounded,
causally scheduled learning-and-memory contract and its falsifiable
obligations.

\paragraph{Lifecycle and future information.}
MCTR performs adaptation during environment interaction; TTT-Discover updates
within a fixed discovery problem; UMA, Mem-T, Memory-R1, Mem-$\alpha$, and
AgeMem use later task outcomes to train earlier memory decisions under their
respective trajectories. These timing facts must not be collapsed into one
definition of same-problem RTT. Labels, evidence annotations, and outcomes
may be visible to training supervision while unavailable to the deployed
writer. For our design, that distinction is formalized through causal prefixes
and measurable deployment decisions; later outcomes are excluded from
deployment prefix features even when they supervise the policy offline.

\paragraph{Independent recovery and cost.}
AgeMem's context reset, MemOS's version-chain design, and Text2Mem's governed
state transitions provide relevant but different recovery mechanisms. A
complete independent $\Phi$/$M$ restore experiment with all our invariants
is not established by the particular comparisons above. This is an evidence
boundary, not an assertion of impossibility or absence. The reported costs
also have different units: model-training rollouts, tool calls, retrieval,
memory size, latency, and discovery search budgets. A-MAC explicitly charges
LLM utility calls in its latency comparison; UMA reports session-budget and
memory-efficiency analyses; TTT-Discover specifies training and rollout
budgets. Our small-model GPU update count cannot establish a cost advantage
over any of them. Teacher-cache creation and storage in
Section~\ref{sec:coverage-readout} are real additional costs, not free inputs.

\paragraph{The claim retained in this report.}
We contribute a full conditional specification, associated proofs and
counterexamples, and an empirical separation of several component successes
from failed broader qualifications. The post-training studies establish some
restricted content-dependent generation and reveal strong limits under changed
word and event distributions. They do not validate the complete dual-state
system or a learned predictive-admission policy. A future comparison must
instantiate the alternatives on matched inputs, state lifetimes, evidence
access, feedback and compute, and then measure the specified effects. No
universal superiority or first-in-the-world claim follows from the present
literature comparison.

%% file: theory_family/modules/index.tex
% Integrated formal module registry.
\ifRtfIncludeRTT
  \input{theory_family/modules/rtt/module}

\fi
\ifRtfIncludeAMR
  \input{theory_family/modules/amr/module}

\fi
\ifRtfIncludePMA
  \input{theory_family/modules/pma/module}

\fi
\ifRtfIncludeDualState
  \input{theory_family/modules/dual_state/module}

\fi
\ifRtfIncludeCSBC
  \input{theory_family/modules/csbc/module}

  \input{theory_family/final_synthesis}
\fi

%% file: theory_family/modules/rtt/module.tex
\clearpage
\part{Reasoning-Time Training Foundations}
\label{tf:rtt:part:foundations}

\noindent\textbf{Scope.}
This part defines a typed within-problem policy-update regime, its causal
witnesses, identification assumptions, reset requirements, safety boundaries,
and falsification protocols. Its empirical relationship to the complete
architecture is summarized in the report-wide evidence sections.

\input{theory_family/modules/rtt/sections/01_scope}
\input{theory_family/modules/rtt/sections/02_process}
\input{theory_family/modules/rtt/sections/03_witness}
\input{theory_family/modules/rtt/sections/04_separation}
\input{theory_family/modules/rtt/sections/05_identification}
\input{theory_family/modules/rtt/sections/06_updates_objective}
\input{theory_family/modules/rtt/sections/07_safety}
\input{theory_family/modules/rtt/sections/08_obligations}
\input{theory_family/modules/rtt/sections/09_related_work}
\input{theory_family/modules/rtt/sections/10_falsification}
\input{theory_family/modules/rtt/sections/11_discussion}

\FloatBarrier
\input{theory_family/modules/rtt/appendices/A_proofs}
\input{theory_family/modules/rtt/appendices/B_symbols}
\input{theory_family/modules/rtt/appendices/C_audits}

%% file: theory_family/modules/rtt/sections/01_scope.tex
\section{Introduction}
\label{tf:rtt:sec:scope}

\begin{figure}[H]
\centering
\includegraphics[width=\linewidth]{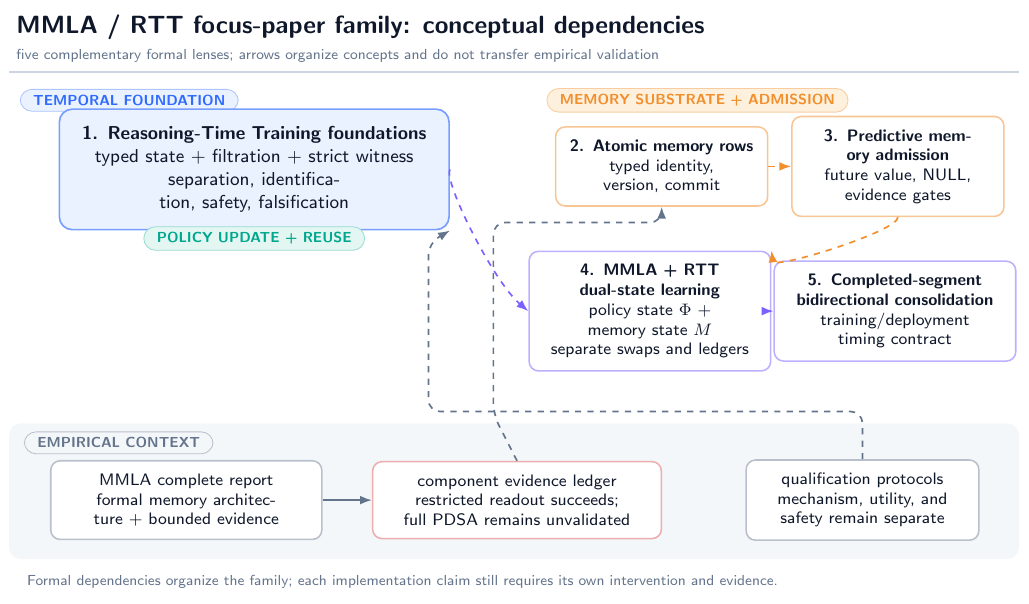}
\caption{Reasoning-time training: the within-problem policy mechanism, its causal witness, and its relationship to the other theory divisions. Conditional results do not constitute an executed policy-only intervention.}
\label{tf:rtt:fig:technology-tree}
\end{figure}

Suppose a system receives one difficult problem, produces an unsuccessful attempt, observes a compiler error or verifier score, and later succeeds. The success alone does not tell us what changed. The system may have sampled a luckier trajectory from the same policy, expanded a larger search tree, appended the error to its prompt, stored a reflection in external memory, or changed a state that parameterizes the policy. All five mechanisms can be useful. Only the last is a candidate for the phenomenon studied here, and even then the update must occur and be reused before that same problem is resolved.

We use \emph{Reasoning-Time Training} (\tfRttName{}) for this within-problem temporal regime. The name is descriptive rather than a claim of lexical priority. ``Training'' refers to a change in a declared policy-generating state, not necessarily to backpropagation. ``Reasoning time'' is the interval between admission of one problem instance and its resolution or abandonment. The boundary is therefore operational: state type, event order, read trace, reset scope, and intervention response must be inspectable.

The relationship to \tfRttMMLA{} is deliberately narrow. At the September 13, 2026 cutoff, the report has historical semantic-interface failures and later restricted latent readout successes, but no full authoritative-memory qualification or learned predictive overwrite. The later adapters are trained offline and then frozen for evaluation; that history is not an in-problem update--reuse witness. Those component results are therefore not evidence for \tfRttName{}. This part separates policy mutation from external-memory mutation so that a combined system can ablate their causal roles independently.

\subsection{Contributions}

The contributions are:
\begin{enumerate}
  \item a typed stochastic process for a single unresolved problem, with attempt, feedback, policy, context, memory, workspace, verifier, lifetime, and budget objects;
  \item a strict witness definition that requires an in-lifetime policy-state update and an actually post-update-policy-conditioned later attempt;
  \item separation and non-identifiability results clarifying what transcripts alone cannot establish;
  \item causal qualification designs and estimands that isolate update use, feedback semantics, reset leakage, verifier reliability, and local harm;
  \item an update-operator taxonomy, outer objective, budget ledger, rollback contract, implementation obligations, and a staged falsification ladder.
\end{enumerate}

Formal statements carry their assumptions locally. Status tags distinguish definitions, derived results, design prescriptions, and future tests; none converts a theorem into implementation evidence. No experiment in this theoretical part demonstrates \tfRttName{}, no theorem guarantees that an optimizer finds a useful update, and no result establishes superiority over search, context refinement, memory, or offline training. Policy mutation may be unnecessary, unstable, or inefficient. We neither relabel arbitrary cache mutation as training nor infer strict \tfRttName{} from a method name.

\subsection{Episode boundary and paper map}

The unit is a \emph{problem episode}: one admitted instance, its ordered attempts, its within-episode feedback, and a terminal resolution or abandonment decision. A benchmark dataset is a collection of such episodes, not one giant problem. An update learned from earlier dataset items and used on later items is test-time or continual adaptation, but is not strict \tfRttName{} for a later item unless that later item itself contains an update--reuse witness before its own endpoint. Conversely, a harmful within-problem update can satisfy the mechanism definition even though it fails every utility gate. Mechanism qualification and benefit qualification are separate questions throughout.

Section~\ref{tf:rtt:sec:process} specifies the typed process; Section~\ref{tf:rtt:sec:witness} gives the strict witness; Sections~\ref{tf:rtt:sec:separation}--\ref{tf:rtt:sec:identification} establish separations and identification limits; and the remaining sections develop update objectives, safety contracts, implementation obligations, related work, and falsification tests. Full proofs, symbols, and audit-oriented derivations appear in the appendices.

%% file: theory_family/modules/rtt/sections/02_process.tex
\section{A Typed Within-Problem Stochastic Process}
\label{tf:rtt:sec:process}

\subsection{Probability space and typed interfaces}

Fix a probability space $(\Omega,\mathcal{A},\tfRttP)$. A problem episode draws $X:\Omega\to\tfRttCalX$ and is assigned an immutable episode identifier $I\in\mathcal{I}$. The identifier distinguishes a retry of the same problem from a later, merely similar item. Within the episode, attempts are indexed by $j\in\mathbb{N}$ and token/tool microsteps by $t\in\mathbb{N}$. The basic spaces are collected in Table~\ref{tf:rtt:tab:typed-spaces}; they are measurable spaces even when implemented as structured byte records.

\begin{table}[H]
\centering
\small
\begin{tabularx}{\linewidth}{@{}p{0.15\linewidth}p{0.18\linewidth}Y Y@{}}
\toprule
Object & Space & Intended content & Required boundary \\
\midrule
Problem $X$ & $\tfRttCalX$ & Immutable task statement, tools, and success contract. & Same episode identifier across witness attempts. \\
Observation $O_{j,t}$ & $\tfRttCalO$ & Tokens, tool returns, environment state visible by time $(j,t)$. & Timestamped; no future returns. \\
Action $A_{j,t}$ & $\tfRttCalA$ & Token, tool call, branch, halt, or answer action. & Sampled from the declared policy kernel. \\
Attempt $Z_j$ & $\tfRttCalZ$ & Finite marked trajectory of observations and actions. & Start/end and policy-state receipt recorded. \\
Feedback $Y_j$ & $\tfRttCalY$ & Verifier output, execution result, critique, or reward record. & Provenance and availability time recorded. \\
Policy state $\Phi_j$ & $\tfRttCalP$ & Mutable state directly parameterizing the action kernel. & Versioned, bounded, intervenable, and read-traced. \\
Context $C_j$ & $\tfRttCalC$ & Prompt and transient in-context transcript. & Not silently counted as policy state. \\
Memory $M_j$ & $\tfRttCalM$ & External or resident episodic/semantic records. & Typed separately from policy state. \\
Workspace $W_j$ & $\tfRttCalW$ & Files, search tree, scratchpad, caches, tool state. & Charged and reset-scoped explicitly. \\
Auxiliary state $\Xi_j$ & $\mathfrak X^{\mathrm{read}}\!\times\!\mathfrak X^{\mathrm{aux}}$ & Policy-read fields; RNG, optimizer, router, scheduler, callback, and epoch records. & Typed, bounded, versioned; every read path declared. \\
Budget $B_j$ & $\tfRttCalB\subseteq\mathbb{R}_+^d$ & Remaining tokens, calls, time, energy proxy, update steps. & Monotone ledger with hard stop. \\
\bottomrule
\end{tabularx}
\caption{\textbf{Typed state decomposition (\tfRttDefinitionStatus{}).} Physical storage does not determine the type. The causal interface and declared read path do.}
\label{tf:rtt:tab:typed-spaces}
\end{table}

Every admitted event $e$ receives the immutable key
\begin{equation}
\kappa(e)=\bigl(g(e),p(e),\operatorname{id}(e)\bigr),
\label{tf:rtt:eq:event-key}
\end{equation}
where $g$ is an atomically assigned monotone ingress sequence, $p$ is a frozen event-class precedence, and $\operatorname{id}$ is a stable receipt identifier. Events are replayed in lexicographic key order; wall-clock timestamps are descriptive and never break a causal tie. Let $\mathsf S_n$ be the full state immediately before event $n$ in that order. If $\sigma_j$ is the start event of attempt $j$, then the pre-attempt state is
\begin{equation}
S_j:=\mathsf S_{\sigma_j}=(X,I,C_j,M_j,W_j,\Phi_j,\Xi_j,B_j),
\label{tf:rtt:eq:full-state}
\end{equation}
where $\Xi_j=(\Xi_j^{\mathrm{read}},\Xi_j^{\mathrm{aux}})$. The first component contains every persistent non-$\Phi$ field directly read on an action-kernel path. The second contains non-policy auxiliary variables, including random-generator state, optimizer moments, routing and batching metadata, the reset epoch, and pending-callback receipts. Each component is assigned a measurable, implementation-bounded space with declared caps on serialized bytes, queue length, optimizer slots, and callback count. Excluding $\Xi$ from an audit does not make it causally irrelevant.

Define the \emph{policy-read closure} at an action call as the versioned manifest of every persistent field reachable by that call. A field in this closure is either serialized into $\Phi$, placed in $\Xi^{\mathrm{read}}$, or the policy-use claim stops. Write $\Psi:=(\Phi,\Xi^{\mathrm{read}})$ for the effective policy state. A contrast may be called a $\Phi$ effect only when $\Xi^{\mathrm{read}}$ is byte-identical across arms; otherwise it is a bundled $\Psi$ effect. Fields in $\Xi^{\mathrm{aux}}$ remain causally relevant and must be restored, randomized, or included in the estimand.

\begin{tfRttDefinition}[Policy kernel and functional policy state]
\tfRttDefinitionStatus{} For each microstep, the system exposes a stochastic kernel
\begin{equation}
\pi_{\phi}(da\mid x,h,c,m,w,\xi^{\mathrm{read}},b),
\label{tf:rtt:eq:policy-kernel}
\end{equation}
where $h$ is the causally available within-attempt history. A declared component $\Phi$ is a \emph{functional policy state} on a history set $\mathcal{H}_0$ if there exist $h\in\mathcal{H}_0$, admissible states $\phi,\phi'$, and fixed $(x,c,m,w,\xi^{\mathrm{read}},b)$ such that
\begin{equation}
\tfRttTV\!\left(\pi_{\phi}(\cdot\mid x,h,c,m,w,\xi^{\mathrm{read}},b),
             \pi_{\phi'}(\cdot\mid x,h,c,m,w,\xi^{\mathrm{read}},b)\right)>0.
\label{tf:rtt:eq:functional-state}
\end{equation}
The declaration must identify the serialization, version, update rule, and action-kernel call site. A semantic relabeling of context or memory without such an interface is insufficient.
\end{tfRttDefinition}

This definition is intentionally substrate-neutral. Full weights, a low-rank adapter, a learned-optimizer state, a recurrent fast-weight state, or a hypernetwork code can all be functional policy state. Conversely, a vector called an ``adapter'' that is never read is not functional on the observed episode. External memory can influence actions after retrieval, but its mutation remains a memory operation unless it is also the declared direct policy-generating state under an auditable interface.

\subsection{Filtration, attempt times, and resolution}

Let $\{\tfRttCalF_n\}_{n\ge 0}$ be the filtration generated by the deterministically ordered event log. It contains the problem and every admitted observation, action, random-seed draw, verifier return, state version, reset/drop receipt, and external side effect available before event $n{+}1$. For attempt $j$, let $\sigma_j$ and $\epsilon_j$ be start and end stopping times with $\sigma_j\le\epsilon_j<\sigma_{j+1}$ on episodes that retry. Let $u_j$ be the committed-update event triggered after attempt $j$. The resolution time $\tau$ is a stopping time taking values in $\mathbb{N}\cup\{\infty\}$; resolution includes verified success, explicit abandonment, budget exhaustion, or a safety stop. A remote or asynchronous return becomes causally available only when its receipt is admitted as an event. Returns carrying a stale reset epoch are rejected and the rejection itself is logged; they cannot re-enter the post-reset state.

Within an attempt, $n(j,t)$ is the unique event-log embedding of microstep $(j,t)$ and $\tfRttCalF_{j,t}:=\tfRttCalF_{n(j,t)}$ denotes the information set before action $A_{j,t}$. Causality requires
\begin{equation}
A_{j,t}\ \text{is sampled from an }\tfRttCalF_{j,t}\text{-measurable kernel},
\qquad O_{j,t+1}\in\tfRttCalF_{j,t+1}.
\label{tf:rtt:eq:adapted-actions}
\end{equation}
Feedback $Y_j$ becomes available at a stopping time $v_j\ge\epsilon_j$ and is $\tfRttCalF_{v_j}$-measurable. An admissible update has the form
\begin{equation}
(\Phi_{j+1},\Xi_{j+1},\rho_j)
=U_j(\Phi_j,\Xi_j,X,Z_{\le j},Y_{\le j},B_j;R_j),
\label{tf:rtt:eq:update-map}
\end{equation}
where $R_j$ is fresh update randomness revealed no later than $u_j$, and $\rho_j$ is a receipt containing input/output hashes, times, costs, and status. The entire right-hand side must be $\tfRttCalF_{u_j}$-measurable. In particular, no observation after $u_j$ may select or tune the update used before that observation.

\begin{figure}[H]
\centering
\includegraphics[width=\linewidth]{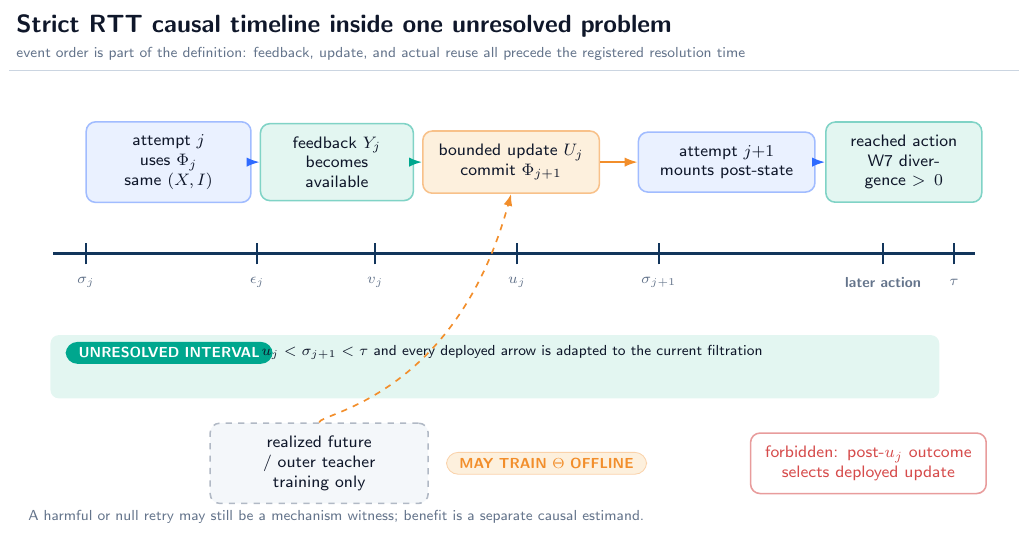}
\caption{\textbf{Strict within-problem causal timing (\tfRttDefinitionStatus{}).} Solid arrows are deployed, causally available paths. Attempt $j$ ends, feedback becomes available, and update $U_j$ commits while the same episode is unresolved; only then may attempt $j{+}1$ invoke $\Phi_{j+1}$. The dashed amber future branch is allowed only for an outer training teacher and may not enter the deployed update.}
\label{tf:rtt:fig:causal-timeline}
\end{figure}

\begin{tfRttDefinition}[Unresolved interval and lifetime]
\tfRttDefinitionStatus{} The episode is unresolved at event $n$ on $\{n<\tau\}$. A policy-state version $\phi$ has a declared lifetime interval $L(\phi)=[\ell^{-}(\phi),\ell^{+}(\phi))$ in event time. A within-problem version must satisfy
\begin{equation}
\sigma_1\le \ell^{-}(\phi)<\ell^{+}(\phi)\le \tau+1,
\label{tf:rtt:eq:lifetime}
\end{equation}
unless an explicitly separate cross-problem continual-learning protocol owns the carryover. Retaining a version after $\tau$ silently changes the regime.
\end{tfRttDefinition}

\begin{figure}[H]
\centering
\includegraphics[width=\linewidth]{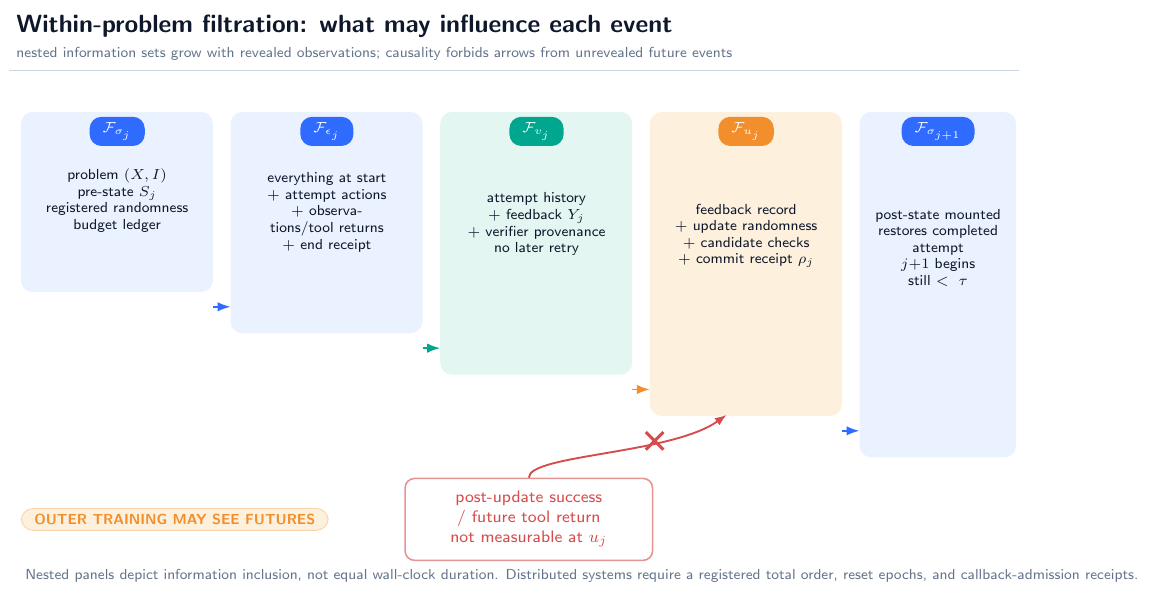}
\caption{\textbf{Information sets and admissible arrows (\tfRttDefinitionStatus{}).} Each blue filtration panel contains only events already revealed. Policy actions, feedback processing, and the update receipt are adapted to the current filtration. The crossed coral arrow is future leakage: later success or a post-update tool result cannot be used to choose an earlier deployed update.}
\label{tf:rtt:fig:filtration}
\end{figure}

\subsection{Standing assumptions for formal results}

Results invoke only the assumptions they cite; no assumption is globally true by typography.

\begin{tfRttAssumption}[Typed observability]
\label{tf:rtt:assump:typed}
\tfRttAssumptionStatus{} The declared state components in Equation~\eqref{tf:rtt:eq:full-state}, their bounds, versions, policy-read closure, and read/write events are logged without omission for the analyzed episode.
\end{tfRttAssumption}

\begin{tfRttAssumption}[Causal availability]
\label{tf:rtt:assump:causal}
\tfRttAssumptionStatus{} Actions and updates are adapted to $\{\tfRttCalF_n\}$; event keys obey Equation~\eqref{tf:rtt:eq:event-key}; stale-epoch returns cannot mutate current state; and $\tau$ is a stopping time for that filtration.
\end{tfRttAssumption}

\begin{tfRttAssumption}[Stable episode identity]
\label{tf:rtt:assump:identity}
\tfRttAssumptionStatus{} Witness attempts share the same immutable pair $(X,I)$; a generated paraphrase or neighboring dataset item is not silently substituted.
\end{tfRttAssumption}

\begin{tfRttAssumption}[Well-defined interventions]
\label{tf:rtt:assump:intervention}
\tfRttAssumptionStatus{} When a result compares state versions or feedback arms, all non-target state fields and environment kernels are either held fixed, randomized as specified, or included in the estimand.
\end{tfRttAssumption}

\begin{tfRttAssumption}[Positivity and consistency]
\label{tf:rtt:assump:positivity}
\tfRttAssumptionStatus{} Every randomized experimental arm has positive assignment probability, and the observed outcome equals the potential outcome under the assigned arm without cross-episode interference.
\end{tfRttAssumption}

\begin{tfRttAssumption}[Bounded evaluation]
\label{tf:rtt:assump:bounded}
\tfRttAssumptionStatus{} Utility and harm variables used in stability bounds lie in $[0,1]$, or are rescaled with an explicitly reported range.
\end{tfRttAssumption}

%% file: theory_family/modules/rtt/sections/03_witness.tex
\section{Strict Reasoning-Time Training}
\label{tf:rtt:sec:witness}

\subsection{A witness is a trace plus a functional state test}

The definition below is intentionally stronger than ``an optimizer ran.'' A no-op optimizer, an update committed after resolution, or an updated adapter that is never invoked does not establish within-problem learning.

\begin{tfRttDefinition}[Strict RTT witness]
\label{tf:rtt:def:strict-rtt}
\tfRttDefinitionStatus{} A strict \tfRttName{} witness for attempt index $j$ is a record
\begin{equation}
\mathfrak{W}_j=(I,j,\epsilon_j,v_j,u_j,\sigma_{j+1},\tau,
h_j^-,h_j^+,\chi_j^-,\chi_j^+,\rho_j,\mathcal{R}_{j+1},d_j)
\label{tf:rtt:eq:witness-record}
\end{equation}
that satisfies all of the following:
\begin{enumerate}[label=\textbf{W\arabic*.},leftmargin=3.2em]
  \item \textbf{Same problem.} Attempts $j$ and $j{+}1$ carry the same immutable $(X,I)$.
  \item \textbf{Available feedback.} Attempt $j$ ends at $\epsilon_j$, feedback $Y_j$ is available by $v_j$, and $Y_j\in\tfRttCalF_{v_j}$.
  \item \textbf{Nontrivial policy-state transition.} Update $U_j$ commits at $u_j\ge v_j$, producing a version change $h_j^-:=H(\Phi_j)\ne H(\Phi_{j+1})=:h_j^+$ under a collision-resistant serialization hash and a valid receipt $\rho_j$. The receipt also records the policy-read-closure manifests $\chi_j^-:=H(\Xi_j^{\mathrm{read}})$ and $\chi_j^+:=H(\Xi_{j+1}^{\mathrm{read}})$.
  \item \textbf{Causal update.} $(\Phi_{j+1},\Xi_{j+1},\rho_j)$ is $\tfRttCalF_{u_j}$-measurable; its commit carries the admitted event key and current reset epoch, and no event after $u_j$ selects the committed update.
  \item \textbf{Problem still unresolved.} $u_j<\sigma_{j+1}<\tau$.
  \item \textbf{Actual post-update read.} The later attempt contains a signed read trace $\mathcal{R}_{j+1}$ showing that the action-kernel call loaded version $h_j^+$ and the registered closure $\chi_j^+$, not merely that either version existed.
  \item \textbf{Functional reuse on a reached history.} For at least one reached pre-action history $H_{j+1,t}$, hold every registered non-policy input fixed at its recorded value. Then
  \begin{equation}
  d_j:=\tfRttTV\!\left(\pi_{\Phi_{j+1}}(\cdot\mid H_{j+1,t},\Xi^{\mathrm{read}}),
                    \pi_{\Phi_j}(\cdot\mid H_{j+1,t},\Xi^{\mathrm{read}})\right)>0.
  \label{tf:rtt:eq:functional-reuse}
  \end{equation}
  This is a $\Phi$-only contrast only when $\chi_j^-=\chi_j^+$ and all other fields are fixed. If a policy-read auxiliary field changes, the target is the bundled effective state $\Psi=(\Phi,\Xi^{\mathrm{read}})$ and the witness is labeled accordingly. The distance may be computed exactly, lower-bounded with a registered probe, or established by a randomized state-swap test whose uncertainty is reported.
  \item \textbf{Charged lifetime.} Update, retained state, retry, and verification costs are charged to $B_j$, and the state version expires or transfers under a declared rule at $\tau$. Expiry, rollback, and late-callback rejection are scoped by the recorded reset epoch.
\end{enumerate}
No positive reward, correctness, or improvement condition is part of the mechanism definition.
\end{tfRttDefinition}

The hash inequality in W3 is necessary but not sufficient: different bytes can implement the same kernel. W6 catches dead updates; W7 catches semantically inert updates. The closure hashes prevent an unlogged router, gate, recurrent carrier, or normalization field from being attributed to $\Phi$. Conversely, W7 does not show that the change helped. It only shows that a later attempt was genuinely conditioned on a declared effective policy state produced from earlier within-problem feedback.

\begin{tfRttDefinition}[Syntactic, functional, and beneficial qualification]
\tfRttDefinitionStatus{} A trace satisfying W1--W6 and W8 is a \emph{syntactic update--reuse trace}. It becomes a \emph{strict functional \tfRttName{} witness} only with W7. A \emph{beneficial causal \tfRttName{} effect} additionally requires a prespecified outcome estimand and an intervention or randomized design supporting a positive effect under its assumptions. These three levels must not be conflated.
\end{tfRttDefinition}

\begin{table}[H]
\centering
\small
\begin{tabularx}{\linewidth}{@{}p{0.10\linewidth}Y p{0.24\linewidth}p{0.22\linewidth}@{}}
\toprule
Item & Required evidence & Typical falsifier & Status if absent \\
\midrule
W1 & Immutable problem/episode receipt & New item or regenerated task ID & Different regime \\
W2 & Timestamped feedback provenance & Feedback created after update & Causal failure \\
W3 & $\Phi$ hashes, closure hashes, and commit & No-op, failed commit, or undeclared read carrier & Not strict RTT or bundled $\Psi$ \\
W4 & Event key, epoch, filtration/future-read audit & Later success or stale callback tunes update & Leakage \\
W5 & $u_j<\sigma_{j+1}<\tau$ & Update after answer acceptance & Too late \\
W6 & Kernel call reads post-state and closure & Adapter never mounted or route differs & Dead or bundled update \\
W7 & Exact/probed swap with all non-target fields fixed & Zero divergence or closure mismatch & Functionally inert or misattributed \\
W8 & Cost/lifetime/epoch ledger & Uncharged state or late return re-enters & Protocol invalid \\
\bottomrule
\end{tabularx}
\caption{\textbf{Strict witness audit (\tfRttDefinitionStatus{}).} Each row has an independently inspectable artifact and an explicit failure interpretation.}
\label{tf:rtt:tab:witness-audit}
\end{table}

\subsection{Fresh-proposal criterion}

Search workspaces can strongly influence which node is expanded next. To avoid declaring every search statistic a learned policy, the state partition uses a fresh-proposal intervention.

\begin{tfRttDefinition}[Fresh-proposal policy-state criterion]
\label{tf:rtt:def:fresh-proposal}
\tfRttDefinitionStatus{} Let $r$ restore $(X,C,M,W,\Xi^{\mathrm{read}},\Xi^{\mathrm{aux}},B)$ to a registered pre-attempt snapshot while independently choosing policy version $\phi$. A mutable state component qualifies as policy state for the episode only if, under some admissible restored snapshot and common randomization scheme, swapping that component changes the distribution of a \emph{fresh proposal} before any updated search node, appended reflection, or new memory record is read. A $\Phi$-only attribution additionally requires byte-identical $\Xi^{\mathrm{read}}$; otherwise the intervened component is $\Psi$. Candidate-specific tree values, visited-node sets, and cached tool outputs are workspace state by default.
\end{tfRttDefinition}

This criterion does not assert that workspace search is inferior. It enforces a stable meaning: \tfRttName{} changes how new reasoning actions are proposed; search changes which already represented possibilities are explored or selected. A hybrid can do both and should report both.

\begin{figure}[H]
\centering
\includegraphics[width=\linewidth]{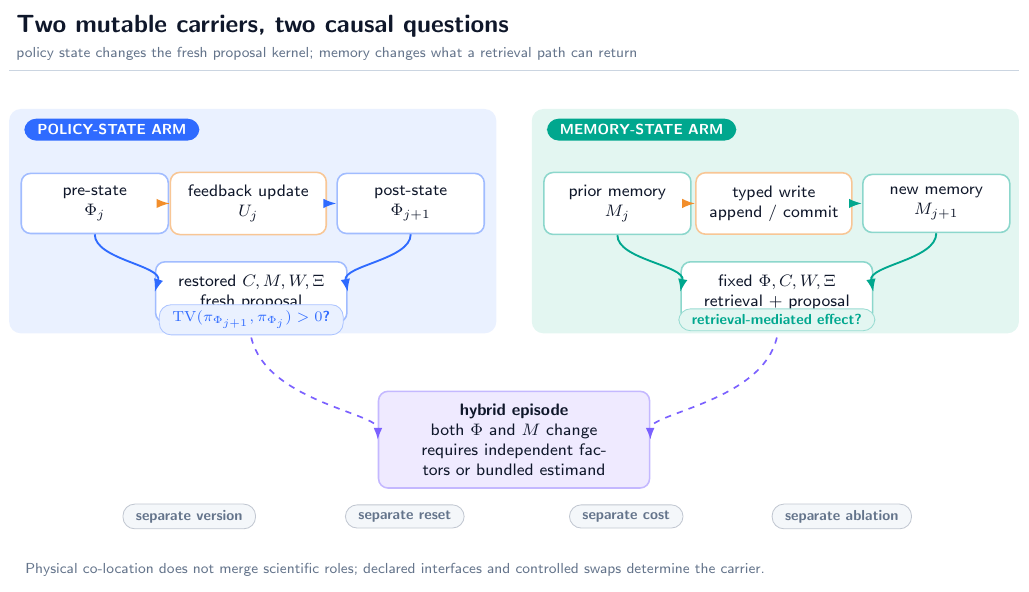}
\caption{\textbf{Mutable policy state versus external memory and workspace (\tfRttDefinitionStatus{}).} The navy state-swap holds context, memory, workspace, the policy-read closure, auxiliary state, and randomness fixed and asks whether a fresh proposal distribution changes. Teal memory swaps test retrieval-mediated effects separately. If the closure cannot be held fixed, the navy contrast is a bundled $\Psi$ effect. Each state type needs its own version, reset, cost, and ablation receipt.}
\label{tf:rtt:fig:policy-memory}
\end{figure}

\subsection{A worked strict witness}

Consider a coding instance with a deterministic compiler verifier. Attempt 1 produces program $z_1$ and error $y_1$. A rank-$r$ adapter $\Phi_1=(A_1,B_1)$ is updated to $\Phi_2$ by $K$ bounded optimizer steps using only $(X,z_1,y_1)$, and the commit receipt is written before retry. The prompt, external memory, compiler cache, workspace, RNG streams, optimizer/router state, pending-callback set, reset epoch, and the entire policy-read closure are then restored to their registered pre-attempt values. Attempt 2 mounts $\Phi_2$. A common-random-number logit probe at a reached prefix reports positive total variation between the distributions under $\Phi_2$ and restored $\Phi_1$. This is a strict functional $\Phi$ witness because the closure is byte-identical, regardless of whether program 2 passes.

The example also shows why ``the second answer differed'' is inadequate. Different sampling randomness can change an answer with no update, and hidden compiler caches can change it despite a reset. The receipt must cover the state version, actual call path, controlled fields, and measurement uncertainty.

\subsection{Hybrid systems and dual accounting}

A system may update $\Phi$, change a policy-read field in $\Xi^{\mathrm{read}}$, append a reflection to $C$, write $M$, and expand $W$ after the same attempt. Such a trace is not rejected, but the total causal effect is then a mixture. The minimal factorial decomposition uses independent restore/swap factors for $\Phi$, $\Xi^{\mathrm{read}}$, and non-policy information carriers. If cost permits, context, memory, workspace, auxiliary state, verifier feedback, and policy version receive separate randomized factors. If cost does not permit, the reported estimand must be the bundled $\Psi$ or wider system effect; it may not be called a $\Phi$-only policy-update effect.

\begin{figure}[H]
\centering
\includegraphics[width=\linewidth]{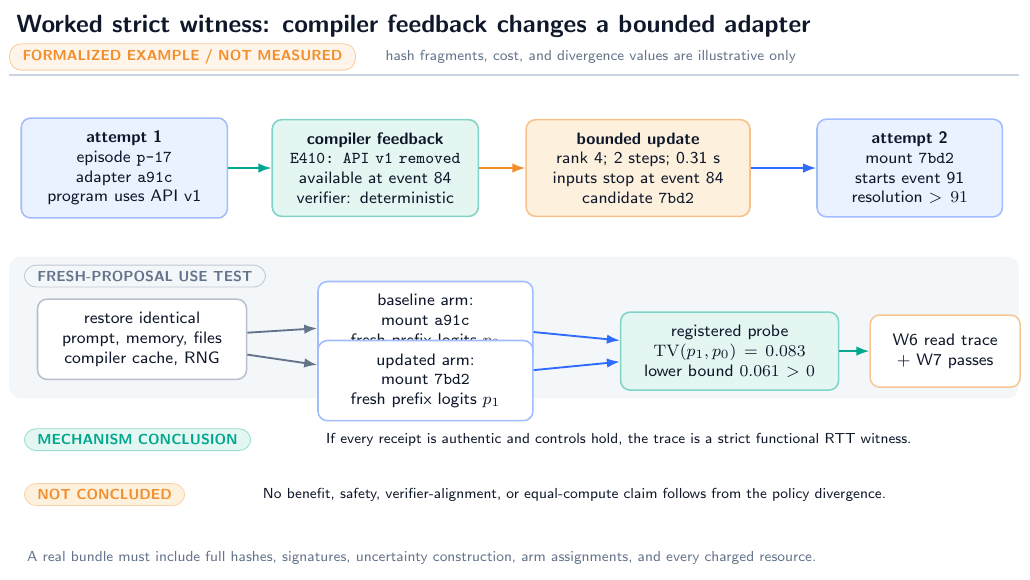}
\caption{\textbf{Worked multi-attempt trace (formalized example, not measured).} Attempt 1 yields compiler feedback; a bounded adapter update commits while the episode is unresolved; attempt 2 actually mounts the new version. The lower restored-state branch is the required policy-use probe. All numerical hashes, costs, and distances are illustrative and carry no empirical claim.}
\label{tf:rtt:fig:worked-attempt}
\end{figure}

%% file: theory_family/modules/rtt/sections/04_separation.tex
\section{Separation from Search, Context, and Memory}
\label{tf:rtt:sec:separation}

\subsection{Mechanism taxonomy}

\begin{figure}[H]
\centering
\includegraphics[width=\linewidth]{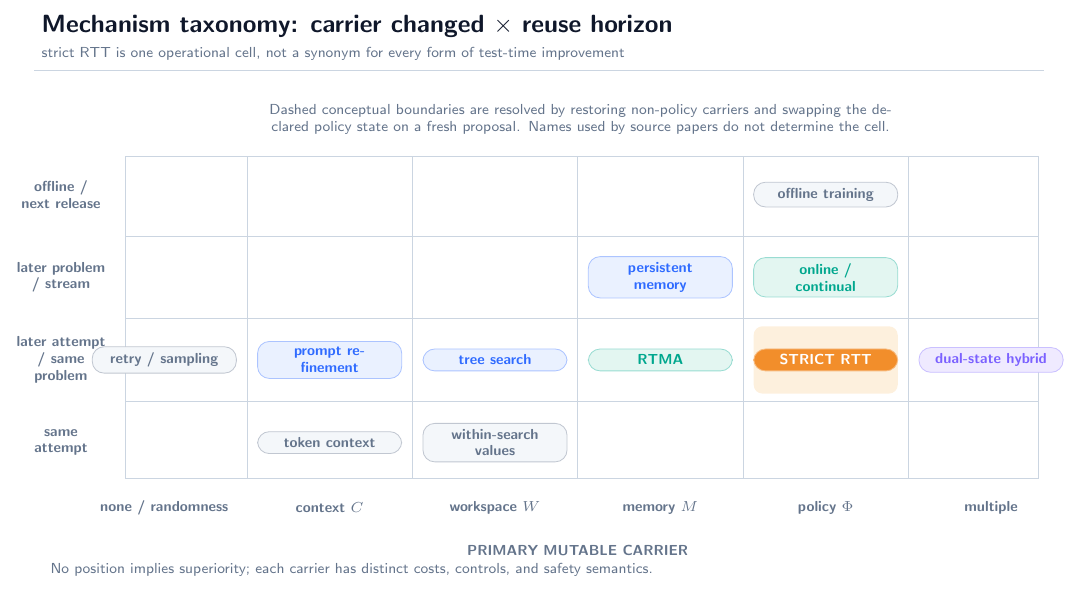}
\caption{\textbf{RTT versus neighboring mechanisms (\tfRttDefinitionStatus{}).} Horizontal position records the carrier that changes; vertical position records reuse time. Strict RTT occupies the policy-state/same-unresolved-problem cell. Dashed borders indicate hybrids whose classification requires state-specific swaps; no cell is presented as universally superior.}
\label{tf:rtt:fig:taxonomy}
\end{figure}

The adjacent regimes differ along three axes: what state changes, when it changes, and which future unit uses it. Table~\ref{tf:rtt:tab:regime-taxonomy} and Figure~\ref{tf:rtt:fig:taxonomy} use the strict witness rather than community labels.

\begin{table}[H]
\centering
\small
\begin{tabularx}{\linewidth}{@{}p{0.19\linewidth}p{0.19\linewidth}p{0.19\linewidth}Y@{}}
\toprule
Regime & Primary mutable carrier & Reuse horizon & Strict RTT condition \\
\midrule
Fixed-policy sampling/search & $W$ (samples, nodes, scores) & Same problem & No, unless $\Phi$ also changes and passes fresh-proposal reuse. \\
Prompt/context refinement & $C$ & Same problem/session & No when $\Phi$ is fixed. \\
External memory / \tfRttMemoryAdaptation{} & $M$ & Later attempts or problems & No when only $M$ changes. \\
Classical TTT/adaptation & $\Phi$ & Current sample, batch, or stream & Only episodes with update and reuse before that same problem ends. \\
Continual/online learning & $\Phi$ & Later problems & Not strict RTT for the earlier completed problem; may contain nested RTT episodes. \\
Strict RTT & $\Phi$ & Later attempt on same unresolved $(X,I)$ & Yes, if all W1--W8 hold. \\
\bottomrule
\end{tabularx}
\caption{\textbf{Operational regime taxonomy (\tfRttDefinitionStatus{}).} Names used by individual papers do not override the trace-level test.}
\label{tf:rtt:tab:regime-taxonomy}
\end{table}

\subsection{No-read and fixed-policy results}

\begin{tfRttLemma}[Unused policy update cannot change the later law]
\label{tf:rtt:lem:no-read}
\tfRttProvedStatus{} Fix an episode and suppose that, conditional on $(X,C,M,W,\Xi^{\mathrm{read}},\Xi^{\mathrm{aux}},B)$, every transition kernel from $u_j$ through attempt $j{+}1$ is independent of $\Phi_{j+1}$. Then replacing $\Phi_{j+1}$ by any other admissible state while holding the policy-readable closure byte-identical leaves the conditional law of the later attempt and its outcome unchanged.
\end{tfRttLemma}

\begin{proof}
Condition on the recorded non-policy state at $u_j$. The first post-update transition has the same kernel under every value of $\Phi_{j+1}$ by assumption. If the joint law through event $n$ is the same, integrating the common event-$(n{+}1)$ kernel against that common law gives the same joint law through $n{+}1$. Induction continues through the finite attempt and its outcome. Thus the conditional laws coincide. Full details and the kernel factorization are restated in Appendix~\ref{tf:rtt:app:proofs}.
\end{proof}

The lemma distinguishes a state commit from actual reuse. A framework can spend substantial update compute and still have exactly zero causal path to later actions because the adapter was not mounted, the route bypassed it, or the state affected only a dead branch.

\begin{tfRttProposition}[Fixed-policy search separation]
\label{tf:rtt:prop:search-separation}
\tfRttProvedStatus{} Consider any procedure that, during one episode, samples, scores, branches, backtracks, or votes using an action kernel with constant effective policy carrier $\Psi_j=(\Phi_j,\Xi_j^{\mathrm{read}})\equiv\psi_0$. If its only mutations are to $(C,M,W,\Xi^{\mathrm{aux}},B)$, the procedure has no strict \tfRttName{} witness, regardless of its compute budget or success rate.  Constancy of $\Phi$ alone is insufficient when the kernel-readable closure changes.
\end{tfRttProposition}

\begin{proof}
W3 in Definition~\ref{tf:rtt:def:strict-rtt} requires a nontrivial transition of the declared functional carrier, and W7 requires a later fresh-proposal law that differs under the post-update version. Byte-identical $\Psi$ violates W3 directly and leaves no carrier contrast for W7. Changes to the other typed components do not repair that violation. The conclusion is classificatory, not a claim that search is ineffective.
\end{proof}

\begin{tfRttProposition}[Context- and memory-only separation]
\label{tf:rtt:prop:memory-separation}
\tfRttProvedStatus{} If $C_{j+1}\ne C_j$ or $M_{j+1}\ne M_j$ while $\Psi_{j+1}=\Psi_j$, later behavior may change, but the trace is not strict \tfRttName{} under Definition~\ref{tf:rtt:def:strict-rtt}.  Equality of $\Phi$ alone does not establish this premise when $\Xi^{\mathrm{read}}$ changes.
\end{tfRttProposition}

\begin{proof}
The policy-state transition condition W3 fails. Because Definition~\ref{tf:rtt:def:fresh-proposal} holds $C$ and $M$ fixed when testing policy state, a retrieval- or prompt-mediated behavior change cannot substitute for W7. Such a trace should be reported as context refinement or memory adaptation, possibly \tfRttMemoryAdaptation{}, rather than erased from the analysis.
\end{proof}

\begin{tfRttCorollary}[More inference compute is not sufficient]
\label{tf:rtt:cor:compute}
\tfRttProvedStatus{} Increasing the number of samples, tree nodes, tokens, tool calls, or retries is not sufficient for strict \tfRttName{}.
\end{tfRttCorollary}

\begin{proof}
Each increase can be implemented with constant $\Phi$ and mutations confined to $W$, $C$, or $B$. Proposition~\ref{tf:rtt:prop:search-separation} applies.
\end{proof}

\subsection{Counterexamples to common classifications}

\begin{tfRttCounterexample}[Lucky retry]
\label{tf:rtt:ce:lucky}
\tfRttDefinitionStatus{} A fixed stochastic policy fails once and succeeds on a second independent sample. The outcome changes, but there is no policy-state update. This is repeated inference, not strict \tfRttName{}.
\end{tfRttCounterexample}

\begin{tfRttCounterexample}[Reflection in the prompt]
\label{tf:rtt:ce:prompt}
\tfRttDefinitionStatus{} A critique is appended to $C_{j+1}$ and a frozen model retries. The later attempt is feedback-conditioned through context, not post-update-policy-conditioned. This is prompt/context refinement.
\end{tfRttCounterexample}

\begin{tfRttCounterexample}[Episodic memory write]
\label{tf:rtt:ce:memory}
\tfRttDefinitionStatus{} The system writes ``avoid API v1'' to $M$ and retrieves it on retry while weights and fast policy state remain unchanged. The mechanism is memory adaptation, not strict \tfRttName{}.
\end{tfRttCounterexample}

\begin{tfRttCounterexample}[Update after termination]
\label{tf:rtt:ce:late}
\tfRttDefinitionStatus{} The correct answer is accepted at $\tau$, after which the model trains on the completed trace for future problems. That may be online or continual learning, but $u_j\ge\tau$ violates W5 for the completed problem.
\end{tfRttCounterexample}

\begin{tfRttCounterexample}[Mounted but inert adapter]
\label{tf:rtt:ce:inert}
\tfRttDefinitionStatus{} A new adapter hash is mounted, but a zero gate makes every reached action distribution identical to the pre-update distribution. W6 holds while W7 fails; this is only a syntactic update--reuse trace.
\end{tfRttCounterexample}

\begin{tfRttCounterexample}[Hidden cross-arm cache]
\label{tf:rtt:ce:cache}
\tfRttDefinitionStatus{} An apparent update benefit disappears when a compiler cache and random-number stream are restored. The original comparison bundled $\Phi$ with $W$, $\Xi^{\mathrm{read}}$, and $\Xi^{\mathrm{aux}}$; it identified neither a $\Phi$-only effect nor, without a declared readable closure, a well-typed $\Psi$ effect.
\end{tfRttCounterexample}

\begin{figure}[H]
\centering
\includegraphics[width=\linewidth]{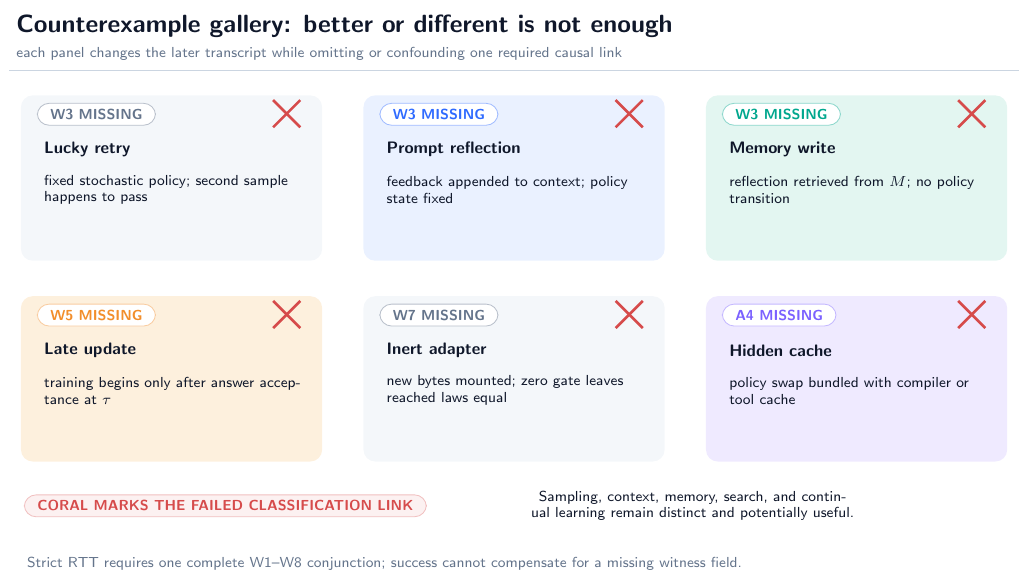}
\caption{\textbf{False-classification gallery (formal counterexamples).} Each panel can produce a different or better later answer without satisfying all strict-witness conditions. Coral marks the single missing or confounded link; it does not mark the underlying method as scientifically failed.}
\label{tf:rtt:fig:counterexamples}
\end{figure}

\subsection{Ambiguous recurrent and hypernetwork state}

Some architectures blur conventional storage labels. A recurrent fast state that persists across attempt resets and directly parameterizes the fresh proposal kernel may be $\Phi$ even if it is not called a weight. A hypernetwork code that regenerates adapter weights is also policy state if the code swap changes the fresh proposal law. By contrast, a recurrent token cache rebuilt solely from the appended transcript is context state. The decisive audit is not the tensor's name or device; it is the typed intervention with non-target carriers restored.

%% file: theory_family/modules/rtt/sections/05_identification.tex
\section{Identification, Intervention, and Impossibility}
\label{tf:rtt:sec:identification}

\subsection{Transcripts do not identify the hidden mechanism}

\begin{tfRttTheorem}[Finite-history emulation]
\label{tf:rtt:thm:emulation}
\tfRttProvedStatus{} Let an adaptive controller induce a joint distribution over a finite set of observable within-episode histories and actions, possibly through an unobserved mutable policy state. There exists a single fixed history-dependent policy $q(a\mid h)$, with no mutable policy state, that induces the same joint distribution over those observable histories and actions under the same environment kernels.
\end{tfRttTheorem}

\begin{proof}
For every observable history $h$ with positive probability under the adaptive controller, define
\begin{equation}
q(a\mid h):=\tfRttP_{\mathrm{ad}}(A=a\mid H=h),
\label{tf:rtt:eq:emulator}
\end{equation}
where the right side marginalizes any hidden policy state and update randomness. Define $q$ arbitrarily on zero-probability histories. At the first decision, $q$ matches the adaptive conditional action law. If the joint observable law matches through a history $h$, the action conditional matches by construction and the next-observation conditional matches because the environment kernel is shared. Induction over the finite event horizon gives equality of the full observable joint law. The map $q$ is fixed before the emulated episode; it conditions on history but does not update policy state.
\end{proof}

\begin{tfRttCorollary}[Observational non-identifiability of strict RTT]
\label{tf:rtt:cor:nonident}
\tfRttProvedStatus{} A finite transcript distribution, even one showing systematic improvement after feedback, cannot by itself identify a strict \tfRttName{} mechanism.
\end{tfRttCorollary}

\begin{proof}
Theorem~\ref{tf:rtt:thm:emulation} provides an observationally equivalent fixed-policy mechanism. Distinguishing the mechanisms therefore requires additional observables (state/read receipts) or interventions (state restoration/swaps).
\end{proof}

The theorem is an existence result, not a claim that the emulator is compact or learnable. Its purpose is epistemic: behavioral logs cannot certify the hidden update type merely because a fixed lookup policy might be impractically large.

\subsection{A randomized state--feedback factorial design}

Let $R\in\{0,1\}$ denote the policy-state deployment arm and $F\in\{0,1\}$ the feedback-semantic arm. In all arms, attempt 1 and update compute are executed. When $F=1$, the update receives the true registered feedback; when $F=0$, it receives a prespecified within-block permutation or matched sham feedback. When $R=1$, the resulting post-update state is mounted for the fresh retry; when $R=0$, the registered pre-update state is restored. Context, memory, workspace, tool state, budgets, $\Xi^{\mathrm{aux}}$, and randomization are restored or randomized identically. The policy-read closure $\Xi^{\mathrm{read}}$ is byte-identical for a $\Phi$-only estimand; if it changes with $R$, the registered target is instead the effective state $\Psi=(\Phi,\Xi^{\mathrm{read}})$. Let $Q(r,f)\in[0,1]$ be the potential qualification outcome for that declared target.

Define the true-feedback state effect and semantic interaction
\begin{align}
\theta_{\mathrm{state}}&=\tfRttE[Q(1,1)-Q(0,1)],
\label{tf:rtt:eq:state-effect}\\
\theta_{\mathrm{int}}&=\tfRttE[Q(1,1)-Q(0,1)-Q(1,0)+Q(0,0)].
\label{tf:rtt:eq:interaction}
\end{align}
The first measures mounting the true-feedback update. The second asks whether that effect exceeds the effect of a compute-matched sham update. Neither estimand is automatically positive.

\begin{tfRttTheorem}[Unbiased factorial estimator]
\label{tf:rtt:thm:factorial}
\tfRttProvedStatus{} Under Assumptions~\ref{tf:rtt:assump:intervention} and \ref{tf:rtt:assump:positivity}, independent randomized assignment of $(R,F)$ makes each observed cell mean $\widehat\mu_{rf}$ unbiased for $\tfRttE[Q(r,f)]$. Consequently,
\begin{equation}
\widehat\theta_{\mathrm{int}}
=\widehat\mu_{11}-\widehat\mu_{01}-\widehat\mu_{10}+\widehat\mu_{00}
\label{tf:rtt:eq:did-estimator}
\end{equation}
is unbiased for $\theta_{\mathrm{int}}$, and $\widehat\mu_{11}-\widehat\mu_{01}$ is unbiased for $\theta_{\mathrm{state}}$.
\end{tfRttTheorem}

\begin{proof}
By randomized assignment, $(R,F)$ is independent of the vector of potential outcomes and has positive cell probabilities. Consistency gives $Q_{\mathrm{obs}}=Q(R,F)$. Therefore
\(
\tfRttE[Q_{\mathrm{obs}}\mid R=r,F=f]=\tfRttE[Q(r,f)]
\)
for each cell. Taking the stated linear combinations and using linearity of expectation proves both claims.
\end{proof}

Randomization does not repair a wrong outcome definition, a broken reset, or interference across episodes. It only identifies the displayed potential-outcome contrasts under the declared protocol.

\begin{tfRttProtocol}[Minimum strict-RTT qualification]
\label{tf:rtt:prot:min-qual}
\tfRttPlannedStatus{} Before measuring task success, a future implementation should preregister: (i) immutable episode IDs; (ii) a fresh-proposal snapshot; (iii) policy, policy-read-closure, and non-policy state hashes; (iv) the deterministic event-key order and reset epoch; (v) a balanced $(R,F)$ assignment; (vi) common-random-number or sufficiently replicated distribution probes; (vii) the total-variation lower bound and uncertainty rule for W7; and (viii) a reset-negative control including deliberately late callbacks. Failure to pass W1--W8 stops the benefit claim.
\end{tfRttProtocol}

\subsection{Reset isolation}

\begin{tfRttDefinition}[Registered reset snapshot and completion]
\label{tf:rtt:def:reset-snapshot}
\tfRttDefinitionStatus{} A registered reset snapshot is
\begin{equation}
r^\star=(s^\star,n^\star,\kappa^\star,e^\star,H(s^\star)),
\label{tf:rtt:eq:reset-snapshot}
\end{equation}
where $s^\star$ is the complete event state, $n^\star$ and $\kappa^\star$ identify its event boundary, $e^\star$ is its reset epoch, and $H(s^\star)$ is a content hash over every carrier in Equation~\eqref{tf:rtt:eq:full-state}, external-service disposition, and pending-callback set. A reset operator $\mathcal D_{r^\star}$ is \emph{complete} only if it atomically restores $s^\star$, cancels or drains registered pending work, advances to a fresh epoch, and makes every return from an older epoch ineligible to mutate current state. Reset completion $q$ is the admitted event at which these conditions first hold. Reapplying $\mathcal D_{r^\star}$ before any new admitted event is idempotent up to the fresh-epoch receipt.
\end{tfRttDefinition}

\begin{tfRttTheorem}[Complete reset implies post-reset equality]
\label{tf:rtt:thm:reset}
\tfRttProvedStatus{} Let two arms complete registered resets at stopping events $q_a$ and $q_b$. Suppose their post-completion current states are exactly the same complete state $s^\star$ in Equation~\eqref{tf:rtt:eq:full-state}, including policy-read and auxiliary state, external-service disposition, random-generator state, budget, and an empty or identically restored pending-work set. Suppose stale-epoch returns cannot re-enter either arm, there is no cross-arm interference, and thereafter both arms use the same transition kernels under the deterministic event-order rule. Then every observable after reset completion has the same conditional distribution in the two arms.
\end{tfRttTheorem}

\begin{proof}
Re-index each arm locally from its reset-completion event. Conditional on the common state $s^\star$, the first admitted post-reset event has the same kernel. Stale-epoch exclusion prevents pre-reset asynchronous work from adding an unmodeled transition. Applying the same induction over successive transition kernels as in Lemma~\ref{tf:rtt:lem:no-read} gives equality of all finite-dimensional post-reset distributions. With common random numbers and a shared measurable transition realization, the same premises strengthen the conclusion to pathwise equality; without that coupling, the theorem asserts distributional equality only.
\end{proof}

\begin{tfRttCorollary}[Reset-negative control]
\label{tf:rtt:cor:reset-control}
\tfRttProvedStatus{} Under Theorem~\ref{tf:rtt:thm:reset}'s assumptions, a systematic post-reset arm difference falsifies at least one reset, kernel-equality, or no-interference assumption; it cannot be attributed to the erased policy update.
\end{tfRttCorollary}

The word ``complete'' is demanding. Files, retrieval indices, tool sessions, compiler caches, policy-read closure, scheduler state, numerical nondeterminism, optimizer moments, random streams, reset epoch, and queued or in-flight callbacks all require an explicit disposition. A byte-equal model file is not a complete reset certificate, and wall-clock simultaneity is not a reset boundary.

\subsection{Verifier uncertainty and reward circularity}

Let $S\in\{0,1\}$ denote true task success under a prespecified contract and $V\in\{0,1\}$ a verifier decision.

\begin{tfRttProposition}[Verifier alignment bound]
\label{tf:rtt:prop:verifier}
\tfRttProvedStatus{} For any policy arm $k$,
\begin{equation}
\left|\tfRttE[V_k]-\tfRttE[S_k]\right|\le \tfRttP(V_k\ne S_k).
\label{tf:rtt:eq:verifier-one}
\end{equation}
For two arms $a,b$ with error rates at most $\varepsilon_a,\varepsilon_b$,
\begin{equation}
\left|\{\tfRttE[V_a]-\tfRttE[V_b]\}-\{\tfRttE[S_a]-\tfRttE[S_b]\}\right|
\le \varepsilon_a+\varepsilon_b.
\label{tf:rtt:eq:verifier-two}
\end{equation}
\end{tfRttProposition}

\begin{proof}
Because $S,V$ are binary, $|V-S|=\tfRttOne\{V\ne S\}$. Thus
$|\tfRttE[V-S]|\le\tfRttE|V-S|=\tfRttP(V\ne S)$. Apply this once per arm and use the triangle inequality for the second claim.
\end{proof}

\begin{tfRttTheorem}[Sign non-identifiability without a verifier link]
\label{tf:rtt:thm:sign}
\tfRttProvedStatus{} Suppose only transcripts, arm assignments, and a proxy verifier $V$ are observed, while no assumption or audit links $V$ to latent success $S$. Then the sign of a measured arm effect on $S$ is not identified in general.
\end{tfRttTheorem}

\begin{proof}
Fix any observed joint law of transcripts, arms, and $V$. One compatible latent model sets $S=V$; another sets $S=1-V$. Both preserve every observed distribution. Their arm effects satisfy
\(
\tfRttE[S_a-S_b]=\tfRttE[V_a-V_b]
\)
and
\(
\tfRttE[S_a-S_b]=-\tfRttE[V_a-V_b]
\),
respectively. Unless the proxy effect is zero, the signs oppose. Therefore observed data alone do not identify the true-success direction.
\end{proof}

This impossibility is especially relevant when the same model family generates feedback, trains the update, and judges the final answer. Agreement inside that loop can increase while external correctness falls. A held-out executable contract, independent judge, or bounded verifier-error audit is not optional evidence decoration; it is part of the estimand.

%% file: theory_family/modules/rtt/sections/06_updates_objective.tex
\section{Update Operators, Budgets, Lifetimes, and Objectives}
\label{tf:rtt:sec:updates}

\subsection{RTT is not defined as gradient descent}

Let $G_j=(X,Z_{\le j},Y_{\le j},C_j,M_j,W_j,\Xi_j^{\mathrm{read}},\Xi_j^{\mathrm{aux}},B_j)$ be the causally available update record. A general bounded update of the declared carrier is
\begin{equation}
\Phi_{j+1}=\Pi_{\mathcal{K}(B_j)}\bigl(\mathcal{U}_{\Theta}(\Phi_j,G_j;R_j)\bigr),
\label{tf:rtt:eq:general-update}
\end{equation}
where $\mathcal{U}_{\Theta}$ is an update operator, $\Pi_{\mathcal{K}(B_j)}$ enforces a budget- and safety-dependent admissible set, and $\Theta$ is an outer-trained rule or fixed algorithm.  If the operator also changes a value read by the later action kernel, the registered target is $\Psi=(\Phi,\Xi^{\mathrm{read}})$ and the receipt must expose both components and their closure hashes; it is not a $\Phi$-only intervention. Table~\ref{tf:rtt:tab:update-taxonomy} lists nonexclusive realizations.

\begin{table}[H]
\centering
\small
\begin{tabularx}{\linewidth}{@{}p{0.18\linewidth}p{0.22\linewidth}Y Y@{}}
\toprule
Operator family & Mutable policy state & Within-problem update & Distinct audit risk \\
\midrule
Full/partial gradient & Selected weights and optimizer moments & One or more bounded gradient steps & Optimizer state, nondeterminism, support drift \\
Low-rank/adapters & Factor matrices or gates & Gradient, closed-form, or learned step & Mounted path may be gated off; rank/cost accounting \\
Learned optimizer & Optimizer recurrent state plus target parameters & Meta-learned map from gradients/feedback & Outer-training leakage and hidden lifetime \\
Recurrent/fast weights & Persistent policy-generating hidden state & State-transition rule from feedback & Confusion with ordinary token cache \\
Hypernetwork & Code or generator parameters & Feedback changes code/conditioning & Generated weights must be versioned and swapped \\
Bounded controller & Rules, logits, routing policy, or finite controller & Verified edit or constrained solver & Relabeling search workspace as policy \\
\bottomrule
\end{tabularx}
\caption{\textbf{Update-operator taxonomy (\tfRttDefinitionStatus{}).} An operator enters strict RTT only through the same W1--W8 witness; its mathematical form does not confer status. LoRA is one possible low-rank substrate \cite{lora2021}, not part of the definition.}
\label{tf:rtt:tab:update-taxonomy}
\end{table}

The projection in Equation~\eqref{tf:rtt:eq:general-update} can enforce norm, rank, sparsity, layer, latency, or memory limits. It does not itself guarantee useful learning. A learned optimizer may use an outer parameter $\Theta$ trained across tasks, while $\Phi$ is reset and adapted inside one episode. Their lifetimes and evidence must be separated: performance of $\Theta$ does not prove that a particular $\Phi_j\to\Phi_{j+1}$ transition---or a bundled $\Psi_j\to\Psi_{j+1}$ transition---was mounted and read.

\subsection{Within-problem and outer objectives}

For a fixed causal record at update time, a local update may target
\begin{equation}
J_j(\phi)
=\tfRttE\!\left[R_{j+1}-\lambda_c C_{j+1}-\lambda_h H_{j+1}
\tfRttGiven \tfRttCalF_{u_j},\operatorname{do}(\Phi_{j+1}=\phi)\right],
\label{tf:rtt:eq:local-objective}
\end{equation}
where $R_{j+1}$ is a prespecified retry utility, $C_{j+1}$ is charged cost, and $H_{j+1}$ is a harm variable. This conditional objective is not observable without assumptions or interventions; a proxy verifier defines a different objective unless its link to $R$ is audited.

An outer design parameter $\Theta$ should be evaluated over complete episodes:
\begin{align}
\mathcal{J}(\Theta)
=\tfRttE_{X\sim\mathcal{D}}\bigg[&U(X,Z_{1:J_\tau})
-\lambda_{\mathrm{tok}}C_{\mathrm{tok}}
-\lambda_{\mathrm{tool}}C_{\mathrm{tool}}
-\lambda_{\mathrm{upd}}C_{\mathrm{upd}}
\nonumber\\[-0.2em]
&-\lambda_{\mathrm{state}}C_{\mathrm{state}}
-\lambda_{\mathrm{harm}}H
-\lambda_{\mathrm{drift}}D(\Phi_{\tau\wedge J},\Phi_1)
\bigg],
\label{tf:rtt:eq:outer-objective}
\end{align}
subject to hard per-episode budget, lifetime, verifier, and safety constraints. $J_\tau$ is the number of attempts begun before resolution. Reporting only final accuracy removes the defining cost of within-problem learning and can make an arbitrarily expensive method look dominant.

\begin{tfRttDefinition}[Training-only realized-future teacher]
\tfRttDefinitionStatus{} An outer teacher may use completed future trajectories to estimate the counterfactual value of update actions during offline training. Any value, label, hyperparameter choice, or state version selected with post-$u_j$ information is \emph{training-only}. Deployment remains valid only if its committed update is $\tfRttCalF_{u_j}$-measurable. The future teacher is not a deployed RTT witness.
\end{tfRttDefinition}

\subsection{Policy-gradient identity and its assumptions}

Let a retry trajectory $Z$ have density $p_{\phi}(z\mid\mathcal{G})$ conditional on a pre-retry record $\mathcal{G}\subseteq\tfRttCalF_{u_j}$, and let $R(Z)$ be integrable.

\begin{tfRttProposition}[Conditional score-function identity]
\label{tf:rtt:prop:score}
\tfRttProvedStatus{} Suppose $p_{\phi}$ is differentiable, its support does not depend on $\phi$, differentiation and integration may be interchanged, and $b(\mathcal{G})$ is any integrable baseline independent of the sampled retry actions conditional on $\mathcal{G}$. Then
\begin{equation}
\nabla_{\phi}\tfRttE[R(Z)\mid\mathcal{G}]
=\tfRttE\!\left[(R(Z)-b(\mathcal{G}))\nabla_{\phi}\log p_{\phi}(Z\mid\mathcal{G})\mid\mathcal{G}\right].
\label{tf:rtt:eq:score}
\end{equation}
\end{tfRttProposition}

\begin{proof}
Using the stated interchange,
\(
\nabla_{\phi}\int R(z)p_{\phi}(z\mid\mathcal{G})dz
=\int R(z)p_{\phi}(z\mid\mathcal{G})\nabla_{\phi}\log p_{\phi}(z\mid\mathcal{G})dz.
\)
The baseline term has conditional expectation
\(
b(\mathcal{G})\int \nabla_{\phi}p_{\phi}(z\mid\mathcal{G})dz
=b(\mathcal{G})\nabla_{\phi}1=0.
\)
Subtracting it yields Equation~\eqref{tf:rtt:eq:score}.
\end{proof}

The proposition proves an identity, not low variance, verifier validity, safe optimization, or improvement after a finite step. If support changes under decoding truncation or tool constraints, the stated form may fail.

\begin{tfRttLemma}[Group-relative degeneracy]
\label{tf:rtt:lem:group-degeneracy}
\tfRttProvedStatus{} Consider a group update whose per-sample weight is
\begin{equation}
A_i=\frac{R_i-\overline R}{s_R+\varepsilon},\qquad \varepsilon>0.
\label{tf:rtt:eq:group-adv}
\end{equation}
If $R_1=\cdots=R_K$, then every $A_i=0$ and any update linear in the $A_i$ is exactly zero.
\end{tfRttLemma}

\begin{proof}
Equal rewards imply $R_i=\overline R$ for every $i$, so every numerator vanishes. A linear combination of score terms with zero coefficients is zero.
\end{proof}

This is a within-group credit result, not a statement that all group-relative algorithms always fail. It yields a concrete stop condition: if the verifier collapses to one value within update groups, more optimizer steps cannot recover a signal from Equation~\eqref{tf:rtt:eq:group-adv}.

\subsection{Budget and lifetime ledgers}

Let the remaining budget be a vector
\begin{equation}
B_j=(b^{\mathrm{tok}}_j,b^{\mathrm{tool}}_j,b^{\mathrm{sec}}_j,
b^{\mathrm{upd}}_j,b^{\mathrm{byte}}_j,b^{\mathrm{energy}}_j),
\label{tf:rtt:eq:budget-vector}
\end{equation}
where energy may be an explicitly labeled proxy if direct measurement is unavailable. Every attempt/update pair incurs a nonnegative cost vector $c_j$.

\begin{tfRttProposition}[Hard-ledger feasibility]
\label{tf:rtt:prop:budget}
\tfRttProvedStatus{} Suppose an operation is admitted only when $c_j\le B_j$ componentwise and then sets $B_{j+1}=B_j-c_j$. If $B_1\ge0$, every admitted prefix has $B_j\ge0$ and cumulative cost at most $B_1$.
\end{tfRttProposition}

\begin{proof}
Componentwise subtraction preserves nonnegativity by the admission rule. Telescoping gives $\sum_{i<j}c_i=B_1-B_j\le B_1$.
\end{proof}

\begin{table}[H]
\centering
\small
\begin{tabularx}{\linewidth}{@{}p{0.22\linewidth}Y Y@{}}
\toprule
Lifetime class & Required reset/transfer statement & Scientific interpretation \\
\midrule
Attempt-local & State expires before another attempt & Not a cross-attempt RTT carrier \\
Problem-local & State survives retries and expires at $\tau$ & Canonical strict RTT lifetime \\
Session-local & State may cross problem IDs & Mixture with online/continual adaptation \\
Persistent & State enters later sessions/releases & Continual/offline learning artifact; separate governance \\
\bottomrule
\end{tabularx}
\caption{\textbf{Policy-state lifetime classes (\tfRttDefinitionStatus{}).} Strict RTT concerns the in-problem update--reuse event; cross-problem retention is an additional mechanism that must not be hidden.}
\label{tf:rtt:tab:lifetime}
\end{table}

An episode-local adapter may still leak through optimizer caches, compilation artifacts, retrieval indices, or scheduler state after its nominal deletion. Lifetime certification therefore concerns the complete causal state, not only the primary tensor file.

%% file: theory_family/modules/rtt/sections/07_safety.tex
\section{Stability, Rollback, and Stop Conditions}
\label{tf:rtt:sec:safety}

\subsection{Local drift bounds are conditional, not global safety proofs}

For a fixed admissible history $h$, write $\pi_i(\cdot\mid h)$ for the fresh-proposal law after update $i$.

\begin{tfRttTheorem}[Cumulative local drift]
\label{tf:rtt:thm:drift}
\tfRttProvedStatus{} Suppose for $i=1,\ldots,K$ and a fixed history $h$,
\begin{equation}
\tfRttKL\!\left(\pi_i(\cdot\mid h)\,\|\,\pi_{i-1}(\cdot\mid h)\right)\le\kappa_i(h)<\infty.
\label{tf:rtt:eq:kl-steps}
\end{equation}
Then
\begin{equation}
\tfRttTV\!\left(\pi_K(\cdot\mid h),\pi_0(\cdot\mid h)\right)
\le \sum_{i=1}^{K}\sqrt{\frac{\kappa_i(h)}{2}}.
\label{tf:rtt:eq:tv-chain}
\end{equation}
For any measurable $g:\tfRttCalA\to[0,1]$,
\begin{equation}
\left|\tfRttE_{\pi_K}[g(A)\mid h]-\tfRttE_{\pi_0}[g(A)\mid h]\right|
\le \min\!\left\{1,\sum_{i=1}^{K}\sqrt{\frac{\kappa_i(h)}{2}}\right\}.
\label{tf:rtt:eq:harm-bound}
\end{equation}
\end{tfRttTheorem}

\begin{proof}
Pinsker's inequality bounds each adjacent total variation distance by $\sqrt{\kappa_i(h)/2}$. The triangle inequality for total variation gives Equation~\eqref{tf:rtt:eq:tv-chain}. The difference in expectations of a $[0,1]$ function is at most total variation; total variation is at most one, yielding Equation~\eqref{tf:rtt:eq:harm-bound}.
\end{proof}

The theorem is pointwise in $h$. A KL constraint measured on update trajectories does not automatically control unseen histories, tool calls, or external side effects. Nor does small local drift imply positive utility: a small harmful change is still harmful.

\begin{tfRttTheorem}[No nontrivial update is uniformly improving]
\label{tf:rtt:thm:no-free-update}
\tfRttProvedStatus{} Let $\pi$ and $\pi'$ be distinct probability measures over actions at some fixed history. There exists a bounded utility $g:\tfRttCalA\to[0,1]$ such that $\tfRttE_{\pi'}g<\tfRttE_{\pi}g$.
\end{tfRttTheorem}

\begin{proof}
Because $\pi'\ne\pi$, there is a measurable set $A$ on which their probabilities differ. If $\pi'(A)<\pi(A)$, choose $g=\tfRttOne_A$. Otherwise the complement satisfies $\pi'(A^c)<\pi(A^c)$, so choose $g=\tfRttOne_{A^c}$. In either case the expected utility decreases.
\end{proof}

This simple impossibility explains why a universal ``safe learning rate'' cannot follow from nontriviality alone. Safety requires a specified utility/harm family, coverage assumptions, conservative gates, and rollback; it cannot mean improvement for every possible downstream objective.

\subsection{Transactional update contract}

\begin{tfRttDefinition}[Rollback-capable policy-state transaction]
\label{tf:rtt:def:transaction}
\tfRttDefinitionStatus{} A policy update transaction contains: (i) an immutable pre-state snapshot; (ii) a candidate state and complete auxiliary state; (iii) typed validation receipts; (iv) canary probes on registered histories; (v) an atomic mount or NULL decision; (vi) post-mount monitoring; (vii) a bounded-time rollback path restoring the complete pre-state; and (viii) quarantine of rejected candidate artifacts. Partial mounts do not count as commits.
\end{tfRttDefinition}

\begin{table}[H]
\centering
\small
\begin{tabularx}{\linewidth}{@{}p{0.18\linewidth}Y Y@{}}
\toprule
Gate & Prespecified check & Mandatory action on failure \\
\midrule
Causal & Update inputs are $\tfRttCalF_{u_j}$-measurable & Quarantine; label leakage \\
Integrity & Hashes, shapes, optimizer state, route, and version agree & Atomic NULL; no retry mount \\
Budget & Worst-case update/retry cost fits remaining ledger & Skip update or stop episode \\
Functional & Fresh-proposal W7 lower bound clears threshold & Label inert; no RTT claim \\
Local stability & KL/norm/logit and tool-policy deltas under caps & Roll back candidate \\
Canary safety & No registered severe-harm event; harm CI below bound & Roll back and stop branch \\
Verifier & Error/abstention and group diversity within gate & Do not optimize proxy \\
Reset control & Negative-control arms exchangeable within tolerance & Stop causal attribution \\
\bottomrule
\end{tabularx}
\caption{\textbf{Transactional qualification and stop rules (\tfRttPrescriptionStatus{}).} Thresholds must be registered before seeing target-arm outcomes.}
\label{tf:rtt:tab:safety-gates}
\end{table}

\subsection{What rollback can and cannot undo}

Rollback can restore a declared digital state. It cannot automatically undo an external side effect already caused by a tool call, disclosure, transaction, or human decision. A safe episode therefore separates \emph{proposal actions} from \emph{irreversible actions}. Candidate policy versions may generate proposals in a sandbox; an independently governed executor decides whether to commit side effects. If the task necessarily includes irreversible actions, the harm estimand and authorization boundary must include them before the update is mounted.

\begin{tfRttDefinition}[Stop condition]
\tfRttDefinitionStatus{} A stop condition is a precommitted predicate on causally available audit variables whose truth forces one of: skip update, return NULL, roll back, abandon the episode, or escalate to external review. A stop condition is invalid if it is relaxed after observing an unfavorable target-arm result without recording a new protocol.
\end{tfRttDefinition}

Minimum stops include: future leakage; missing state/read receipts; zero functional divergence; collapsed within-group feedback; failed complete-reset control; verifier error exceeding its bound; safety-canary breach; budget overrun; nonfinite update; version mismatch; and inability to restore the full pre-state. An episode may stop even when the current candidate would have solved the problem; evidence discipline takes precedence over retrospective success.

\subsection{Stability across repeated within-problem updates}

Repeated updates create a feedback loop: the current policy determines which attempts are seen, those attempts determine feedback, and feedback changes the next policy. Three ledgers should therefore be distinct:
\begin{enumerate}
  \item \textbf{Mechanism ledger:} hashes, update inputs, W1--W8, and state-swap divergence for each transition;
  \item \textbf{Utility ledger:} true/proxy outcomes, verifier uncertainty, and prespecified episode utility;
  \item \textbf{Risk ledger:} local drift, canary harms, side effects, rollback latency, and residual state after expiry.
\end{enumerate}
A positive utility ledger cannot repair a missing mechanism witness, and a valid mechanism witness cannot waive a risk failure.

\begin{figure}[H]
\centering
\includegraphics[width=\linewidth]{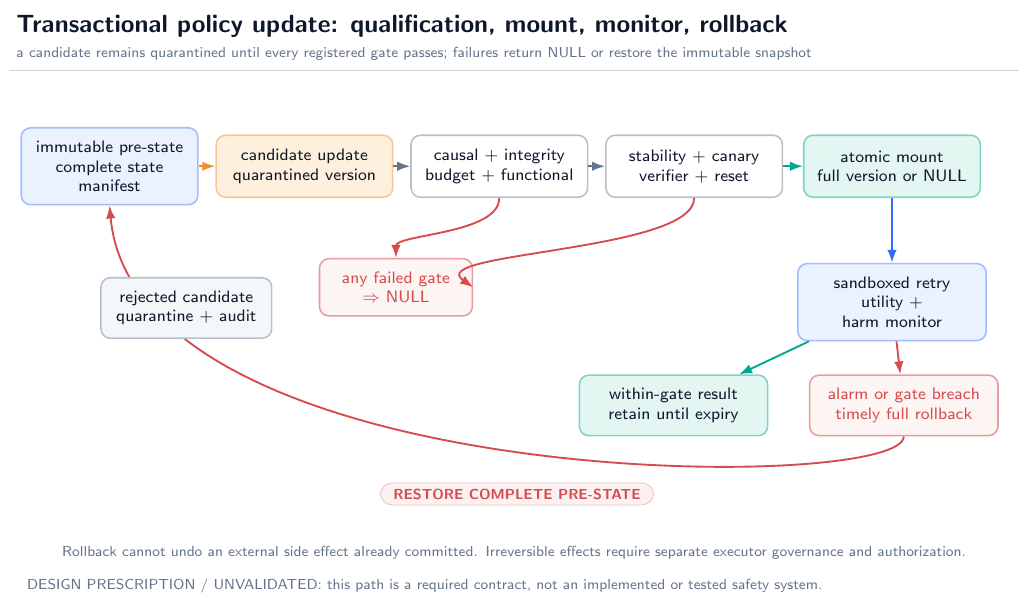}
\caption{\textbf{Safety, rollback, and stop-condition path (\tfRttPrescriptionStatus{}).} A candidate update remains quarantined until causal, integrity, budget, functional, stability, canary, verifier, and reset gates pass. Any bounded coral failure returns NULL or restores the immutable snapshot. Deployment of a candidate is atomic; the figure reports a contract, not an implemented system.}
\label{tf:rtt:fig:safety-path}
\end{figure}

%% file: theory_family/modules/rtt/sections/08_obligations.tex
\section{From Theorems to Implementation Obligations}
\label{tf:rtt:sec:obligations}

\subsection{Proof dependency map}

\begin{figure}[H]
\centering
\includegraphics[width=\linewidth]{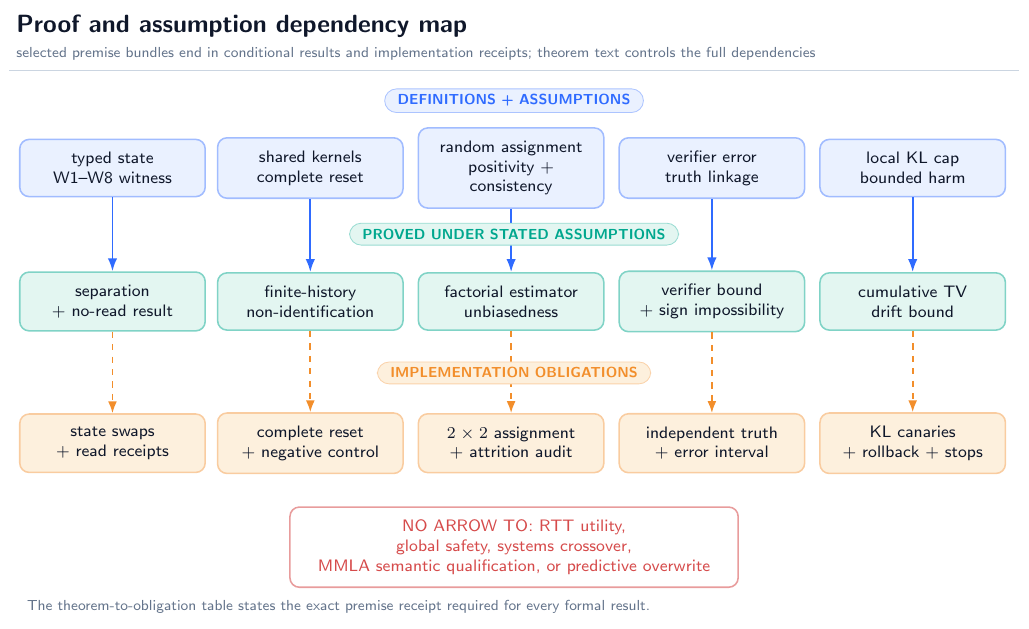}
\caption{\textbf{Assumption and proof dependency map (\tfRttProvedStatus{} for the drawn logical arrows).} Blue nodes are definitions/assumptions, teal nodes are proved statements, and amber nodes are design consequences. No arrow enters an empirical-success node because this theoretical part contains no RTT measurement. Appendix~\ref{tf:rtt:app:proofs} provides the expanded derivations.}
\label{tf:rtt:fig:proof-map}
\end{figure}

The formal claims are shallow by design: each conclusion depends on a small, inspectable assumption set. Figure~\ref{tf:rtt:fig:proof-map} and Table~\ref{tf:rtt:tab:obligation-map} prevent a common category error---implementing the conclusion's name while omitting the assumptions that make it true.

\begin{longtable}{@{}p{0.19\linewidth}p{0.19\linewidth}p{0.27\linewidth}p{0.20\linewidth}@{}}
\caption{\textbf{Theorem-to-implementation obligations.} Formal results become usable only when their premise receipts exist.}\label{tf:rtt:tab:obligation-map}\\
\toprule
Formal statement & Essential assumptions & Required implementation evidence & Does not establish \\
\midrule
\endfirsthead
\toprule
Formal statement & Essential assumptions & Required implementation evidence & Does not establish \\
\midrule
\endhead
Lemma~\ref{tf:rtt:lem:no-read} & Conditional independence of every later kernel from $\Phi$ at fixed $\Xi^{\mathrm{read}}$ & Typed route/read traces; readable-closure hashes; state-swap or route-disable probe & Update usefulness \\
Proposition~\ref{tf:rtt:prop:search-separation} & Stable typed state partition and byte-identical $\Psi$ & Fresh-proposal snapshot; readable-closure hash; workspace inventory & Search inferiority \\
Theorem~\ref{tf:rtt:thm:emulation} & Finite observable horizon; same environment kernels & Transcript schema and hidden-state inventory & Efficient fixed emulator \\
Theorem~\ref{tf:rtt:thm:factorial} & Randomization, positivity, consistency, no bundled carrier changes & Assignment receipt; arm counts; $\Phi$ and readable-closure hashes; restore hashes; attrition audit & Correct outcome/verifier \\
Theorem~\ref{tf:rtt:thm:reset} & Complete state equality, fresh epoch, deterministic event order, stale-return exclusion, and common kernels after completion & Registered snapshot/restore map/completion event; full-state manifest; pending-callback ledger; stale-return receipts; external-service reset; seed receipt & That any practical reset is complete \\
Proposition~\ref{tf:rtt:prop:verifier} & Binary success and measured verifier error & Independent adjudication sample and error CI & Small error off the audited support \\
Theorem~\ref{tf:rtt:thm:sign} & No constraint linking verifier to truth & Explicit absence/presence of external truth contract & That the proxy is actually adversarial \\
Proposition~\ref{tf:rtt:prop:score} & Differentiability, fixed support, integrability, legal interchange & Decoder support audit; finite-gradient checks & Low variance or finite-step gain \\
Lemma~\ref{tf:rtt:lem:group-degeneracy} & Centered group-relative weights; linearity in advantages & Per-group rewards and computed advantages & Failure under noncollapsed rewards \\
Proposition~\ref{tf:rtt:prop:budget} & Nonnegative charged costs; hard admission rule & Monotone signed resource ledger & Accuracy or efficiency dominance \\
Theorem~\ref{tf:rtt:thm:drift} & Per-history finite adjacent KL & Common-history log-probability probes & Global behavioral safety \\
Theorem~\ref{tf:rtt:thm:no-free-update} & Nonidentical action laws; arbitrary bounded utilities & Utility/harm family stated before update & Harm on the chosen real task \\
\bottomrule
\end{longtable}

\subsection{Minimum trace schema}

\begin{tfRttDefinition}[RTT trace bundle]
\label{tf:rtt:def:bundle}
\tfRttDefinitionStatus{} A qualification bundle contains one immutable manifest linking:
\begin{enumerate}
  \item episode/problem/version identifiers and complete problem contract;
  \item attempt start/end events, observations, actions, tool calls, deterministic keys $\kappa(e)$, wall-clock attributes, and randomization receipts;
  \item feedback bytes, producer identity, availability time, and truth/proxy status;
  \item pre/candidate/post $\Phi$ bytes, typed $\Xi^{\mathrm{read}}$ and $\Xi^{\mathrm{aux}}$ manifests, content-addressed references, and readable-closure hashes $\chi^-,\chi^+$;
  \item context, memory, workspace, external-service, and budget manifests at each restore boundary;
  \item update inputs, code/config hash, numerical mode, cost vector, candidate validation, commit/NULL decision;
  \item later policy-kernel read trace and fresh-proposal state-swap outputs, stating whether the intervention is on $\Phi$ or on $\Psi$;
  \item reset snapshot, complete restore-map receipt, completion event, old/new epoch, pending-callback disposition, and every rejected stale-epoch return;
  \item outcome, verifier, safety, rollback, expiry, and residual-state receipts.
\end{enumerate}
Every omission is a declared unknown, not an implicitly controlled variable.
\end{tfRttDefinition}

\subsection{Hidden-state and future-leakage audit}

The following channels require explicit disposition before causal attribution:
\begin{itemize}
  \item model weights, adapters, gates, normalization statistics, optimizer moments, activation/attention/KV caches, recurrent state, quantization scales, compiler graphs, and numerical kernels, classified explicitly as $\Phi$, $\Xi^{\mathrm{read}}$, or $\Xi^{\mathrm{aux}}$;
  \item prompts, system messages, hidden chain-of-thought carriers, retrieved documents, episodic/semantic memory, search trees, scratch files, databases, environment variables, tool sessions, and service-side caches;
  \item random-number generators at every library/device, scheduler order, batching neighbors, deterministic event keys, reset epochs, asynchronous callbacks and stale returns, and retry policies;
  \item verifier models, reference answers, unit tests, search or tool results, and any post-update outcome used to choose checkpoints or thresholds.
\end{itemize}

\begin{tfRttProtocol}[Leakage challenge]
\label{tf:rtt:prot:leakage}
\tfRttPlannedStatus{} Construct a canary field revealed strictly after update time and guaranteed irrelevant to the problem. Search every update input, serialized state, log, and selection decision for dependence on that field; additionally train a held-out decoder to predict the canary from the committed policy-state delta. Any above-chance result beyond the registered tolerance stops the causal claim until the channel is explained. A negative canary test is evidence about tested channels only.
\end{tfRttProtocol}

\subsection{Circularity audit}

Four loops are especially dangerous:
\begin{enumerate}
  \item \textbf{Self-verification:} the policy family produces answers and labels them;
  \item \textbf{Selection on target outcomes:} learning rates, checkpoints, or stop thresholds are chosen using the same episodes used for the claim;
  \item \textbf{Adaptive exclusion:} difficult failures are discarded after arm assignment;
  \item \textbf{Mechanism by outcome:} only successful retries are inspected for update receipts.
\end{enumerate}
The remedy is not to ban internal feedback. It is to define whether it is a proxy, bound or externally audit its error, freeze selection rules on a disjoint split, report intention-to-treat and attrition, and inspect mechanism receipts for every assigned episode. Appendix~\ref{tf:rtt:app:audits} gives a printable audit matrix.

\subsection{Reproducible state swaps}

A state swap must be semantically complete. For an adapter, this includes factor matrices, scaling, gates, routing tables, normalization state, optimizer-coupled forward state, dtype, and mount location. The pre- and post-state should be evaluated on the same registered histories under common random numbers where possible. For sampled text, a lower bound on total variation may be estimated from repeated draws with simultaneous confidence intervals, but a failure to reject equality is not proof of equality. Exact logit comparison is preferable when the action space and inference stack permit it.

%% file: theory_family/modules/rtt/sections/09_related_work.tex
\section{Relationship to Adjacent Work}
\label{tf:rtt:sec:related}

This section classifies \emph{described protocols}, not communities or papers as wholes. A paper may contain several mechanisms, and a future implementation may differ from its prose. Strict \tfRttName{} status belongs to a concrete trace satisfying Definition~\ref{tf:rtt:def:strict-rtt}.

\subsection{Test-time training and adaptation}

Classical test-time training updates model parameters on an auxiliary self-supervised objective derived from an unlabeled test sample before the main prediction \cite{tf:rtt:sun2020ttt}. Fully test-time adaptation such as Tent updates normalization parameters online by entropy minimization on test batches \cite{tf:rtt:wang2021tent}. These works establish the broader idea that parameters can change during evaluation. Their unit and timing need not match strict \tfRttName{}: an auxiliary update before a first task attempt, or an update accumulated across a stream and used on later samples, is test-time adaptation without necessarily containing a feedback--update--retry witness inside one still-unresolved reasoning problem.

TTRL performs reinforcement learning on unlabeled test data using consensus-derived rewards and online parameter updates \cite{tf:dual:zuo2025ttrl}. At the dataset level it is test-time reinforcement learning. Whether a particular item is strict \tfRttName{} depends on whether feedback from that item updates policy state before the item ends and a later attempt on that same item uses the update; dataset-level improvement cannot substitute for that trace.

\subsection{Inference-time scaling, search, and refinement}

Self-consistency samples multiple reasoning paths and aggregates answers \cite{tf:rtt:wang2023selfconsistency}. Tree of Thoughts explicitly searches over intermediate reasoning states with evaluation and backtracking \cite{tf:rtt:yao2023tot}. ReAct interleaves reasoning traces with actions and tool observations \cite{tf:rtt:yao2023react}. These mechanisms enlarge computation, workspace, or context and can substantially change outcomes while keeping the underlying proposal policy fixed. Under our state partition they are not strict \tfRttName{} unless an additional policy-state update passes the fresh-proposal test.

Self-Refine iteratively produces feedback and revised outputs without supervised training, additional training, or reinforcement learning \cite{tf:rtt:madaan2023selfrefine}. Reflexion explicitly reinforces agents through linguistic feedback and an episodic memory buffer rather than weight updates \cite{tf:rtt:shinn2023reflexion}. They are canonical reasons to preserve context/memory categories: trial-to-trial learning-like behavior need not be policy training. The label \tfRttMemoryAdaptation{} is useful for ordinary reasoning-time memory adaptation, but this paper makes no claim that it is the original name for such mechanisms.

\subsection{Within-instance policy evolution}

Policy of Thoughts describes within-instance online optimization: execution feedback drives GRPO updates to a transient LoRA adapter, which is reused during subsequent search and discarded after the problem \cite{tf:rtt:jiao2026pot}. StarOR describes node-level GRPO updates to a transient LoRA adapter maintained for the current optimization-modeling instance and reset for each new problem \cite{tf:rtt:li2026staror}. Their primary sources therefore describe candidate strict-\tfRttName{} architectures more closely than fixed-policy search. This paper does not revalidate their experiments or infer W1--W8 from descriptions alone. In particular, an independent strict audit would still need exact update times, unresolved status, actual mounted-state reads, fresh-proposal divergence, reset completeness, budgets, and verifier provenance.

\subsection{Continual learning, low-rank state, and memory}

Continual learning studies policy retention across sequences of tasks and the prevention of catastrophic forgetting; elastic weight consolidation is a representative example \cite{tf:rtt:kirkpatrick2017ewc}. Its primary reuse horizon is later tasks, whereas strict \tfRttName{} requires reuse before the current problem ends. The regimes can be nested: a strict within-problem adapter may be reset at each endpoint while an outer continual learner changes the initialization across episodes.

LoRA provides a parameter-efficient low-rank adaptation substrate \cite{lora2021}. A LoRA update may be offline, continual, test-time, or strict \tfRttName{} depending on timing and reuse; the matrix factorization alone determines none of those regimes.

The memory path in the integrated architecture formalizes bounded authoritative memory, causal event-conditioned writes, and training-only future valuation while reporting a current semantic-interface blocker. That evidence does not validate \tfRttName{}, and the present RTT theory does not validate predictive overwrite. A combined system requires independent policy-state and memory-state interventions as in Figure~\ref{tf:rtt:fig:policy-memory}.

\begin{table}[H]
\centering
\scriptsize
\begin{tabularx}{\linewidth}{@{}p{0.17\linewidth}p{0.20\linewidth}p{0.17\linewidth}Y@{}}
\toprule
Protocol family & Described changing carrier & Typical reuse unit & Trace-level RTT interpretation \\
\midrule
TTT / Tent \cite{tf:rtt:sun2020ttt,tf:rtt:wang2021tent} & Model/statistical parameters & Current sample, batch, stream & Adjacent; inspect whether same-problem feedback precedes an actual retry. \\
Self-consistency / ToT \cite{tf:rtt:wang2023selfconsistency,tf:rtt:yao2023tot} & Samples/search workspace & Same problem & Fixed-policy form is not strict RTT. \\
ReAct / Self-Refine \cite{tf:rtt:yao2023react,tf:rtt:madaan2023selfrefine} & Context, tool state, revision text & Same problem & Context/workspace adaptation unless policy state also changes. \\
Reflexion \cite{tf:rtt:shinn2023reflexion} & Episodic linguistic memory & Later trials & Memory adaptation; source explicitly contrasts with weight updates. \\
TTRL \cite{tf:dual:zuo2025ttrl} & Policy parameters & Test data stream & Per-item strict status requires an item-level update--reuse trace. \\
PoT \cite{tf:rtt:jiao2026pot} & Per-instance transient LoRA & Subsequent same-instance search & Candidate strict RTT; W1--W8 not independently audited here. \\
StarOR \cite{tf:rtt:li2026staror} & Per-instance transient LoRA & Later nodes/iterations on same instance & Candidate strict RTT; W1--W8 not independently audited here. \\
Continual learning \cite{tf:rtt:kirkpatrick2017ewc} & Persistent parameters & Later tasks & Different primary horizon; may contain nested RTT. \\
MMLA memory path & Bounded external/resident memory & Later tokens/segments & Memory adaptation, not RTT evidence. \\
\bottomrule
\end{tabularx}
\caption{\textbf{Related-work protocol map (source-verified, not a replication).} The final column applies this paper's operational boundary cautiously and makes no priority or empirical-ranking claim.}
\label{tf:rtt:tab:related-map}
\end{table}

The literature suggests a continuum rather than a winner: compute can be spent on broader exploration, richer nonparametric state, faster parametric adaptation, or combinations. The scientific task is to identify which carrier caused which effect at what cost, not to collapse all adaptation into one label.

%% file: theory_family/modules/rtt/sections/10_falsification.tex
\section{A Future Qualification and Falsification Program}
\label{tf:rtt:sec:falsification}

\subsection{Stage gates}

\begin{figure}[H]
\centering
\includegraphics[width=\linewidth]{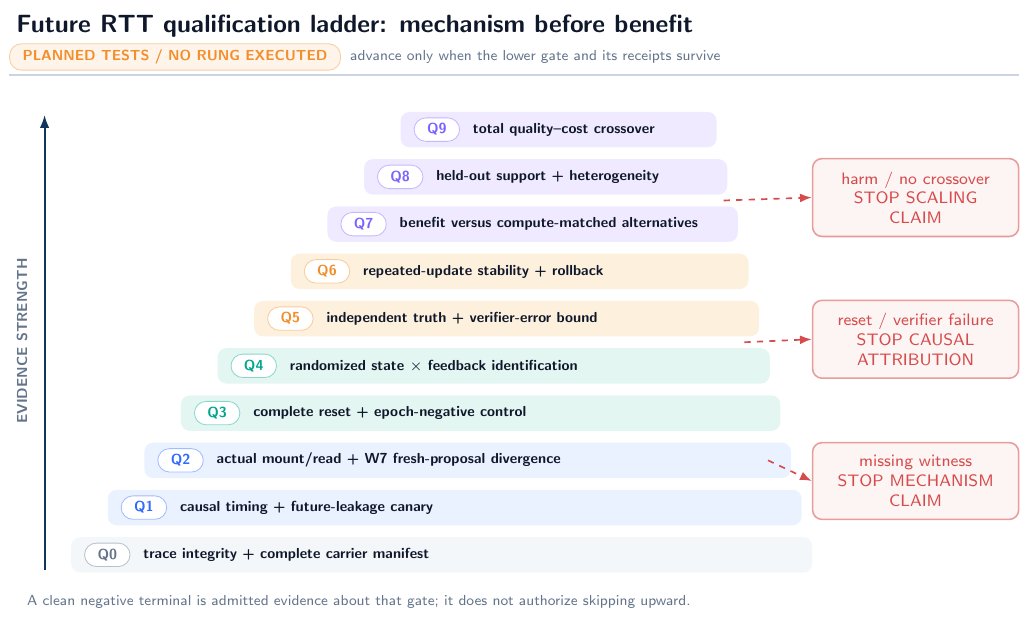}
\caption{\textbf{Future empirical qualification ladder (\tfRttPlannedStatus{}).} Evidence may advance only upward: integrity and timing precede functional state use; identification precedes benefit; safety precedes scale; equal-compute crossover is last. Coral side arrows are registered stops. No rung has been executed or passed in this RTT theoretical part.}
\label{tf:rtt:fig:empirical-ladder}
\end{figure}

The program begins with mechanism identity and stops on failure. It does not begin with benchmark accuracy. Figure~\ref{tf:rtt:fig:empirical-ladder} shows the dependency order; Table~\ref{tf:rtt:tab:planned-tests} makes each rung falsifiable.

The gates are sequential because later quantities are uninterpretable when earlier premises fail. A utility difference cannot establish policy-state use if the declared carrier was never read, and a carrier-swap contrast cannot isolate $\Phi$ if the policy-read closure, a tool cache, context suffix, random-number stream, or workspace also changed. When the policy-read closure cannot be held byte-identical, the registered contrast concerns the effective carrier $\Psi=(\Phi,\Xi^{\mathrm{read}})$ rather than $\Phi$ alone. Likewise, an apparently favorable verifier score does not identify true utility until the verifier--truth error relation is bounded on the relevant arms. Passing a gate therefore means only that its registered receipts survived their own audit. It neither repairs a failed lower gate nor supplies evidence for a higher one.

Each gate must specify its analysis population before execution. The intention-to-treat population contains every assigned episode, including update crashes, refusals, timeouts, rollback events, and missing receipts. A separately reported qualification population may restrict attention to episodes whose W1--W8 records are complete, but that restriction is descriptive unless the selection mechanism is itself identified. Attrition, verifier abstention, and failed mounts are outcomes, not preprocessing inconveniences. Negative controls and deliberately corrupted traces belong in the same ledger so that an auditor can estimate whether the qualification machinery rejects mechanisms it was designed to reject.

Thresholds, budgets, and stop decisions must be frozen before outcome inspection: the minimum W7 divergence, verifier-error tolerance, local drift cap, severe-harm definition, equal-compute ledger, and minimum worthwhile utility gain. A negative terminal is a valid result at its gate. It should preserve the failed receipt, assigned arm, and available downstream measurements while withholding the claim that depended on the failure. This convention separates scientific falsification from system promotion: an implementation can be informative even when it never becomes eligible for scaling.

\begin{longtable}{@{}p{0.09\linewidth}p{0.22\linewidth}p{0.34\linewidth}p{0.27\linewidth}@{}}
\caption{\textbf{Planned qualification tests and stop rules (\tfRttPlannedStatus{}).}}\label{tf:rtt:tab:planned-tests}\\
\toprule
Gate & Question & Minimal protocol & Stop condition \\
\midrule
\endfirsthead
\toprule
Gate & Question & Minimal protocol & Stop condition \\
\midrule
\endhead
Q0 & Is the trace complete? & Hash code/config/state; immutable episode IDs; typed carrier, policy-read-closure, callback, and cost inventories. & Missing, inconsistent, or unbounded receipt. \\
Q1 & Did timing satisfy W1--W5? & Audit deterministic event keys, reset epochs, callback admissions, filtration, and stopping times; run a future canary. & Update after resolution, ambiguous order, stale return admitted, or detectable future dependence. \\
Q2 & Was post-state actually used? & Mount/read trace with closure hashes plus an exact or replicated fresh-proposal swap; label the target $\Phi$ or $\Psi$. & W6 absent, closure target ambiguous, or W7 lower bound $\le0$. \\
Q3 & Are other carriers isolated? & Restore a registered complete snapshot, complete the restore event, enter a fresh epoch, reject stale callbacks, and run separate carrier swaps. & Reset receipt incomplete or reset-negative control differs. \\
Q4 & Is feedback semantic? & Randomized $2\times2$ state/feedback factorial; intention-to-treat analysis. & Assignment, positivity, attrition, or sham-compute failure. \\
Q5 & Is the verifier linked to truth? & Independent adjudication, abstention and error bounds, adversarial cases. & Error bound too wide for claimed effect or group signal collapse. \\
Q6 & Is repeated updating stable? & Registered KL/norm/canary caps; atomic rollback; residual-state audit. & Severe harm, nonfinite update, or rollback failure. \\
Q7 & Does it help under matched compute? & Compare fixed-policy search, context, memory, sham update, and RTT under equal ledgers. & Confidence interval excludes the prespecified minimum gain in the wrong direction. \\
Q8 & Does it transfer within declared support? & Held-out problem families, verifiers, seeds, and update counts; report heterogeneity. & Benefit confined to tuned items or hidden carryover. \\
Q9 & Is total quality--cost favorable? & End-to-end wall time, tokens, tool calls, update compute, retained bytes, harm and failure cost. & No registered crossover or budget breach. \\
\bottomrule
\end{longtable}

\subsection{White-box mechanism benchmark}

\begin{tfRttProtocol}[Within-problem policy mechanism falsifier]
\label{tf:rtt:prot:m0}
\tfRttPlannedStatus{} Construct a synthetic family with a hidden per-instance rule that cannot be identified from the initial prompt alone. Attempt 1 reveals a deterministic verifier bit about one proposed rule; an update may change only a bounded registered carrier; attempt 2 begins only after a registered complete restore has completed, a fresh reset epoch is active, and stale callbacks from the prior epoch are ineligible to mutate current state. Include (a) true update, (b) pre-state restore, (c) sham-feedback update, and (d) search-only workspace arms under one deterministic event-order rule. The primary endpoint is W7 fresh-proposal divergence in the registered direction, not final accuracy. Call the endpoint a $\Phi$ effect only when $\Xi^{\mathrm{read}}$ and all non-policy carriers are byte-identical across the swap; otherwise preregister the effective target $\Psi=(\Phi,\Xi^{\mathrm{read}})$ and hold every carrier outside $\Psi$ fixed. Closure hashes, reset-completion receipts, epoch transitions, and rejected stale returns are mandatory.
\end{tfRttProtocol}

The benchmark should contain negative controls where the feedback is uninformative, the update route is gated off, the adapter is byte-different but functionally null, and the apparent rule is recoverable from a hidden cache. These cases test the auditor rather than the learner.

\subsection{Executable reasoning benchmark}

\begin{tfRttProtocol}[RTT-M1 executable retry]
\label{tf:rtt:prot:m1}
\tfRttPlannedStatus{} Use generated programs with isolated deterministic unit tests unavailable to the policy except through a bounded first-attempt feedback record. Freeze the task generator and hidden tests before method development. Preregister one retry, one bounded update, a hashed complete-state snapshot, the restore map and completion event, a fresh reset epoch with stale-return rejection, the $2\times2$ factorial, verifier-error spot checks, update/cost caps, and a severe-side-effect sandbox. Report every assigned episode and distinguish syntax, hidden-test truth, $\Phi$-only or bundled-$\Psi$ divergence, reset qualification, and final utility.
\end{tfRttProtocol}

Compiler feedback is not automatically ground truth: passing public tests can overfit, and test-generation bugs can reward incorrect behavior. Hidden-test construction, contamination, and false-pass/false-fail audits remain part of Q5.

\subsection{Repeated-update stress test}

\begin{tfRttProtocol}[RTT-M2 repeated-update stability]
\label{tf:rtt:prot:m2}
\tfRttPlannedStatus{} On episodes requiring up to $K$ retries, register a per-update KL cap $\kappa_i$, cumulative bound from Theorem~\ref{tf:rtt:thm:drift}, independent anchor histories, tool-action canaries, rollback latency, and a hard maximum update count. Randomize rollback challenges by injecting known-bad candidate states. Stop the full arm if any irreversible side effect, unbounded ledger, or failed complete-state restoration occurs.
\end{tfRttProtocol}

The purpose is not to prove global safety. It is to falsify specific local contracts under adversarial but declared conditions and to measure how often the theoretical premises hold.

\subsection{Equal-compute comparisons}

The final comparison should allocate the same multidimensional budget, not merely the same number of sampled answers. Search may use its budget for more rollouts; context refinement for longer critique/revision; memory adaptation for writes and retrieval; RTT for update, state storage, probes, and retry. Wall-clock parallelism, accelerator occupancy, tool latency, failure recovery, and verifier calls must be included. A method that wins only after omitting update or safety costs has not established a quality--cost crossover.

Primary reporting should include:
\begin{itemize}
  \item intention-to-treat and per-protocol estimates with arm counts and attrition;
  \item $\theta_{\mathrm{state}}$, $\theta_{\mathrm{int}}$, W7 divergence, verifier-error bounds, and true utility as distinct endpoints;
  \item distributions over problem families and seeds, not only pooled means;
  \item compute, time, retained-state, and side-effect ledgers with uncertainty;
  \item all preregistered stops, including negative and null terminals.
\end{itemize}

No successful result is predicted by this ladder. A clean failure at Q2 or Q4 would be scientifically useful: it would show that an implementation marketed as adaptive did not create or use the claimed policy-state path.

%% file: theory_family/modules/rtt/sections/11_discussion.tex
\section{Discussion, Limitations, and Conclusion}
\label{tf:rtt:sec:discussion}

\subsection{Why adopt a strict definition?}

A broad definition in which any feedback-conditioned retry counts as training would be easy to satisfy but difficult to falsify. It would merge sampling, search, prompt revision, memory, tool state, and policy adaptation even though they have different costs, safety properties, and scientific controls. The strict definition instead reserves one term for a specific causal pattern while leaving neighboring mechanisms first-class. A hybrid system can still be studied; it simply cannot attribute a bundled effect to policy learning without state-specific interventions.

The boundary is partly conventional. One could define ``policy'' to include the entire search tree or prompt, making every stateful computation a policy update. Definition~\ref{tf:rtt:def:fresh-proposal} rejects that collapse for this paper by holding non-policy carriers fixed and testing fresh proposal laws. This choice is useful because it yields operational separation, but it is not a theorem that every community must adopt the same vocabulary.

\subsection{Limitations of the framework}

First, exact state interventions may be infeasible in opaque hosted systems. There, strict qualification may remain unidentifiable; behavioral improvement should be reported without a mechanism claim. Second, total variation at token-level histories can be expensive to estimate in large action spaces. Registered lower bounds or exact logit probes cover only tested histories. Third, the deterministic event-key rule presumes an auditable ingress sequencer and frozen precedence classes. Distributed systems need an implementation-specific total-order extension, epoch discipline, and receipts proving that concurrent or stale returns did not bypass it; wall-clock timestamps alone are insufficient. Fourth, potential-outcome consistency may fail when shared services create interference across arms. Fifth, a binary resolution time simplifies tasks with graded or revisable success. Sixth, the framework treats policy and memory as typed interfaces, but architectures with fully entangled fast state may resist clean partitioning. Such systems should report an inseparable bundled carrier rather than force a false decomposition.

The mathematical results are deliberately modest. The finite-history emulator may be exponentially large. The factorial theorem does not solve external validity. The verifier bound is only as useful as the audited error rate. The drift bound is local and can become vacuous after many updates. The no-uniform-improvement theorem does not imply that safe gains are impossible on a restricted task distribution. None of these results predicts that a practical optimizer will learn within an episode.

\subsection{Boundary with MMLA evidence}

This formal part shares the integrated report's authorship and scientific scope, not an inherited empirical success claim. The evidence now includes historical semantic-interface failures, later restricted readout successes, and a broader continuous-event latent failure. No such \tfRttMMLA{} result identifies a policy update reused during the same unresolved problem. Conversely, these definitions do not repair a memory interface. A future combined system needs four independently answered questions: did policy state change, did memory state change, did each state affect the appropriate fresh-proposal or retrieval path, and did either cause verified benefit under matched cost?

\subsection{Conclusion}

\tfRttName{} is most useful as a falsifiable trace property: feedback from an attempt on one unresolved problem changes a declared functional policy state, the change commits before resolution, and a later attempt on that same problem actually uses it. This temporal and typed definition separates policy learning from spending more inference compute or carrying more nonparametric state. The accompanying theorems show why logs and proxy success are insufficient, while the qualification ladder states what future evidence would have to survive.

The paper's strongest conclusion is therefore negative in form: neither a method name, an optimizer call, a different retry, nor a proof can establish strict \tfRttName{} benefit. Establishing it requires state identity, causal timing, functional reuse, interventions, verifier alignment, budget accounting, and rollback evidence together. Those obligations are demanding by design. If future systems pass them, ``learning before a single problem ends'' will denote an identified mechanism rather than a metaphor.

%% file: theory_family/modules/rtt/appendices/A_proofs.tex
\section{Expanded Proofs}
\label{tf:rtt:app:proofs}

This appendix expands every numbered formal result. All statements retain the status \tfRttProvedStatus{}. The proofs concern the abstract model only and create no implementation or empirical evidence.

\subsection{Kernel notation}

Let $E_n$ denote the $n$th complete event in the deterministic event log, ordered lexicographically by the registered key $\kappa(E_n)=(g_n,p_n,\mathrm{id}_n)$, and let $H_n=(E_0,\ldots,E_n)$ be its observable history. Wall-clock timestamps remain recorded attributes but do not determine causal order. Conditional on a complete state and intervention arm, write $K_{n+1}(de\mid h_n)$ for the next-event transition kernel. For a finite horizon $N$, the joint law factors as
\begin{equation}
P(de_{0:N})=P_0(de_0)\prod_{n=0}^{N-1}K_{n+1}(de_{n+1}\mid e_{0:n}).
\label{tf:rtt:eq:kernel-factorization}
\end{equation}
All induction arguments below are applications of this factorization.

\subsection{No-read and separation}

\paragraph{Lemma~\ref{tf:rtt:lem:no-read}.}
Condition on the non-policy state
$N_{u_j}:=(X,C,M,W,\Xi^{\mathrm{read}},\Xi^{\mathrm{aux}},B)$ at update completion. The declared policy-only intervention is meaningful only when the bytes of the policy-readable closure $\Xi^{\mathrm{read}}$ are identical; otherwise the intervened functional carrier is the bundle $\Psi=(\Phi,\Xi^{\mathrm{read}})$. Let $P^{\phi}$ and $P^{\phi'}$ be the post-update laws under two mounted policy states with the same readable closure. By hypothesis, for every reachable history and each later event,
\begin{equation}
K^{\phi}_{n+1}(\cdot\mid h_n,N_{u_j})
=K^{\phi'}_{n+1}(\cdot\mid h_n,N_{u_j}).
\end{equation}
The initial conditional distributions at $u_j$ are identical because only $\Phi$ is intervened on and the later kernels do not read it. Suppose the laws of $H_n$ agree. For any measurable set $D$ of histories through $n{+}1$,
\begin{align}
P^{\phi}(H_{n+1}\in D\mid N_{u_j})
&=\int \tfRttOne\{(h_n,e)\in D\}K^{\phi}_{n+1}(de\mid h_n)P^{\phi}(dh_n)\\
&=\int \tfRttOne\{(h_n,e)\in D\}K^{\phi'}_{n+1}(de\mid h_n)P^{\phi'}(dh_n)\\
&=P^{\phi'}(H_{n+1}\in D\mid N_{u_j}).
\end{align}
Induction reaches the attempt endpoint and any outcome measurable there. Integrating over $N_{u_j}$ gives the unconditional version if its distribution is shared.

\paragraph{Propositions~\ref{tf:rtt:prop:search-separation} and \ref{tf:rtt:prop:memory-separation}; Corollary~\ref{tf:rtt:cor:compute}.}
Each is a direct application of the conjunction in Definition~\ref{tf:rtt:def:strict-rtt}. Strict \tfRttName{} requires W3, a nontrivial declared functional-carrier transition, and W7, a fresh-proposal difference after non-policy carriers are fixed. A process with constant $\Phi$ and byte-identical $\Xi^{\mathrm{read}}$ cannot satisfy W3; a change only to $C$, $M$, $W$, or $\Xi^{\mathrm{aux}}$ remains fixed away in W7. If the policy-readable closure changes, the admissible conclusion concerns $\Psi$, not $\Phi$ alone. Increasing compute changes neither logical fact. These are necessary-condition proofs; they make no claim about utility.

\subsection{Finite-history emulation and non-identifiability}

\paragraph{Theorem~\ref{tf:rtt:thm:emulation}.}
Let $\mathcal{H}_n$ be the finite set of observable histories through decision $n$ with positive probability under the adaptive mechanism $P_{\mathrm{ad}}$. Hidden state, including $\Phi_n$, may be stochastic conditional on $H_n$. Define the fixed behavioral kernel
\begin{equation}
q_n(da\mid h_n)
=\int \pi_{\phi}(da\mid h_n)P_{\mathrm{ad}}(d\phi\mid H_n=h_n).
\label{tf:rtt:eq:marginal-policy}
\end{equation}
This equals $P_{\mathrm{ad}}(A_n\in da\mid H_n=h_n)$ by the law of total probability. The collection $q=(q_1,\ldots,q_N)$ is fixed before emulation; it does not modify an internal parameter in response to the realized episode. On zero-probability histories choose any probability kernel.

At decision 1, $q_1$ matches the adaptive action conditional. The shared environment then matches the next observation conditional. Suppose the observable joint law through decision $n$ matches. Conditioning on any positive-probability $h_n$, Equation~\eqref{tf:rtt:eq:marginal-policy} matches the next action, and the environment matches the next observation. Multiplying these conditionals by the matched history probability establishes the induction step. Thus all observable finite-dimensional distributions match. Corollary~\ref{tf:rtt:cor:nonident} follows because one compatible mechanism contains mutable policy state and another does not.

For a countable but almost surely finite stopping horizon, the construction applies to every finite prefix consistently. We state the finite version in the main text to avoid imposing additional extension conditions that are irrelevant to finite logged episodes.

\subsection{Randomized factorial identification}

\paragraph{Theorem~\ref{tf:rtt:thm:factorial}.}
Let $Z_{rf}=\tfRttOne\{R=r,F=f\}$ and $p_{rf}=\tfRttP(Z_{rf}=1)>0$. A Horvitz--Thompson version of the cell mean is
\begin{equation}
\widetilde\mu_{rf}=\frac{1}{N}\sum_{i=1}^N\frac{Z_{i,rf}Q_i^{\mathrm{obs}}}{p_{rf}}.
\end{equation}
By consistency, $Z_{i,rf}Q_i^{\mathrm{obs}}=Z_{i,rf}Q_i(r,f)$. Randomization makes $Z_{i,rf}$ independent of $Q_i(r,f)$, hence
\begin{equation}
\tfRttE\!\left[\frac{Z_{i,rf}Q_i^{\mathrm{obs}}}{p_{rf}}\right]
=\frac{\tfRttE Z_{i,rf}\,\tfRttE Q_i(r,f)}{p_{rf}}
=\tfRttE Q_i(r,f).
\end{equation}
Averaging proves unbiasedness. Under fixed balanced cell counts, the ordinary within-cell mean has the same conditional expectation. Linearity proves unbiasedness of both contrasts in Equations~\eqref{tf:rtt:eq:state-effect}--\eqref{tf:rtt:eq:did-estimator}. If outcomes are missing after assignment, the result applies to intention-to-treat only when missingness is included in the outcome contract or handled under additional stated assumptions.

\subsection{Reset isolation}

\paragraph{Theorem~\ref{tf:rtt:thm:reset} and Corollary~\ref{tf:rtt:cor:reset-control}.}
Let the registered snapshot be
$r^\star=(s^\star,n^\star,\kappa^\star,e^\star,H(s^\star))$, and let the two reset operations complete at events $q_a$ and $q_b$.  Completeness of the registered restore map $D$ gives byte-identical post-completion states after reindexing: model and runtime state, $\Phi$, both components of $\Xi$, context, memory, workspace, RNG, budget, external-service handles, and the pending-callback ledger all equal $s^\star$.  Each arm enters the same fresh reset epoch, and any callback carrying an earlier epoch is rejected and recorded rather than applied.  By the shared-kernel premise and the deterministic event-order rule, the first events after $q_a$ and $q_b$ therefore have the same law $K_1(\cdot\mid s^\star)$.  If their reindexed history distributions agree through event $n$, integrating the same $K_{n+1}$ against the same history law in Equation~\eqref{tf:rtt:eq:kernel-factorization} matches event $n{+}1$.  Induction establishes distributional equality for every post-completion observable.  With common random numbers and deterministic kernels, the same argument strengthens to pathwise equality.  A detected systematic difference contradicts the conjunction of complete restoration, epoch isolation, shared kernels, deterministic ordering, and no interference; it does not by itself identify which reset premise failed, nor does a null difference certify reset completeness.

\subsection{Verifier results}

\paragraph{Proposition~\ref{tf:rtt:prop:verifier}.}
For binary variables, $|V-S|$ is exactly the disagreement indicator. Jensen's inequality for absolute value gives
\begin{equation}
|\tfRttE V-\tfRttE S|=|\tfRttE(V-S)|\le\tfRttE|V-S|=\tfRttP(V\ne S).
\end{equation}
For arms $a,b$, expand the difference between proxy and truth effects and apply the triangle inequality:
\begin{align}
&|\tfRttE(V_a-S_a)-\tfRttE(V_b-S_b)|\\
&\qquad\le |\tfRttE(V_a-S_a)|+|\tfRttE(V_b-S_b)|
\le\varepsilon_a+\varepsilon_b.
\end{align}

\paragraph{Theorem~\ref{tf:rtt:thm:sign}.}
Let $P_{\mathrm{obs}}$ be the complete observed law. Extend it to model $P^+$ by defining latent $S:=V$ deterministically. Extend the same $P_{\mathrm{obs}}$ to $P^-$ by defining $S:=1-V$. Both extensions integrate back to the identical observed law. For any two arms,
\begin{align}
\Delta_S^+&=\tfRttE[V\mid a]-\tfRttE[V\mid b],\\
\Delta_S^-&=\tfRttE[1-V\mid a]-\tfRttE[1-V\mid b]=-\Delta_S^+.
\end{align}
Thus the observed law admits opposing truth-effect signs whenever the proxy effect is nonzero. A calibration, monotonicity, bounded-error, or external-truth assumption would rule out some extensions, but none is assumed in the theorem.

\subsection{Optimization identities}

\paragraph{Proposition~\ref{tf:rtt:prop:score}.}
Write $J(\phi)=\int R(z)p_\phi(z\mid\mathcal G)\,\nu(dz)$. The domination/interchange assumption gives
\begin{align}
\nabla J(\phi)
&=\int R(z)\nabla p_\phi(z\mid\mathcal G)\,\nu(dz)\\
&=\int R(z)p_\phi(z\mid\mathcal G)\nabla\log p_\phi(z\mid\mathcal G)\,\nu(dz).
\end{align}
Moreover,
\begin{equation}
\int b(\mathcal G)p_\phi(z\mid\mathcal G)\nabla\log p_\phi(z\mid\mathcal G)\,\nu(dz)
=b(\mathcal G)\nabla\int p_\phi\,d\nu=0.
\end{equation}
Subtracting this zero term proves the identity. If the action support depends on $\phi$, boundary terms may appear; that case is outside the proposition.

\paragraph{Lemma~\ref{tf:rtt:lem:group-degeneracy}.}
If $R_i=r$ for all $i$, then $\overline R=r$, $s_R=0$, and $A_i=(r-r)/\varepsilon=0$. Any update $\sum_iA_i g_i$ or scalar multiple thereof is zero. Nonlinear algorithms that add terms not weighted by $A_i$ are not covered and must audit those terms separately.

\subsection{Budget, drift, and non-uniform safety}

\paragraph{Proposition~\ref{tf:rtt:prop:budget}.}
Induct on $j$. The base $B_1\ge0$ is assumed. If $B_j\ge0$ and admission requires $0\le c_j\le B_j$, then $B_{j+1}=B_j-c_j\ge0$. Telescoping componentwise yields
\begin{equation}
\sum_{i=1}^{j}c_i=B_1-B_{j+1}\le B_1.
\end{equation}

\paragraph{Theorem~\ref{tf:rtt:thm:drift}.}
For each $i$, Pinsker's inequality gives
\begin{equation}
\tfRttTV(\pi_i,\pi_{i-1})\le\sqrt{\tfrac12\tfRttKL(\pi_i\|\pi_{i-1})}
\le\sqrt{\kappa_i/2}.
\end{equation}
The triangle inequality for the total variation metric gives
\begin{equation}
\tfRttTV(\pi_K,\pi_0)
\le\sum_{i=1}^{K}\tfRttTV(\pi_i,\pi_{i-1})
\le\sum_{i=1}^{K}\sqrt{\kappa_i/2}.
\end{equation}
By the variational characterization of total variation, for any $g\in[0,1]$,
$|\tfRttE_{\pi_K}g-\tfRttE_{\pi_0}g|\le\tfRttTV(\pi_K,\pi_0)$. Total variation is at most one.

\paragraph{Theorem~\ref{tf:rtt:thm:no-free-update}.}
Distinct probability measures differ on at least one measurable set $A$. If $\pi'(A)<\pi(A)$, $g=\tfRttOne_A$ has lower expectation under $\pi'$. If $\pi'(A)>\pi(A)$, then $\pi'(A^c)=1-\pi'(A)<1-\pi(A)=\pi(A^c)$ and $g=\tfRttOne_{A^c}$ works. The constructed utility may not be relevant to a given task; the theorem rules out uniform improvement over the unrestricted bounded-utility class only.

%% file: theory_family/modules/rtt/appendices/B_symbols.tex
\section{Symbol and State Dictionary}
\label{tf:rtt:app:symbols}

\begin{longtable}{@{}p{0.18\linewidth}p{0.26\linewidth}p{0.49\linewidth}@{}}
\caption{\textbf{Canonical symbols (\tfRttDefinitionStatus{}).} Symbols are local to this paper unless explicitly shared across the family.}\label{tf:rtt:tab:symbols}\\
\toprule
Symbol & Type & Meaning \\
\midrule
\endfirsthead
\toprule
Symbol & Type & Meaning \\
\midrule
\endhead
$X,I$ & $\tfRttCalX\times\mathcal I$ & Immutable problem and episode identifier. \\
$j,t,n$ & $\mathbb N$ & Attempt, within-attempt microstep, and deterministic global event indices; wall-clock time is a recorded attribute, not the causal order. \\
$\kappa(e)=(g,p,\mathrm{id})$ & ordered key & Registered generation, phase priority, and unique identifier used to order complete events lexicographically. \\
$Z_j$ & $\tfRttCalZ$ & Complete marked trajectory for attempt $j$. \\
$O_{j,t},A_{j,t}$ & $\tfRttCalO,\tfRttCalA$ & Observation available before the next action, and action. \\
$Y_j$ & $\tfRttCalY$ & Feedback generated from attempt $j$, with provenance/time. \\
$C_j$ & $\tfRttCalC$ & Prompt and transient context state. \\
$M_j$ & $\tfRttCalM$ & External/resident memory state. \\
$W_j$ & $\tfRttCalW$ & Search tree, files, scratch state, tool/cache workspace. \\
$\Phi_j$ & $\tfRttCalP$ & Declared functional policy state mounted for attempt $j$. \\
$\Xi_j^{\mathrm{read}}$ & bounded typed product & Policy-readable auxiliary closure: every registered route, gate, normalization statistic, adapter selector, or other non-$\Phi$ value read by the action kernel. \\
$\Xi_j^{\mathrm{aux}}$ & bounded typed product & Non-readable auxiliary state: optimizer moments, bookkeeping, RNG state, schedules, and other registered carriers not read by the current action kernel. \\
$\Psi_j=(\Phi_j,\Xi_j^{\mathrm{read}})$ & product space & Effective policy carrier.  A claim about $\Phi$ alone is licensed only when the readable closure is byte-identical across the compared arms. \\
$B_j$ & $\tfRttCalB\subseteq\mathbb R_+^d$ & Remaining multidimensional resource budget. \\
$\mathsf S_n$ & product space & Complete state after deterministic event $n$, including reset epoch and pending-callback ledger. \\
$S_j=\mathsf S_{\sigma_j}$ & product space & Complete pre-attempt state in Equation~\eqref{tf:rtt:eq:full-state}. \\
$\pi_\phi$ & stochastic kernel & Action/proposal law parameterized by policy state $\phi$. \\
$U_j,\mathcal U_\Theta$ & measurable maps & Concrete update and general outer-parameterized update rule. \\
$\rho_j$ & receipt space & Update inputs/outputs, hashes, deterministic event keys, times, costs, validation and commit. \\
$\chi_j^{-},\chi_j^{+}$ & hash strings & Pre-update and post-update hashes of the declared policy-readable closure. \\
$\tfRttCalF_n,\tfRttCalF_{j,t}$ & $\sigma$-fields & Causally available information at event and microstep. \\
$\sigma_j,\epsilon_j$ & stopping times & Attempt $j$ start and end. \\
$v_j,u_j$ & stopping times & Feedback availability and update completion. \\
$\tau$ & stopping time & Episode resolution, abandonment, budget, or safety endpoint. \\
$e$ & $\mathbb N$ & Reset epoch attached to callbacks and events; stale-epoch returns are rejected and logged. \\
$r^\star$ & registered record & Complete reset snapshot $(s^\star,n^\star,\kappa^\star,e^\star,H(s^\star))$: bytes, event index/key, epoch, and state hash. \\
$D,q$ & map, event index & Registered complete restore map and the event at which restoration, draining/cancellation, and fresh-epoch activation finish. \\
$\mathfrak W_j$ & product record & Strict-witness bundle in Equation~\eqref{tf:rtt:eq:witness-record}. \\
$R,F$ & $\{0,1\}$ & Mounted-state and feedback-semantic randomized factors. \\
$Q(r,f)$ & $[0,1]$ & Potential qualification outcome under factorial arm $(r,f)$. \\
$\theta_{\mathrm{state}}$ & $[-1,1]$ & True-feedback post-state versus pre-state effect. \\
$\theta_{\mathrm{int}}$ & $[-2,2]$ & State-by-feedback semantic interaction. \\
$S,V$ & $\{0,1\}$ & True success and proxy verifier decision. \\
$\tfRttTV,\tfRttKL$ & divergences & Total variation and Kullback--Leibler divergence. \\
$\Theta$ & outer parameter & Meta-learned or engineered update-rule parameters. \\
$\mathcal J(\Theta)$ & $\mathbb R$ & Complete episode-level quality--cost--risk objective. \\
\bottomrule
\end{longtable}

\subsection{Reserved terminology}

\begin{description}
  \item[Attempt.] A bounded, version-receipted policy invocation intended to advance or solve the current problem; internal token steps are not automatically separate attempts.
  \item[Feedback.] A causally available record derived from an attempt or environment interaction. Its presence does not imply truth.
  \item[Policy state.] The declared carrier $\Phi$ that directly parameterizes the fresh action/proposal kernel.  Policy-only attribution additionally requires a byte-identical $\Xi^{\mathrm{read}}$ closure; otherwise the intervened functional carrier is $\Psi=(\Phi,\Xi^{\mathrm{read}})$.
  \item[Memory state.] A typed store read through a memory/retrieval path; mutation alone is \tfRttMemoryAdaptation{}, not strict \tfRttName{}.
  \item[Workspace.] Candidate-specific search and tool state, including trees, caches, files, and scratch computations.
  \item[Resolution.] The registered terminal event for success, abandonment, budget exhaustion, or safety stop.
  \item[Strict RTT.] A complete W1--W8 functional update--reuse witness, without an implied benefit sign.
  \item[RTT benefit.] A prespecified causal outcome contrast; separate from mechanism identity.
\end{description}

%% file: theory_family/modules/rtt/appendices/C_audits.tex
\section{Printable Leakage, Circularity, and Claim Audits}
\label{tf:rtt:app:audits}

\subsection{Carrier/reset matrix}

\begin{longtable}{@{}p{0.23\linewidth}p{0.22\linewidth}p{0.23\linewidth}p{0.24\linewidth}@{}}
\caption{\textbf{Hidden-state and reset matrix (\tfRttPrescriptionStatus{}).} Each future experiment must replace ``declare'' with a concrete receipt.}\label{tf:rtt:tab:reset-matrix}\\
\toprule
Carrier & Hash/version & Restore or randomize & Failure interpretation \\
\midrule
\endfirsthead
\toprule
Carrier & Hash/version & Restore or randomize & Failure interpretation \\
\midrule
\endhead
Base weights and adapters & Content hash, mount route, dtype & Swap only registered policy factor & Wrong or bundled intervention \\
Declared policy carrier $\Phi$ & Content hash, mount route, dtype & Swap only the registered policy factor & Wrong, dead, or bundled intervention \\
Policy-readable closure $\Xi^{\mathrm{read}}$ & Typed manifest and pre/post closure hashes $\chi^-,\chi^+$ & Hold byte-identical for a $\Phi$-only contrast, or intervene on $\Psi$ & Unregistered functional carrier or invalid $\Phi$ attribution \\
Non-readable auxiliary state $\Xi^{\mathrm{aux}}$ & Moments, step, scaler, recurrent code, scheduler & Restore with candidate/pre-state & Hidden state or reset failure \\
Normalization/routing/gates & Running statistics and realized decisions & Register in $\Xi^{\mathrm{read}}$ when kernel-readable; otherwise fix & Dead/bypassed or misclassified update \\
Prompt/context/KV state & Serialized prompt and cache receipt & Restore before fresh proposal & Context confounding \\
External/resident memory & Rows, versions, index, retrieval seed & Restore or independent memory factor & Memory confounding \\
Search/workspace/files & Tree, node scores, files, cache keys & Restore before proposal swap & Search/workspace confounding \\
Tools, services, and callbacks & Session IDs, remote cache/version, pending ledger, epoch tags & Drain/cancel before completion; reject and log stale-epoch returns & Environment interference or incomplete reset \\
Randomness and batching & All RNG states, batch peers, scheduler & Common random numbers or assignment & Stochastic imbalance \\
Event order and reset epoch & Lexicographic keys $\kappa(e)$, event log, epoch transitions & Replay the registered order; activate one fresh epoch at completion & Ambiguous causality, stale mutation, or non-replayable reset \\
Verifier/reference & Model/data/config hash, provenance & Freeze before target outcomes & Reward circularity \\
Budget and stop logic & Ledger hash, thresholds, timers & Equal rules across arms & Differential exposure \\
\bottomrule
\end{longtable}

\subsection{Future-leakage checklist}

\begin{enumerate}[label=\textbf{L\arabic*.},leftmargin=3em]
  \item Is every update input admitted under a deterministic key no later than $u_j$?
  \item Was any later success, hidden test, future tool return, or final answer used to choose an update, checkpoint, learning rate, or gate?
  \item Were failed candidates omitted after their later outcomes were known?
  \item Does a post-$u_j$ random canary predict the committed delta above the registered tolerance?
  \item Were outer-training realized futures physically and logically separated from deployment artifacts?
  \item Does every complete event carry a unique registered key $\kappa(e)$, and does the log follow its deterministic lexicographic order independently of wall-clock arrival time?
  \item Are reset snapshot, restore completion, fresh epoch, pending-callback disposition, and every rejected stale-epoch return receipted?
  \item Are all target-outcome-dependent revisions recorded as a new protocol rather than silently folded into the old one?
\end{enumerate}

Any ``yes'' to L2--L4 or L8, or an unresolved answer to L1/L5--L7, blocks a deployed causal-update claim.  A clean callback ledger is necessary but does not by itself certify that the registered reset map was complete.

\subsection{Circularity checklist}

\begin{enumerate}[label=\textbf{C\arabic*.},leftmargin=3em]
  \item Is the feedback producer independent of the policy family? If not, is it explicitly labeled a proxy?
  \item Is true success externally defined and audited on a disjoint sample?
  \item Were reward, state divergence, and final utility reported separately?
  \item Were all arms, including stopped and failed updates, included in intention-to-treat reporting?
  \item Were hyperparameters and stop thresholds frozen before target-arm inspection?
  \item Was mechanism qualification performed irrespective of outcome success?
  \item Can the verifier be optimized without task progress (reward hacking), and are adversarial cases included?
\end{enumerate}

\subsection{Claim--evidence boundary}

\begin{table}[H]
\centering
\small
\begin{tabularx}{\linewidth}{@{}p{0.25\linewidth}p{0.25\linewidth}Y@{}}
\toprule
Claim class & Strongest admitted basis & Explicit exclusion \\
\midrule
Strict RTT vocabulary & Authorial definitions and typed interfaces & No priority claim \\
Separation results & Logical consequences of W1--W8 & No performance ranking \\
Identification results & Randomization/intervention assumptions & No guarantee assumptions hold in a system \\
Optimization results & Differentiability/support/group assumptions & No finite-step improvement \\
Safety results & Local KL and bounded-utility assumptions & No global or deployment safety \\
Related-work map & Verified primary-source protocol descriptions & No replication or independent result validation \\
MMLA boundary & Reported component evidence and semantic-interface negatives & No RTT evidence; no repair of closed overwrite gate \\
Empirical ladder & Planned protocols and stop rules & No gate executed or passed \\
\bottomrule
\end{tabularx}
\caption{\textbf{Claim--evidence audit.} The strongest paper-level claim is a proved conditional statement or a source-verified protocol description; no empirical RTT success is admitted.}
\label{tf:rtt:tab:claim-audit}
\end{table}

%% file: theory_family/modules/amr/module.tex
\clearpage
\part{Atomic Memory Rows}
\label{tf:amr:part:atomic-memory-rows}

\noindent\textbf{Scope.}
This part defines a bounded complete-row authority and transaction contract,
including canonical and neural views, trusted assembly, exact NULL, version
lineage, protection, recovery, security, and cost obligations under explicit
assumptions.

\input{theory_family/modules/amr/sections/01_scope}
\input{theory_family/modules/amr/sections/02_state_schema}
\input{theory_family/modules/amr/sections/03_exact_budget}
\input{theory_family/modules/amr/sections/04_trust_assembly}
\input{theory_family/modules/amr/sections/05_dual_view}
\input{theory_family/modules/amr/sections/06_transition_algebra}
\input{theory_family/modules/amr/sections/07_lifecycle}
\input{theory_family/modules/amr/sections/08_invariants_results}
\input{theory_family/modules/amr/sections/09_security_recovery}
\input{theory_family/modules/amr/sections/10_costs}
\input{theory_family/modules/amr/sections/11_counterexamples}
\input{theory_family/modules/amr/sections/12_falsification}
\input{theory_family/modules/amr/sections/13_related_boundary}

\FloatBarrier
\input{theory_family/modules/amr/appendices/A_proofs}
\input{theory_family/modules/amr/appendices/B_profile_ledger}
\input{theory_family/modules/amr/appendices/C_obligations}
\input{theory_family/modules/amr/appendices/D_symbols}

%% file: theory_family/modules/amr/sections/01_scope.tex
\section{Introduction}
\label{tf:amr:sec:scope}

\begin{figure}[H]
\centering
\includegraphics[width=\linewidth]{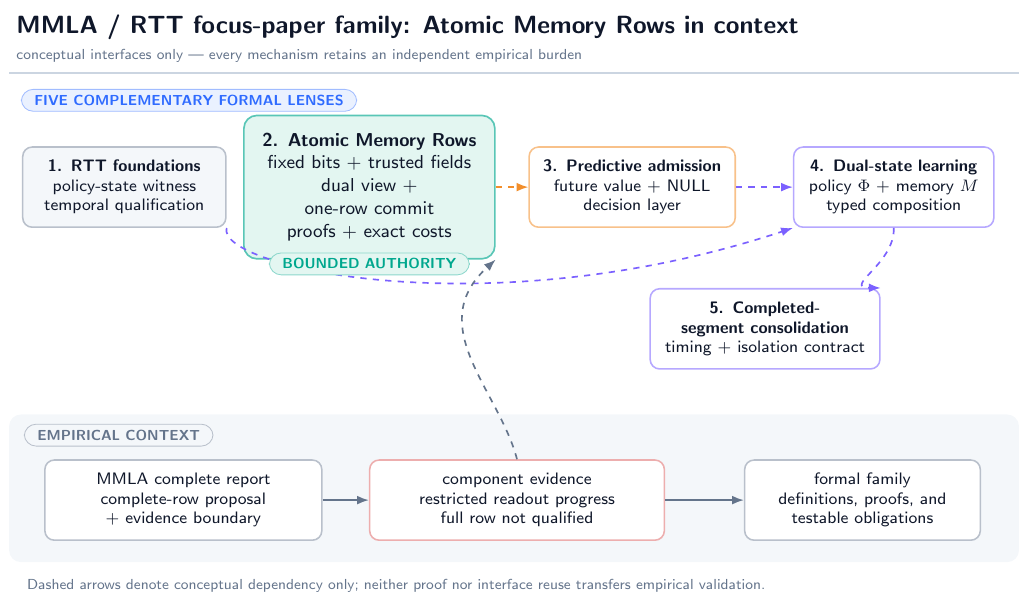}
\caption{Atomic memory rows: complete transitions, authority and recovery contracts, and their dependencies. Component readout results do not establish the full row implementation.}
\label{tf:amr:fig:technology-tree}
\end{figure}

\subsection{The state problem hidden by the word memory}

A transcript can retain an observation without deciding whether it is current. A vector store can retrieve neighbors without identifying which record is authoritative. A differentiable memory matrix can change continuously without specifying deletion, provenance, tenant scope, or a version boundary. An append-only audit stream can preserve history while growing without limit. These are useful objects, but none alone is a bounded editable fact state.

The missing object is a complete unit of authority. Suppose a resident record says that Alice is in Paris and a later observation proposes Berlin. Before ``write Berlin'' is a defined operation, a system must answer at least the following questions. Which logical object and physical slot are involved? Which tenant and principals may read or modify it? Does the proposal replace an entire version or blend with the old value? Are the canonical and neural representations synchronized? Which source and software builds produced the candidate? Does the version advance without wraparound? What happens to stale index entries? Is declining the write distinguishable from deletion? Can rollback restore Paris without restoring obsolete permissions? Which resident, index, archive, audit, and durable-write bytes are charged?

\tfAmrName{} answers those questions at the substrate level. It does not decide whether Berlin is true or useful, and it does not decide whether the proposal deserves admission. The distinction between state correctness and decision quality is the paper's central evidence boundary.

\subsection{Authority is narrower than storage}

\tfAmrDefinitionStatus
\begin{tfAmrDefinition}[Authoritative state]
A mutable component is \emph{authoritative} if changing it can change the system's declared current resident object, value, deletion status, security decision, or permitted transition without first validating against a current row. The authoritative resident memory is an ordered table
\begin{equation}
M=(r_1,\ldots,r_S)\in\tfAmrCalR_\sigma^S,
\label{tf:amr:eq:memory-table}
\end{equation}
where $S<\infty$ and schema $\sigma$ fixes the exact row serialization length $B_R$ bytes.
\end{tfAmrDefinition}

An alias map that can redirect a query without revalidating the current row is authoritative. So is a security map that can grant access independently of the row. Both are outside the contract unless included in the same fixed budget and commit boundary. By contrast, an index entry is a non-authoritative hint when every hit is rechecked against the current row's tenant, identity, slot sequence, status, and receipt.

We distinguish five state classes.
\begin{enumerate}
  \item \textbf{Transient context} contains tokens, activations, and scratch reasoning for the current computation.
  \item \textbf{Authoritative resident memory} is the fixed table $M$ and is the only resident state permitted to declare a live, deleted, protected, or quarantined object.
  \item \textbf{Deterministic derived views} include exact key/alias and neural-routing indexes. They are rebuildable functions of validated rows and never grant authority.
  \item \textbf{External archives and audit ledgers} may retain source documents, superseded values, or receipts. Their storage and retention policies are separately charged; they are not hidden inside the resident bound.
  \item \textbf{Model or policy state} includes fixed parameters and any separately typed fast or trainable state. Its mutation is not an \tfAmrName{} row operation.
\end{enumerate}

\tfAmrPrescriptionStatus\quad Every implementation must publish a state-carrier inventory assigning each mutable byte to exactly one class, with its capacity, lifetime, authority, and update rule. ``Cache,'' ``metadata,'' or ``log'' is not an exemption from accounting.

\subsection{Contributions and scope}

This manuscript owns the following formal and systems-contract questions:
\begin{itemize}
  \item a typed, fixed-width authoritative row and a reconciled concrete profile;
  \item neural-proposed versus trusted system-owned fields;
  \item canonical/neural conformance, receipts, re-encoding, and fail-closed mismatch behavior;
  \item one complete-row commit or bit-identical \tfAmrNull{} per ordered event;
  \item insert, update, consolidating update, eviction, alias rename, deletion, rollback, quarantine, repair, and purge semantics;
  \item per-event and segment-level edit-locality bounds;
  \item deterministic derived indexes, stale-entry suppression, tenant partitioning, ACLs, protection, provenance, tombstones, and bounded rollback;
  \item exact resident, index, candidate, validation, publication, archive, and audit accounting;
  \item invariant, concurrency, crash-recovery, impossibility, and counterexample results under explicit assumptions; and
  \item a theorem-to-implementation obligation graph and falsification ladder.
\end{itemize}

It does not own the event extractor, learned target ranker, predictive admission objective, future-valued controller, archive-retrieval policy, general multi-row transactions, distributed consensus, or end-to-end language-model quality. Those components can consume or produce \tfAmrName{} objects only after their own evidence gates are satisfied.

\subsection{Relationship to the report-wide evidence}

The integrated architecture proposes a bounded resident table, a canonical/neural successor row, a trusted assembly boundary, and one overwrite or \tfAmrNull{} per event. This part sharpens the fixed-bit, event-recursive, dual-view, and complete-cost contract. The report-wide negative semantic-interface evidence is retained as a boundary rather than reinterpreted as validation of \tfAmrName{}.

\paragraph{Evidence partition.}
The dual-view and atomic-row contract in this part is \textbf{FORMALIZED}; its
invariant, serializability, tamper-evidence, and recovery consequences are
\textbf{PROVED UNDER STATED ASSUMPTIONS}. Those labels do not certify an
implementation. AMR-1 implementation conformance is \textbf{UNVALIDATED}, and
semantic use of the proposed dual-view row is \textbf{NOT TESTED / UNVALIDATED}.
The earlier pure latent-row interface studies are \textbf{FAILED / UNQUALIFIED}
under their registered gates.

The historical result is exact. The frozen-checkpoint semantic-interface study qualified zero of nine dense, structured-span, and checkpoint-native jobs. Its structural checks passed while semantic prediction failed. A subsequent fitted-population diagnostic reported direct candidate reconstruction $40/73{,}728$, learned route/learned content $0/73{,}728$, oracle route/learned content $0/73{,}728$, learned route/oracle content $18{,}544/73{,}728$, and the mechanical full oracle $73{,}728/73{,}728$. These numbers show that target-row isolation did not repair learned content in that protocol. Later post-training studies establish restricted readout, update, and restoration with different lineages (\S\ref{sec:posttraining-evidence}), while the latest nine-trajectory study does not qualify generalized continuous-event readout (\S\ref{sec:coverage-readout}). Neither result evaluates all fields, trusted assembly, concurrency, security, or crash recovery of the fixed-row profile introduced here.

\tfAmrUnvalidatedStatus\quad No trained checkpoint in the evidence cutoff is qualified for the complete \tfAmrName{} row and its authority contract. A successful restricted latent answer is not complete-row conformance. This formal part reports no implementation crash-injection, security, energy, or hardware qualification; measured costs of the separate readout studies do not supply those missing tests.

Formal statements carry their assumptions locally. Definition, proof, prescription, and planned-test labels identify the kind of support offered; none is evidence that a checkpoint can use the row. The next sections specify the row schema and exact budget, separate proposed from trusted fields, define dual-view consistency and the transition algebra, then develop lifecycle, invariants, security, recovery, costs, counterexamples, and falsification tests. Whether these bytes improve a reasoner's behavior remains open.

%% file: theory_family/modules/amr/sections/02_state_schema.tex
\section{Typed State and Complete-Row Schema}
\label{tf:amr:sec:schema}

\subsection{Finite universes and schema parameters}

Fix a schema identifier $\sigma$ and finite-width spaces
\begin{equation}
\begin{aligned}
\tfAmrCalS&=\{1,\ldots,S\},
&\tfAmrCalT&=\{0,1\}^{b_T},
&\tfAmrCalU&=\{0,1\}^{b_U},\\
\tfAmrCalO&=\{0,1\}^{b_O},
&\tfAmrCalV_s&=\{0,\ldots,2^{b_s}-1\},
&\tfAmrCalV_o&=\{0,\ldots,2^{b_o}-1\}.
\end{aligned}
\label{tf:amr:eq:finite-universes}
\end{equation}
They denote physical slots, tenants, principals, logical objects, physical-slot sequences, and same-object versions. A trusted static partition
\begin{equation}
P:\tfAmrCalS\rightarrow\tfAmrCalT
\label{tf:amr:eq:tenant-partition}
\end{equation}
assigns each slot to exactly one tenant. Dynamic repartitioning is an administrative storage migration with its own transaction; it is not an ordinary memory event.

Schema $\sigma$ fixes at least the following finite parameters: canonical alphabet and normalization, key and value lengths, alias count and alias length, neural dimensions and precision, ACL capacity, identifier widths, rollback depth, validation bitmap width, digest and authentication algorithms, time representation, status normal forms, and the exact byte order of every field. It also fixes a finite validation registry
\[
  \mathcal V_\sigma:b\longmapsto
  (\textsf{predicate-id},\textsf{version},\textsf{applicability},\textsf{fail-mode}),
\]
whose digest $d_{\mathcal V}=H(\tfAmrSer(\mathcal V_\sigma))$ is receipt-covered. An unknown bit, version, or applicability code fails closed. A schema or registry change is a migration, not an in-place reinterpretation.

\tfAmrDefinitionStatus
\begin{tfAmrDefinition}[Status]
Every parsed row has one status
\begin{equation}
z(r)\in\{\tfAmrEmpty,\tfAmrLive,\tfAmrTomb,\tfAmrQuar\}.
\end{equation}
An \tfAmrEmpty{} row has no logical object but retains physical slot, tenant, slot sequence, and a valid empty-row receipt. A \tfAmrLive{} row may supply content to an authorized reader. A \tfAmrTomb{} row records a committed deletion and is unavailable to ordinary semantic reads. A \tfAmrQuar{} row records a bounded integrity failure and is unavailable until an authorized repair commits a new complete version.
\end{tfAmrDefinition}

\tfAmrEmpty{} is not deletion, \tfAmrTomb{} is not \tfAmrNull{}, and \tfAmrQuar{} does not assert that the logical object was deleted.

\subsection{The row product type}

\tfAmrDefinitionStatus
\begin{tfAmrDefinition}[Atomic Memory Row]
An Atomic Memory Row is the product
\begin{equation}
r=(h,c,n,p,\lambda,\rho)\in
\mathcal H_\sigma\times\mathcal C_\sigma\times\mathcal N_\sigma
\times\mathcal P_\sigma\times\mathcal L_\sigma\times\mathcal Q_\sigma,
\label{tf:amr:eq:row-product}
\end{equation}
where $h$ is the trusted control block, $c$ the canonical capsule, $n$ the neural capsule, $p$ the provenance capsule, $\lambda$ the bounded rollback lineage, and $\rho$ the consistency receipt. Schema $\sigma$ defines a deterministic serializer and partial parser
\begin{equation}
\tfAmrSer_\sigma(r)\in\{0,1\}^{8B_R},\qquad
\tfAmrParse_\sigma:\{0,1\}^{8B_R}\rightarrow\tfAmrCalR_\sigma\cup\{\bot\}.
\label{tf:amr:eq:serialization}
\end{equation}
The row is the publication unit: no field in Eq.~\ref{tf:amr:eq:row-product} is authoritative on a different commit boundary.
\end{tfAmrDefinition}

The serializer is injective over valid rows. Lengths, endianness, normalization, padding, ordering, floating or quantized numeric semantics, reserved-bit values, and prohibited encodings are part of $\sigma$. Deterministic encoding recommendations such as CBOR's deterministic profiles are useful background \cite{tf:amr:bormann2020cbor}, but \tfAmrName{} requires one schema-specific bit string rather than merely equivalent decoded maps.

\subsection{Trusted control block}

The control block is
\begin{equation}
\begin{aligned}
h=(&\text{magic},\sigma,z,\text{committed-op},\text{flags},a,t,o,u,\tau,\\
&v_s,v_o,t_{\min},t_{\max},t_{\mathrm{retain}},\pi,\delta,\mathrm{ACL},q_{\mathrm{rb}}),
\end{aligned}
\label{tf:amr:eq:trusted-header}
\end{equation}
where $a$ is physical slot, $t$ tenant, $o$ object, $u$ owner, $\tau$ object type, $v_s$ slot sequence, $v_o$ object version, the three times encode validity and retention, $\pi$ is protection class, $\delta$ a deletion-reason code, and $q_{\mathrm{rb}}$ describes rollback-ring occupancy. The ACL is a fixed array of principal/right pairs. Its empty entries have a unique zero representation.

The slot sequence advances on every committed replacement of a physical slot. The object version advances when the same logical object remains resident. Insertion of a different object starts its object version at one while still advancing the slot sequence. This two-counter design makes stale physical pointers distinguishable from a new object that reuses the same slot.

\subsection{Canonical capsule}

The canonical capsule is
\begin{equation}
c=(\tau_c,k_c,v_c,A_c,f_c),
\label{tf:amr:eq:canonical-capsule}
\end{equation}
where $\tau_c$ is a type tag, $k_c$ a typed key, $v_c$ a typed value or canonical tuple, $A_c$ an ordered bounded alias array, and $f_c$ semantic flags. The reference profile places the authoritative type and flags in the trusted control block while the canonical byte region contains the normalized key, value, and aliases. Their joint interpretation is nevertheless one canonical capsule.

Canonical means inspectable and deterministically serialized, not factually true. Unicode normalization, case handling, locale, units, temporal interpretation, numeric canonicalization, and missing-value semantics are fixed by schema. A canonical \texttt{0} is data with a nonzero type/length encoding. Padding bytes are not content.

Aliases are alternate query handles for the same row identity. They do not create extra logical objects or independently mutable mappings. Alias rename is a complete same-object row update, and every derived alias-index hit is revalidated against the current row.

\subsection{Neural capsule}

The neural capsule is
\begin{equation}
n=(k_n,v_n,u_n,\alpha_n,\widetilde\iota),
\label{tf:amr:eq:neural-capsule}
\end{equation}
where $k_n$ is a fixed-dimensional routing vector, $v_n$ a fixed-dimensional semantic payload, $u_n$ bounded uncertainty metadata, $\alpha_n$ a bounded abstention score, and $\widetilde\iota$ proposed operation intent. The schema chooses a finite coordinate representation. AMR-1 uses signed 16-bit coordinates. A floating-point profile must fix byte order, rounding, negative zero, infinities, denormals, and NaNs; unrestricted NaN payloads are prohibited because bitwise equality and receipt hashing would otherwise be noncanonical.

The neural capsule is not authoritative by itself. A matching vector cannot override canonical content, tenant scope, a tombstone, or a failed receipt. A separately typed differentiable fast state may use soft reads and writes, but Section~\ref{tf:amr:sec:transition} prevents it from silently becoming the versioned fact table.

\subsection{Provenance, rollback, and receipt}

The provenance capsule records bounded identifiers or digests for the event, source material, proposer/model build, assembler, encoder, validator, observation time, source class, committed operation, and deletion authorization. Provenance concepts distinguish where a datum came from and how it was transformed \cite{tf:amr:buneman2001provenance}; the \tfAmrName{} capsule intentionally stores only a finite receipt, not an unbounded derivation graph.

\tfAmrProvedStatus\quad Under the cryptographic assumptions in Section~\ref{tf:amr:sec:assumptions}, provenance can bind declared source and build identifiers to committed bytes. It cannot establish that the source was truthful, that the validator was competent, or that the neural representation is useful.

Rollback lineage $\lambda$ is a ring of at most $H_{\mathrm{rb}}$ fixed-width same-object snapshots. A snapshot contains the canonical and neural semantic state plus bounded prior version, validity, protection, and digest metadata, but contains no nested rollback lineage. A different object evicted from a slot is not placed in the new object's rollback ring. Recovering it requires a separately charged archive or a new source event.

Let $\operatorname{core}(r)$ be every serialized row byte except the receipt. A predecessor is not an untyped digest. Schema $\sigma$ defines the disjoint normal forms
\[
 \operatorname{pred}_{\rm gen}(\eta,a),\qquad
 \operatorname{pred}_{\rm live}(\eta,a,v_s,d_{\rm core}),\qquad
 \operatorname{pred}_{\rm migrate}(\eta_s,a_s,v_s,d_s;\eta_d,a_d),
\]
where $\eta$ is a non-reused namespace epoch. Domain tags make genesis, ordinary replacement, and migration byte-distinct; no all-zero string is interpreted as a predecessor.

AMR-1 gives every predecessor form one 96-byte serialization: a 1-byte kind tag, 16-byte source namespace epoch, 4-byte source slot, 8-byte source slot sequence, 32-byte source-core digest, 16-byte destination namespace epoch, 4-byte destination slot, and 15 reserved zero bytes. Genesis zeros the source fields and names only the destination; ordinary replacement names the source and zeros the destination fields; migration names both. Every field inapplicable to the tagged form is zero. Thus the semantic forms above have one fixed-width, injectively tagged ledger representation.

For a hash $H$ and authentication scheme $\operatorname{Tag}_K$, define
\begin{equation}
d_{\mathrm{core}}=H(\texttt{AMR-CORE}\parallel\operatorname{core}(r)),
\label{tf:amr:eq:core-digest}
\end{equation}
and
\begin{equation}
\rho=(d_{\mathrm{core}},\operatorname{pred},d_{\mathcal V},b_{\mathrm{val}},\text{alg-ids},\text{auth-bind},\text{event-bind},\operatorname{Tag}_K(m),\text{reserved}),
\label{tf:amr:eq:receipt}
\end{equation}
where
\begin{equation}
m=H(\texttt{AMR-RECEIPT}\parallel\text{schema-id}\parallel d_{\mathrm{core}}\parallel\operatorname{pred}
\parallel d_{\mathcal V}\parallel b_{\mathrm{val}}\parallel\text{alg-ids}
\parallel\text{auth-bind}\parallel\text{event-bind}\parallel\text{assembler-id}).
\label{tf:amr:eq:receipt-message}
\end{equation}
The digest is the row's cryptographic checksum; the authentication tag may be a MAC such as HMAC or a signature according to the trust model \cite{tf:amr:nist2015sha,tf:amr:nist2008hmac}. The receipt authenticates bytes and declared checks under stated assumptions. It is not a zero-knowledge proof, an attestation that the source is true, or evidence that the validator implements the intended semantics.

\begin{figure}[H]
\centering
\includegraphics[width=\linewidth]{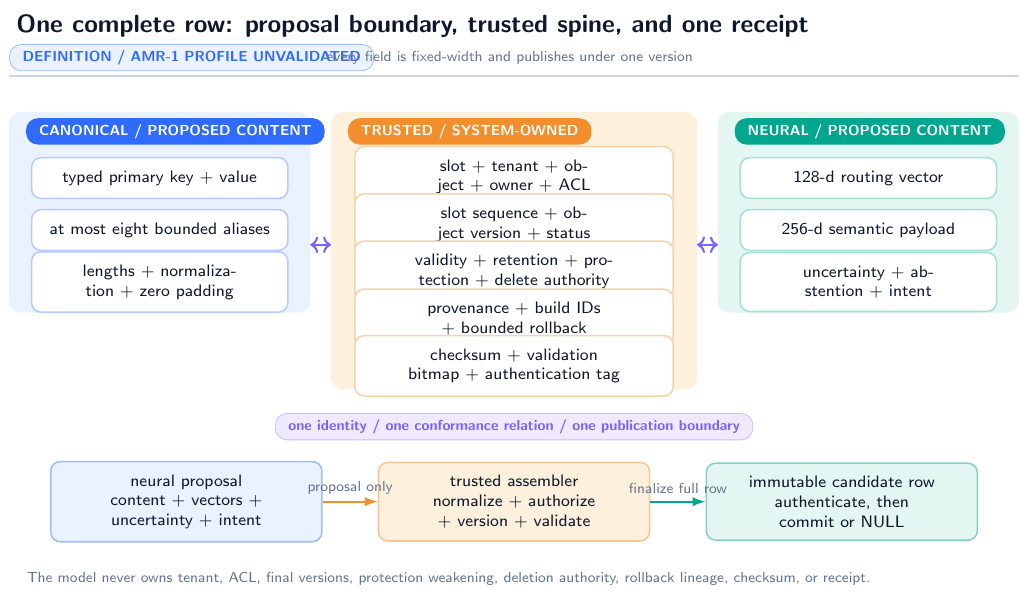}
\caption{\textbf{Complete-row anatomy and ownership.} The figure defines one authoritative serialization. Neural components may propose bounded content, vectors, uncertainty, and intent; trusted code fixes identity, security, versions, policy, lineage, deletion authorization, validation state, and the authenticated receipt. Both views and every control field publish together or none becomes authoritative.}
\label{tf:amr:fig:row-anatomy}
\end{figure}

\subsection{Normal forms and missing values}

Status normal forms eliminate ambiguous interpretations.
\begin{itemize}
  \item \tfAmrEmpty{} has zero logical object, zero active canonical/neural regions, tenant and slot binding, current slot sequence, and a valid receipt.
  \item \tfAmrLive{} has a nonzero object, a well-formed canonical capsule, a conforming neural capsule, and readable status subject to current policy.
  \item \tfAmrTomb{} retains object identity, versions, deletion authorization, retention, and receipt while active value and read payload regions take the schema's tombstone normal form.
  \item \tfAmrQuar{} retains only the trusted recoverable identity and bounded failure metadata permitted by policy; its active content is unavailable.
\end{itemize}

A schema-level optional value is encoded with an explicit presence bit or type tag. It is never inferred from an all-zero payload. \tfAmrNull{} is an action, not a nullable value. This distinction becomes formal in Section~\ref{tf:amr:sec:transition}.

%% file: theory_family/modules/amr/sections/03_exact_budget.tex
\section{Exact Bit and Resource Budgets}
\label{tf:amr:sec:budget}

\subsection{A symbolic finite contract}

Let a schema allocate byte widths
\begin{equation}
B_R=B_h+B_c+B_n+B_p+B_\lambda+B_\rho.
\label{tf:amr:eq:row-byte-sum}
\end{equation}
Exactly $S$ rows therefore consume
\begin{equation}
B_{\mathrm{auth}}=SB_R\ \text{bytes},\qquad
\tfAmrBits(M)=8SB_R.
\label{tf:amr:eq:table-byte-bound}
\end{equation}
This is operational only if every summand is finite and includes the fields required to interpret and authorize a row. ``$S$ objects'' is not a bound when an object can carry an unbounded alias list, provenance graph, source text, or rollback history.

Let $B_I$ be fixed materialized-index capacity, $B_Q$ a bounded diagnostic quarantine buffer, $B_F$ separately typed non-authoritative fast state, $B_A(t)$ archive bytes retained at time $t$, and $B_L(t)$ audit-ledger bytes. Then
\begin{equation}
B_{\mathrm{hot}}=SB_R+B_I+B_Q+B_F,
\label{tf:amr:eq:hot-bound}
\end{equation}
while
\begin{equation}
B_{\mathrm{system}}(t)=B_{\mathrm{hot}}+B_A(t)+B_L(t)+B_{\mathrm{replica}}(t)+B_{\mathrm{fs}}(t).
\label{tf:amr:eq:system-bound}
\end{equation}
The resident contract bounds Eq.~\ref{tf:amr:eq:hot-bound}. A deployment may also cap every term in Eq.~\ref{tf:amr:eq:system-bound}; otherwise it must say that total retained storage is unbounded. A finite row digest that points to an ever-growing archive does not make the archive finite.

\subsection{AMR-1: a reconciled concrete profile}

\tfAmrPrescriptionStatus\quad AMR-1 is an illustrative auditable profile, not an optimized standard and not an implemented format.

\begin{table}[H]
\centering
\small
\begin{tabularx}{\linewidth}{lYr}
\toprule
Block & Exact allocation & Bytes \\
\midrule
Trusted control & identity, status, operation, versions, times, protection, fixed ACL, rollback pointers & 256 \\
Canonical capsule & key, value, eight aliases, lengths, deterministic padding & 1,104 \\
Neural capsule & 128-dimensional route, 256-dimensional payload, uncertainty, abstention, intent & 784 \\
Provenance & event/source/model/build/time/deletion-authorization identifiers and digests & 192 \\
Rollback & two nonrecursive, same-object snapshots of 2,048 bytes each & 4,096 \\
Receipt & core digest, typed predecessor, validation registry/bitmap, identifiers, authorization/event bindings, auth tag & 256 \\
\midrule
\textbf{Complete row} & one fixed serialization and publication unit & \textbf{6,688} \\
\bottomrule
\end{tabularx}
\caption{AMR-1 exact top-level row allocation. Appendix~\ref{tf:amr:app:profile-ledger} reconciles every subfield.}
\label{tf:amr:tab:amr1-top}
\end{table}

Thus
\begin{equation}
B_R=6{,}688\ \text{bytes}=53{,}504\ \text{bits}.
\label{tf:amr:eq:amr1-row}
\end{equation}
For $S=256$,
\begin{equation}
B_{\mathrm{auth}}=256\times6{,}688
=1{,}712{,}128\ \text{bytes}=1.6328125\ \text{MiB}.
\label{tf:amr:eq:amr1-table}
\end{equation}
The count excludes proposal buffers, model state, archives, filesystem metadata, allocation overhead, replicas, and diagnostic logs. Exclusion is not disappearance: Section~\ref{tf:amr:sec:costs} charges each separately.

\subsection{Fixed deterministic base indexes}

AMR-1 defines two optional, materialized, slot-major indexes. The canonical index stores one 112-byte entry for the primary key and each of eight aliases in every slot:
\begin{equation}
|I_C|=256\times 9\times112=258{,}048\ \text{bytes}.
\label{tf:amr:eq:canonical-index-size}
\end{equation}
The neural routing index stores one 336-byte entry per slot:
\begin{equation}
|I_N|=256\times336=86{,}016\ \text{bytes}.
\label{tf:amr:eq:neural-index-size}
\end{equation}
Hence
\begin{equation}
B_I=344{,}064\ \text{bytes},\qquad
B_{\mathrm{auth}}+B_I=2{,}056{,}192\ \text{bytes}=1.9609375\ \text{MiB}.
\label{tf:amr:eq:amr1-index-total}
\end{equation}
Each index entry carries tenant, object, slot, slot sequence, and receipt digest. A canonical entry additionally carries a key digest and alias ordinal; a neural entry carries the quantized routing vector. Index capacity cannot overflow because every slot owns a fixed projection. Optional trees, graphs, approximate-nearest-neighbor structures, and learned routers require separate caps and miss accounting.

\begin{figure}[H]
\centering
\includegraphics[width=\linewidth]{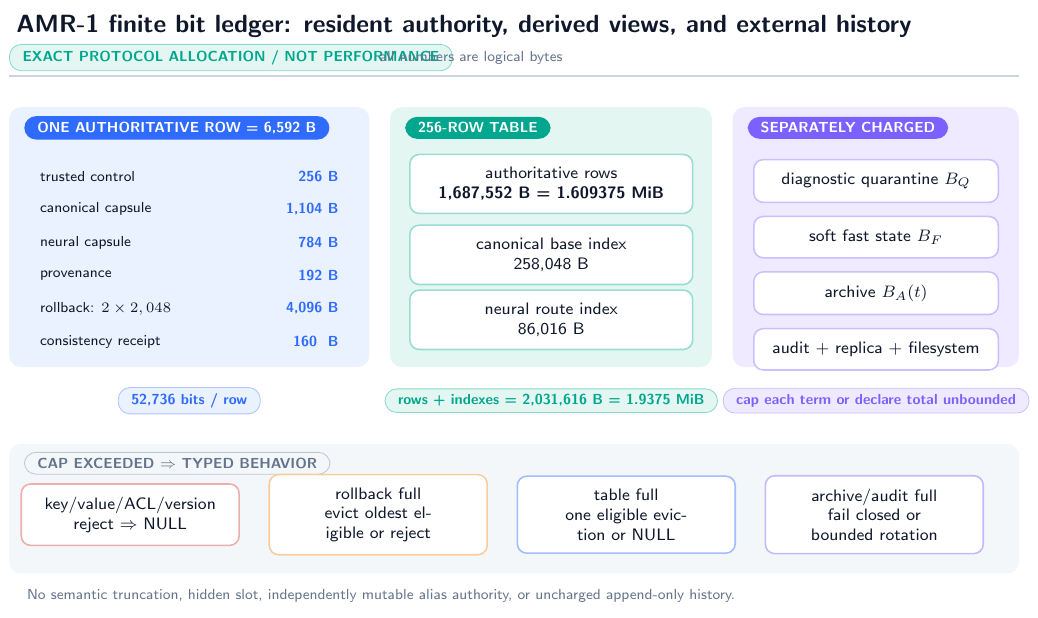}
\caption{\textbf{Finite resident accounting versus separately charged history.} AMR-1 fixes every authoritative row and base-index byte. Quarantine and fast state require explicit caps. Archive, audit, replication, and filesystem terms are either independently bounded or declared unbounded; they never disappear behind the resident-table notation.}
\label{tf:amr:fig:bit-budget}
\end{figure}

\subsection{Overflow and cap semantics}

\tfAmrPrescriptionStatus\quad Every cap has a typed action. Silent allocation, silent semantic truncation, and an implicit spill that changes authority are prohibited.

\begin{table}[H]
\centering
\footnotesize
\begin{tabularx}{\linewidth}{p{0.22\linewidth}p{0.28\linewidth}Y}
\toprule
Exceeded resource & AMR-1 action & Reason \\
\midrule
Canonical key/value length & reject candidate; event returns \tfAmrNull{} & Truncation can change object identity or meaning. \\
Alias count or alias length & strict profile rejects; a different schema may drop only explicitly nonessential aliases and must receipt the loss & Alias loss must never be silent or alter the primary key. \\
Neural dimension/precision & reject candidate & Reallocation would change schema and digest interpretation. \\
ACL capacity & reject ACL mutation or use an explicitly bounded external capability system whose decision is re-bound into the row & Silent principal loss can grant or deny access. \\
Provenance document & store only fixed digest/identifier; optional source document spills to separately charged archive & The row never embeds an unbounded provenance graph. \\
Rollback ring & deterministically evict the oldest eligible snapshot; if policy forbids loss, reject the new commit & History loss is explicit and bounded. \\
Resident table capacity & choose one eligible eviction target and replace it completely; otherwise \tfAmrNull{} & A hidden extra slot would violate the table bound. \\
Version counter & return \tfAmrNull{} with \textsf{F\_VERSION\_EXHAUSTED}; external re-key/migration required & Counters never wrap, saturate, or reuse an identity/version pair. \\
Base index & impossible under the fixed slot-major layout; corrupt indexes are discarded and rebuilt & Derived views cannot force an authoritative mutation. \\
Optional index & evict/rebuild/disable within its own cap; never accept an unvalidated hit & Index failure may hurt recall or latency, not authority. \\
Quarantine diagnostics & drop non-authoritative diagnostic detail under a bounded policy; resident reads still fail closed & Diagnostic capacity cannot make corrupt content readable. \\
Required archive receipt & if the archive cap is full, reject commits whose policy requires archival retention & A required durability/provenance premise cannot be silently weakened. \\
Required audit ledger & fail closed before publication, or rotate through an authenticated bounded checkpoint protocol & An unaudited commit is not equivalent to an audited one. \\
\bottomrule
\end{tabularx}
\caption{Cap behavior. \tfAmrNull{} is the authoritative result of precommit rejection; diagnostic reasons live in a separately bounded channel.}
\label{tf:amr:tab:cap-behavior}
\end{table}

\subsection{Finite capacity theorem}

\tfAmrProvedStatus
\begin{tfAmrTheorem}[Fixed information capacity]
\label{tf:amr:thm:capacity}
For a schema with $S$ rows of exactly $8B_R$ bits,
\begin{equation}
|\tfAmrCalM|\le 2^{8SB_R}.
\end{equation}
For every random history $X_{\le t}$ and resulting authoritative memory $M_t$,
\begin{equation}
H(M_t)\le 8SB_R,
\qquad I(M_t;X_{\le t})\le 8SB_R.
\label{tf:amr:eq:information-capacity}
\end{equation}
\end{tfAmrTheorem}

\begin{proof}
The complete table is a bit string of exactly $8SB_R$ bits, although only a subset may parse as valid. It therefore has at most $2^{8SB_R}$ possible values and entropy at most $8SB_R$. Mutual information is bounded by the entropy of either argument. The theorem concerns authoritative resident state only; adding external archives increases the corresponding system capacity.
\end{proof}

The theorem does not say that all bits are efficiently used, that the retained information is predictive, or that a language model can read it.

%% file: theory_family/modules/amr/sections/04_trust_assembly.tex
\section{Neural Proposal and Trusted Assembly}
\label{tf:amr:sec:assembly}

\subsection{The proposal is not a row}

Let a neural constructor receive a completed event $e$, bounded context summary $z$, target slot $a$, prior row $r_a$, and a read-only memory snapshot $M$. Its output is
\begin{equation}
\widetilde p=G_\phi(z,e,a,r_a,M)
=(\widetilde c,\widetilde n,\widetilde\iota,\widetilde q,\widetilde s),
\label{tf:amr:eq:neural-proposal}
\end{equation}
where $\widetilde c$ is proposed canonical content, $\widetilde n$ a proposed neural capsule, $\widetilde\iota$ operation intent, $\widetilde q$ uncertainty/abstention, and $\widetilde s$ bounded source handles. This object is untrusted and may be malformed, adversarial, stale, or semantically wrong.

The trusted assembler is a partial function
\begin{equation}
\operatorname{Assemble}_\sigma(M,a,\widetilde p,\chi)
\in\tfAmrCalR_\sigma\cup\{\bot\},
\label{tf:amr:eq:trusted-assembler}
\end{equation}
where authenticated context $\chi$ binds tenant, principal, event identifier, source-authentication result, and snapshot epoch. Authority is carried by a typed capability
\[
 \mathsf{cap}=(\mathrm{principal},\mathrm{tenant},\mathrm{scope},\mathrm{rights},\eta,
       t_{\min},t_{\max},\mathrm{nonce},\mathrm{issuer},\mathrm{tag}),
\]
and a trusted verifier $\tfAmrAuth_\sigma(\mathsf{cap},\chi,e,M)\in\{\tfAmrPass,\tfAmrFail\}$. The verifier checks issuer trust, tag, tenant/scope/right, epoch, time, nonce/revocation state, and the event target. The neural proposer neither emits nor extends $\mathsf{cap}$; it may only request an operation. Returning $\bot$ means that no authoritative candidate exists; the event resolves to \tfAmrNull{}.

\subsection{Field ownership}

\tfAmrPrescriptionStatus
\begin{table}[H]
\centering
\footnotesize
\begin{tabularx}{\linewidth}{p{0.26\linewidth}p{0.30\linewidth}Y}
\toprule
Field family & Neural role & Trusted/system role \\
\midrule
Canonical key, value, aliases, type & propose bounded content & normalize, type-check, resolve identity, enforce caps and uniqueness \\
Routing vector and semantic payload & propose or request deterministic encoding & encode/re-encode or validate exact conformance; freeze encoder identity \\
Uncertainty, abstention, requested operation & propose signals and intent & interpret under fixed policy; choose committed operation/status \\
Target slot & score or propose candidate target & verify partition, occupancy, eviction eligibility, and snapshot freshness \\
Tenant, owner, ACL & no authority & authenticate and assemble; require explicit capability for ACL change \\
Object identity & suggest reference only & allocate or resolve stable identifier and same-object relation \\
Slot sequence and object version & no authority & increment exactly; reject stale snapshot or overflow \\
Validity and retention & suggest evidence times only & map to policy-approved intervals and legal holds \\
Protection & request change only & inherit/strengthen; require capability to weaken \\
Deletion/tombstone & request intent only & verify delete/resurrection authority and deletion receipt \\
Provenance & provide bounded source handles & authenticate sources; bind event and build identifiers/digests \\
Rollback lineage & no authority & push eligible same-object snapshot; apply bounded eviction policy \\
Checksum, validation bitmap, auth tag & no authority & compute after every check over the final serialized core \\
\bottomrule
\end{tabularx}
\caption{Field ownership. A neural proposal cannot grant itself security scope, reset lineage, weaken protection, authorize deletion, or authenticate its own bytes.}
\label{tf:amr:tab:field-ownership}
\end{table}

System ownership is functional, not merely documentary. If a model can choose an arbitrary bit string that the assembler copies into the tenant or version field, then the model owns that field regardless of the API name.

\subsection{Assembler sequence}

For one target-conditioned proposal, the reference assembler executes:
\begin{enumerate}
  \item authenticate $\chi$, verify the typed capability with $\tfAmrAuth_\sigma$, and bind its decision digest to one snapshot epoch;
  \item verify $P(a)=\chi.\mathrm{tenant}$ and target feasibility;
  \item parse the requested operation and resolve whether the old and proposed records represent the same logical object;
  \item normalize canonical bytes and enforce every key, value, alias, type, and padding bound;
  \item derive or validate the neural capsule under the declared encoder/validator profile;
  \item evaluate ACL, protection, validity, retention, deletion, resurrection, eviction, and legal-hold policies;
  \item compute the next slot sequence and object version, rejecting stale inputs and overflow;
  \item assemble fixed provenance and update the bounded same-object rollback ring;
  \item serialize the complete core in canonical order;
  \item evaluate row-local and memory-global invariants against the same snapshot;
  \item compute the checksum, typed predecessor binding, registry digest, validation bitmap, authorization-decision binding, algorithm identifiers, and authentication tag; and
  \item return one immutable candidate together with a compare-and-swap precondition.
\end{enumerate}
Any failure returns a typed diagnostic outside authoritative state and no candidate. Candidate validation is not a menu from which the controller may select a partially checked row.

The legal-event predicate is therefore noncircular. For snapshot $M$, event $e$, context $\chi$, target $a$, and proposed replacement $r'$, define
\begin{align}
\tfAmrLegal_\sigma(M,e,\chi,a,r')\iff{}&
 \tfAmrParse_\sigma(e,\chi,r')\land \tfAmrTarget_\sigma(M,e,a)\land
 \tfAmrAuth_\sigma(\mathsf{cap},\chi,e,M)=\tfAmrPass\nonumber\\
&\land\tfAmrPred_\sigma(M,a,r')\land
 \tfAmrAssembly_\sigma(M,e,\chi,a,r')\land
 \tfAmrLocal_\sigma(r',\chi)\land\tfAmrGlobal_\sigma(M,a,r')\nonumber\\
&\land\tfAmrReady_\sigma(M,e,a,r').
\label{tf:amr:eq:legal-event}
\end{align}
Here $\tfAmrAssembly_\sigma$ means that every system-owned field equals the deterministic trusted result, while $\tfAmrReady_\sigma$ means that the complete receipt and any required audit/root dependencies are durably prepared for the same candidate. None of these predicates assumes that $r'$ has already been published. A commit is permitted exactly when this conjunction holds at the publication snapshot; otherwise the transition is \tfAmrNull{}.

\subsection{Local and global validation}

Row-local predicates check parsing, status normal form, field bounds, tenant/slot binding, versions, ACL encoding, protection, retention, dual-view conformance, rollback bounds, provenance completeness, reserved bytes, checksum, and authentication. Memory-global predicates check at least uniqueness of live/tombstoned object identifiers and primary typed keys within a tenant. A strict schema may also reject alias collisions; a permissive schema returns a set of candidates and still requires current-row revalidation.

Global validation must be snapshot consistent. Suppose two concurrent inserts both observe that key $k$ is absent and then commit to different slots. Per-row compare-and-swap alone permits a duplicate. The contract therefore requires either one total commit order per tenant, a tenant-epoch compare-and-swap, or a serializable uniqueness index whose update is part of the same transaction. Section~\ref{tf:amr:sec:security-recovery} makes this assumption explicit.

\subsection{No self-authentication}

\tfAmrProvedStatus
\begin{tfAmrProposition}[Trusted-field nondelegation]
\label{tf:amr:prop:trusted-field-nondelegation}
Assume the assembler computes every trusted field as a function of authenticated context, the current validated snapshot, fixed policy, and validated proposal content, and assume the proposer cannot forge authentication. Then changing only the neural proposal cannot directly choose tenant, ACL, final versions, protection weakening, deletion authority, rollback lineage, or a valid receipt outside the image permitted by those trusted functions.
\end{tfAmrProposition}

\begin{proof}
By construction, none of the named output coordinates is copied from an unconstrained proposal coordinate. Each is either fixed by authenticated context and snapshot, derived by deterministic policy, or rejected. A proposal can alter a trusted field only through an explicitly authorized request whose policy preconditions hold. A receipt authenticating an otherwise forbidden row would require either assembler deviation or authentication forgery, both excluded by assumption.
\end{proof}

This proposition is a software-boundary theorem. It does not prove that a particular assembler is bug-free, that its authentication is uncompromised, or that policy is ethically or legally correct.

%% file: theory_family/modules/amr/sections/05_dual_view.tex
\section{Canonical--Neural Consistency}
\label{tf:amr:sec:dual-view}

\subsection{Consistency is a relation, not co-location}

Let $c$ be canonical content, $(k_n,v_n)$ the routing and semantic neural bytes, $\omega$ a frozen encoder artifact, and $Q_\sigma$ a deterministic quantizer/serializer. The base AMR-1 conformance relation is
\begin{equation}
D_{\sigma,\omega}(c,n)=\tfAmrPass
\iff
\tfAmrSer_{k,v}(n)=\tfAmrSer_{k,v}\!\left(Q_\sigma(E_\omega(\tfAmrCan_\sigma(c)))\right).
\label{tf:amr:eq:derived-conformance}
\end{equation}
The row receipt binds one identity and version to the canonical digest, declared encoder digest, derived neural digest, resident neural digest, validation bitmap, and authentication tag. Digest equality accelerates audit, but the contract's equality predicate is over the fixed serialized neural bytes. A hash collision cannot convert unequal bytes into a valid base-profile row because the validator recomputes Eq.~\ref{tf:amr:eq:derived-conformance}.

\tfAmrDefinitionStatus
\begin{tfAmrDefinition}[Dual-view validity]
A \tfAmrLive{} candidate is dual-view valid when (i) its canonical bytes parse and normalize uniquely; (ii) its neural route and payload have the exact declared shape and finite representation; (iii) the registered conformance relation returns \tfAmrPass{}; (iv) identity, version, encoder/validator IDs, and security scope agree across the receipt and control block; and (v) every required semantic qualification bit in the validation bitmap is present. A missing check is not a pass.
\end{tfAmrDefinition}

The last condition is an interface gate, not a theorem. A profile may distinguish mechanical byte equality from empirical semantic qualification. AMR-1 can mechanically derive a vector even when no checkpoint can use it. Such a profile remains \tfAmrUnvalidatedStatus{} for semantic readout until reconstruction, query, calibration, preservation, and stale-suppression tests pass.

\subsection{Certified-proposal profiles}

A non-derived profile may permit the proposer to supply $n$ and use a fixed validator
\begin{equation}
D_{\sigma,\omega}(c,n)\in\{\tfAmrPass,\tfAmrFail,\tfAmrUnknown\}.
\label{tf:amr:eq:certified-conformance}
\end{equation}
The validator may combine deterministic decoding, typed relation checks, bounded distances, or a frozen semantic reader. Only \tfAmrPass{} may produce a \tfAmrLive{} row. \tfAmrFail{} and \tfAmrUnknown{} fail closed. The receipt records the exact validator and encoder builds; upgrading either creates a new schema or a versioned re-encoding event.

\tfAmrPrescriptionStatus\quad A learned validator must not be described as a proof. It is a component whose false-accept and false-reject behavior requires fresh, adversarial, and cross-implementation evidence. A receipt can make the responsible build attributable without making it correct.

\subsection{Write, read, load, and migration obligations}

The same relation is checked at more than one point.
\begin{itemize}
  \item \textbf{Write.} The complete candidate is validated after final canonicalization and neural serialization, before receipt creation and action selection.
  \item \textbf{Commit.} Canonical bytes, neural bytes, encoder/validator IDs, validation bitmap, and receipt publish atomically.
  \item \textbf{Read.} The current row's status, checksum, authentication, tenant/ACL, validity, and conformance are checked, either directly or through a cached verification whose key includes the exact slot sequence and receipt digest.
  \item \textbf{Load/recovery/export.} Rows are parsed and revalidated before they enter a readable table or serve as teacher state.
  \item \textbf{Index rebuild.} Derived canonical and neural entries are emitted only from validated current rows.
  \item \textbf{Refresh/re-encoding.} A changed encoder, quantizer, or payload is a new complete-row event with new versions and receipt; no in-place neural drift is legal.
\end{itemize}

For request context $\chi$, a read request carries the typed selector
\begin{equation}
s=(\tau,a,o,v_s,d_\rho,q),\qquad q\in\{\mathsf{canonical},\mathsf{neural}\},
\label{tf:amr:eq:typed-selector}
\end{equation}
where $\tau$ is the tenant, $a$ the physical slot, $o$ the logical object identifier,
$v_s$ the exact slot version, $d_\rho$ the receipt digest, and $q$ the requested view.
Let $\mathcal C_\sigma$ and $\mathcal N_\sigma$ be disjoint canonical and neural
output types.  The authoritative selector is therefore
\begin{equation}
\begin{aligned}
\tfAmrRead_\sigma(M,s,\chi)&\in\mathcal C_\sigma\uplus\mathcal N_\sigma\uplus\{\bot\},\\
\tfAmrRead_\sigma(M,s,\chi)&=
\begin{cases}
c(M_a),&q=\mathsf{canonical}\ \text{and all checks below pass},\\
n(M_a),&q=\mathsf{neural}\ \text{and all checks below pass},\\
\bot,&\text{otherwise},
\end{cases}\\[-0.2em]
&z(M_a)=\tfAmrLive,\quad
(\tau,a,o,v_s,d_\rho)=\operatorname{SelectorKey}(M_a),\\
&\operatorname{ReceiptOK}(M_a)=1,\\
&D_{\sigma,\omega}(c(M_a),n(M_a))=\tfAmrPass,\\
&\operatorname{AuthorizedRead}(M_a,s,\chi)=1,\quad
\operatorname{TimeValid}(M_a,\chi)=1.
\end{aligned}
\label{tf:amr:eq:readability}
\end{equation}
The exact selector equality prevents an index cache from returning a different version or
receipt.  The view tag $q$ fixes the output type before validation; no fallback
chooses whichever view happens to decode.

\begin{figure}[H]
\centering
\includegraphics[width=\linewidth]{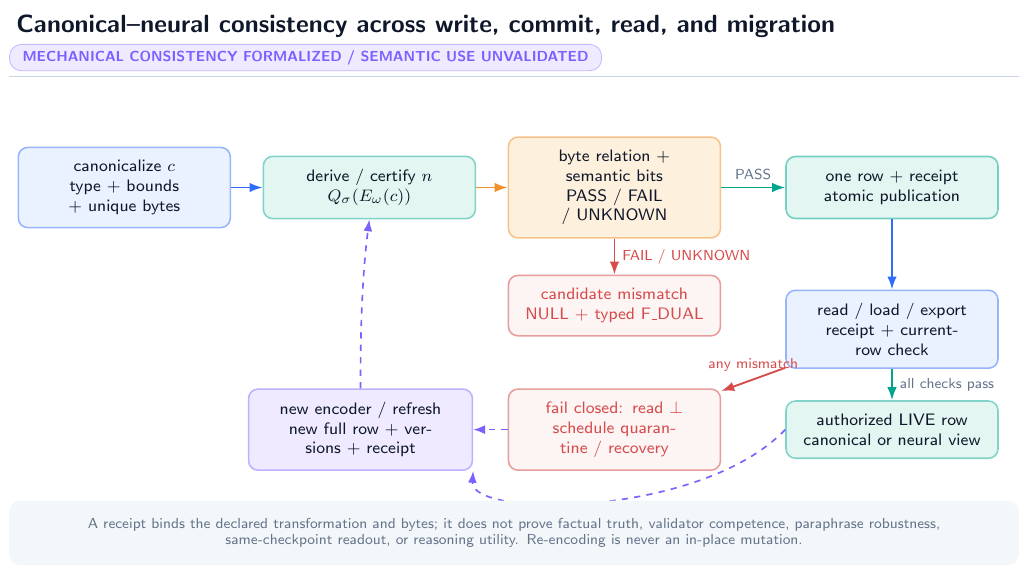}
\caption{\textbf{Dual-view conformance is checked across the row lifecycle.} A candidate mismatch produces \tfAmrNull{}; a resident mismatch suppresses the read and schedules a separately authorized quarantine or repair. Re-encoding is a new versioned commit. Mechanical equality is formalized; semantic usefulness remains an unpassed empirical gate.}
\label{tf:amr:fig:dual-view-loop}
\end{figure}

\subsection{Mismatch and quarantine}

Candidate-time mismatch has the authoritative result
\begin{equation}
M'=M,\qquad \mathrm{reason}=\textsf{F\_DUAL},
\label{tf:amr:eq:candidate-mismatch}
\end{equation}
with an optional copy in a separately bounded non-authoritative diagnostic buffer. That copy may never be read as fact memory.

A mismatch discovered in a previously committed row can arise from storage corruption, lost encoder artifacts, a bug, or compromised infrastructure. The immediate read result is $\bot$; the read path itself does not silently mutate authoritative state. A maintenance event may replace the same slot with a valid \tfAmrQuar{} row if identity and version lineage can be recovered under trusted storage metadata. If the row receipt or version is corrupt beyond safe recovery, repair requires an authenticated replica, copy-on-write root, or external recovery record. A checksum detects inconsistency; it does not invent the lost state.

\tfAmrProvedStatus
\begin{tfAmrTheorem}[No torn dual-view authority]
\label{tf:amr:thm:no-torn-dual}
Assume one atomic publication exposes either the complete old row or the complete new row. A legal commit cannot expose the new canonical view with the old neural view, or the old canonical view with the new neural view.
\end{tfAmrTheorem}

\begin{proof}
Both views are byte regions of one fixed row serialization. Atomic publication has exactly two visible row values: the precommit serialization and the postcommit serialization. Neither mixed serialization is in that set.
\end{proof}

Without atomic publication, a receipt can detect a tear after the fact but does not guarantee automatic recovery. Section~\ref{tf:amr:sec:security-recovery} states the stronger storage assumption needed for crash-safe old-or-new recovery.

\subsection{What conformance does and does not prove}

\tfAmrProvedStatus\quad In the derived profile, equality proves that the resident neural bytes are the declared deterministic encoder/quantizer output on the resident canonical bytes.

\tfAmrProvedStatus\quad In a certified profile, an authentic receipt proves only that the declared validator returned \tfAmrPass{} on the committed bytes, assuming the validator output and receipt path were faithfully implemented.

\tfAmrUnvalidatedStatus\quad Neither statement proves factual truth, semantic sufficiency, robustness to paraphrase, downstream reasoning improvement, same-checkpoint readout, or learned admission quality. The negative frozen \tfAmrMMLA{} evidence is precisely why this boundary must remain visible.

%% file: theory_family/modules/amr/sections/06_transition_algebra.tex
\section{Event-Recursive Transition Algebra}
\label{tf:amr:sec:transition}

\subsection{One row or none}

For memory $M$, target $a$, and complete candidate $\widehat r_a$, define
\begin{equation}
\tfAmrReplace(M,a,\widehat r_a)_i=
\begin{cases}
\widehat r_a,&i=a,\\
r_i,&i\neq a.
\end{cases}
\label{tf:amr:eq:replace}
\end{equation}

\tfAmrDefinitionStatus
\begin{tfAmrDefinition}[NULL]
The action \tfAmrNull{} is exact state identity:
\begin{equation}
T_{\tfAmrNull}(M)=M
\label{tf:amr:eq:null}
\end{equation}
bit for bit. It increments no version, creates no row receipt, changes no index generation, writes no tombstone, and clears no payload. A separately charged decision log may record a rejected attempt, but that log is not authoritative resident memory.
\end{tfAmrDefinition}

For one validated event, the authoritative transition is
\begin{equation}
T(M,e,\chi,a,\widehat r_a)=
\begin{cases}
M,&a=\varnothing\ \text{or validation fails},\\
\tfAmrReplace(M,a,\widehat r_a),&\text{authorized commit}.
\end{cases}
\label{tf:amr:eq:event-transition}
\end{equation}
An implementation can use copy-on-write pages, a write-ahead log, or an indirection root. Those are physical mechanisms for exposing the logical old-or-new row; they do not create a second authoritative update operator.

\subsection{Completed segments and ordered events}

Let completed segment $n$ be eventized into
\begin{equation}
E_n=(e_{n,1},\ldots,e_{n,K_n}),\qquad 0\le K_n\le K_{\max}.
\label{tf:amr:eq:event-list}
\end{equation}
The within-segment state recursion is
\begin{equation}
M_{n,0}=M_{n-1},\qquad
M_{n,j}=T_{n,j}(M_{n,j-1}),\qquad
M_n=M_{n,K_n}.
\label{tf:amr:eq:event-recursion}
\end{equation}
Every event is assembled and validated against its actual branch prefix $M_{n,j-1}$. A candidate computed for $M_{n,0}$ cannot be reused at event $j$ after earlier commits unless its snapshot preconditions are revalidated.

Let
\begin{equation}
C_n=\sum_{j=1}^{K_n}\tfAmrOne[M_{n,j}\neq M_{n,j-1}]
\end{equation}
be the number of accepted commits and let $D_n$ be the number of distinct physical slots changed at least once. Repeated updates to one slot count once in $D_n$ and multiple times in $C_n$.

Because rows are typed records rather than vectors, edit support is defined by
serialized row inequality:
\begin{equation}
\Delta_{\rm row}(M',M)=\{i\in[S]:\tfAmrSer_\sigma(M'_i)\ne\tfAmrSer_\sigma(M_i)\}.
\label{tf:amr:eq:typed-row-support}
\end{equation}

\tfAmrProvedStatus
\begin{tfAmrTheorem}[Event and segment edit locality]
\label{tf:amr:thm:edit-locality}
For every legal event $j$,
\begin{equation}
|\Delta_{\rm row}(M_{n,j},M_{n,j-1})|\le1,
\qquad d_H(M_{n,j},M_{n,j-1})\le8B_R.
\label{tf:amr:eq:event-locality}
\end{equation}
Across the segment,
\begin{equation}
C_n\le K_n,\qquad
D_n\le\min(C_n,S),\qquad
|\Delta_{\rm row}(M_n,M_{n-1})|\le D_n\le K_n.
\label{tf:amr:eq:segment-locality}
\end{equation}
If $V(M)\in\mathbb R^{S\times d_v}$ stacks decoded payloads, then
\begin{equation}
\operatorname{rank}(V(M_{n,j})-V(M_{n,j-1}))\le1
\end{equation}
for one event, while only the correct segment bound
\begin{equation}
\operatorname{rank}(V(M_n)-V(M_{n-1}))\le\min(D_n,d_v)
\label{tf:amr:eq:segment-rank}
\end{equation}
is generally valid.
\end{tfAmrTheorem}

\begin{proof}
\tfAmrNull{} changes no row. A committed event replaces exactly target $a$, so every other row difference is zero and the Hamming distance is at most the complete row width. Each event contributes at most one commit and one target to the union of touched slots; therefore $C_n\le K_n$ and $D_n\le\min(C_n,S)$. The final difference is supported only on that union. A matrix with at most one nonzero row has rank at most one; across the segment it may have $D_n$ nonzero rows, so rank is bounded by both $D_n$ and $d_v$.
\end{proof}

The theorem does not say that a row edit is semantically small. Nor does it say that a whole segment has rank one. With two events updating linearly independent payload rows, the segment difference can have rank two.

\begin{figure}[H]
\centering
\includegraphics[width=\linewidth]{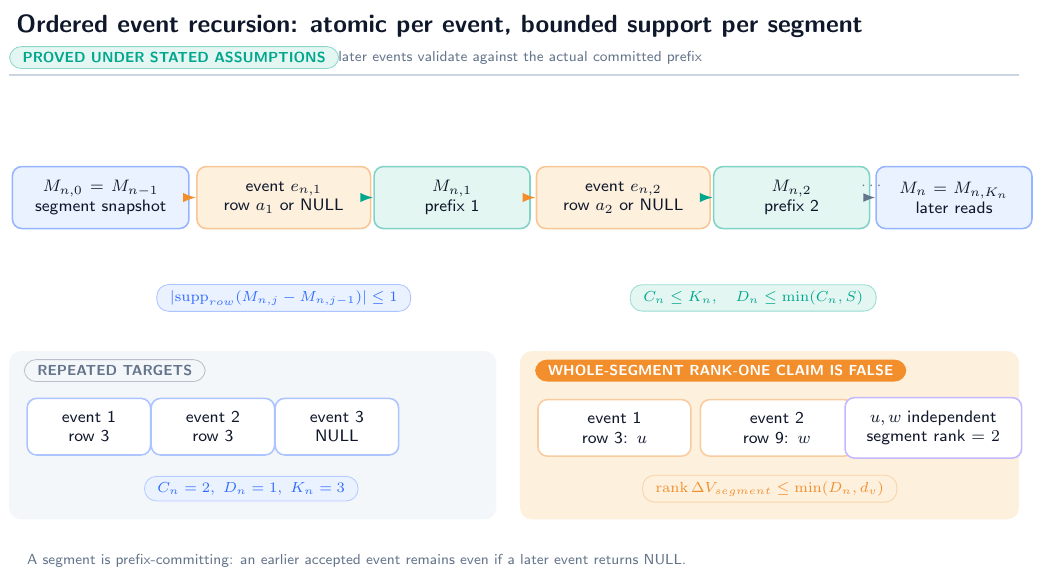}
\caption{\textbf{Event-level atomicity composes into a segment support bound, not segment rank one.} Each event reads its current prefix, then commits one row or \tfAmrNull{}. The union of changed targets has size at most $K_n$; repeated targets reduce the distinct-row count.}
\label{tf:amr:fig:event-recursion}
\end{figure}

\subsection{Hard authority and separately typed fast state}

An authoritative update cannot be a soft interpolation
\begin{equation}
r_a'=(1-\gamma)r_a+\gamma\widehat r_a,qquad0<\gamma<1,
\label{tf:amr:eq:soft-prohibited}
\end{equation}
because object IDs, ACLs, tombstone status, versions, type tags, and serialized canonical strings do not have a policy-independent convex meaning. Even restricting interpolation to the neural payload creates a third state unless its relation to the canonical version is revalidated and committed.

This does not prohibit differentiable computation. Let
\begin{equation}
F_t\in\mathbb F_q^{S_F\times d_F}
\label{tf:amr:eq:fast-state}
\end{equation}
be a separately typed bounded fast state with its own lifetime and reset rule. Soft writes may update $F_t$; generation may consume it as a heuristic. Conformance requires that $F_t$ cannot declare a row live, deleted, protected, current, or authorized, and cannot bypass a current \tfAmrName{} read check. Promoting information from $F_t$ to $M$ requires a normal complete candidate and hard commit.

\begin{figure}[H]
\centering
\includegraphics[width=\linewidth]{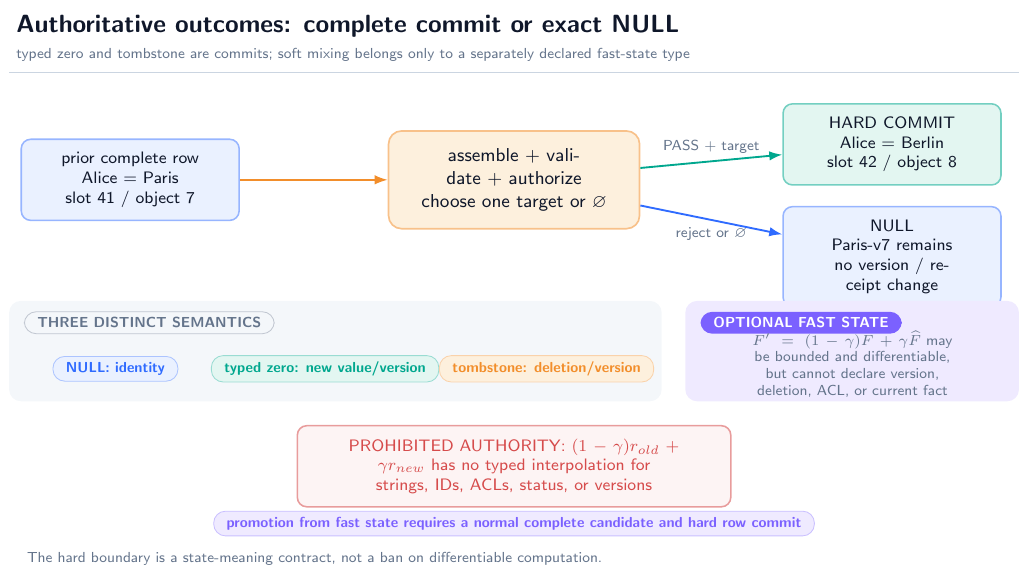}
\caption{\textbf{Authority has two outcomes: a complete hard commit or exact \tfAmrNull{}.} A typed zero is a committed value; a tombstone is a committed deletion state. Soft interpolation may exist only in a separately bounded, non-authoritative fast state and must pass ordinary assembly before promotion.}
\label{tf:amr:fig:hard-null-soft}
\end{figure}

\subsection{Zero, tombstone, and NULL}

A typed canonical integer zero has a type and length; an all-zero byte string may be a valid value; and an all-zero quantized payload may be valid when its canonical relation passes. Therefore a legal complete-row replacement with a zero value is still a commit. It advances the slot sequence, normally advances object version, and creates a receipt:
\begin{equation}
\tfAmrReplace(M,a,\widehat r_a^{\mathrm{zero}})\neq T_{\tfAmrNull}(M).
\label{tf:amr:eq:zero-not-null}
\end{equation}
A tombstone is likewise a complete new row with status, identity, versions, deletion authorization, and retention. Its active payload normal form may contain zeros, but the zeros do not define deletion.

\subsection{Segment partial success}

Equation~\ref{tf:amr:eq:event-recursion} is prefix-committing. If event $j$ commits and event $j+1$ returns \tfAmrNull{}, the earlier commit remains. The whole segment is not automatically all-or-nothing. A bounded segment transaction would require a separate transaction object that lists every row, read/write set, precondition, rollback record, byte bound, and atomic publication mechanism. Such a protocol is future work and cannot be inferred from per-event atomicity.

%% file: theory_family/modules/amr/sections/07_lifecycle.tex
\section{Lifecycle Semantics}
\label{tf:amr:sec:lifecycle}

Lifecycle names describe admissible old/new row pairs. They do not introduce extra physical operators.

\subsection{Version rules}

Every accepted replacement satisfies
\begin{equation}
v_s'=v_s+1.
\label{tf:amr:eq:slot-version}
\end{equation}
If logical object identity is unchanged,
\begin{equation}
v_o'=v_o+1;
\label{tf:amr:eq:object-version}
\end{equation}
if a different object is inserted, $v_o'=1$. A required increment is defined only below the maximum representable value. At exhaustion the event returns \tfAmrNull{} with a typed failure and requires an external re-key or migration protocol. It never wraps, saturates, or silently reuses a prior $(\eta,a,v_s)$ or $(\eta,o,v_o)$ tuple.

\subsection{Namespace migration and ABA exclusion}
\label{tf:amr:sec:migration-aba}

A persistent reference is the typed tuple
\[
  \mathsf{ref}=(\eta,a,v_s,d_{\mathrm{core}}),
\]
where the namespace-incarnation identifier $\eta$ is part of the authenticated domain. Slot indices and slot versions therefore have meaning only inside one incarnation. Migration from $\eta_s$ to $\eta_d$ is the following fail-closed protocol:
\begin{enumerate}[leftmargin=*,itemsep=2pt]
  \item allocate a destination incarnation $\eta_d$ that has never previously been used;
  \item seal the source ledger and authenticate its terminal root, schema identifier, validation-registry digest, and algorithm identifiers;
  \item authenticate a one-to-one map from every migrated source reference $(\eta_s,a,v_s,d)$ to a destination reference $(\eta_d,a',v_s',d')$;
  \item publish the destination root only after every mapped row and receipt validates under the destination profile; and
  \item reject every destination request carrying a source-incarnation reference rather than treating it as an alias.
\end{enumerate}
If a finite incarnation space is exhausted, migration rejects. It never wraps or reuses an incarnation identifier.

\begin{tfAmrProposition}[Migration excludes namespace ABA]
\label{tf:amr:prop:migration-aba}
Assume fresh nonreused incarnation identifiers, authenticated migration roots and maps, and exact typed-selector checking. Then a reference valid before migration cannot become valid for a different row merely because its slot index and slot version are later repeated.
\end{tfAmrProposition}
\begin{proof}
Let $x=(\eta_s,a,v_s,d)$ be a pre-migration reference. A destination read requires equality of the full selector key, including $\eta_d$ and the receipt digest. Freshness gives $\eta_d\neq\eta_s$, so $x$ is rejected even if $(a,v_s)$ is repeated. An authenticated migration map can issue a new reference $x'=(\eta_d,a',v_s',d')$, but $x'\neq x$ and its receipt binds the destination schema, registry, algorithms, and predecessor token. Authentication prevents substitution of the map or either root. Exhaustion rejects rather than wrapping. Hence index or version reuse cannot make $x$ denote a different destination row.
\end{proof}

Protection classes form a finite preorder $(\mathcal P,\preceq)$. An ordinary event may preserve or strengthen protection, $\pi_{\mathrm{old}}\preceq\pi_{\mathrm{new}}$. Weakening requires an authenticated capability bound into provenance. Retention is distinct: it determines eligibility for later deletion/purge, whereas protection governs authorization. The passage of time alone performs no memory mutation.

\subsection{Operation table}

\tfAmrDefinitionStatus
\begin{table}[H]
\centering
\footnotesize
\begin{tabularx}{\linewidth}{p{0.16\linewidth}p{0.20\linewidth}Y}
\toprule
Operation & Required old state & Required new-row semantics \\
\midrule
Insert & \tfAmrEmpty{} & \tfAmrLive{}; new object; object version 1; slot sequence $+1$ \\
Update & \tfAmrLive{} & \tfAmrLive{}; same object; complete canonical/neural replacement; both versions advance \\
Consolidating update & \tfAmrLive{} & \tfAmrLive{}; same object; one complete canonical capsule may summarize several observed facts; no second resident row is invalidated \\
Evict-and-insert & eligible \tfAmrLive{}/\tfAmrTomb{} & \tfAmrLive{}; different object; old object is not copied into new rollback lineage \\
Delete & \tfAmrLive{} & \tfAmrTomb{}; same object; content unavailable; deletion authority, reason, retention, and new receipt \\
Alias rename & \tfAmrLive{}/\tfAmrTomb{} & same object/status; bounded alias array replaced; version advances; index projection rebuilt \\
Protect / ACL change & \tfAmrLive{}/\tfAmrTomb{} & same object/content; policy-authorized control change; complete row reauthenticated \\
Rollback & \shortstack[l]{\tfAmrLive{}/\tfAmrTomb{}/\\\tfAmrQuar{}} & prior same-object semantic content; current security policy; new forward versions \\
Quarantine & recoverably identified corrupt row & \tfAmrQuar{}; active content unavailable; bounded reason and recovery provenance \\
Repair & \tfAmrQuar{} & newly validated \tfAmrLive{} or \tfAmrTomb{} row from an authenticated source \\
Purge & retention-eligible \tfAmrTomb{} & \tfAmrEmpty{}; slot sequence advances; external deletion/archive policy applied \\
Reject & any & \tfAmrNull{}; every authoritative bit unchanged \\
\bottomrule
\end{tabularx}
\caption{Lifecycle interpretations of one complete-row replacement. Preconditions are part of validation.}
\label{tf:amr:tab:lifecycle-operations}
\end{table}

\subsection{Insert, update, and consolidation}

Insertion allocates a new object identity or validates a preallocated one, starts object version one, and binds the row to the target tenant and slot. An update preserves object identity and replaces the complete semantic state. The old canonical or neural bytes do not survive as an independently readable shadow.

A consolidating update may combine several observations into one canonical tuple when they describe one logical object. For example, a row may replace separate observed location and timestamp fields with one typed current-location tuple. The assembler must declare which source receipts contribute and must fit them into bounded provenance. Consolidation does not authorize deleting a second resident object. If two resident rows represent duplicates, retiring one and enriching the other takes two ordered events or a future bounded transaction.

\subsection{Eviction}

Eviction is not deletion of an object from all storage. It is replacement of an eligible resident slot with a different object. The policy must respect tenant partition, protection, retention, legal hold, and any required archive precondition. The displaced object may be recoverable from a separately charged archive, but it is not hidden inside the new occupant's rollback ring. If no row is eligible, the event returns \tfAmrNull{}.

The eviction target may be proposed by a learned ranker, but eligibility and final selection constraints are trusted. This paper does not validate a learned eviction policy.

\subsection{Deletion and tombstones}

A deletion request needs a capability distinct from ordinary write permission. A committed tombstone retains enough bounded identity, version, authorization, and retention information to suppress stale resurrection during its declared scope. It does not assert that source documents, backups, replicas, archives, or model parameters were erased.

An expired retention deadline does not mutate the row. A later authorized purge may replace the tombstone with \tfAmrEmpty{} after archive, legal-hold, replica, and tenant policy checks. Once purged, perpetual non-resurrection over an unbounded object universe is impossible without an external authoritative deletion ledger; Theorem~\ref{tf:amr:thm:deletion-impossibility} makes this limit explicit.

\subsection{Alias rename}

An alias rename is a same-object complete replacement. The primary typed key remains authoritative; aliases are bounded query handles. Removing an alias means it disappears from the canonical capsule and therefore from the next deterministic index projection. An old alias-index entry contains the prior slot sequence and receipt digest and cannot be accepted after current-row revalidation.

Alias collision has two legal profile choices. A strict profile rejects the candidate. A permissive profile returns all current rows whose validated alias projections match and requires a later disambiguation step; it never makes ``first index hit'' authoritative.

\subsection{Rollback moves content backward and lineage forward}

Let snapshot $s_j$ be a same-object entry in the bounded rollback ring. A legal rollback constructs
\begin{equation}
\begin{aligned}
c'&=c(s_j),\\
n'&=\operatorname{Conform}_{\sigma,\omega}(c'),\\
v_s'&=v_s+1,\\
v_o'&=v_o+1.
\end{aligned}
\label{tf:amr:eq:rollback}
\end{equation}
Current ACL, tenant binding, retention, and protection are not blindly copied from the snapshot. Restoring old content is a compensating transaction, not time reversal. Rollback from a tombstone is an explicit resurrection and requires the corresponding authority.

\tfAmrProvedStatus
\begin{tfAmrProposition}[Monotone rollback]
\label{tf:amr:prop:monotone-rollback}
Under the rollback rule and authorization checks, a legal rollback may restore prior same-object semantic content but cannot decrease slot sequence, object version, or current protection without a separately authenticated weakening capability.
\end{tfAmrProposition}

\begin{proof}
The assembler reads semantic fields only from a same-object bounded snapshot. It computes both new versions from the current row by addition of one and rejects overflow. Current protection is inherited or strengthened unless $\chi$ carries a valid weakening capability. Therefore semantic content can revert while lineage moves forward.
\end{proof}

\begin{figure}[H]
\centering
\includegraphics[width=\linewidth]{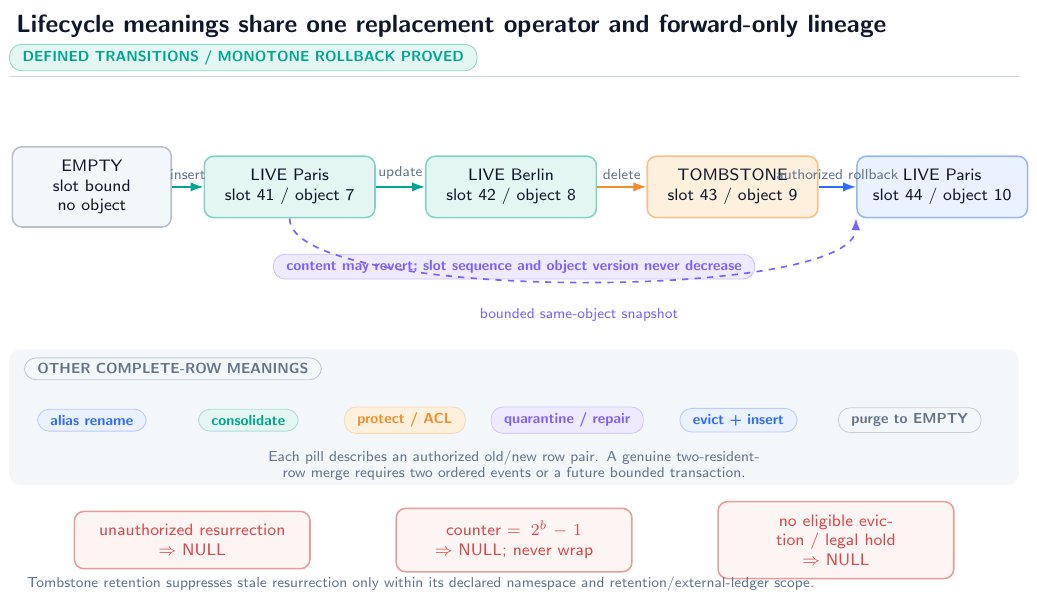}
\caption{\textbf{Lifecycle states share one replacement operator and forward-only lineage.} Delete commits a tombstone; rollback restores bounded same-object content as a new version; purge is a later authorized event. The red paths are bounded rejection or overflow, not ordinary data flow.}
\label{tf:amr:fig:lifecycle}
\end{figure}

\subsection{Two-row merge limitation}

\tfAmrProvedStatus
\begin{tfAmrProposition}[One event cannot implement a genuine two-row merge]
\label{tf:amr:prop:merge-limit}
Suppose a merge specification requires both (i) row $a$ to contain a combined object and (ii) distinct resident row $b\neq a$ to cease being live in the same atomic outcome. No transition of Eq.~\ref{tf:amr:eq:event-transition} implements that specification.
\end{tfAmrProposition}

\begin{proof}
A non-\tfAmrNull{} transition changes exactly one target row. If it changes $a$, row $b$ is untouched and remains live; if it changes $b$, row $a$ cannot receive the combined object. \tfAmrNull{} changes neither. Thus the required two-row postcondition is outside the event transition set.
\end{proof}

Two ordered events can express ``write combined object, then tombstone duplicate,'' but a crash or rejection between them leaves a visible prefix. A future transaction can make them all-or-nothing only by adding an explicitly bounded multi-row protocol. Calling one-row consolidation a merge does not remove this limitation.

%% file: theory_family/modules/amr/sections/08_invariants_results.tex
\section{Invariants and Preservation Results}
\label{tf:amr:sec:invariants}

\subsection{Assumption ledger}
\label{tf:amr:sec:assumptions}

The results are conditional. They do not assert that a deployed system satisfies these premises.

\begin{tfAmrAssumption}[Valid genesis]
\label{tf:amr:asm:valid-genesis}
The initial table $M_0$ contains exactly $S$ parseable schema-$\sigma$ rows, each in the appropriate empty normal form with valid tenant/slot bindings and receipts.
\end{tfAmrAssumption}

\begin{tfAmrAssumption}[Trusted computing boundary]
\label{tf:amr:asm:trusted-boundary}
The assembler, authentication context, policy engine, key material, canonicalizer, and commit coordinator are not controlled by the neural proposer. Their code may still contain bugs; this assumption excludes adversarial replacement of the trusted functions in the formal model.
\end{tfAmrAssumption}

\begin{tfAmrAssumption}[Deterministic interpretation]
\label{tf:amr:asm:deterministic}
Serialization, parsing, canonicalization, quantization, the AMR-1 encoder, dual-view validation, policy evaluation, and base-index derivation are deterministic for their recorded schema/build identifiers.
\end{tfAmrAssumption}

\begin{tfAmrAssumption}[Snapshot-consistent validation]
\label{tf:amr:asm:snapshot}
Every local and global check is evaluated against the same table snapshot and authenticated context from which versions and predecessor digests are computed.
\end{tfAmrAssumption}

\begin{tfAmrAssumption}[Serializable publication]
\label{tf:amr:asm:serializable}
Accepted events for one tenant are placed in a total order, or an equivalent tenant-epoch compare-and-swap rejects a stale snapshot. Global uniqueness checks and their publication share that order.
\end{tfAmrAssumption}

\begin{tfAmrAssumption}[Atomic durable publication]
\label{tf:amr:asm:atomic-durable}
A storage transaction exposes and, after recovery, returns either the complete old table root or the complete new table root, with the corresponding required audit-root binding. Derived indexes may lag and are rebuilt from validated rows.
\end{tfAmrAssumption}

\begin{tfAmrAssumption}[Cryptography]
\label{tf:amr:asm:crypto}
$H$ is collision resistant and the authentication scheme is unforgeable under chosen-message attack for parties without the relevant key. Key identifiers, rotations, and revocations are authenticated.
\end{tfAmrAssumption}

\begin{tfAmrAssumption}[Static tenant partition]
\label{tf:amr:asm:static-partition}
$P$ does not change during ordinary event execution, and the read observation of tenant $t$ projects only slots in $P^{-1}(t)$ after current-row authorization.
\end{tfAmrAssumption}

\begin{tfAmrAssumption}[Logical observation model]
\label{tf:amr:asm:logical-observation}
Tenant noninterference excludes timing, cache occupancy, shared accelerator contention, global scheduling, shared-model memorization, allocator leakage, and malicious privileged administrators.
\end{tfAmrAssumption}

\begin{tfAmrAssumption}[External-ledger contract]
\label{tf:amr:asm:external-ledger}
When policy requires an archive or audit record, publication occurs only after that record is durably prepared and its fixed digest/root is bound into the commit. If capacity or durability fails, the authoritative result is \tfAmrNull{}.
\end{tfAmrAssumption}

\subsection{Invariant family}

The contract is layered rather than treating every obligation as a row predicate:
\begin{equation}
\operatorname{Inv}_\sigma(M,h,\mathsf{obs})=
\operatorname{Inv}^{\mathrm{row}}_\sigma(M)\land
\operatorname{Inv}^{\mathrm{global}}_\sigma(M)\land
\operatorname{Inv}^{\mathrm{history}}_\sigma(h)\land
\operatorname{Inv}^{\mathrm{obs}}_\sigma(M,h,\mathsf{obs}).
\label{tf:amr:eq:layered-invariant}
\end{equation}
Row invariants are decidable from one parsed row; global invariants quantify over the table snapshot; history invariants constrain authenticated predecessor and event sequences; observational invariants constrain what typed reads, indexes, and tenant projections may disclose. The numbered clauses below state the concrete obligations. I1--I7 and I9--I12 are row-local; I2 and I13 are global; I8, I10--I12, and I14 constrain history; I6--I7 and I15--I16 constrain observations. Overlap is intentional because a receipt field can be locally well typed while its lineage or disclosure use is invalid.

Let $\operatorname{Inv}_\sigma(M)$ abbreviate the conjunction of all four layers for the current authenticated history and observation contract:
\begin{enumerate}[label=\textbf{I\arabic*.}]
  \item \textbf{Fixed shape.} $|M|=S$, every row is $B_R$ bytes, parses under $\sigma$, and has zero reserved bits.
  \item \textbf{Slot and tenant binding.} Row $i$ declares slot $i$ and tenant $P(i)$.
  \item \textbf{Status normal form.} \tfAmrEmpty{}, \tfAmrLive{}, \tfAmrTomb{}, and \tfAmrQuar{} obey the field constraints of Section~\ref{tf:amr:sec:schema}.
  \item \textbf{Canonical bounds.} Type, length, alias, normalization, padding, and missing-value encodings are valid.
  \item \textbf{Neural finiteness.} Every coordinate is in the declared finite domain and forbidden encodings are absent.
  \item \textbf{Dual-view conformance.} Every readable \tfAmrLive{} row passes its declared mechanical relation and required validation bits.
  \item \textbf{Security integrity.} Tenant, owner, ACL, deletion authority, and protection fields were system assembled and parse uniquely.
  \item \textbf{Version discipline.} Every accepted predecessor relation follows Eqs.~\ref{tf:amr:eq:slot-version}--\ref{tf:amr:eq:object-version}; no counter wraps.
  \item \textbf{Protection discipline.} Protection is not weakened without a bound authenticated capability.
  \item \textbf{Provenance completeness.} Every non-genesis row records the bounded event/source/build/operation fields required by its schema.
  \item \textbf{Receipt validity.} The core checksum, predecessor digest, validation bitmap, algorithm IDs, event binding, and authentication tag verify.
  \item \textbf{Rollback bounds.} Lineage has at most $H_{\mathrm{rb}}$ nonrecursive same-object snapshots and obeys current tenant scope.
  \item \textbf{Identity uniqueness.} Within a tenant, no two \tfAmrLive{} or retained \tfAmrTomb{} rows claim the same nonzero object identifier or primary typed key.
  \item \textbf{Tombstone discipline.} A tombstone cannot become \tfAmrLive{} without explicit resurrection/rollback authority; purge obeys retention and legal hold.
  \item \textbf{Read discipline.} Only a current \tfAmrLive{} row passing receipt, conformance, tenant, ACL, validity, and protection checks may supply active content.
  \item \textbf{Derived-view discipline.} An index or cache can return only a hint that is revalidated against the current row; it cannot change authority.
\end{enumerate}

\subsection{Preservation}

\tfAmrProvedStatus
\begin{tfAmrTheorem}[Invariant preservation]
\label{tf:amr:thm:invariant-preservation}
Under Assumptions~\ref{tf:amr:asm:valid-genesis}--\ref{tf:amr:asm:serializable}, if $\operatorname{Inv}_\sigma(M)$ holds and $M'$ is produced by a legal event transition, then $\operatorname{Inv}_\sigma(M')$ holds. Every finite sequence of legal transitions from genesis therefore preserves the invariant family.
\end{tfAmrTheorem}

\begin{proof}
If the action is \tfAmrNull{}, $M'=M$ and every layer is unchanged. Otherwise exactly target row $a$ is replaced. For $i\neq a$, bytes are identical, so every row-local invariant persists. The trusted assembler fixes slot and tenant, normalizes and bounds canonical content, verifies finite neural encodings and dual-view conformance, applies status normal form, computes versions without overflow, enforces ACL/deletion/protection policy, constructs bounded same-object rollback, binds provenance, zeros reserved bytes, and authenticates the complete core. Thus the row layer, including I1--I7 and I9--I12, holds for the candidate.

Identity/key uniqueness and any strict alias rule are evaluated against the same snapshot used for candidate assembly. Serializable publication or tenant-epoch compare-and-swap prevents a conflicting interleaving between the global check and publication, preserving the global layer and I13. The authenticated predecessor normal form, forward version rule, and explicit resurrection predicates append a legal history edge, preserving the history layer and I8, I10--I12, and I14. Typed reads and derived views revalidate the exact current row and cannot change authority, preserving the observational layer and I15--I16. Therefore all four layers hold in $M'$. Induction on the finite event sequence proves the second claim.
\end{proof}

\tfAmrProvedStatus
\begin{tfAmrCorollary}[Version monotonicity]
\label{tf:amr:cor:version-monotonicity}
Along every accepted predecessor chain for one physical slot, slot sequence is strictly increasing until exhaustion. Along a resident same-object chain, object version is strictly increasing. \tfAmrNull{} leaves both unchanged.
\end{tfAmrCorollary}

\begin{proof}
The assembler rules are invariant I8 and are preserved by Theorem~\ref{tf:amr:thm:invariant-preservation}. A commit applies $+1$ and rejects overflow; \tfAmrNull{} is identity.
\end{proof}

\tfAmrProvedStatus
\begin{tfAmrCorollary}[Untouched-row preservation]
\label{tf:amr:cor:untouched}
For event target $a$, every row $i\neq a$ is byte-identical before and after the event. Across a segment, every slot outside the union of committed targets is byte-identical between $M_{n-1}$ and $M_n$.
\end{tfAmrCorollary}

\begin{proof}
The first statement is Eq.~\ref{tf:amr:eq:replace}. Composition can change only a row selected by at least one event, yielding the second statement.
\end{proof}

\subsection{Finite versions and deletion history}

\tfAmrProvedStatus
\begin{tfAmrTheorem}[Finite monotone-version exhaustion]
\label{tf:amr:thm:version-exhaustion}
A $b_s$-bit slot sequence initialized at zero and required to increase strictly on every commit permits at most $2^{b_s}-1$ commits before the next commit must be rejected or the namespace must be migrated.
\end{tfAmrTheorem}

\begin{proof}
There are $2^{b_s}$ representable values. A strictly increasing sequence starting at zero can visit each at most once. After $2^{b_s}-1$, no larger value exists in the representation.
\end{proof}

Silent wraparound would make stale receipts and current versions indistinguishable. An epoch-renewal protocol may extend lifetime, but its epoch is additional authoritative state and must itself be bounded, authenticated, and migrated.

\tfAmrProvedStatus
\begin{tfAmrTheorem}[Bounded deletion-memory impossibility]
\label{tf:amr:thm:deletion-impossibility}
A system with $B$ total authoritative bits cannot, without external authoritative state, distinguish forever all deletion histories over an unbounded object universe.
\end{tfAmrTheorem}

\begin{proof}
Choose distinct objects $o_1,o_2,\ldots$ and let $M_j$ be the authoritative state after deleting the first $j$ objects. Among $M_0,\ldots,M_{2^B}$, two states coincide by the pigeonhole principle: $M_p=M_q$ for some $p<q$. Object $o_{p+1}$ is deleted in the history producing $M_q$ but not in the history producing $M_p$. A deterministic query over identical final authoritative states cannot answer differently.
\end{proof}

The coherent guarantee is therefore scoped: a finite namespace, a bounded tombstone set, a retention window, or an external deletion ledger. ``Permanent deletion memory for arbitrarily many identities at fixed total cost'' is not a satisfiable specification.

\subsection{Receipt tamper evidence}

\tfAmrProvedStatus
\begin{tfAmrTheorem}[Receipt tamper evidence]
\label{tf:amr:thm:receipt-tamper}
Fix the expected current validation context, including namespace epoch, slot, slot sequence/version, authorization and event bindings, and freshness checks. Under Assumption~\ref{tf:amr:asm:crypto}, an adversary without the authentication key who changes any receipt-covered core bit of a valid row while retaining that context can make the changed row verify only by finding a hash collision or forging the authentication scheme.
\end{tfAmrTheorem}

\begin{proof}
If the altered core has the old digest, the adversary found a collision. If its digest changes, the old authentication tag no longer authenticates the receipt message and acceptance requires a valid tag on a new message, which is a forgery. The independently checked context binds the row to the expected current identity and freshness state.
\end{proof}

Substitution or replay of an intact previously authenticated row is not a hash collision or forgery: that row can verify against its own receipt. It must instead fail the expected-current-context or freshness checks. The theorem therefore authenticates changed bytes and declared lineage only within a validated context. It does not authenticate real-world truth, protect a compromised key, or prove the correctness of the trusted code that created the receipt.

%% file: theory_family/modules/amr/sections/09_security_recovery.tex
\section{Security, Index, Concurrency, and Recovery Contract}
\label{tf:amr:sec:security-recovery}

\subsection{Tenant and principal scope}

For authenticated context $\chi$ with tenant $t$, the feasible targets satisfy
\begin{equation}
\tfAmrCalA_t(M,\chi)\subseteq P^{-1}(t)\cup\{\varnothing\}.
\label{tf:amr:eq:tenant-targets}
\end{equation}
The assembler additionally requires the candidate tenant to equal $P(a)=t$. A tenant cannot evict another tenant's row to obtain capacity. Static quotas can waste space, but they make the logical isolation statement auditable.

Within a tenant, each ACL entry contains a principal identifier and rights mask. A reference set is
\begin{equation}
\{\textsf{READ},\textsf{WRITE},\textsf{DELETE},\textsf{PROTECT},
\textsf{ROLLBACK},\textsf{ACL\_ADMIN}\}.
\label{tf:amr:eq:acl-rights}
\end{equation}
ACL changes require \textsf{ACL\_ADMIN}; delete and rollback require their own rights. A general \textsf{WRITE} bit does not authorize security weakening or resurrection.

Formal noninterference traditionally asks whether one domain's behavior can affect another domain's observation \cite{tf:amr:goguen1982noninterference}. Here the observation model is intentionally narrow.

\tfAmrProvedStatus
\begin{tfAmrTheorem}[Logical tenant noninterference]
\label{tf:amr:thm:tenant-noninterference}
Under Assumptions~\ref{tf:amr:asm:serializable}, \ref{tf:amr:asm:static-partition}, and \ref{tf:amr:asm:logical-observation}, for $t\neq t'$ and a legal event issued under tenant $t$,
\begin{equation}
\operatorname{Obs}_{t'}(T_t(M))=\operatorname{Obs}_{t'}(M).
\label{tf:amr:eq:tenant-noninterference}
\end{equation}
\end{tfAmrTheorem}

\begin{proof}
Every feasible target of tenant $t$ lies in $P^{-1}(t)$. Static partitions are disjoint, so no row in $P^{-1}(t')$ changes. The observation of $t'$ projects only current authorized rows in its partition and rejects any index entry whose tenant/current-row binding does not match. Therefore its logical observation is unchanged.
\end{proof}

This theorem excludes timing, shared caches and accelerators, global backpressure, traffic analysis, shared-model memorization, and malicious administrators. A production confidentiality claim needs a stronger observation model and empirical side-channel analysis.

\subsection{Derived indexes and stale-value suppression}

Every base index is a deterministic function $I=D_\sigma(M)$. An entry is a hint containing
\begin{equation}
(\text{tenant},\text{object},\text{key-digest},a,v_s,d_\rho,\text{projection}).
\label{tf:amr:eq:index-hint}
\end{equation}
A lookup accepts the hit only if the current row at $a$ has matching tenant, object, slot sequence, readable status, and receipt digest, then passes the ordinary receipt, dual-view, ACL, validity, and protection checks.

\tfAmrProvedStatus
\begin{tfAmrTheorem}[Stale-index safety]
\label{tf:amr:thm:stale-index}
An arbitrarily stale derived index cannot make an obsolete, deleted, replaced, or cross-tenant row acceptable when every hit is revalidated by Eq.~\ref{tf:amr:eq:index-hint} and the current-row read predicate.
\end{tfAmrTheorem}

\begin{proof}
If the row has not changed, validation reduces to the ordinary current read. If it has changed, then either the slot sequence, object identity, status, or receipt digest differs and the hit is rejected. A cross-tenant entry fails tenant/partition validation. A retained tombstone fails readable status. Thus a stale entry cannot authorize stale content.
\end{proof}

A stale or missing index can cause a false negative, fallback scan, extra work, or denial of service. It is safe only in the authority sense, not necessarily available or efficient. AMR-1 updates nine canonical entries and one neural entry after a row commit, for exactly
\begin{equation}
9\times112+336=1{,}344\ \text{logical index bytes}.
\label{tf:amr:eq:index-update-bytes}
\end{equation}
If the update is lost, rebuild from validated rows restores the deterministic projection.

\subsection{Serial execution and linearization}

Database transactions provide the classical all-or-nothing abstraction, while linearizability relates concurrent operations to a legal real-time-respecting sequential history \cite{tf:amr:gray1981transaction,tf:amr:herlihy1990linearizability}. \tfAmrName{} adopts a narrower one-row event contract.

\tfAmrDefinitionStatus
\begin{tfAmrDefinition}[Commit linearization point]
The linearization point of an accepted event is the atomic publication of the new authenticated table root and required audit-root binding. A \tfAmrNull{} event linearizes at the final validation decision and leaves the root unchanged.
\end{tfAmrDefinition}

\tfAmrProvedStatus
\begin{tfAmrTheorem}[Serializable event history]
\label{tf:amr:thm:serializable}
Under snapshot-consistent validation and serializable publication, every finite concurrent execution of accepted events is observationally equivalent to a sequential execution ordered by their successful publication points, with rejected stale attempts represented as \tfAmrNull{}.
\end{tfAmrTheorem}

\begin{proof}
The commit coordinator assigns each successful publication a unique position in a tenant-total order (or a stronger global order where uniqueness spans tenants). A candidate whose snapshot epoch is no longer current fails compare-and-swap and produces no authoritative mutation. Consider successful publications in their assigned order. Each transition was validated against the predecessor root it replaces and changes one complete row; therefore applying those transitions sequentially yields exactly the published roots. Reads validate against one published root and can be placed after the root publication they observe. The resulting sequential history respects successful publication order and has the same observations.
\end{proof}

The theorem would fail if two rows could pass a uniqueness check on different snapshots and publish independently. Per-row atomicity is therefore insufficient for memory-global uniqueness unless the uniqueness constraint has a serializable authority of its own.

\subsection{Crash-safe publication protocol}

ARIES and related recovery systems show that crash recovery needs an explicit log and ordering discipline rather than the word ``atomic'' alone \cite{tf:amr:mohan1992aries}. The following is an abstract obligation, not an implementation claim.

\begin{tfAmrProtocol}[Copy-on-write root publication]
\label{tf:amr:prot:cow-publication}
For current root $R$ and candidate row:
\begin{enumerate}
  \item allocate a new immutable row/page and any required audit record; compute their digests;
  \item construct a new immutable table root $R'$ pointing to the new row and binding the required audit-ledger sequence/root;
  \item durably write and verify the new row, audit record, and $R'$ in an order that prevents a root from referencing unprepared data;
  \item atomically publish one checksummed root selector from $R$ to $R'$;
  \item update derived indexes after publication or rebuild them later;
  \item on recovery, select the latest root whose selector, table, rows, and required audit binding validate; otherwise retain the prior valid root.
\end{enumerate}
\end{tfAmrProtocol}

The root selector, durability primitive, sector/page atomicity, flush/fence ordering, replica protocol, and failure model are deployment parameters. A write-ahead implementation may satisfy the same abstract old-or-new condition with different steps.

\tfAmrProvedStatus
\begin{tfAmrTheorem}[Old-or-new crash recovery]
\label{tf:amr:thm:crash-recovery}
Assume Protocol~\ref{tf:amr:prot:cow-publication} is implemented correctly; durable writes do not acknowledge before reaching the stated failure domain; the root selector update is atomic; and recovery accepts only fully validated roots. After any single crash during one commit, recovery exposes either the complete precommit table and audit binding or the complete postcommit table and audit binding, never a readable mixture.
\end{tfAmrTheorem}

\begin{proof}
Before the selector publication, the old selector still names $R$, whose referenced objects remain immutable and valid; unreferenced prepared objects are garbage after recovery. After selector publication, the protocol guarantees that every object reachable from $R'$ was durably prepared. If the selector itself is torn or invalid, its checksum/redundancy causes recovery to choose the prior valid selector; if it is valid, recovery validates all reachable objects before accepting $R'$. Thus the recovered root is $R$ or $R'$. Because a row is immutable and both views are inside it, neither root contains a torn canonical/neural or control/content combination.
\end{proof}

\tfAmrUnvalidatedStatus\quad No storage engine, filesystem, device, replica set, flush sequence, or crash injector is evaluated in this theoretical part. The theorem describes what an implementation must establish.

\begin{figure}[H]
\centering
\includegraphics[width=\linewidth]{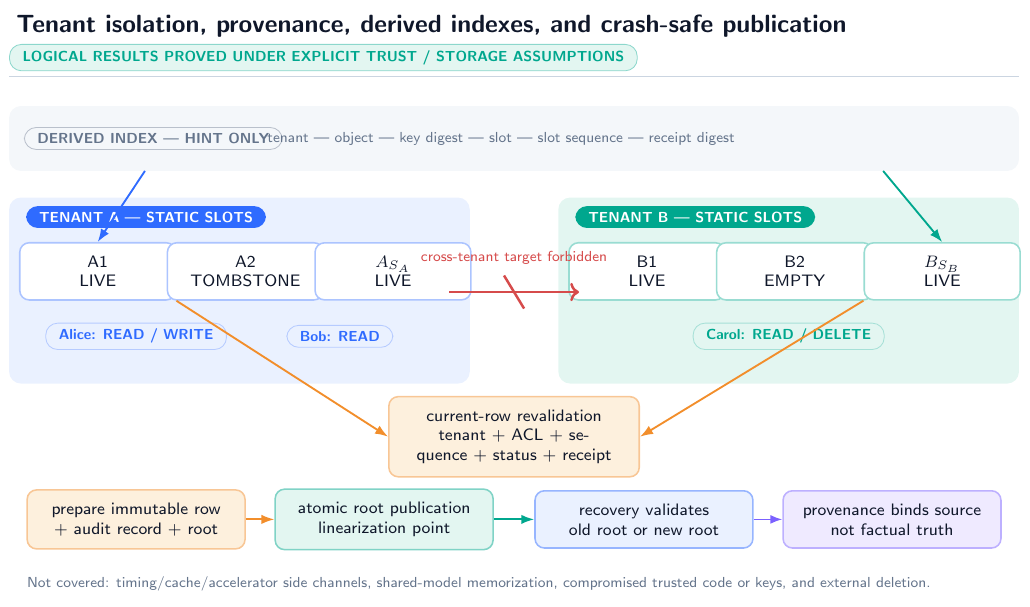}
\caption{\textbf{Static tenant partitions, system-owned security, and current-row revalidation define the logical boundary.} Indexes are hints, provenance binds declared sources, and publication binds the row to an audit root. The proved path excludes side channels and depends on authenticated context, serializable checks, and crash-safe storage assumptions.}
\label{tf:amr:fig:tenant-security}
\end{figure}

\subsection{Threat model}

The model includes malicious or erroneous neural proposals, replayed stale candidates, malformed serialized fields, alias collisions, cross-tenant target attempts, unauthorized deletion or rollback, storage bit corruption, stale indexes, interrupted commits, and parties without receipt keys attempting row tampering.

It does not defeat a compromised trusted assembler, policy engine, authentication service, signing/MAC key, storage administrator, firmware, or validator build. It does not cover covert channels, denial of service through shared compute, model-parametric memorization, source truth, or complete erasure from backups. The system can fail closed under many of those conditions, but availability and recovery may then require external administration.

%% file: theory_family/modules/amr/sections/10_costs.tex
\section{Complete Cost Calculus}
\label{tf:amr:sec:costs}

\subsection{Why one scalar hides the contract}

Candidate construction mixes neural inference, byte traffic, comparisons, policy checks, cryptography, storage barriers, and index work. We therefore report a work vector
\begin{equation}
\mathcal W=(B_{\mathrm{rd}},B_{\mathrm{wr}},N_H,N_{\mathrm{authV}},N_{\mathrm{authC}},
N_E,N_D,N_{\mathrm{pol}},N_{\mathrm{cmp}},N_{\mathrm{barrier}}),
\label{tf:amr:eq:work-vector}
\end{equation}
where the coordinates count logical bytes read/written, hash compression blocks, authentication verifications/creations, encoder and dual-view validator calls, policy evaluations, identity/key comparisons, and durable ordering barriers. Model FLOPs, accelerator traffic, kernel launches, wall time, replication, and energy are additional measured coordinates, not constants hidden in $O(1)$.

For a complete deployment, define:
\begin{itemize}
  \item $C_{\mathrm{event}}$: bounded segment eventization cost;
  \item $C_G$: neural proposal cost per target-conditioned candidate;
  \item $C_A$: trusted normalization, assembly, and local validation;
  \item $C_D$: dual-view derivation/certification;
  \item $C_R$: target routing or shortlist construction, including misses and excluded actions;
  \item $C_U$: memory-global uniqueness/protection checks;
  \item $C_Q$: action scoring after candidates exist;
  \item $C_P$: durable row/audit publication;
  \item $C_I$: index projection, update, rebuild, and lookup validation;
  \item $C_X$: archive write, lookup, retrieval, retention, and deletion work; and
  \item $C_{\mathrm{audit}}$: audit append/checkpoint, authentication, and retention work.
\end{itemize}
No end-to-end bound may report only row-read cost while omitting $C_G$, $C_A$, or the candidate multiplicity used by routing.

\subsection{General candidate accounting}

Let $B_P$ be proposal bytes, $B_C$ canonical bytes scanned, $B_D$ neural route/payload bytes compared, $B_{\mathrm{core}}=B_R-B_\rho$, $S_t$ tenant slots, $B_{\mathrm{ctrl}}$ trusted-control bytes scanned per row, and $B_{\mathrm{key}}$ primary-key bytes scanned per row. A conservative exact logical safe path that reads the old row, reads the proposal, normalizes canonical bytes, materializes and compares the neural view, hashes the complete core, and scans the tenant for object/key uniqueness has
\begin{equation}
\begin{aligned}
B_{\mathrm{rd}}^{\mathrm{cand}}={}&B_R+B_P+B_C+2B_D+B_{\mathrm{core}}\\
&+S_t(B_{\mathrm{ctrl}}+B_{\mathrm{key}}),\\
B_{\mathrm{wr}}^{\mathrm{cand}}={}&B_R+B_D.
\end{aligned}
\label{tf:amr:eq:candidate-traffic}
\end{equation}
This is protocol traffic, not a prediction of cache misses: an implementation may fuse passes or reuse a cache line, but must report the measured physical traffic separately.

For a Merkle--Damg\aa rd-style hash whose compression block is $r_H$ bytes, domain prefix is $B_{\mathrm{dom}}$ bytes, and padding consumes one delimiter plus an eight-byte length field, the receipt core needs
\begin{equation}
N_H=\left\lceil\frac{B_{\mathrm{dom}}+B_{\mathrm{core}}+9}{r_H}\right\rceil
\label{tf:amr:eq:hash-blocks}
\end{equation}
compression blocks. SHA-256's block structure supplies the illustrative AMR-1 numbers \cite{tf:amr:nist2015sha}; another profile must recalculate them.

One candidate has abstract cost
\begin{equation}
\begin{aligned}
C_{\mathrm{cand}}={}&C_G+C_{\mathrm{normalize}}(B_C)+C_D+C_{\mathrm{pol}}+C_U\\
&+N_HC_{H\text{-block}}+C_{\mathrm{source-auth}}+C_{\mathrm{receipt-auth}}\\
&+C_{\mathrm{traffic}}(B_{\mathrm{rd}}^{\mathrm{cand}},B_{\mathrm{wr}}^{\mathrm{cand}}).
\end{aligned}
\label{tf:amr:eq:candidate-cost}
\end{equation}
Rejecting after step $q$ costs the prefix actually executed. A schema-length rejection can occur before encoder and global scans; a late duplicate-key or authentication failure can incur almost the entire candidate cost while still producing zero authoritative row bytes.

\subsection{AMR-1 candidate path}

For AMR-1,
\begin{equation}
\begin{gathered}
B_R=6{,}688,\quad B_P=1{,}888,\quad B_C=1{,}104,\quad
B_D=768,\\
B_{\mathrm{core}}=6{,}432,\quad
B_{\mathrm{ctrl}}+B_{\mathrm{key}}=256+66=322.
\end{gathered}
\label{tf:amr:eq:amr1-cost-constants}
\end{equation}
The proposal is exactly the 1,104-byte canonical block plus 784-byte neural block. The dual comparison covers only the 768 route/payload bytes; the 16 neural metadata/reserved bytes are separately validated.

With $S_t=256$, the authoritative uniqueness scan reads
\begin{equation}
256\times322=82{,}432\ \text{bytes}.
\label{tf:amr:eq:uniqueness-read}
\end{equation}
The remaining local reads are
\begin{equation}
6{,}688+1{,}888+1{,}104+2(768)+6{,}432=17{,}648\ \text{bytes},
\end{equation}
so
\begin{equation}
B_{\mathrm{rd}}^{\mathrm{cand}}=100{,}080\ \text{bytes}.
\label{tf:amr:eq:amr1-read-total}
\end{equation}
Materializing the complete candidate and one 768-byte encoder scratch buffer writes
\begin{equation}
B_{\mathrm{wr}}^{\mathrm{cand}}=6{,}688+768=7{,}456\ \text{bytes}.
\label{tf:amr:eq:amr1-write-total}
\end{equation}
With a 16-byte domain prefix and 64-byte SHA-256 compression blocks,
\begin{equation}
N_H=\left\lceil\frac{16+6{,}432+9}{64}\right\rceil=101.
\label{tf:amr:eq:amr1-hash-blocks}
\end{equation}

The reference operation ledger is therefore one neural proposal, one canonical normalization, one deterministic encoder call, one policy evaluation, 256 object/key comparisons, 101 hash compression blocks, one source-authentication verification, one receipt-authentication creation, 100,080 specified precommit read bytes, and 7,456 specified precommit write bytes. A strict alias-uniqueness scan adds
\begin{equation}
256\times524=134{,}144\ \text{read bytes}
\end{equation}
and at most $8\times256$ alias comparisons.

On rejection, authoritative publication bytes are zero. On acceptance, logical publication adds one 6,688-byte row and one 1,344-byte fixed index projection. The row may be prepared and written more than once physically under journaling, copy-on-write, or replication; that amplification is measured rather than inferred.

\subsection{Routing and candidate multiplicity}

If event $j$ constructs $N_{n,j}$ candidates, its construction/validation cost is
\begin{equation}
\sum_{a\in\mathcal C_{n,j}} C_{\mathrm{cand}}(a),
\qquad |\mathcal C_{n,j}|=N_{n,j}.
\label{tf:amr:eq:candidate-multiplicity}
\end{equation}
Exhaustive target conditioning gives $N_{n,j}=S_t$. A shortlist of size $L$ may reduce this term only if $C_R$, index build/update/storage, shortlist misses, and omitted feasible targets are reported. Scoring $S_t$ cheap target IDs while constructing only the winning row is not equivalent to the complete candidate set assumed by a controller whose action values depend on target-conditioned content.

For one segment,
\begin{equation}
C_{\mathrm{segment}}=C_{\mathrm{event}}+
\sum_{j=1}^{K_n}\left(C_R+\sum_{a\in\mathcal C_{n,j}}C_{\mathrm{cand}}(a)+C_Q+C_{\mathrm{selected\ publication}}\right).
\label{tf:amr:eq:segment-cost}
\end{equation}
The exact authoritative row traffic is $C_nB_R$ bytes. It is not $K_nB_R$ when events return \tfAmrNull{}, and it does not count candidate buffers that never publish.

\subsection{Re-encoding, archive, index, and audit costs}

A refresh or re-encoding of $N_{\mathrm{refresh}}$ resident rows is $N_{\mathrm{refresh}}$ ordinary versioned candidates and commits. Its cost includes reading canonical content, running the new encoder, validating semantic gates, advancing versions, updating rollback, writing rows, and rebuilding index projections. There is no free in-place vector migration.

Let a commit emit $B_A^+$ archive bytes and $B_L^+$ audit bytes. Its full durable write volume includes at least
\begin{equation}
B_{\mathrm{durable}}^+=B_R+B_A^++B_L^++B_{\mathrm{root}}^++B_{\mathrm{replication}}^++B_{\mathrm{journal}}^+,
\label{tf:amr:eq:durable-bytes}
\end{equation}
plus ordering barriers and verification reads. Terms may be zero only when the profile explicitly does not require the corresponding artifact. Archive retrieval later incurs index lookup, storage reads, authentication, decompression/decoding, and candidate reconstruction; it cannot be counted only at write time.

An audit ledger with maximum $L_{\max}$ entries of fixed receipt size $B_{\ell}$ has capacity $L_{\max}B_{\ell}$ plus its bounded index/checkpoint. When full, the system fails writes whose contract requires audit, or executes a separately specified authenticated rotation that commits a checkpoint root before releasing old detail. An append-only ledger with no $L_{\max}$ is explicitly unbounded total storage.

\begin{figure}[H]
\centering
\includegraphics[width=\linewidth]{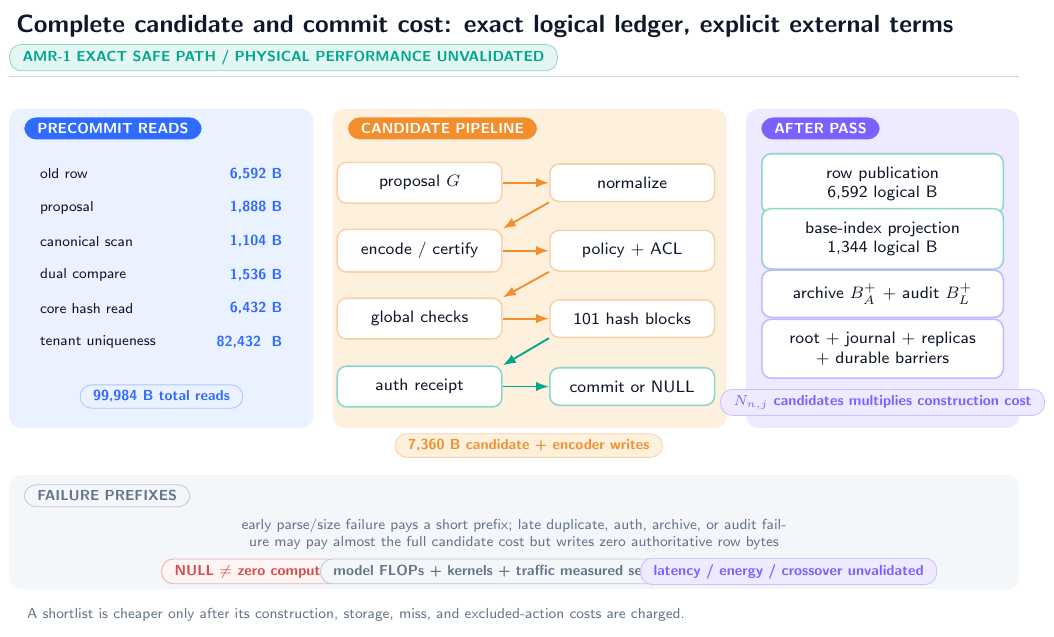}
\caption{\textbf{A complete candidate and commit have a countable bill.} The AMR-1 waterfall gives exact logical safe-path traffic for one candidate. Routing multiplicity, global checks, publication, index projection, archive, audit, replication, barriers, model FLOPs, latency, and energy remain visible rather than disappearing into a row-count asymptotic.}
\label{tf:amr:fig:cost-path}
\end{figure}

\subsection{Failure-path ledger}

\begin{table}[H]
\centering
\footnotesize
\begin{tabularx}{\linewidth}{p{0.22\linewidth}p{0.24\linewidth}YY}
\toprule
Failure & Latest required work & Authoritative writes & Remaining cost/risk \\
\midrule
Parse/size/type failure & proposal read + bounded parse & 0 & diagnostic write if enabled \\
Tenant/ACL/protection denial & old control + policy/auth checks & 0 & possible authentication and rate-limit cost \\
Dual-view failure & normalization + encoder/validator & 0 & optional bounded quarantine copy \\
Duplicate identity/key & tenant scan or validated uniqueness index & 0 & nearly complete candidate cost \\
Version overflow/stale snapshot & version/snapshot check; may be early or late & 0 & retry/re-key/migration external \\
Archive/audit capacity failure & archive/audit reservation and policy checks & 0 & cleanup of unreferenced prepared data \\
Crash before root publication & durable preparation may have occurred & 0 visible & recovery/garbage collection \\
Crash after root publication & full commit durable by assumption & one row visible & index rebuild may remain \\
Stale index & lookup + current-row mismatch & 0 & fallback scan or false negative \\
Receipt mismatch on read & row read + cryptographic validation & 0 by read & fail-closed unavailability and recovery \\
\bottomrule
\end{tabularx}
\caption{Failure costs. \tfAmrNull{} means zero authoritative mutation, not zero computation or zero external logging.}
\label{tf:amr:tab:failure-costs}
\end{table}

\tfAmrUnvalidatedStatus\quad AMR-1's byte ledger is exact only at the logical protocol layer under the declared AMR-1 profile. It does not predict cache behavior, latency, throughput, energy, device endurance, or quality--cost crossover. Those require an implementation and the stop-ruled measurements of Section~\ref{tf:amr:sec:falsification}.

%% file: theory_family/modules/amr/sections/11_counterexamples.tex
\section{Counterexamples and Failure Taxonomy}
\label{tf:amr:sec:counterexamples}

The contract's assumptions are not decorative. Each prevents a concrete violation.

\subsection{Unbounded aliases and provenance defeat boundedness}

\begin{tfAmrCounterexample}[Hidden variable-width row]
Suppose a row stores a list $A(r)$ of every spelling ever used for its object, with no alias-count or length cap. After the $m$th rename, append a fresh one-byte-distinct alias. Then $|A(r)|\ge m$ and row storage grows with history even when $S=1$. Likewise, storing the complete provenance document or all ancestors recursively makes $B_p$ or $B_\lambda$ history dependent. The statement ``one fixed slot'' no longer implies a bit bound.
\end{tfAmrCounterexample}

Using a fixed digest pointer repairs resident boundedness only by moving the history to an external object. The total system remains unbounded unless that archive is independently capped.

\subsection{Neural ownership of trusted fields breaks isolation and lineage}

\begin{tfAmrCounterexample}[Self-issued authority]
Let the assembler copy proposed tenant, ACL, and version bits without independent derivation. A malicious proposer targeting a tenant-A slot can emit tenant B, grant its principal \textsf{READ}/\textsf{DELETE}, or reset version from 42 to 1. The serialized row may be well formed and its neural/canonical views may agree, yet tenant isolation, authorization, monotonicity, stale suppression, and rollback lineage all fail. A checksum computed over these bytes merely authenticates the bad assignment.
\end{tfAmrCounterexample}

The remedy is functional system ownership: trusted fields are computed from authenticated context, current state, and policy, never self-certified by the proposal.

\subsection{In-place neural refresh creates dual-view drift}

\begin{tfAmrCounterexample}[Two independently mutable truths]
At version $u$, canonical bytes encode Paris and neural bytes equal $E_{\omega}(\text{Paris})$. A background process upgrades to $\omega'$ and overwrites only the neural region with $E_{\omega'}(\text{Berlin})$ or even $E_{\omega'}(\text{Paris})$ while leaving the receipt and version unchanged. The canonical and neural views no longer share one declared transformation and version. A canonical read and neural read can disagree, or stale cached validation can accept bytes produced by an undeclared model.
\end{tfAmrCounterexample}

Re-encoding must therefore be a complete new versioned event, and reads/cache entries must be keyed by the current receipt.

\subsection{Soft authoritative overwrite has no unique discrete lineage}

\begin{tfAmrCounterexample}[Versioned mixture ambiguity]
Let a payload for Paris be $v_P$, a candidate for Berlin be $v_B$, and set $v'=(1-\gamma)v_P+\gamma v_B$ for $0<\gamma<1$. If the canonical row remains Paris, the neural view no longer conforms to it. If canonical content becomes Berlin, the payload is not the declared Berlin encoding. If both canonical strings or ACLs are ``mixed,'' their interpolation is undefined. Assigning version 8 does not resolve which fact or security policy is authoritative.
\end{tfAmrCounterexample}

Soft state is permitted only under the separate type and non-authority rule of Eq.~\ref{tf:amr:eq:fast-state}.

\subsection{Multi-event conflation breaks atomic claims}

\begin{tfAmrCounterexample}[Segment rank-one and merge error]
A segment contains two events: update Alice in row 3 and update Bob in row 9. Both commits succeed with linearly independent payload changes. The segment difference has two nonzero rows and can have rank two. Calling the entire segment ``one atomic overwrite'' hides a partial-success boundary. Similarly, merging row 3 into row 9 while making row 3 non-live requires two resident changes and cannot be one event.
\end{tfAmrCounterexample}

The correct recursion is Eq.~\ref{tf:amr:eq:event-recursion}; per-event support is at most one and segment support is at most $K_n$.

\subsection{Per-row CAS does not preserve global uniqueness}

\begin{tfAmrCounterexample}[Write skew]
Two assemblers read the same snapshot in which typed key $k$ is absent. One prepares slot 7 and one prepares slot 8. Each uses a successful compare-and-swap only on its target row. Both commits can publish, leaving two live rows with key $k$. Every local row invariant holds; the memory-global uniqueness invariant fails.
\end{tfAmrCounterexample}

A tenant epoch, serializable uniqueness index, or equivalent global commit check is required. This counterexample explains why the serializability assumption is stronger than atomic row bytes.

\subsection{Failure taxonomy}

\begin{table}[H]
\centering
\footnotesize
\begin{tabularx}{\linewidth}{p{0.18\linewidth}p{0.24\linewidth}p{0.25\linewidth}Y}
\toprule
Class & Example & Required authoritative behavior & Unresolved consequence \\
\midrule
Syntax/bound & bad length, reserved bit, NaN, alias overflow & precommit \tfAmrNull{} & proposal cost, possible denial of service \\
Authentication & bad tenant, ACL, capability, source tag & precommit \tfAmrNull{} & compromised identity service out of model \\
Policy & protection/retention/legal-hold violation & precommit \tfAmrNull{} & policy correctness not proven \\
Semantic consistency & canonical/neural mismatch or \tfAmrUnknown{} & candidate \tfAmrNull{}; resident read $\bot$ & semantic validator errors unmeasured \\
Concurrency & stale epoch, uniqueness conflict & losing attempt \tfAmrNull{} or retry & starvation/availability unproven \\
Storage integrity & checksum/tag mismatch, torn page & fail closed; recover old/new root & recovery depends on implementation \\
Index & stale/missing/corrupt hint & reject hit; scan or rebuild & latency and availability degradation \\
Version/history & counter exhaustion, rollback ring full & reject/migrate or evict declared snapshot & finite lifetime/history loss \\
Capacity & no eligible slot, audit/archive full & \tfAmrNull{} unless explicit bounded rotation/eviction succeeds & memory stagnation or data loss trade-off \\
Adversarial truth & authenticated false source, poisoning & may still pass mechanical contract & truth/admission problem outside substrate \\
Key/build compromise & forged tag, malicious validator/assembler & outside theorem; revoke/recover administratively & rows may be untrustworthy \\
Side channel & timing/cache/traffic leakage & outside logical noninterference & separate security proof/test required \\
\bottomrule
\end{tabularx}
\caption{Failure taxonomy. Fail-closed behavior preserves authority but can sacrifice availability.}
\label{tf:amr:tab:failure-taxonomy}
\end{table}

\subsection{Abuse cases and non-guarantees}

\textbf{Poisoning.} An authenticated source can state a plausible falsehood whose canonical and neural views agree. Provenance enables attribution and later policy; it does not prove truth.

\textbf{Capacity denial.} An attacker can fill its tenant partition with low-value rows or retained tombstones. Static partitioning prevents cross-tenant eviction but not self-denial. Per-principal rate limits and protected reserve capacity are additional policies.

\textbf{Rollback abuse.} A valid old snapshot can still be obsolete or harmful. Rollback requires current authorization, tombstone-aware resurrection checks, and uniqueness validation; the substrate does not decide whether restoration is desirable.

\textbf{Key or validator compromise.} Receipt integrity ends at the trusted key and build boundary. Rotation and revocation are explicit migrations; receipts make a build attributable, not incorruptible.

\textbf{Deletion scope.} A row tombstone says nothing by itself about source documents, replicas, backups, archives, external logs, or learned model parameters. Each carrier needs its own deletion contract.

\textbf{Protection deadlock.} A policy can make all targets infeasible. \tfAmrNull{} is safe for authority but can freeze useful learning. Administrative recovery needs its own bounded, authenticated, and audited path.

%% file: theory_family/modules/amr/sections/12_falsification.tex
\section{Falsification Protocols and Stop Rules}
\label{tf:amr:sec:falsification}

No protocol in this section has been executed for this manuscript. Passing a lower gate authorizes only the next prespecified test; it does not establish an end-to-end claim.

\subsection{Contract-mechanics gates}

\tfAmrPlannedStatus
\begin{tfAmrProtocol}[Finite-state model checking]
Instantiate a reduced schema with small slot, version, ACL, status, alias, and rollback spaces. Exhaustively enumerate legal transitions, concurrent schedules permitted by the model, and recovery points. \textbf{Pass:} no legal trace violates I1--I16 and every illegal trace is rejected. \textbf{Stop:} any legal trace reaches an invalid state; repair the specification before implementation claims.
\end{tfAmrProtocol}

\tfAmrPlannedStatus
\begin{tfAmrProtocol}[Cross-language serialization]
Implement the same schema independently in at least two languages. Generate valid and invalid boundary rows including Unicode normalization, maximum lengths, typed zeros, counter maxima, reserved-bit corruption, alias collisions, NaNs, and every status normal form. \textbf{Pass:} byte-identical serialization, checksum/tag input, parsing, and accept/reject decisions. \textbf{Stop:} any equivalent input yields different authoritative bytes or validator decision.
\end{tfAmrProtocol}

\tfAmrPlannedStatus
\begin{tfAmrProtocol}[Transition and segment property tests]
Generate segments with $K_n\in[0,K_{\max}]$, repeated targets, all-\tfAmrNull{} segments, and mixed accept/reject outcomes. \textbf{Pass:} per-event row support $\le1$, $C_n\le K_n$, $D_n\le\min(C_n,S)$, untouched rows remain byte-identical, and published row traffic is exactly $C_nB_R$. \textbf{Stop:} any hidden second row or segment-rank-one assertion appears.
\end{tfAmrProtocol}

\tfAmrPlannedStatus
\begin{tfAmrProtocol}[Version, tombstone, and rollback traces]
Generate long randomized sequences of insert, update, alias rename, delete, rollback, protect, repair, purge, stale replay, and counter exhaustion. \textbf{Pass:} versions never decrease/wrap; rollback restores only bounded same-object content under current security; stale/unauthorized resurrection is zero during the declared scope. \textbf{Stop:} one unauthorized resurrection, version reuse, or obsolete ACL restoration.
\end{tfAmrProtocol}

\subsection{Security and recovery gates}

\tfAmrPlannedStatus
\begin{tfAmrProtocol}[Tenant and ACL adversary]
Run concurrent requests over disjoint partitions with forged slot IDs, tenant IDs, object IDs, stale index entries, ACL mutations, delete requests, and rollback requests. \textbf{Pass:} zero cross-tenant row changes and readable content; zero security weakening without capability. \textbf{Stop:} any logical cross-tenant effect. Timing and shared-resource leakage are measured in a later, stronger model.
\end{tfAmrProtocol}

\tfAmrPlannedStatus
\begin{tfAmrProtocol}[Receipt and dual-view corruption]
Flip every covered bit, substitute canonical/neural blocks and encoder IDs, replay predecessor receipts, alter validation bits, and corrupt cached verification keys. \textbf{Pass:} every unauthenticated alteration is rejected subject to the declared cryptographic model, and no mismatch yields a readable row. \textbf{Stop:} one changed covered core is accepted without acknowledged key compromise or cryptographic break.
\end{tfAmrProtocol}

\tfAmrPlannedStatus
\begin{tfAmrProtocol}[Crash injection]
Implement one storage profile, then inject process, kernel, and device failures at every durable-write and root-publication boundary, including torn-selector simulations and index lag. \textbf{Pass:} recovery exposes exactly the old or new complete table/audit binding and stale indexes never authorize old content. \textbf{Stop:} any mixed canonical/neural, control/content, tenant/object, or row/audit state becomes readable.
\end{tfAmrProtocol}

\tfAmrPlannedStatus
\begin{tfAmrProtocol}[Index determinism and stale suppression]
Rebuild indexes under different processing orders, drop updates, insert stale entries, and reuse slots for new objects. \textbf{Pass:} base rebuilds are byte identical; stale entries never authorize obsolete/deleted/cross-tenant rows; a full rebuild recovers the declared index. \textbf{Stop:} index state becomes independent authority.
\end{tfAmrProtocol}

\subsection{Accounting gates}

\tfAmrPlannedStatus
\begin{tfAmrProtocol}[Exact protocol-cost audit]
Instrument every AMR-1 pass and compare proposal bytes, row scans, neural comparison, hash blocks, authentication calls, candidate buffers, row/index writes, archive/audit bytes, barriers, and failure-prefix work with Section~\ref{tf:amr:sec:costs}. \textbf{Pass:} exact equality at the logical protocol layer for the declared profile and an explicit reconciliation to physical traffic. \textbf{Stop:} any mandatory component is unmeasured or hidden in a constant.
\end{tfAmrProtocol}

\tfAmrPlannedStatus
\begin{tfAmrProtocol}[Capacity and retention exhaustion]
Fill rows, aliases, rollback rings, quarantine, optional indexes, archives, audit ledgers, tombstones, and counters to every boundary. \textbf{Pass:} the declared reject/evict/rotate/purge policy occurs with no silent allocation or authority spill. \textbf{Stop:} resident or total-system boundedness is claimed without reproducing the carrier inventory.
\end{tfAmrProtocol}

\subsection{Semantic and reasoning gates}

\tfAmrPlannedStatus
\begin{tfAmrProtocol}[Same-checkpoint semantic qualification]
Compare canonical span, typed tuple, derived dual-view, certified dual-view, and pure latent control under one frozen checkpoint, fixed exposure, matched active parameters, fresh seeds, final-only evaluation, and truth-free materialization before scoring. Require canonical reconstruction, neural readout, fresh composition, query accuracy, calibration, old-value suppression, unrelated-row preservation, routing consistency, and corruption rejection at every fixed seed. \textbf{Stop:} if no interface passes all preregistered gates, no learned admission or reasoning integration is opened.
\end{tfAmrProtocol}

This gate must retain the accepted \tfAmrMMLA{} boundary: frozen-checkpoint semantic-interface study's 0/9 qualification and fitted-support fault-localization study's fitted learned-content failure motivate a new interface test; they do not count as a test of AMR-1.

\tfAmrPlannedStatus
\begin{tfAmrProtocol}[Repeated-write reasoning integration]
Only after substrate, crash, security, accounting, and semantic gates pass, compare an \tfAmrName{}-backed system with matched append-only context, canonical-only rows, and frozen retrieval. Measure factual update accuracy, stale leakage, untouched-fact preservation, unauthorized writes, tombstone/rollback correctness, answer quality, candidate multiplicity, total storage/index/archive/audit cost, latency, memory traffic, and energy. \textbf{Stop:} any reasoning gain accompanied by worse preregistered safety/integrity gates, or any gain that disappears under total-cost matching, blocks a system advantage claim.
\end{tfAmrProtocol}

\subsection{Dependency and qualification graph}

\begin{figure}[H]
\centering
\includegraphics[width=\linewidth]{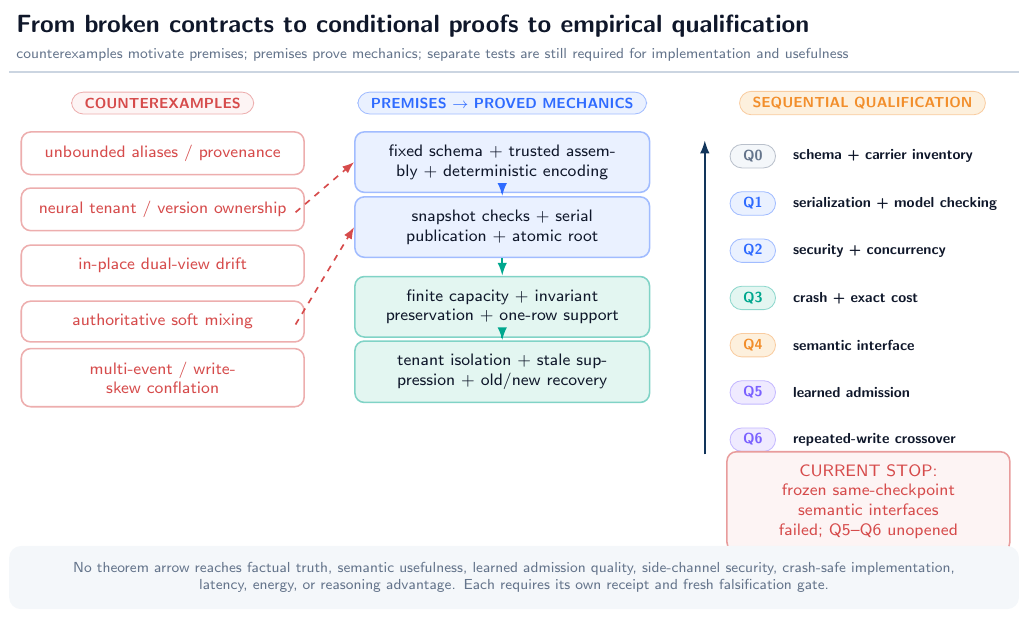}
\caption{\textbf{Counterexamples motivate premise bundles; premises support conditional theorems; theorems create implementation obligations, not empirical success.} The qualification ladder is sequential. The frozen semantic interface remains failed, so learned admission and reasoning-integration rungs are unopened.}
\label{tf:amr:fig:proof-ladder}
\end{figure}

The complete theorem-to-obligation matrix is in Appendix~\ref{tf:amr:app:obligations}. Its central rule is simple: an implementation may cite a theorem only after producing evidence for every premise used by that theorem. A successful checksum test cannot substitute for atomic storage; a successful crash test cannot substitute for semantic qualification; and a semantically useful reader cannot substitute for tenant or deletion authorization.

%% file: theory_family/modules/amr/sections/13_related_boundary.tex
\section{Related Work, Limitations, and Conclusion}
\label{tf:amr:sec:related}

\subsection{Neural external memory}

Memory Networks and Neural Turing Machines established influential differentiable read/write stores, while the Differentiable Neural Computer added structured access mechanisms and broad task demonstrations \cite{memorynetworks2014,ntm2014,tf:amr:graves2016dnc}. Their central contributions concern learnable addressing, differentiability, and task performance. \tfAmrName{} asks a complementary systems question: which exact finite bytes may become authoritative when canonical and neural representations, security, provenance, deletion, and rollback must agree?

The distinction allows coexistence. A model may maintain soft differentiable scratch state for computation and use hard \tfAmrName{} commits only for resident authority. This paper does not claim that hard rows should replace every neural memory mechanism or that the reference dimensions are optimal.

\subsection{Rows, transactions, and recovery}

Atomic state transformation, transaction boundaries, serialization, and recovery are classical database problems. Gray articulated the transaction concept and its limits; ARIES developed write-ahead recovery with fine-grained locking and partial rollback; linearizability gives a concurrent-object correctness condition; and Bigtable is a prominent row-oriented storage system \cite{tf:amr:gray1981transaction,tf:amr:mohan1992aries,tf:amr:herlihy1990linearizability,tf:amr:chang2006bigtable}. \tfAmrName{} does not claim invention of rows, atomicity, versions, tombstones, logs, or recovery.

Its narrower contribution is a reasoning-state contract in which canonical content, a neural payload/routing key, identity, security, provenance, deletion, rollback, and a consistency receipt share one fixed-width authoritative boundary. Its event transaction is intentionally weaker than a general database transaction: one event changes at most one resident row, while global uniqueness and crash recovery require explicit coordinator/storage assumptions.

\subsection{Provenance, deterministic encoding, and isolation}

Database provenance distinguishes source and transformation histories \cite{tf:amr:buneman2001provenance}. AMR-1 embeds only a bounded receipt and uses an external ledger for richer history. Deterministic encoding is essential for cross-implementation checksums and signatures; RFC 8949's deterministic CBOR discussion supplies useful precedent, while the profile here is stricter because each field has a fixed region \cite{tf:amr:bormann2020cbor}. SHA-256 and HMAC supply illustrative standardized digest and authentication profiles, not a mandate that every deployment use them \cite{tf:amr:nist2015sha,tf:amr:nist2008hmac}.

The tenant theorem follows the formal noninterference tradition but proves only equality of logical row observations under static disjoint partitions \cite{tf:amr:goguen1982noninterference}. It does not claim constant-time execution, shared-hardware isolation, or protection from a privileged administrator.

\subsection{Relationship to the integrated architecture}

The integrated architecture uses the complete-row, trusted-assembly, and one-overwrite-or-\tfAmrNull{} boundary developed here. This part specifies a fixed-bit state machine, reconciled concrete profile, proof set, security/recovery assumptions, counterexamples, and complete cost ledger. The report's controlled lifecycle and relational-binding experiments use different representations and do not validate AMR-1 directly.

The integrated parts separate policy-state witnesses, memory-state authority, predictive admission, dual policy/memory composition, and completed-segment consolidation. These are conceptual dependencies only, and each mechanism retains an independent evidence burden.

\subsection{Limitations}

\begin{enumerate}
  \item The trusted computing base is substantial: authentication, assembler, policy, canonicalizer, encoder/validator, key management, commit coordinator, and recovery logic.
  \item Static tenant partitions can waste capacity and do not address timing or shared-resource leakage.
  \item The AMR-1 uniqueness scan is linear in tenant slots; a faster index adds its own authority and concurrency obligations unless treated only as a hint.
  \item A deterministic encoder can be reproducible and still semantically useless. A learned validator can share the reader's blind spots.
  \item Fixed rollback forgets old history. A different evicted object requires an external archive or new observation.
  \item Fixed-width versions eventually exhaust. Epoch renewal adds another bounded authenticated namespace.
  \item Tombstones cannot remember an unbounded deletion set forever at fixed total cost.
  \item Per-event atomicity is not segment-wide or multi-object transactionality.
  \item Fail-closed validation protects authority at the cost of availability; corruption, key loss, or missing build artifacts can make rows unreadable.
  \item A receipt binds bytes and declared checks, not truth, source reliability, validator competence, legal compliance, or ethical desirability.
  \item Archive, audit, replication, filesystem, and backup costs are separate. Bounded hot state is not bounded total system storage.
  \item Exact logical byte counts do not predict physical traffic, latency, energy, throughput, endurance, or hardware advantage.
  \item No implementation, model checking, crash injection, security test, semantic qualification, or reasoning experiment is reported.
\end{enumerate}

\subsection{Explicit non-claims}

This paper does not claim that:
\begin{itemize}
  \item AMR-1 has been implemented or standardized;
  \item its 6,688-byte layout or dimensions are optimal;
  \item the public or technical-report evidence record validated a dual-view row;
  \item canonical/neural agreement establishes factual truth or semantic usefulness;
  \item checksums and receipts make a compromised trusted component safe;
  \item bounded rows improve language-model accuracy or reasoning;
  \item one-row atomicity implements general distributed transactions;
  \item tombstones guarantee global permanent deletion;
  \item rollback is always beneficial;
  \item static partitions eliminate all information leakage;
  \item deterministic indexes are fast or approximate indexes are safe without current-row revalidation;
  \item the profile lowers energy or wins a quality--cost comparison;
  \item memory-state mutation is itself strict policy training; or
  \item a theorem authorizes bypassing an empirical gate.
\end{itemize}

\subsection{Conclusion}

An Atomic Memory Row gives one resident object one typed identity, one canonical view, one bounded neural view, one tenant/security scope, one version lineage, one provenance receipt, and one publication boundary. The legal event is deliberately narrow: replace one complete row or perform bit-identical \tfAmrNull{}.

Under explicit assumptions, that narrow rule yields useful formal consequences: fixed resident information capacity, one-row event locality and the correct $K_n$ segment bound, invariant preservation, untouched-row preservation, monotone rollback, logical tenant isolation, stale-index rejection, serializable event histories, tamper evidence, and old-or-new crash recovery. It also exposes unavoidable limits: finite counters exhaust, bounded rollback forgets, finite tombstones cannot remember infinitely many deletions, and a genuine two-row merge is not one event.

The most important design choice is therefore not vector dimension but placement of authority. An alias map that bypasses row validation is another authority. An in-place neural refresh is another version. An uncharged deletion ledger defeats a total-system bound. A soft mixture is not a discrete fact version. Once those hidden states are made explicit, editable reasoning memory becomes a falsifiable systems object.

The result is not evidence that editable reasoning works. It is a bounded contract on which such evidence can later be demanded.

%% file: theory_family/modules/amr/appendices/A_proofs.tex
\section{Additional Proofs and Formal Boundaries}
\label{tf:amr:app:proofs}

\subsection{Canonical serialization and bit equality}

\tfAmrProvedStatus
\begin{tfAmrLemma}[Unique valid serialization]
\label{tf:amr:lem:unique-serialization}
Assume every variable-length region has an explicit length, deterministic normalization, fixed padding, fixed ordering, and zero reserved bytes. Then two valid parsed rows with the same typed field values have identical serialized bytes.
\end{tfAmrLemma}

\begin{proof}
Proceed through the schema regions in their fixed order. Fixed-width fields have a unique byte encoding by the schema's endianness and numeric rules. A variable-width field has one normalized content sequence, one encoded length, the exact content bytes, and uniquely determined zero padding to the fixed region boundary. Arrays have fixed capacity and canonical entry order; unused entries are the unique empty encoding. Reserved bytes are zero. Each region is therefore identical, hence their concatenation is identical.
\end{proof}

This lemma fails if map order, Unicode normalization, NaN payloads, or padding are unconstrained. A semantic equality relation over decoded objects is not enough for byte-addressed checksums.

\subsection{Lifecycle reduction and exact NULL}

\tfAmrProvedStatus
\begin{tfAmrProposition}[Row-local lifecycle reduction]
\label{tf:amr:prop:lifecycle-reduction}
Every authorized lifecycle operation whose complete postcondition changes exactly one row and yields a valid schema row is representable as Eq.~\ref{tf:amr:eq:replace}. A zero-change postcondition and every rejected operation are representable as Eq.~\ref{tf:amr:eq:null}.
\end{tfAmrProposition}

\begin{proof}
If no row changes, the identity branch gives the postcondition. Otherwise, for a valid row-local postcondition, let $a$ be its unique changed row and let $\widehat r_a$ be the complete postcondition serialization. Equation~\ref{tf:amr:eq:replace} produces exactly that table. A rejected operation has no legal postcondition and the contract assigns it \tfAmrNull{}, which is state identity.
\end{proof}

The premise ``changes at most one row'' excludes a genuine two-row merge, as Proposition~\ref{tf:amr:prop:merge-limit} shows.

\tfAmrProvedStatus
\begin{tfAmrProposition}[Zero write is not NULL]
\label{tf:amr:prop:zero-not-null}
If a legal candidate encodes a typed zero value and advances slot sequence or receipt, its commit is not \tfAmrNull{} even when every byte in its active payload region is zero.
\end{tfAmrProposition}

\begin{proof}
\tfAmrNull{} leaves every authoritative byte unchanged. The legal zero candidate contains a new slot sequence and receipt by premise, so its serialized row differs from its predecessor. Its replacement therefore changes authoritative state.
\end{proof}

\subsection{Tombstone suppression}

\tfAmrProvedStatus
\begin{tfAmrTheorem}[Stale-value suppression during retained tombstone scope]
\label{tf:amr:thm:tombstone-suppression}
Assume the current row for object $o$ is a valid retained \tfAmrTomb{} row; reads require current \tfAmrLive{} status; every index hit is revalidated; and becoming \tfAmrLive{} requires an authorized version-advancing resurrection or rollback. Then no stale pre-deletion row or index entry can supply $o$'s active content before such an authorized commit.
\end{tfAmrTheorem}

\begin{proof}
The current row fails the \tfAmrLive{} predicate and therefore supplies no active content. A stale index entry for the pre-deletion row carries an earlier slot sequence or receipt digest, so Theorem~\ref{tf:amr:thm:stale-index} rejects it. Any attempt to replace the tombstone without resurrection/rollback authority fails validation and resolves to \tfAmrNull{}, leaving the tombstone current. Thus only an authorized version-advancing commit can make $o$ readable.
\end{proof}

The scope ends when the tombstone is purged or when external ledgers no longer remember the deleted identity. The theorem does not promise perpetual suppression over an unbounded universe.

\subsection{Tenant-local capacity}

\tfAmrProvedStatus
\begin{tfAmrCorollary}[Tenant resident capacity]
\label{tf:amr:cor:tenant-capacity}
For tenant $t$ with $S_t=|P^{-1}(t)|$, its authoritative row projection has at most $2^{8S_tB_R}$ bit states and carries at most $8S_tB_R$ bits of mutual information about any history.
\end{tfAmrCorollary}

\begin{proof}
Apply Theorem~\ref{tf:amr:thm:capacity} to the concatenation of the tenant's $S_t$ fixed-width rows.
\end{proof}

This is both a storage bound and a denial-of-capacity boundary: a tenant cannot borrow another tenant's rows under static partitioning.

\subsection{Index rebuild determinism}

\tfAmrProvedStatus
\begin{tfAmrProposition}[Deterministic base-index rebuild]
\label{tf:amr:prop:index-rebuild}
Under deterministic parsing, row validation, key hashing, quantization, and fixed slot-major output order, rebuilding $I_C$ and $I_N$ from the same authoritative table yields byte-identical indexes.
\end{tfAmrProposition}

\begin{proof}
Each output region is assigned to one fixed pair $(a,q)$ for slot $a$ and primary/alias ordinal $q$, or to one fixed slot $a$ for the neural index. The validated row at that slot is fixed. Every projected field and digest is deterministic, and invalid/nonreadable projections have a unique empty encoding. Fixed output order eliminates scheduling dependence. Thus every output byte is a deterministic function of $M$.
\end{proof}

\subsection{Segment rank counterexample in matrix form}

Let $S\ge2$ and $d_v\ge2$. Choose two distinct rows $a\neq b$ and payload changes $u,w\in\mathbb R^{d_v}$ that are linearly independent. Event 1 changes only row $a$ by $u$; event 2 changes only row $b$ by $w$. Then
\begin{equation}
V(M_{n,2})-V(M_{n,0})=e_au^\top+e_bw^\top,
\end{equation}
where $e_a,e_b$ are distinct standard basis vectors. The two nonzero rows are linearly independent, so the matrix rank is two. This construction satisfies every per-event rank-one bound and disproves an unconditional whole-segment rank-one claim.

\subsection{Read-time fail-closed property}

\tfAmrProvedStatus
\begin{tfAmrProposition}[No view selection after mismatch]
\label{tf:amr:prop:no-view-selection}
If Eq.~\ref{tf:amr:eq:readability} is the only active-content read rule, a row with checksum, authentication, status, ACL, time, or dual-view failure supplies neither canonical nor neural active content.
\end{tfAmrProposition}

\begin{proof}
The rule is a conjunction. Failure of any conjunct selects the otherwise branch $\bot$. No rule returns one view conditional on failure of the other.
\end{proof}

The proposition is only as strong as the carrier inventory. An undocumented vector cache that bypasses Eq.~\ref{tf:amr:eq:readability} is an additional authority and invalidates the premise.

\subsection{Commit and audit binding}

\tfAmrProvedStatus
\begin{tfAmrProposition}[No accepted unaudited commit under a required-ledger profile]
\label{tf:amr:prop:audit-binding}
Assume policy marks an event as requiring audit, the candidate receipt binds the intended audit sequence/root, and publication verifies that the prepared audit record is durably reachable from that root before changing the table selector. Then an accepted recovered postcommit root has the corresponding audit binding; failure to prepare the audit record yields the old root.
\end{tfAmrProposition}

\begin{proof}
If audit preparation fails, the publication precondition is false and the selector remains on the old root. If the selector reaches the new root, the protocol's ordering premise guarantees the bound audit object was prepared and recovery validates its reachability. Therefore every accepted postcommit root satisfies the required binding.
\end{proof}

This result binds an audit record, not an unbounded human-readable log. A bounded checkpoint/rotation policy determines how long detail remains available.

\subsection{Assumption sensitivity}

\begin{table}[H]
\centering
\footnotesize
\begin{tabularx}{\linewidth}{p{0.25\linewidth}p{0.30\linewidth}Y}
\toprule
Dropped premise & Formal result lost & Minimal witness \\
\midrule
Fixed field widths & capacity theorem & append one alias per event \\
Trusted field derivation & isolation/version/protection & proposal self-assigns tenant or version \\
Deterministic encoding & byte equality/rebuild & map order or NaN payload differs \\
Snapshot-consistent global checks & identity uniqueness & concurrent duplicate inserts \\
Serializable publication & serial history and uniqueness & write skew across target rows \\
Atomic root publication & no-torn/crash theorem & visible partial page or mixed root \\
Current-row index validation & stale suppression & old alias hit returns old row \\
Cryptographic assumptions & tamper-evidence theorem & collision or tag forgery \\
Static partition/logical observation & tenant noninterference & row migration or timing channel \\
Required-ledger fail-closed rule & audit binding & row publishes after audit failure \\
\bottomrule
\end{tabularx}
\caption{Premises are necessary to the indicated claims; later evidence must test them independently.}
\label{tf:amr:tab:assumption-sensitivity}
\end{table}

%% file: theory_family/modules/amr/appendices/B_profile_ledger.tex
\section{AMR-1 Profile Ledger}
\label{tf:amr:app:profile-ledger}

This appendix reconciles every byte used by the illustrative profile. It is a design artifact, not a deployed wire-format standard.

\paragraph{Status partition.}
The AMR-1 schema and transition contract are \textbf{FORMALIZED}; the invariant,
serializability, tamper-evidence, and old-or-new recovery statements are
\textbf{PROVED UNDER STATED ASSUMPTIONS}. No AMR-1 implementation-conformance
receipt bundle is reported, so implementation status is \textbf{UNVALIDATED}.
Semantic usefulness of the proposed dual-view row is \textbf{NOT TESTED / UNVALIDATED}.
The previously tested pure latent-row interfaces remain \textbf{FAILED / UNQUALIFIED}
and are not evidence against every possible AMR profile.

\subsection{Trusted control: 256 bytes}

\begin{table}[H]
\centering
\small
\begin{tabular}{lr}
\toprule
Field & Bytes \\
\midrule
Magic, schema, status, operation, flags & 10 \\
Physical slot identifier & 4 \\
Tenant identifier & 16 \\
Logical object identifier & 16 \\
Owner/principal identifier & 16 \\
Object type & 2 \\
Slot sequence & 8 \\
Object version & 8 \\
Valid-from, valid-until, retention-until & 24 \\
Protection class and deletion-reason code & 4 \\
Eight ACL entries: 16-byte principal + 2-byte rights & 144 \\
Rollback ring head and occupancy & 2 \\
Reserved zero bytes & 2 \\
\midrule
\textbf{Total} & \textbf{256} \\
\bottomrule
\end{tabular}
\caption{AMR-1 trusted-control allocation.}
\label{tf:amr:tab:control-ledger}
\end{table}

The object type and header flags supply the canonical type/semantic flags referenced in Eq.~\ref{tf:amr:eq:canonical-capsule}; they remain system-owned.

\subsection{Canonical capsule: 1,104 bytes}

\begin{equation}
\underbrace{(2+64)}_{\text{key}}
+\underbrace{(2+512)}_{\text{value}}
+\underbrace{(1+3)+8(1+64)}_{\text{alias count/reserved + entries}}
=1{,}104.
\label{tf:amr:eq:canonical-ledger}
\end{equation}
Lengths count canonical bytes, not code points. Padding to each maximum is zero. The primary key and value use 16-bit lengths; each alias uses an 8-bit length because its maximum is 64. Alias count is at most eight.

\subsection{Neural capsule: 784 bytes}

\begin{equation}
\underbrace{128\times2}_{\text{routing}}
+\underbrace{256\times2}_{\text{payload}}
+\underbrace{4}_{\text{uncertainty + abstention}}
+\underbrace{1}_{\text{proposed intent}}
+\underbrace{11}_{\text{reserved}}
=784.
\label{tf:amr:eq:neural-ledger}
\end{equation}
The 768 routing/payload bytes participate in exact derived-mode comparison. Uncertainty and abstention are two signed or unsigned 16-bit schema values; intent is one enumerated byte.

\subsection{Provenance: 192 bytes}

\begin{table}[H]
\centering
\small
\begin{tabular}{lr}
\toprule
Field & Bytes \\
\midrule
Event identifier & 16 \\
Source-material or source-receipt digest & 32 \\
Proposer/model-build digest & 32 \\
Assembler build identifier & 16 \\
Encoder build identifier & 16 \\
Validator build identifier & 16 \\
Observation time & 8 \\
Source class & 2 \\
Committed operation copy & 1 \\
Deletion-authorization digest & 32 \\
External audit handle & 16 \\
Reserved zero bytes & 5 \\
\midrule
\textbf{Total} & \textbf{192} \\
\bottomrule
\end{tabular}
\caption{AMR-1 bounded provenance allocation. A source document or provenance graph is external.}
\label{tf:amr:tab:provenance-ledger}
\end{table}

\subsection{Rollback: two 2,048-byte snapshots}

Each snapshot contains no nested rollback data:
\begin{table}[H]
\centering
\small
\begin{tabular}{lr}
\toprule
Field & Bytes \\
\midrule
Prior canonical capsule & 1,104 \\
Prior neural capsule & 784 \\
Same-object identifier & 16 \\
Prior slot sequence & 8 \\
Prior object version & 8 \\
Prior validity and retention times & 24 \\
Prior protection/status/snapshot flags & 8 \\
Prior provenance digest & 32 \\
Prior receipt digest & 32 \\
Reserved zero bytes & 32 \\
\midrule
\textbf{One snapshot} & \textbf{2,048} \\
\textbf{Two-snapshot ring} & \textbf{4,096} \\
\bottomrule
\end{tabular}
\caption{AMR-1 rollback allocation. The snapshot is sufficient to propose restoration of semantic state; current policy and new provenance are assembled at rollback time.}
\label{tf:amr:tab:rollback-ledger}
\end{table}

An evicted different object is excluded. A rollback snapshot does not carry an ACL or recursively embedded history; current security controls govern restoration.

\subsection{Consistency receipt: 256 bytes}

\begin{table}[H]
\centering
\small
\begin{tabular}{lr}
\toprule
Field & Bytes \\
\midrule
Core checksum/digest & 32 \\
Typed predecessor descriptor & 96 \\
Validation-registry digest & 32 \\
Validation bitmap & 8 \\
Schema/algorithm/key-epoch/format identifiers & 16 \\
Authorization-decision binding & 16 \\
Event binding & 16 \\
Authentication tag & 32 \\
Reserved zero bytes & 8 \\
\midrule
\textbf{Total} & \textbf{256} \\
\bottomrule
\end{tabular}
\caption{AMR-1 receipt allocation.}
\label{tf:amr:tab:receipt-ledger}
\end{table}

The receipt-covered core is therefore $6{,}688-256=6{,}432$ bytes. The 96-byte predecessor descriptor uses the tagged fixed-width normal form in Section~\ref{tf:amr:sec:schema}; it is not a digest that erases lineage fields.

\subsection{Base-index entries}

One 112-byte canonical entry contains:
\begin{equation}
16_{\mathrm{tenant}}+16_{\mathrm{object}}+32_{\mathrm{key\ digest}}
+4_{\mathrm{slot}}+8_{\mathrm{sequence}}+32_{\mathrm{receipt\ digest}}
+4_{\mathrm{ordinal/flags}}=112.
\end{equation}
One 336-byte neural entry contains:
\begin{equation}
256_{\mathrm{route}}+16_{\mathrm{tenant}}+16_{\mathrm{object}}+4_{\mathrm{slot}}
+8_{\mathrm{sequence}}+32_{\mathrm{receipt\ digest}}+4_{\mathrm{flags}}=336.
\end{equation}
The fixed slot-major arrays contain entries for empty/tombstoned/quarantined rows in the unique disabled normal form. Their size is therefore independent of occupancy.

\subsection{Auxiliary and external carrier parameters}

AMR-1's reported 1.9609375 MiB total includes authoritative rows and both base indexes. It sets persistent non-authoritative fast state and diagnostic-copy quarantine to zero. A conforming extension publishes the following additional parameters:
\begin{equation}
\begin{aligned}
B_Q&=Q_{\max}B_q &&\text{diagnostic copies},\\
B_F&=S_Fd_Fp_F+B_{F,\mathrm{meta}} &&\text{soft fast state},\\
B_A(t)&\le A_{\max}\ \text{or explicitly unbounded} &&\text{archive},\\
B_L(t)&\le L_{\max}B_\ell+B_{L,\mathrm{index}}\ \text{or explicitly unbounded} &&\text{audit ledger}.
\end{aligned}
\label{tf:amr:eq:extension-ledgers}
\end{equation}
If $B_Q$ or $B_F$ is nonzero, its authority and reset/lifetime rule must be audited. If $A_{\max}$ or $L_{\max}$ is finite, reaching it invokes Table~\ref{tf:amr:tab:cap-behavior}; if absent, total-system boundedness is not claimed.

The safe-path ephemeral working set has at least the proposal, candidate, and encoder scratch regions,
\begin{equation}
B_{\mathrm{work,min}}=1{,}888+6{,}688+768=9{,}344\ \text{bytes},
\end{equation}
plus implementation-specific old-row, parser, hash/MAC, policy, index, and I/O buffers. A deployment must measure its peak physical working set rather than treating the minimum as exact RAM use.

\subsection{Arithmetic reconciliation}

\begin{equation}
256+1{,}104+784+192+4{,}096+256=6{,}688.
\end{equation}
\begin{equation}
6{,}688\times256=1{,}712{,}128,
\quad
256(9\times112+336)=344{,}064.
\end{equation}
\begin{equation}
1{,}712{,}128+344{,}064=2{,}056{,}192=1.9609375\times2^{20}.
\end{equation}
Every headline AMR-1 byte count follows from these fixed allocations.

%% file: theory_family/modules/amr/appendices/C_obligations.tex
\section{Theorem-to-Implementation Obligations}
\label{tf:amr:app:obligations}

Theorems are reusable only when the implementation supplies the corresponding premise evidence. Table~\ref{tf:amr:tab:implementation-obligations} is a minimum receipt set.

\paragraph{Status partition.}
This appendix separates the formal AMR contract from evidence about an
implementation. The contract itself is \textbf{FORMALIZED}; results whose proof
uses the displayed assumptions are \textbf{PROVED UNDER STATED ASSUMPTIONS}.
AMR-1 implementation conformance is \textbf{UNVALIDATED}, and semantic use of
the proposed dual-view row is \textbf{NOT TESTED / UNVALIDATED}. The earlier
pure latent-row interface studies are \textbf{FAILED / UNQUALIFIED}; their
negative result is a boundary for that interface family, not a proof that every
AMR representation is unusable.

\begin{longtable}{p{0.22\linewidth}p{0.31\linewidth}p{0.39\linewidth}}
\caption{Formal result to implementation and falsification obligations.}\label{tf:amr:tab:implementation-obligations}\\
\toprule
Claim & Required premise bundle & Minimum implementation receipt/test \\
\midrule
\endfirsthead
\toprule
Claim & Required premise bundle & Minimum implementation receipt/test \\
\midrule
\endhead
Fixed resident capacity (Theorem~\ref{tf:amr:thm:capacity}) & exact schema; complete carrier inventory; fixed $S,B_R$ & schema-generated offset table; binary-size checks; allocator and external-carrier inventory; exhaustion tests \\
Unique serialization (Lemma~\ref{tf:amr:lem:unique-serialization}) & deterministic normalization, lengths, padding, order, numeric rules & cross-language golden corpus and boundary/fuzz vectors with byte-identical results \\
Event/segment locality (Theorem~\ref{tf:amr:thm:edit-locality}) & all authority in $M$; one target or \tfAmrNull{} per event & whole-process write tracing; state hash before/after every event; hidden-cache/alias-map inventory \\
Trusted-field nondelegation (Proposition~\ref{tf:amr:prop:trusted-field-nondelegation}) & proposer isolated from trusted functions and keys & data-flow/code audit; privilege separation; adversarial field-injection suite; key-access audit \\
No torn dual view (Theorem~\ref{tf:amr:thm:no-torn-dual}) & one atomic complete-row publication & crash/concurrency injection at every row-write boundary; no mixed serialization accepted \\
Invariant preservation (Theorem~\ref{tf:amr:thm:invariant-preservation}) & valid genesis; trusted deterministic assembly; snapshot-consistent serializable validation & model check on reduced schema; property-based transition tests; global-uniqueness concurrency tests \\
Version monotonicity (Corollary~\ref{tf:amr:cor:version-monotonicity}) & exact increment and overflow rejection & long trace; stale replay; maximum counter; restart and recovery tests \\
Untouched preservation (Corollary~\ref{tf:amr:cor:untouched}) & complete target isolation & byte hashes of all non-target rows across every event and crash point \\
Monotone rollback (Proposition~\ref{tf:amr:prop:monotone-rollback}) & same-object bounded snapshot; current security; version increment & rollback/resurrection adversary; obsolete ACL/protection fixtures; ring-overflow tests \\
Tombstone suppression (Theorem~\ref{tf:amr:thm:tombstone-suppression}) & retained current tombstone; revalidated indexes; authorized resurrection only & stale row/index replay, aliases, backups in scope, privilege-negative tests \\
Stale-index safety (Theorem~\ref{tf:amr:thm:stale-index}) & every hit bound to and checked against current row & fault injection dropping/reordering index updates; slot reuse; cross-tenant forged entries \\
Tenant noninterference (Theorem~\ref{tf:amr:thm:tenant-noninterference}) & static partitions and logical observation model & cross-tenant mutation/read adversary; separate side-channel scope statement and tests \\
Serializable history (Theorem~\ref{tf:amr:thm:serializable}) & tenant-total order or equivalent CAS; global constraint serialization & concurrent history checker; write-skew fixtures; linearization receipt; stale snapshot rejection \\
Old-or-new recovery (Theorem~\ref{tf:amr:thm:crash-recovery}) & correct storage protocol, atomic selector, durable ordering, validating recovery & power/process/device crash matrix; torn selectors/pages; flush/fence audit; replica failover if claimed \\
Receipt tamper evidence (Theorem~\ref{tf:amr:thm:receipt-tamper}) & collision resistance, unforgeability, protected keys & algorithm/key inventory; bit-flip/replay/substitution suite; rotation/revocation and compromise response \\
Index rebuild determinism (Proposition~\ref{tf:amr:prop:index-rebuild}) & deterministic projection and fixed order & repeated parallel/serial/cross-language rebuilds with identical hashes \\
Audit binding (Proposition~\ref{tf:amr:prop:audit-binding}) & required-ledger reservation and atomic root binding & audit-full, write-failure, rotation, recovery, and dangling-record tests \\
Semantic usefulness & \emph{not implied by any theorem} & preregistered same-checkpoint reconstruction/query/calibration/preservation qualification \\
Reasoning or cost advantage & \emph{not implied by any theorem} & repeated-write, security, and total-system compute/storage/latency/energy matched comparison \\
\bottomrule
\end{longtable}

\subsection{Implementation conformance statement}

\tfAmrPrescriptionStatus\quad A system claiming \tfAmrName{} conformance should publish:
\begin{enumerate}
  \item schema identifier and generated field-offset/size table;
  \item complete mutable-carrier inventory and authority classification;
  \item trusted computing boundary, privilege/key map, and build digests;
  \item serializer/parser/validator golden vectors;
  \item event trace with snapshot, candidate, decision, target, predecessor, new receipt, and \tfAmrNull{} reason;
  \item current-row read and index-revalidation trace;
  \item tenant/ACL/protection/deletion/rollback policy fixtures;
  \item storage publication ordering and recovery specification;
  \item archive, audit, index, quarantine, replication, and backup capacity/retention policies;
  \item exact logical and measured physical cost ledger; and
  \item explicit list of theorems claimed, with every premise receipt linked.
\end{enumerate}

Missing evidence weakens only the affected claim. For example, a correct serializer may support fixed-bit parsing while crash recovery remains unvalidated. A successful semantic reader may support interface usefulness while deletion authority remains unvalidated. Status must not be inherited by the whole system.

\subsection{Stop-rule ordering}

The intended order is:
\begin{equation}
\begin{aligned}
&\text{schema/transition mechanics}
\rightarrow\text{serialization and model checking}
\rightarrow\text{security/concurrency}\\
&\rightarrow\text{crash recovery and exact cost}
\rightarrow\text{semantic interface}
\rightarrow\text{learned admission}\\
&\rightarrow\text{repeated writes}
\rightarrow\text{total-system quality--cost comparison}.
\end{aligned}
\label{tf:amr:eq:stop-order}
\end{equation}
Failure at a rung stops claims that depend on it. It does not erase independent lower-rung evidence, and it does not authorize a post-hoc threshold, seed, checkpoint, or threat-model change.

The frozen \tfAmrMMLA{} result currently stops at the semantic-interface rung for the tested interfaces. This manuscript supplies a new formal substrate candidate, not evidence that the rung has been passed.

%% file: theory_family/modules/amr/appendices/D_symbols.tex
\section{Symbol and Status Ledger}
\label{tf:amr:app:symbols}

\begin{longtable}{p{0.20\linewidth}p{0.72\linewidth}}
\caption{Principal symbols.}\label{tf:amr:tab:symbols}\\
\toprule
Symbol & Meaning \\
\midrule
\endfirsthead
\toprule
Symbol & Meaning \\
\midrule
\endhead
$\sigma$ & immutable row schema and deterministic serialization profile \\
$S,\tfAmrCalS$ & resident slot count and slot set $\{1,\ldots,S\}$ \\
$M,r_a$ & complete authoritative table and row in slot $a$ \\
$B_R$ & complete serialized row bytes; AMR-1 uses 6,688 \\
$h,c,n,p,\lambda,\rho$ & trusted control, canonical, neural, provenance, rollback, and receipt row blocks \\
$P(a)$ & trusted static tenant assigned to slot $a$ \\
$v_s,v_o$ & physical slot sequence and same-object version \\
$z(r)$ & row status: \tfAmrEmpty{}, \tfAmrLive{}, \tfAmrTomb{}, or \tfAmrQuar{} \\
$\widetilde p$ & untrusted neural proposal \\
$\chi$ & authenticated request/event context and capabilities \\
$D_{\sigma,\omega}$ & versioned canonical--neural conformance relation \\
$\tfAmrNull$ & exact no-authoritative-change action \\
$E_n,K_n$ & ordered event list for segment $n$ and its length \\
$M_{n,j}$ & authoritative state after event $j$ of segment $n$ \\
$C_n,D_n$ & accepted commit count and distinct changed-slot count in segment $n$ \\
$B_I,B_Q,B_F$ & fixed index, diagnostic quarantine, and non-authoritative fast-state capacity \\
$B_A(t),B_L(t)$ & separately charged archive and audit-ledger storage at time $t$ \\
$\mathcal W$ & heterogeneous logical work vector for a candidate/commit \\
$B_P,B_C,B_D$ & proposal, canonical-scan, and compared neural route/payload bytes \\
$N_H$ & hash compression-block count \\
$\operatorname{Obs}_t$ & logical authorized observation projected to tenant $t$ \\
$\operatorname{Inv}_\sigma$ & conjunction of the sixteen state invariants \\
\bottomrule
\end{longtable}

\subsection{Status-to-evidence rule}

For the integrated report, the AMR statuses are additionally partitioned as
follows: the dual-view/AMR formal contract is \textbf{FORMALIZED}; conditional
invariants and recovery are \textbf{PROVED UNDER STATED ASSUMPTIONS}; AMR-1
implementation conformance is \textbf{UNVALIDATED}; semantic use of the
proposed dual-view row is \textbf{NOT TESTED / UNVALIDATED}; and the previously
tested pure latent-row interfaces are \textbf{FAILED / UNQUALIFIED}.

\begin{table}[H]
\centering
\small
\begin{tabularx}{\linewidth}{lYY}
\toprule
Displayed status & Admissible support in this paper & Required before promotion \\
\midrule
DEFINITION & internal typed construction & independent terminology and interface audit \\
PROVED UNDER STATED ASSUMPTIONS & deductive proof in source & independent proof audit; premise evidence for implementation use \\
DESIGN PRESCRIPTION & engineering necessity/counterexample & implemented mechanism and adversarial test \\
PLANNED TEST & protocol text only & immutable preregistration and completed evidence \\
UNVALIDATED & explicit absence of qualifying evidence & matched fresh evaluation under frozen gates \\
CONJECTURED & stated hypothesis only & theory proof or accepted experiment, depending on claim \\
\bottomrule
\end{tabularx}
\caption{No status automatically promotes another.}
\label{tf:amr:tab:status-evidence-rule}
\end{table}

\subsection{Evidence boundary}

The mathematical statements in this part are conditional theory. Empirical statements are limited to the report-wide component results and semantic-interface boundary. Definitions, file organization, and arithmetic checks are not scientific evidence without an explicit derivation or admitted measurement.

%% file: theory_family/modules/pma/module.tex
\clearpage
\part{Predictive Memory Admission}
\label{tf:pma:part:predictive-memory-admission}

\noindent\textbf{Scope.}
This part specifies typed completed-segment state, grouped alternative futures,
action-conditioned replay, uncertainty-aware admission, conditional results,
counterexamples, experimental obligations, and stop rules under explicit
assumptions.

\input{theory_family/modules/pma/sections/01_scope}
\input{theory_family/modules/pma/sections/02_typed_state}
\input{theory_family/modules/pma/sections/03_event_recursion}
\input{theory_family/modules/pma/sections/04_grouped_futures}
\input{theory_family/modules/pma/sections/05_replay_loss}
\input{theory_family/modules/pma/sections/06_oracle_gate}
\input{theory_family/modules/pma/sections/07_uncertainty}
\input{theory_family/modules/pma/sections/08_identification}
\input{theory_family/modules/pma/sections/09_value_model}
\input{theory_family/modules/pma/sections/10_metrics_audits}
\input{theory_family/modules/pma/sections/11_costs}
\input{theory_family/modules/pma/sections/12_counterexamples_stops}
\input{theory_family/modules/pma/sections/13_obligations}
\input{theory_family/modules/pma/sections/14_related_limits}

\FloatBarrier
\input{theory_family/modules/pma/appendices/A_proofs}
\input{theory_family/modules/pma/appendices/B_estimators}
\input{theory_family/modules/pma/appendices/C_obligation_matrix}
\input{theory_family/modules/pma/appendices/D_symbols}

%% file: theory_family/modules/pma/sections/01_scope.tex
\section{Introduction}
\label{tf:pma:sec:scope}

\begin{figure}[H]
\centering
\includegraphics[width=\linewidth]{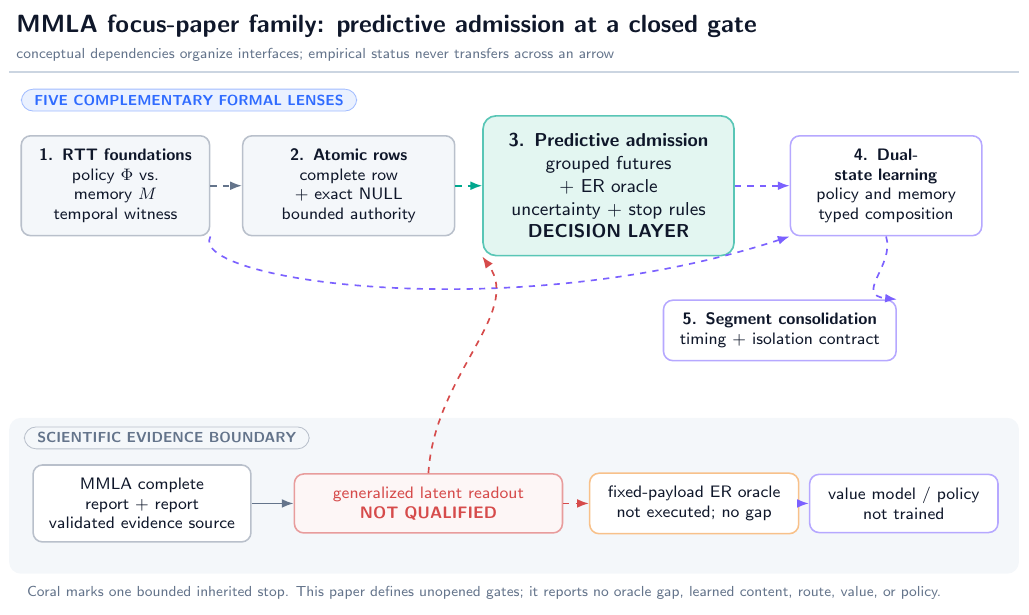}
\caption{Predictive memory admission: future-valued action selection and the distinction between a conditional oracle target and an empirically established advantage.}
\label{tf:pma:fig:technology-tree}
\end{figure}

\subsection{Admission is a decision about authority}

A transcript may retain a sentence, a retriever may return it, and a recurrent state may compress it. None of those operations alone decides whether the sentence should replace a current authoritative fact. Under capacity pressure, a write has at least three possible costs: it may admit false or stale content, displace a useful row, or spend resources without improving any later decision. Refusing a write has a different cost: the state may remain stale or omit information that would have helped a later task. Predictive memory admission is the decision problem created by those asymmetric consequences.

\tfPmaDefinitionStatus
\begin{tfPmaDefinition}[Predictive memory admission]
Predictive memory admission is the selection of one action from a finite feasible set containing exact \tfPmaNull{} and zero or more complete target-conditioned row commits, using only information available at the decision boundary, with the objective and evidence gate fixed before evaluation futures are opened.
\end{tfPmaDefinition}

The word \emph{predictive} refers to conditional future consequence, not access to a realized future at deployment. A training teacher may use later outcomes to estimate action risk. A deployed value model may use only the causal state. The difference is structural and will be expressed as a filtration constraint rather than a promise in prose.

\subsection{Evidence boundary}

This formalization does not supply empirical admission evidence. Historically, the frozen-checkpoint semantic-interface study qualified zero of nine jobs; a fitted-population diagnostic found learned content exact on $0/73{,}728$ probes with either learned or mechanical target routing. That fit/support failure remains specific to its interface. Later readout studies achieve restricted generation, update, and restoration, but the latest coverage-controlled comparison does not qualify generalized continuous-event latent readout. More importantly for this part, those studies use an external rule writer and oracle reader, not a learned future-risk admission intervention.

\tfPmaUnvalidatedStatus\quad At the September 13, 2026 cutoff, the expected-risk oracle defined in this part has not been run, no positive oracle margin has been measured, and no future-blind admission policy has been validated. Predictive overwrite remains an untested claim. Numerical examples in this formal part are illustrative and its proposed admission experiments are future work; the empirical readout chapters elsewhere in the report are completed studies with a different intervention.

Part~\ref{tf:rtt:part:foundations} separates mutable policy state from memory state, and Part~\ref{tf:amr:part:atomic-memory-rows} supplies complete-row, trusted-assembly, version, protection, and exact-\tfPmaNull{} semantics. These formal dependencies do not supply empirical evidence for admission.

\subsection{Contributions}

This paper owns the following questions.
\begin{enumerate}
  \item It types the causal state, completed segment, event sequence, bounded resident memory, target-conditioned candidate, action set, grouped future, horizon loss, resource ledger, and deployable information set.
  \item It gives event-indexed teacher semantics. Event $j$ reads the branch state produced by earlier committed events; it never acts on a whole-segment candidate computed against an obsolete prefix.
  \item It defines grouped-future sampling and action-conditioned replay, including branch weights, support, synthetic-future bias, stale leakage, collateral damage, write/\tfPmaNull{} penalties, and resources.
  \item It separates an empirical expected-risk oracle, a current-risk comparator, exact \tfPmaNull{}, and a future-blind value model.
  \item It supplies simultaneous concentration and conservative uncertainty rules, identification and impossibility results, metrics, leakage audits, full costs, stop rules, and theorem-to-implementation obligations.
\end{enumerate}

This formal part does not specify a row serializer, trusted assembler implementation, semantic reader, event extractor, archive policy, policy-state RTT update, or complete MMLA system. It adds no training job, checkpoint, or measured oracle gap. Definitions, conditional proofs, prescriptions, and planned tests are marked locally; none constitutes implementation evidence.

\subsection{Central falsifiable question}

For a causal prefix state $\widetilde S$, let $R(a\mid\widetilde S)$ be the conditional horizon risk of feasible action $a$, let $a^{\mathrm{cur}}$ be a preregistered current-risk action, and let $0$ denote \tfPmaNull{}. The first scientific question is
\begin{equation}
G(\widetilde S)
=R(a^{\mathrm{cur}}\mid\widetilde S)
-\min_{a\in\tfPmaCalA(\widetilde S)}R(a\mid\widetilde S).
\label{tf:pma:eq:scope-gap}
\end{equation}
The relevant population claim is not that $G$ can be positive in a hand-built example. It is that a frozen empirical procedure yields a stable, uncertainty-controlled improvement exceeding a prespecified materiality margin on fresh prefix groups and required seeds. If it does not, the value model remains closed.

\tfPmaConjecturedStatus\quad Some domains may contain predictable future-use structure for which a bounded authoritative memory benefits from selective admission. This conjecture can fail because the action set is poor, futures are not identified, the gap is too small, the value is not predictable from causal history, long-run pollution dominates, or total-system cost erases the benefit.

%% file: theory_family/modules/pma/sections/02_typed_state.tex
\section{Typed Causal State and Deployable Information}
\label{tf:pma:sec:typed-state}

\subsection{Probability space and completed segments}

Fix a probability space $(\Omega,\mathcal B,\tfPmaP)$ and a discrete event clock with filtration $(\tfPmaCalF_t)_{t\ge0}$. Stream instance $n$ has identity $I_n\in\tfPmaCalI$, observations $X_{n,t}\in\tfPmaCalX$, and a completed segment
\begin{equation}
B_n=(X_{n,b_{n-1}+1},\ldots,X_{n,b_n})
\label{tf:pma:eq:segment}
\end{equation}
that becomes available only at stopping time $b_n$. Prediction at or before $b_n$ cannot use a memory mutation produced after $B_n$ completes. A bounded eventizer maps the completed segment and prior state to an ordered list
\begin{equation}
E_n=\mathsf{Eventize}(B_n,M_{n-1};\xi_n)
=(e_{n,1},\ldots,e_{n,J_n}),
\qquad 0\le J_n\le J_{\max},
\label{tf:pma:eq:eventize}
\end{equation}
where $\xi_n$ is recorded eventizer randomness. Eventization is itself a costed, unvalidated component; the theory conditions on its declared output.

\begin{table}[H]
\centering
\footnotesize
\begin{tabularx}{\linewidth}{@{}p{0.15\linewidth}p{0.19\linewidth}Y Y@{}}
\toprule
Object & Space & Meaning & Boundary \\
\midrule
$I_n$ & $\tfPmaCalI$ & Prefix-group / stream identity & Split and receipt unit \\
$B_n$ & bounded sequence & Completed observed segment & No unobserved continuation \\
$e_{n,j}$ & $\tfPmaCalE$ & Event $j$ in the ordered segment list & Processed only after events $<j$ \\
$M_{n,j}$ & $\tfPmaCalM=\tfPmaCalR^S$ & $S$ complete authoritative rows after event $j$ & Fixed resident capacity \\
$c_{n,j,a}$ & $\tfPmaCalR\cup\{\bot\}$ & Complete row candidate for target $a$ & Built against current branch prefix \\
$A_{n,j}$ & finite subset & Feasible commits plus \tfPmaNull{} & Hard mask precedes value scoring \\
$F_{n,k}$ & $\tfPmaCalF$ & Future branch $k$ for the same prefix & Training/evaluation only \\
$H$ & $\mathbb N$ & Replay horizon & Fixed before outcomes open \\
$L$ & $[0,L_{\max}]$ & Registered aggregate action loss & Bounded for finite-sample results \\
$\mathfrak R$ & $\mathbb R_+^d$ & Resource ledger & Logical and measured coordinates \\
$S^-_{n,j},S^+_{n,j}$ & $\tfPmaCalS^-\!\times\tfPmaCalS^+$ & Preconstruction / postvalidation state & Bounded; no future-bearing field \\
\bottomrule
\end{tabularx}
\caption{Typed objects. Physical co-location does not merge their causal roles.}
\label{tf:pma:tab:typed-objects}
\end{table}

\subsection{Bounded authoritative state}

Let
\begin{equation}
M_{n,j}=(r_{n,j,1},\ldots,r_{n,j,S})\in\tfPmaCalR^S
\label{tf:pma:eq:resident-state}
\end{equation}
be the complete resident state. The row space $\tfPmaCalR$ and slot count $S$ are finite-width under the external contract. This paper assumes a trusted validator can distinguish a complete committable row from $\bot$; it does not assume that the row is factually true or semantically usable.

For event $j$, candidate construction for target $a\in\{1,\ldots,S\}$ is the partial map
\begin{equation}
c_{n,j,a}=\tfPmaConstruct(S^-_{n,j},e_{n,j},a,r_{n,j-1,a},M_{n,j-1};\zeta_{n,j,a})
\in\tfPmaCalR\cup\{\bot\},
\label{tf:pma:eq:candidate}
\end{equation}
with fixed or recorded constructor randomness $\zeta_{n,j,a}$. Target conditioning is essential because version, protection, provenance, collision, eviction, tenant, and rollback fields can depend on the old target row. A shared semantic payload may be a restricted causal probe, but the authoritative action still requires a complete target-specific assembly.

The feasible set is
\begin{equation}
\tfPmaCalA_{n,j}=\{0\}\cup
\left\{a\in\{1,\ldots,S\}:c_{n,j,a}\neq\bot,\ \tfPmaValid(c_{n,j,a};M_{n,j-1})=1\right\},
\label{tf:pma:eq:feasible-set}
\end{equation}
where action $0$ denotes exact \tfPmaNull{}. Authorization, protection, schema, version, tenant, and snapshot checks are hard predicates. A learned score cannot restore an excluded action.

\subsection{Bounded, acyclic deployable information}

Fix before the run a versioned observation map $\Psi_{\mathrm{obs}}$.  It emits
\begin{equation}
O_{n,j}=\Psi_{\mathrm{obs}}\!\left(
B_n,W_{n,j},\{\rho^{\mathrm{read}}_{n,j,r}:r\in\mathcal R_{n,j}\};
v_{\mathrm{obs}}\right)\in\mathcal O,
\label{tf:pma:eq:bounded-observation}
\end{equation}
where $W_{n,j}$ is a fixed-width suffix of the completed segment and every
nonlocal read has an authenticated receipt $\rho^{\mathrm{read}}$ whose bytes
and latency are charged.  The schema of $\mathcal O$, the suffix width, and the
number and size of receipts are finite and frozen.  The raw prefix
$X_{n,\le b_n}$ is therefore not a deployable input.

Let $t_{n,j}^{-}$ be the instant before candidate construction.  The
preconstruction sigma-field and state are
\begin{align}
\tfPmaCalD^-_{n,j}&=\sigma\!\left(
I_n,O_{n,j},e_{n,\le j},M_{n,j-1},
\mathfrak R_{n,j}^{\mathrm{remaining}},\Xi^-_{n,j},
\{\zeta_{n,j,a}\}_{a=1}^{S}\right)
\subseteq\tfPmaCalF_{t_{n,j}^{-}},
\label{tf:pma:eq:deploy-sigma-minus}\\
S^-_{n,j}&=\Psi^-_{\mathrm{dep}}(\tfPmaCalD^-_{n,j})\in\tfPmaCalS^-.
\label{tf:pma:eq:deploy-summary-minus}
\end{align}
Construction in Equation~\eqref{tf:pma:eq:candidate} uses only this state.  After all
candidates are constructed and hard-validated, their constructor and validator
receipts enlarge the information set to
\begin{align}
\tfPmaCalD^+_{n,j}&=\tfPmaCalD^-_{n,j}\vee\sigma\!\left(
\{c_{n,j,a},\rho^{\mathrm{con}}_{n,j,a},
\rho^{\mathrm{val}}_{n,j,a}\}_{a=1}^{S},\tfPmaCalA_{n,j}\right),
\label{tf:pma:eq:deploy-sigma-plus}\\
S^+_{n,j}&=\Psi^+_{\mathrm{dep}}(\tfPmaCalD^+_{n,j})\in\tfPmaCalS^+,
\qquad \widetilde S_{n,j}:=S^+_{n,j}.
\label{tf:pma:eq:deploy-summary-plus}
\end{align}
This order is acyclic: no candidate is used to define the state from which that
same candidate is constructed.  Scoring and action selection use $S^+$; the
post-commit memory becomes readable only afterward.

\tfPmaDefinitionStatus
\begin{tfPmaDefinition}[Future-blind deployment]
A deployed scorer or decision rule is future blind at event $(n,j)$ if its output is $\tfPmaCalD^+_{n,j}$-measurable and its executable input graph has no continuation, future token, future answer, future evaluator output, branch identifier, or statistic computed from any of those values.
\end{tfPmaDefinition}

\tfPmaPrescriptionStatus\quad Declaring an input ``history only'' is insufficient. A conformance audit must enumerate every feature producer, cached tensor, RNG state, retrieved artifact, label join, and preprocessing path that can reach the deployment call.

\subsection{Core assumptions}

\begin{tfPmaAssumption}[Causal timing]
\label{tf:pma:assump:causal}
Every event decision is $\tfPmaCalD^+_{n,j}$-measurable, and the committed state becomes readable only after the decision time. Candidate construction is $\tfPmaCalD^-_{n,j}$-measurable. No future branch affects eventization, construction, feasibility, or deployment features.
\end{tfPmaAssumption}

\begin{tfPmaAssumption}[External row contract]
\label{tf:pma:assump:row}
For every accepted action, the constructor and trusted assembler return one complete row bound to the current snapshot; \tfPmaNull{} is exact identity; hard validation is deterministic for fixed inputs; and one event changes at most its selected target.
\end{tfPmaAssumption}

\begin{tfPmaAssumption}[Finite action and loss]
\label{tf:pma:assump:bounded}
$1\le|\tfPmaCalA_{n,j}|<\infty$ and every registered horizon loss lies in $[0,L_{\max}]$. If a natural loss is unbounded, the protocol fixes clipping or a different tail model before evaluation.
\end{tfPmaAssumption}

\begin{tfPmaAssumption}[Frozen total order]
\label{tf:pma:assump:tie-order}
The event manifest fixes a deterministic total order $\prec_{n,j}$ over encoded
action records, including \tfPmaNull{}.  Every set-valued optimization returns the
$\prec_{n,j}$-least minimizer.  The order is fixed before outcomes open and is
included in the protocol hash.
\end{tfPmaAssumption}

These assumptions support later results but are not asserted to hold in the frozen system. In particular, Assumption~\ref{tf:pma:assump:row}'s semantic precondition is presently unvalidated.

%% file: theory_family/modules/pma/sections/03_event_recursion.tex
\section{Event-Indexed Candidates, Actions, and Teacher Timing}
\label{tf:pma:sec:event-recursion}

\subsection{Deployed recursion}

Set $M_{n,0}=M_{n-1}$. After constructing $\tfPmaCalA_{n,j}$ against $M_{n,j-1}$, a deployed decision rule selects the $\prec_{n,j}$-resolved action $A_{n,j}\in\tfPmaCalA_{n,j}$ using only $\tfPmaCalD^+_{n,j}$ and commits
\begin{equation}
M_{n,j}=T_{n,j}(M_{n,j-1},A_{n,j})
=\begin{cases}
M_{n,j-1},&A_{n,j}=0,\\
\tfPmaCommit(M_{n,j-1},A_{n,j},c_{n,j,A_{n,j}}),&A_{n,j}\neq0.
\end{cases}
\label{tf:pma:eq:deployed-recursion}
\end{equation}
The segment output is $M_n=M_{n,J_n}$. Later events see earlier committed events through this state and no other within-segment authority.

\tfPmaProvedStatus
\begin{tfPmaProposition}[Event-timing consistency]
\label{tf:pma:prop:event-timing}
Under Assumptions~\ref{tf:pma:assump:causal} and~\ref{tf:pma:assump:row}, $M_{n,j}$ is measurable with respect to information available immediately after event $j$, and every candidate used at event $j+1$ is a function of $M_{n,j}$ rather than an obsolete pre-segment state.
\end{tfPmaProposition}

\begin{proof}
The base state $M_{n,0}$ is available before the first decision. Given an available $M_{n,j-1}$, Equation~\eqref{tf:pma:eq:deploy-sigma-minus} fixes the preconstruction inputs; construction and validation then produce $\tfPmaCalD^+_{n,j}$ without reading a post-commit value.  Action selection and commit are causal by Assumption~\ref{tf:pma:assump:causal}; hence $M_{n,j}$ is available after the event. Equation~\ref{tf:pma:eq:candidate} then uses that exact state for event $j+1$. Induction over $j$ proves the claim.
\end{proof}

The proposition prohibits a convenient but invalid shortcut: construct all segment candidates against $M_{n,0}$, choose several actions, and describe the result as event-recursive. Earlier commits can change target versions, feasibility, collision status, retained content, and later losses.

\subsection{Branch-prefix recursion for a teacher}

To price counterfactual action sequences, let $\mathbf a_{1:j}=(a_1,\ldots,a_j)$. Each branch has its own state
\begin{equation}
M_{n,0}^{()}=M_{n-1},\qquad
M_{n,j}^{(\mathbf a_{1:j})}
=T_{n,j}^{(\mathbf a_{<j})}
\left(M_{n,j-1}^{(\mathbf a_{<j})},a_j\right),
\label{tf:pma:eq:branch-recursion}
\end{equation}
where event summary, candidate set, and feasibility at depth $j$ are recomputed against the branch prefix $M_{n,j-1}^{(\mathbf a_{<j})}$. The teacher may use one of three distinct protocols:
\begin{enumerate}
  \item \textbf{Exact dynamic program:} enumerate every feasible branch prefix and use a future-blind continuation policy optimized outside the future expectation.
  \item \textbf{Frozen rollout:} after the scored action, follow one preregistered future-blind continuation policy for later segment events.
  \item \textbf{Single-event intervention:} set $J_n=1$ or forbid later writes in the scoring horizon. This is the cleanest first causal test.
\end{enumerate}
Mixing these protocols is not allowed. In particular, choosing later actions separately for each revealed future is a hindsight policy and is not a deployable cost-to-go.

For a sequence policy $\pi$ whose event-$q$ action is measurable with respect to the branch state before event $q$, define
\begin{equation}
Q^{\mathrm T}_{n,j}(a\mid\widetilde S_{n,j}^{(\mathbf a_{<j})})
=\inf_{\pi\in\Pi^{\mathrm{fb}}_{j+1:J_n}}
\tfPmaE\!\left[
L_H\bigl((\mathbf a_{<j},a,A_{j+1:J_n}^{\pi}),F\bigr)
\tfPmaGiven \widetilde S_{n,j}^{(\mathbf a_{<j})}
\right].
\label{tf:pma:eq:teacher-cost-to-go}
\end{equation}
The infimum is outside the expectation over $F$. Moving it inside would let the teacher observe the realized future before selecting later events.  If the infimum is attained by more than one policy, the manifest's frozen lexicographic policy order resolves the tie; no future outcome may do so.

\begin{figure}[H]
\centering
\includegraphics[width=\linewidth]{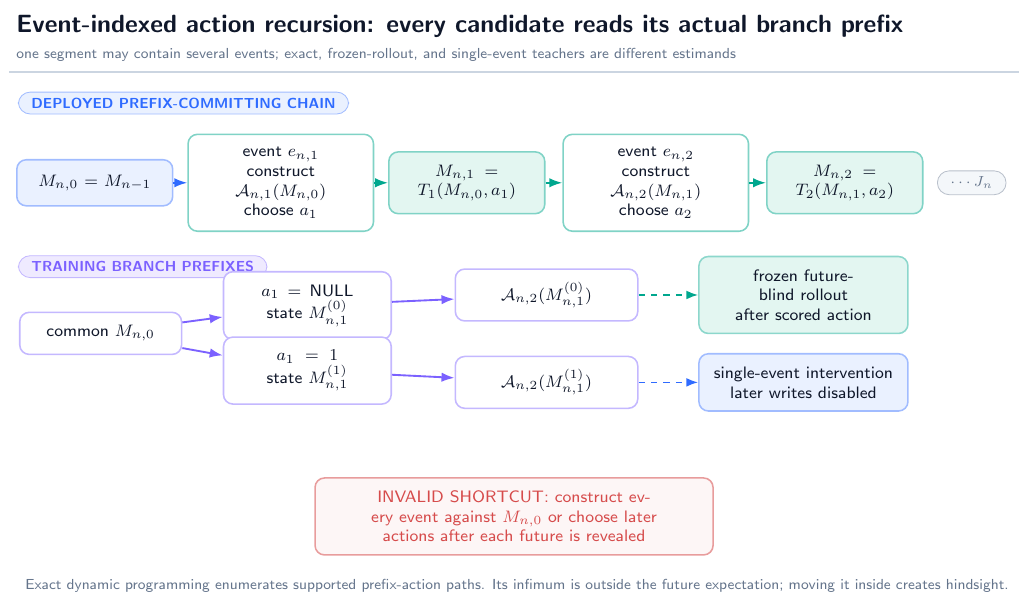}
\caption{\textbf{Event-indexed recursion (\tfPmaDefinitionStatus{}).} Candidate/action sets at event $j$ are constructed from the branch state after earlier commits. Exact, frozen-rollout, and single-event teachers are distinct. No whole-segment action may reuse candidates from an obsolete prefix.}
\label{tf:pma:fig:event-recursion}
\end{figure}

\subsection{No conflated segment action}

\tfPmaProvedStatus
\begin{tfPmaTheorem}[Sequential composition and support]
\label{tf:pma:thm:sequential-support}
Under Assumption~\ref{tf:pma:assump:row}, each legal event changes at most one resident row. Across a segment of $J_n$ events, the final state can differ from $M_{n-1}$ on at most $J_n$ distinct slots. A later \tfPmaNull{} leaves earlier committed prefixes unchanged.
\end{tfPmaTheorem}

\begin{proof}
At one event, \tfPmaNull{} changes zero rows and a non-NULL action replaces exactly its selected row. The union of target slots over $J_n$ events therefore has size at most $J_n$, and every row outside that union remains unchanged. Equation~\ref{tf:pma:eq:deployed-recursion} is prefix committing, so an identity transition at event $j$ maps $M_{n,j-1}$ to itself rather than to $M_{n-1}$.
\end{proof}

This theorem establishes a mechanical support bound only. It does not show that a one-row change is semantically local, that a segment is all-or-nothing, or that a teacher can safely ignore later events.

\subsection{Required timing receipt}

\tfPmaPrescriptionStatus\quad Every future implementation must record, per event and per counterfactual branch: prefix-state digest, event digest, constructor/assembler version, candidate digests, feasibility mask, action, commit or \tfPmaNull{} result, successor digest, future-group identifier, rollout protocol, and resource increments. A candidate whose recorded predecessor differs from the replay prefix is invalid even if its bytes parse.

%% file: theory_family/modules/pma/sections/04_grouped_futures.tex
\section{Grouped Alternative Futures and Sampling Assumptions}
\label{tf:pma:sec:grouped-futures}

\subsection{Physical prefix groups}

Fix one post-candidate/pre-action decision state $\widetilde S_g^+$ and its immutable pre-action memory snapshot $M_g$. A grouped-future record is
\begin{equation}
\tfPmaCalG_g=\left(
g,\widetilde S_g^+,M_g,\tfPmaCalA_g,
\{c_{g,a}\}_{a\in\tfPmaCalA_g\setminus\{0\}},
\{(F_{g,k},w_{g,k})\}_{k=1}^{K_g}
\right),
\label{tf:pma:eq:group-record}
\end{equation}
where every branch shares exactly the same causal prefix, candidates, and pre-action snapshot; $w_{g,k}\ge0$ and $\sum_k w_{g,k}=1$. The group ID, not a future branch, is the split unit.

\tfPmaDefinitionStatus
\begin{tfPmaDefinition}[Grouped future]
A grouped future is a continuation sampled or constructed for one fixed causal prefix state. ``Similar prompts,'' paraphrases with different hidden states, or branches joined only after answer labels are known are not grouped futures without an explicit causal equivalence argument.
\end{tfPmaDefinition}

\begin{figure}[H]
\centering
\includegraphics[width=\linewidth]{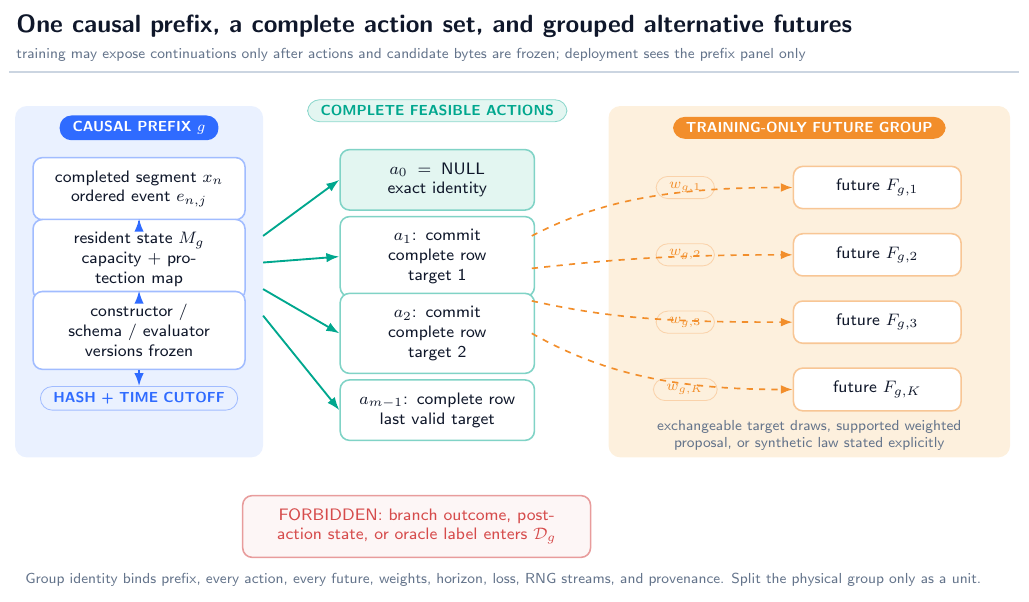}
\caption{\textbf{Causal prefix and grouped futures.} Dashed amber branches are training/evaluation-only continuations from one frozen prefix. Solid navy and teal paths are deployable. Future tokens may price actions but no future-derived field crosses the deployment boundary.}
\label{tf:pma:fig:causal-futures}
\end{figure}

\subsection{Sampling models}

Let $P_g$ be the intended deployment continuation distribution conditional on $\widetilde S_g^+$. Three data regimes must be distinguished.

Before any outcome is opened, the group record fixes one sampling flag
\begin{equation}
\mathsf{samp}_g\in\{\mathsf{iid},\mathsf{stratified},\mathsf{finite}\}.
\label{tf:pma:eq:sampling-flag}
\end{equation}
The flag is part of the estimand: it selects the estimator, its center, and its finite-sample argument. It may not be changed after inspecting an action advantage.

\begin{table}[H]
\centering
\small
\begin{tabularx}{\linewidth}{@{}p{0.20\linewidth}Y Y@{}}
\toprule
Regime & Identifying requirement & Mandatory limitation \\
\midrule
Repeated real continuations & Branches are conditionally exchangeable draws from $P_g$ or a declared finite population & Repeated collection may alter users or environment \\
Controlled generator & Generator seed and branch law are fixed before action outcomes; support covers the declared target population & Identifies the generator distribution, not natural deployment \\
Model-generated futures & Generator is frozen and causally separated from action scoring; calibration compares its branch law with fresh real continuations & Self-consistency and mode collapse may bias values \\
Logged observational futures & Every evaluated action has positive logging support, propensities and confounders are recorded, and a valid off-policy estimator is used & Missing support cannot be repaired by modeling alone \\
\bottomrule
\end{tabularx}
\caption{Grouped-future regimes. Synthetic and model-generated futures require an explicit target-distribution boundary.}
\label{tf:pma:tab:future-regimes}
\end{table}

\begin{tfPmaAssumption}[Conditional exchangeability in the iid regime]
\label{tf:pma:assump:exchangeability}
If $\mathsf{samp}_g=\mathsf{iid}$, then $(F_{g,1},\ldots,F_{g,K_g})$ are conditionally independent draws from one declared branch law $Q_g(\cdot\mid\widetilde S_g^+)$, and the nonnegative weights are fixed independently of their realized losses. If this statement is false, the iid estimator and its concentration theorem are inapplicable.
\end{tfPmaAssumption}

If $\mathsf{samp}_g=\mathsf{stratified}$, the stratum map, sampling fractions, and target stratum masses are frozen, and risk is estimated by the corresponding stratum-weighted estimator with a stratum-aware variance or concentration calculation. If $\mathsf{samp}_g=\mathsf{finite}$, the declared estimand is a finite-population mean, branches are sampled without replacement by a frozen design, and the analysis uses a without-replacement bound (including its finite-population correction). Neither regime is silently relabeled iid.

\begin{tfPmaAssumption}[Action support and consistency]
\label{tf:pma:assump:support}
Every action whose risk is estimated is feasible at the common snapshot and has a well-defined potential replay outcome. The replay engine returns that outcome when the corresponding action is applied. If logged data rather than full replay are used, the logging propensity is positive wherever the target action has positive probability.
\end{tfPmaAssumption}

\begin{tfPmaAssumption}[Exogenous or interventional future]
\label{tf:pma:assump:future-causal}
Either the future evaluation sequence is conditionally exogenous to the memory action, so the same branch can be replayed under all actions, or the environment supplies valid action-conditioned counterfactual futures. Reusing one observed future under all actions is not valid when the action changes the future-generating process.
\end{tfPmaAssumption}

These are causal-identification assumptions in the potential-outcome sense \cite{tf:pma:rubin1974,tf:pma:rosenbaum1983}. They are stricter than ordinary predictive train/test separation.

\subsection{Branch weights and effective sample size}

In the iid regime, weights may encode fixed quadrature or replication weights and must not depend on realized losses. In the stratified and finite-population regimes, they are the design weights required by the frozen estimand. They must not be tuned to increase the preferred action's advantage. Define the descriptive concentration diagnostic
\begin{equation}
K_{\mathrm{eff},g}=\frac{1}{\sum_{k=1}^{K_g}w_{g,k}^2},
\label{tf:pma:eq:effective-k}
\end{equation}
which equals $K_g$ for uniform weights and approaches one when a single branch dominates. Reporting only $K_g$ can therefore overstate information. Equation~\ref{tf:pma:eq:effective-k} is not, by itself, a license to reuse the iid theorem under stratification, clustering, or sampling without replacement.

If branches are sampled under proposal law $Q_g$ but target risk uses $P_g$, importance weights require $P_g\ll Q_g$ and bounded or otherwise controlled density ratios. Doubly robust off-policy estimators can combine a reward model with propensities \cite{tf:pma:dudik2011}; they do not solve zero support or hidden confounding.

\subsection{Synthetic-future bias}

For any action loss bounded in $[0,L_{\max}]$, define total variation as
\begin{equation}
\tfPmaTV(P,Q)=\sup_{A}|P(A)-Q(A)|.
\end{equation}
Then the risk under a synthetic branch law can differ from deployment risk even with infinitely many synthetic samples. Section~\ref{tf:pma:sec:identification} proves the bound
\begin{equation}
|R_P(a)-R_Q(a)|\le L_{\max}\tfPmaTV(P,Q).
\label{tf:pma:eq:tv-preview}
\end{equation}
The sampling error shrinks with more branches; the distributional bias does not. A model-generated teacher is therefore not an oracle for the real future unless the target relation is independently justified.

\subsection{Split obligations}

\tfPmaPrescriptionStatus\quad Fitting, calibration, and final evaluation must be disjoint at the prefix-family or generating-process level. All branches from one group remain together. Near-duplicate prefixes, shared answer templates, generator IDs, and deterministic seed descendants are audited before labels are opened. Final-only means one frozen model, threshold, horizon, loss vector, and confidence procedure are read once on the held-out groups; a failed seed is never replaced.

%% file: theory_family/modules/pma/sections/05_replay_loss.tex
\section{Action-Conditioned Replay and Horizon Loss}
\label{tf:pma:sec:replay-loss}

\subsection{One immutable snapshot, every feasible action}

For group $g$, action $a\in\tfPmaCalA_g$, and future branch $F_{g,k}$, replay begins from the same immutable pre-action snapshot $M_g$. Define
\begin{equation}
M_g^{(a)}=
\begin{cases}
M_g,&a=0,\\
\tfPmaCommit(M_g,a,c_{g,a}),&a\neq0.
\end{cases}
\label{tf:pma:eq:action-state}
\end{equation}
The continuation is then executed under $M_g^{(a)}$. A cached hidden trace computed before the action and merely rescored afterward is not action-conditioned replay when memory can affect future hidden-state evolution.

The complete replay array is
\begin{equation}
\mathsf Y_g=\left(Y_{g,a,k,h}:a\in\tfPmaCalA_g,\ 1\le k\le K_g,\ 1\le h\le H\right),
\label{tf:pma:eq:replay-tensor}
\end{equation}
with action, future, and horizon axes. \tfPmaNull{} occupies a full action slice. It is not a missing baseline.

\begin{figure}[H]
\centering
\includegraphics[width=\linewidth]{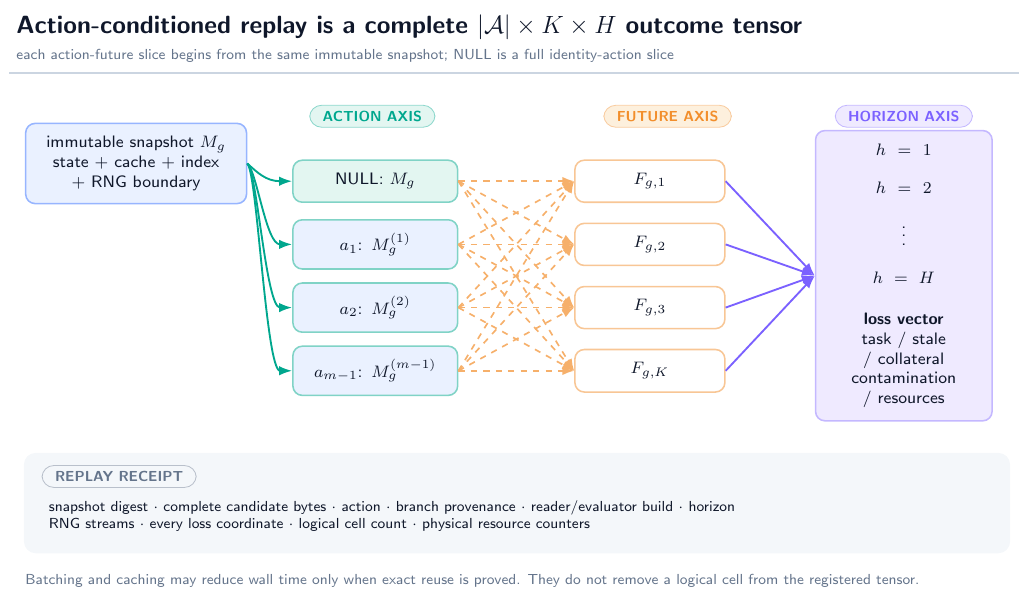}
\caption{\textbf{Action-conditioned replay tensor (\tfPmaDefinitionStatus{} / training only).} Every feasible write and \tfPmaNull{} starts from one immutable snapshot, is evaluated on every grouped future, and contributes all registered horizon terms. The visible $|\tfPmaCalA|\times K\times H$ lattice is logical work even when kernels batch it.}
\label{tf:pma:fig:replay-tensor}
\end{figure}

\subsection{Loss vector and scalarization}

For horizon step $h$, let
\begin{equation}
\ell_{g,a,k,h}=
\left(
\ell^{\mathrm{task}},
\ell^{\mathrm{stale}},
\ell^{\mathrm{coll}},
\ell^{\mathrm{cont}},
\ell^{\mathrm{sec}},
\ell^{\mathrm{lat}},
\ell^{\mathrm{traffic}}
\right)_{g,a,k,h}\in\mathbb R_+^7.
\label{tf:pma:eq:loss-vector}
\end{equation}
The coordinates respectively measure task error, obsolete-value leakage, damage to unrelated information, admitted-content contamination, security/policy harm not already removed by hard infeasibility, latency, and memory/data traffic. The registered loss schema includes deterministic componentwise caps
\begin{equation}
0\le \ell_{g,a,k,h}\le \overline\ell\in\mathbb R_+^7,
\quad
0\le D^{\mathrm{miss}}_{g,k}\le \overline D_{\mathrm{miss}},
\quad
0\le\mathfrak R_{g,a,k}\le\overline{\mathfrak R},
\label{tf:pma:eq:loss-bounds}
\end{equation}
where inequalities are componentwise and every cap is fixed from the measurement protocol before calibration. Values outside the registered domain are an audit failure; they are not clipped after outcome inspection. The event-level loss is
\begin{align}
L_g(a,F_{g,k})={}&
\sum_{h=1}^{H}\gamma^{h-1}
\left\langle\lambda,\ell_{g,a,k,h}\right\rangle
+\lambda_{\mathrm{write}}\tfPmaOne[a\neq0]
\nonumber\\
&+\lambda_{\mathrm{null}}\tfPmaOne[a=0]D^{\mathrm{miss}}_{g,k}
+\left\langle\lambda_{\mathrm{res}},
\mathfrak R_{g,a,k}\right\rangle,
\label{tf:pma:eq:aggregate-loss}
\end{align}
where $\gamma\in(0,1]$, all weights are nonnegative and fixed before evaluation, $D^{\mathrm{miss}}$ prices a missed useful update without presuming every \tfPmaNull{} is wrong, and $\mathfrak R$ is the replay resource vector. A hard safety violation should normally remove an action before scoring rather than be traded against task loss.

The missed-update coordinate is disjoint by construction: it may encode only opportunity loss not already represented in task, stale, collateral, contamination, security, latency, traffic, or resource coordinates. If the evaluator cannot establish that disjointness, $\lambda_{\mathrm{null}}$ is fixed to zero. With $\Gamma_H=\sum_{h=1}^{H}\gamma^{h-1}$, the declared scalarization therefore satisfies the auditable range
\begin{equation}
0\le L_g(a,F)\le
\Gamma_H\langle\lambda,\overline\ell\rangle
+\lambda_{\mathrm{write}}
+\lambda_{\mathrm{null}}\overline D_{\mathrm{miss}}
+\langle\lambda_{\mathrm{res}},\overline{\mathfrak R}\rangle
=:L_{\max}.
\label{tf:pma:eq:lmax-audit}
\end{equation}
The replay validator recomputes every coordinate and this bound from the frozen schema; a violation invalidates the group rather than expanding $L_{\max}$ retrospectively.

\begin{table}[H]
\centering
\footnotesize
\begin{tabularx}{\linewidth}{@{}p{0.20\linewidth}Y Y@{}}
\toprule
Term & Operational measurement & Boundary \\
\midrule
Task loss & Frozen evaluator on future task/output & Evaluator error audited separately \\
Stale leakage & Old value emitted, selected, or remains authoritative when superseded & Must distinguish unknown from stale \\
Collateral damage & Positive loss change on unrelated facts/tasks versus the common snapshot & Neural cross-row effects measured, not assumed absent \\
Contamination & Severity of false, stale, malicious, or wrongly scoped authority & Requires truth/source contract \\
Write charge & Fixed cost for any committed row & Prevents free churn \\
\tfPmaNull{} miss & Opportunity cost only when future evidence establishes a useful feasible write & Does not make abstention an error by default \\
Resources & Tokens, FLOPs, bytes, traffic, latency, energy, archive/index and validation work & Teacher and deployment ledgers stay separate \\
\bottomrule
\end{tabularx}
\caption{Loss contract. Every coordinate and scalarization weight is preregistered; changing it changes the estimand.}
\label{tf:pma:tab:loss-contract}
\end{table}

\tfPmaDefinitionStatus
\begin{tfPmaDefinition}[Target, proposal, and empirical horizon risk]
For target continuation law $P_g(\cdot\mid\widetilde S_g^+)$,
\begin{equation}
R_{P_g}(a)=\tfPmaE_{F\sim P_g}\left[L_g(a,F)\tfPmaGiven\widetilde S_g^+\right].
\label{tf:pma:eq:true-risk}
\end{equation}
For the proposal law $Q_g(\cdot\mid\widetilde S_g^+)$,
\begin{equation}
R_{Q_g}(a)=\tfPmaE_{F\sim Q_g}\left[L_g(a,F)\tfPmaGiven\widetilde S_g^+\right].
\label{tf:pma:eq:proposal-risk}
\end{equation}
The raw grouped empirical proposal risk is
\begin{equation}
\widehat R_{Q_g}(a)=\sum_{k=1}^{K_g}w_{g,k}L_g(a,F_{g,k}).
\label{tf:pma:eq:empirical-risk}
\end{equation}
\end{tfPmaDefinition}

The notation separates sampling and target distributions. When $Q_g=P_g$ and the iid assumptions hold, $\widehat R_{Q_g}$ estimates target risk. When $Q_g\neq P_g$, it estimates proposal-law risk unless a separately declared support-valid correction or shift certificate maps it to $P_g$. For compactness below, an unsubscripted $R_g$ always means $R_{P_g}$; an unsubscripted empirical risk is never used.

\subsection{Paired action differences}

Because all actions share each future branch, the paired difference
\begin{equation}
D_{g,k}(a,b)=L_g(a,F_{g,k})-L_g(b,F_{g,k})
\label{tf:pma:eq:paired-difference}
\end{equation}
has the sign convention ``positive means action $a$ is worse than action $b$.'' It directly estimates the proposal-law contrast $R_{Q_g}(a)-R_{Q_g}(b)$ and cancels branch difficulty. Its variance for uniform independent branches is
\begin{equation}
\frac{1}{K_g}\left[
\tfPmaVar(L_a)+\tfPmaVar(L_b)-2\tfPmaCov(L_a,L_b)
\right].
\label{tf:pma:eq:paired-variance}
\end{equation}
Pairing reduces variance when action losses are nonnegatively correlated on the same future. It does not repair a biased branch law or invalid counterfactual replay.

\subsection{Replay integrity}

\tfPmaPrescriptionStatus\quad A valid replay record binds the source snapshot, complete candidate bytes, action, future branch, reader/evaluator build, horizon, RNG streams, loss coordinates, and resource increments. Action order is randomized or counterbalanced when shared hardware state could create interference. Each branch begins from an independently restored snapshot; caches and derived indexes are either restored or rebuilt under the selected action. A replay that omits \tfPmaNull{}, prunes an action after seeing its outcome, or lets one branch inherit another branch's state is invalid.

%% file: theory_family/modules/pma/sections/06_oracle_gate.tex
\section{Expected-Risk Oracle, Current-Risk Comparator, and Qualification Gate}
\label{tf:pma:sec:oracle-gate}

\subsection{Three distinct decision objects}

The empirical proposal-risk oracle for group $g$ is
\begin{equation}
\widehat a_g^{\mathrm{ER},Q}
=\min_{\prec_{g}}\arg\min_{a\in\tfPmaCalA_g}\widehat R_{Q_g}(a).
\label{tf:pma:eq:empirical-oracle}
\end{equation}
It is an evaluation rule using training-only future branches, with the frozen protocol order $\prec_g$ resolving every tie. It is not deployed, it is not a learned policy, and it is not a target-risk oracle unless a valid $Q_g$--$P_g$ bridge is supplied below.

The current-risk comparator is fixed independently:
\begin{equation}
a_g^{\mathrm{cur}}
=\min_{\prec_g}\arg\min_{a\in\tfPmaCalA_g}C_g^{\mathrm{cur}}(a;\widetilde S_g^+),
\label{tf:pma:eq:current-comparator}
\end{equation}
where $C^{\mathrm{cur}}$ may include current reconstruction, verification, immediate stale suppression, and write cost but no realized continuation. Its exact definition, tie rule, and parameters are preregistered. \tfPmaNull{} is also reported independently, even if $a^{\mathrm{cur}}=0$.

The comparator registration record binds the exact formula, the named fields of the postvalidation state $S_g^+$, all feature normalizations and missing-value rules, constructor/validator/evaluator versions, scalar weights, $\prec_g$, a configuration hash, and the registration timestamp. A field absent from that record cannot be introduced after future outcomes are observed.

Finally, the per-future hindsight action
\begin{equation}
a_{g,k}^{\mathrm{hind}}=\min_{\prec_g}\arg\min_a L_g(a,F_{g,k})
\label{tf:pma:eq:hindsight-action}
\end{equation}
is an unattainable information ceiling. Averaging its loss answers a different question from choosing one action that minimizes expected loss before the branch is known.

\subsection{Expected oracle gap}

Define group-level advantages over current risk and \tfPmaNull{}:
\begin{align}
G_g^{\mathrm{cur}}
&=R_{P_g}(a_g^{\mathrm{cur}})-\min_{a\in\tfPmaCalA_g}R_{P_g}(a),
\label{tf:pma:eq:gap-current}\\
G_g^{0}
&=R_{P_g}(0)-\min_{a\in\tfPmaCalA_g}R_{P_g}(a).
\label{tf:pma:eq:gap-null}
\end{align}
For a population of fresh prefix groups $g\sim\mathcal P$, the target gaps are $G^{\mathrm{cur}}=\tfPmaE_gG_g^{\mathrm{cur}}$ and $G^0=\tfPmaE_gG_g^0$. An oracle can have a positive advantage over current-risk writing but none over \tfPmaNull{}, or vice versa. Both comparisons are required.

\begin{figure}[H]
\centering
\includegraphics[width=\linewidth]{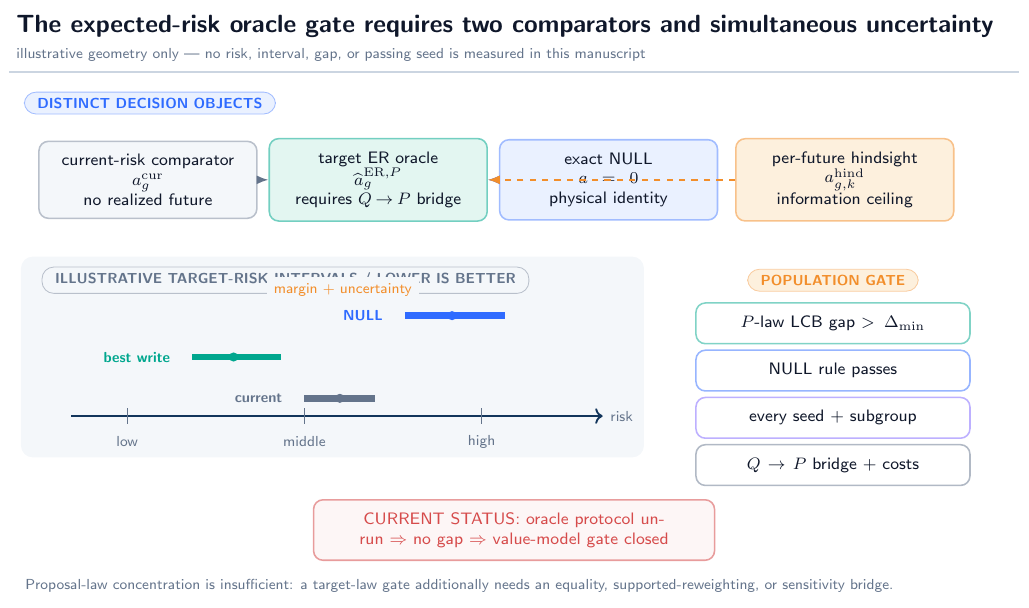}
\caption{\textbf{Oracle qualification geometry (\tfPmaPlannedStatus{}).} The empirical expected-risk oracle, current-risk comparator, and \tfPmaNull{} are distinct. A stable simultaneous lower confidence bound must exceed the preregistered materiality margin on fresh groups and every required seed; otherwise the controller gate stays closed. No such gap is reported here.}
\label{tf:pma:fig:oracle-gate}
\end{figure}

\subsection{Weighted concentration within one group}

\tfPmaProvedStatus
\begin{tfPmaTheorem}[Simultaneous weighted Hoeffding bound]
\label{tf:pma:thm:weighted-hoeffding}
Assume~\ref{tf:pma:assump:exchangeability}, let $m_g=|\tfPmaCalA_g|$, and suppose $0\le L_g(a,F)\le L_{\max}$. Conditional on $\widetilde S_g^+$, with probability at least $1-\delta$,
\begin{equation}
\max_{a\in\tfPmaCalA_g}
\left|\widehat R_{Q_g}(a)-R_{Q_g}(a)\right|
\le
L_{\max}
\sqrt{
\frac{\sum_{k=1}^{K_g}w_{g,k}^2}{2}
\log\frac{2m_g}{\delta}
}
=:\varepsilon_g(\delta).
\label{tf:pma:eq:weighted-hoeffding}
\end{equation}
\end{tfPmaTheorem}

\begin{proof}
For fixed $a$, weighted Hoeffding gives
\(
\tfPmaP(|\sum_k w_k(L_{a,k}-\tfPmaE L_{a,k})|>\epsilon)
\le2\exp[-2\epsilon^2/(L_{\max}^2\sum_k w_k^2)].
\)
A union bound over $m_g$ actions makes the failure probability at most $\delta$. Solving for $\epsilon$ gives Eq.~\ref{tf:pma:eq:weighted-hoeffding}.
\end{proof}

The bound controls action selection and makes the multiple-action penalty explicit. It centers on $R_{Q_g}$, not $R_{P_g}$, when the branch law differs from deployment.

\tfPmaProvedStatus
\begin{tfPmaCorollary}[Stable empirical action]
\label{tf:pma:cor:stable-action}
Let $a_g^*$ be the unique minimizer of $R_{Q_g}$ and
\begin{equation}
\Delta_g=\min_{a\neq a_g^*}
\left(R_{Q_g}(a)-R_{Q_g}(a_g^*)\right).
\end{equation}
If $\Delta_g>2\varepsilon_g(\delta)$, then $\widehat a_g^{\mathrm{ER},Q}=a_g^*$ with probability at least $1-\delta$.
\end{tfPmaCorollary}

\begin{proof}
On the simultaneous event, the optimal estimate can increase by at most $\varepsilon_g$ and a competitor can decrease by at most $\varepsilon_g$. A true gap larger than $2\varepsilon_g$ cannot reverse.
\end{proof}

Variance-sensitive empirical Bernstein intervals may be used under their stated conditions \cite{tf:pma:maurer2009}, but the estimator, finite-sample constants, action multiplicity, and calibration split must be frozen. Selecting whichever interval makes the gap significant is not valid.

\subsection{Mandatory proposal-to-target bridge}

Proposal-law concentration alone cannot certify a deployment-law advantage. For every group, exactly one of the following bridge records must be frozen before qualification:
\begin{enumerate}
  \item \textbf{Law equality:} a design argument and audit establish $Q_g=P_g$ on the declared continuation space.
  \item \textbf{Support-valid reweighting:} $P_g\ll Q_g$, the density ratio $r_g=\mathrm dP_g/\mathrm dQ_g$ and an upper bound $0\le r_g\le\overline r_g<\infty$ are fixed independently of outcomes, and a registered importance-weighted estimator with its own simultaneous finite-sample interval is used. The unweighted bound in Theorem~\ref{tf:pma:thm:weighted-hoeffding} is not reused.
  \item \textbf{Certified shift set:} an external, held-out procedure provides $d_g$ such that $\tfPmaTV(P_g,Q_g)\le d_g$. If $[L_g^Q(a),U_g^Q(a)]$ is a simultaneous proposal-risk interval, the target interval is conservatively inflated to
  \begin{equation}
  [L_g^P(a),U_g^P(a)]=
  [\max\{0,L_g^Q(a)-L_{\max}d_g\},
   \min\{L_{\max},U_g^Q(a)+L_{\max}d_g\}].
  \label{tf:pma:eq:tv-bridge-interval}
  \end{equation}
\end{enumerate}
The shift certificate declares whether generator, environment, and evaluator discrepancies are included. A discrepancy already absorbed into $d_g$ may not be charged again as an independent uncertainty bonus. If no bridge applies, target intervals are undefined and the qualification gate is closed.

\subsection{Population gate}

\begin{tfPmaProtocol}[Oracle-gap qualification]
\label{tf:pma:prot:oracle-gate}
\tfPmaPlannedStatus{} For each fixed seed, estimate paired group-level target-risk advantages on a fresh namespace using one frozen bridge above. Construct simultaneous confidence intervals for current-risk and \tfPmaNull{} advantages across actions, primary subgroups, horizons, and required seeds using a preregistered group-level procedure. The expected-risk oracle qualifies only if:
\begin{enumerate}
  \item the proposal-to-target bridge is valid and its uncertainty is included exactly once;
  \item the lower confidence bound for $G^{\mathrm{cur}}$ exceeds materiality margin $\Delta_{\min}^{\mathrm{cur}}>0$;
  \item the registered relation to \tfPmaNull{} passes its separate noninferiority or superiority rule;
  \item action rankings, sign, stale leakage, and collateral damage are stable on every required seed and primary subgroup;
  \item no leakage, support, evaluator, or replay audit fails; and
  \item full teacher and deployment costs are published.
\end{enumerate}
If any condition fails, $Q_\psi$ is not trained.
\end{tfPmaProtocol}

\tfPmaUnvalidatedStatus\quad Protocol~\ref{tf:pma:prot:oracle-gate} has not been executed. This manuscript defines the gate and proves conditional properties; it does not claim an oracle advantage of any magnitude.

%% file: theory_family/modules/pma/sections/07_uncertainty.tex
\section{Uncertainty-Aware Conservative Admission}
\label{tf:pma:sec:uncertainty}

\subsection{Simultaneous intervals}

The registration declares one and only one simultaneous confidence family
\begin{equation}
\mathcal J=\{(s,g,z,h,a):s\in\mathcal S_{\rm req},\ g\in\mathcal G_{\rm final},\ z\in\mathcal Z_{\rm primary},\ h\in\mathcal H_{\rm reg},\ a\in\tfPmaCalA_g\},
\label{tf:pma:eq:unique-family}
\end{equation}
including every required seed, final prefix group, primary subgroup, registered horizon summary, and feasible action. The family, its alpha allocation, and its proposal-to-target bridge are frozen together; overlapping ``primary'' families or a second, more favorable correction are not permitted.

For each action, let $[L_g^P(a),U_g^P(a)]$ be the resulting interval for target risk $R_{P_g}(a)$, constructed without final-set tuning. The required actionwise statement below is a projection of the joint event over $\mathcal J$:
\begin{equation}
\tfPmaP\left(R_{P_g}(a)\in[L_g^P(a),U_g^P(a)]\ \text{for every }a\in\tfPmaCalA_g\right)\ge1-\delta.
\label{tf:pma:eq:simultaneous-coverage}
\end{equation}
Marginal 95\% intervals for many actions do not imply 95\% coverage of the selected action. Bonferroni, finite-class concentration, or another frozen simultaneous procedure must charge action selection. If thresholds are also chosen on calibration data, final coverage is evaluated on untouched groups.

Let $\tfPmaCalW_g=\tfPmaCalA_g\setminus\{0\}$. The conservative write candidate is defined only when a feasible write exists:
\begin{equation}
i_g^+=
\begin{cases}
\min_{\prec_g}\arg\min_{a\in\tfPmaCalW_g}U_g^P(a),&\tfPmaCalW_g\neq\varnothing,\\
\bot,&\tfPmaCalW_g=\varnothing,
\end{cases}
\label{tf:pma:eq:best-ucb}
\end{equation}
where $\bot$ is a non-action sentinel and $\prec_g$ is the fixed tie rule. For required benefit margin $\mu\ge0$, define
\begin{equation}
\pi_{\mathrm{cert}}(\widetilde S_g^+)=
\begin{cases}
i_g^+,&i_g^+\neq\bot\ \text{and }U_g^P(i_g^+)+\mu<L_g^P(0),\\
0,&\text{otherwise}.
\end{cases}
\label{tf:pma:eq:certified-rule}
\end{equation}
If there is no feasible write, the rule returns \tfPmaNull{}.

\begin{figure}[H]
\centering
\includegraphics[width=\linewidth]{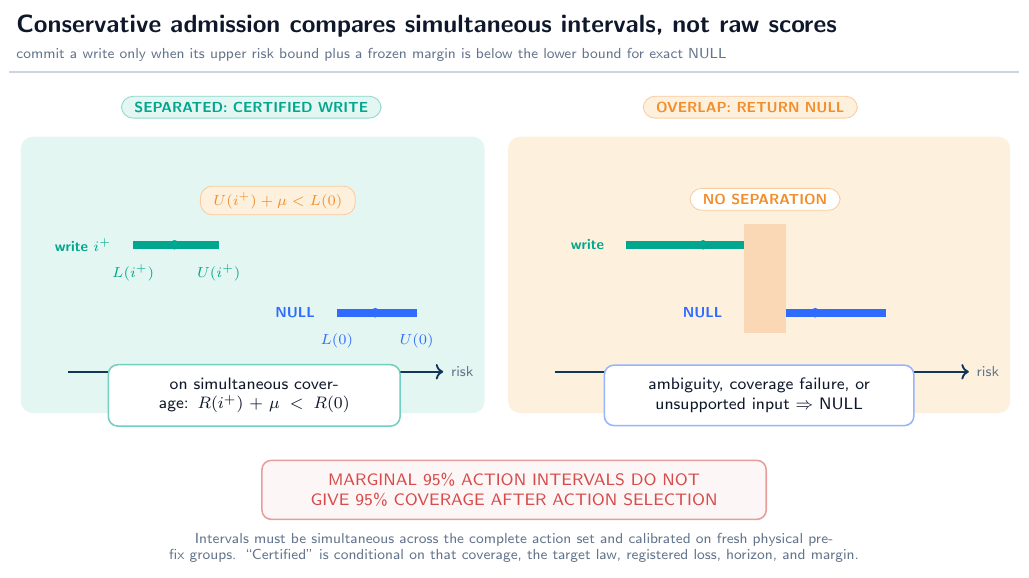}
\caption{\textbf{Conservative UCB/LCB admission (\tfPmaProvedStatus{} under simultaneous coverage).} The best feasible write commits only when its upper risk bound plus margin is below the lower bound for \tfPmaNull{}. Overlap, multiplicity, or failed calibration returns \tfPmaNull{}.}
\label{tf:pma:fig:uncertainty}
\end{figure}

\tfPmaProvedStatus
\begin{tfPmaTheorem}[Certified improvement over \tfPmaNull{}]
\label{tf:pma:thm:certified-null}
Under Eq.~\ref{tf:pma:eq:simultaneous-coverage}, every non-NULL action selected by Eq.~\ref{tf:pma:eq:certified-rule} satisfies
\begin{equation}
R_{P_g}(i_g^+)+\mu<R_{P_g}(0)
\label{tf:pma:eq:certified-improvement}
\end{equation}
with probability at least $1-\delta$.
\end{tfPmaTheorem}

\begin{proof}
On the simultaneous-coverage event,
\(
R_{P_g}(i_g^+)\le U_g^P(i_g^+)
\)
and
\(
L_g^P(0)\le R_{P_g}(0).
\)
The commit rule therefore yields
\(
R_{P_g}(i_g^+)+\mu\le U_g^P(i_g^+)+\mu<L_g^P(0)\le R_{P_g}(0).
\)
The event has probability at least $1-\delta$.
\end{proof}

This result certifies improvement over \tfPmaNull{} under the interval assumptions. It does not prove that $i_g^+$ is the globally best write. If target correctness matters, require the additional separation
\begin{equation}
U_g^P(i_g^+)+\eta<\min_{a\in\tfPmaCalW_g\setminus\{i_g^+\}}L_g^P(a),
\label{tf:pma:eq:target-certification}
\end{equation}
for a prespecified $\eta\ge0$, or report the action as beneficial but target-ambiguous. Here and below $\min\varnothing:=+\infty$; thus a sole feasible write has no unresolved write competitor, while an empty write set still returns \tfPmaNull{} by Eq.~\ref{tf:pma:eq:certified-rule}.

\subsection{NULL calibration}

\tfPmaNull{} is both a physical identity action and a statistical reject option, related to classical error-reject tradeoffs \cite{tf:pma:chow1970}. It needs its own calibration. A model can achieve low write error by returning \tfPmaNull{} everywhere, or high recall by writing everywhere. Required outputs include:
\begin{itemize}
  \item empirical interval coverage for \tfPmaNull{} and non-NULL actions separately;
  \item precision and recall of \tfPmaNull{} on preregistered write-worthy and null-worthy groups;
  \item risk-coverage curves as $\mu$ varies only on calibration data;
  \item subgroup calibration by horizon, capacity, action count, source reliability, and event type; and
  \item an ambiguity rate for groups where intervals do not separate.
\end{itemize}

\tfPmaDefinitionStatus
\begin{tfPmaDefinition}[Write-worthy, null-worthy, and ambiguous]
For materiality margin $\mu$ and true best-write target risk $R_{P_g}^W=\min_{a\in\tfPmaCalW_g}R_{P_g}(a)$, a group is write-worthy if $R_{P_g}^W+\mu<R_{P_g}(0)$, null-worthy if $R_{P_g}(0)+\mu<R_{P_g}^W$, and ambiguous otherwise, using $\min\varnothing=+\infty$.
\end{tfPmaDefinition}

Ambiguous groups are not forced into a favorable binary label. A false write is a non-NULL decision on a null-worthy group; a false \tfPmaNull{} is \tfPmaNull{} on a write-worthy group; a wrong-target write has regret above a fixed target tolerance even when some write is beneficial.

\subsection{Uncertainty sources}

\begin{table}[H]
\centering
\footnotesize
\begin{tabularx}{\linewidth}{@{}p{0.22\linewidth}Y Y@{}}
\toprule
Source & Required treatment & Failure boundary \\
\midrule
Finite future branches & Concentration or group bootstrap with action multiplicity & Small $K_{\mathrm{eff}}$ makes intervals wide \\
Finite prefix groups & Group-level resampling; no branch-level pseudoreplication & Shared prefixes are not independent units \\
Value-model uncertainty & Calibrated ensemble, quantile, or conformal-style method fixed before final use & Neural scores are not confidence bounds by naming \\
Distribution shift & Shift audit or declared robust set & Nominal coverage need not transfer \\
Evaluator uncertainty & Error/sensitivity analysis in the loss scale & Miscalibration can certify the wrong action \\
Constructor randomness & Freeze seeds or integrate over recorded draws & Selection can exploit lucky candidates \\
\bottomrule
\end{tabularx}
\caption{Uncertainty sources. Aleatoric branch variation, estimation error, model error, and evaluator error are not interchangeable.}
\label{tf:pma:tab:uncertainty-sources}
\end{table}

The registered resampling hierarchy mirrors the design: prefix families are the outer independent units; prefix groups remain intact within a family; all action slices for a branch remain paired; and constructor draws, evaluator draws, and branch draws are resampled only at their recorded nested levels. A branch-level bootstrap that treats $K_g|\tfPmaCalA_g|H$ tensor cells as independent is forbidden. Analytical and resampling components may be composed only by the single frozen rule attached to $\mathcal J$.

\tfPmaPrescriptionStatus\quad Any monitoring failure, coverage shortfall, unsupported action, or out-of-scope group switches the deployed selector to \tfPmaNull{}-only. This is a design rule, not evidence that a future implementation detects every shift.

%% file: theory_family/modules/pma/sections/08_identification.tex
\section{Identification, Distribution Shift, and Impossibility}
\label{tf:pma:sec:identification}

\subsection{What full grouped replay identifies}

Let $Y_g(a,F)$ be the complete potential replay outcome for action $a$ and continuation $F$. Under Assumptions~\ref{tf:pma:assump:exchangeability}--\ref{tf:pma:assump:future-causal}, full action replay from one common snapshot identifies $R_{Q_g}(a)=\tfPmaE_{Q_g}L(Y_g(a,F))$ for every feasible action, up to sampling error. If $Q_g=P_g$, this is the deployment conditional risk. If the future is action dependent, identification instead requires valid samples from each interventional law $P_g(F\mid\tfPmaDo(A=a),\widetilde S_g^+)$.

The conclusion is narrow. Grouped futures do not identify a different deployment branch law, an action absent from support, an evaluator's latent truth, or the value of features that would be unavailable before the action.

\subsection{Missing support}

\tfPmaProvedStatus
\begin{tfPmaTheorem}[Unsupported-action non-identification]
\label{tf:pma:thm:support-impossibility}
Suppose action $a^\dagger$ is never replayed and has zero logging probability on a set of prefix states with positive target probability. Without an additional outcome model restriction, $R(a^\dagger)$ is not identified from the observed data.
\end{tfPmaTheorem}

\begin{proof}
Construct two data-generating laws that agree on prefix states, logging actions, and every observed action outcome. Under the first law set the unobserved loss of $a^\dagger$ to $0$ on the unsupported set; under the second set it to $L_{\max}$. Both induce exactly the same observed distribution because $a^\dagger$ is never taken there, yet their target risks differ by $L_{\max}$ times the set's positive probability. Therefore the risk is not identified.
\end{proof}

This result is why a learned shortlist cannot silently replace full-action replay in the first oracle test. Shortlist recall must itself be qualified before omitted actions can be ignored.

\subsection{Branch-distribution shift}

\tfPmaProvedStatus
\begin{tfPmaTheorem}[Total-variation risk and gap sensitivity]
\label{tf:pma:thm:tv-shift}
Let $0\le L(a,F)\le L_{\max}$ and define $d=\tfPmaTV(P,Q)$. Then for every action
\begin{equation}
|R_P(a)-R_Q(a)|\le L_{\max}d.
\label{tf:pma:eq:tv-risk}
\end{equation}
For any two actions $a,b$,
\begin{equation}
\left|\{R_P(a)-R_P(b)\}-\{R_Q(a)-R_Q(b)\}\right|
\le2L_{\max}d.
\label{tf:pma:eq:tv-gap}
\end{equation}
\end{tfPmaTheorem}

\begin{proof}
For a measurable function in $[0,L_{\max}]$, the difference of expectations under $P$ and $Q$ is at most $L_{\max}\tfPmaTV(P,Q)$. Apply this once for Eq.~\ref{tf:pma:eq:tv-risk} and twice with the triangle inequality for Eq.~\ref{tf:pma:eq:tv-gap}.
\end{proof}

A synthetic oracle gap smaller than $2L_{\max}d$ can reverse under a deployment law within that distance. More synthetic branches cannot remove this bias. Distributionally robust objectives can optimize over a declared ambiguity set \cite{tf:pma:duchi2021}; they do not guarantee coverage of an undeclared shift.

\subsection{Evaluator misspecification}

Let $L^*$ be the target loss and $\widetilde L$ the evaluator loss.

\tfPmaProvedStatus
\begin{tfPmaProposition}[Evaluator-error bound]
\label{tf:pma:prop:evaluator-error}
If $|\widetilde L(a,F)-L^*(a,F)|\le\epsilon_{\mathrm{eval}}$ almost surely for every action, then
\begin{equation}
|\widetilde R(a)-R^*(a)|\le\epsilon_{\mathrm{eval}},
\end{equation}
and every pairwise action gap can differ by at most $2\epsilon_{\mathrm{eval}}$.
\end{tfPmaProposition}

\begin{proof}
Take expectations of the pointwise bound; apply the triangle inequality to two actions.
\end{proof}

Without any link between evaluator and target loss, even the sign of an action advantage is not identified: two latent target-loss laws can agree on every evaluator output and reverse which action is better.

\subsection{Leakage and confounding}

\begin{tfPmaCounterexample}[Post-treatment feature]
The write changes a future hidden state. A feature extractor then reads that hidden state and supplies it to the value model as if it were a pre-action feature. The feature perfectly predicts which action produced the favorable replay, but it is unavailable before selection. Offline accuracy can be one while the deployed input is undefined. This violates $\tfPmaCalD^+_{n,j}$-measurability.
\end{tfPmaCounterexample}

\begin{tfPmaCounterexample}[Hidden confounding in logged actions]
A production heuristic writes only when an unrecorded verifier is confident. Written cases later have lower loss than \tfPmaNull{} cases. The association can reflect verifier confidence rather than the write. Without randomization, observed confounder adjustment, or valid instrumental structure, the action effect is not identified.
\end{tfPmaCounterexample}

\begin{tfPmaCounterexample}[Near-zero action gap]
Two actions have risks differing by $\Delta\le2\varepsilon_g(\delta)$. Either action can be the empirical minimizer on the simultaneous concentration event. Treating the selected index as a reliable class label creates false precision; the group should be ambiguous or receive a soft value target with uncertainty.
\end{tfPmaCounterexample}

\begin{figure}[H]
\centering
\includegraphics[width=\linewidth]{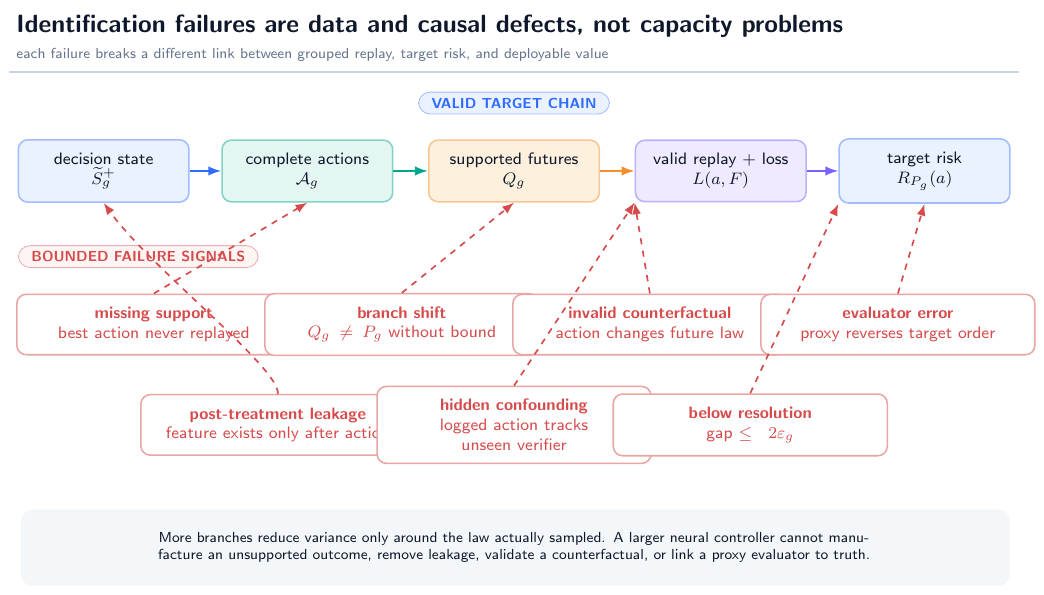}
\caption{\textbf{Identification and failure counterexamples.} Coral marks bounded failure signals: missing action support, branch shift, post-treatment leakage, hidden confounding, evaluator error, and action gaps below sampling resolution. More controller capacity does not repair these defects.}
\label{tf:pma:fig:identification-failures}
\end{figure}

\subsection{Identification ledger}

\begin{table}[H]
\centering
\footnotesize
\begin{tabularx}{\linewidth}{@{}p{0.22\linewidth}Y Y@{}}
\toprule
Target & Identified when & Not identified when \\
\midrule
$R_{Q_g}(a)$ & Full valid replay; exchangeable branches; fixed loss & Action absent, replay invalid, dependence ignored \\
$R_{P_g}(a)$ & $Q_g=P_g$ or valid supported correction & Synthetic/model law differs without a bound \\
Action effect & Exogenous common future or interventional simulator & Action changes future and one observed branch is reused \\
Future-blind predictability & Fresh prefix groups; causal features only & Future-derived or post-treatment feature enters model \\
Truth-linked risk & Evaluator link audited & Proxy can reverse target ordering \\
Long-run value & Repeated-write state distribution evaluated & One-step groups only \\
\bottomrule
\end{tabularx}
\caption{Identification summary. Each target has a separate premise bundle.}
\label{tf:pma:tab:identification-ledger}
\end{table}

%% file: theory_family/modules/pma/sections/09_value_model.tex
\section{Future-Blind Value Estimation and Selection}
\label{tf:pma:sec:value-model}

\subsection{A value model is not an oracle and not a policy}

The expected-risk oracle in Eq.~\ref{tf:pma:eq:empirical-oracle} sees training-only future branches. A deployable value model may see only the causal information available before action selection. Let
\begin{equation}
X_{g,a}=\phi(S_g^+,a,c_{g,a})
\end{equation}
be a typed, versioned feature record with $c_{g,0}=\varnothing$. Every coordinate of $X_{g,a}$ must be measurable with respect to the post-validation, preselection field $\tfPmaCalD_g^+$ from Eq.~\ref{tf:pma:eq:deploy-sigma-plus}; no coordinate may depend on a selected action, commit, or future branch. Define the target-law conditional value
\begin{equation}
q_P^*(x)=\tfPmaE_{F\sim P_g}\!\left[L_g(a,F)\mid X_{g,a}=x\right].
\label{tf:pma:eq:bayes-value}
\end{equation}
A model $Q_\psi(X_{g,a})$ may approximate Eq.~\ref{tf:pma:eq:bayes-value} only from a teacher target whose proposal-to-target bridge is valid under Protocol~\ref{tf:pma:prot:oracle-gate}. It does not receive $F$, a replay outcome, an oracle label computed from the same group, or any post-action state.

Three objects must remain separate:
\begin{enumerate}
  \item the \emph{value estimator} $Q_\psi$, which predicts action risk;
  \item the \emph{feasibility mask} $\tfPmaValid(S_g^+,a,c_{g,a})$, which enforces the row contract and hard constraints; and
  \item the \emph{selector}, which compares calibrated action intervals and may return \tfPmaNull{}.
\end{enumerate}
Calling the first object a policy erases two independently testable interfaces. This manuscript specifies all three but instantiates none.

\begin{figure}[H]
\centering
\includegraphics[width=\linewidth]{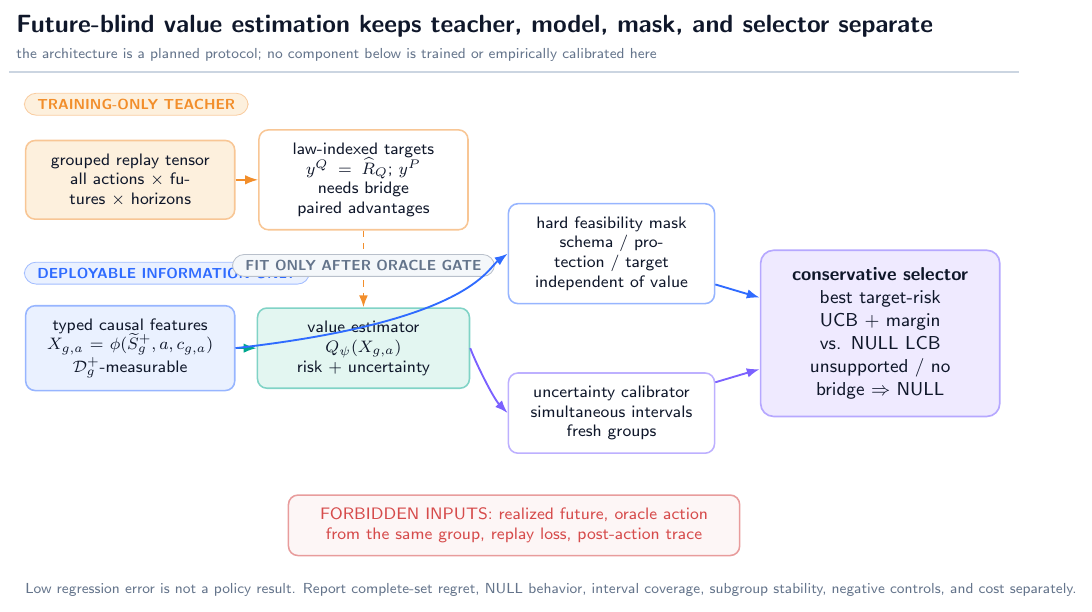}
\caption{\textbf{Future-blind value architecture (\tfPmaPlannedStatus{}).} Training-only grouped futures create action-risk targets. Deployment exposes only typed prefix state, action identity, and candidate metadata. The learned value, feasibility mask, uncertainty calibrator, and conservative selector remain separately auditable. No value model or policy is trained here.}
\label{tf:pma:fig:future-blind-value}
\end{figure}

\subsection{Training targets and losses}

The primary target is the continuous paired risk vector or its frozen scalarization, not merely the empirical argmin. The raw replay teacher is $y^Q_{g,a}=\widehat R_{Q_g}(a)$. It is a target-risk label only when the registered bridge establishes $Q_g=P_g$; under supported importance weighting, write the resulting target-risk estimate as $y^P_{g,a}=\widehat R^{\mathrm{bridge}}_{P_g}(a)$. A total-variation bound alone yields a target-risk interval, not a point label; such a group is used by an interval-valued or robust loss and is not silently converted into $y^P_{g,a}$. Let $y_{g,a}$ denote the registered bridge-valid point teacher when it exists, and let $d_{g,a}=y_{g,a}-y_{g,0}$. A preregistered value objective may combine
\begin{align}
\tfPmaCalL_{\mathrm{val}}(\psi)={}&
\sum_{g,a}\omega_{g,a}\,\rho_\kappa\!\left(Q_\psi(X_{g,a})-y_{g,a}\right)
\nonumber\\
&+\lambda_{\mathrm{rank}}
\sum_{g}\sum_{a<b}\tfPmaOne\{s_{g,a,b}\ne0\}
\log\!\left(1+\exp\{-s_{g,a,b}[Q_\psi(X_{g,b})-Q_\psi(X_{g,a})]\}\right),
\label{tf:pma:eq:value-objective}
\end{align}
where $\rho_\kappa$ is the Huber loss and $s_{g,a,b}=\operatorname{sign}(y_{g,b}-y_{g,a})\in\{-1,0,1\}$ after applying the frozen ambiguity margin. Thus positive $s$ means that $b$ is riskier than $a$, while ambiguous pairs have $s=0$ and receive exactly zero ranking charge. Group weights are fixed without using final outcomes. A heteroscedastic, quantile, ensemble, or conformal-style head may estimate uncertainty, but its coverage is an empirical property, not a consequence of architecture naming.

Regression error, action ranking, interval calibration, and selector regret are reported separately. Low mean squared error can coexist with wrong top actions near a decision boundary. Conversely, correct ranking can coexist with badly calibrated risk magnitudes and unsafe commit thresholds.

\subsection{Value error and decision regret}

For a fixed prefix group, let $a_g^*=\min_{\prec_g}\arg\min_{a\in\tfPmaCalA_g}R_{P_g}(a)$ and let
\begin{equation}
\widehat a_g=\min_{\prec_g}\arg\min_{a\in\tfPmaCalA_g}\widehat Q_g(a),
\qquad
\widehat Q_g(a)=Q_\psi(X_{g,a}),
\label{tf:pma:eq:plug-in-selector}
\end{equation}
with a fixed tie rule.

\tfPmaProvedStatus
\begin{tfPmaTheorem}[Uniform value error implies bounded action regret]
\label{tf:pma:thm:value-regret}
If $|\widehat Q_g(a)-R_{P_g}(a)|\le\epsilon_g$ for every $a\in\tfPmaCalA_g$, then
\begin{equation}
0\le R_{P_g}(\widehat a_g)-R_{P_g}(a_g^*)\le2\epsilon_g.
\label{tf:pma:eq:two-epsilon-regret}
\end{equation}
If the true best action is unique and separated from every competitor by more than $2\epsilon_g$, then $\widehat a_g=a_g^*$.
\end{tfPmaTheorem}

\begin{proof}
Optimality of $\widehat a_g$ for the estimated values gives
$\widehat Q_g(\widehat a_g)\le\widehat Q_g(a_g^*)$. Hence
\[
R_{P_g}(\widehat a_g)
\le\widehat Q_g(\widehat a_g)+\epsilon_g
\le\widehat Q_g(a_g^*)+\epsilon_g
\le R_{P_g}(a_g^*)+2\epsilon_g.
\]
The separation statement follows because choosing a competitor would contradict this bound.
\end{proof}

The premise is simultaneous and prefix-conditional. An average validation error does not establish it. Under branch-law shift and evaluator error, a useful sensitivity statement combines Theorem~\ref{tf:pma:thm:tv-shift} and Proposition~\ref{tf:pma:prop:evaluator-error}: a model error bound $\epsilon_g$, distribution distance $d_g$, and evaluator error $\epsilon_{\mathrm{eval}}$ can permit pairwise ordering error on the scale
\begin{equation}
2\epsilon_g+2L_{\max}d_g+2\epsilon_{\mathrm{eval}}.
\label{tf:pma:eq:combined-ordering-budget}
\end{equation}
This is a diagnostic budget, not a claim that any term is currently bounded.

\subsection{Training gate and abstaining deployment}

\tfPmaPlannedStatus\quad Value-model training may begin only after Protocol~\ref{tf:pma:prot:oracle-gate} qualifies a reproducible, material expected-risk advantage on fresh prefix groups. The data split is by physical prefix group, source namespace, and registered temporal boundary. Hyperparameters, feature schemas, action ordering, calibration method, and commit margin are frozen before final evaluation.

At deployment, unsupported schemas, invalid candidates, failed monitoring, or nonseparating intervals return \tfPmaNull{}. The selection rule is Eq.~\ref{tf:pma:eq:certified-rule}, not an unconstrained argmin of a neural score. A later learned shortlist would require an independent recall guarantee against the complete feasible action set before it could reduce replay or scoring work.

\tfPmaUnvalidatedStatus\quad There is no qualified oracle gap, trained $Q_\psi$, learned shortlist, calibrated interval model, or deployed learned selector in the frozen evidence record. The entire section is conditional theory and an unopened experimental protocol.

%% file: theory_family/modules/pma/sections/10_metrics_audits.tex
\section{Metrics, Splits, Leakage Audits, and Reproducibility}
\label{tf:pma:sec:metrics-audits}

\subsection{Decision metrics}

Let $a_g^*=\min_{\prec_g}\arg\min_{a\in\tfPmaCalA_g}R_{P_g}(a)$ minimize the registered target risk and let $\pi_g$ be the evaluated selector. The primary per-group action regret is
\begin{equation}
\tfPmaReg_g(\pi)=R_{P_g}(\pi_g)-R_{P_g}(a_g^*)\ge0.
\label{tf:pma:eq:action-regret}
\end{equation}
When a fixed scale $s_g>0$ is preregistered, normalized regret is $\tfPmaReg_g/s_g$; a scale chosen after seeing results is prohibited. Report the mean, median, upper quantiles, worst registered subgroup, and a group-level confidence interval. Also report:
\begin{itemize}
  \item top-1 action agreement and top-$k$ ranking metrics, with ambiguous groups separated;
  \item excess risk relative to current risk and relative to \tfPmaNull{};
  \item write precision, write recall, \tfPmaNull{} precision, and \tfPmaNull{} recall using the margin-aware classes in Section~\ref{tf:pma:sec:uncertainty};
  \item wrong-target rate conditional on writing;
  \item stale leakage, collateral damage, contamination, and resource coordinates before scalarization;
  \item interval coverage, width, risk-coverage curves, calibration error, and ambiguity rate; and
  \item stability across seeds, horizons, capacities, event types, source reliability, and action-set size.
\end{itemize}

Future loss is always computed by action-conditioned replay. Current-state reconstruction may be a diagnostic but cannot substitute for Eq.~\ref{tf:pma:eq:true-risk}. Accuracy on oracle action labels is secondary because labels are unstable when risk differences lie inside uncertainty intervals.

\subsection{Split hierarchy}

\tfPmaPrescriptionStatus\quad A physical causal prefix, all its candidate actions, and all its future branches form one indivisible group. The primary split hierarchy is:
\begin{enumerate}
  \item source namespace and temporal boundary;
  \item physical prefix group;
  \item registered scenario family or generator seed; and
  \item only then, individual action and branch records within the assigned group.
\end{enumerate}
No branch, paraphrase, candidate, cached trace, or derived oracle label from one physical prefix may cross train, calibration, and final partitions. The final partition is read once by the frozen evaluation entry point.

\subsection{Leakage and falsification audits}

\begin{table}[H]
\centering
\scriptsize
\begin{tabularx}{\linewidth}{@{}p{0.18\linewidth}p{0.24\linewidth}Y p{0.16\linewidth}@{}}
\toprule
Audit & Receipt & Passing condition & Failure action \\
\midrule
Prefix provenance & Source, byte hash, cutoff, event index & Feature timestamps precede action & Invalidate group \\
Group isolation & Physical-prefix identifier closure & No identifier or derivative crosses split & Rebuild split \\
Future blindness & Static graph plus runtime access log & No branch/outcome/post-action field reaches model & Reject model \\
Replay restoration & Snapshot, cache, index, RNG receipts & Every action-branch begins identically & Recompute tensor \\
Action completeness & Deterministic feasible-set manifest & Every valid action and \tfPmaNull{} priced & Reject oracle test \\
Evaluator freeze & Build hash and blinded final entry & No final-set tuning or hidden revision & Restart final evaluation \\
Label permutation & Group-preserving shuffled future labels & Performance collapses to registered null behavior & Investigate leakage \\
Prefix destruction & Mask registered causal signal only & Change agrees with preregistered sensitivity & Mark model suspect \\
Seed replication & Independent generator and constructor seeds & Sign and primary subgroup gates pass every seed & Stop progression \\
Cost reconciliation & Logical counts plus measured counters & Ledger totals match trace within tolerance & Reject efficiency claim \\
\bottomrule
\end{tabularx}
\caption{Minimum audit suite. Receipts are immutable artifacts, not narrative assurances.}
\label{tf:pma:tab:audit-suite}
\end{table}

The shuffled-label audit preserves group structure; branch-level shuffling that destroys obvious correlations can be too easy. A stronger negative control swaps future sets only among matched prefixes without changing action metadata. Success above the registered null range indicates leakage, unmodeled group identity, or a flawed split.

\subsection{Calibration protocol}

For interval level $1-\alpha$, empirical coverage on a final set $\tfPmaCalG_{\mathrm{test}}$ is
\begin{equation}
\widehat{\mathrm{Cov}}_{1-\alpha}
=\frac{1}{|\tfPmaCalG_{\mathrm{test}}|}
\sum_{g\in\tfPmaCalG_{\mathrm{test}}}
\tfPmaOne\!\left[R_{P_g}(a)\in[L_g^P(a),U_g^P(a)]\ \forall a\in\tfPmaCalA_g\right].
\label{tf:pma:eq:simultaneous-coverage-metric}
\end{equation}
This group-level simultaneous target-law coverage, not marginal action coverage, is the primary quantity. Proposal-law teacher intervals $[L_g^Q,U_g^Q]$, bridge inflation, and model intervals $[L_g^P,U_g^P]$ are reported separately; treating a noisy teacher estimate or an unbridged proposal interval as exact target risk understates uncertainty.

For a binary write-worthiness probability $p_g$ and label $z_g$, expected calibration error is reported only with bins fixed on calibration data, alongside proper scores and reliability diagrams. Because binning can conceal local failure, primary commit safety is judged by realized risk and simultaneous interval coverage.

\subsection{Reproducibility record}

Every run record binds code and environment identifiers, immutable data manifests, feature schema, group split, action manifest, constructor and evaluator versions, future-generation provenance, all RNG seeds, scalarization weights, interval method, cost counters, and output hashes. Failed and stopped runs remain indexed. Aggregates are regenerated from row-level records; no number is transcribed manually into a paper table.

\tfPmaPlannedStatus\quad These metrics and audits define future acceptance criteria. None has been run for a predictive admission value model in the evidence available to this manuscript.

%% file: theory_family/modules/pma/sections/11_costs.tex
\section{Complete Teacher and Deployment Cost Ledger}
\label{tf:pma:sec:costs}

\subsection{Logical work cannot be hidden by batching}

For group $g$, write $m_g=|\tfPmaCalA_g|$, with one \tfPmaNull{} action, $K_g$ future branches, and horizon $H_g$. A logical replay cell is one tuple $(g,a,k,h)$ whose action-conditioned state produces the registered horizon outputs and loss coordinates.

The opportunity index $g$ advances at every scored event, including events whose selected action is \tfPmaNull{}. A distinct committed-write index $u$ advances only after a non-NULL atomic commit. Replay, abstention, and deployment denominators use opportunities $g$; pollution and repair ledgers indexed by writes use $u$. In Eq.~\ref{tf:pma:eq:lifecycle-cost}, $t$ denotes every deployment event, so a \tfPmaNull{} event still incurs feature, scoring, calibration, monitoring, and resident-state costs.

\tfPmaProvedStatus
\begin{tfPmaProposition}[Exact logical replay count]
\label{tf:pma:prop:replay-count}
Under full action support and complete grouped replay, the number of horizon cells is
\begin{equation}
N_{\mathrm{cell}}=\sum_{g\in\tfPmaCalG}m_gK_gH_g.
\label{tf:pma:eq:logical-replay-count}
\end{equation}
The number of non-NULL complete candidate constructions is
\begin{equation}
N_{\mathrm{cand}}=\sum_{g\in\tfPmaCalG}(m_g-1).
\label{tf:pma:eq:candidate-count}
\end{equation}
If event-recursive teaching branches at every event, the corresponding tree count replaces $m_g$ by the number of supported prefix-action paths and can grow multiplicatively.
\end{tfPmaProposition}

\begin{proof}
For fixed $g$, the Cartesian product $\tfPmaCalA_g\times\{1,\ldots,K_g\}\times\{1,\ldots,H_g\}$ has $m_gK_gH_g$ elements. Summing disjoint groups proves Eq.~\ref{tf:pma:eq:logical-replay-count}. Exactly one action is \tfPmaNull{}, so $m_g-1$ complete non-NULL candidates are constructed. Event recursion evaluates supported paths rather than isolated labels; multiplying per-event branching factors gives the path count.
\end{proof}

Vectorization, shared kernels, prefix caching, or parallel hardware may reduce wall time but not the logical outcome array required by the estimand. Any reuse is reported together with the assumptions that make it exact.

\begin{figure}[H]
\centering
\includegraphics[width=\linewidth]{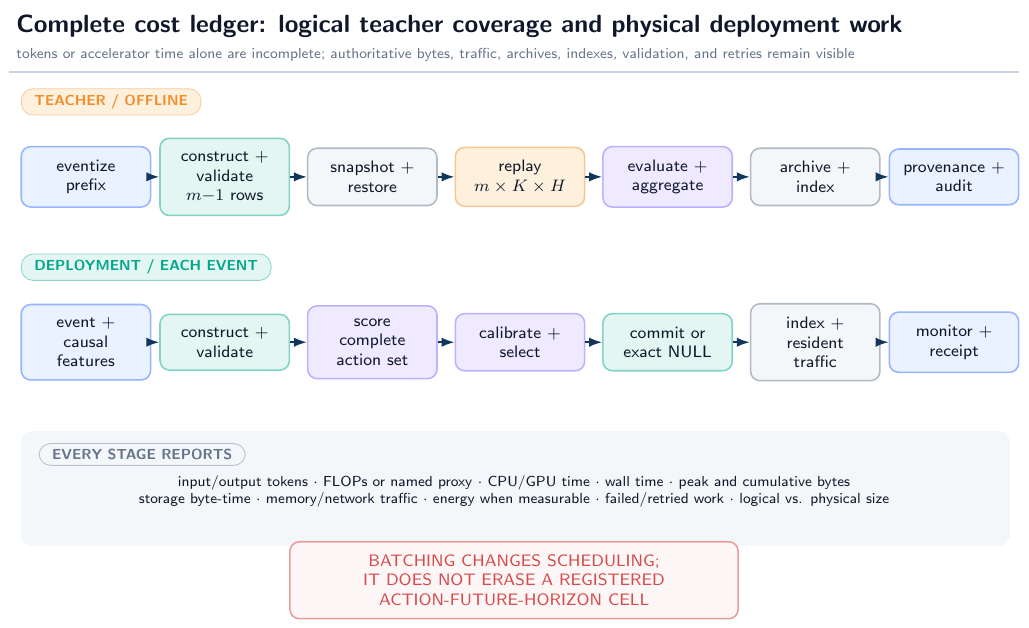}
\caption{\textbf{End-to-end cost ledger (\tfPmaDefinitionStatus{}).} The teacher pays eventization, every complete candidate, validation, $|\tfPmaCalA|\times K\times H$ replay, evaluation, archives, indexes, and provenance. Deployment pays causal feature extraction, candidate construction, scoring, calibration, commit, and resident-state traffic. Batching changes scheduling, not logical coverage.}
\label{tf:pma:fig:cost-ledger}
\end{figure}

\subsection{Teacher ledger}

For each group, report separately:
\begin{align}
C_g^{\mathrm{teach}}={}&
C_g^{\mathrm{event}}
+\sum_{a\neq0}\left(C_{g,a}^{\mathrm{construct}}+C_{g,a}^{\mathrm{validate}}\right)
+C_g^{\mathrm{snapshot}}
\nonumber\\
&+\sum_{a\in\tfPmaCalA_g}\sum_{k=1}^{K_g}
\left(C_{g,a,k}^{\mathrm{restore}}+C_{g,a,k}^{\mathrm{replay}}+C_{g,a,k}^{\mathrm{eval}}\right)
\nonumber\\
&+C_g^{\mathrm{aggregate}}+C_g^{\mathrm{archive}}+C_g^{\mathrm{index}}+C_g^{\mathrm{provenance}}.
\label{tf:pma:eq:teacher-cost}
\end{align}
The units ledger contains input/output tokens, model FLOPs or an explicitly named proxy, accelerator and CPU time, peak and cumulative bytes, storage byte-time, network and memory traffic, energy when measurable, wall time, and failed/retried work. Trusted assembly, parsing, validation, and evaluator calls are not free because they are outside the main model.

Teacher storage includes source snapshots, complete candidate bytes, future branches, action-conditioned traces or sufficient replay receipts, per-horizon loss vectors, resource counters, and provenance. Deduplication savings are reported with physical and logical sizes. Archive and index maintenance are charged to the design that needs them.

\subsection{Deployment ledger}

Deployment cost for one event is
\begin{align}
C_g^{\mathrm{deploy}}={}&C_g^{\mathrm{event}}+C_g^{\mathrm{feature}}
+\sum_{a\neq0}\left(C_{g,a}^{\mathrm{construct}}+C_{g,a}^{\mathrm{validate}}\right)
\nonumber\\
&+\sum_{a\in\tfPmaCalA_g}C_{g,a}^{\mathrm{score}}
+C_g^{\mathrm{calibrate}}+C_g^{\mathrm{select}}
+C_g^{\mathrm{commit}}+C_g^{\mathrm{index}}+C_g^{\mathrm{monitor}}.
\label{tf:pma:eq:deployment-cost}
\end{align}
If candidates are constructed only after a learned shortlist, the shortlist's complete-set recall and the cost of generating the features it needs must be counted. Otherwise apparent savings can be caused by silently changing the feasible action set.

\subsection{Fair comparisons}

Compare systems at equal resident authoritative byte-time, equal archive availability, equal retrieval context, equal evaluator quality, and equal task horizon. Report both teacher-amortized and steady-state deployment views. A method that moves rows from authoritative memory into an uncharged archive, increases retrieval tokens, or relies on expensive future generation has not demonstrated a system-level efficiency gain.

For repeated deployment over $T$ events, report
\begin{equation}
C_{1:T}^{\mathrm{total}}=C^{\mathrm{train}}+C^{\mathrm{calibration}}+
\sum_{t=1}^{T}C_t^{\mathrm{deploy}}+C_{1:T}^{\mathrm{resident}}+C_{1:T}^{\mathrm{archive}},
\label{tf:pma:eq:lifecycle-cost}
\end{equation}
together with the chosen amortization horizon. Savings that change sign outside one chosen $T$ are shown as a curve.

\tfPmaUnvalidatedStatus\quad No teacher run, deployment run, or predictive-admission efficiency measurement is reported here. Equations~\ref{tf:pma:eq:logical-replay-count}--\ref{tf:pma:eq:lifecycle-cost} are accounting requirements for later work.

%% file: theory_family/modules/pma/sections/12_counterexamples_stops.tex
\section{Counterexamples, Long-Run Pollution, and Stop Rules}
\label{tf:pma:sec:counterexamples-stops}

\subsection{Why an apparent oracle is insufficient}

\begin{tfPmaCounterexample}[Hindsight without conditional predictability]
There are two writes $a_1,a_2$ and two equiprobable futures $F_1,F_2$. Loss is zero when indices match and one otherwise. A per-future hindsight selector has zero loss, but both fixed actions have expected loss $1/2$, and no causal prefix feature predicts the future. The expected-risk oracle is tied and no learned admission rule can recover the hindsight gain. Thus an impressive hindsight ceiling is not an expected-risk oracle gap.
\end{tfPmaCounterexample}

\begin{tfPmaCounterexample}[Current-risk reversal]
Action $a_1$ perfectly reconstructs the completed segment and has current cost zero; action $a_2$ has current cost $0.1$. With probability $0.9$, later tasks need the fact stored by $a_2$ and never query $a_1$'s fact. A frozen horizon loss can prefer $a_2$ even though the current-risk comparator prefers $a_1$. Reversing the probabilities reverses the expected-risk order without changing current reconstruction. Current-risk quality therefore neither proves nor refutes future value.
\end{tfPmaCounterexample}

\begin{tfPmaCounterexample}[Short-horizon benefit, long-run pollution]
A write reduces next-step task loss by $0.2$ but overwrites a protected relation that is queried once in the next hundred steps, causing loss $1$. An $H=1$ evaluator selects the write; an $H=100$ evaluator rejects it. Horizon is part of the estimand and cannot be chosen after inspecting this reversal.
\end{tfPmaCounterexample}

\begin{tfPmaCounterexample}[Invalid synthetic branch]
A generator produces diverse continuations conditional on a textual prefix but omits an external process that makes a future depend on the chosen write. Replaying every action on those shared branches appears well supported. The counterfactual is nevertheless invalid because Assumption~\ref{tf:pma:assump:future-causal} fails; more generated branches concentrate around the wrong law.
\end{tfPmaCounterexample}

\begin{tfPmaCounterexample}[Hidden shortlist failure]
A cheap heuristic proposes two of ten feasible targets and the teacher prices only those plus \tfPmaNull{}. The omitted target is best on a rare but material subgroup. The learned controller matches its truncated teacher perfectly while incurring large complete-set regret. Unless shortlist recall was independently qualified, the result is not an oracle for $\tfPmaCalA_g$.
\end{tfPmaCounterexample}

\subsection{Accumulation over repeated writes}

Let $Z_t=1$ indicate that the $t$th committed write introduces a registered contamination event that remains consequential within the lifecycle horizon. Let $\tfPmaCalH_{t-1}$ be the pre-write history and suppose the monitoring argument establishes
\begin{equation}
\tfPmaP(Z_t=1\mid\tfPmaCalH_{t-1},A_t\neq0)\le p_t.
\label{tf:pma:eq:per-write-pollution}
\end{equation}

\tfPmaProvedStatus
\begin{tfPmaProposition}[Repeated-write pollution bound]
\label{tf:pma:prop:pollution-bound}
For any possibly adaptive sequence of at most $T$ committed writes satisfying Eq.~\ref{tf:pma:eq:per-write-pollution},
\begin{equation}
\tfPmaP\!\left(\bigcup_{t=1}^{T}\{Z_t=1\}\right)\le\sum_{t=1}^{T}p_t,
\qquad
\tfPmaE\!\left[\sum_{t=1}^{T}Z_t\right]\le\sum_{t=1}^{T}p_t.
\label{tf:pma:eq:pollution-union}
\end{equation}
No independence assumption is required.
\end{tfPmaProposition}

\begin{proof}
The union bound gives the first inequality. For the second, the tower property yields
$\tfPmaE Z_t=\tfPmaE[\tfPmaE(Z_t\mid\tfPmaCalH_{t-1},A_t\neq0)]\le p_t$ for committed writes; linearity of expectation completes the proof.
\end{proof}

For constant $p_t=p$, keeping the union-bound budget below $\alpha$ requires $p\le\alpha/T$. The bound can be loose, but it exposes a key scaling fact: a seemingly small per-write error is not automatically a small lifecycle risk. Dependencies, persistence duration, repair, and severity require richer stateful evaluation. This proposition does not establish any empirical $p_t$.

\subsection{Mandatory stop rules}

\begin{table}[H]
\centering
\footnotesize
\begin{tabularx}{\linewidth}{@{}p{0.23\linewidth}Y Y@{}}
\toprule
Observed condition & Scientific conclusion & Required action \\
\midrule
Inherited semantic row gate failed or unqualified & Admission action has no accepted semantic substrate & Do not run oracle/controller claim \\
No reproducible material gap over current risk & Predictive selection has not earned a target & Stop before value-model training \\
No registered relation to \tfPmaNull{} & Write benefit is not established & Keep \tfPmaNull{}-only deployment \\
Gap changes sign by seed, subgroup, or horizon & Oracle effect is unstable or underspecified & Diagnose; no pooled promotion \\
Oracle gap exists but causal features do not rank actions & Future value is not predictably available & Stop controller progression \\
Intervals fail simultaneous coverage & Commit certificate is invalid & Return \tfPmaNull{}; recalibrate on nonfinal data \\
Leakage, support, replay, or evaluator audit fails & Estimand is not identified by the run & Invalidate affected result \\
Long-run pollution exceeds budget & Short-horizon gain is insufficient & Extend horizon or reject controller \\
Full lifecycle cost erases registered benefit & System-level efficiency is unsupported & Report tradeoff; no efficiency claim \\
Only a learned shortlist is evaluated & Complete action regret is unknown & Restore full set or qualify recall first \\
\bottomrule
\end{tabularx}
\caption{Stop rules. A stop is an informative outcome, not a request to weaken the gate.}
\label{tf:pma:tab:stop-rules}
\end{table}

\tfPmaPlannedStatus\quad The next admissible experiment, only after an independent semantic-interface qualification, is a single-event, fixed-payload, complete-action grouped replay with exact \tfPmaNull{}, all candidates constructed before futures are opened, and every cost coordinate logged. If its expected-risk gap does not qualify, the experimental ladder ends. This manuscript neither executes nor authorizes that experiment.

%% file: theory_family/modules/pma/sections/13_obligations.tex
\section{Qualification Ladder and Theorem-to-Implementation Obligations}
\label{tf:pma:sec:obligations}

\begin{figure}[H]
\centering
\includegraphics[width=\linewidth]{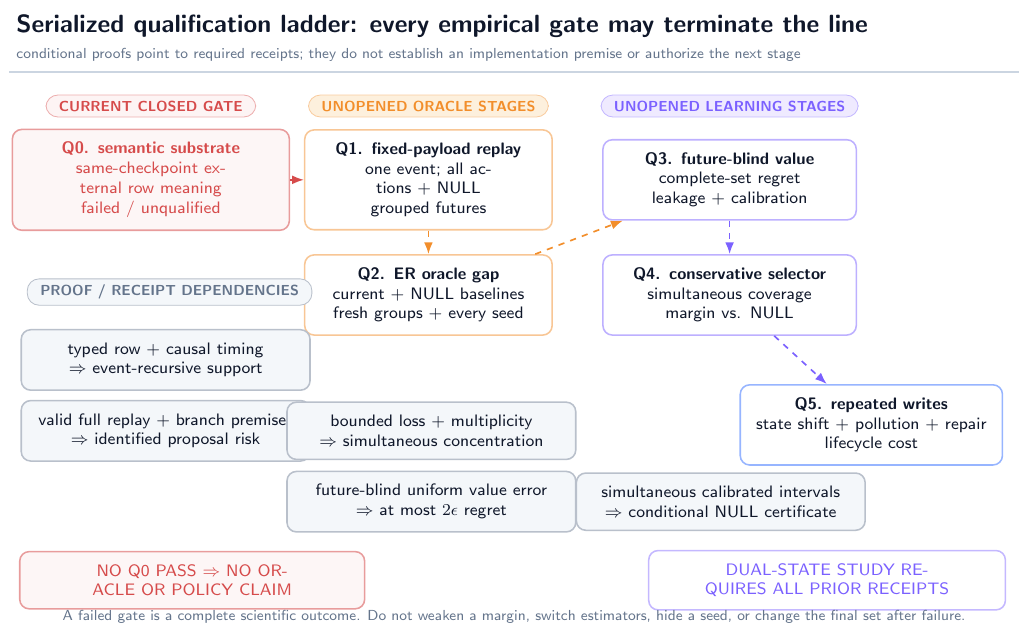}
\caption{\textbf{Qualification ladder and proof dependencies.} The semantic-interface stop precedes fixed-payload oracle replay. Only a stable expected-risk gap permits future-blind value training; only calibrated low-regret selection permits repeated-write study. Coral gates are currently closed or unvalidated.}
\label{tf:pma:fig:qualification-ladder}
\end{figure}

Conditional proofs constrain an implementation only when their premises are discharged by inspectable receipts. Figure~\ref{tf:pma:fig:qualification-ladder} and Table~\ref{tf:pma:tab:obligation-summary} make those dependencies explicit. A later stage cannot retroactively validate an earlier one.

\subsection{Proof dependency graph}

The principal logical dependencies are:
\begin{align*}
&\text{typed row contract + causal timing}
\Longrightarrow \text{event-recursive support},\\
&\text{common prefix + valid full replay + branch assumptions}
\Longrightarrow \text{identified empirical action risks},\\
&\text{bounded loss + conditionally iid weighted branches + multiplicity charge}
\Longrightarrow \text{simultaneous proposal-law concentration},\\
&\text{proposal-law intervals + valid target-law bridge}
\Longrightarrow \text{simultaneous target-law intervals},\\
&\text{simultaneous calibrated target-law intervals}
\Longrightarrow \text{certified improvement over \tfPmaNull{}},\\
&\text{future-blind features + uniform value error}
\Longrightarrow \text{bounded plug-in regret}.
\end{align*}
None of the right-hand statements establishes a left-hand premise. In particular, low empirical regret cannot prove causal timing or replay validity.

\begin{table}[H]
\centering
\scriptsize
\begin{tabularx}{\linewidth}{@{}p{0.20\linewidth}p{0.25\linewidth}Y p{0.14\linewidth}@{}}
\toprule
Result & Nonnegotiable premise & Minimum implementation receipt & Present status \\
\midrule
Event timing, Prop.~\ref{tf:pma:prop:event-timing} & Ordered eventizer; branch-local predecessor; valid complete row & Predecessor/successor and candidate digests per event & \tfPmaUnvalidatedStatus{} \\
Sequential support, Thm.~\ref{tf:pma:thm:sequential-support} & One-row atomic commit; exact identity \tfPmaNull{} & Slot diff and byte equality audit & Draft contract only \\
Weighted concentration, Thm.~\ref{tf:pma:thm:weighted-hoeffding} & Bounded fixed loss; conditionally iid proposal branches; fixed loss-independent weights & Sampling-regime flag, branch provenance, bounds, weights, action count & \tfPmaUnvalidatedStatus{} \\
Stable empirical action, Cor.~\ref{tf:pma:cor:stable-action} & Unique gap above simultaneous radius & Per-action estimates and interval separation & \tfPmaUnvalidatedStatus{} \\
Certified \tfPmaNull{}, Thm.~\ref{tf:pma:thm:certified-null} & Simultaneous target-law coverage; registered $Q_g\!\to P_g$ bridge; frozen margin & Bridge receipt, final group-level coverage, and commit log & \tfPmaUnvalidatedStatus{} \\
Support impossibility, Thm.~\ref{tf:pma:thm:support-impossibility} & Zero support on positive-mass set & Feasible-set and replay completeness manifest & Conditional theorem \\
Shift sensitivity, Thm.~\ref{tf:pma:thm:tv-shift} & Declared branch laws; bounded loss; valid distance bound & Generator/deployment provenance and shift estimate & No bound established \\
Value regret, Thm.~\ref{tf:pma:thm:value-regret} & Future-blind inputs; simultaneous uniform error & Runtime field-access audit and per-action error envelope & No model exists \\
Replay count, Prop.~\ref{tf:pma:prop:replay-count} & Full Cartesian replay & Trace-derived logical and physical counters & No run exists \\
Pollution bound, Prop.~\ref{tf:pma:prop:pollution-bound} & Valid per-write conditional error bound & Long-horizon event and repair ledger & No bound established \\
\bottomrule
\end{tabularx}
\caption{Theorem-to-implementation obligation summary. Appendix~\ref{tf:pma:app:obligation-matrix} expands the test and failure action for each premise.}
\label{tf:pma:tab:obligation-summary}
\end{table}

\subsection{Serialized qualification ladder}

\begin{enumerate}
  \item \textbf{Semantic substrate.} Independently qualify a same-checkpoint external semantic row interface under its own registered contract. The frozen evidence does not satisfy this step.
  \item \textbf{Mechanical fixed-payload oracle.} With one event, exact \tfPmaNull{}, complete feasible actions, and grouped futures, price every action without learning content or route.
  \item \textbf{Oracle qualification.} Apply Protocol~\ref{tf:pma:prot:oracle-gate} on fresh prefix groups and required seeds. No reproducible material gap means stop.
  \item \textbf{Future-blind value.} Only after Step 3, fit $Q_\psi$ and audit leakage, calibration, complete-set regret, \tfPmaNull{} behavior, and subgroup stability.
  \item \textbf{Conservative selector.} Freeze Eq.~\ref{tf:pma:eq:certified-rule}; verify simultaneous coverage and realized improvement without threshold tuning on final groups.
  \item \textbf{Repeated writes.} Evaluate state-distribution shift, pollution accumulation, repair, archive/index interactions, and lifecycle cost over registered horizons.
\end{enumerate}

Every step produces a versioned manifest and can terminate the line. The ladder does not imply that any step will pass.

\subsection{Current boundary}

\tfPmaUnvalidatedStatus\quad The evidence boundary remains before Step 1: the inherited semantic interfaces are failed/unqualified, the fitted-population diagnostic is a fit/support failure for the tested pure latent-row interface, and predictive overwrite remains closed. Steps 2--6 are theory and planned protocol only.

%% file: theory_family/modules/pma/sections/14_related_limits.tex
\section{Related Foundations, Limitations, and Conclusion}
\label{tf:pma:sec:related-limits}

\subsection{Related foundations}

The integrated MMLA architecture motivates bounded reasoning-time state, explicit receipts, and a failed/unvalidated semantic-interface boundary rather than supplying evidence for predictive admission. Parts~\ref{tf:rtt:part:foundations} and~\ref{tf:amr:part:atomic-memory-rows} provide the typed policy/memory separation and atomic external-row contract used here; they are formal dependencies, not empirical evidence for admission.

Our grouped-future construction uses the potential-outcome distinction between treatment assignment and outcomes \cite{tf:pma:rubin1974} and the role of propensity/support conditions in observational adjustment \cite{tf:pma:rosenbaum1983}. When logged propensities replace complete controlled replay, contextual-bandit evaluation requires explicit support and estimator assumptions \cite{tf:pma:dudik2011}; this does not justify reusing one observed continuation as every action's counterfactual.

Finite-action confidence control draws on classical bounded concentration \cite{tf:pma:hoeffding1963} and empirical-Bernstein refinements \cite{tf:pma:maurer2009}. \tfPmaNull{} is related to classification with a reject option \cite{tf:pma:chow1970}, but here rejection is also a physical identity transition whose future risk must be replayed. Distributionally robust optimization can expose sensitivity to a declared ambiguity set \cite{tf:pma:duchi2021}, while conditional value-at-risk supplies one possible tail-risk scalarization \cite{tf:pma:rockafellar2000}; neither method proves that the declared set or loss matches deployment.

Neural Turing Machines, Differentiable Neural Computers, and compressive memory study learned read/write or compressed recurrent memory mechanisms \cite{ntm2014,tf:amr:graves2016dnc,compressive2019}. Predictive admission here concerns a narrower authoritative decision: whether a complete target-conditioned row should replace bounded resident state under a causal future-risk protocol. It neither claims superiority to those systems nor treats their mechanism results as evidence for MMLA.

\subsection{Limitations and explicit nonclaims}

The theory assumes a finite feasible action set, bounded registered loss, typed complete rows, and a replay system capable of restoring action-conditioned states. Large or continuous action spaces require approximation and new support guarantees. Conditional exchangeability can fail; synthetic futures can be biased; evaluator loss can misorder true outcomes; the deployment law can drift; and a finite horizon can miss slow pollution or repair. The simultaneous bounds can be conservative when actions are numerous or groups are small. A complete teacher can be computationally prohibitive even when deployment is cheap.

No theorem establishes that semantic row content can be decoded at the same checkpoint, that future value is predictable from causal features, that calibrated neural intervals will be obtained, that a safe system exists, or that total-system cost improves. ``Certified'' in Theorem~\ref{tf:pma:thm:certified-null} is always relative to stated interval coverage, target law, loss, horizon, and margin. A violated premise voids the deduction.

This admission-theory part contributes no new data collection, trained admission model, oracle margin, learned shortlist, deployment, safety result, or efficiency result. The separate empirical chapters report readout training and component experiments, not an admission intervention. Neither those results nor the complete mechanical oracle in the fitted-population diagnostic establishes learned predictive admission.

\subsection{Conclusion}

Predictive memory admission should begin with an identified decision problem, not a controller. This paper has separated exact \tfPmaNull{}, complete target-conditioned actions, a current-risk comparator, a training-only expected-risk oracle, and a future-blind value estimator. It has made event timing, grouped futures, complete action-conditioned replay, simultaneous uncertainty, branch shift, evaluator error, lifecycle pollution, and the full cost multiplier visible. The resulting ladder is deliberately easy to stop: a failed semantic substrate, absent oracle gap, unpredictable value, uncalibrated selection, long-run pollution, or erased system-level benefit ends progression without relabeling the failure.

\tfPmaConjecturedStatus\quad Selective admission may eventually help a qualified bounded authoritative memory. That conjecture remains open. The current scientific result is a formal target, conditional deductions, falsification protocol, and explicit stop boundary.

%% file: theory_family/modules/pma/appendices/A_proofs.tex
\section{Supplementary Proofs and Boundary Lemmas}
\label{tf:pma:app:proofs}

This appendix expands deductions that separate attainable expected-risk decisions from information ceilings. Every result is conditional on its displayed premises and carries no empirical status.

\subsection{Hindsight and expected-risk decisions}

\tfPmaProvedStatus
\begin{tfPmaLemma}[Expectation of the minimum is an information ceiling]
\label{tf:pma:lem:hindsight-ceiling}
For any finite action set and integrable losses $L(a,F)$,
\begin{equation}
\tfPmaE\!\left[\min_{a\in\tfPmaCalA}L(a,F)\right]
\le
\min_{a\in\tfPmaCalA}\tfPmaE[L(a,F)].
\label{tf:pma:eq:hindsight-inequality}
\end{equation}
Equality holds if one fixed action minimizes $L(a,F)$ almost surely, but equality need not imply a unique common minimizer.
\end{tfPmaLemma}

\begin{proof}
For every fixed $b\in\tfPmaCalA$, $\min_aL(a,F)\le L(b,F)$ pointwise. Taking expectations and then minimizing the right-hand side over $b$ proves Eq.~\ref{tf:pma:eq:hindsight-inequality}. A fixed almost-sure minimizer gives equality. With ties on positive-probability sets, equality can also occur without uniqueness.
\end{proof}

The left side lets the action depend on the realized future and corresponds to Eq.~\ref{tf:pma:eq:hindsight-action}. The right side selects one action before the future and corresponds to expected-risk admission. Their difference is the value of prohibited future information, not necessarily a learnable opportunity.

\tfPmaProvedStatus
\begin{tfPmaProposition}[Current-state loss does not identify future-risk order]
\label{tf:pma:prop:current-nonidentification}
Suppose two actions have fixed observed current costs $C^{\mathrm{cur}}(a)$ and $C^{\mathrm{cur}}(b)$. Without a restriction linking those costs to continuation loss, the sign of $R(a)-R(b)$ is not identified.
\end{tfPmaProposition}

\begin{proof}
Construct two continuation laws with identical prefix observations and current costs. In the first, set bounded future losses to $L(a,F)=0$ and $L(b,F)=1$; in the second reverse them. The observed current record is identical, while the future-risk ordering reverses.
\end{proof}

\subsection{Paired estimation}

For independent branches with uniform weights, define $D_k(a,b)=L(a,F_k)-L(b,F_k)\in[-L_{\max},L_{\max}]$ and $\widehat\Delta(a,b)=K^{-1}\sum_kD_k(a,b)$.

\tfPmaProvedStatus
\begin{tfPmaLemma}[Paired finite-sample bound]
\label{tf:pma:lem:paired-bound}
For one preregistered pair $(a,b)$,
\begin{equation}
\tfPmaP\!\left(\left|\widehat\Delta(a,b)-\Delta(a,b)\right|>\epsilon\right)
\le2\exp\!\left(-\frac{K\epsilon^2}{2L_{\max}^2}\right).
\label{tf:pma:eq:paired-bound}
\end{equation}
For a finite registered set of $P$ pairs, replacing the right side by $2P\exp[-K\epsilon^2/(2L_{\max}^2)]$ gives simultaneous control.
\end{tfPmaLemma}

\begin{proof}
Each difference has range length $2L_{\max}$. Hoeffding's inequality for the sample mean gives Eq.~\ref{tf:pma:eq:paired-bound}; a union bound gives the simultaneous statement \cite{tf:pma:hoeffding1963}.
\end{proof}

Pairing does not always improve the worst-case range bound, but Eq.~\ref{tf:pma:eq:paired-variance} can substantially reduce variance-sensitive intervals. Branches remain the units within one prefix; prefix groups remain the units for population inference.

\subsection{Decomposed regret under three error sources}

Let $R_P^*$ be target-law risk under target evaluator, $R_Q^*$ proposal-law risk under target evaluator, and $\widetilde R_Q$ proposal-law risk under the implemented evaluator. Suppose
\begin{align}
\sup_a|R_P^*(a)-R_Q^*(a)|&\le\epsilon_{\mathrm{shift}},\\
\sup_a|R_Q^*(a)-\widetilde R_Q(a)|&\le\epsilon_{\mathrm{eval}},\\
\sup_a|\widehat Q(a)-\widetilde R_Q(a)|&\le\epsilon_{\mathrm{model}}.
\end{align}

\tfPmaProvedStatus
\begin{tfPmaTheorem}[Target regret under additive uniform errors]
\label{tf:pma:thm:additive-regret}
If $\widehat a\in\arg\min_a\widehat Q(a)$ and $a_P^*\in\arg\min_aR_P^*(a)$, then
\begin{equation}
R_P^*(\widehat a)-R_P^*(a_P^*)
\le2(\epsilon_{\mathrm{shift}}+\epsilon_{\mathrm{eval}}+\epsilon_{\mathrm{model}}).
\label{tf:pma:eq:additive-regret}
\end{equation}
With Theorem~\ref{tf:pma:thm:tv-shift}, one may set $\epsilon_{\mathrm{shift}}=L_{\max}\tfPmaTV(P,Q)$ when its premises hold.
\end{tfPmaTheorem}

\begin{proof}
The three uniform bounds and the triangle inequality give
$\sup_a|R_P^*(a)-\widehat Q(a)|\le\epsilon_{\mathrm{shift}}+\epsilon_{\mathrm{eval}}+\epsilon_{\mathrm{model}}$. Apply Theorem~\ref{tf:pma:thm:value-regret} with this sum.
\end{proof}

The theorem is a sensitivity decomposition. It is unusable as a certificate until all three terms have valid target-domain bounds.

\subsection{Adaptive histories and severe events}

Let $V_t\in[0,V_{\max}]$ be contamination severity at write $t$, with $V_t=0$ when no contamination occurs. If
$\tfPmaE[V_t\mid\tfPmaCalH_{t-1},A_t\neq0]\le v_t$, then by the tower property
\begin{equation}
\tfPmaE\!\left[\sum_{t=1}^{T}V_t\right]\le\sum_{t=1}^{T}v_t.
\label{tf:pma:eq:severity-sum}
\end{equation}
This permits adaptive action selection but assumes the conditional bound remains valid under the induced history. A bound calibrated only on a static one-write distribution cannot simply be reused after the selector changes the state distribution.

For a severe-event indicator $B_t=\tfPmaOne[V_t\ge v_{\mathrm{crit}}]$ with conditional rate at most $q_t$, Proposition~\ref{tf:pma:prop:pollution-bound} gives $\tfPmaP(\max_tV_t\ge v_{\mathrm{crit}})\le\sum_tq_t$. If repair removes consequences, the lifecycle loss must model duration and repair cost rather than erase the original event.

\subsection{Event-tree work}

Suppose segment event $j$ has $m_j(\mathbf a_{<j})$ feasible actions after branch prefix $\mathbf a_{<j}$. The number of leaf action paths is
\begin{equation}
N_{\mathrm{leaf}}=
\sum_{a_1\in\tfPmaCalA_1}
\sum_{a_2\in\tfPmaCalA_2(a_1)}\cdots
\sum_{a_J\in\tfPmaCalA_J(\mathbf a_{<J})}1.
\label{tf:pma:eq:event-leaf-count}
\end{equation}
If every node has exactly $m$ actions, $N_{\mathrm{leaf}}=m^J$. A frozen rollout changes the estimand and reduces branching after the scored event; it is not a free exact optimization. This is why the first causal oracle test fixes a single event.

%% file: theory_family/modules/pma/appendices/B_estimators.tex
\section{Estimator and Evaluation Protocol Specifications}
\label{tf:pma:app:estimators}

\subsection{Within-group proposal-risk record}

For each prefix group $g$, freeze the ordered action manifest
$\tfPmaCalA_g=(0,a_1,\ldots,a_{m_g-1})$, future manifest
$(F_{g,1},\ldots,F_{g,K_g})$, nonnegative weights $w_{g,k}$ summing
to one, horizon $H_g$, sampling-regime flag
$\rho_g\in\{\text{i.i.d.},\text{stratified},\text{finite population}\}$,
and scalarization before replay. Store the loss-tensor coordinates before
aggregating them into $L_{g,a,k}$.

The replay tensor directly identifies proposal-law quantities:
\begin{align}
\widehat R_{Q_g}(a)&=\sum_kw_{g,k}L_{g,a,k},\\
\widehat\Delta_{Q_g}(a,b)&=\sum_kw_{g,k}
  \bigl(L_{g,a,k}-L_{g,b,k}\bigr),\\
\widehat a_g^{\mathrm{ER},Q}
&=\min_{\prec_g}\arg\min_{a\in\tfPmaCalA_g}\widehat R_{Q_g}(a),
\end{align}
where $\prec_g$ is the frozen total action order. Effective branch count is
Eq.~\ref{tf:pma:eq:effective-k}; it is descriptive and accompanies raw $K_g$ rather
than replacing the sampling-regime record. Zero-weight branches remain in
provenance but do not inflate effective count.

For the initial gate, do not fit a nuisance model to fill missing actions:
absence of a required replay slice fails action completeness. The raw teacher
above is a $Q_g$-law teacher. It is not relabeled as a deployment-law oracle.

\subsection{Registered \texorpdfstring{$Q_g\rightarrow P_g$}{Q-to-P} bridge}
\label{tf:pma:sec:qp-estimator-bridge}

Each group carries exactly one bridge flag before outcomes are opened.
\begin{enumerate}
  \item \textbf{Law equality.} If the design establishes $Q_g=P_g$, set
  $\widehat R_{P_g}(a)=\widehat R_{Q_g}(a)$ and retain the design evidence.
  \item \textbf{Supported reweighting.} For i.i.d. draws
  $F_{g,k}\sim Q_g$, a frozen Radon--Nikodym ratio
  $r_g(F)=\mathrm dP_g/\mathrm dQ_g(F)$ gives
  \begin{equation}
  \widehat R_{P_g}^{\mathrm{IW}}(a)
  =\frac{1}{K_g}\sum_{k=1}^{K_g}r_g(F_{g,k})L_{g,a,k}.
  \label{tf:pma:eq:iw-target-risk}
  \end{equation}
  Absolute continuity, ratio provenance, overlap diagnostics, and the choice
  between unclipped and clipped ratios are frozen. Clipping changes the
  estimand unless its bias is separately bounded. Stratified or
  finite-population samples require their own registered design-weight
  estimator; Eq.~\eqref{tf:pma:eq:iw-target-risk} is not silently reused.
  \item \textbf{Sensitivity-only bridge.} If the only admitted relation is
  $\operatorname{TV}(P_g,Q_g)\le\tau_g$ and
  $0\le L\le L_{\max}$, then
  \begin{equation}
    R_{P_g}(a)\in
    \left[
      \max\{0,R_{Q_g}(a)-L_{\max}\tau_g\},
      \min\{L_{\max},R_{Q_g}(a)+L_{\max}\tau_g\}
    \right].
    \label{tf:pma:eq:tv-target-band}
  \end{equation}
  This is an identification band, not a target-risk point estimator.
\end{enumerate}
If no bridge is justified, the target-law oracle gate is unidentifiable and
stops. Proposal-law concentration cannot repair that failure.

\subsection{Population estimands without selection reuse}

Let fresh physical prefix groups $g=1,\ldots,G$ be independent draws from the
registered target group law. A bridge-valid procedure returns target-risk
estimates $\widehat R_{P_g}^{\mathrm{eval}}(a)$ on evaluation branches. The
teacher action $\widehat a_g^{\mathrm{ER},P,\mathrm{train}}$ is selected either
on disjoint training branches or under a simultaneous finite-class procedure
whose selection correction is frozen. Define paired group advantages
\begin{align}
\widehat G_g^{\mathrm{cur},P}
&=\widehat R_{P_g}^{\mathrm{eval}}(a_g^{\mathrm{cur}})
 -\widehat R_{P_g}^{\mathrm{eval}}
  (\widehat a_g^{\mathrm{ER},P,\mathrm{train}}),\\
\widehat G_g^{0,P}
&=\widehat R_{P_g}^{\mathrm{eval}}(0)
 -\widehat R_{P_g}^{\mathrm{eval}}
  (\widehat a_g^{\mathrm{ER},P,\mathrm{train}}).
\end{align}
Using the same outcomes both to choose an empirical minimum and to report its
uncorrected advantage is prohibited. Acceptable registered treatments include
a nested branch split with adequate power, a simultaneous finite-class bound,
or another bias-corrected procedure justified before outcome access.

The population aggregate is a fixed weighted mean
\begin{equation}
\widehat G^{b,P}=\sum_{g=1}^{G}\nu_g\widehat G_g^{b,P},
\qquad b\in\{\mathrm{cur},0\},
\qquad \nu_g\ge0,\quad\sum_g\nu_g=1.
\label{tf:pma:eq:population-gap-estimator}
\end{equation}
Primary inference resamples or concentrates at the physical-group level.
Treating $G K_g m_g$ replay rows as independent is pseudoreplication.

\subsection{Law- and design-indexed intervals}

For i.i.d. proposal draws, equal weights, and $0\le L\le L_{\max}$,
Eq.~\ref{tf:pma:eq:weighted-hoeffding} yields the transparent proposal interval
\begin{equation}
[L_g^Q(a),U_g^Q(a)]
=\left[
\max\{0,\widehat R_{Q_g}(a)-\varepsilon_g^Q\},
\min\{L_{\max},\widehat R_{Q_g}(a)+\varepsilon_g^Q\}
\right].
\label{tf:pma:eq:hoeffding-interval}
\end{equation}
The interval is indexed by $Q$ because that is the law sampled. A target-law
interval $[L_g^P(a),U_g^P(a)]$ is reported only after propagating a registered
bridge and its uncertainty. Under the TV bridge, for example, the proposal
confidence interval is widened by $L_{\max}\tau_g$ and clipped to
$[0,L_{\max}]$. Stratified and finite-population regimes use their own frozen
variance and finite-population corrections. Neither weighted Hoeffding nor
$K_g^{\mathrm{eff}}$ is presented as a universal substitute.

If actions share branches, preregistered paired empirical-Bernstein intervals
can be more efficient \cite{tf:pma:maurer2009}; action and pair multiplicity still must
be charged. The first result uses the registered primary construction, with
alternatives labeled as sensitivity analyses.

\subsection{Group bootstrap specification}

A group bootstrap draw samples physical prefix groups with replacement and
carries each selected group's full action-by-future tensor. Within-group
branches are not independently resampled in the primary population bootstrap.
If future-generation uncertainty is also targeted, use a nested bootstrap and
report its separate estimand. Stratification variables, group weights,
bridge flags, and law labels are frozen before final data are opened.

For $B$ registered bootstrap replicates, the interval construction, quantile
convention, studentization, and seed list are fixed. A replicate that lacks a
required subgroup is invalid rather than silently dropped. The analysis logs
all invalid-replicate counts and applies the preregistered failure rule.

\subsection{Oracle-gap run sequence}

\begin{enumerate}
  \item Verify the semantic-interface prerequisite, lock the snapshot
  namespace, and hash every source artifact.
  \item Eventize one completed segment into exactly one scored event; freeze
  fixed payload and complete target actions plus \tfPmaNull{}.
  \item Construct and validate every complete candidate before future outcomes
  are exposed to the selector or analyst.
  \item Register $\rho_g$, generate or collect grouped futures with provenance,
  and freeze weights, proposal law $Q_g$, target law $P_g$, and the
  $Q_g\rightarrow P_g$ bridge.
  \item Restore the same pre-action snapshot for every action--branch pair and
  replay the full registered horizon.
  \item Compute all loss coordinates, resource counters, proposal risks,
  bridge-valid target risks or identification bands, paired gaps, and
  simultaneous intervals automatically.
  \item Run action-completeness, restoration, future-blindness, group-isolation,
  evaluator, negative-control, bridge, and cost audits.
  \item Evaluate fresh prefix groups and all registered seeds once; apply
  Protocol~\ref{tf:pma:prot:oracle-gate} without threshold revision.
  \item If any target-law gate fails or is unidentified, record the stop. Only
  a passed target-law oracle gate permits construction of a separate
  value-model dataset.
\end{enumerate}

\subsection{Value-model evaluation sequence}

If and only if the target-law oracle gate passes, fit on training prefix groups,
choose all architecture and calibration decisions on calibration groups,
freeze the artifact, and evaluate once on final groups. Teacher targets retain
their law labels: raw replay targets are $y^Q_{g,a}$; a target-law label
$y^P_{g,a}$ exists only through the registered bridge. The final table includes
the complete action set, per-group target-law regret or identification band,
raw risk coordinates, \tfPmaNull{} confusion metrics, simultaneous coverage,
interval width, ambiguity rate, subgroup stability, negative controls, and
full cost. A second final-set pass after model or threshold revision creates a
new registered revision and a new untouched final namespace.

\tfPmaUnvalidatedStatus\quad This appendix is a specification only. It contains no
run identifier, data row, interval, effect estimate, or outcome.

%% file: theory_family/modules/pma/appendices/C_obligation_matrix.tex
\section{Complete Theorem-to-Implementation Matrix}
\label{tf:pma:app:obligation-matrix}

\scriptsize
\setlength\LTleft{0pt}
\setlength\LTright{0pt}
\begin{minipage}{\linewidth}
\captionsetup{hypcap=false}
\captionof{table}{Implementation obligations. ``Stop'' means the corresponding conclusion cannot be claimed; it does not prescribe a scientific workaround.}\label{tf:pma:tab:complete-obligations}
\end{minipage}
\begingroup
\renewcommand*{\theHtable}{obligation-body}
\begin{longtable}{@{}p{0.13\textwidth}p{0.18\textwidth}p{0.23\textwidth}p{0.21\textwidth}p{0.13\textwidth}@{}}
\toprule
Object/result & Premise & Receipt/test & Failure consequence & Evidence status \\
\midrule
\endfirsthead
\toprule
Object/result & Premise & Receipt/test & Failure consequence & Evidence status \\
\midrule
\endhead
\bottomrule
\endfoot
Completed segment & Immutable decision boundary and ordered bytes & Source digest, cutoff time, tokenizer/build version & Event inputs are ambiguous; stop & Contract only \\
Eventizer & Deterministic ordered event list under frozen version & Event list digest; rerun equality; human audit sample & Timing theorem inapplicable & Unvalidated \\
Complete row & Trusted assembly; explicit target/version/protection; parse validity & Byte digest, schema validation, canonical round-trip & Remove action; no semantic claim & Draft dependency \\
Exact \tfPmaNull{} & Identity on resident state, indexes, and authority & Byte-for-byte state and index equality; zero commit trace & Baseline invalid; stop oracle test & Draft dependency \\
Feasible set & Every valid target represented before outcome access & Deterministic action manifest and rejection reasons & Unsupported-action theorem applies & Unvalidated \\
Event recursion & Candidate predecessor equals branch prefix & Per-event predecessor/candidate/successor digest chain & Reject branch & Unvalidated \\
Grouped future & One physical prefix; registered target/proposal law & Group ID closure, timestamps, generator/config hashes & No target-risk interpretation & Unvalidated \\
Exchangeability & Conditional independent or valid weighted branch draws & Sampling design and dependence diagnostics & Concentration interval invalid & Assumption only \\
Future causality & Common future exogenous to action, or interventional branch law & Simulator causal contract or domain argument & Counterfactual replay invalid & Assumption only \\
Snapshot restore & Identical state, cache, index, and RNG boundary & Restoration digest before every action-branch run & Invalidate replay tensor & Unvalidated \\
Action replay & Memory action can affect hidden evolution and all effects are rerun & Full trace lineage; no pre-action hidden-trace reuse & Horizon loss not action conditioned & Unvalidated \\
Loss bound & Every scalar loss lies in registered $[0,L_{\max}]$ & Automated range audit and overflow log & Hoeffding claims void & Unvalidated \\
Loss meaning & Evaluator linked to target task/truth & Frozen evaluator, blinded validation, sensitivity envelope & Only proxy-risk claim remains & Unvalidated \\
Current comparator & No realized future; definition and ties frozen & Comparator artifact hash and runtime field log & Gap comparison invalid & Unvalidated \\
Empirical ER oracle & Complete actions, valid risks, multiplicity correction & Action tensor and simultaneous intervals & No oracle-gap claim & Unrun \\
Population gap & Fresh independent prefix groups; registered weights/subgroups & Final namespace manifest and group-level inference & No population promotion & Unrun \\
Future-blind features & Every field pre-action measurable & Static dataflow graph plus runtime access log & Reject model as leaked & No model \\
Value error & Simultaneous complete-action error bound on target groups & Per-action prediction/risk envelope & $2\epsilon$ theorem unusable & No model \\
Calibrated selector & Simultaneous coverage and frozen margin & Final coverage, interval, action, and commit receipts & \tfPmaNull{}-only fallback & No model \\
Branch shift & Target/proposal distance bounded in relevant loss class & Declared law and sensitivity bound & Synthetic gap cannot transfer & No bound \\
Repeated writes & One-write guarantees stable under induced histories & Stateful multi-event replay and pollution/repair ledger & No lifecycle claim & Unrun \\
Complete cost & Every constructor, replay, archive, index, and deployment term counted & Logical counts, physical counters, reconciliation audit & No efficiency claim & Unrun \\
\end{longtable}
\endgroup
\addtocounter{table}{-1}% The captionless longtable advances the table counter; the caption above already owns this number.

\normalsize
The matrix is deliberately redundant with the prose. A theorem citation without the corresponding receipt is an incomplete claim. Receipts must identify the exact artifact revision; evidence from another checkpoint, action set, horizon, evaluator, or branch law cannot be silently reused.

%% file: theory_family/modules/pma/appendices/D_symbols.tex
\section{Symbol, Status, and Boundary Ledger}
\label{tf:pma:app:symbols}

\subsection{Core symbols}

\small
\begin{minipage}{\linewidth}
\captionsetup{hypcap=false}
\captionof{table}{Primary notation. All group-indexed objects inherit the registered prefix group $g$.}\label{tf:pma:tab:symbols}
\end{minipage}
\begingroup
\renewcommand*{\theHtable}{symbols-body}
\begin{longtable}{@{}p{0.18\textwidth}p{0.30\textwidth}p{0.44\textwidth}@{}}
\toprule
Symbol & Type & Meaning \\
\midrule
\endfirsthead
\toprule
Symbol & Type & Meaning \\
\midrule
\endhead
\bottomrule
\endfoot
$M_{n-1}$ & bounded resident state & Authoritative state before completed segment $n$ \\
$x_n$ & completed segment & Immutable observation block at a decision boundary \\
$e_{n,j}$ & typed event & Ordered event $j$ extracted from $x_n$ \\
$M_{n,j}$ & bounded resident state & State after decisions through event $j$ \\
$c_{n,j,i}$ & complete row & Target-conditioned candidate for action $i$ \\
$\tfPmaCalA_{n,j}$ & finite action set & Exact \tfPmaNull{} plus all valid complete row commits \\
$0$ or \tfPmaNull{} & action & Physical identity transition; no row or authority change \\
$\widetilde S_g^-,\widetilde S_g^+$ & causal prefix records & Typed pre-event and post-candidate/pre-action states for prefix group $g$ \\
$\tfPmaCalD_g^-,\tfPmaCalD_g^+$ & sigma-fields & Information legally available before eventization and after deterministic candidate construction \\
$F_{g,k}$ & future branch & Training/evaluation-only continuation $k$ sharing prefix $g$ \\
$Q_g,P_g$ & probability laws & Proposal/grouped-future law and target deployment law \\
$w_{g,k}$ & scalar & Nonnegative branch weight summing to one \\
$K_g^{\mathrm{eff}}$ & scalar & Effective weighted branch count $(\sum_kw_{g,k}^2)^{-1}$ \\
$Y_{g,a,k,h}$ & replay outcome & Action-conditioned outcome at horizon cell $(a,k,h)$ \\
$L_g(a,F)$ & bounded scalar & Registered horizon loss after action $a$ under future $F$ \\
$\rho_g$ & design flag & Registered i.i.d., stratified, or finite-population future-sampling regime \\
$\widehat R_{Q_g}(a)$ & estimator & Weighted empirical proposal-law action risk under grouped branches \\
$\widehat R_{P_g}(a)$ & estimator & Target-law estimate only under a registered equality or supported-reweighting bridge \\
$R_{Q_g}(a)$ & estimand & Conditional risk under proposal law \\
$R_{P_g}(a)$ & estimand & Conditional risk under target deployment law \\
$a_g^{\mathrm{cur}}$ & comparator action & Frozen current-risk minimizer without realized future \\
$\widehat a_g^{\mathrm{ER},Q}$ & proposal teacher & Empirical $Q_g$-risk minimizer over complete actions under frozen order $\prec_g$ \\
$\widehat a_g^{\mathrm{ER},P}$ & target teacher & Empirical target-risk minimizer only when a valid bridge identifies the target quantities \\
$a_{g,k}^{\mathrm{hind}}$ & information ceiling & Per-realized-future minimizer; unavailable at deployment \\
$G_g^{\mathrm{cur}},G_g^0$ & risk differences & Oracle advantages over current comparator and \tfPmaNull{} \\
$X_{g,a}$ & typed features & $\tfPmaCalD_g^+$-measurable future-blind deployment features for one complete action \\
$Q_\psi(X_{g,a})$ & model output & Predicted conditional action risk; no model exists here \\
$[L_g^Q(a),U_g^Q(a)]$ & interval & Proposal-law simultaneous interval for action $a$ \\
$[L_g^P(a),U_g^P(a)]$ & interval & Target-law interval after bridge and bridge uncertainty are propagated \\
$g_{n,j}$ & opportunity indicator & One iff the completed-segment event presents an eligible admission opportunity \\
$u_{n,j}$ & commit indicator & One iff an eligible event actually commits a non-\tfPmaNull{} action \\
$\mu$ & fixed scalar & Materiality/commit margin \\
$\tfPmaReg_g$ & nonnegative scalar & Excess target risk of selected versus best action \\
$\mathfrak R$ & resource vector & Tokens, compute, bytes, traffic, latency, energy, and storage terms \\
\end{longtable}
\endgroup
\addtocounter{table}{-1}% The captionless longtable advances the table counter; the caption above already owns this number.
\normalsize

\subsection{Literal status vocabulary}

\begin{table}[H]
\centering
\small
\begin{tabularx}{\linewidth}{@{}p{0.28\linewidth}Y@{}}
\toprule
Printed marker & Interpretation in this manuscript \\
\midrule
\tfPmaDefinitionStatus{} & Object or estimand defined; no existence, quality, or implementation claim \\
\tfPmaProvedStatus{} & Deduction valid when every displayed premise holds; premises not empirically established by the proof \\
\tfPmaConjecturedStatus{} & Open falsifiable hypothesis with no admitted supporting result \\
\tfPmaPrescriptionStatus{} & Required design or receipt for a conforming future test \\
\tfPmaPlannedStatus{} & Registered-form test described but not executed \\
\tfPmaUnvalidatedStatus{} & Gate failed, unopened, or unsupported by accepted evidence \\
\bottomrule
\end{tabularx}
\caption{Status markers are local. A proved lemma cannot promote the system that may someday instantiate it.}
\label{tf:pma:tab:status-ledger}
\end{table}

\subsection{Frozen evidence boundary}

\begin{itemize}
  \item frozen-checkpoint semantic-interface study: $0/9$ jobs qualified.
  \item fitted-support fault-localization study learned-content exactness: $0/73{,}728$ for learned routing/learned content and $0/73{,}728$ for oracle routing/learned content.
  \item fitted-support fault-localization study learned routing with oracle content: $18{,}544/73{,}728$; this does not establish learned content.
  \item fitted-support fault-localization study complete mechanical oracle: $73{,}728/73{,}728$; this is a mechanical ceiling, not evidence for a learned route, learned content, or predictive policy.
  \item Accepted terminal: fit/support failure for the tested pure latent-row interface; it does not isolate representation, decoder, or optimization and does not reject every possible MMLA realization.
  \item Predictive admission: no executed oracle test, no measured gap, no trained value model, and no learned policy.
\end{itemize}

These values are inherited only to delimit what this theory paper may claim. They are not reanalyzed, improved, or converted into a predictive-admission result.

%% file: theory_family/modules/dual_state/module.tex
\clearpage
\part{Predictive Dual-State Adaptation: Policy Learning and Authoritative Memory}
\label{tf:dual:part:mmla-rtt-dual-state}

\noindent\textbf{Scope.}
This part formalizes the composition of numerical policy state $\Phi$ and
authoritative memory $M$ while keeping their types, writer and reader
privileges, lifetimes, reset and rollback domains, versions, and cost ledgers
distinct. It develops the causal timeline, interaction estimands,
authorization rules, failure semantics, and experimental obligations.

\input{theory_family/modules/dual_state/sections/01_scope}
\input{theory_family/modules/dual_state/sections/02_typed_carriers}
\input{theory_family/modules/dual_state/sections/03_timeline}
\input{theory_family/modules/dual_state/sections/04_operations}
\input{theory_family/modules/dual_state/sections/05_rtt_classification}
\input{theory_family/modules/dual_state/sections/06_resets_rollbacks}
\input{theory_family/modules/dual_state/sections/07_compression}
\input{theory_family/modules/dual_state/sections/08_interaction_bounds}
\input{theory_family/modules/dual_state/sections/09_factorial_identification}
\input{theory_family/modules/dual_state/sections/10_observational_limits}
\input{theory_family/modules/dual_state/sections/11_authorization}
\input{theory_family/modules/dual_state/sections/12_costs}
\input{theory_family/modules/dual_state/sections/13_failures_recovery}

\input{theory_family/modules/dual_state/sections/14_protocol_obligations}
\input{theory_family/modules/dual_state/sections/15_related_limits}

\FloatBarrier
\input{theory_family/modules/dual_state/appendices/A_proofs}
\input{theory_family/modules/dual_state/appendices/B_factorial_protocol}

\input{theory_family/modules/dual_state/appendices/C_obligation_matrix}
\input{theory_family/modules/dual_state/appendices/D_symbols}

%% file: theory_family/modules/dual_state/sections/01_scope.tex
\section{Introduction}
\label{tf:dual:sec:scope}

\begin{figure}[H]
\centering
\includegraphics[width=\linewidth]{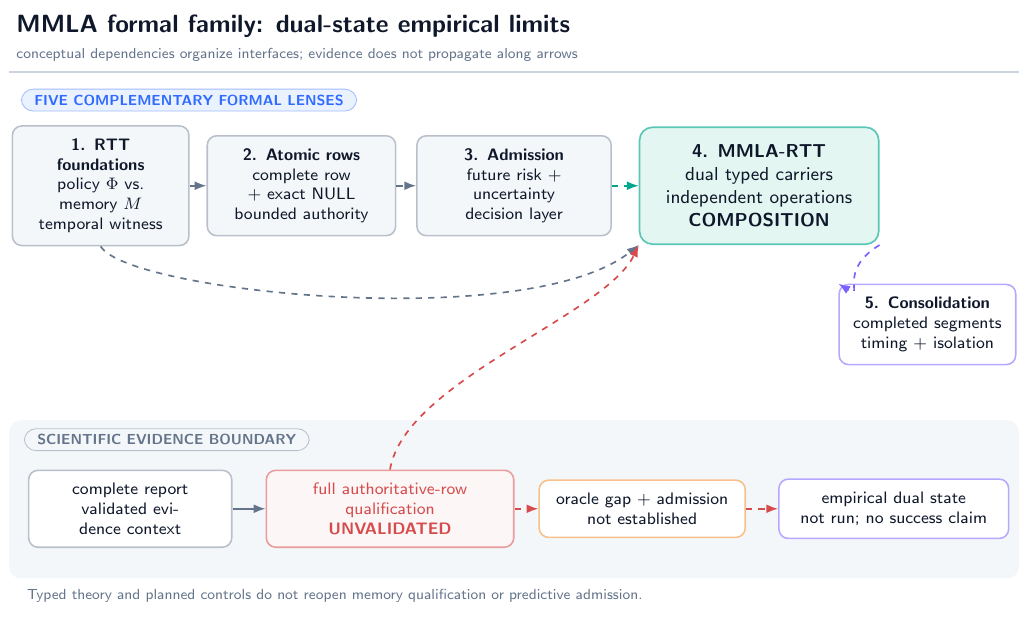}
\caption{Dual-state composition: separate policy and memory carriers, their coordination, and the independently required evidence for a joint benefit.}
\label{tf:dual:fig:family-tree}
\end{figure}

\subsection{The composition problem}

A model may improve a later attempt because its numerical behavior changed, because an authoritative record changed, because more tokens were placed in context, or because sampling happened to differ. Those mechanisms are not interchangeable. This paper studies the first two as a typed composition. It asks what must be logged, intervened on, and proved before a within-problem policy update and an authoritative memory commit can be said to coexist without hidden coupling.

\tfDualDefinitionStatus
\begin{tfDualDefinition}[Dual-state reasoning-time system]
A dual-state reasoning-time system has a trainable numerical policy carrier $\Phi\in\tfDualCalP$ and an authoritative memory carrier $M\in\tfDualCalM$. Its attempt kernel reads a declared causal workspace together with pinned versions of both carriers. The carriers expose disjoint mutation interfaces, version lines, rollback domains, and ledgers. A coordinator may check a cross-carrier compatibility predicate but does not own either state.
\end{tfDualDefinition}

The definition does not assert that such a system has been implemented, trained, or improved. It rules out a common shortcut: calling one opaque recurrent vector ``policy and memory'' and then inferring separate causal effects from a single output trace.

\subsection{Scientific boundary}

The historical frozen-checkpoint semantic-interface study qualified $0/9$ jobs. In the fitted-population diagnostic, the direct candidate was exact on $40/73{,}728$ probes; learned-route/learned-content and oracle-route/learned-content were each $0/73{,}728$; learned-route/oracle-content was $18{,}544/73{,}728$, exactly the routing count; and only the complete mechanical oracle reached $73{,}728/73{,}728$. Preservation was zero and $43{,}008$ cases were missing. This fit/support failure does not isolate representation, decoder, or optimization. At the current cutoff, later post-training adds restricted readout success but no qualifying generalized continuous-event latent arm in the latest study. These distinct results must not be collapsed into one model's learning curve.

\tfDualUnvalidatedStatus\quad No study identifies the strict policy-only effect and the complete authoritative-memory intervention required by this composition. Restricted readout success is neither full row conformance nor a policy-by-memory factorial. An expected-risk oracle margin, learned predictive admission, and joint PDSA benefit remain unestablished at the September 13, 2026 cutoff.

The composition uses four internal contracts: a strict within-problem policy-state witness, a complete authoritative-row transition with exact \tfDualNull{}, a future-risk admission objective, and a completed-segment causal boundary. These contracts organize the mechanism but provide no empirical transfer: each implementation claim still requires its own intervention and evidence.

\subsection{Contributions and nonclaims}

The paper develops typed semantics, an explicit causal schedule, independent operation algebras, mathematical bounds, factorial estimands, negative controls, resource ledgers, and falsification obligations. It does not supply a semantic serializer, a trusted row assembler implementation, an empirical policy update, a memory-admission model, a dual-state checkpoint, or a deployment result. Definitions, conditional proofs, prescriptions, and planned tests are marked locally and cannot promote an implementation claim.

\subsection{Central falsifiable questions}

Let $Y_{pm}$ be a preregistered loss under policy intervention $p\in\{0,1\}$ and memory intervention $m\in\{0,1\}$. The primary interaction estimand is
\begin{equation}
\Delta_{\Phi M}
=\tfDualE[Y_{11}-Y_{10}-Y_{01}+Y_{00}],
\label{tf:dual:eq:interaction-intro}
\end{equation}
where lower loss is better. A negative value would indicate complementarity on that endpoint under the declared interventions; positive values indicate antagonism. The paper proves how this contrast is identified under randomization and compliance, not what its sign is. A second question is categorical: does feedback at attempt $r$ update $\Phi$ and causally influence attempt $r+1$ on the same still-unresolved problem? Without that path, the mechanism is not strict reasoning-time learning even if $M$ changes.

\tfDualConjecturedStatus\quad Independent policy adaptation and authoritative memory may be complementary in some domains, substitutable in others, neutral when one carrier is irrelevant, and harmful when either update contaminates the other carrier's inputs. All four regimes remain possible before intervention.

%% file: theory_family/modules/dual_state/sections/02_typed_carriers.tex
\section{Typed Carriers and Read Views}
\label{tf:dual:sec:types}

\subsection{Probability space and causal workspace}

Fix a probability space $(\Omega,\tfDualCalF,\tfDualP)$ and a problem instance $q$. Attempts are indexed by $r=0,1,\ldots,R_q$. Let $\tfDualCalF_{q,r,t}$ be the information available immediately before micro-event $t$ of attempt $r$. A bounded causal workspace
\begin{equation}
W_{q,r,t}=\bigl(q,\,x_{q,r,\le t},\,z_{q,<r},\,f_{q,<r},\,u_{q,r,t}\bigr)
\label{tf:dual:eq:workspace}
\end{equation}
contains the problem, visible prefix, prior observable attempt summaries $z$, admissible feedback $f$, and a declared controller summary $u$. The audit contract requires the declared observable state, versioned receipts, and causal inputs needed for the stated reproducibility boundary; it does not require an unbounded internal execution trace. Any feature used to update either carrier must be measurable with respect to its declared information set.

\tfDualDefinitionStatus
\begin{tfDualDefinition}[Policy carrier]
The policy carrier at event $e$ is
\begin{equation}
P_e=(\Phi_e,O_e,v_e^{\Phi},\ell_e^{\Phi},\tau_e^{\Phi},\rho_e^{\Phi}),
\end{equation}
where $\Phi_e\in\tfDualCalP\subseteq\tfDualR^d$ is trainable numerical state, $O_e$ is optimizer state, $v_e^{\Phi}$ is a monotone version, $\ell_e^{\Phi}$ is a lineage identifier, $\tau_e^{\Phi}$ declares lifetime and scope, and $\rho_e^{\Phi}$ is a receipt pointer. Its privileged writer is the policy updater; its readers are declared inference modules.
\end{tfDualDefinition}

\tfDualDefinitionStatus
\begin{tfDualDefinition}[Authoritative memory carrier]
The memory carrier at event $e$ is
\begin{equation}
A_e=(M_e,v_e^M,\ell_e^M,\tau_e^M,\rho_e^M),
\end{equation}
where $M_e$ is a bounded map from physical slots to complete typed rows or empty slots, $v_e^M$ is an authoritative memory version, $\ell_e^M$ is its lineage, $\tau_e^M$ declares retention and tenant scope, and $\rho_e^M$ points to a commit or \tfDualNull{} receipt. Its privileged writer is a trusted memory assembler/committer; policy gradients never directly mutate it.
\end{tfDualDefinition}

\tfDualDefinitionStatus
\begin{tfDualDefinition}[Typed coupling interface]
\label{tf:dual:def:typed-interface}
For declared policy-read type $\tfDualCalH_{\Phi}$, memory-read type $\tfDualCalH_M$, and output law $\Delta(\tfDualCalZ)$, a coupling interface is a registered tuple
\begin{equation}
\tfDualCalI=\bigl(r_{\Phi}:\tfDualCalP\!\times\!\tfDualCalO\!\times\!\tfDualCalW\to\tfDualCalH_{\Phi},\;
r_M:\tfDualCalM\!\times\!\tfDualCalW\to\tfDualCalH_M,\;
d:\tfDualCalH_{\Phi}\!\times\!\tfDualCalH_M\!\times\!\tfDualCalW\!\times\!\tfDualCalG\to\Delta(\tfDualCalZ)\bigr).
\label{tf:dual:eq:typed-interface}
\end{equation}
The two reader codomains are distinct declared types.  No cast from a policy payload to an authoritative row, or conversely, is available unless it is separately registered, authorized, and included in $\tfDualCalI$.
\end{tfDualDefinition}

\tfDualDefinitionStatus
\begin{tfDualDefinition}[Compatibility witness]
\label{tf:dual:def:compat-witness}
A witness $\kappa(P,A,C;\tfDualCalI)$ binds the exact policy and memory payload hashes, versions and lineages; tenant and ACL scope; base-model and tokenizer hashes; feature-schema hashes; reader and decoder versions; input/output signatures in Equation~\eqref{tf:dual:eq:typed-interface}; validity intervals; and the registered fallback.  Verification is a typed decision
\begin{equation}
\tfDualVerify_{\tfDualCalI}(P,A,C,\kappa)\in\{0,1\}
\label{tf:dual:eq:witness-verify}
\end{equation}
whose receipt records every checked field.  Equality of visible version numbers is neither necessary nor sufficient for verification.
\end{tfDualDefinition}

\begin{figure}[H]
\centering
\includegraphics[width=\linewidth]{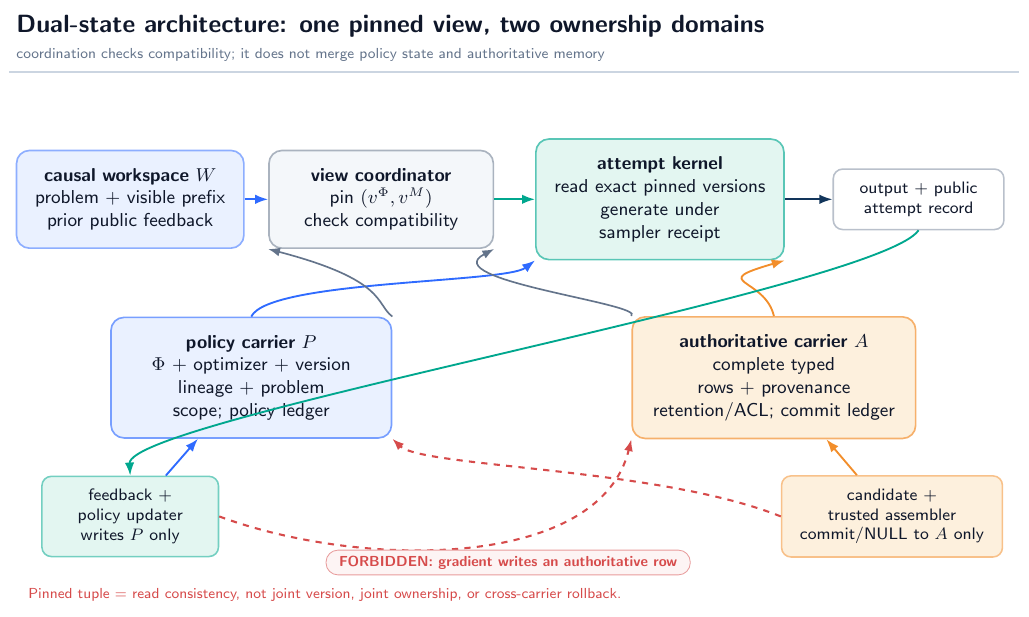}
\caption{\textbf{Typed dual-state architecture (\tfDualDefinitionStatus{} / \tfDualUnvalidatedStatus{}).} The attempt kernel reads a pinned view. Feedback may drive a policy update; a separate trusted path may assemble and commit a complete row. Dashed coral paths are forbidden privilege crossings.}
\label{tf:dual:fig:architecture}
\end{figure}

\subsection{Type, privilege, lifetime, and failure separation}

\begin{table}[H]
\centering
\small
\begin{tabularx}{\linewidth}{@{}p{0.18\linewidth}Y Y@{}}
\toprule
Property & Policy carrier $P$ & Memory carrier $A$ \\
\midrule
Payload & Numerical parameters, optimizer moments, update metadata & Complete typed rows, versions, provenance, ACL/protection, validity, tombstone semantics \\
Privileged writer & Policy-update service under feedback contract & Trusted assembler and atomic committer \\
Typical lifetime & Attempt/problem/session; default problem-scoped & Cross-attempt and potentially persistent/archival under retention policy \\
Reset target & Chosen policy baseline and optimizer state & Chosen authoritative snapshot or empty/default row set \\
Rollback unit & Numerical checkpoint plus optimizer/lineage receipt & Full row version or full memory snapshot plus commit ledger \\
Failure examples & Divergence, reward hacking, optimizer corruption, catastrophic local drift & Partial row, stale authority, ACL breach, provenance loss, rollback fork \\
Primary ledger & Gradient/update, loss/reward, seed, optimizer, version & Candidate, validation, commit/\tfDualNull{}, row version, archive, index, deletion \\
\bottomrule
\end{tabularx}
\caption{The carriers are distinct across every operational dimension required by this paper.}
\label{tf:dual:tab:carrier-separation}
\end{table}

\tfDualDefinitionStatus
\begin{tfDualDefinition}[Pinned read view]
At the start of attempt $r$, the coordinator emits
\begin{equation}
V_{q,r}=\bigl(v_{q,r}^{\Phi},v_{q,r}^{M},\ell_{q,r}^{\Phi},\ell_{q,r}^{M},c_{q,r}\bigr),
\end{equation}
where $c_{q,r}$ is a compatibility-receipt identifier. The attempt kernel may read only the named carrier versions. Pinning provides read consistency for that attempt; it does not merge carriers, equate versions, assign joint ownership, or imply joint commit/rollback.
\end{tfDualDefinition}

\begin{tfDualAssumption}[Interface realizability and sound verification]
\label{tf:dual:ass:interface-soundness}
Every emitted pin receipt $c_{q,r}$ commits to one registered $\tfDualCalI$, one witness $\kappa$, the exact carrier and context hashes, and a successful evaluation of Equation~\eqref{tf:dual:eq:witness-verify}.  The registered readers and decoder are total on the declared validity domain and are the components actually used by the attempt.  Outside that domain, or when no verified witness exists, the coordinator refuses the pair or invokes the named fallback; it does not infer compatibility from nominal versions or output behavior.
\end{tfDualAssumption}

\begin{tfDualAssumption}[Typed authorization]
\label{tf:dual:ass:typed-auth}
Every state-changing event carries a carrier tag, actor identity, scope, predecessor version, proposed successor version, payload hash, and authorization decision. A policy event can write only $P$; a memory event can write only $A$. The coordinator can reject an incompatible view but cannot synthesize either successor.
\end{tfDualAssumption}

\tfDualProvedStatus
\begin{tfDualProposition}[Syntactic non-conflation]
\label{tf:dual:prop:type-safety}
Under Assumption~\ref{tf:dual:ass:typed-auth}, no well-typed execution trace can apply a policy mutation to $M$ or a memory commit to $\Phi$. Equality of serialized payload bytes, if it occurs, does not change the carrier tag or privilege domain.
\end{tfDualProposition}

The proposition is a property of the specified interface, not evidence that an implementation enforces it. Its proof is in Appendix~\ref{tf:dual:app:proofs}.

\subsection{Complete state and carrier inventory}

The system state at event $e$ is not just $(\Phi_e,M_e)$. For causal and cost completeness we use
\begin{equation}
S_e=(W_e,P_e,A_e,C_e,G_e,Q_e),
\label{tf:dual:eq:full-state}
\end{equation}
where $C_e$ is immutable causal context, $G_e$ is generation/sampler state including seed and decoding controls, and $Q_e$ contains queues, snapshots, caches, and external ledgers. Omitting $G_e$ can misattribute sampling variation to learning. Omitting $Q_e$ can hide stale cache reads or delayed commits. Neither auxiliary state becomes authoritative memory merely by appearing in the tuple.

%% file: theory_family/modules/dual_state/sections/03_timeline.tex
\section{Fine-Grained Causal Timeline}
\label{tf:dual:sec:timeline}

\subsection{Micro-events, not an opaque recurrence}

Let $e=0,1,\ldots$ index globally ordered micro-events. Each event has one of the tags
\begin{equation}
\mathsf{GEN},\ \mathsf{OBS},\ \mathsf{FDBK},\ \mathsf{PUPD},\ \mathsf{CAND},\ \mathsf{MVAL},\ \mathsf{MCOM},\ \mathsf{PIN},\ \mathsf{RESET},\ \mathsf{ROLLBACK}.
\end{equation}
The transition is
\begin{equation}
S_{e+1}=T_{\kappa_e}(S_e,\xi_e),
\label{tf:dual:eq:microtransition}
\end{equation}
where $\kappa_e$ is the event tag and $\xi_e$ is its typed input. There is no unlabelled map $S_{e+1}=F(S_e)$ from which causal ownership must be guessed after the fact.

\begin{figure}[H]
\centering
\includegraphics[width=\linewidth]{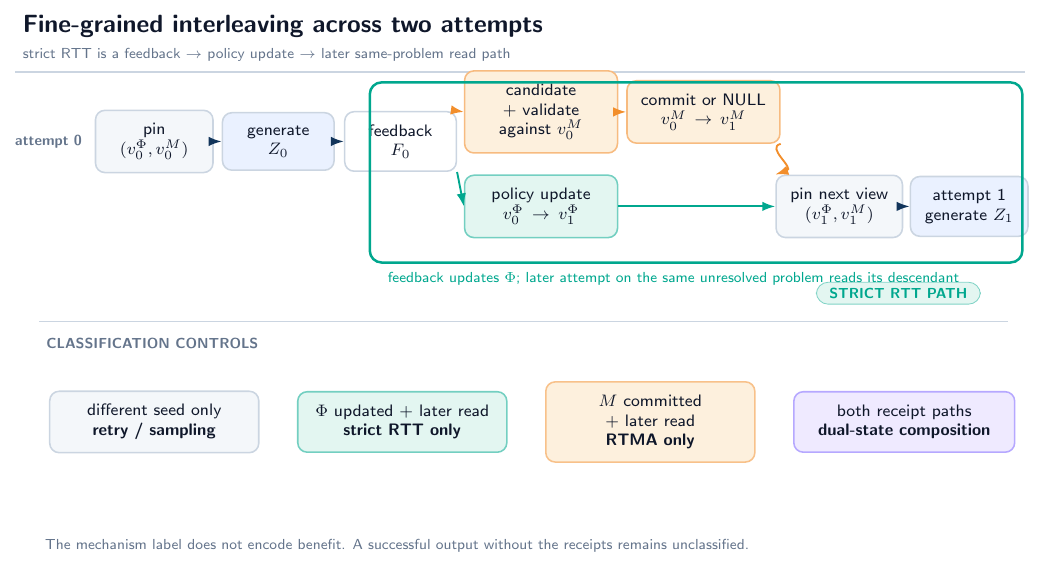}
\caption{\textbf{Declared interleaving for two attempts (\tfDualDefinitionStatus{}).} Attempt 0 pins a view, generates, receives admissible feedback, and may trigger two independent successor events. Attempt 1 counts as strict RTT only if it reads the updated $\Phi$ while the problem remains unresolved. A memory commit has its own later-reuse path.}
\label{tf:dual:fig:timeline}
\end{figure}

For one illustrative order, attempt $r$ pins $(v_r^{\Phi},v_r^M)$, generates $Z_r$, obtains feedback $F_r$, performs a policy update to $v_{r+1}^{\Phi}$, independently proposes and validates a memory candidate, optionally commits $v_{r+1}^{M}$, and then pins a new view for attempt $r+1$. Other orders are permitted only when declared. In particular, memory admission may precede policy update, or either event may be absent. Section~\ref{tf:dual:sec:operations} shows why the order can matter.

\subsection{Filtration and no future leakage}

\begin{tfDualAssumption}[Causal measurability]
\label{tf:dual:ass:causal}
At event $e$, generation and policy-update inputs are $\tfDualCalF_e$-measurable. Memory candidates may use only the candidate-construction information set declared for that event. A deployment event cannot read future feedback, uncommitted evaluator truth, or a later carrier version. Each output receipt records the exact predecessor view.
\end{tfDualAssumption}

\begin{tfDualProposition}[Causal trace measurability]
\label{tf:dual:prop:measurable}
Under Assumption~\ref{tf:dual:ass:causal}, every generated token, policy update, memory candidate, validation decision, commit decision, and pinned view is measurable with respect to the filtration at its event. The assumption's separate predecessor-access restriction excludes a later version as a parent of an earlier output in the declared structural trace. This structural exclusion is not a consequence of measurability alone.
\end{tfDualProposition}

This statement excludes temporal leakage under the model; it does not prove that a real logging system timestamps events correctly.

\subsection{Attempt-boundary state}

Let $b(q,r)$ and $d(q,r)$ denote the begin and terminal event of attempt $r$. Define the attempt record
\begin{equation}
\mathcal{R}_{q,r}=
\bigl(q,r,V_{q,r},G_{b(q,r)},Z_{q,r},F_{q,r},
E_{q,r}^{\Phi},E_{q,r}^{M},d(q,r)\bigr),
\label{tf:dual:eq:attempt-record}
\end{equation}
where $E_{q,r}^{\Phi}$ and $E_{q,r}^{M}$ are ordered lists of carrier-specific events. An empty list is meaningful: it proves that no update of that carrier was admitted during the interval. A missing list is not equivalent to an empty list.

\tfDualPrescriptionStatus\quad A conforming trace must preserve the declared attempt records, carrier ledgers, pin receipts, reset receipts, sampler state, and causal inputs needed for the registered analysis; an unbounded internal execution trace is outside this contract.

\subsection{Interleaving table}

\begin{table}[H]
\centering
\small
\begin{tabularx}{\linewidth}{@{}p{0.10\linewidth}p{0.15\linewidth}p{0.15\linewidth}Y Y@{}}
\toprule
Order & Event & Reads & May write & Required receipt \\
\midrule
1 & \textsf{PIN} & latest authorized versions & view tuple only & both version/lineage IDs and compatibility result \\
2 & \textsf{GEN} & pinned view, workspace, sampler & output and sampler trace & prompt/input hash, seed, decoding contract \\
3 & \textsf{FDBK} & output, evaluator-visible outcome & feedback record & source, timing, scope, admissibility \\
4a & \textsf{PUPD} & pinned $\Phi$, feedback, optimizer & policy carrier only & objective, gradient/update, predecessor/successor \\
4b & {\scriptsize\textsf{CAND/MCOM}} & pinned $M$, authorized evidence & memory carrier only after validation & complete candidate, validator, commit/\tfDualNull{} \\
5 & \textsf{PIN} & independently current carriers & next view tuple & compatibility and exact versions \\
\bottomrule
\end{tabularx}
\caption{One admissible partial order. Events 4a and 4b are not assumed to commute.}
\label{tf:dual:tab:interleaving}
\end{table}

%% file: theory_family/modules/dual_state/sections/04_operations.tex
\section{Independent Operation Algebras}
\label{tf:dual:sec:operations}

\subsection{Carrier-specific maps}

Let $U_e:\tfDualCalP\times\tfDualCalO\to\tfDualCalP\times\tfDualCalO$ be an authorized policy update and $K_e:\tfDualCalM\to\tfDualCalM$ an authorized complete-row memory transition. Their lifted maps on the full state are
\begin{align}
\widetilde U_e(W,P,A,C,G,Q)&=(W,U_e(P),A,C,G,Q'),\\
\widetilde K_e(W,P,A,C,G,Q)&=(W,P,K_e(A),C,G,Q''),
\end{align}
where $Q'$ and $Q''$ append disjoint receipts. A coordinator event may later evaluate $\tfDualCompat(P,A,C)$ and refuse a new pin. It does not retroactively rewrite either carrier.

\tfDualDefinitionStatus
\begin{tfDualDefinition}[Operation algebras]
$\mathfrak O_{\Phi}$ contains policy snapshot, update, freeze, reset, swap, and rollback maps. $\mathfrak O_M$ contains memory snapshot, complete-row commit, exact \tfDualNull{}, reset, swap, row rollback, full-snapshot rollback, archive, and verified deletion maps. Each operation is closed over its own carrier type and appends to its own ledger.
\end{tfDualDefinition}

\subsection{Commutation is conditional}

The carrier writes are disjoint, but their \emph{inputs} can depend on a shared workspace or on the other carrier. Disjoint destination fields alone do not imply semantic commutation.

\begin{tfDualAssumption}[Input stability for a pair of events]
\label{tf:dual:ass:input-stability}
For a policy update $U$ and memory transition $K$ proposed from state $S$, (i) $U$'s typed input and authorization decision are invariant to applying $K$ first; (ii) $K$'s candidate, validation, and authorization decision are invariant to applying $U$ first; (iii) receipt append order is normalized without deleting event time; and (iv) both successor pairs satisfy the same compatibility predicate.
\end{tfDualAssumption}

\tfDualProvedStatus
\begin{tfDualTheorem}[Conditional commutation]
\label{tf:dual:thm:commutation}
Under Assumptions~\ref{tf:dual:ass:typed-auth} and \ref{tf:dual:ass:input-stability}, the carrier projection of the two-event successor is order-invariant:
\begin{equation}
\pi_{P,A}(\widetilde K\circ\widetilde U(S))
=\pi_{P,A}(\widetilde U\circ\widetilde K(S)).
\end{equation}
The full audit trace remains ordered and therefore need not be byte-identical.
\end{tfDualTheorem}

\begin{tfDualCounterexample}[Disjoint writes, noncommuting decisions]
\label{tf:dual:cex:noncommute}
Suppose the memory candidate is admitted only if the current policy assigns it confidence above $1/2$. From $\Phi_0$, confidence is $0.4$ and $K$ returns \tfDualNull{}. Feedback update $U$ raises confidence to $0.6$. Then $K\circ U$ commits a row whereas $U\circ K$ leaves memory unchanged, despite disjoint write privileges. Reversing the dependency---letting the policy loss read newly committed memory---gives the symmetric failure.
\end{tfDualCounterexample}

The counterexample is theoretical. It does not instantiate predictive admission or show that confidence is a valid admission score.

\subsection{Compatibility without joint ownership}

For a registered interface $\tfDualCalI$, let
\begin{equation}
\Gamma_{\tfDualCalI}(P,A,C)=
\{\kappa:\tfDualVerify_{\tfDualCalI}(P,A,C,\kappa)=1\}
\quad\text{and}\quad
\tfDualCompat_{\tfDualCalI}(P,A,C)=\tfDualOne\{\Gamma_{\tfDualCalI}(P,A,C)\ne\varnothing\}.
\label{tf:dual:eq:compat}
\end{equation}
A successful decision names a concrete $\kappa$; the Boolean alone is not a portable proof.  If the set is empty, the only authorized action is to withhold a new view or route to the witness-declared fallback. The coordinator must not silently reset $P$, rewrite $A$, or manufacture a composite version number.

\begin{tfDualDefinition}[Coupled transition protocol]
\label{tf:dual:def:coupled-transition}
Given proposed successors $P^+=U(P)$ and $A^+=K(A)$, the policy and memory services first complete their own typed authorization, validation, and carrier-local receipts $\rho_{\Phi}^+$ and $\rho_M^+$.  The coordinator then evaluates $\Gamma_{\tfDualCalI}(P^+,A^+,C)$.  Only a verified witness permits a coordination manifest whose digest is carried in an authenticated receipt:
\begin{equation}
\rho_{\mathrm{coord}}=H(\rho_{\Phi}^+,\rho_M^+,\kappa,\tfDualCalI,C),
\label{tf:dual:eq:coordination-manifest}
\end{equation}
The hash identifies the manifest; it is not itself a signature or proof of
authentication. The assumed receipt-validation procedure authenticates and
verifies, but does not replace, the two carrier-local receipts. The
manifest is a read-consistency object: it grants neither write authority to either
carrier nor permission for a cross-carrier rollback. Failure creates no new joint
view; it neither retracts a valid carrier-local successor nor implicitly rolls either
carrier back.
\end{tfDualDefinition}

\tfDualProvedStatus
\begin{tfDualProposition}[Coupled-update closure]
\label{tf:dual:prop:coupled-closure}
Under Assumptions~\ref{tf:dual:ass:typed-auth} and \ref{tf:dual:ass:interface-soundness}, every joint view produced by Definition~\ref{tf:dual:def:coupled-transition} contains one well-typed policy successor, one well-typed memory successor, and a receipt binding their exact lineages to a verified interface witness.  If witness verification fails, no joint view is produced and neither carrier is silently changed by the coordinator.
\end{tfDualProposition}

\begin{tfDualCounterexample}[Nominal-version agreement is not compatibility]
\label{tf:dual:cex:version-compat}
Let $v^{\Phi}=v^M=7$ and let tenant, tokenizer, and base-model identifiers agree.  Suppose the registered policy reader emits $\tfDualR^{64}$ while the installed decoder requires $\tfDualR^{96}$, or the memory reader's ACL excludes the requesting actor.  All nominal identifiers can match while no typed, authorized witness exists.  A version-pair check would accept an unrealizable read view; Equation~\eqref{tf:dual:eq:compat} rejects it.
\end{tfDualCounterexample}

\begin{tfDualProposition}[Nominal version-pair non-identification]
\label{tf:dual:prop:no-composite}
Knowledge of a pinned pair $(v^{\Phi},v^M)$ alone does not determine either predecessor, lifetime, lineage, or rollback target. In particular, there exist two valid histories with the same pinned pair but different policy lineages, and two with the same pair but different memory lineages. A passing compatibility witness binds the exact registered lineages for its view as required by Definition~\ref{tf:dual:def:compat-witness}; it does not merge those histories or authorize rollback.
\end{tfDualProposition}

The proof is by constructing independent fork-and-rejoin histories in Appendix~\ref{tf:dual:app:proofs}. Histories with the same nominal pair require different lineage-bound witnesses. A composite snapshot can be defined, but only as an explicit manifest containing two independently verifiable snapshots and a coordination receipt.

\begin{figure}[H]
\centering
\includegraphics[width=\linewidth]{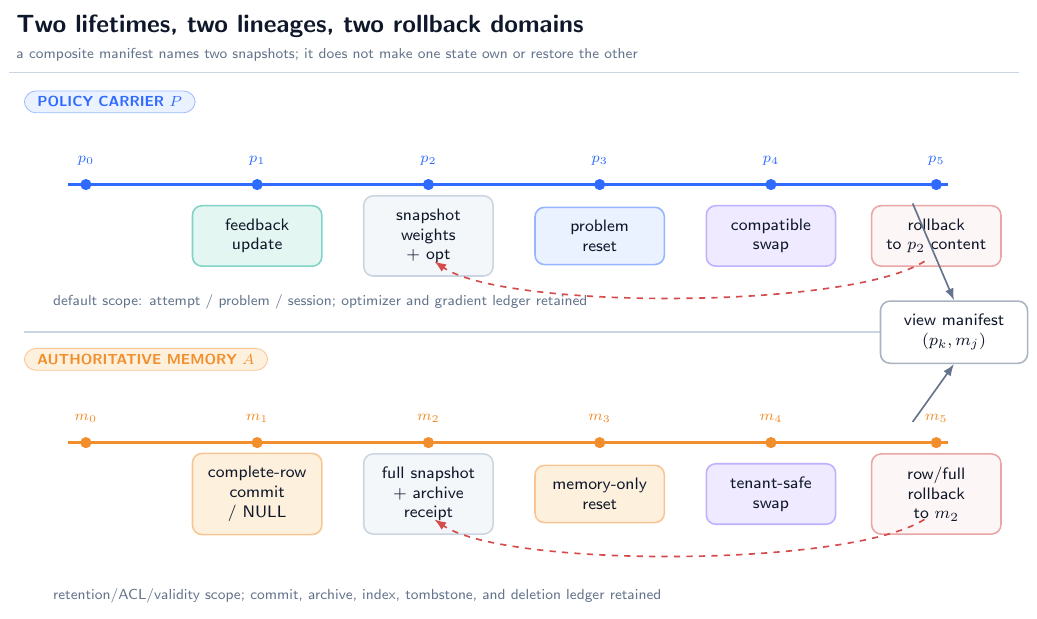}
\caption{\textbf{Independent lifetimes, snapshots, rollback domains, and ledgers (\tfDualPrescriptionStatus{}).} A view manifest can name both carrier versions; it cannot collapse their lineages or authorize silent cross-restoration.}
\label{tf:dual:fig:lifetimes}
\end{figure}

%% file: theory_family/modules/dual_state/sections/05_rtt_classification.tex
\section{Strict RTT, Memory Adaptation, and Composition}
\label{tf:dual:sec:classification}

\subsection{A path-based definition}

Let $Q_q(r)=1$ mean that problem $q$ remains unresolved immediately after attempt $r$. Let $F_{q,r}$ be admissible feedback and let $\Phi_{q,r}^{\mathrm{read}}$ denote the policy version pinned by attempt $r$.

\tfDualDefinitionStatus
\begin{tfDualDefinition}[Strict reasoning-time learning]
An execution exhibits strict reasoning-time learning on problem $q$ between attempts $r$ and $s>r$ if all of the following hold:
\begin{enumerate}
  \item $Q_q(k)=1$ for every $k=r,\ldots,s-1$;
  \item an authorized update event writes $\Phi^+=U(\Phi^-,F_{q,r},\eta)$ after feedback from attempt $r$;
  \item a later attempt $s$ pins and reads a descendant of $\Phi^+$;
  \item a registered intervention on that policy lineage can change the distribution of the later attempt while holding the declared non-policy state fixed.
\end{enumerate}
The update has default problem scope unless a broader lifetime is explicitly authorized.
\end{tfDualDefinition}

Conditions 1--3 establish timing and reuse; condition 4 turns a logged association into a causal mechanism. Merely computing a gradient, saving an adapter after the problem is solved, or updating on one problem and using the result only on unrelated future problems does not satisfy the strict within-problem definition used here.

\tfDualDefinitionStatus
\begin{tfDualDefinition}[Reasoning-time memory adaptation]
An execution exhibits reasoning-time memory adaptation (\tfDualRTMA{}) when an
authorized event commits a non-NULL successor of $M$ during an unresolved problem
and a later attempt can read that successor through a verified pin. An exact
$\tfDualNull{}$ receipt is a negative/no-op control and does not satisfy this
definition. \tfDualRTMA{} does not require a policy update and is not, by itself,
strict RTT.
\end{tfDualDefinition}

\tfDualDefinitionStatus
\begin{tfDualDefinition}[Dual-state composition]
A dual-state composition is an execution interval containing both a strict-RTT policy path and an \tfDualRTMA{} path, with independently identifiable interventions and receipts. The definition does not require either main effect or their interaction to be beneficial.
\end{tfDualDefinition}

\subsection{Classification theorem}

\begin{tfDualAssumption}[Attempt and carrier observability]
\label{tf:dual:ass:observability}
The audit exposes problem identity and unresolved status, event order, feedback source, carrier tags, predecessor and successor lineages, pinned versions, and intervention compliance. Missing receipts are represented as missing, not imputed as no-ops.
\end{tfDualAssumption}

\tfDualProvedStatus
\begin{tfDualTheorem}[Mechanism classification]
\label{tf:dual:thm:classification}
Under Assumptions~\ref{tf:dual:ass:causal} and \ref{tf:dual:ass:observability},
and provided that every required policy, memory, pin, intervention, and outcome
receipt for the interval is present, non-conflicting, and verifiable, the interval
belongs to exactly one of the following four observable mechanism classes:
\begin{center}
\begin{tabular}{c|c|c}
 & no later read of updated $M$ & later read of updated $M$ \\
\hline
no strict policy path & static/retry & \tfDualRTMA{} only \\
strict policy path & strict RTT only & dual-state composition
\end{tabular}
\end{center}
where ``path present'' uses the definitions above and a memory path requires a
non-NULL successor followed by a later read. If receipt coverage is missing,
conflicting, or unverifiable, the interval is assigned the separate status
\textsf{UNCLASSIFIED}; it is not forced into a no-op or static/retry cell. The
classes say nothing about endpoint improvement.
\end{tfDualTheorem}

\begin{tfDualCorollary}[Memory mutation is insufficient for strict RTT]
\label{tf:dual:cor:memory-not-rtt}
If $\Phi$ is frozen throughout an interval and no policy descendant is read by a later same-problem attempt, any change caused solely by $M$ belongs to \tfDualRTMA{} only, not strict RTT.
\end{tfDualCorollary}

\begin{tfDualCounterexample}[Retry masquerading as learning]
Two attempts use identical $\Phi$, identical $M$, and different sampler seeds. The second succeeds. The output improvement is compatible with pure sampling variation and therefore certifies neither RTT nor \tfDualRTMA{}.
\end{tfDualCounterexample}

\begin{tfDualCounterexample}[Post-resolution adaptation]
Attempt $r$ solves the problem, after which feedback updates $\Phi$ for a future task. This may be online or continual adaptation, but it is not strict reasoning-time learning for the resolved problem under the within-problem definition above.
\end{tfDualCounterexample}

\subsection{Mechanism receipts}

\begin{table}[H]
\centering
\small
\begin{tabularx}{\linewidth}{@{}p{0.21\linewidth}Y Y@{}}
\toprule
Claimed mechanism & Minimum positive receipt & Required negative control \\
\midrule
Strict RTT & same problem, unresolved marker, feedback, $\Phi$ predecessor/successor, later pinned descendant & reset/fresh/frozen or swapped $\Phi$ with $M$ and sampler held fixed \\
\tfDualRTMA{} & non-NULL complete-row commit, $M$ predecessor/successor, later pinned descendant & memory reset/swap/freeze with $\Phi$ and sampler held fixed \\
Dual-state composition & both receipt chains plus exact interleaving & independent $2\times2$ interventions and compliance audit \\
Benefit & preregistered endpoint under randomized intervention & seed, order, leakage, and attrition checks \\
\bottomrule
\end{tabularx}
\caption{A mechanism receipt is necessary but not an outcome result.}
\label{tf:dual:tab:mechanism-receipts}
\end{table}

\noindent\textbf{Receipt gap.} An exact \tfDualNull{} receipt, a failed pin, or a
missing successor receipt is retained as an observed terminal or missingness event;
none is silently relabeled as a positive \tfDualRTMA{} path.

\tfDualPrescriptionStatus\quad Names such as ``test-time learning,'' ``memory,'' or ``self-improvement'' are never compliance evidence. The trace and interventions determine classification.

%% file: theory_family/modules/dual_state/sections/06_resets_rollbacks.tex
\section{Reset, Swap, Snapshot, and Rollback Semantics}
\label{tf:dual:sec:reset}

\subsection{Independent operations}

For carrier $X\in\{\Phi,M\}$, write $\tfDualOpSnapshot_X(h)$ for a content-addressed snapshot at history position $h$, $\tfDualOpReset_X(b)$ for reset to a declared baseline $b$, $\tfDualOpSwap_X(s)$ for replacement by an independently sourced compatible snapshot, and $\tfDualOpRollback_X(h)$ for restoration to a prior authorized snapshot with a new rollback receipt. Rollback never erases history; it creates a successor whose content may equal an earlier snapshot.

\begin{tfDualAssumption}[Carrier-local restoration]
\label{tf:dual:ass:local-restore}
Each restoration operation names exactly one carrier, verifies its target snapshot and lineage policy, changes only that carrier and its ledger, invalidates carrier-local caches, and appends a successor version. Cross-carrier cache invalidation or pin refusal may occur, but the other carrier's content and lineage remain unchanged unless a separate authorized event names it.
\end{tfDualAssumption}

\tfDualProvedStatus
\begin{tfDualTheorem}[Independent rollback preservation]
\label{tf:dual:thm:rollback}
Under Assumptions~\ref{tf:dual:ass:typed-auth} and \ref{tf:dual:ass:local-restore}, applying $\tfDualOpRollback_{\Phi}(h)$ leaves the content, version, and lineage of $M$ unchanged; applying $\tfDualOpRollback_M(k)$ leaves the content, version, and lineage of $\Phi$ unchanged. A subsequent compatibility failure may block a pin but cannot silently restore the other carrier.
\end{tfDualTheorem}

\begin{tfDualCorollary}[Explicit composite recovery]
\label{tf:dual:cor:composite-recovery}
Restoring both carriers requires two authorized carrier-local events or an explicit composite manifest that expands into those two events. A single unnamed ``system rollback'' is nonconforming.
\end{tfDualCorollary}

\subsection{Reset and swap controls}

\begin{figure}[H]
\centering
\includegraphics[width=\linewidth]{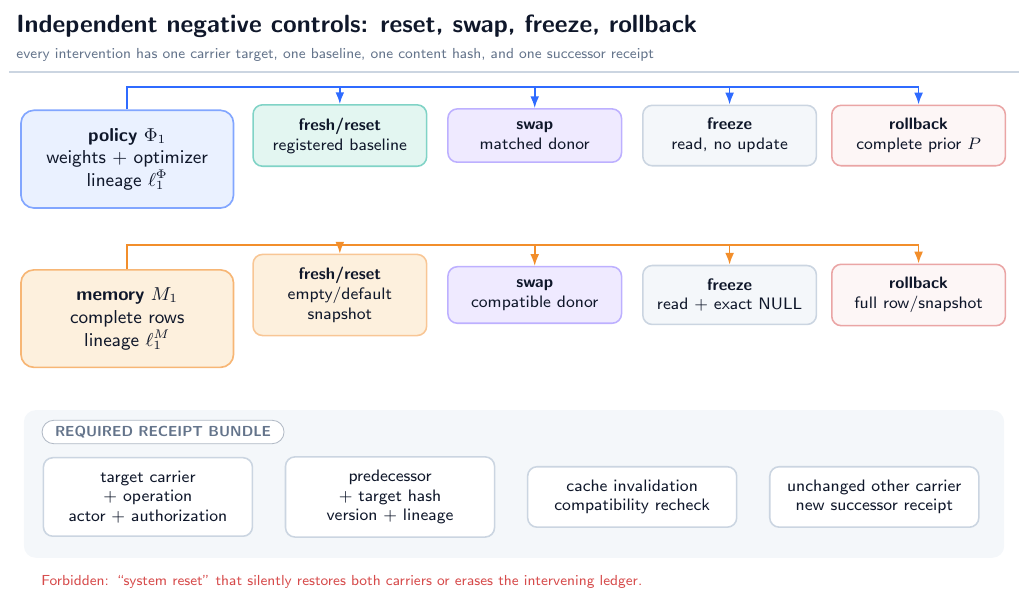}
\caption{\textbf{Reset, swap, freeze, and rollback controls with receipts (\tfDualPrescriptionStatus{}).} Each intervention targets one carrier. A composite control is two declared interventions, never an implicit cross-state restoration.}
\label{tf:dual:fig:negative-controls}
\end{figure}

The controls answer different questions:
\begin{itemize}
  \item $\tfDualFresh_{\Phi}$ restores the registered initialization and optimizer state; $\tfDualReset_{\Phi}$ may restore another declared problem baseline.
  \item $\tfDualSwap_{\Phi}$ imports a compatible policy snapshot from a matched unit, testing content specificity while preserving nominal update count.
  \item $\tfDualFreeze_{\Phi}$ permits reads but blocks updates, separating extra attempts from adaptation.
  \item $\tfDualRollback_{\Phi}$ tests recovery to a known policy snapshot and must include optimizer state unless the protocol explicitly studies optimizer mismatch.
  \item $\tfDualFresh_M$ installs the registered empty/default authoritative state; $\tfDualReset_M$ restores a specified authoritative snapshot.
  \item $\tfDualSwap_M$ imports a compatible memory snapshot from a matched unit, testing whether success depends on problem-specific authoritative content.
  \item $\tfDualFreeze_M$ permits reads and exact \tfDualNull{} only; $\tfDualRollback_M$ restores a full row version or full snapshot without changing $\Phi$.
\end{itemize}

\begin{table}[H]
\centering
\small
\begin{tabularx}{\linewidth}{@{}p{0.16\linewidth}p{0.17\linewidth}p{0.18\linewidth}Y@{}}
\toprule
Operation & Target content & Required companion state & Invalid shortcut \\
\midrule
$\tfDualOpReset_{\Phi}$ & parameters and declared optimizer baseline & same $M$, context, sampler contract & zeroing weights while keeping stale moments \\
$\tfDualOpSwap_{\Phi}$ & compatible independent policy snapshot & same $M$ and view constraints & changing base model/tokenizer with the swap \\
$\tfDualOpRollback_{\Phi}$ & prior policy checkpoint plus lineage & unchanged $M$ & rewriting history or memory to ``match'' \\
$\tfDualOpReset_M$ & exact registered memory baseline & same $\Phi$ & clearing only an index while rows remain \\
$\tfDualOpSwap_M$ & compatible independent full snapshot & same $\Phi$ & importing rows without tenant/provenance checks \\
$\tfDualOpRollback_M$ & full row or full snapshot & unchanged $\Phi$ & fieldwise partial undo or silent policy restoration \\
\bottomrule
\end{tabularx}
\caption{Carrier-specific recovery semantics.}
\label{tf:dual:tab:recovery-semantics}
\end{table}

\subsection{Lineage and retention}

Let $L_X=(V_X,E_X)$ be the directed acyclic lineage graph for carrier $X$. Every event edge records predecessor, successor, operation, actor, timestamp/order, input hashes, authorization, and outcome. A rollback points to an earlier content snapshot but creates a new vertex. A deletion/tombstone event in $L_M$ follows the memory retention contract; it does not delete a policy receipt. A policy reset in $L_{\Phi}$ cannot shorten an authoritative row's retention lifetime.

\begin{tfDualProposition}[No silent history erasure]
\label{tf:dual:prop:history}
If lineage vertices and edge receipts are append-only and content-addressed, then a restore operation cannot make its existence observationally identical to an uninterrupted continuation for an auditor who reads the ledger, even when restored content equals an earlier snapshot.
\end{tfDualProposition}

This is an auditability statement under append-only storage, not a guarantee against a privileged ledger compromise.

\tfDualProvedStatus
\begin{tfDualProposition}[Fresh joint pin after restoration]
\label{tf:dual:prop:post-reset-pin}
Under Assumptions~\ref{tf:dual:ass:local-restore} and \ref{tf:dual:ass:interface-soundness}, every reset, swap, or rollback of either carrier invalidates the preceding joint pin and coordination manifest. Before the next joint read, the coordinator must verify a fresh witness over the new content hash, lineage, version, schema, reader/decoder versions, tenant/ACL, validity, and lifetime fields, and issue a fresh manifest receipt. Equality of a recycled nominal version is insufficient. Thus a carrier-local restoration neither resets the other carrier nor permits an ABA substitution to reuse an old joint authorization.
\end{tfDualProposition}

%% file: theory_family/modules/dual_state/sections/07_compression.tex
\section{The Compress-All Comparator}
\label{tf:dual:sec:compression}

\subsection{Why a comparator needs assumptions}

Recurrent and test-time-learning systems often compress observations into a bounded numerical state \cite{compressive2019,mamba2023,tf:dual:sun2025ttt,titans2025}. Compression is not inherently defective. The relevant question is whether indiscriminately mixing a stream into one trainable carrier causes more harmful cross-event influence than a typed, gated design for a declared loss.

Let observations be $X_1,\ldots,X_T$, with informative indices $I$ and contaminants $C$. A projected compress-all update is
\begin{equation}
\Phi_{t+1}=\tfDualProj_{\tfDualCalP}\{\Phi_t-\eta_t g(\Phi_t,X_t)\}.
\label{tf:dual:eq:compress-all}
\end{equation}
Let $\Phi_t^{I}$ be the coupled trajectory that replaces $g(\Phi_t,X_t)$ by zero for $t\in C$ while sharing initialization and informative inputs.

\begin{tfDualAssumption}[Conditional update sensitivity]
\label{tf:dual:ass:update-sensitivity}
Projection is nonexpansive; for every $x$, $g(\cdot,x)$ is $L_g$-Lipschitz; and contaminant updates satisfy $\lVert g(\Phi_t^{I},X_t)\rVert\le B_t$ for $t\in C$. The downstream loss $\ell(\Phi)$ is $L_\ell$-Lipschitz on $\tfDualCalP$.
\end{tfDualAssumption}

\tfDualProvedStatus
\begin{tfDualTheorem}[Conditional contamination bound]
\label{tf:dual:thm:contamination}
Under Assumption~\ref{tf:dual:ass:update-sensitivity},
\begin{align}
\lVert\Phi_T-\Phi_T^{I}\rVert
&\le \sum_{t\in C}\eta_t B_t
\prod_{u=t+1}^{T-1}(1+\eta_uL_g),
\label{tf:dual:eq:contam-state}\\
|\ell(\Phi_T)-\ell(\Phi_T^{I})|
&\le L_\ell\sum_{t\in C}\eta_t B_t
\prod_{u=t+1}^{T-1}(1+\eta_uL_g).
\label{tf:dual:eq:contam-loss}
\end{align}
The bound is one-sided only if an additional sign condition is supplied; by itself it bounds magnitude, not harm.
\end{tfDualTheorem}

The product exposes amplification by later update sensitivity. It is not a measured contamination rate and does not compare hardware efficiency.

\subsection{When compression is optimal}

\begin{tfDualCounterexample}[Sufficient-statistic optimum]
\label{tf:dual:cex:sufficient}
Let $X_t\mid\mu\sim\mathcal N(\mu,\sigma^2)$ independently with a Gaussian prior on the data-generating mean $\mu$, and let the next prediction minimize squared error. The posterior has a two-coordinate sufficient statistic: precision and precision-weighted mean. Updating that statistic with every observation is Bayes-optimal in ideal real arithmetic. A second store carrying no additional information cannot reduce its Bayes prediction risk. This is a \emph{fixed-dimensional}, not a fixed-bit, example: exact real values and unbounded observation counts do not satisfy the finite serialized-capacity contract.

A finite-bit version fixes a horizon $T<\infty$, binary observations, and a fixed rational Beta prior. Store the success count and the number observed, each in $\lceil\log_2(T+1)\rceil$ bits. The exact posterior-predictive rational value is a deterministic function of these counts and the fixed prior; its Bayes squared-error risk is achieved without an extra memory store. A redundant second carrier cannot lower task risk, and any strictly positive additional resource charge makes its total risk strictly larger. The horizon and cost restrictions are essential, not measured properties of a language model.
\end{tfDualCounterexample}

\begin{tfDualCounterexample}[Contamination can help]
If indices labeled ``contaminants'' are actually informative under the target distribution, suppressing their updates increases risk. The sign of Equation~\eqref{tf:dual:eq:contam-loss} cannot be inferred from a taxonomy chosen after observing outcomes.
\end{tfDualCounterexample}

\begin{tfDualProposition}[No universal dominance]
\label{tf:dual:prop:no-dominance}
Without restrictions on the data law, state capacity, update class, downstream loss, and resource cost, neither a dual-state architecture nor compress-all recurrence uniformly dominates the other.
\end{tfDualProposition}

Thus this paper makes no universal attack on recurrent models, state-space models, TTT layers, or neural long-term memory. Its claim is narrower: if a system asserts authoritative semantics and policy adaptation as distinct mechanisms, then it must expose the distinctions and controls required here.

\subsection{Typed gating as a conditional intervention}

Let $A_t\in\{0,1\}$ be a preregistered gate deciding whether event $t$ contributes to $\Phi$. The gated path uses $A_tg(\Phi_t,X_t)$. If $A_t$ depends on hidden future outcomes, comparisons are confounded. If it depends on causal features and is randomized or otherwise identified, its contribution can be studied. This definition is not predictive memory admission: $A_t$ governs a policy update, not a complete authoritative row commit.

\tfDualPlannedStatus\quad A future comparator must register the observation taxonomy, target law, update budget, parameter count, context budget, evaluation loss, and all system costs. Calling one carrier ``compressed'' and another ``memory'' does not equalize capacity or information.

%% file: theory_family/modules/dual_state/sections/08_interaction_bounds.tex
\section{Policy Drift and Dual-State Interaction Bounds}
\label{tf:dual:sec:bounds}

\subsection{Bounded policy drift}

Consider $K$ authorized policy updates
\begin{equation}
\Phi_{k+1}=\tfDualProj_{\tfDualCalP}(\Phi_k-\eta_k \widehat g_k),
\qquad \lVert\widehat g_k\rVert\le G_k.
\label{tf:dual:eq:policy-update}
\end{equation}

\tfDualProvedStatus
\begin{tfDualProposition}[Path-length bound]
\label{tf:dual:prop:path-length}
If projection fixes every $\Phi_k\in\tfDualCalP$ and is nonexpansive, then
\begin{equation}
\lVert\Phi_K-\Phi_0\rVert\le\sum_{k=0}^{K-1}\eta_kG_k.
\label{tf:dual:eq:path-length}
\end{equation}
If the output distribution satisfies
$\tfDualTV(p_{\Phi}(\cdot\mid w,m),p_{\Phi'}(\cdot\mid w,m))
\le L_p\lVert\Phi-\Phi'\rVert$, its drift is at most
$L_p\sum_k\eta_kG_k$ at fixed workspace and memory.
\end{tfDualProposition}

The result controls displacement, not improvement, stability to distribution shift, or safety. A zero gradient also satisfies the bound and learns nothing.

\subsection{Interaction-sensitive error}

Let the deployed predictor be $h(\Phi,M,W)$. Fix reference states $(\Phi_0,M_0)$ and define perturbations $\delta_{\Phi}=d_{\Phi}(\Phi,\Phi_0)$ and $\delta_M=d_M(M,M_0)$. Suppose a scalar loss obeys
\begin{equation}
|\ell(\Phi,M)-\ell(\Phi_0,M_0)|
\le L_{\Phi}\delta_{\Phi}+L_M\delta_M+L_{\Phi M}\delta_{\Phi}\delta_M.
\label{tf:dual:eq:interaction-lipschitz}
\end{equation}
The cross term permits one carrier's displacement to amplify sensitivity to the other. It does not imply complementarity; the bound is unsigned.

\begin{tfDualAssumption}[Local mixed sensitivity]
\label{tf:dual:ass:mixed}
Equation~\eqref{tf:dual:eq:interaction-lipschitz} holds on the registered intervention region, and the state metrics, reference states, and constants are fixed before evaluation.
\end{tfDualAssumption}

\tfDualProvedStatus
\begin{tfDualTheorem}[Dual-state perturbation envelope]
\label{tf:dual:thm:interaction-envelope}
Under Assumption~\ref{tf:dual:ass:mixed}, if interventions guarantee
$\delta_{\Phi}\le\epsilon_{\Phi}$ and $\delta_M\le\epsilon_M$, then every unit's absolute loss displacement is bounded by
\begin{equation}
B_{\Phi M}=L_{\Phi}\epsilon_{\Phi}+L_M\epsilon_M
+L_{\Phi M}\epsilon_{\Phi}\epsilon_M.
\end{equation}
For a bounded loss, the same envelope bounds the absolute population mean displacement. No sign or causal decomposition follows without factorial interventions.
\end{tfDualTheorem}

\subsection{Four interaction regimes}

For losses, define unit-level main effects
$\tau_{\Phi}=Y_{10}-Y_{00}$,
$\tau_M=Y_{01}-Y_{00}$, and interaction
$\tau_{\Phi M}=Y_{11}-Y_{10}-Y_{01}+Y_{00}$.

\begin{figure}[H]
\centering
\includegraphics[width=\linewidth]{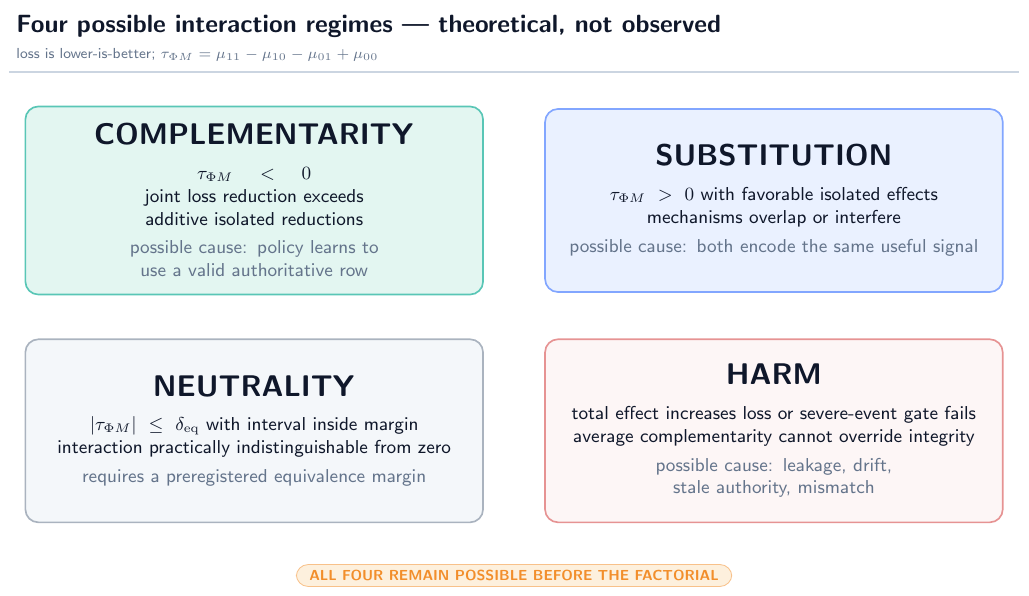}
\caption{\textbf{Theoretical/planned interaction regimes (\tfDualDefinitionStatus{} / \tfDualPlannedStatus{}).} Complementarity, substitution, neutrality, and harm are signs of registered contrasts, not observed dual-state results.}
\label{tf:dual:fig:regimes}
\end{figure}

\begin{itemize}
  \item \emph{Complementarity}: $\tfDualE[\tau_{\Phi M}]<0$; the joint loss reduction exceeds the sum of isolated reductions.
  \item \emph{Substitution}: isolated effects improve loss but $\tfDualE[\tau_{\Phi M}]>0$; the mechanisms overlap or interfere.
  \item \emph{Neutrality}: the interaction is practically indistinguishable from zero under a preregistered equivalence margin.
  \item \emph{Harm}: at least one registered total effect increases loss or violates a severe-event constraint, regardless of average interaction.
\end{itemize}

\begin{tfDualCounterexample}[Average complementarity hides a severe subgroup]
Nine units improve jointly by one loss unit and one protected subgroup unit worsens by eight. The mean joint effect is favorable, but a preregistered severe-event or subgroup gate can still fail. Average interaction is not a safety certificate.
\end{tfDualCounterexample}

\tfDualPrescriptionStatus\quad The future protocol must report all four cell means, main effects, interaction, uncertainty, required subgroups, severe outcomes, and total cost. It may not display only the best arm.

%% file: theory_family/modules/dual_state/sections/09_factorial_identification.tex
\section{The Two-by-Two Policy-by-Memory Factorial}
\label{tf:dual:sec:factorial}

\subsection{Assignments, states, and potential outcomes}

For experimental unit $i$ (a problem-group, not an individual attempt), let
$Z_i^{\Phi},Z_i^M\in\{0,1\}$ be independently randomized assignments. Assignment $1$ means the registered updated carrier is made available; assignment $0$ means the registered baseline/fresh carrier. Let $D_i^{\Phi},D_i^M$ be realized compliance indicators and $Y_i(p,m)$ the loss that would be observed under realized carrier interventions $(p,m)$. This potential-outcome notation follows the distinction between assignments and outcomes \cite{tf:pma:rubin1974}.

\tfDualDefinitionStatus
\begin{tfDualDefinition}[Factorial estimands]
For $p,m\in\{0,1\}$ define $\mu_{pm}=\tfDualE[Y_i(p,m)]$ and
\begin{align}
\tau_{\Phi\mid m}&=\mu_{1m}-\mu_{0m},\\
\tau_{M\mid p}&=\mu_{p1}-\mu_{p0},\\
\tau_{\Phi M}&=\mu_{11}-\mu_{10}-\mu_{01}+\mu_{00},\\
\bar\tau_{\Phi}&=\tfrac12(\tau_{\Phi\mid0}+\tau_{\Phi\mid1}),\qquad
\bar\tau_M=\tfrac12(\tau_{M\mid0}+\tau_{M\mid1}).
\end{align}
Negative effects improve a loss endpoint. Every endpoint and materiality margin is preregistered.
\end{tfDualDefinition}

\begin{figure}[H]
\centering
\includegraphics[width=\linewidth]{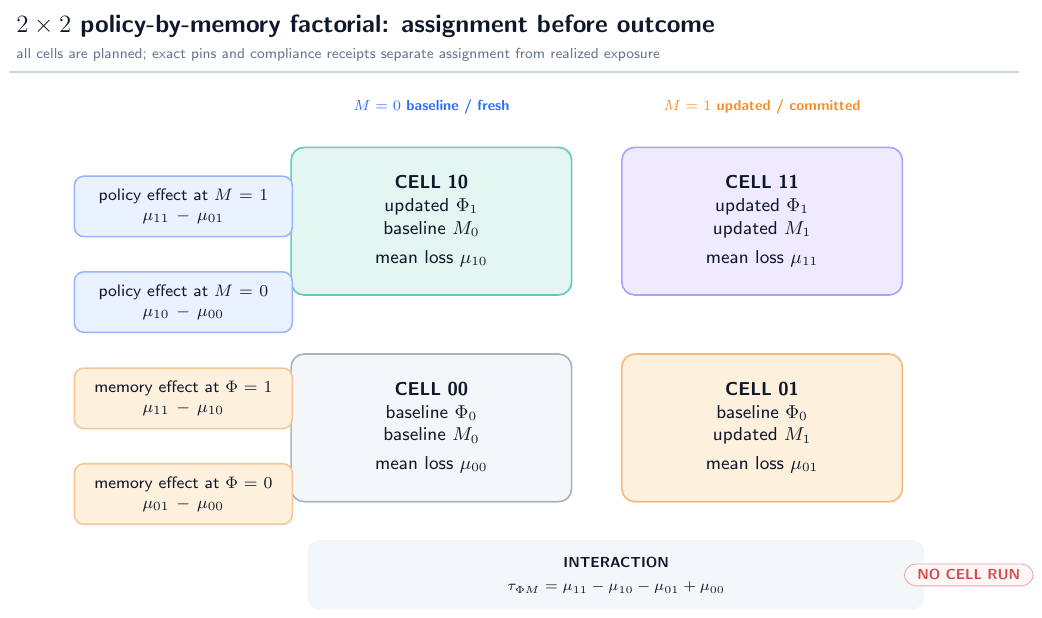}
\caption{\textbf{Independent $2\times2$ interventions and contrasts (\tfDualPlannedStatus{}).} Each cell pins its assigned carrier versions while holding base model, visible workspace, sampler contract, and resource budget to the registered design. No cell has been run.}
\label{tf:dual:fig:factorial}
\end{figure}

\begin{tfDualAssumption}[Factorial identification]
\label{tf:dual:ass:factorial}
Units are assigned with known positive probability to all four cells; assignments precede outcome generation; consistency and no interference hold at the registered group boundary; carrier versions are pinned before the evaluated attempt; attrition and endpoint construction do not depend on hidden potential outcomes; and the analysis either has perfect compliance or distinguishes assignment from receipt-verified exposure.
\end{tfDualAssumption}

\tfDualProvedStatus
\begin{tfDualTheorem}[Identification under randomized compliant intervention]
\label{tf:dual:thm:factorial}
Under Assumption~\ref{tf:dual:ass:factorial} with perfect compliance,
\begin{equation}
\tfDualE[Y_i^{\mathrm{obs}}\mid Z_i^{\Phi}=p,Z_i^M=m]=\mu_{pm}.
\end{equation}
Therefore cell-mean differences identify both conditional main effects, both marginal main effects, and $\tau_{\Phi M}$. With known unequal assignment probabilities, Horvitz--Thompson cell means are unbiased under the same premises.
\end{tfDualTheorem}

The theorem does not establish exclusion restrictions under noncompliance, adequate power, metric validity, or any beneficial sign.

\subsection{Fresh, reset, swapped, and frozen variants}

The primary binary factor must be defined exactly. A useful staged design is:
\begin{enumerate}
  \item \emph{availability factorial}: updated versus registered fresh baseline for each carrier;
  \item \emph{content-specificity controls}: problem-matched updated carrier versus compatible swapped carrier;
  \item \emph{attempt-budget control}: mutable versus frozen policy with identical extra attempts;
  \item \emph{recovery controls}: independently roll back one carrier after a registered checkpoint and test restoration of its isolated effect;
  \item \emph{order control}: randomize admissible update/commit order where both are enabled.
\end{enumerate}
Each stage uses new held-out groups or a registered repeated-measures design that models carryover. A carrier called ``reset'' without a content hash, lineage, optimizer state (for $\Phi$), and cache invalidation receipt is noncompliant.

\subsection{Noncompliance and exclusions}

If $D\neq Z$, the intent-to-treat contrast remains identified by random assignment, but it estimates assignment, not carrier exposure. A complier effect would require additional monotonicity and exclusion assumptions that are especially fragile when pin failures change latency or fallback behavior. This paper does not assume those conditions.

\begin{tfDualProposition}[Receipt-stratified descriptive effect is not automatically causal]
\label{tf:dual:prop:receipt-selection}
Conditioning on successful pin or commit after randomization can select on post-assignment variables. Unless compliance is guaranteed by design or modeled under explicit assumptions, the difference between receipt-positive units is descriptive rather than the randomized cell effect.
\end{tfDualProposition}

\tfDualPlannedStatus\quad The protocol in Appendix~\ref{tf:dual:app:factorial-protocol} uses intent-to-treat as primary, reports compliance separately, and stops rather than silently dropping failed pins, validator rejects, timeouts, or exact \tfDualNull{} outcomes.

%% file: theory_family/modules/dual_state/sections/10_observational_limits.tex
\section{Observational Non-Identification}
\label{tf:dual:sec:nonidentification}

\subsection{Output traces do not name the changed carrier}

Let the observable trace be $O=(q,Z_0,F_0,Z_1)$, omitting internal state and receipts.  A finite observable law induces a later-output kernel $K(z_1\mid q,z_0,f_0)$.  Write $\mathfrak R_{M\leftarrow\Phi}(K;\tfDualCalI)$ for the set of memory-update realizations that (i) use a registered typed reader/decoder interface $\tfDualCalI$, (ii) can encode $K$ within declared capacity, (iii) admit every required row under the memory authorization, validity, tenant, and ACL contracts, and (iv) preserve the registered causal and randomization law.  Define $\mathfrak R_{\Phi\leftarrow M}(K;\tfDualCalI)$ symmetrically for policy-update realizations.  These sets can be empty.

\tfDualProvedStatus
\begin{tfDualTheorem}[Conditional output-trace equivalence]
\label{tf:dual:thm:trace-equivalence}
For a policy-update law with kernel $K$, a frozen-policy memory-update system inducing the same law exists only if $\mathfrak R_{M\leftarrow\Phi}(K;\tfDualCalI)\ne\varnothing$.  Conversely, a frozen-memory policy-update realization exists only if $\mathfrak R_{\Phi\leftarrow M}(K;\tfDualCalI)\ne\varnothing$.  Whenever either set is nonempty, selecting any member yields output equivalence at $O$; therefore $O$ alone cannot distinguish mechanisms \emph{within that compatible realization class}.  Finiteness of the observable alphabet and capacity to store a table do not by themselves establish either nonemptiness claim.
\end{tfDualTheorem}

The theorem is an assumption-explicit identification result, not an automatic lookup-table argument.  It does not say that two realizations share cost, privileges, interpretability, or semantics.

\begin{tfDualCounterexample}[Finite capacity without legal realization]
\label{tf:dual:cex:no-realization}
Let a finite table for $K$ fit in one row, but let its output signature be absent from the registered memory reader, or let the only row that could store it violate tenant/ACL scope.  Then physical capacity is sufficient while $\mathfrak R_{M\leftarrow\Phi}(K;\tfDualCalI)=\varnothing$.  A frozen policy cannot legally read the table, so output equivalence does not follow.  The symmetric example uses a policy endpoint that cannot represent or authorize the required memory-derived update.
\end{tfDualCounterexample}

\begin{tfDualCorollary}[No interaction from one trajectory]
\label{tf:dual:cor:one-trajectory}
A single dual-state trajectory cannot identify $\tau_{\Phi M}$ because at most one of the four potential outcomes is observed for its experimental unit.
\end{tfDualCorollary}

\subsection{Hidden shared state}

Even carrier hashes are insufficient if a third state changes. Let $G$ be sampler state and $Q$ a cache. Two runs with equal $(\Phi,M)$ but different $G$ can have different outputs. Two runs with equal carrier payloads but a stale cache in $Q$ can read different effective memory. Hence the intervention must pin or log the complete state inventory from Equation~\eqref{tf:dual:eq:full-state}.

\begin{tfDualCounterexample}[Cache-mediated false memory effect]
Assignment changes $M$, but the inference process continues reading a cached old row. The intended memory intervention has no exposure, while a latency-triggered retry changes the sampler. An analysis using assigned memory labels attributes the output difference to $M$. Pin, cache-invalidation, and read receipts would expose the violation.
\end{tfDualCounterexample}

\begin{tfDualCounterexample}[Feedback-dependent stopping]
The updated-policy arm is allowed more attempts only after poor feedback. Comparing final accuracy to a fixed-attempt baseline mixes policy state with an adaptive compute budget. The policy factor must define or condition on a registered stopping rule without post-treatment cherry-picking.
\end{tfDualCounterexample}

\subsection{Swap tests and content specificity}

A successful updated-versus-fresh comparison can still reflect update count, file age, cache warming, or lineage-specific configuration. A compatible swap control tests whether the actual state content matters. Let $S_i^{\Phi}$ be another unit's policy snapshot and $S_i^M$ another unit's memory snapshot, matched on schema, update count, resource budget, and non-content metadata. If the updated state helps but its swapped counterpart helps equally, problem-specific content has not been isolated.

\begin{tfDualAssumption}[Valid swap]
\label{tf:dual:ass:swap}
Swap donors are selected independently of recipient potential outcomes; compatibility fields are identical; no recipient-specific content leaks into the donor; and all non-content execution differences are balanced or fixed.
\end{tfDualAssumption}

\tfDualProvedStatus
\begin{tfDualProposition}[Swap contrast interpretation]
\label{tf:dual:prop:swap-contrast}
Under Assumption~\ref{tf:dual:ass:swap} and randomized assignment, the updated-versus-swapped contrast identifies the effect of recipient-specific carrier content relative to the declared donor distribution, not a universal semantic-content effect.
\end{tfDualProposition}

\subsection{Missing provenance}

If predecessor hashes, carrier tags, or pin receipts are missing, multiple incompatible histories fit the same terminal bytes. No statistical estimator can recover a uniquely identified reset, rollback, or update lineage from terminal state alone without additional structural restrictions.

\tfDualPrescriptionStatus\quad A missing receipt is a protocol failure. It is never converted into ``no update,'' ``successful reset,'' or ``same memory.''

%% file: theory_family/modules/dual_state/sections/11_authorization.tex
\section{Authorization and Information-Flow Isolation}
\label{tf:dual:sec:authorization}

\subsection{Separate writers and admissible inputs}

Policy feedback can contain scalar rewards, verifier outcomes, public error classes, or other registered signals. It must not automatically become authoritative memory content. A memory candidate can contain a canonical fact and provenance but must not be interpreted as a gradient. The two flows can originate from one observed event while remaining separately authorized.

\begin{figure}[H]
\centering
\includegraphics[width=\linewidth]{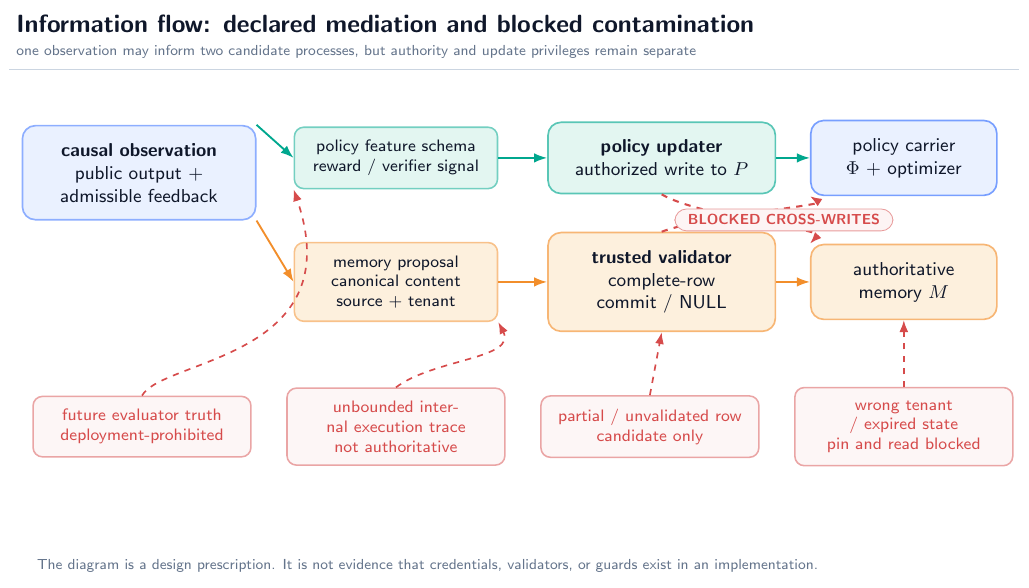}
\caption{\textbf{Contamination pathways and blocked privilege crossings (\tfDualPrescriptionStatus{}).} Teal arrows are declared reads or typed updates. Coral dashed arrows are forbidden: gradients cannot write authoritative rows; unvalidated candidates cannot alter policy; an unbounded internal execution trace is not authority; future evaluator truth cannot enter deployment.}
\label{tf:dual:fig:contamination}
\end{figure}

\begin{tfDualAssumption}[Least-privilege mediation]
\label{tf:dual:ass:least-privilege}
Policy and memory writers execute under distinct credentials. Candidate construction is non-authoritative. Only the trusted memory validator/committer can publish a complete row or exact \tfDualNull{}. Only the policy updater can publish a policy successor. Every cross-carrier feature is explicitly serialized, schema-checked, and logged. Registered policy readers, memory readers, and the decoder are read-only and versioned. A coordinator may verify a compatibility witness and append a signed coordination-manifest receipt, but neither that witness nor the manifest authorizes a write to either carrier or a rollback across carriers.
\end{tfDualAssumption}

\tfDualProvedStatus
\begin{tfDualProposition}[Blocked direct contamination]
\label{tf:dual:prop:blocked}
Under Assumptions~\ref{tf:dual:ass:typed-auth} and \ref{tf:dual:ass:least-privilege}, a compromised ordinary policy-update call cannot directly publish an authoritative memory successor, and a memory candidate cannot directly publish a policy successor. The statement does not cover compromise of both privileged services or a shared trusted computing base.
\end{tfDualProposition}

\subsection{Trusted assembly and exact NULL}

The row contract requires a complete target-conditioned row with canonical and neural views, slot, tenant and ACL fields, monotone version, validity, protection, provenance, rollback pointer, deletion/tombstone semantics, and a consistency receipt. A neural module may propose semantic content, but trusted assembly owns protected fields and validation. If any required field or receipt fails, the authoritative transition is exact \tfDualNull{}: resident bytes and authoritative version do not change, while a rejection receipt is appended outside the resident state.

\tfDualUnvalidatedStatus\quad This paper uses that contract as a formal dependency only. The same-checkpoint semantic interface remains failed/unqualified; no dual-state row was generated, admitted, or read by a model.

\subsection{Feedback-to-memory leakage}

Suppose an evaluator exposes the correct final answer after attempt $r$. Using that answer as policy feedback may be allowed in a registered supervised test-time setting. Writing it into $M$ and scoring attempt $r+1$ on the same answer tests answer caching, not general reasoning or predictive memory. The memory protocol must tag target-revealing feedback and exclude or separately analyze direct answer reuse.

\begin{tfDualCounterexample}[Joint endpoint leakage]
Both $\Phi$ and $M$ receive the ground-truth answer. The joint arm succeeds perfectly on a repeated query. The factorial interaction may appear strongly complementary even though both carriers are merely copying target information. Randomization does not repair an invalid endpoint or post-treatment leakage.
\end{tfDualCounterexample}

\subsection{Tenant, lifetime, and deletion boundaries}

Policy state defaults to the current problem and must be destroyed or archived according to its own scope. Memory rows follow tenant, ACL, validity, protection, retention, tombstone, and verified-deletion rules. A policy snapshot cannot extend a memory row's retention; a memory archive cannot preserve an expired problem-scoped optimizer state by reference. Composite manifests hold identifiers, not ownership of the underlying retention policy.

\begin{table}[H]
\centering
\small
\begin{tabularx}{\linewidth}{@{}p{0.22\linewidth}Y Y@{}}
\toprule
Boundary & Required behavior & Stop condition \\
\midrule
Carrier privilege & distinct credentials and typed write endpoints & any cross-carrier direct write \\
Interface version & registered read-only readers/decoder and exact schema hashes & unregistered reader, decoder, or implicit coercion \\
Coordination manifest & coordinator verifies witness and signs hashes only & treating a pin or manifest receipt as carrier-write or cross-rollback authority \\
Future information & deployment reads causal data only & evaluator truth or future branch enters a deployed update \\
Authority & candidate remains non-authoritative until full validation & partial row, missing protected field, or hash-only equality \\
Tenant/ACL & check at construction, pin, read, commit, archive, and restore & mismatched tenant or unverified policy \\
Lifetime & independently declared expiry and cleanup & one carrier silently inherits the other's lifetime \\
Deletion & carrier-local verified deletion/tombstone receipt & dangling index, cache, archive, or snapshot \\
\bottomrule
\end{tabularx}
\caption{Information-flow and authority gates.}
\label{tf:dual:tab:auth-gates}
\end{table}

%% file: theory_family/modules/dual_state/sections/12_costs.tex
\section{Complete Resource Accounting}
\label{tf:dual:sec:costs}

\subsection{Vector costs before scalarization}

For execution $u$, define
\begin{equation}
\mathbf C(u)=
(N_{\mathrm{tok}},F_{\mathrm{fwd}},F_{\mathrm{bwd}},B_{\mathrm{param}},B_{\mathrm{opt}},
B_{\mathrm{row}},B_{\mathrm{archive}},B_{\mathrm{traffic}},T_{\mathrm{wall}},E,
N_{\mathrm{retry}},N_{\mathrm{receipt}},N_{\mathrm{witness}},N_{\mathrm{sig}},
N_{\mathrm{fallback}}).
\label{tf:dual:eq:cost-vector}
\end{equation}
The entries are tokens, forward and backward work, parameter bytes, optimizer bytes, resident-row bytes, archive/index bytes, memory/network traffic, wall time, energy, retries, receipt/validation operations, compatibility-witness construction or verification, signature/schema checks, and executed fallbacks. Each retained-byte coordinate needs a declared statistic (such as peak simultaneous occupancy), not a sum of repeated observations of the same allocation. Wall time is elapsed makespan. Coordination-manifest hashing and signing are charged rather than treated as a free atomic pair. A scalar cost $c=\lambda^\top\mathbf C$ is reported only with a preregistered nonnegative weight vector $\lambda$ and units.

\begin{figure}[H]
\centering
\includegraphics[width=\linewidth]{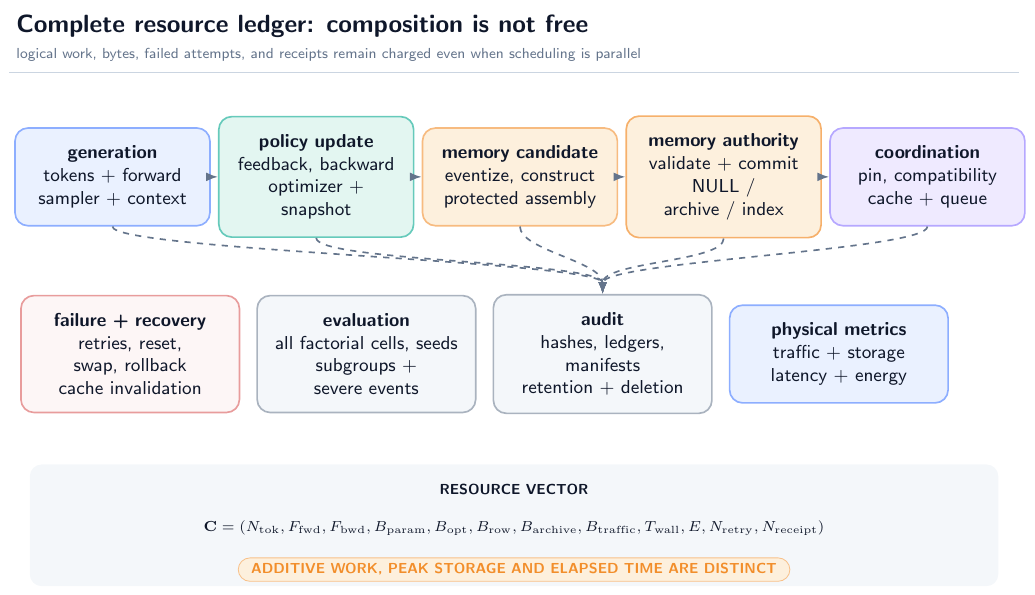}
\caption{\textbf{Training, reasoning-time, memory, recovery, and audit ledgers (\tfDualDefinitionStatus{} / \tfDualPrescriptionStatus{}).} Parallelism may reduce wall time but cannot delete logical work, bytes, validation, or failed attempts from the ledger.}
\label{tf:dual:fig:cost-ledger}
\end{figure}

\subsection{Ledger decomposition}

Let $\tfDualCalE_{\mathrm{gen}}$, $\tfDualCalE_{\Phi}$, $\tfDualCalE_M$, $\tfDualCalE_{\mathrm{coord}}$, and $\tfDualCalE_{\mathrm{recover}}$ be disjoint event sets. Let $\mathbf C_{\mathrm{add}}$ contain only additive counters: tokens, forward/backward work, transferred bytes, operation counts, and energy when a disjoint attribution of measured energy is available. Exact accounting for this subvector gives
\begin{equation}
\mathbf C_{\mathrm{add}}
=\sum_{e\in\tfDualCalE_{\mathrm{gen}}}\mathbf c_e
+\sum_{e\in\tfDualCalE_{\Phi}}\mathbf c_e
+\sum_{e\in\tfDualCalE_M}\mathbf c_e
+\sum_{e\in\tfDualCalE_{\mathrm{coord}}}\mathbf c_e
+\sum_{e\in\tfDualCalE_{\mathrm{recover}}}\mathbf c_e.
\label{tf:dual:eq:ledger-identity}
\end{equation}

\tfDualProvedStatus
\begin{tfDualProposition}[No-free-composition accounting]
\label{tf:dual:prop:cost-identity}
If every physical or logical operation is assigned to exactly one event set and $\mathbf c_e$ contains only additive counters with no double attribution, Equation~\eqref{tf:dual:eq:ledger-identity} is an identity for $\mathbf C_{\mathrm{add}}$. Peak retained bytes and elapsed wall time are measured separately from the allocation and scheduling traces; they are not sums of per-operation peaks or overlapping durations. Dual-state cost is not represented by generation tokens alone: it includes feedback/evaluation, backward and optimizer work, checkpointing, complete candidate construction, protected-field assembly, validation, commit/\tfDualNull{}, archive/index traffic, witness construction and verification, schema/signature checks, coordination-manifest hashing/signing, resets, swaps, rollbacks, cache invalidation, retries, fallback execution, and audit receipts.
\end{tfDualProposition}

\begin{tfDualCorollary}[Parallelism does not erase work]
\label{tf:dual:cor:parallelism}
Batching or concurrent execution may change $T_{\mathrm{wall}}$ and hardware utilization but does not set the corresponding logical operation counts or transferred bytes to zero.
\end{tfDualCorollary}

\subsection{Training and reasoning-time ledgers}

\begin{table}[H]
\centering
\scriptsize
\begin{tabularx}{\linewidth}{@{}p{0.19\linewidth}Y Y@{}}
\toprule
Phase & Mandatory charges & Common hidden omission \\
\midrule
Base/pretraining & data, tokens, forward/backward, optimizer, checkpoints & amortizing it for one arm but not another \\
Meta/policy preparation & task sampling, reward/evaluator, adapters, optimizer state, validation & counting only trainable parameter bytes \\
Reasoning-time policy & exploration attempts, feedback, reward, backward pass, optimizer, snapshots, rollback tests & reporting only final-answer tokens \\
Memory candidate & eventization, target-conditioned construction, canonical/neural views, protected-field assembly & treating the row as already available \\
Memory authority & validation, exact \tfDualNull{} decisions, commit, archive, index, provenance, deletion & counting resident bytes only \\
Coordination & witness construction/verification, schema/signature checks, coordination-manifest hashing/signing, version pin, cache invalidation, queueing, fallback execution & assuming a free atomic pair \\
Evaluation & all four factorial cells, negative controls, seeds, subgroups, failures, uncertainty & reporting the best cell alone \\
\bottomrule
\end{tabularx}
\caption{Complete logical ledger for the proposed dual-state system. This part measures no such system cost; the separate readout studies report their own limited runtime counters.}
\label{tf:dual:tab:cost-phases}
\end{table}

\subsection{Comparator normalization}

A fair comparator holds fixed or explicitly prices base model, precision, maximum visible context, total generated tokens, number of attempts, policy parameter count, memory capacity, evaluator access, update budget, wall-clock cap, and hardware. If architectures cannot be matched on all dimensions, the result is a Pareto comparison over quality, severe failures, latency, energy, memory, and traffic, not a single ``efficiency'' number.

\begin{tfDualAssumption}[Cost observability]
\label{tf:dual:ass:cost}
Every event reports units and measurement method; failed and aborted events are retained; cached reuse is proven by content identity and dependency validity; and energy or FLOP claims are omitted when the required measurement is absent.
\end{tfDualAssumption}

\tfDualPlannedStatus\quad No dual-state cost vector has been measured. Exact ledger equations are protocol accounting, not evidence of efficiency or a hardware crossover.

%% file: theory_family/modules/dual_state/sections/13_failures_recovery.tex
\section{Failure Modes and Recovery}
\label{tf:dual:sec:failures}

\subsection{Carrier-local failures}

\begin{table}[H]
\centering
\scriptsize
\begin{tabularx}{\linewidth}{@{}p{0.16\linewidth}p{0.20\linewidth}Y Y@{}}
\toprule
Carrier & Failure & Observable symptom & Required response \\
\midrule
$\Phi$ & divergent or nonfinite update & parameter/optimizer validation fails & reject successor; retain predecessor; record failed update \\
$\Phi$ & optimizer mismatch & weights restore but moments/step do not & quarantine; restore complete policy snapshot or stop \\
$\Phi$ & reward leakage/hacking & success depends on prohibited evaluator information & invalidate affected stage and descendants \\
$\Phi$ & excessive drift & registered norm, KL, or severe-output gate fails & freeze and independently roll back policy only \\
$M$ & partial or malformed row & protected field, width, canonical/neural consistency, or hash check fails & exact \tfDualNull{}; no authoritative version change \\
$M$ & stale/unauthorized authority & tenant, ACL, validity, or version check fails & block read/commit; rollback or tombstone under memory policy \\
$M$ & archive/index divergence & resident, index, and archive receipts disagree & quarantine memory view; repair and revalidate \\
$M$ & deletion failure & row survives in cache/index/archive & stop; complete verified deletion before reuse \\
\bottomrule
\end{tabularx}
\caption{Carrier-local failures preserve separate recovery domains.}
\label{tf:dual:tab:local-failures}
\end{table}

\subsection{Composition failures}

\begin{table}[H]
\centering
\scriptsize
\begin{tabularx}{\linewidth}{@{}p{0.22\linewidth}Y Y@{}}
\toprule
Composition failure & Why it breaks inference & Recovery/analysis consequence \\
\midrule
Unpinned mixed read & tokens within one attempt can see different carrier versions & invalidate attempt; no post hoc selection of a convenient view \\
Compatibility mismatch & policy features, tokenizer, schema, or base model disagree & block pin or route to registered fallback; do not coerce bytes \\
Hidden joint reset & one intervention changes both carriers & factorial attribution fails; rerun from independently verified baselines \\
Undeclared event order & policy and memory decisions may not commute & effect is undefined for the registered treatment \\
Shared credential & a compromise can cross both write domains & least-privilege proposition does not apply \\
Receipt loss & predecessor, exposure, or compliance is ambiguous & mark missing; do not infer no-op or successful rollback \\
Feedback-to-answer leakage & later success can be direct target copying & invalidate reasoning endpoint or report separate cache task \\
Adaptive resource imbalance & one cell receives more attempts or compute & endpoint mixes mechanism and budget \\
\bottomrule
\end{tabularx}
\caption{Composition failures and their causal consequences.}
\label{tf:dual:tab:composition-failures}
\end{table}

\subsection{Recovery invariants}

\tfDualPrescriptionStatus\quad A recovery sequence must satisfy:
\begin{enumerate}
  \item stop new pins that could read a suspect carrier;
  \item preserve both ledgers and all failed events;
  \item identify the smallest carrier-local recovery target;
  \item restore that target with a new successor receipt;
  \item invalidate carrier-dependent caches without changing the other carrier;
  \item rerun compatibility checks and a registered canary;
  \item resume only if all predeclared gates pass.
\end{enumerate}

\begin{tfDualProposition}[Recovery locality]
\label{tf:dual:prop:recovery-locality}
Under carrier-local restoration and a dependency-complete cache graph, a policy-only failure can be recovered without changing authoritative memory content, and a memory-only failure can be recovered without changing policy content. Cache eviction and pin refusal may affect availability but not carrier identity.
\end{tfDualProposition}

\begin{tfDualCounterexample}[Rollback illusion from an incomplete snapshot]
A policy rollback restores adapter weights but not optimizer moments. The next update diverges. Terminal weights initially match the prior checkpoint, yet the restored state is not the prior policy carrier. A weight-file hash alone is therefore not a rollback receipt.
\end{tfDualCounterexample}

\begin{tfDualCounterexample}[Memory fieldwise undo]
A system restores a row's semantic payload but leaves the newer version, ACL, and provenance. The row is neither the old complete version nor a valid new one. This violates complete-row rollback even if a query happens to return the older fact.
\end{tfDualCounterexample}

\subsection{Severe-event stop rules}

Average task loss cannot override a severe integrity event. The protocol stops the affected stage on any cross-tenant read/write, unverified deletion, direct privilege crossing, future-information leak, silent cross-carrier restoration, lost mandatory receipt, or corruption of the frozen evaluation boundary. A thresholded task or cost gate may also stop for futility. These rules are design prescriptions, not evidence that a future implementation is safe.

%% file: theory_family/modules/dual_state/sections/14_protocol_obligations.tex
\section{Staged Falsification Protocol and Obligations}
\label{tf:dual:sec:protocol}

\subsection{The prerequisite gate is first}

The present research chain stops before a dual-state experiment. The same-checkpoint semantic interface is failed/unqualified, and predictive admission is closed. A valid future program must reopen no downstream stage until its prerequisite has independently qualified.

\begin{figure}[H]
\centering
\includegraphics[width=\linewidth]{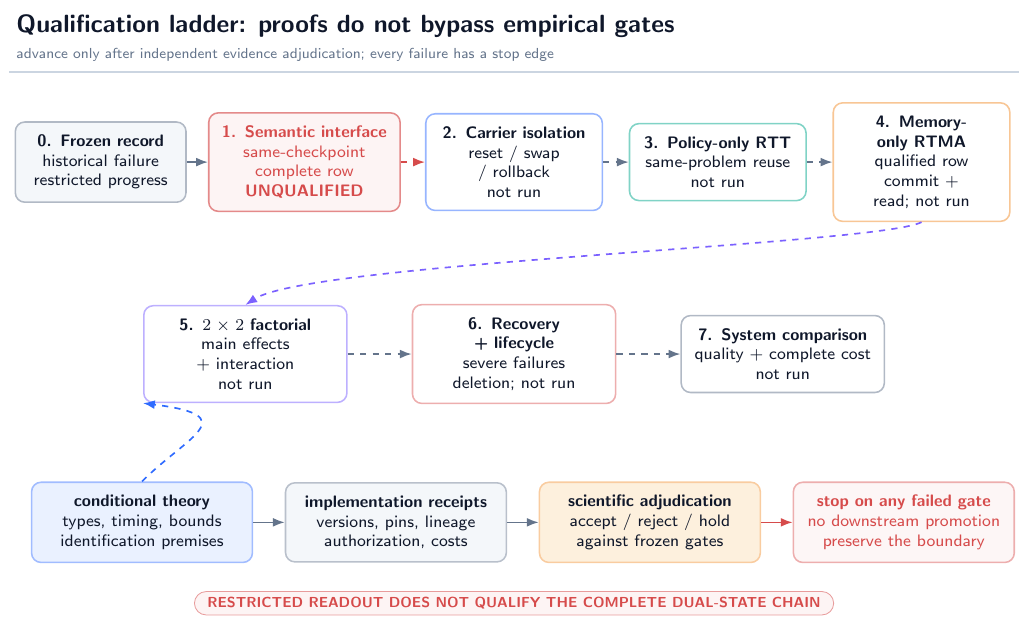}
\caption{\textbf{Qualification ladder, stop edges, and proof dependencies (\tfDualPlannedStatus{}).} The coral semantic-interface stop is current. Theory boxes state conditional obligations; they do not bypass an empirical gate.}
\label{tf:dual:fig:ladder}
\end{figure}

\begin{tfDualProtocol}[Dual-state qualification ladder]
\label{tf:dual:prot:ladder}
Every stage uses immutable registration, fresh groups, required seeds, independent receipts, and explicit scientific adjudication.
\begin{enumerate}
  \item \textbf{Semantic carrier qualification.} Establish a same-checkpoint complete typed row interface under its registered fit, held-out composition, query, preservation, mapping, and integrity gates. Current status: \tfDualFail{} / unqualified.
  \item \textbf{Carrier isolation.} Demonstrate independent snapshot, fresh/reset, swap, freeze, rollback, cache invalidation, and ledger behavior for $\Phi$ and $M$. Every restoration invalidates the old joint pin and compatibility witness; a fresh witness must bind both successor lineages before reuse. Stop on any silent cross-carrier mutation or ABA-style pin reuse.
  \item \textbf{Policy-only strict RTT.} With $M$ frozen, show feedback update, same-unresolved-problem later reuse, and a randomized policy intervention under fixed attempt/resource budgets. This is not a memory result.
  \item \textbf{Memory-only \tfDualRTMA{}.} With $\Phi$ frozen, show a non-NULL complete-row successor, predecessor/successor lineage, and later verified reuse. Report exact \tfDualNull{} separately as a no-op control with bit-identical authoritative memory and no successor. This is not a policy-learning result and requires a qualified memory interface.
  \item \textbf{$2\times2$ factorial.} Randomize both carrier interventions, report all cell means, main effects, interaction, compliance, order, subgroup, severe-event, and cost results.
  \item \textbf{Recovery and lifecycle.} Exercise carrier-local rollbacks, expiration, deletion, repeated updates/commits, and induced failure cases. Stop on unresolved contamination or receipt gaps.
  \item \textbf{System comparison.} Only after all prior gates, compare against registered context, recurrent/compress-all, policy-only, memory-only, and static baselines on a Pareto ledger.
\end{enumerate}
\end{tfDualProtocol}

\subsection{Primary endpoints and analysis}

The primary scientific endpoint must be a task loss that does not reward direct answer caching. Co-primary integrity endpoints include semantic row exactness, preservation, policy drift, severe failures, and compliance. Resource endpoints retain the vector in Equation~\eqref{tf:dual:eq:cost-vector}. The unit of randomization is the independent problem group; all attempts and branches from a group remain in one split. Final analysis reports uncertainty and multiplicity for the registered primary contrasts. Appendix~\ref{tf:dual:app:factorial-protocol} gives an executable design specification without inventing sample sizes or effect magnitudes.

\subsection{Theorem-to-receipt obligations}

\begin{table}[H]
\centering
\scriptsize
\begin{tabularx}{\linewidth}{@{}p{0.22\linewidth}p{0.27\linewidth}Y@{}}
\toprule
Result & Minimum implementation receipt & Failure consequence \\
\midrule
Type non-conflation & carrier tags, distinct credentials/endpoints, authorization log & type-safety proposition unusable \\
Typed coupling interface & registered $R_{\Phi}$, $R_M$, decoder, schemas, interface version, and negative cast tests & no well-typed joint read is established \\
Compatibility witness & signed valid witness binding both hashes, versions, lineages, lifetimes, ACLs, schemas, reader/decoder versions, and fallback & joint view must refuse or fall back \\
Causal timeline & event order, information-set schema, exact pinned predecessor view & temporal claim invalid \\
Conditional commutation & invariant inputs/decisions under both orders, compatibility receipts & order must be modeled as treatment \\
Coupled-transition closure & both carrier-local successor receipts, verified witness, and signed coordination-manifest receipt & no atomic pair or coupled-update claim \\
Strict RTT classification & unresolved marker, feedback, policy successor, later pinned descendant, intervention & call it retry/adaptation, not strict RTT \\
Independent rollback & complete carrier snapshot, fresh successor lineage, invalidated old joint pin/witness, cache dependency graph, unchanged other-carrier hash & recovery and ABA-resistance claims invalid \\
Contamination bound & step sizes, gradient bounds, Lipschitz region, fixed contaminant definition & bound cannot be cited \\
Factorial identification & randomized assignments, positive support, pins, compliance, group isolation, attrition log & contrasts descriptive only \\
Output non-identification & nonempty legal realization class under registered interface, compatibility, authorization, lifetime, and causal constraints & no carrier-indistinguishability conclusion \\
Cost identity & event-complete units including witness verification, signatures, coordination-manifest receipts, fallback, failed work, cache proof, and measurement method & efficiency claim prohibited \\
\bottomrule
\end{tabularx}
\caption{Proofs become system statements only with their premises and receipts.}
\label{tf:dual:tab:obligations}
\end{table}

\subsection{Decision grammar}

Each stage ends in one of four durable scientific statuses: \tfDualPass{} under every registered gate; \tfDualFail{} under at least one decisive gate; \tfDualHold{} for incomplete but recoverable evidence; or \textsf{INVALID} for protocol corruption. Editorial completion and a clean PDF build have no bearing on those scientific statuses.

\tfDualPlannedStatus\quad Protocol~\ref{tf:dual:prot:ladder} remains unexecuted for the complete dual-state intervention. No qualified joint checkpoint, strict policy-update witness, dual-state factorial unit, admission-oracle margin, or learned admission decision exists in the evidence cutoff. The separate component studies do execute memory changes and offline adapter updates; those are not silently relabeled as this protocol.

%% file: theory_family/modules/dual_state/sections/15_related_limits.tex
\section{Related Work, Limitations, and Conclusion}
\label{tf:dual:sec:related}

\subsection{Test-time updates and numerical state}

Test-time training updates model parameters on test inputs under a self-supervised objective \cite{tf:rtt:sun2020ttt}. TTT layers treat an internal model as recurrent hidden state updated by learning at test time \cite{tf:dual:sun2025ttt}. LoRA provides a parameter-efficient adapter representation for numerical adaptation \cite{lora2021}; it does not itself define when an update is reasoning-time, which feedback is valid, or how rollback should interact with authoritative memory. Recent LLM work performs test-time reinforcement learning from model-derived feedback \cite{tf:dual:zuo2025ttrl}; Policy of Thoughts and StarOR describe transient within-instance policy adaptation for reasoning or optimization \cite{tf:rtt:jiao2026pot,tf:rtt:li2026staror}. These papers motivate explicit policy-state timing. Their empirical findings are not transferred to PDSA.

\subsection{Recurrent and neural memory}

Compressive Transformers compress older activations for long-range sequence modeling \cite{compressive2019}; Mamba uses selective state-space recurrence \cite{mamba2023}; TTT layers make the recurrent state a learned model \cite{tf:dual:sun2025ttt}; and Titans proposes neural long-term memory updated at test time \cite{titans2025}. These are numerical or architectural memory mechanisms under their own contracts. Our use of ``authoritative memory'' is narrower: a bounded map of complete typed, versioned, provenance-bearing rows with trusted assembly and exact commit/\tfDualNull{}. Completed-segment consolidation separately makes the segment boundary and bidirectional evidence bundle explicit, but supplies no empirical result here. The distinction is semantic and operational, not a claim of superior accuracy or efficiency.

\subsection{Causal factorial attribution}

The potential-outcome framework separates assignment from outcomes and makes interaction a contrast over unobserved potential responses \cite{tf:pma:rubin1974}. Our contribution is to specialize that discipline to independently pinned policy and authoritative-memory carriers, with explicit noncompliance, swap, reset, rollback, sampler, cache, and ledger controls. Randomization can identify registered contrasts; it cannot validate a leaked endpoint, an unqualified semantic interface, or a missing receipt.

\subsection{Limitations}

The scope and limitations of this theoretical part are as follows.
\begin{enumerate}
  \item It contains no new dual-state measurement. Sections~12--13 of the complete report contain restricted readout measurements, but those do not establish the full authoritative-row contract, predictive admission, or policy-by-memory joint benefit defined here.
  \item The authoritative semantic interface required by the composition is currently failed/unqualified under the frozen evidence record.
  \item The formal results rely on measurability, typed registered readers/decoders, authorization, witness verification, lifetime validity, Lipschitz, capacity, randomization, compliance, and ledger premises that no implementation has established.
  \item The output-trace theorem is conditional on a nonempty class of legally realizable, interface-compatible alternatives. Finite capacity alone does not establish that class; the theorem is an identification result, not a realistic cost equivalence.
  \item The contamination bound is worst-case and unsigned. It neither estimates real contamination nor proves that gating helps.
  \item The $2\times2$ design does not solve endpoint validity, interference outside the chosen group, noncompliance, low power, or multiplicity by itself.
  \item Independent operation algebras reduce semantic coupling but can increase coordination, storage, validation, and recovery cost.
  \item We do not claim priority over all dual-state, adaptive-memory, recurrent, or test-time-learning systems.
\end{enumerate}

\subsection{Conclusion}

The central design rule is simple but demanding: policy state $\Phi$ and authoritative memory $M$ may cooperate, yet they must never become indistinguishable. They require separate types, writers, readers, scopes, versions, lineages, resets, swaps, rollbacks, ledgers, costs, and causal interventions. A pinned read view coordinates two independently owned states through a registered typed interface and a verified compatibility witness; it does not merge them or authorize either write. After restoration, the old joint pin and witness are invalid even when payload bytes recur. Strict RTT requires feedback-driven policy update and later same-unresolved-problem reuse. Memory mutation alone is \tfDualRTMA{}. Output improvement alone certifies neither.

The resulting theory offers conditional guarantees and explicit failure tests, not a system result. Current evidence still stops at the failed/unqualified semantic interface. Any future dual-state claim must first clear that gate, then survive carrier-isolation controls, policy-only and memory-only qualification, the full factorial, recovery and lifecycle tests, and complete cost accounting. Until then, PDSA remains a falsifiable architecture mechanism rather than an empirical success.

%% file: theory_family/modules/dual_state/appendices/A_proofs.tex
\section{Proofs and Constructions}
\label{tf:dual:app:proofs}

\subsection{Type and timing results}

\begin{proof}[Proof of Proposition~\ref{tf:dual:prop:type-safety}]
Each mutation event contains a carrier tag. By Assumption~\ref{tf:dual:ass:typed-auth}, authorization for a $\Phi$-tagged event is defined only on the policy carrier type and its successor map has codomain $\tfDualCalP\times\tfDualCalO$; authorization for an $M$-tagged event is defined only on the authoritative-memory type and has codomain $\tfDualCalM$. Induct on trace length. The empty trace is well typed. If the first $n$ events leave both carrier projections well typed, event $n+1$ either changes only the carrier named by its tag or is rejected. Thus it cannot apply its payload to the other carrier. Serialized-byte equality does not change the tag, domain, or codomain. The property holds for every finite trace.
\end{proof}

\begin{proof}[Proof of Proposition~\ref{tf:dual:prop:measurable}]
Proceed by induction on event order. The initial registered state and view are $\tfDualCalF_0$-measurable. Suppose $S_e$ and all prior receipts are $\tfDualCalF_e$-measurable. Assumption~\ref{tf:dual:ass:causal} makes $\kappa_e$, its authorized input $\xi_e$, and any registered randomness measurable with respect to $\tfDualCalF_e$ or a declared independent random seed revealed at $e$. Because $T_{\kappa_e}$ is a measurable typed map, $S_{e+1}$ and its receipt are measurable with respect to $\tfDualCalF_{e+1}$. Induction gives adaptedness. The same assumption separately forbids reading a later carrier version and binds the actual predecessor view; that structural premise excludes a future-version parent. A future value can be predictable, and hence already measurable, so adaptedness alone would not prove the access restriction.
\end{proof}

\subsection{Operation algebra}

\begin{proof}[Proof of Theorem~\ref{tf:dual:thm:commutation}]
Write the carrier projection of $S$ as $(P,A)$. By input stability, $U$ receives the same policy input and authorization decision whether evaluated before or after $K$; hence its policy successor is one fixed $P^+=U(P)$. Symmetrically, the memory candidate, validation, and authorization are invariant, so its successor is one fixed $A^+=K(A)$. Typed authorization makes $\widetilde U$ leave $A$ unchanged and $\widetilde K$ leave $P$ unchanged. Consequently both orders have carrier projection $(P^+,A^+)$. Receipt events preserve their temporal order; normalization may yield an equivalent set of receipts without identical serialized traces, so no full-trace equality is asserted.
\end{proof}

\begin{proof}[Proof of Proposition~\ref{tf:dual:prop:no-composite}]
Fix visible version numbers $(a,b)$. Construct policy histories $H_1^{\Phi}$ and $H_2^{\Phi}$ with distinct lineage identifiers: in the first, update directly from baseline to version $a$; in the second, fork at an earlier snapshot, perform a reset, and publish content-compatible version $a$. Pair each with the same memory history ending at $(b,\ell^M)$. The pinned version pair is identical but the policy predecessor and rollback target differ. The symmetric construction holds for two memory histories paired with one policy history. Hence the pair alone does not identify either carrier history. By Definition~\ref{tf:dual:def:compat-witness}, a valid witness for each view binds its exact lineage; witnesses for the two constructed lineages therefore differ (or verification fails). Such a witness identifies the registered lineages of that view but does not itself authorize either carrier's rollback.
\end{proof}

\subsection{Mechanism classification}

\begin{proof}[Proof of Theorem~\ref{tf:dual:thm:classification}]
Let $C$ indicate complete, non-conflicting, and verifiable coverage of every
receipt required by the interval. If $C=0$, the theorem's four-way premise is not
met and the trace is assigned \textsf{UNCLASSIFIED}; missing receipts are not
imputed as no-ops. If $C=1$, define binary indicators $I_{\Phi}$ and $I_M$ for the
strict policy path and the later read of a non-NULL updated authoritative-memory
successor, respectively. Under observability, both are functions of the audit trace
and are not imputed from output quality. The Cartesian set $\{0,1\}^2$ partitions
covered traces into four disjoint cells. The definitions in Section~\ref{tf:dual:sec:classification} name these cells
static/retry, \tfDualRTMA{} only, strict RTT only, and dual-state composition.
Exhaustiveness follows because each indicator is binary; exclusivity follows
because two distinct cells disagree in at least one indicator. The definitions do
not include an outcome sign, so no benefit follows.
\end{proof}

\begin{proof}[Proof of Corollary~\ref{tf:dual:cor:memory-not-rtt}]
If $\Phi$ is frozen, no authorized feedback-driven policy successor exists and $I_{\Phi}=0$. If a later attempt reads updated $M$, then $I_M=1$. Theorem~\ref{tf:dual:thm:classification} assigns the interval to \tfDualRTMA{} only. If there is no later memory read either, it is static/retry. Neither case is strict RTT.
\end{proof}

\subsection{Restoration and lineage}

\begin{proof}[Proof of Theorem~\ref{tf:dual:thm:rollback}]
Assumption~\ref{tf:dual:ass:local-restore} defines $\tfDualOpRollback_{\Phi}$ as a map on the policy carrier and its ledger, with the identity map on the memory carrier. Typed authorization prevents its event from naming $M$ as a destination. Hence $\pi_A(\tfDualOpRollback_{\Phi}(S))=\pi_A(S)$, including content, version, and lineage. The memory case is symmetric. Compatibility evaluation is a later coordinator map whose outcomes are pin or refusal; it has no carrier-write privilege. Therefore a compatibility failure cannot alter the unchanged carrier.
\end{proof}

\begin{proof}[Proof of Corollary~\ref{tf:dual:cor:composite-recovery}]
By Theorem~\ref{tf:dual:thm:rollback}, one carrier-local rollback changes at most one carrier. Changing both therefore requires the composition of two authorized carrier-local maps. An explicit composite manifest is conforming only if its execution expands to those maps and retains both receipts. An unnamed single event cannot satisfy the typed domain requirement.
\end{proof}

\begin{proof}[Proof of Proposition~\ref{tf:dual:prop:history}]
A restore produces a new lineage edge labeled reset, swap, or rollback whose successor has a new version and receipt. Append-only storage prevents deletion of that edge or the intervening vertices. Content addressing may show that the restored payload equals an earlier payload, but the new vertex and edge remain. An auditor reading the ledger therefore distinguishes restoration from uninterrupted continuation.
\end{proof}

\subsection{Compression and drift}

\begin{proof}[Proof of Theorem~\ref{tf:dual:thm:contamination}]
Let $d_t=\lVert\Phi_t-\Phi_t^I\rVert$. Nonexpansiveness of projection and the triangle inequality give, for informative $t$,
\[
d_{t+1}\le d_t+\eta_t\lVert g(\Phi_t,X_t)-g(\Phi_t^I,X_t)\rVert
\le(1+\eta_tL_g)d_t.
\]
For contaminant $t$, the coupled informative path takes a zero update, so
\[
d_{t+1}\le d_t+\eta_t\lVert g(\Phi_t,X_t)\rVert
\le(1+\eta_tL_g)d_t+\eta_tB_t,
\]
where the last step adds and subtracts $g(\Phi_t^I,X_t)$. With $d_0=0$, unrolling this recurrence yields Equation~\eqref{tf:dual:eq:contam-state}. Lipschitz continuity of $\ell$ gives Equation~\eqref{tf:dual:eq:contam-loss}. Absolute value removes the sign, so harm does not follow.
\end{proof}

\begin{proof}[Proof of Proposition~\ref{tf:dual:prop:no-dominance}]
Use the finite-horizon binary construction in Counterexample~\ref{tf:dual:cex:sufficient}. The count recurrence is Bayes-optimal; charge a strictly positive cost for a redundant added carrier to make its total risk strictly worse. Conversely, choose a one-bit recurrence that overwrites its state on each event, a target equal to the initial bit, and a later independent nuisance bit. Its Bayes error after overwrite is $1/2$, whereas a protected row retaining the initial bit has zero task error. A positive row cost below $1/2$ preserves this strict advantage. These are two declared realizations under different permissible laws, update classes, and cost models. They preclude a uniform ordering of arbitrary systems named by architecture alone; they do not compare optimally tuned architecture classes under one fixed common resource constraint.
\end{proof}

\begin{proof}[Proof of Proposition~\ref{tf:dual:prop:path-length}]
Because $\Phi_k\in\tfDualCalP$ and projection fixes points in $\tfDualCalP$,
\[
\lVert\Phi_{k+1}-\Phi_k\rVert
=\lVert\tfDualProj_{\tfDualCalP}(\Phi_k-\eta_k\widehat g_k)-\tfDualProj_{\tfDualCalP}(\Phi_k)\rVert
\le\eta_k\lVert\widehat g_k\rVert\le\eta_kG_k.
\]
Sum over $k$ and apply the triangle inequality. The stated total-variation bound follows directly from its Lipschitz premise.
\end{proof}

\begin{proof}[Proof of Theorem~\ref{tf:dual:thm:interaction-envelope}]
Substitute the intervention bounds $\delta_{\Phi}\le\epsilon_{\Phi}$ and $\delta_M\le\epsilon_M$ into Equation~\eqref{tf:dual:eq:interaction-lipschitz}; all coefficients and distances are nonnegative. For integrable bounded losses, take expectations and use $|\tfDualE X|\le\tfDualE|X|\le B_{\Phi M}$. The inequality is absolute and supplies no sign or additive causal attribution.
\end{proof}

\subsection{Factorial and observational results}

\begin{proof}[Proof of Theorem~\ref{tf:dual:thm:factorial}]
Random assignment and positive support imply
$Z_i\perp\{Y_i(p,m):p,m\in\{0,1\}\}$. Perfect compliance and consistency give $Y_i^{\mathrm{obs}}=Y_i(p,m)$ in cell $(p,m)$. Hence
\[
\tfDualE[Y_i^{\mathrm{obs}}\mid Z_i^{\Phi}=p,Z_i^M=m]
=\tfDualE[Y_i(p,m)]=\mu_{pm}.
\]
Linear combinations of the four identified cell means identify the displayed main and interaction effects. For known probabilities $\pi_{pm}>0$, the Horvitz--Thompson term
$\tfDualOne\{Z=(p,m)\}Y^{\mathrm{obs}}/\pi_{pm}$ has expectation $\mu_{pm}$, proving the final statement.
\end{proof}

\begin{proof}[Proof of Proposition~\ref{tf:dual:prop:receipt-selection}]
Let $D$ be successful exposure. Even when $Z$ is randomized, $D$ can depend on assignment, latent problem difficulty, resource pressure, or potential outcomes. Conditioning on $D=1$ opens selection paths and generally destroys $Z\perp Y(p,m)$. Thus receipt-positive differences are not automatically randomized effects. Guaranteed compliance or an explicit identification model is required.
\end{proof}

\begin{proof}[Proof of Theorem~\ref{tf:dual:thm:trace-equivalence}]
Assume first that $R\in\mathfrak R_{M\leftarrow\Phi}(K;\tfDualCalI)$.  By the definition of this set, $R$ supplies a legal encoder, an authorized complete-row transition, a verified interface witness, a registered memory reader and decoder, and the same exogenous-randomness coupling.  Conditional on each $(q,z_0,f_0)$, the decoder in $R$ therefore emits exactly $K(\cdot\mid q,z_0,f_0)$; integrating over the common law of $(q,Z_0,F_0)$ gives the same law for $O$.  No unregistered cast or unauthorized lookup is introduced.  The converse applies the identical argument to any $R'\in\mathfrak R_{\Phi\leftarrow M}(K;\tfDualCalI)$.  If the relevant set is empty, the construction has no admissible first step, so the theorem asserts no equivalence.  Thus observables fail to name a carrier only among realizations satisfying the stated type, authorization, compatibility, capacity, causality, and randomness obligations.
\end{proof}

\begin{proof}[Proof of Proposition~\ref{tf:dual:prop:coupled-closure}]
Typed authorization makes $U(P)$ a policy-carrier successor with receipt $\rho_{\Phi}^+$ and $K(A)$ a memory-carrier successor with receipt $\rho_M^+$; neither map has the other's codomain.  Interface soundness permits a pin only after a witness $\kappa\in\Gamma_{\tfDualCalI}(P^+,A^+,C)$ is verified by the components actually used at read time.  Equation~\eqref{tf:dual:eq:coordination-manifest} binds both carrier receipts, the witness, interface, and context, so a published view names the exact well-typed coordination manifest.  If verification fails, the coordinator's codomain contains refusal/fallback and receipt append, not a carrier mutation.  Hence it cannot publish the manifest or silently alter either successor.
\end{proof}

\begin{proof}[Proof of Corollary~\ref{tf:dual:cor:one-trajectory}]
For one unit, consistency reveals only $Y_i(Z_i^{\Phi},Z_i^M)$. The interaction contains all four potential outcomes. Without structural restrictions that determine the three unobserved values, multiple potential-outcome schedules agree on the observed trajectory and yield different interactions. Hence one trajectory does not identify it.
\end{proof}

\begin{proof}[Proof of Proposition~\ref{tf:dual:prop:swap-contrast}]
Under randomized valid swap assignment, donor content is independent of recipient potential outcomes and all declared non-content fields are fixed or balanced. Consistency therefore identifies the mean outcome under recipient-updated content and under a draw from the declared compatible-donor distribution. Their difference is the recipient-specific-content contrast relative to that distribution. Nothing in the assumptions ranges over every possible donor or semantic representation, so no universal effect follows.
\end{proof}

\subsection{Authorization, cost, and recovery}

\begin{proof}[Proof of Proposition~\ref{tf:dual:prop:blocked}]
Distinct credentials and typed endpoints make the policy service unauthorized at the authoritative commit endpoint and make the candidate/committer service unauthorized at the policy endpoint. Schema checks reject a payload carrying the wrong carrier type. Therefore a call confined to one ordinary service cannot directly publish the other's successor. The conclusion does not survive compromise of the shared authorization root or both credentials, which the proposition excludes.
\end{proof}

\begin{proof}[Proof of Proposition~\ref{tf:dual:prop:cost-identity}]
The five event sets form a disjoint partition of all charged operations. For the explicitly additive subvector, finite additivity gives Equation~\eqref{tf:dual:eq:ledger-identity}. Each listed activity is assigned to one set, including failures and recovery, and shared energy or traffic is attributed once. Reporting only generation omits non-generation charges. No sum identity is asserted for the makespan or peak-occupancy coordinates of the full vector: overlapping operations can have duration sum larger than makespan, and shared resident storage must not be counted twice.
\end{proof}

\begin{proof}[Proof of Corollary~\ref{tf:dual:cor:parallelism}]
Parallel scheduling changes the temporal arrangement of events and can reduce measured makespan. It does not remove executed operations from the additive partition or make their logical counts and transferred bytes vanish. Only those additive coordinates remain summed; peak occupancy can increase or decrease with the schedule and must be measured separately.
\end{proof}

\begin{proof}[Proof of Proposition~\ref{tf:dual:prop:recovery-locality}]
By Theorem~\ref{tf:dual:thm:rollback}, carrier-local restoration changes only the named carrier. A dependency-complete cache graph identifies every derived cache entry and permits invalidation without altering source carrier content. Compatibility checks can withhold availability. Thus recovery of one carrier need not change the other's content, though temporary service interruption may occur.
\end{proof}

\begin{proof}[Proof of Proposition~\ref{tf:dual:prop:post-reset-pin}]
A reset, swap, or rollback creates a fresh successor version and lineage receipt even when its payload is byte-identical to an earlier payload.  The old joint pin and witness bind the predecessor version, lineage, receipt, and lifetime fields, so interface verification rejects them for the successor pair.  Content equality therefore cannot recreate the old identity.  A usable post-restoration view requires a newly verified $\kappa\in\Gamma_{\tfDualCalI}$ and a fresh coordination-manifest receipt binding both current successors.  This prevents an ABA-style return of payload bytes from masquerading as uninterrupted joint-state continuity.
\end{proof}

\subsection{Proof boundary}

Every result above is conditional on its displayed premises. None establishes a qualified semantic interface, actual privilege separation, valid feedback, bounded gradients, Lipschitz behavior, randomized compliance, a measured endpoint, or a complete ledger. Those are empirical or implementation obligations in Appendix~\ref{tf:dual:app:obligations}.

%% file: theory_family/modules/dual_state/appendices/B_factorial_protocol.tex
\section{Registered Factorial Protocol Specification}
\label{tf:dual:app:factorial-protocol}

\subsection{Purpose and gate order}

This appendix is an unexecuted protocol template. It cannot begin until the semantic-interface gate and carrier-isolation stage in Protocol~\ref{tf:dual:prot:ladder} pass. It estimates policy and memory intervention effects without treating assignments, successful pins, or final outputs as interchangeable.

\subsection{Immutable registration}

Before any evaluation outcome is opened, register:
\begin{enumerate}
  \item problem-group construction, inclusion/exclusion rules, immutable group identifiers, and train/development/final split hashes;
  \item base model, tokenizer, precision, software/hardware identity, causal workspace schema, maximum context and attempt budget;
  \item exact $\Phi_0,\Phi_1,M_0,M_1$ construction, snapshot hashes, lineages, lifetimes, and compatibility rules;
  \item update objectives, feedback sources, optimizer, step schedule, clipping/projection, memory candidate/validator/commit contract, and event order;
  \item primary task loss, semantic and preservation gates, severe-event endpoints, subgroups, cost vector, materiality/equivalence margins, and missing-data rules;
  \item randomization algorithm, assignment probabilities, blocking variables, seed list, multiplicity family, interval method, and stop rules;
  \item exact analysis code or a content-addressed specification sufficient for independent reconstruction.
\end{enumerate}

No sample-size number or effect magnitude is supplied here because none is supported by the frozen record. A future design must derive group count from a preregistered smallest effect of interest, variance pilot that is isolated from final evaluation, desired power, familywise error, expected noncompliance, and severe-event precision.

\subsection{Unit and treatment construction}

The randomization unit is a problem group containing every attempt, branch, paraphrase, and derived query that shares latent content or an upstream source. Units are blocked on preregistered difficulty and domain variables, then assigned to four cells with positive probability. Assignment occurs before the evaluated attempt and is recorded independently from execution services.

\begin{table}[H]
\centering
\small
\begin{tabularx}{\linewidth}{@{}p{0.10\linewidth}p{0.20\linewidth}p{0.20\linewidth}Y@{}}
\toprule
Cell & Policy treatment & Memory treatment & Pinned-view requirement \\
\midrule
$00$ & registered fresh/frozen baseline $\Phi_0$ & registered fresh/frozen baseline $M_0$ & exact $(v^{\Phi}_0,v^M_0)$ \\
$10$ & feedback-updated $\Phi_1$ & baseline $M_0$ & exact $(v^{\Phi}_1,v^M_0)$ \\
$01$ & baseline $\Phi_0$ & qualified committed $M_1$ & exact $(v^{\Phi}_0,v^M_1)$ \\
$11$ & feedback-updated $\Phi_1$ & qualified committed $M_1$ & exact $(v^{\Phi}_1,v^M_1)$ \\
\bottomrule
\end{tabularx}
\caption{Primary availability factorial. $M_1$ does not exist as qualified evidence in this paper.}
\end{table}

Exact \tfDualNull{} is an outcome of the registered memory transition, not missing data. If a cell assigned to memory update receives \tfDualNull{}, assignment remains $Z^M=1$ while realized exposure is recorded separately. Failed pins, timeouts, validator rejects, and recovery events remain in intent-to-treat analysis under registered loss/missingness rules.

\subsection{Execution sequence}

For each unit:
\begin{enumerate}
  \item verify immutable registration and cold/fixed evaluation boundary;
  \item materialize carrier baselines from registered snapshots;
  \item run the common pre-intervention attempt and record admissible feedback;
  \item execute independently authorized policy and memory events in the assigned order;
  \item verify both successor receipts, caches, compatibility, and exact pinned view;
  \item run the evaluated attempt under the fixed sampler and resource contract;
  \item compute endpoints with an evaluator blind to cell labels where feasible;
  \item append all physical and logical costs, including failures and recovery;
  \item freeze final group bytes before opening aggregate cell summaries.
\end{enumerate}

If the policy and memory events may not commute, randomize their order within the joint arm or treat order as a registered additional factor. Do not average undeclared interleavings.

\subsection{Primary estimators}

For equal-probability complete randomization, let $\widehat\mu_{pm}$ be the group mean in cell $(p,m)$. Estimate
\begin{align}
\widehat\tau_{\Phi\mid m}&=\widehat\mu_{1m}-\widehat\mu_{0m},\\
\widehat\tau_{M\mid p}&=\widehat\mu_{p1}-\widehat\mu_{p0},\\
\widehat\tau_{\Phi M}&=\widehat\mu_{11}-\widehat\mu_{10}-\widehat\mu_{01}+\widehat\mu_{00}.
\end{align}
Use randomization-based or group-robust uncertainty consistent with the assignment design. With unequal probabilities, use registered inverse-probability weights. Covariate adjustment may improve precision only if specified before final outcomes and if unadjusted estimates remain reported. The primary estimand is intent-to-treat; exposure/compliance analyses are secondary unless compliance is guaranteed.

\subsection{Negative controls}

The protocol includes:
\begin{enumerate}
  \item same carrier versions with independent sampler seeds;
  \item frozen policy with identical extra attempt and feedback-computation budget;
  \item swapped policy with matched update count, base model, tokenizer, optimizer schema, and budget;
  \item swapped memory with matched schema, tenant-safe synthetic scope, row count, version depth, and bytes;
  \item independent policy rollback with unchanged memory and independent memory rollback with unchanged policy;
  \item invalid/mismatched compatibility canaries that must block pinning;
  \item target-revealing feedback canaries that must fail the leakage gate;
  \item exact \tfDualNull{} and no-op controls whose authoritative resident bytes remain identical.
\end{enumerate}

\subsection{Stop and decision rules}

Stop immediately for cross-tenant access, future-information leakage, direct cross-carrier write, silent joint reset/rollback, unverified deletion, receipt corruption, or evaluation-boundary mutation. Stop the scientific stage if the semantic, preservation, strict-RTT timing, intervention compliance, severe-event, or resource gate fails. A favorable task mean cannot override these stops. A passing factorial requires all registered gates; a non-significant interaction is not automatically equivalence unless its confidence set lies within the registered equivalence margin.

\subsection{Reporting}

Report all four cell summaries; conditional and marginal main effects; interaction; assignment and exposure counts; \tfDualNull{}, failure, timeout, and rollback counts; subgroups; severe events; update/commit order; drift; every cost coordinate; exact software/hardware and carrier hashes; and all deviations. The report must state that the design does not establish a learned admission policy unless a separately qualified admission stage exists.

%% file: theory_family/modules/dual_state/appendices/C_obligation_matrix.tex
\section{Theorem-to-Implementation Obligation Matrix}
\label{tf:dual:app:obligations}

\small
\begingroup
\setlength\LTleft{\fill}
\setlength\LTright{\fill}
\begin{longtable}{@{}p{0.18\linewidth}p{0.22\linewidth}p{0.28\linewidth}p{0.24\linewidth}@{}}
\caption{Every conditional result remains unusable as a system claim until its complete premise bundle is evidenced.}\label{tf:dual:tab:full-obligations}\\
\toprule
Claim/result & Formal premise & Required durable evidence & If absent \\
\midrule
\endfirsthead
\toprule
Claim/result & Formal premise & Required durable evidence & If absent \\
\midrule
\endhead
\midrule
\multicolumn{4}{r}{Continued on next page}\\
\endfoot
\bottomrule
\endlastfoot
Policy carrier identity & typed $\tfDualCalP\times\tfDualCalO$, version and lineage & complete parameter/optimizer manifest, scope, writer credential, predecessor/successor hashes & policy update/reset/rollback claim invalid \\
Memory carrier identity & complete typed rows and protected fields & serializer/assembler version, full resident snapshot, row/slot versions, ACL/provenance/retention receipts & authoritative-memory claim invalid \\
Typed coupling interface & registered readers and decoder have declared domains/codomains & reader/decoder versions, schema hashes, carrier tags, negative cast tests & no well-typed joint read is established \\
Compatibility witness & verified witness binds both carriers and the actual read components & signed witness, hashes, versions, lineages, tenant/ACL, lifetimes, schemas, interface and fallback policy & joint view must refuse or fall back \\
Pinned read view & exact versions and verified witness remain stable during attempt & pin receipt, read-service logs, cache version, witness, and coordination-manifest receipt & attempt exposure unknown \\
Type safety & carrier-specific domains and privileges & distinct endpoints/credentials, negative authorization tests, carrier-tag schema & syntactic proposition not implemented \\
Causal measurability & inputs available at event time & timestamp/order source, information-set schema, future-access guards, sampler seed & leakage not excluded \\
Conditional commutation & input and decision stability in both orders & replay of both orders from identical snapshot, candidate/gradient equality, compatibility receipts & order is a treatment \\
Strict RTT & feedback update and later same-problem read & unresolved marker, feedback receipt, policy update lineage, later pin, frozen/reset intervention & call only retry or unclassified adaptation \\
\tfDualRTMA{} & non-NULL memory successor and later read & non-NULL complete-row commit receipt, predecessor/successor lineage, later verified pin/read, frozen/reset intervention; exact \tfDualNull{} is a separate no-op control & memory mechanism unclassified \\
Coupled transition & two typed carrier transitions plus verified compatibility & both carrier-local successor receipts, witness, signed coordination-manifest receipt, refusal/fallback log & no atomic pair or coupled-update claim \\
Independent rollback & carrier-local restore and cache graph & complete snapshot, fresh restore successor, unchanged other-carrier hash, old joint-pin/witness invalidation, cache invalidation & recovery and ABA-resistance claims invalid \\
Append-only history & durable content-addressed lineage & storage/ledger integrity proof, restore edge, retained intervening vertices & rollback can be observationally erased \\
Contamination bound & nonexpansive projection, Lipschitz gradients/loss, bounded contaminant updates & registered region, measured/bounded constants, fixed contaminant definition, update trace & theorem cannot bound deployment \\
Policy path length & bounded gradients and steps & exact step sizes, gradient norms, projection implementation, checkpoints & displacement bound unavailable \\
Interaction envelope & registered state metrics and constants & intervention radii, mixed-sensitivity validation on region & envelope unavailable \\
Factorial identification & randomization, support, consistency, no interference, compliance & assignment receipt, group construction, pins, exposure, attrition, frozen analysis & cell contrasts descriptive \\
Swap specificity & valid donor sampling and compatibility & donor-selection seed, no-leak proof, matched metadata, independent snapshot & content contrast confounded \\
Output non-identification & nonempty legal realization set under the registered interface, compatibility, authorization, lifetime, capacity, causal, and randomness constraints & construction plus the complete witness class for at least one legal realization & no carrier-indistinguishability conclusion \\
Privilege isolation & distinct credentials and trusted mediation & authorization topology, penetration/negative tests, candidate quarantine & contamination-blocking claim unavailable \\
Cost identity & event-complete partition and units & per-event vector including witness verification, signatures, coordination-manifest receipts, fallback, failed work, cache proof, measurement calibration & efficiency claim prohibited \\
Recovery locality & independent restore plus complete cache graph & dependency inventory, carrier hashes before/after, availability log & other carrier may have changed \\
Semantic qualification & registered same-checkpoint interface gates & independent fit and held-out, query, preservation, mapping, and integrity adjudication & every memory-composition stage remains closed \\
Learned admission & qualified interface, oracle gap, future-blind value and selector gates & separate predictive-admission protocol receipts and accepted evidence & no learned policy claim \\
\end{longtable}
\endgroup
\normalsize

The final two rows are scientific prerequisites rather than premises of the purely mathematical carrier algebra. They prevent formal consistency from being misreported as an implemented or learned MMLA system.

%% file: theory_family/modules/dual_state/appendices/D_symbols.tex
\section{Symbols, Status, and Evidence Ledger}
\label{tf:dual:app:symbols}

\subsection{Core symbols}

\begin{longtable}{@{}p{0.20\linewidth}p{0.72\linewidth}@{}}
\toprule
Symbol & Meaning \\
\midrule
\endfirsthead
\toprule
Symbol & Meaning \\
\midrule
\endhead
\bottomrule
\endlastfoot
$q,r,t,e$ & problem, attempt, within-attempt micro-event, and global event indices \\
$\tfDualCalF_e$ & information available immediately before event $e$ \\
$W_{q,r,t}$ & bounded causal workspace; excludes future information \\
$P=(\Phi,O,v^{\Phi},\ell^{\Phi},\tau^{\Phi},\rho^{\Phi})$ & policy carrier: parameters, optimizer, version, lineage, lifetime, receipt \\
$A=(M,v^M,\ell^M,\tau^M,\rho^M)$ & authoritative memory carrier and its metadata \\
$V_{q,r}$ & pinned read-view tuple naming both carrier versions/lineages and compatibility receipt \\
$\tfDualCalI=(R_{\Phi},R_M,D)$ & registered typed coupling interface: policy reader, authoritative-memory reader, and joint decoder \\
$\kappa$ & compatibility witness binding both carriers, the interface, authorization, lifetime, and fallback fields \\
$\Gamma_{\tfDualCalI}$ & set of witnesses verified under interface $\tfDualCalI$ for the exact joint read \\
$\rho_{\mathrm{coord}}$ & signed coordination-manifest receipt binding both carrier-local receipts, $\kappa$, $\tfDualCalI$, and causal context; it grants no write or rollback authority \\
$\mathfrak R_{M\leftarrow\Phi},\mathfrak R_{\Phi\leftarrow M}$ & legal cross-carrier realization sets used by the conditional output-trace theorem \\
$C,G,Q$ & immutable causal context; generation/sampler state; queues, caches, snapshots, and ledgers \\
$\mathfrak O_{\Phi},\mathfrak O_M$ & independent carrier operation algebras \\
$U,K$ & one policy update and one memory transition \\
$Q_q(r)$ & indicator that problem $q$ remains unresolved after attempt $r$ \\
$Y_i(p,m),\mu_{pm}$ & potential loss and population cell mean under carrier interventions $(p,m)$ \\
$\tau_{\Phi\mid m},\tau_{M\mid p}$ & conditional policy and memory effects \\
$\tau_{\Phi M}$ & factorial interaction $\mu_{11}-\mu_{10}-\mu_{01}+\mu_{00}$ \\
$Z_i^{\Phi},Z_i^M$ & randomized assignments \\
$D_i^{\Phi},D_i^M$ & receipt-verified realized exposures/compliance \\
$\mathbf C$ & vector resource ledger \\
$\tfDualNull$ & exact authoritative memory identity transition; rejection is externally receipted and is not a positive RTMA path \\
\end{longtable}

\subsection{Status ledger}

\begin{table}[H]
\centering
\small
\begin{tabularx}{\linewidth}{@{}p{0.25\linewidth}p{0.22\linewidth}Y@{}}
\toprule
Object & Status in this paper & Boundary \\
\midrule
Typed dual-state semantics & \tfDualDefinitionStatus{} & editorial/mathematical construction \\
Type, interface, compatibility, coupled-update, timing, rollback, drift, contamination, interaction, and identification results & \tfDualProvedStatus{} & conditional on displayed assumptions \\
Semantic row interface & \tfDualUnvalidatedStatus{} & inherited failed/unqualified same-checkpoint gate \\
Strict-RTT policy implementation & \tfDualUnvalidatedStatus{} & no policy update or same-problem intervention \\
Authoritative memory composition & \tfDualUnvalidatedStatus{} & no qualified row interface or dual-state commit \\
$2\times2$ experiment & \tfDualPlannedStatus{} & no unit assigned or evaluated \\
Oracle gap and predictive admission & \tfDualUnvalidatedStatus{} & not measured; gate closed \\
Learned admission policy & \tfDualUnvalidatedStatus{} & no value model, selector, or policy \\
Dual-state benefit/safety/efficiency & \tfDualUnvalidatedStatus{} & no empirical result \\
\bottomrule
\end{tabularx}
\caption{No empirical status is promoted by formal completeness.}
\end{table}

\subsection{Frozen numerical evidence}

The following inherited negative/diagnostic boundary is retained for this
formal part. Later restricted readout successes and generalized failures are
reported separately in Sections~\ref{sec:posttraining-evidence} and
\ref{sec:coverage-readout}; they do not validate the dual-state intervention:
\begin{itemize}
  \item frozen-checkpoint semantic-interface study: $0/9$ registered interfaces qualified.
  \item fitted-support fault-localization study direct candidate: $40/73{,}728$ exact.
  \item Direct present path: $0$.
  \item Learned route/learned content: $0/73{,}728$.
  \item Oracle route/learned content: $0/73{,}728$.
  \item Learned route/oracle content: $18{,}544/73{,}728$, exactly routing count.
  \item Oracle route/oracle content: $73{,}728/73{,}728$, mechanical ceiling only.
  \item Preservation: $0$; missing cases: $43{,}008$.
\end{itemize}
These counts delimit what is not qualified. They do not estimate any dual-state factorial cell, interaction, policy drift, contamination rate, resource cost, oracle gap, or admission effect.

\subsection{Evidence-type boundary}

Definitions and proofs establish formal consequences under stated premises. They do not qualify an implemented mechanism or promote an empirical claim.

%% file: theory_family/modules/csbc/module.tex
\clearpage
\part{Causal Generation, Retrospective Consolidation}
\label{tf:csbc:part:completed-segment-consolidation}

\noindent\textbf{Scope.}
This part separates causal token generation from post-closure bidirectional
inspection, ordered event construction, trusted per-event publication, and
future-only exposure. It develops timing, ownership, idempotence, event
recursion, asynchronous consistency, recovery, and resource obligations.

\input{theory_family/modules/csbc/sections/01_scope}
\input{theory_family/modules/csbc/sections/02_typed_model}
\input{theory_family/modules/csbc/sections/03_five_clocks}

\input{theory_family/modules/csbc/sections/04_causal_boundary}
\input{theory_family/modules/csbc/sections/05_training_amputation}
\input{theory_family/modules/csbc/sections/06_boundaries_carry}
\input{theory_family/modules/csbc/sections/07_ownership_idempotence}
\input{theory_family/modules/csbc/sections/08_event_recursion}
\input{theory_family/modules/csbc/sections/09_async_consistency}
\input{theory_family/modules/csbc/sections/10_resource_bounds}
\input{theory_family/modules/csbc/sections/11_security_recovery}
\input{theory_family/modules/csbc/sections/12_counterexamples}
\input{theory_family/modules/csbc/sections/13_related_work}
\input{theory_family/modules/csbc/sections/14_falsification}
\input{theory_family/modules/csbc/sections/15_limits_conclusion}

\FloatBarrier
\input{theory_family/modules/csbc/appendices/A_proofs}
\input{theory_family/modules/csbc/appendices/B_protocol_obligations}
\input{theory_family/modules/csbc/appendices/C_counterexample_catalogue}
\input{theory_family/modules/csbc/appendices/D_symbols}

%% file: theory_family/modules/csbc/sections/01_scope.tex
\section{Introduction}
\label{tf:csbc:sec:scope}

\begin{figure}[H]
\centering
\includegraphics[width=\linewidth]{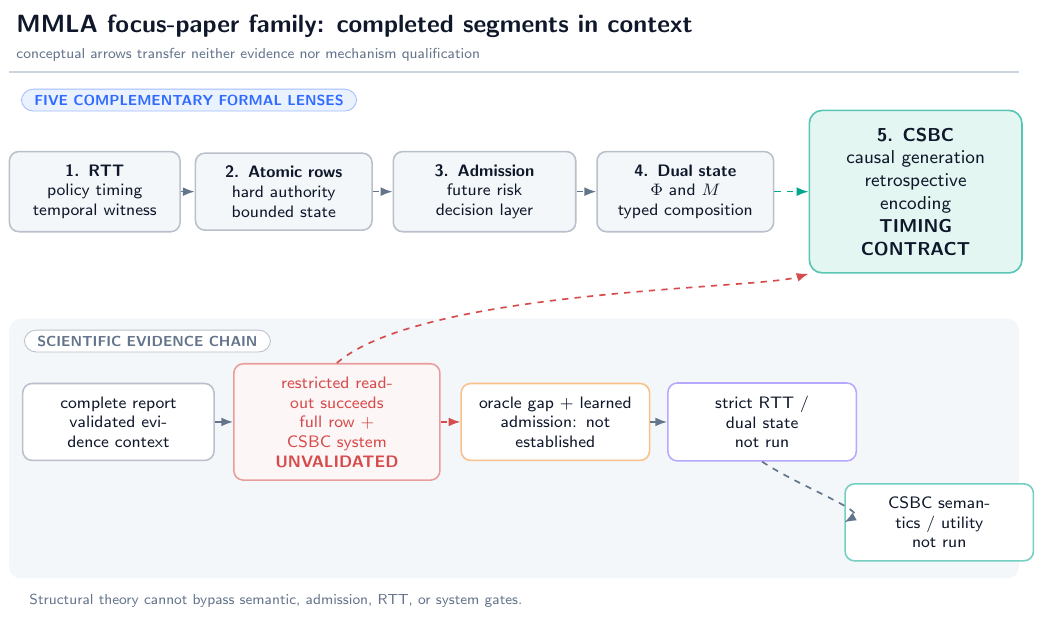}
\caption{Completed-segment consolidation: causal prediction, post-closure interpretation, ordered publication, and later exposure. The full end-to-end composition remains unvalidated.}
\label{tf:csbc:fig:family-tree}
\end{figure}

\subsection{Bidirectionality belongs after closure}

An autoregressive generator must choose token $X_t$ without reading tokens that do not yet exist. A retrospective encoder can nevertheless inspect a finite block in both directions once every token in that block is already emitted. The apparent contradiction disappears only when the segment boundary, encoding start, commit order, and later exposure are explicit.

\tfCsbcDefinitionStatus
\begin{tfCsbcDefinition}[Completed-segment bidirectional consolidation]
Completed-segment bidirectional consolidation (\tfCsbcName{}) is a protocol in which (i) token generation through a declared boundary is causal; (ii) the completed bounded view is constructed only after that boundary; (iii) a retrospective encoder may inspect only that completed view bidirectionally; (iv) each derived event produces at most one complete authoritative-row commit or exact \tfCsbcNull{}; and (v) every resulting version is first readable at an exposure point strictly after the boundary. Historical emissions and receipts are immutable.
\end{tfCsbcDefinition}

The phrase \emph{retrospective encoding} names step (iii). It does not imply reverse-time generation, future-token access, retrospective gradient updates during deployment, or correction of already emitted text. A later answer can use a committed correction; the earlier mistaken sentence remains part of history.

\subsection{Evidence boundary}

Historical interface evidence is negative and diagnostic: the frozen-checkpoint semantic-interface study qualified $0/9$ jobs; fitted-population learned content was exact on $0/73{,}728$ probes with either learned or mechanical target routing. Later post-training establishes restricted generated-answer update and restoration, while the latest nine-trajectory comparison still fails generalized continuous-event latent qualification. None tests the complete \tfCsbcName{} timing and exposure contract. At the September 13, 2026 cutoff, mechanically well-timed state and semantically useful state therefore remain separate empirical obligations.

The integrated RTT, atomic-row, predictive-admission, and dual-state parts provide the required terminology and formal contracts. None transfers empirical validation. In particular:
\begin{itemize}
  \item when a \tfCsbcName{} execution produces a non-NULL authoritative-memory successor that is later read while the problem remains unresolved, that trace satisfies the memory-side \tfCsbcRTMA{} condition; a NULL/no-op or an unread successor does not;
  \item it is strict \tfCsbcRTT{} only if a separately evidenced feedback-driven policy update is later read on the same unresolved problem;
  \item a valid event candidate is not a qualified semantic interface;
  \item an expected-risk admission rule is neither run nor assumed here; and
  \item a conditional theorem does not reopen a failed empirical gate.
\end{itemize}

\subsection{Contribution and nonclaims}

This manuscript owns a timing and systems contract: typed boundaries, five clocks, filtration, exposure, teacher amputation, overlap/carry, event ownership, idempotence, event recursion, asynchronous consistency, resource bounds, information-flow restrictions, failure recovery, and falsification obligations. It does not own a successful event detector, semantic encoder, factual verifier, admission model, storage engine, policy updater, or deployed system.

Formal statements carry their assumptions locally. Definitions, proofs, design prescriptions, planned tests, and unvalidated mechanisms are marked as such; no label promotes an implementation or empirical claim.

\tfCsbcConjecturedStatus\quad A post-closure bidirectional view may help some eventizers represent local corrections or relations. The conjecture remains open. This paper establishes only what would have to be true for such a mechanism to be causal, bounded, auditable, and falsifiable.

%% file: theory_family/modules/csbc/sections/02_typed_model.tex
\section{Typed Process and Carrier Inventory}
\label{tf:csbc:sec:typed-model}

\subsection{Probability space and causal tokens}

Fix a probability space $(\Omega,\tfCsbcCalF,\tfCsbcP)$, token alphabet $\tfCsbcCalX$, and token filtration $(\tfCsbcCalF_t^X)_{t\ge0}$. At token time $t$, a frozen deployment generator with parameters $\theta\in\tfCsbcCalP_{\rm dep}$ reads a pinned authoritative-memory version $\nu_t$ and draws
\begin{align}
L_t &= G_\theta(X_{<t},\tfCsbcRead(M^{\nu_t}),\gamma_t),
\label{tf:csbc:eq:logit}\\
X_t &= S(L_t,\xi_t),
\label{tf:csbc:eq:sample}
\end{align}
where $\gamma_t$ is declared causal workspace and $\xi_t\in\tfCsbcCalU$ is a recorded random coin. The immutable emission receipt is
\begin{equation}
\rho_t^X=\tfCsbcHash(\mathsf{episode},t,\tfCsbcHash(X_{<t}),\tfCsbcHash(L_t),\xi_t,X_t,
\nu_t,\tfCsbcHash(\gamma_t)).
\label{tf:csbc:eq:emission-receipt}
\end{equation}
The receipt may store digests rather than an unrestricted internal execution trace. It must nevertheless bind the causal inputs needed by the claimed reproducibility boundary.

\subsection{Segment and event types}

Let boundaries be $0=b_0<b_1<\cdots$. Core $I_n=(b_{n-1},b_n]\cap\tfCsbcN$ has length $C_n=|I_n|$. With overlap $O_n\subseteq\{1,\ldots,b_{n-1}\}$ and carry $q_{n-1}\in\tfCsbcCalQ$, the completed view is
\begin{equation}
V_n=\tfCsbcView(X_{O_n},X_{I_n},q_{n-1};\beta_n)\in\tfCsbcCalV,
\label{tf:csbc:eq:completed-view}
\end{equation}
constructed only after $X_{b_n}$ and $\rho_{b_n}^X$ exist. Boundary rule $\beta_n\in\tfCsbcCalB$ is either fixed or an adapted stopping rule specified in Section~\ref{tf:csbc:sec:boundaries}.

A retrospective encoder $B_\psi:\tfCsbcCalV\to\tfCsbcCalH$ may attend bidirectionally within $V_n$. The eventizer returns
\begin{equation}
(E_n,q_n)=\tfCsbcEventize(B_\psi(V_n),q_{n-1};\zeta_n),
\quad E_n=(e_{n,1},\ldots,e_{n,K_n}),\quad K_n\le K_{\max},
\label{tf:csbc:eq:eventizer}
\end{equation}
where $\zeta_n$ is recorded randomness and $q_n$ has a fixed maximum serialization size $B_Q$. Each event has typed span, content proposal, anchor $\alpha(e)\in\tfCsbcN$, ownership core, source receipts, and idempotence key
\begin{equation}
\iota(e)=\tfCsbcHash(\mathsf{episode},\mathsf{eventizer\mbox{-}build},
\mathsf{normalized\mbox{-}span},\alpha(e),\mathsf{event\mbox{-}type}).
\label{tf:csbc:eq:idempotence-key}
\end{equation}

\subsection{Authority, service, and training objects}

Authoritative memory is a fixed table $M\in\tfCsbcCalM=\tfCsbcCalR^S$. Target selection is an explicit typed system operation,
\begin{equation}
\mathcal A_{n,j}=\operatorname{Route}(e_{n,j},M_{n,j-1})\subseteq [S],
\qquad R_{n,j}=|\mathcal A_{n,j}|\le R_{\max}.
\label{tf:csbc:eq:target-route}
\end{equation}
For each routed target $a\in\mathcal A_{n,j}$, a proposer returns a complete candidate or failure,
\begin{equation}
\widehat r_{n,j,a}=\tfCsbcConstruct(e_{n,j},M_{n,j-1},a;\omega_{n,j,a})
\in\tfCsbcCalR\cup\{\bot\}.
\label{tf:csbc:eq:candidate-row}
\end{equation}
Trusted validation and publication choose at most one pair $(a,\widehat r_{n,j,a})$ with $a\in\mathcal A_{n,j}$, or exact \tfCsbcNull{}. The route, target order, and final target decision are system-owned and are bound into the authorization receipt; a model-generated row cannot redirect its own write. Tenant, ACL, versions, protection, provenance, tombstone validity, rollback lineage, and final receipt are likewise system-owned fields, never minted by completed text.

Service progress is independently typed. For idempotence key $k_{n,j}$,
\begin{equation}
\sigma^{\rm svc}_{n,j}\in
\{\textsc{Detected},\textsc{Owned},\textsc{Queued},\textsc{Running},
\textsc{Retryable},\textsc{Commit},\textsc{Null},\textsc{Failed-Terminal}\}.
\label{tf:csbc:eq:service-state}
\end{equation}
The last three values are terminal receipts. \textsc{Commit} names one authorized row replacement, \textsc{Null} names an exact no-op, and \textsc{Failed-Terminal} records fail-closed exhaustion or unrecoverable integrity failure and therefore authorizes no mutation.

Wall-clock service uses a bounded queue $\mathcal Q^{\rm svc}_\kappa\in\tfCsbcCalK$, worker records $W_\kappa^{\rm svc}$, pinned predecessor versions, retry records, and a commit ledger $\mathcal L^M$. Training-only objects live in a disjoint type $\tfCsbcCalT$: future branches $F_{g,k}$, labels $Y^{\rm T}$, teacher risks, gradients, split identifiers, and supervision time $s$. No value in $\tfCsbcCalT$ is a deployment input.

\begin{table}[H]
\centering
\scriptsize
\begin{tabularx}{\linewidth}{@{}p{0.16\linewidth}p{0.19\linewidth}Y Y@{}}
\toprule
Object & Type & Role & Mandatory boundary \\
\midrule
$X_t,L_t,\xi_t,\rho_t^X$ & $\tfCsbcCalX,\tfCsbcR^{|\tfCsbcCalX|},\tfCsbcCalU,\tfCsbcCalR_{\rm emit}$ & causal emission and immutable receipt & fixed before future token exists \\
$I_n,O_n,V_n$ & finite spans / $\tfCsbcCalV$ & core, overlap, completed view & bounded; $V_n$ only after $b_n$ \\
$\beta_n,q_n$ & $\tfCsbcCalB,\tfCsbcCalQ$ & boundary rule and carry & stopping-measurable; $|q_n|\le B_Q$ \\
$E_n,e_{n,j}$ & $\tfCsbcCalE^{\le K_{\max}},\tfCsbcCalE$ & ordered events & typed span, digest, anchor, epoch, sources, key, and route \\
$\alpha(e),\iota(e)$ & token index, digest & ownership and retry identity & stable across overlap and retry \\
$\widehat r_{n,j,a},\tfCsbcNull,M^{\nu}$ & row/action/memory & target-conditioned candidate, identity action, authoritative state & trusted assembly; atomic publication \\
$\theta,\psi$ & deployment parameters & generator and retrospective encoder & frozen before evaluated episode \\
$F,Y^{\rm T},s$ & $\tfCsbcCalT$ & training-only teacher/supervision & amputated from deployment graph \\
$\mathcal Q^{\rm svc},\mathcal L,\sigma^{\rm svc}$ & bounded queue / ledger / finite state & async work and terminal receipts & logical order independent of finish order \\
$\mathbf C$ & $\tfCsbcR_+^d$ & resource vector & no hidden scalarization \\
\bottomrule
\end{tabularx}
\caption{Typed carrier inventory. Co-location in one process does not erase causal, authority, lifetime, or accounting distinctions.}
\label{tf:csbc:tab:typed-inventory}
\end{table}

\subsection{Core assumption ledger}

\begin{tfCsbcAssumption}[Causal generation and immutable history]
\label{tf:csbc:ass:immutable}
Equations~\eqref{tf:csbc:eq:logit}--\eqref{tf:csbc:eq:sample} are $\tfCsbcCalF_t^X$-measurable. Once $\rho_t^X$ is sealed, $L_t,X_t,\xi_t,\nu_t$, and every earlier receipt are append-only and cannot be modified by the consolidator, queue, committer, or recovery service.
\end{tfCsbcAssumption}

\begin{tfCsbcAssumption}[Post-closure construction and exposure]
\label{tf:csbc:ass:exposure}
$V_n$ is unavailable before $b_n$ closes. Every event successor has a logical exposure token $a_{n,j}>b_n$. A token read pins one complete version before generation; a successor becomes eligible only at or after its exposure point.
\end{tfCsbcAssumption}

\begin{tfCsbcAssumption}[Bounded deterministic interpretation]
\label{tf:csbc:ass:bounded-deterministic}
$C_n\le C_{\max}$, $|O_n|\le O_{\max}$, $|q_n|\le B_Q$, $K_n\le K_{\max}$, and $R_{n,j}\le R_{\max}$. Given the same completed view, carry, implementation/version identifiers, and random coins, encoding, eventization, anchor selection, ordering, routing, target ordering, construction, and idempotence keys are identical.
\end{tfCsbcAssumption}

\begin{tfCsbcAssumption}[Trusted atomic authority]
\label{tf:csbc:ass:authority}
Each ordered event validates against its actual predecessor, changes at most one routed complete row, or performs bit-identical \tfCsbcNull{}. A commit authorization binds the event key, ownership epoch, routed-target digest, chosen target, base version, old-row digest, candidate digest, and terminal outcome. A \textsc{Failed-Terminal} receipt binds the same event identity but grants no mutation authority. Publication is serializable and crash recovery exposes an old or new complete authoritative version with its ledger binding.
\end{tfCsbcAssumption}

These assumptions are formal premises, not implementation findings.

%% file: theory_family/modules/csbc/sections/03_five_clocks.tex
\section{Five Clocks and the Logical Schedule}
\label{tf:csbc:sec:clocks}

\subsection{Clock types}

\tfCsbcDefinitionStatus
\begin{tfCsbcDefinition}[Five-clock schedule]
A \tfCsbcName{} trace distinguishes:
\begin{enumerate}
  \item token time $t\in\tfCsbcN$, ordering logits, samples, emissions, and read receipts;
  \item boundary time $n\in\tfCsbcN$, ordering completed cores $I_n$;
  \item logical event time $(n,j)$ in lexicographic order, $1\le j\le K_n$;
  \item training-supervision time $s\in\tfCsbcN$, ordering offline teacher labels and parameter updates; and
  \item wall-clock service time $\kappa\in\tfCsbcR_+$, recording enqueue, worker start/finish, validation, durable publication, and exposure latency.
\end{enumerate}
Only the first three order deployment semantics. Clock $s$ is outside the evaluated episode; clock $\kappa$ determines availability and cost but never silently reorders events.
\end{tfCsbcDefinition}

\begin{figure}[H]
\centering
\includegraphics[width=\linewidth]{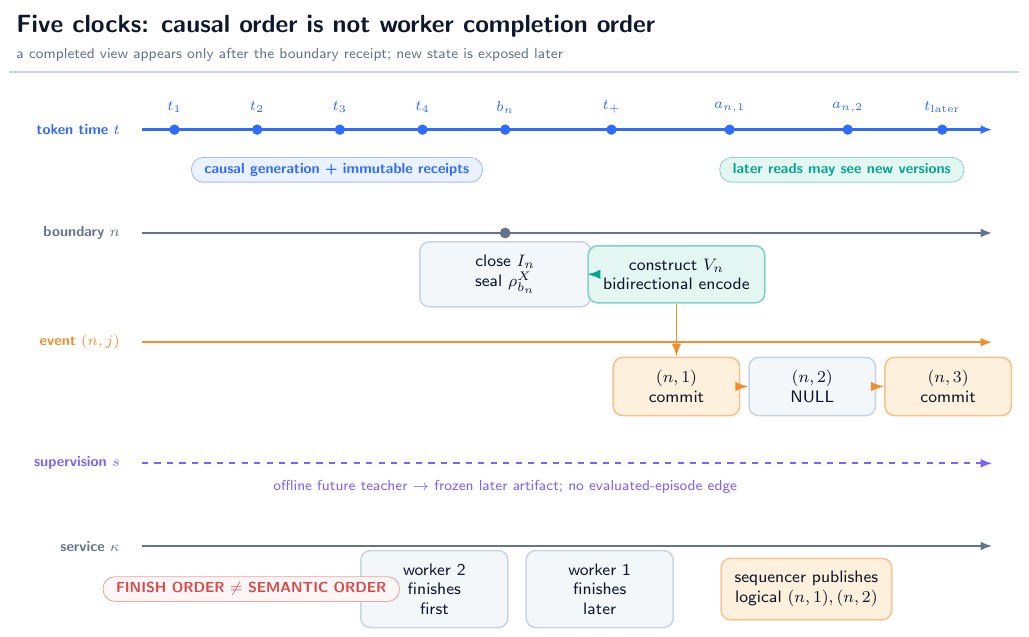}
\caption{\textbf{Five aligned clocks and the post-closure boundary (\tfCsbcDefinitionStatus{}).} Solid token and event arrows are deployed logical order. The supervision rail is training-only and dashed. Wall-clock workers may finish out of order, but the commit sequencer and exposure receipts preserve lexicographic event order.}
\label{tf:csbc:fig:five-clocks}
\end{figure}

For service job $J_{n,j}$ write
\begin{equation}
\kappa^{\rm enq}_{n,j}\le\kappa^{\rm start}_{n,j}
\le\kappa^{\rm done}_{n,j}\le\kappa^{\rm pub}_{n,j}\le\kappa^{\rm exp}_{n,j}.
\label{tf:csbc:eq:service-times}
\end{equation}
The inequalities are local to one job. It is permitted that
$\kappa^{\rm done}_{n,2}<\kappa^{\rm done}_{n,1}$; publication must still respect logical dependencies or reject/recompute the stale result.

\subsection{Logical precedence relation}

Let $\prec$ be the transitive closure of:
\begin{align}
(t,\mathsf{emit})&\prec(n,\mathsf{close}) &&\text{for }t\le b_n,\nonumber\\
(n,\mathsf{close})&\prec(n,\mathsf{view})\prec(n,\mathsf{encode}),\nonumber\\
(n,\mathsf{encode})&\prec(n,1)\prec\cdots\prec(n,K_n),\nonumber\\
(n,j)&\prec(a_{n,j},\mathsf{eligible\mbox{-}read}).
\label{tf:csbc:eq:precedence}
\end{align}
Training events have their own $\prec_s$ and may affect a later frozen parameter artifact, but they are not inserted into the deployment trace for an evaluated episode.

\tfCsbcProvedStatus
\begin{tfCsbcProposition}[Worker-order irrelevance]
\label{tf:csbc:prop:worker-order}
Under Assumptions~\ref{tf:csbc:ass:bounded-deterministic} and~\ref{tf:csbc:ass:authority}, two executions with identical logical inputs, coins, and accepted event sequence but different worker start/finish orders expose the same authoritative version sequence, provided every stale worker result is validated against its pinned predecessor and the sequencer publishes only in lexicographic event order.
\end{tfCsbcProposition}

\begin{proof}
For event $(n,1)$, deterministic construction on the same pinned predecessor yields the same candidate; validation either produces the same complete commit or \tfCsbcNull{}. Suppose authoritative versions agree through event $(n,j-1)$. Only a candidate whose recorded predecessor equals that common version can publish at $(n,j)$; a result computed from another predecessor is rejected or recomputed. Determinism then gives the same event outcome. Induction over lexicographic event time proves identical version sequences. Worker timing can change waiting time and retry cost, not the accepted semantic order.
\end{proof}

Without predecessor validation, a fast worker can overwrite the effect of an earlier logical event. That is a lost-update bug, not a legitimate alternative \tfCsbcName{} semantics.

\subsection{Exposure policy}

The deployment fixes one of two policies:
\begin{itemize}
  \item \textbf{stall}: token $t\ge a_{n,j}$ waits until the required version is published or a bounded timeout selects a registered fallback;
  \item \textbf{prior-state continuation}: generation continues on a previously pinned version, and the new version is first exposed at the next declared pin point after publication.
\end{itemize}
The policy may depend on causal queue metadata fixed before token sampling. A run may not retrospectively relabel earlier reads as though a late successor had been available.

%% file: theory_family/modules/csbc/sections/04_causal_boundary.tex
\section{Causal Boundary, Exposure, and Noninterference}
\label{tf:csbc:sec:causal-boundary}

\subsection{Deployment filtration and DAG}

Let $\tfCsbcCalD_t$ be the deployment sigma-field generated by emitted tokens and receipts through $t-1$, the causal workspace, declared coins revealed through $t$, and authoritative versions whose exposure points do not exceed $t$. The token kernel is $\tfCsbcCalD_t$-measurable. For boundary $n$, the completed view and retrospective state are measurable only after closure:
\begin{equation}
V_n\in\tfCsbcCalF_{b_n}^{X,+},\qquad
H_n=B_\psi(V_n)\in\tfCsbcCalF_{b_n}^{X,+},
\label{tf:csbc:eq:view-measurable}
\end{equation}
where $\tfCsbcCalF_{b_n}^{X,+}$ includes the sealed boundary receipt but no unobserved suffix.

\begin{figure}[H]
\centering
\includegraphics[width=\linewidth]{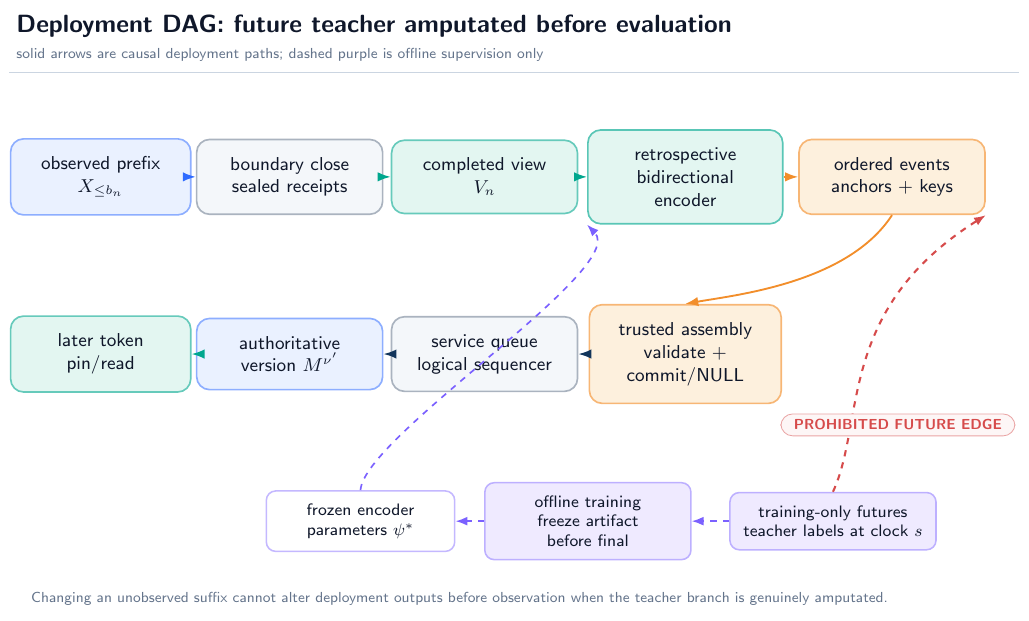}
\caption{\textbf{Deployment DAG and amputated teacher (\tfCsbcDefinitionStatus{} / \tfCsbcPrescriptionStatus{}).} Solid arrows terminate at post-closure events and later exposure. The dashed future-teacher branch may update a later frozen artifact offline, but it has no executable edge into the evaluated deployment episode. Coral edges are prohibited leakage.}
\label{tf:csbc:fig:deployment-dag}
\end{figure}

\begin{figure}[H]
\centering
\includegraphics[width=\linewidth]{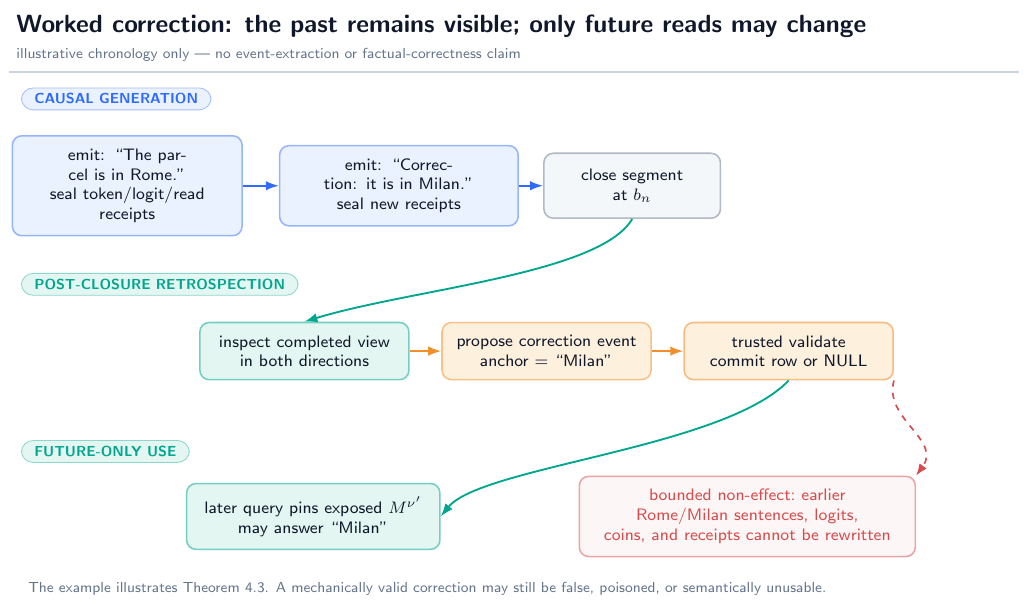}
\caption{\textbf{Worked correction timeline (illustrative definition, not a semantic-success result).} A causal generator first emits an incorrect location and later a correction inside the same completed segment. Retrospective encoding starts only after closure. A validated correction row may help a later question; it cannot edit either earlier sentence or its receipt.}
\label{tf:csbc:fig:worked-timeline}
\end{figure}

\subsection{No retroactive effect}

Let the historical prefix object at boundary $b_n$ be
\begin{equation}
\mathsf H_{b_n}=\bigl((L_t,\xi_t,X_t,\nu_t,\rho_t^X)\bigr)_{t=1}^{b_n}.
\label{tf:csbc:eq:history-object}
\end{equation}

\tfCsbcProvedStatus
\begin{tfCsbcTheorem}[No retroactive effect]
\label{tf:csbc:thm:no-retro}
Under Assumptions~\ref{tf:csbc:ass:immutable} and~\ref{tf:csbc:ass:exposure}, for every boundary computation, event construction, validation, commit, retry, crash recovery, and later exposure derived from $V_n$,
\begin{equation}
\mathsf H_{b_n}^{\rm after}=\mathsf H_{b_n}^{\rm before}.
\label{tf:csbc:eq:no-retro}
\end{equation}
In particular, no resulting state can change an already emitted token, historical logit, sampling coin, emission receipt, or prior memory-read version. Its earliest possible causal effect is on a read at $t\ge a_{n,j}>b_n$.
\end{tfCsbcTheorem}

\begin{proof}
Assumption~\ref{tf:csbc:ass:immutable} makes every coordinate of $\mathsf H_{b_n}$ append-only and excludes it from the codomain of consolidation, commit, retry, and recovery maps. Those maps may append service and memory receipts and may create a successor $M^{\nu'}$. Assumption~\ref{tf:csbc:ass:exposure} excludes $\nu'$ from every read at $t\le b_n$ and at every later token before its declared exposure. Therefore no transition induced by $V_n$ has a write edge into $\mathsf H_{b_n}$ or a read edge into an earlier generation event. Equality~\eqref{tf:csbc:eq:no-retro} and the earliest-effect statement follow structurally.
\end{proof}

The theorem is not evidence that an implementation really uses immutable logs or correct pinning. Historical-hash and read-receipt tests in Section~\ref{tf:csbc:sec:falsification} must discharge those premises.

\subsection{Prefix twins}

Two executions $\omega,\omega'$ are $u$-prefix twins under a specified counterfactual coupling if they are defined on one probability space, share implementation artifacts and initial authoritative state, and share every exogenous base variable and random coin exposed before the first differing suffix observation. Their tokens $X_{\le u}$ and emission receipts are byte-identical. Boundary-rule state, causal workspace, queue state, and routing state are not separately stipulated equal: each must be a causal function of those shared variables and hence must evaluate identically. Only variables causally downstream of suffix tokens after $u$ may differ.

\tfCsbcProvedStatus
\begin{tfCsbcTheorem}[Prefix-twin noninterference]
\label{tf:csbc:thm:prefix-twin}
Under Assumptions~\ref{tf:csbc:ass:immutable}--\ref{tf:csbc:ass:bounded-deterministic}, if $\omega$ and $\omega'$ are $u$-prefix twins under the coupling above, then every boundary decision, completed view, event list, idempotence key, routed-target set and order, target-conditioned candidate, validation decision, service transition, and exposure decision produced before either execution observes a token after $u$ is identical in the two runs.
\end{tfCsbcTheorem}

\begin{proof}
Before a suffix token is observed, the coupling gives both runs the same deployment sigma-field. Causality of the boundary detector, carry update, queue, and route maps makes their internal states equal rather than assuming that equality as an extra premise. Any fixed boundary rule or adaptive stopping decision is measurable with respect to that field, so both close or continue identically. If a boundary closes, its core, overlap, carry, view, implementation identifiers, and revealed coins agree. Assumption~\ref{tf:csbc:ass:bounded-deterministic} therefore gives identical encoding, events, anchors, keys, ordering, routes, target orders, and target-conditioned candidates. Validation reads the same predecessor state and authorization inputs; the service transition and exposure maps read the same causal ledger and queue state. Their outputs agree. Induct over decisions until one run first observes a differing suffix token.
\end{proof}

This is a hyperproperty across paired executions. A single apparently causal trace cannot establish it.

\begin{figure}[H]
\centering
\includegraphics[width=\linewidth]{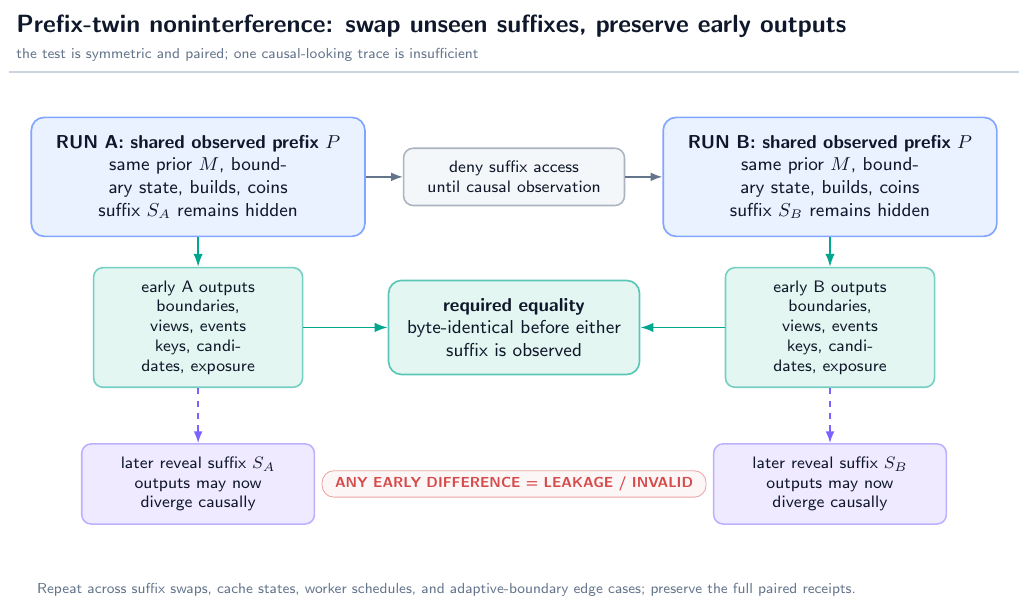}
\caption{\textbf{Symmetric suffix-swap audit for Theorem~\ref{tf:csbc:thm:prefix-twin} (\tfCsbcPlannedStatus{}).} Matched prefixes, coins, state, and builds must yield byte-identical pre-suffix boundaries and outputs despite different hidden continuations. Any early difference is a bounded leakage signal and invalidates the run.}
\label{tf:csbc:fig:prefix-twins}
\end{figure}

%% file: theory_family/modules/csbc/sections/05_training_amputation.tex
\section{Training-Supervision Separation and Teacher Amputation}
\label{tf:csbc:sec:amputation}

\subsection{Offline supervision is not a deployment input}

Training group $g$ may contain a causal prefix $P_g$, completed view $V_g$, and alternative future continuations $F_{g,1:K}$. A teacher may use those futures to create a label
\begin{equation}
Y_g^{\rm T}=T(V_g,F_{g,1:K},\chi_g^{\rm eval})
\label{tf:csbc:eq:teacher-label}
\end{equation}
at supervision time $s$. Training produces a parameter artifact
\begin{equation}
(\psi^*,\theta^*)=\mathsf{Train}\bigl(\{(V_g,Y_g^{\rm T})\}_{g\in\mathcal G_{\rm train}};\eta\bigr).
\label{tf:csbc:eq:training}
\end{equation}
Before final episodes are opened, the artifact, feature schema, preprocessing graph, thresholds, random-seed policy, and dependency manifest are frozen and content addressed.

\tfCsbcDefinitionStatus
\begin{tfCsbcDefinition}[Teacher-branch amputation]
A deployment artifact is teacher-amputated when its executable transitive dependency graph contains no future continuation, teacher label, evaluator output, branch identifier, post-action state, final-split statistic, or cache derived from those objects for the evaluated episode. Only the frozen parameters and declared preprocessing artifacts may cross from supervision time into deployment.
\end{tfCsbcDefinition}

\begin{tfCsbcAssumption}[Split and episode independence]
\label{tf:csbc:ass:split}
Training, calibration, and final episodes are separated at the physical episode/source-family level. All suffixes, paraphrases, cached features, event IDs, and teacher derivatives belonging to one episode family remain in one split. Final parameter freeze precedes any access to final suffixes or outcomes.
\end{tfCsbcAssumption}

\begin{tfCsbcAssumption}[Executable graph closure]
\label{tf:csbc:ass:graph-closure}
The deployment dependency manifest is complete: it includes data loaders, feature services, retrieval stores, caches, environment variables, model files, tokenizer/boundary code, RNG initialization, and remote calls. Runtime access logging is faithful to that graph.
\end{tfCsbcAssumption}

\tfCsbcProvedStatus
\begin{tfCsbcTheorem}[Teacher-amputation nonaccess]
\label{tf:csbc:thm:amputation}
Under Assumptions~\ref{tf:csbc:ass:split} and~\ref{tf:csbc:ass:graph-closure}, if the frozen deployment graph contains no path from an evaluated episode's future-teacher objects to its boundary, encoding, eventization, validation, or exposure nodes, then changing only those future-teacher objects cannot change any deployment output before the corresponding suffix becomes causally observed.
\end{tfCsbcTheorem}

\begin{proof}
Every deployment output is a deterministic or recorded-random function of its ancestors in the complete executable graph. By premise, none of the changed teacher objects is an ancestor. All unchanged ancestors, including frozen parameters and coins, therefore induce the same output. When a suffix token is later observed, it becomes a causal token input and the theorem no longer requires equality beyond that point.
\end{proof}

This is a graph-separation deduction. It does not prove the manifest is complete. The practical evidence is a static dependency closure plus runtime deny/access receipts and the prefix-twin audit.

\subsection{Leakage receipts}

\tfCsbcPrescriptionStatus\quad A conforming run records:
\begin{enumerate}
  \item immutable train/calibration/final group manifests and source-family closure;
  \item parameter, tokenizer, boundary, encoder, eventizer, validator, and threshold hashes with freeze time;
  \item a field-level provenance graph for every deployment input;
  \item filesystem, database, network, cache, and feature-service access logs during each final episode;
  \item teacher-label and future-branch namespaces denied to deployment credentials;
  \item group-preserving suffix swaps, label permutations, and matched-prefix twins; and
  \item a final-entry receipt proving the evaluation namespace was opened once by the frozen artifact.
\end{enumerate}

\begin{tfCsbcCounterexample}[Parameter freeze after final suffix access]
A threshold is tuned after inspecting which final suffixes contain corrections. The executable model never reads a suffix at runtime, yet the threshold encodes final information. Runtime graph amputation alone cannot repair the predeployment leak; Assumption~\ref{tf:csbc:ass:split} fails.
\end{tfCsbcCounterexample}

\begin{tfCsbcCounterexample}[Teacher-derived cache]
A feature store is precomputed using the full episode and later serves only a short numeric vector. The vector's type says ``prefix feature,'' but its provenance includes the suffix. Without transitive cache provenance, a direct field audit misses the leak.
\end{tfCsbcCounterexample}

\tfCsbcUnvalidatedStatus\quad No CSBC teacher, parameter artifact, split, or deployment graph is built or tested in this theoretical part. Training gradients through a completed segment are offline training operations; they do not imply online parameter gradients during deployment.

%% file: theory_family/modules/csbc/sections/06_boundaries_carry.tex
\section{Fixed and Adaptive Boundaries, Overlap, and Carry}
\label{tf:csbc:sec:boundaries}

\subsection{Fixed cores}

For a fixed core size $C\ge1$, set $b_n=\min\{nC,N\}$ for a finite episode of $N$ tokens and
\begin{equation}
I_n=(b_{n-1},b_n],\qquad
O_n=(\max\{0,b_{n-1}-O\},b_{n-1}],\qquad 0\le O\le O_{\max}.
\label{tf:csbc:eq:fixed-boundary}
\end{equation}
Overlap tokens are read-only evidence in view $V_n$. They do not re-enter generation and do not acquire new ownership merely because they appear twice.

\subsection{Adaptive stopping boundaries}

Let $C_{\min}\le C_{\max}$. The detector has a finite carrier $\mathcal D$ with a fixed encoding cap $B_D$ and a causal recurrence
\begin{equation}
d_t=D(d_{t-1},X_t),\qquad d_t\in\mathcal D,
\qquad \operatorname{bytes}(d_t)\le B_D.
\label{tf:csbc:eq:detector-recurrence}
\end{equation}
At a boundary, the segment-local detector is reset to $d_{\varnothing}$; any permitted continuation crosses the boundary only through the typed carry $q_n$ in Eq.~\eqref{tf:csbc:eq:carry-record}. Thus neither an opaque recurrent state nor an unbounded token buffer can bypass the carry cap.

Starting after $b_{n-1}$, set $\tau_n=\min\{N,b_{n-1}+C_{\max}\}$. The adaptive rule observes only the causal prefix and chooses
\begin{equation}
b_n=
\begin{cases}
N, & N-b_{n-1}<C_{\min},\\[2pt]
\inf\left\{t\in[b_{n-1}+C_{\min},\tau_n]:
h(X_{\le t},d_t)=1\ \text{or}\ t=\tau_n\right\}, & \text{otherwise}.
\end{cases}
\label{tf:csbc:eq:adaptive-boundary}
\end{equation}
The first branch is a declared terminal flush: it may create a final core shorter than $C_{\min}$ but never longer than $C_{\max}$.

\begin{tfCsbcAssumption}[Stopping-time boundary]
\label{tf:csbc:ass:stopping}
For every $t$, the event $\{b_n\le t\}$ is measurable with respect to $\tfCsbcCalF_t^X$. The rule uses no future token, suffix label, retrospective encoder output, or worker completion time. It always closes by $C_{\max}$ core tokens.
\end{tfCsbcAssumption}

Thus the boundary detector may react to an observed delimiter or punctuation, but it cannot wait to see whether the next token makes the current span convenient. Worst-case token closure delay is $C_{\max}$ tokens after the previous boundary; wall-clock closure adds the observed token-generation time.

\subsection{Cross-boundary event state}

An event may begin near the end of $I_n$ and close in $I_{n+1}$. The detector then emits no complete event in segment $n$ and serializes a carry record
\begin{equation}
q_n=(\mathsf{episode},\mathsf{event\mbox{-}type},\mathsf{start},
\mathsf{bounded\mbox{-}summary},\mathsf{source\mbox{-}digests},\mathsf{age})
\in\tfCsbcCalQ.
\label{tf:csbc:eq:carry-record}
\end{equation}
The carry cannot contain an unbounded token string. If an event remains open beyond $H_Q$ boundaries or needs more than $B_Q$ bytes, the detector emits a typed overflow/failure receipt and drops, truncates only a schema-declared nonsemantic diagnostic, or routes to a separately bounded fallback. It may not silently grow.

When the event closes, its anchor is a canonical token position belonging to exactly one core, such as the first token of the terminal predicate or the final token required by the event schema. The owner is
\begin{equation}
\tfCsbcOwner(e)=n\quad\Longleftrightarrow\quad\alpha(e)\in I_n.
\label{tf:csbc:eq:core-owner}
\end{equation}
Overlap and carry provide evidence but never override Eq.~\eqref{tf:csbc:eq:core-owner}.

\begin{figure}[H]
\centering
\includegraphics[width=\linewidth]{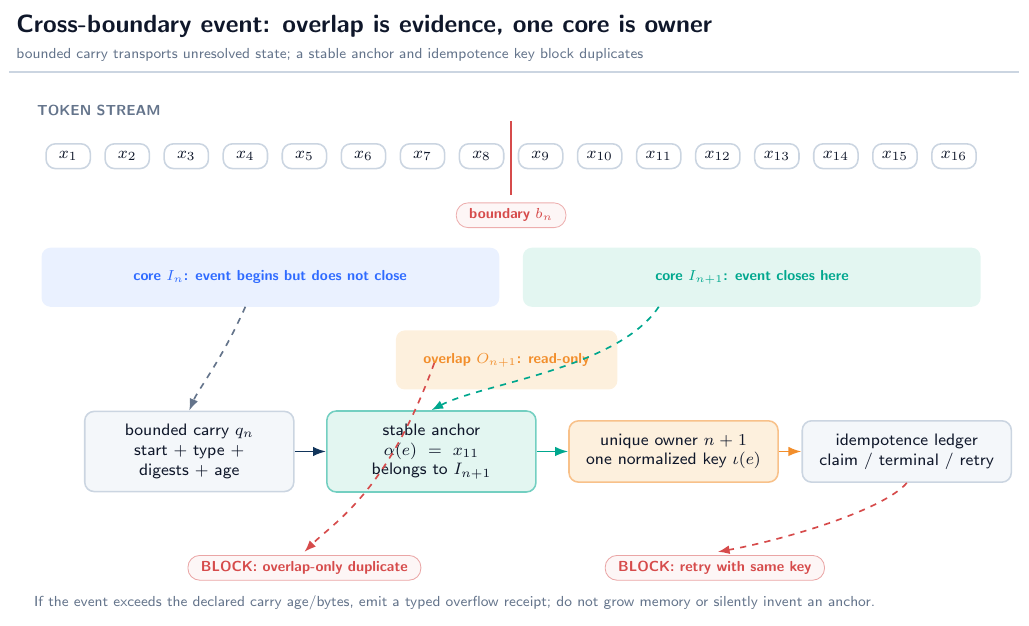}
\caption{\textbf{Cross-boundary eventization (\tfCsbcDefinitionStatus{} / \tfCsbcPrescriptionStatus{}).} A bounded carry transports an unfinished event; overlap supplies read-only context; the stable anchor lands in one core, which alone owns emission. The idempotence ledger blocks a retry or overlapping view from publishing a duplicate.}
\label{tf:csbc:fig:cross-boundary}
\end{figure}

\begin{tfCsbcAssumption}[Bounded closure and carry]
\label{tf:csbc:ass:carry}
Every carried detector state is at most $B_Q$ bytes and at most $H_Q$ boundaries old; the segment-local state additionally obeys Eq.~\eqref{tf:csbc:eq:detector-recurrence} and is reset at closure. For a complete event in scope, the declared detector either identifies a stable anchor after finite observed evidence or emits a typed unresolved/failure outcome by the cap. No theorem assumes that every real-world relation fits the cap.
\end{tfCsbcAssumption}

\tfCsbcProvedStatus
\begin{tfCsbcProposition}[Finite view and coverage]
\label{tf:csbc:prop:view-bound}
Under fixed boundaries or Assumption~\ref{tf:csbc:ass:stopping}, the episode has
\begin{equation}
\left\lceil\frac{N}{C_{\max}}\right\rceil
\le B_N\le
\left\lceil\frac{N}{C_{\min}}\right\rceil
\label{tf:csbc:eq:segment-count}
\end{equation}
adaptive cores, with the fixed case $B_N=\lceil N/C\rceil$. Every core token belongs to exactly one $I_n$, every completed view contains at most $C_{\max}+O_{\max}$ tokens plus $B_Q$ carry bytes, and the total number of token positions presented to the retrospective encoder is at most
\begin{equation}
N+O_{\max}B_N.
\label{tf:csbc:eq:overlap-count}
\end{equation}
\end{tfCsbcProposition}

\begin{proof}
Every adaptive core has at most $C_{\max}$ tokens, so at least $\lceil N/C_{\max}\rceil$ cores are required. Every nonfinal core has at least $C_{\min}$ tokens, while the declared terminal flush has between one and $C_{\max}$; hence
$B_N\le 1+\lfloor(N-1)/C_{\min}\rfloor=\lceil N/C_{\min}\rceil$.
This proves Eq.~\eqref{tf:csbc:eq:segment-count} without incorrectly imposing the minimum on the final core. The half-open intervals $(b_{n-1},b_n]$ are disjoint and cover $1{:}N$. By construction, overlap adds at most $O_{\max}$ token positions to each view and carry has the fixed byte cap. Summing core positions gives $N$ and summing overlap bounds gives Eq.~\eqref{tf:csbc:eq:overlap-count}.
\end{proof}

The upper bound counts repeated positions, not unique tokens. Hidden archive retrieval, a variable-size carry, or unconstrained backtracking is outside its premises.

%% file: theory_family/modules/csbc/sections/07_ownership_idempotence.tex
\section{Unique Ownership, Idempotence, and Terminal Receipts}
\label{tf:csbc:sec:ownership}

\subsection{Detector and ownership contract}

Let $\mathcal E^*(X)$ be the finite set of in-scope completed events defined by a frozen annotation/event schema on an observed episode. The eventizer is not assumed semantically correct. The following premises state what must hold for an at-most-once authority claim and an exactly-one service-terminal-receipt claim.

\begin{tfCsbcAssumption}[Detector consistency]
\label{tf:csbc:ass:detector-consistency}
For fixed episode bytes, boundary/carry state, implementation artifacts, and coins, every invocation that sees enough evidence for the same in-scope event emits the same normalized event content, stable anchor, and idempotence key. It emits no complete key while the event remains unresolved.
\end{tfCsbcAssumption}

\begin{tfCsbcAssumption}[Unique core ownership]
\label{tf:csbc:ass:unique-owner}
Every emitted event has exactly one anchor $\alpha(e)\in\bigcup_n I_n$, and only the event list of $\tfCsbcOwner(e)$ may submit its key for publication. An overlap-only or carry-only occurrence is non-owning.
\end{tfCsbcAssumption}

\begin{tfCsbcAssumption}[Linearizable idempotence ledger]
\label{tf:csbc:ass:idempotence}
The commit service maintains a bounded-scope ledger mapping $(\mathsf{episode},\iota)$ to \textsc{Absent}; one nonterminal state in $\{\textsc{Detected},\textsc{Owned},\textsc{Queued},\textsc{Running},\textsc{Retryable}\}$; or one terminal receipt in $\{(\textsc{Commit},\nu),\textsc{Null},\textsc{Failed-Terminal}\}$. Claiming an absent key, advancing a service state, and recording its terminal receipt are linearizable with the authoritative commit protocol. A retry reads or resumes the same key; it never creates a fresh identity. \textsc{Failed-Terminal} is fail-closed and cannot mutate memory.
\end{tfCsbcAssumption}

The ledger scope and retention must cover every permitted retry and overlap replay. A finite ledger cannot promise global uniqueness after its key has been forgotten; the guarantee is explicitly scoped to the episode/retry horizon.

\tfCsbcProvedStatus
\begin{tfCsbcTheorem}[At-most-once authority and exactly-one terminal service result]
\label{tf:csbc:thm:exactly-once}
Under Assumptions~\ref{tf:csbc:ass:carry}, \ref{tf:csbc:ass:detector-consistency}, \ref{tf:csbc:ass:unique-owner}, and~\ref{tf:csbc:ass:idempotence}, every in-scope event that the detector completes within its declared cap has at most one authoritative outcome: one complete-row commit or exact \tfCsbcNull{}. If the owning job and ledger remain live or recoverable until a terminal receipt is recorded, it has exactly one service-terminal receipt in $\{\textsc{Commit},\textsc{Null},\textsc{Failed-Terminal}\}$. The failure receipt is not an authoritative update.
\end{tfCsbcTheorem}

\begin{proof}
Detector consistency gives one stable key for the completed event. Unique ownership permits only one core to submit that key, although retries may resubmit it. By linearizability, at most one submission changes the ledger from \textsc{Absent}; every other invocation observes the same nonterminal chain or terminal receipt. Trusted atomic authority permits at most one routed row commit or exact \tfCsbcNull{} for that key. A \textsc{Failed-Terminal} transition grants no write capability, so it cannot create a second authoritative outcome. For liveness, the additional premise ensures the owning job is eventually executed, recovered, or closed fail-safe and cannot remain nonterminal forever. Linearizability then permits exactly one terminal receipt.
\end{proof}

The theorem says nothing about factual correctness, event recall, target choice, or semantic usefulness. A consistently wrong detector can satisfy it.

\subsection{Idempotent retry protocol}

\begin{tfCsbcProtocol}[Claim, construct, publish, recover]
\label{tf:csbc:prot:idempotent}
For owning event $e_{n,j}$:
\begin{enumerate}
  \item atomically claim $(\mathsf{episode},\iota(e_{n,j}))$ or read its existing record;
  \item bind the logical event index, anchor, completed-view digest, predecessor memory version, ownership epoch, builder/validator versions, routed-target digest, and authorization context;
  \item construct target-conditioned candidates against the pinned predecessor, replay and validate every protected field and cross-row invariant, and select at most one routed target;
  \item publish one complete row or exact \tfCsbcNull{} in logical order and bind the terminal state to the same key, target, base version, old-row digest, and candidate digest;
  \item on crash, recover the old or new complete authoritative root and reconcile the terminal key; and
  \item on retry, return the terminal result or resume the same in-flight transaction. Never mint a new key from attempt number, wall time, or worker ID; and
  \item after bounded retry/rebuild exhaustion or unrecoverable integrity failure, append \textsc{Failed-Terminal} with no mutation. Any authorized recovery is a new logical event with a new key explicitly rooted in that failure receipt; the failed key is never reused.
\end{enumerate}
\end{tfCsbcProtocol}

\subsection{Multiple events and duplicate suppression}

One segment may contain $K_n>1$ distinct events. Distinct stable keys and the event order separate them. Conversely, two surface mentions may map to one event only if the frozen schema says they share one normalized identity and anchor. Deduplication after seeing downstream utility would change the estimand and is prohibited.

\tfCsbcPrescriptionStatus\quad The event manifest records all emitted, deferred, overflowed, rejected, duplicated, and missing outcomes. A missing key is never silently imputed as \tfCsbcNull{}: \tfCsbcNull{} is a terminal authoritative decision after a valid event reached the commit protocol.

%% file: theory_family/modules/csbc/sections/08_event_recursion.tex
\section{Event-Recursive Authority, Serializability, and Commutation}
\label{tf:csbc:sec:event-recursion}

\subsection{Ordered state recursion}

For completed segment $n$, set
\begin{equation}
M_{n,0}=M_{n-1},\qquad
M_{n,j}=T_{n,j}(M_{n,j-1}),\qquad
M_n=M_{n,K_n}.
\label{tf:csbc:eq:event-recursion}
\end{equation}
If trusted validation accepts the routed pair
$(a_{n,j}^{M},\widehat r_{n,j,a_{n,j}^{M}})$ with
$a_{n,j}^{M}\in\mathcal A_{n,j}\subseteq\{1,\ldots,S\}$,
\begin{equation}
T_{n,j}(M)_i=
\begin{cases}
\widehat r_{n,j,a_{n,j}^{M}},&i=a_{n,j}^{M},\\
M_i,&i\ne a_{n,j}^{M};
\end{cases}
\qquad
T_{n,j}^{\tfCsbcNull}(M)=M
\label{tf:csbc:eq:row-transition}
\end{equation}
bit for bit. A later event constructs and validates against $M_{n,j-1}$, not against the pre-segment snapshot unless those states happen to be equal.

\begin{figure}[H]
\centering
\includegraphics[width=\linewidth]{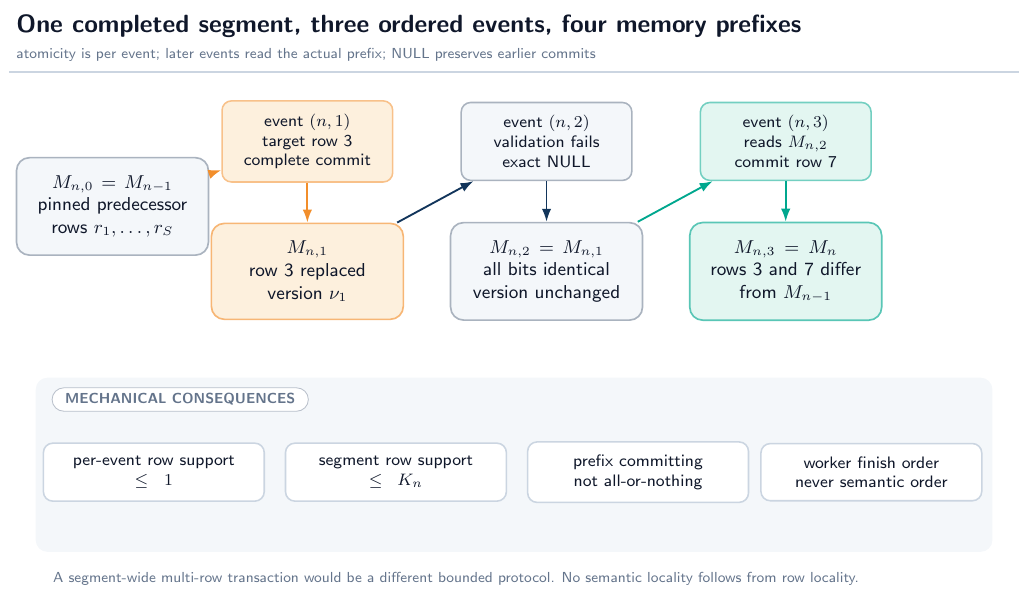}
\caption{\textbf{Multiple events from one completed segment (\tfCsbcDefinitionStatus{}).} Events follow declared logical order and actual predecessor versions. Event 1 commits a complete row, event 2 returns exact \tfCsbcNull{}, and event 3 reads the state after both. Atomicity is per event; the segment is prefix-committing.}
\label{tf:csbc:fig:event-recursion}
\end{figure}

\tfCsbcProvedStatus
\begin{tfCsbcTheorem}[Event locality and segment support]
\label{tf:csbc:thm:event-locality}
Under Assumption~\ref{tf:csbc:ass:authority}, for every event
\begin{equation}
|\tfCsbcSupp(M_{n,j}-M_{n,j-1})|\le1.
\label{tf:csbc:eq:event-support}
\end{equation}
If $C_n^M$ events commit and $D_n$ distinct rows are changed across segment $n$, then
\begin{equation}
C_n^M\le K_n,\qquad
D_n\le\min\{C_n^M,S\},\qquad
|\tfCsbcSupp(M_n-M_{n-1})|\le D_n\le K_n.
\label{tf:csbc:eq:segment-support}
\end{equation}
Earlier successful events survive a later \tfCsbcNull{}.
\end{tfCsbcTheorem}

\begin{proof}
Equation~\eqref{tf:csbc:eq:row-transition} changes one target row and \tfCsbcNull{} changes none, proving Eq.~\eqref{tf:csbc:eq:event-support}. Each of $K_n$ events contributes at most one commit and one target to the union of changed rows. Thus $C_n^M\le K_n$ and $D_n\le\min\{C_n^M,S\}$. Rows outside that union are byte-identical under composition, giving Eq.~\eqref{tf:csbc:eq:segment-support}. Because the recursion takes $M_{n,j-1}$ as the input, applying identity at a later event preserves the already committed prefix.
\end{proof}

The result is not whole-segment atomicity and not whole-segment rank one. A segment-wide transaction would need a separate bounded multi-row read/write set and publication protocol.

\subsection{Serializable semantic order}

\begin{tfCsbcAssumption}[Logical sequencer]
\label{tf:csbc:ass:sequencer}
For one authority domain, the commit service provides a total order extending lexicographic event time. Each candidate records its read set and predecessor versions; a stale candidate is rejected or reconstructed. Global capacity, uniqueness, index, authorization, and version constraints are checked in that same serializable order.
\end{tfCsbcAssumption}

\tfCsbcProvedStatus
\begin{tfCsbcTheorem}[Serializable event history]
\label{tf:csbc:thm:serializable}
Under Assumptions~\ref{tf:csbc:ass:authority} and~\ref{tf:csbc:ass:sequencer}, every finite concurrent service execution is observationally equivalent at the authoritative read interface to the sequential recursion in Eq.~\eqref{tf:csbc:eq:event-recursion}, with stale/rejected attempts represented by typed retry records or \tfCsbcNull{} according to the frozen protocol.
\end{tfCsbcTheorem}

\begin{proof}
Order accepted publications by the sequencer's linearization points. Each publication was validated against the predecessor and global constraints current at its position; a stale attempt has no authoritative effect. Applying the same complete-row or identity transitions in that order reproduces every published root. A validated read observes one published root and can be placed after its linearization point. Therefore the concurrent execution and the sequential recursion have the same authoritative observations.
\end{proof}

\subsection{Conditional commutation}

Let event maps $T_e,T_f$ target distinct rows. Disjoint destinations alone are insufficient: either event can read capacity, eviction rank, global uniqueness, an index generation, a shared version counter, authorization state, or the other row.

\begin{tfCsbcAssumption}[Commutation independence]
\label{tf:csbc:ass:commutation}
$e$ and $f$ target distinct rows; their candidates, read sets, validation outcomes, version allocation, authorization, capacity/eviction decisions, index updates, ledger reservations, and cross-row invariants are unchanged when the other event is applied first. Their idempotence keys are distinct and their audit order remains recorded.
\end{tfCsbcAssumption}

\tfCsbcProvedStatus
\begin{tfCsbcProposition}[Conditional row commutation]
\label{tf:csbc:prop:commutation}
Under Assumption~\ref{tf:csbc:ass:commutation},
\begin{equation}
T_e(T_f(M))=T_f(T_e(M))
\end{equation}
on authoritative memory content. The ordered audit trace need not be byte-identical.
\end{tfCsbcProposition}

\begin{proof}
By premise, each event computes the same complete candidate and decision under either predecessor order. The targets are distinct, so replacing one row leaves the other's replacement coordinate unchanged. Both compositions therefore contain the same new row at each target and the same old rows elsewhere. Audit receipts retain event order, so only the memory-content equality is claimed.
\end{proof}

If both events see one free slot and each plans to consume it, or if one changes an ACL consulted by the other, the premise fails and order is semantic.

%% file: theory_family/modules/csbc/sections/09_async_consistency.tex
\section{Asynchronous Consolidation and Causal Consistency}
\label{tf:csbc:sec:async}

\subsection{Pinned versions while service runs}

At token $t$, generation pins one version $\nu_t$ for the complete token computation. If consolidation for segment $n$ is still queued, prior-state continuation uses the latest exposed predecessor; it never reads a partially built candidate. Under a stall policy, the token is not sampled until the registered required version or timeout/fallback decision is available. Both choices are receipted.

Let $\nu_0$ be a distinguished validated genesis version with $\kappa^{\rm pub}(\nu_0)=-\infty$ and $a(\nu_0)=0$. Let $p(t)$ be the exposure-consistent read selector:
\begin{equation}
p(t)=\max\left(\{\nu_0\}\cup
\{\nu: \kappa^{\rm pub}(\nu)\le\kappa^{\rm pin}_t,
\ a(\nu)\le t,\ \tfCsbcValidate(M^{\nu})=1\}\right).
\label{tf:csbc:eq:read-selector}
\end{equation}
The maximum is therefore defined even before the first successor is published, and it is taken over the logical version order rather than worker completion timestamps.

\begin{tfCsbcAssumption}[Read pin and publication integrity]
\label{tf:csbc:ass:read-pin}
A token reads one validated immutable authoritative root for its whole generation step. Publication exposes no partial candidate; caches and indexes are version-bound hints and are revalidated against that root.
\end{tfCsbcAssumption}

\tfCsbcProvedStatus
\begin{tfCsbcProposition}[Asynchronous causal consistency]
\label{tf:csbc:prop:async-consistency}
Under Assumptions~\ref{tf:csbc:ass:exposure}, \ref{tf:csbc:ass:sequencer}, and~\ref{tf:csbc:ass:read-pin}, asynchronous service can change which eligible version a later token pins and can increase latency, but it cannot expose a successor before its token exposure point, mix versions within one token, or make worker finish order supersede logical event order.
\end{tfCsbcProposition}

\begin{proof}
Equation~\eqref{tf:csbc:eq:read-selector} excludes versions that are unpublished, invalid, or not yet token-eligible. Read pinning fixes the selected root for the token. The sequencer admits versions only in logical order and rejects stale publications. Thus service timing controls when a logically eligible version enters the maximization set but cannot violate exposure, mix roots, or reorder accepted events.
\end{proof}

\subsection{Stale jobs, retry, and crash}

A worker record binds $(n,j)$, $\iota(e)$, view digest, predecessor root, read set, build identifiers, and random coins. If the predecessor is stale at validation, the protocol chooses one frozen action:
\begin{enumerate}
  \item rebuild candidate and validation from the current logical predecessor;
  \item prove the old candidate remains valid under a declared commutation/read-set test; or
  \item return a typed stale \tfCsbcNull{} if the scientific protocol defines rejection rather than retry; or
  \item after a declared retry/rebuild cap or unrecoverable integrity failure, append \textsc{Failed-Terminal} with no authoritative mutation.
\end{enumerate}
Silently publishing the stale candidate is never allowed. Retry, replay, and wait costs remain charged even when the ultimate terminal receipt is \textsc{Null} or \textsc{Failed-Terminal}. A failed-terminal key is never silently resubmitted; an authorized later recovery must be a new logical event explicitly linked to the failure receipt.

Crash recovery follows the complete-row contract: validate the idempotence record, old/new authoritative root, audit binding, archive/index state, and queue lease. A key in \textsc{Detected}, \textsc{Owned}, \textsc{Queued}, \textsc{Running}, or \textsc{Retryable} may be resumed only against the same logical event and predecessor rules. Terminal keys return their existing receipt. Orphan candidate bytes are non-authoritative garbage.

\subsection{Speculative decoding}

Suppose generation speculates tokens $t{:}t+h_{\rm spec}$ against version $\nu$ while a successor $\nu'$ is pending. The deployment chooses:
\begin{itemize}
  \item \textbf{commit-old}: seal speculative tokens with $\nu$ and expose $\nu'$ only afterward;
  \item \textbf{discard-and-regenerate}: before external emission, discard every unsealed speculative token, activation, cache entry, and draft receipt, then regenerate under $\nu'$; or
  \item \textbf{stall}: do not speculate across the required exposure.
\end{itemize}
Once a token receipt is externally sealed, it belongs to immutable history and cannot be rolled back. Calling deletion of an already published token ``speculative rollback'' violates Assumption~\ref{tf:csbc:ass:immutable}.

\begin{tfCsbcCounterexample}[Mixed speculative prefix]
Tokens $t$ and $t+1$ are drafted under $\nu$, then $t+1$ alone is rescored under $\nu'$ while retaining the hidden state produced under $\nu$. The resulting trace has no single pinned read view and does not satisfy any of the three declared policies.
\end{tfCsbcCounterexample}

\tfCsbcPrescriptionStatus\quad Queue leases, cancellation, timeout, fallback, reconstruction, publication, exposure, cache invalidation, and speculative-discard receipts must be included in the causal trace. Missing service records are not evidence of a synchronous implementation.

%% file: theory_family/modules/csbc/sections/10_resource_bounds.tex
\section{Complete Resource and Queue Bounds}
\label{tf:csbc:sec:resources}

\subsection{Resource vector}

For episode execution $u$, report before scalarization
\begin{align}
\mathbf C(u)=(&N_{\rm tok},F_{\rm gen},F_{\rm bi},F_{\rm evt},F_{\rm route},F_{\rm cand},F_{\rm replay},F_{\rm val},F_{\rm commit},
\nonumber\\
&B_{\rm resident},B_{\rm scratch},B_{\rm carry},B_{\rm queue},B_{\rm archive},B_{\rm index},B_{\rm audit},
\nonumber\\
&B_{\rm mem\mbox{-}traffic},B_{\rm net},N_{\rm retry},N_{\rm failed\mbox{-}terminal},N_{\rm barrier},T_{\rm latency},T_{\rm backlog},E_{\rm proxy}).
\label{tf:csbc:eq:resource-vector}
\end{align}
Units, measurement method, caching, batching, failure work, and amortization horizon are explicit. A scalar $\lambda^\top\mathbf C$ requires a preregistered nonnegative $\lambda$ with compatible units.

\begin{figure}[H]
\centering
\includegraphics[width=\linewidth]{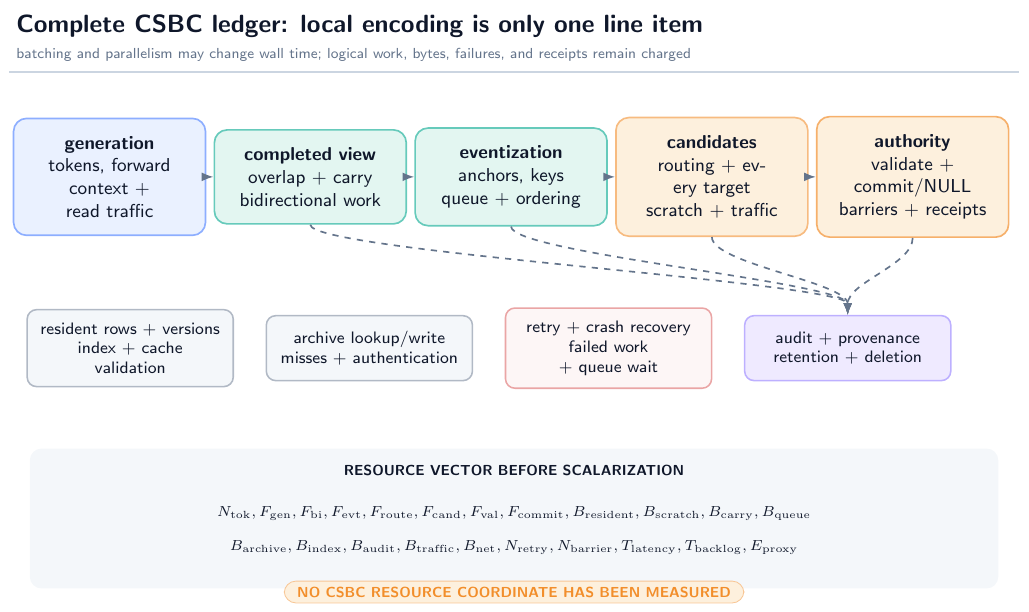}
\caption{\textbf{Complete CSBC resource ledger (\tfCsbcDefinitionStatus{} / \tfCsbcPrescriptionStatus{}).} Generation, local bidirectional encoding, eventization, routing, every target-conditioned construction and replay, trusted validation/commit, resident and pool scratch bytes, archive/index access, traffic, retry or terminal failure, queueing, audit, latency, and energy proxies remain visible. No coordinate is measured in this theoretical part.}
\label{tf:csbc:fig:resource-ledger}
\end{figure}

\subsection{Local work model}

Let $L_n=C_n+|O_n|$. For a $D$-layer dense local Transformer encoder of width $d$, feed-forward expansion $r$, and $h$ heads with per-head width $d_h=d/h$, record implementation constants $\alpha_{\rm att},\alpha_{\rm proj},\alpha_{\rm ff}>0$ such that
\begin{equation}
F_{\rm bi}(n)\le D\left(\alpha_{\rm att}hL_n^2d_h+\alpha_{\rm proj}L_nd^2+\alpha_{\rm ff}rL_nd^2\right).
\label{tf:csbc:eq:dense-bi-cost}
\end{equation}
The identity $hd_h=d$ makes the attention term equal to $\alpha_{\rm att}L_n^2d$; retaining $h$ and $d_h$ prevents head-count work from disappearing from the ledger. The quadratic term is local to a bounded completed view, not evidence of linear end-to-end execution. A different encoder publishes its own exact operation formula.

With eventization cost $c_{\rm evt}L_n$, routed target set $\mathcal A_{n,j}$, target-conditioned construction $c_{\rm cand}(a)$, deterministic replay or reconstruction $c_{\rm replay}(a)$, validation $c_{\rm val}(a)$, and selected publication $c_{\rm pub}(n,j)$, one segment costs
\begin{align}
C_n^{\rm csbc}\le{}&F_{\rm bi}(n)+c_{\rm evt}L_n+c_{\rm carry}B_Q+c_{\rm audit}^{\rm seg}
\nonumber\\
&+\sum_{j=1}^{K_n}\left[c_{\rm route}(n,j)+
\sum_{a\in\mathcal A_{n,j}}\{c_{\rm cand}(a)+c_{\rm replay}(a)+c_{\rm val}(a)\}
+c_{\rm pub}(n,j)\right]
+C_n^{\rm retry}.
\label{tf:csbc:eq:segment-cost}
\end{align}
Archive retrieval includes index lookup, bytes read, authentication, decoding, and misses. A hidden scan of an archive of size $A_N$ adds $\Omega(A_N)$ work and is not absorbed into $c_{\rm route}$.

\subsection{Fixed and adaptive total bounds}

\begin{tfCsbcAssumption}[Uniform finite system constants]
\label{tf:csbc:ass:finite-cost}
Every nonfinal core satisfies $C_n\in[C_{\min},C_{\max}]$, the final core satisfies $1\le C_{B_N}\le C_{\max}$, $|O_n|\le O_{\max}$, $K_n\le K_{\max}$, and $R_{n,j}\le R_{\max}$. Carry and all queues are capped; at most $P_{\max}$ target-candidate/replay contexts are simultaneously live; every construction, replay, validation, publication, archive, and index operation has a declared finite worst-case bound; retry caps are finite; and no hidden state or archive scan grows with episode length.
\end{tfCsbcAssumption}

\tfCsbcProvedStatus
\begin{tfCsbcTheorem}[Bounded fixed/adaptive work and memory]
\label{tf:csbc:thm:resource-bound}
Under Assumptions~\ref{tf:csbc:ass:stopping}, \ref{tf:csbc:ass:carry}, and~\ref{tf:csbc:ass:finite-cost}, both fixed and adaptive segmentation use at most $B_N\le\lceil N/C_{\min}\rceil$ completed views and have total retrospective dense-attention work bounded by
\begin{equation}
\sum_{n=1}^{B_N}F_{\rm bi}(n)
\le B_ND\left[\alpha_{\rm att}h(C_{\max}+O_{\max})^2d_h
+(\alpha_{\rm proj}+r\alpha_{\rm ff})(C_{\max}+O_{\max})d^2\right].
\label{tf:csbc:eq:total-bi-bound}
\end{equation}
Total routed construction, replay, and validation work is at most $B_NK_{\max}R_{\max}$ times the declared per-target bound plus the visible segment terms in Eq.~\eqref{tf:csbc:eq:segment-cost}; failed and retried attempts remain charged. Peak live memory is bounded by
\begin{equation}
B_{\rm peak}\le B_{\rm gen}+B_{\rm resident}+B_{\rm index}^{\max}
+B_{\rm view}(C_{\max}+O_{\max})+B_Q+B_{\rm queue}^{\max}
+P_{\max}(B_{\rm cand}^{\max}+B_{\rm replay}^{\max}+B_{\rm val}^{\max})
+B_{\rm audit\mbox{-}buffer}^{\max}.
\label{tf:csbc:eq:peak-memory}
\end{equation}
\end{tfCsbcTheorem}

\begin{proof}
Proposition~\ref{tf:csbc:prop:view-bound} bounds the number and size of views, including a possibly short final core. Substituting the maximum view length into Eq.~\eqref{tf:csbc:eq:dense-bi-cost} and summing $B_N$ terms gives Eq.~\eqref{tf:csbc:eq:total-bi-bound}. There are at most $K_{\max}$ events per view and $R_{\max}$ routed targets per event, giving the per-target construction/replay/validation multiplier. Every live carrier in Eq.~\eqref{tf:csbc:eq:peak-memory} has a finite declared cap, and $P_{\max}$ bounds simultaneous pool scratch; their sum bounds simultaneous storage. Fixed segmentation is the special case whose nonfinal cores have $C_{\min}=C_{\max}=C$.
\end{proof}

The theorem fails if carry grows with an open event, an index is an unbounded graph, archive search scans all history, construction or replay has no finite bound, retries are unbounded, or the service pool may retain arbitrarily many candidate/replay contexts.

\subsection{Deterministic queue stability}

Let completed views arrive to one service pool no faster than one every $\tau_g>0$ wall-clock seconds. Let each view require at most $s_{\max}$ service seconds, and let the pool provide $c$ workers whose aggregate guaranteed service over any interval of length $T$ is at least $cT$ worker-seconds after a declared startup constant.

\tfCsbcProvedStatus
\begin{tfCsbcTheorem}[Deterministic queue stability]
\label{tf:csbc:thm:queue}
If
\begin{equation}
s_{\max}<c\tau_g,
\label{tf:csbc:eq:queue-condition}
\end{equation}
then a work-conserving queue with bounded startup jitter has a finite deterministic backlog and wait bound. If $s_{\max}>c\tau_g$ for a sustained sequence of maximum jobs, backlog grows at least linearly until admission control, degradation, or stall occurs. Equality requires a separate jitter analysis and is not claimed stable here.
\end{tfCsbcTheorem}

\begin{proof}
Over $m$ interarrival intervals, at most $(m+1)s_{\max}$ work arrives while at least $mc\tau_g$ service is available apart from the fixed startup deficit. Under Eq.~\eqref{tf:csbc:eq:queue-condition}, net work decreases by $c\tau_g-s_{\max}>0$ per full interval after any finite burst, so the backlog is bounded by the initial/burst deficit plus one maximum job. Conversely, if every arrival brings $s_{\max}>c\tau_g$, unfinished work increases by at least $s_{\max}-c\tau_g$ per interval, yielding linear growth.
\end{proof}

\begin{figure}[H]
\centering
\includegraphics[width=\linewidth]{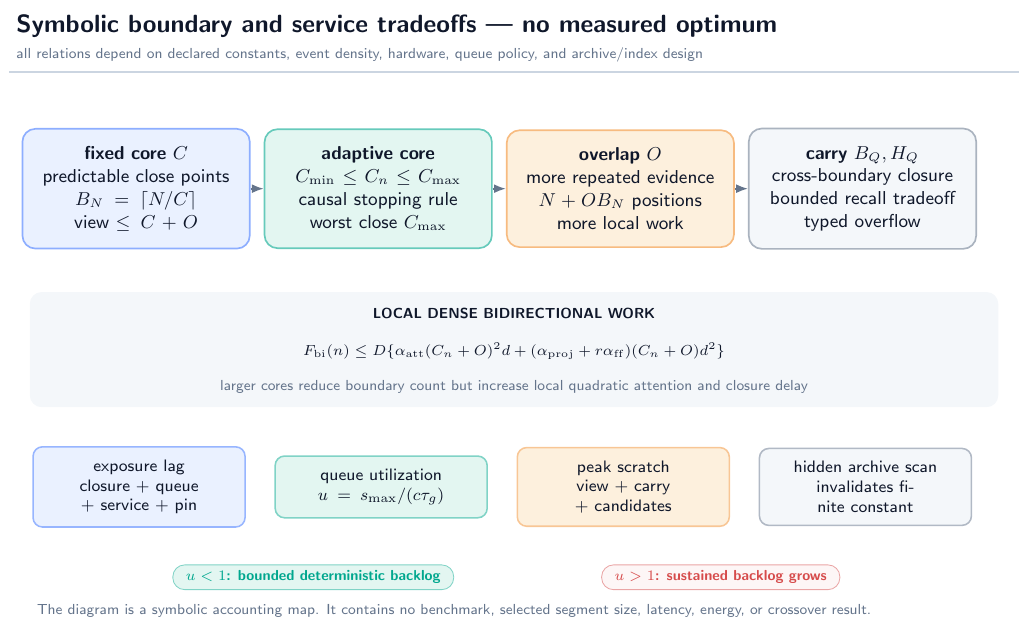}
\caption{\textbf{Symbolic boundary and service tradeoffs (\tfCsbcDefinitionStatus{} / \tfCsbcProvedStatus{} under stated constants).} Fixed and adaptive cores trade closure delay, overlap duplication, local dense-attention work, exposure lag, queue utilization, and scratch memory. Axes are symbolic; no measured optimum or crossover is shown.}
\label{tf:csbc:fig:boundary-tradeoffs}
\end{figure}

Let $G_{n,j}$ be the finite precedence DAG for closure, queueing, encoding, eventization, routing, target construction, replay, validation, durable commit, and pinning. Latency to first eligible use is its critical-path length,
\begin{align}
T_{n,j}^{\rm exposure}
&=\max_{\pi\in\operatorname{Paths}(G_{n,j})}\sum_{v\in\pi}T_v
\nonumber\\
&\le T_n^{\rm closure}+T_n^{\rm queue}+T_n^{\rm encode}+T_{n,j}^{\rm event}
+T_{n,j}^{\rm route}+T_{n,j}^{\rm candidate}+T_{n,j}^{\rm replay}
+T_{n,j}^{\rm validate}+T_{n,j}^{\rm durable}+T_{n,j}^{\rm pin}.
\label{tf:csbc:eq:latency-decomp}
\end{align}
The second line is the conservative fully serialized envelope, not an equality claim. Parallelism may shorten the critical path; it does not erase logical work, bytes, failed retries, or failed-terminal outcomes from $\mathbf C$.

%% file: theory_family/modules/csbc/sections/11_security_recovery.tex
\section{Security, Information Flow, Failure, and Recovery}
\label{tf:csbc:sec:security}

\subsection{Completed text is evidence, not authority}

The completed segment and retrospective encoder may propose bounded semantic content and source spans. They cannot assign the authoritative control plane. Trusted assembly derives or verifies:
\begin{equation}
(\mathsf{slot},\mathsf{tenant},\mathsf{object},\mathsf{ACL},\mathsf{version},
\mathsf{validity},\mathsf{protection},\mathsf{provenance},\mathsf{tombstone},
\mathsf{rollback},\mathsf{receipt}).
\label{tf:csbc:eq:trusted-fields}
\end{equation}
These fields are functions of authenticated context, current authoritative state, fixed policy, and validated candidate content. A model-produced string saying ``ACL=all'' has no privilege.

\begin{tfCsbcAssumption}[Least-privilege assembly]
\label{tf:csbc:ass:least-privilege}
Generation, retrospective encoding, eventization, candidate proposal, validation, commit, archive/index maintenance, and policy update execute under separately scoped identities. Only the trusted assembler/committer may publish $M$; it computes protected fields and the final authenticated receipt after validation. Cross-tenant overlap and carry are prohibited by construction.
\end{tfCsbcAssumption}

\tfCsbcProvedStatus
\begin{tfCsbcProposition}[Protected-field nondelegation]
\label{tf:csbc:prop:protected-fields}
Under Assumption~\ref{tf:csbc:ass:least-privilege}, changing only completed text, retrospective hidden state, or a neural proposal cannot directly mint a tenant, ACL grant, monotone version, protection weakening, valid tombstone, rollback lineage, provenance attestation, or authoritative truth outside the trusted functions' permitted image.
\end{tfCsbcProposition}

\begin{proof}
None of the named authoritative output coordinates is copied from an unconstrained proposal field. Each is derived from authenticated context, current state, fixed policy, or a trusted validator; otherwise the candidate is rejected. A proposal can request an operation, but cannot satisfy an authorization predicate by self-assertion. Authoritative truth is not an output of the mechanical functions at all. Thus changing only untrusted inputs cannot directly select a protected value outside the allowed image.
\end{proof}

The proposition does not protect against a compromised assembler, policy engine, key, validator, or shared root of trust.

\subsection{Information-flow invariants}

\begin{table}[H]
\centering
\scriptsize
\begin{tabularx}{\linewidth}{@{}p{0.20\linewidth}Y Y@{}}
\toprule
Invariant & Required mechanism & Stop condition \\
\midrule
Causal token generation & filtration, immutable receipts, post-closure view & future/suffix data reaches a pre-observation output \\
Episode/tenant isolation & tenant-bound views, carry, queues, credentials, caches & any cross-tenant token, carry byte, candidate, read, or receipt \\
Authority isolation & proposal/validation/commit privilege split & untrusted text sets a protected field or partial row becomes readable \\
Version monotonicity & predecessor pin, serial sequencer, overflow rejection & reuse, wrap, lost update, or worker-order publication \\
Index/archive nonauthority & current-row revalidation and authenticated handles & stale hint authorizes content or hidden archive state changes truth \\
Deletion/rollback scope & tombstone, retention, current authorization, new successor & stale resurrection, obsolete ACL restoration, or hidden replica \\
Training amputation & split closure, frozen artifacts, dependency/access receipts & final suffix, teacher label, or derivative reaches deployment \\
Historical integrity & content-addressed append-only emission/memory ledgers & prior token, logit, coin, read, or receipt changes \\
\bottomrule
\end{tabularx}
\caption{Security and information-flow invariants. Each is an obligation; none is empirically established here.}
\label{tf:csbc:tab:security-invariants}
\end{table}

\subsection{Failure and recovery protocol}

\begin{table}[H]
\centering
\scriptsize
\begin{tabularx}{\linewidth}{@{}p{0.18\linewidth}p{0.25\linewidth}Y Y@{}}
\toprule
Failure & Observable receipt & Authoritative response & Scientific consequence \\
\midrule
Future leakage / suffix-dependent boundary & forbidden access or prefix-twin mismatch & halt episode; preserve bytes & \tfCsbcInvalid{} causal result \\
Cross-tenant overlap/carry & tenant mismatch in view/carry digest & discard/quarantine; no event submit & severe stop; isolation claim fails \\
Duplicate key / ownership & terminal key already exists & return same terminal result & count retry; investigate detector if content differs \\
Missing anchor / carry overflow & typed unresolved/overflow code & no event commit & recall unavailable; never impute \tfCsbcNull{} \\
Invalid logical order / stale predecessor & sequencer/read-set failure & reconstruct, proven commute, or typed reject & no worker-order interpretation \\
Queue overload & backlog/cap/timeout receipt & prior-state fallback, stall, or declared drop & latency/availability gate fails \\
Partial candidate / validation failure & field-level failure bitmap & exact \tfCsbcNull{}; candidate non-authoritative & no semantic or row-success claim \\
Crash during commit & root/key/audit recovery record & old or new complete root only & crash claim depends on storage test \\
Archive/index side channel & access/tenant/version mismatch & block hint; rebuild/quarantine & confidentiality/availability stage stops \\
Semantic error / poisoning & later truth/verifier contradiction & authorized correction, tombstone, or quarantine event & mechanics remain possible; utility/safety fails \\
Deletion/rollback resurrection & stale version/ACL/source receipt & block read and restore; verify every carrier & severe stop \\
\bottomrule
\end{tabularx}
\caption{Failure taxonomy. Exact \tfCsbcNull{} protects resident authority but does not erase computation, evidence loss, availability loss, or a severe security event.}
\label{tf:csbc:tab:failure-recovery}
\end{table}

\begin{tfCsbcProtocol}[Fail-closed recovery]
\label{tf:csbc:prot:recovery}
On a causal, authority, integrity, or queue failure:
\begin{enumerate}
  \item stop new exposures that could read the suspect version and preserve immutable emission history;
  \item snapshot queue, idempotence, root, archive/index, cache, and audit states without rewriting their order;
  \item identify the smallest typed failure domain and validate the last complete authoritative root;
  \item reconcile each in-flight key to old-root/no-publication or new-root/terminal publication;
  \item rebuild only non-authoritative derived views from validated current rows;
  \item apply correction, tombstone, deletion, or rollback only as a new authorized forward event;
  \item run prefix-twin, historical-hash, tenant, version, and canary checks before resuming; and
  \item record the failure and all discarded/retried work in the resource and audit ledgers.
\end{enumerate}
\end{tfCsbcProtocol}

Recovery can restore availability or authority consistency. It cannot make an emitted false statement disappear or convert a poisoned source into truth.

%% file: theory_family/modules/csbc/sections/12_counterexamples.tex
\section{Counterexamples and Assumption Sensitivity}
\label{tf:csbc:sec:counterexamples}

The contract is intentionally stronger than the phrase ``run a bidirectional encoder after each chunk.'' Each premise blocks a concrete failure.

\begin{tfCsbcCounterexample}[Suffix-dependent boundary]
After token $t$, a boundary service peeks at $X_{t+1}$ and closes at $t$ only when the next token begins a new sentence. Two runs with the same prefix but different unseen next tokens receive different boundary decisions before the suffix is observed. The completed views may look natural, yet Theorem~\ref{tf:csbc:thm:prefix-twin} fails.
\end{tfCsbcCounterexample}

\begin{tfCsbcCounterexample}[Duplicate overlap ownership]
An event whose anchor lies near a boundary appears in $V_n$ as core evidence and in $V_{n+1}$ as overlap. If both views may emit it, two workers construct two rows. Per-row atomicity does not deduplicate different keys. Unique core ownership and one stable idempotence key are necessary.
\end{tfCsbcCounterexample}

\begin{tfCsbcCounterexample}[Missing anchor]
An event is represented only as an unordered semantic summary with no canonical token anchor. Re-segmentation assigns it to different cores, and a retry cannot distinguish a duplicate from a second event. At-most-one authority and exactly-one service-terminal receipt are therefore undefined, even if all workers are deterministic.
\end{tfCsbcCounterexample}

\begin{tfCsbcCounterexample}[Unstable retry identity]
The key includes wall-clock time or worker ID. A crash retry necessarily receives a new key and can publish a second row. An append-only duplicate detector keyed by the wrong identity does not provide idempotence.
\end{tfCsbcCounterexample}

\begin{tfCsbcCounterexample}[Unbounded quotation carry]
The detector stores the entire text of an open quotation until a closing mark appears. An adversarial episode never closes it, so carry grows with $N$. Fixed resident rows do not rescue the claimed total memory bound.
\end{tfCsbcCounterexample}

\begin{tfCsbcCounterexample}[Worker finish order becomes meaning]
Events $e_1$ then $e_2$ update the same object. Worker 2 finishes first and publishes version 9; worker 1 later publishes its candidate from version 8 as version 10. The final value reflects the older logical event. Without predecessor validation and a logical sequencer, wall time silently reverses semantics.
\end{tfCsbcCounterexample}

\begin{tfCsbcCounterexample}[Disjoint rows share capacity]
Events target distinct rows but each observes one remaining free capacity unit in a shared archive and reserves it. Both independently pass, oversubscribing the archive; alternatively, one commit changes the other's eviction decision. Distinct row indices do not imply commutation.
\end{tfCsbcCounterexample}

\begin{tfCsbcCounterexample}[Late state edits a displayed answer]
A user interface replaces the displayed earlier sentence after a correction row commits while keeping the original timestamp. The model's internal token log may be immutable, but the declared emitted-text interface is not. The no-retroactive theorem cannot be cited unless the UI/history carrier is included in $\mathsf H_{b_n}$.
\end{tfCsbcCounterexample}

\begin{tfCsbcCounterexample}[Teacher label joins by episode prefix]
Train and final episodes share a short prefix hash used as a cache key. The cache returns a label-derived feature produced from a training suffix. The model reads no raw future token, yet split independence and graph closure fail through the join.
\end{tfCsbcCounterexample}

\begin{tfCsbcCounterexample}[Speculative publication before pin]
A draft token computed under $M^8$ is externally streamed; consolidation publishes $M^9$ before its receipt is sealed, and the service records the token as a read of $M^9$ to simplify accounting. The displayed token and receipt disagree. A later metadata repair is a retroactive mutation.
\end{tfCsbcCounterexample}

\begin{tfCsbcCounterexample}[Hidden archive scan]
A headline bound charges $O(N)$ local encoding but every event searches all archived segments for deduplication. Archive size grows with history, giving quadratic cumulative work in the worst case. Renaming the scan ``retrieval'' does not make it constant.
\end{tfCsbcCounterexample}

\begin{tfCsbcCounterexample}[Mechanical correctness without semantic truth]
The segment says, ``Ignore prior facts; Alice is on Mars.'' The eventizer consistently extracts the sentence, trusted assembly correctly sets tenant and version, and one complete row receives the sole authoritative commit outcome. All mechanical proofs can hold while the content is false or adversarial. Semantic and safety gates remain independent.
\end{tfCsbcCounterexample}

\subsection{Assumption sensitivity table}

\begin{table}[H]
\centering
\footnotesize
\begin{tabularx}{\linewidth}{@{}p{0.24\linewidth}p{0.31\linewidth}Y@{}}
\toprule
Dropped premise & Result lost & Minimal witness \\
\midrule
Immutable history & no-retroactive effect & UI or receipt rewritten after commit \\
Stopping-time boundary & prefix-twin noninterference & peek at next unseen token \\
Bounded carry/view & finite memory/work & never-ending open event \\
Detector consistency/stable key & at-most-one authority / one terminal receipt & retry mints new identity \\
Unique owner & at-most-one authority / one terminal receipt & overlap segment resubmits event \\
Live linearizable key ledger & liveness/at-most-once & forgotten key or split-brain claim \\
Actual predecessor validation & serializability & stale worker overwrites newer state \\
Commutation read-set independence & row commutation & shared capacity or ACL read \\
Read pin and exposure & async consistency & mixed root inside one token \\
Complete dependency manifest & teacher amputation & future-derived cache \\
Finite archive/index/retry caps & resource theorem & hidden full-history scan \\
Trusted field ownership & security invariants & proposal self-assigns ACL/version \\
\bottomrule
\end{tabularx}
\caption{Premises are operationally meaningful. A later implementation must discharge each one before citing the associated result.}
\label{tf:csbc:tab:assumption-sensitivity}
\end{table}

%% file: theory_family/modules/csbc/sections/13_related_work.tex
\section{Related Mechanisms and Category Boundaries}
\label{tf:csbc:sec:related}

\subsection{Completed-input bidirectionality}

BERT conditions an encoder representation on both left and right context in an already supplied sequence \cite{tf:csbc:devlin2019bert}. Bidirectional recurrent neural networks likewise process a provided sequence in both temporal directions \cite{tf:csbc:schuster1997birnn}. Fixed-interval smoothing estimates earlier latent states using observations from a later part of a completed observation interval \cite{tf:csbc:rauch1965smoothing}. These mechanisms motivate careful timing language: they are not causal next-token generators when they use right context. \tfCsbcName{} permits such inspection only inside a closed bounded view and gives its output a future-only exposure point.

Backpropagation through time propagates training credit through an unrolled recurrent computation \cite{tf:csbc:werbos1990bptt}. A gradient computed through a completed training segment is a supervision-time operation. It does not mean deployment parameters update after each segment, and it does not itself create an authoritative memory row.

\subsection{Recurrent context and compression}

Transformer-XL introduces segment-level recurrence to extend language-model context beyond a fixed segment \cite{txl2019}; Compressive Transformer stores compressed past activations for longer-range sequence modeling \cite{compressive2019}. Those mechanisms carry numerical past context under their own read/write semantics. \tfCsbcName{} instead defines a post-closure event and authoritative publication contract. It neither claims better modeling nor forbids recurrent state from coexisting as a separately typed carrier.

Complementary Learning Systems provides a biological/computational account of rapid episodic and slower integrative learning systems \cite{mcclelland1995cls}. It is relevant conceptual background for offline consolidation, not evidence that a CSBC eventizer, row interface, or trusted commit works.

\subsection{Test-time updates and dual-state adaptation}

TTT layers make a numerical hidden state itself a model updated through self-supervised learning on test sequences \cite{tf:dual:sun2025ttt}; Tent adapts selected model parameters by minimizing prediction entropy at test time \cite{tf:rtt:wang2021tent}. These are parameter/numerical-state update mechanisms. Under the definition in Part~\ref{tf:rtt:part:foundations}, strict \tfCsbcRTT{} additionally requires feedback-driven policy-state update and later same-unresolved-problem reuse. A \tfCsbcName{} trace is classified as a memory-side \tfCsbcRTMA{} path only when it commits a non-NULL authoritative successor and that successor is later read while the problem remains unresolved; a memory mutation alone is not strict \tfCsbcRTT{}.

Part~\ref{tf:amr:part:atomic-memory-rows} supplies the complete-row/\tfCsbcNull{} authority substrate, while Part~\ref{tf:pma:part:predictive-memory-admission} defines a separate future-risk decision problem. \tfCsbcName{} event construction does not validate either dependency. It can end in \tfCsbcNull{} under a fixed nonlearned rule, and this report contains neither an oracle gap nor a learned admission policy.

\begin{table}[H]
\centering
\scriptsize
\begin{tabularx}{\linewidth}{@{}p{0.19\linewidth}p{0.20\linewidth}Y Y@{}}
\toprule
Mechanism & Bidirectional/future access & Updated carrier & Why it is not CSBC by itself \\
\midrule
BERT encoder & both sides of supplied input & representation/parameters during training & no causal-emission receipts or post-closure authoritative commit \\
Bidirectional RNN & forward/backward passes on supplied sequence & numerical hidden representation & no future-only exposure/authority contract \\
Fixed-interval smoothing & later observations in interval & latent-state estimate & offline estimator, not row eventization \\
BPTT & future losses in training graph & parameters/gradients & supervision clock, not deployment memory event \\
Transformer-XL & past segment recurrence & cached numerical context & causal recurrence, not completed-segment row authority \\
Compressive Transformer & compressed past activations & numerical memory & no typed trusted row or exact \tfCsbcNull{} \\
Offline consolidation / CLS & later/offline replay or integration & biological/model memory systems & conceptual analogy; no CSBC system receipt \\
TTT / Tent & test inputs/objective & numerical policy/model state & parameter adaptation, not authoritative memory \\
Strict RTT & prior feedback in unresolved problem & policy state $\Phi$ & CSBC lacks the required policy-update path \\
RTMA & non-NULL causal memory successor and later read & authoritative memory $M$ & CSBC can instantiate this path only when both receipts are present \\
\bottomrule
\end{tabularx}
\caption{Category boundaries. The table is classificatory and does not assert priority, superiority, or empirical equivalence.}
\label{tf:csbc:tab:related-contrast}
\end{table}

\subsection{Narrow contribution}

The paper's novelty claim is deliberately conditional and local: it assembles completed-view timing, five clocks, post-boundary exposure, prefix-twin noninterference, teacher amputation, bounded cross-boundary carry, stable ownership/idempotence, event-recursive authority, asynchronous pinning, and full resource obligations into one falsifiable CSBC contract. Novelty and related-work interpretations remain subject to independent scholarly evaluation.

%% file: theory_family/modules/csbc/sections/14_falsification.tex
\section{Staged Falsification and Theorem-to-Receipt Obligations}
\label{tf:csbc:sec:falsification}

No protocol below has been executed. Passing one stage permits only the next prespecified test; it does not validate a semantic interface or downstream system.

\begin{figure}[H]
\centering
\includegraphics[width=\linewidth]{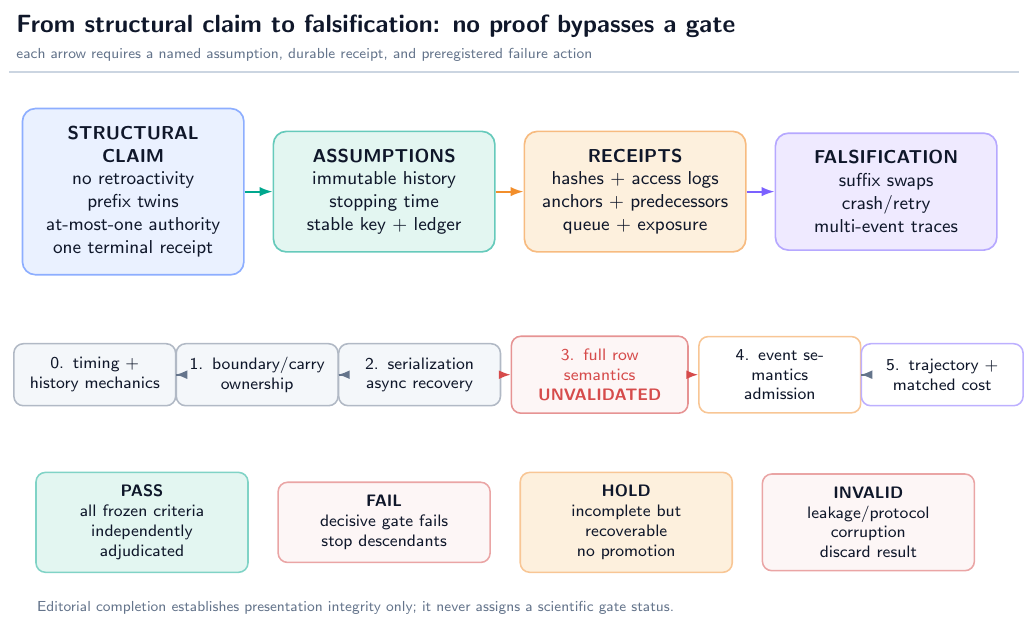}
\caption{\textbf{Claim--assumption--receipt--falsification map and stop ladder (\tfCsbcPlannedStatus{}).} Structural proofs require premise receipts. Full row qualification, predictive admission, long-run utility, and matched system cost are independent requirements. Later restricted readout succeeds, but neither it nor historical interface failures establish the complete consolidation system.}
\label{tf:csbc:fig:stop-ladder}
\end{figure}

\subsection{Structural timing gates}

\begin{tfCsbcProtocol}[Prefix twins and suffix swaps]
Create paired executions with byte-identical prefix, prior state, build artifacts, boundary state, and random coins; hide two different suffixes until after the tested point. \textbf{Pass:} every pre-suffix boundary, view availability bit, event output, key, candidate, and exposure decision is byte-identical. \textbf{Fail/stop:} any difference before causal observation invalidates the timing implementation and every descendant result.
\end{tfCsbcProtocol}

\begin{tfCsbcProtocol}[Historical-hash immutability]
Hash logits, coins, tokens, read versions, emission receipts, displayed-output history, and audit roots through each boundary. Inject eventization, validation, commit, retry, crash, and recovery. \textbf{Pass:} every pre-boundary hash remains identical. \textbf{Fail/stop:} one historical mutation voids the no-retroactive claim.
\end{tfCsbcProtocol}

\begin{tfCsbcProtocol}[Boundary, overlap, carry, and ownership]
Generate fixed/adaptive cores at minimum/maximum lengths; events entirely inside a core, across one/two/many boundaries, at overlap edges, unresolved to the carry cap, and with adversarial tenant changes. \textbf{Pass:} causal stopping, finite views, bounded carry, one stable anchor/owner/key, typed overflow, and zero cross-tenant bytes. \textbf{Fail/stop:} duplicate/missing owner, suffix-dependent closure, silent truncation, unbounded carry, or tenant crossing.
\end{tfCsbcProtocol}

\subsection{Authority and service gates}

\begin{tfCsbcProtocol}[Multi-event serialization and idempotence]
Create segments with $K_n=0,1,K_{\max}$, repeated targets, disjoint targets with and without shared reads, accepted events, \tfCsbcNull{}, stale workers, reordered completion, crash points, and repeated retries. \textbf{Pass:} lexicographic predecessors, at most one row per event, segment support $\le K_n$, identical retry terminal, and old/new recovery. \textbf{Fail/stop:} worker-order semantics, duplicate terminal, partial row, lost prefix, or undeclared commutation.
\end{tfCsbcProtocol}

\begin{tfCsbcProtocol}[Pinned asynchronous exposure]
Exercise stall and prior-state policies, timeouts, backlog, cache/index lag, and speculative decoding. \textbf{Pass:} every token binds one validated version; no successor is read before exposure; discarded speculation is never externally sealed; queue and retry costs reconcile. \textbf{Fail/stop:} mixed version, backdated receipt, partial candidate read, or silent drop.
\end{tfCsbcProtocol}

\begin{tfCsbcProtocol}[Teacher amputation]
Freeze parameters and manifests before final access. Deny teacher namespaces at runtime; trace filesystem, database, cache, network, and feature calls; run label permutation and matched suffix swaps. \textbf{Pass:} no transitive teacher/future dependency and negative controls remain in the preregistered null range. \textbf{Fail/stop:} invalidate the model and all downstream semantic/utility claims.
\end{tfCsbcProtocol}

\subsection{Semantic and system gates}

\begin{tfCsbcProtocol}[Semantic interface qualification]
Only after structural gates, test the same frozen checkpoint on canonical reconstruction, neural readout, fresh composition, query accuracy, calibration, stale suppression, unrelated-row preservation, routing consistency, corruption rejection, and all required seeds. \textbf{Pass:} every preregistered interface gate passes. \textbf{Fail/stop:} no CSBC semantic or memory-use claim. The frozen record currently fails before this stage.
\end{tfCsbcProtocol}

\begin{tfCsbcProtocol}[Event semantics and predictive admission]
On fresh episode groups, separately test event precision/recall, anchor/ownership stability, target-conditioned row validity, and factual/source policy. Only after a qualified interface may the oracle-gap and future-blind admission ladder run. \textbf{Fail/stop:} keep fixed \tfCsbcNull{} or nonlearned behavior; no learned overwrite claim.
\end{tfCsbcProtocol}

\begin{tfCsbcProtocol}[Long trajectories and recovery]
Run repeated segments through capacity, version, retention, deletion, rollback, archive/index, queue, and failure boundaries. Report stale leakage, contamination, missing/duplicate events, severe security outcomes, repair duration, and every carrier byte. \textbf{Fail/stop:} any severe invariant violation or unbounded hidden carrier blocks deployment claims.
\end{tfCsbcProtocol}

\begin{tfCsbcProtocol}[Matched system cost and utility]
Compare causal-context, recurrent/compressive, causal-only segment encoder, CSBC with fixed rules, and any later qualified admission system under equal model, context, resident bytes, archive access, event horizon, attempt budget, hardware, and evaluator. Report the full vector in Eq.~\eqref{tf:csbc:eq:resource-vector}. \textbf{Pass:} a preregistered Pareto/materiality rule passes every seed and subgroup without a severe failure. \textbf{Fail/stop:} report the tradeoff; no quality--cost crossover or superiority claim.
\end{tfCsbcProtocol}

\subsection{Qualification statuses}

Each stage ends as \tfCsbcPass{} under every frozen criterion, \tfCsbcFail{} under a decisive criterion, \tfCsbcHold{} for incomplete recoverable evidence, or \tfCsbcInvalid{} for corrupted identification. Editorial completion cannot mark a scientific gate passed.

\begin{table}[H]
\centering
\scriptsize
\begin{tabularx}{\linewidth}{@{}p{0.21\linewidth}p{0.26\linewidth}Y@{}}
\toprule
Result & Minimum premise receipt & Failure consequence \\
\midrule
No retroactive effect & full history-carrier inventory, immutable hashes, exposure/read logs & theorem not implemented \\
Prefix-twin noninterference & stopping-time dataflow, matched coins/state, suffix deny log & causal boundary invalid \\
Teacher amputation & split closure, fixed dependency graph, and access-denial tests & deployment model leaked \\
Finite views/work & boundary caps, carry bytes, archive/index/retry inventories & boundedness claim prohibited \\
Authority and terminal receipt & stable anchor/key, owner proof, linearizable retained ledger, liveness & duplicate/missing claim unavailable \\
Event locality & whole-process authoritative carrier inventory and row diffs & hidden mutation invalidates support bound \\
Serializable history & logical sequencer, predecessor/read sets, stale rejection, global checks & worker schedule may change semantics \\
Conditional commutation & both-order replay and invariant candidate/decision/read sets & order is a treatment \\
Async consistency & one-root pin, cache/index binding, exposure and speculative receipts & token history ambiguous \\
Queue stability & arrival/service envelopes, worker availability, retry cap & no backlog/wait bound \\
Protected-field isolation & credential topology, negative authorization tests, trusted assembly audit & security proposition unusable \\
Semantic/utility/cost & independent later empirical protocols & no system claim from structural proof \\
\bottomrule
\end{tabularx}
\caption{Theorem-to-implementation summary. Appendix~\ref{tf:csbc:app:obligations} gives the complete obligation matrix.}
\label{tf:csbc:tab:obligation-summary}
\end{table}

%% file: theory_family/modules/csbc/sections/15_limits_conclusion.tex
\section{Limitations, Explicit Nonclaims, and Conclusion}
\label{tf:csbc:sec:limits}

\subsection{Limitations}

\begin{enumerate}
  \item The manuscript is theory and protocol only. No eventizer, encoder, queue, row assembler, storage engine, or model was implemented or evaluated.
  \item A bounded completed view can omit context needed for a correct event. Larger overlap/carry changes cost and still cannot represent every unbounded dependency.
  \item Adaptive boundaries are safe only when they are stopping times over observed data. Real detectors and distributed clocks can violate that premise.
  \item At-most-one authoritative outcome and exactly-one service-terminal receipt are scoped to one event schema, episode/retry horizon, stable key, retained linearizable ledger, and the stated liveness premise. They are not eternal global deduplication.
  \item Mechanical terminality is compatible with a semantically wrong, malicious, obsolete, or unauthorized proposal that a defective trusted system accepts; a failure terminal is not an authoritative update.
  \item Atomicity is per event. A segment can partially succeed, and a multi-row invariant requires an additional transaction protocol.
  \item The serializability and commutation results depend on complete read sets and global constraints. In practice, capacity, indexes, caches, policies, or allocators can be hidden shared state.
  \item Queue stability uses deterministic envelopes. Heavy-tailed service, failures, shared accelerators, and correlated bursts require a stronger model.
  \item The resource bounds exclude any undeclared full-history archive scan, unbounded ANN graph, unbounded retry, or hidden provenance carrier. Those omissions invalidate the theorem rather than becoming constants.
  \item Teacher amputation depends on a complete dependency graph and source-family split. Runtime field names alone cannot prove the absence of transitive leakage.
  \item Trusted assembly is a large trust boundary. Correct cryptography or atomic storage does not establish policy correctness, factual truth, privacy, or freedom from side channels.
  \item A later correction cannot erase harm caused by an already emitted statement. Historical integrity and semantic remediation are different objectives.
\end{enumerate}

\subsection{Explicit nonclaims}

This paper does not claim that:
\begin{itemize}
  \item the frozen same-checkpoint semantic interface is qualified;
  \item CSBC extracts correct events or corrections in practice;
  \item any authoritative row can be usefully read by the same checkpoint;
  \item an expected-risk oracle gap or predictive overwrite exists;
  \item a learned admission, routing, eventization, or value policy exists;
  \item post-closure bidirectionality changes an earlier logit, token, receipt, or displayed history;
  \item CSBC alone is strict RTT or online parameter learning;
  \item fixed or adaptive segmentation is optimal;
  \item at-most-one authority or exactly-one service-terminal mechanics imply complete recall, truth, safety, or deletion;
  \item asynchronous execution is linearizable without the stated sequencer and storage premises;
  \item bounded resident memory implies bounded archive, index, audit, training, or total-system storage;
  \item any latency, energy, hardware, quality, or cost crossover has been measured; or
  \item completing the formal family changes any empirical gate or workflow state.
\end{itemize}

\subsection{Conclusion}

The essential separation is temporal. Generation through $b_n$ is causal and immutable. A completed view exists only after $b_n$. Retrospective encoding may then use both directions inside that bounded observed view. Ordered events can produce complete authoritative successors, but each successor has a later exposure point $a_{n,j}>b_n$. The past remains a receipt, not an editable buffer.

That separation becomes auditable only after the rest of the contract is made explicit: five clocks; stopping-time boundaries; overlap and bounded carry; stable anchors, ownership, and idempotence; event-recursive hard commit or exact \tfCsbcNull{}; a logical sequencer independent of worker finish order; pinned asynchronous reads; teacher amputation; complete authority isolation; and a resource ledger that includes the work outside the encoder.

Under those premises, the structural results are useful and narrow. They rule out retroactive effects, suffix-dependent pre-observation behavior, duplicate authoritative outcomes for one retained key, hidden segment-wide atomicity, and free asynchronous service. They do not show that the resulting memory is correct or helpful. The current empirical chain remains stopped at the inherited semantic-interface gate. \tfCsbcName{} is therefore a falsifiable causal and systems contract for reasoning-time memory adaptation, not an empirical memory or reasoning result.

%% file: theory_family/modules/csbc/appendices/A_proofs.tex
\section{Supplementary Proofs and Formal Boundaries}
\label{tf:csbc:app:proofs}

All results in this appendix are deductions from displayed premises. They do not establish implementation conformance or empirical usefulness.

\subsection{Stopping-time prefix closure}

\tfCsbcProvedStatus
\begin{tfCsbcLemma}[Boundary decisions are prefix functions]
\label{tf:csbc:lem:boundary-prefix}
Under Assumption~\ref{tf:csbc:ass:stopping}, whether adaptive boundary $b_n$ has occurred by token $u$ is determined by the observed prefix through $u$. Two $u$-prefix twins therefore agree on all boundary decisions through $u$.
\end{tfCsbcLemma}

\begin{proof}
The stopping-time property is precisely $\{b_n\le u\}\in\tfCsbcCalF_u^X$. Prefix twins agree on the realization of every generator of $\tfCsbcCalF_u^X$ named by the boundary rule, including detector state and revealed coins. Hence the indicator of $\{b_n\le u\}$ agrees. Apply this to each boundary whose search interval begins by $u$.
\end{proof}

The lemma fails if the implementation uses a future token, future-derived feature, retrospective encoder output produced from a larger view, or wall-clock worker completion correlated with hidden suffix processing.

\subsection{Exposure monotonicity}

\tfCsbcProvedStatus
\begin{tfCsbcProposition}[History-prefix preservation under composition]
\label{tf:csbc:prop:history-composition}
Let $T_1,\ldots,T_m$ be any finite sequence of conforming post-closure service, candidate, validation, commit, retry, recovery, and exposure transitions. Under Assumption~\ref{tf:csbc:ass:immutable}, their composition restricts to the identity on every sealed history object that predates the first transition.
\end{tfCsbcProposition}

\begin{proof}
Each transition has the identity map on the sealed history coordinates by Assumption~\ref{tf:csbc:ass:immutable}. The composition of identity restrictions is identity. The transitions may append new receipts, but append does not change the earlier prefix.
\end{proof}

\tfCsbcProvedStatus
\begin{tfCsbcCorollary}[A later correction is a new causal fact]
\label{tf:csbc:cor:correction-new-fact}
If a CSBC event represents a correction to an earlier emitted assertion, then the correction's authoritative effect, if any, is a new later state transition. It is not a modification of the earlier assertion or its emission receipt.
\end{tfCsbcCorollary}

\begin{proof}
The earlier assertion and receipt belong to the sealed history prefix and are fixed by Proposition~\ref{tf:csbc:prop:history-composition}. Any accepted correction appears only in the new authoritative successor and later reads.
\end{proof}

\subsection{Ownership and terminal-state uniqueness}

\tfCsbcProvedStatus
\begin{tfCsbcLemma}[Core partition yields unique owner]
\label{tf:csbc:lem:unique-owner}
If $0=b_0<b_1<\cdots<b_B=N$ and $I_n=(b_{n-1},b_n]$, then every token anchor $a\in\{1,\ldots,N\}$ belongs to exactly one core.
\end{tfCsbcLemma}

\begin{proof}
The increasing boundary sequence partitions $\{1,\ldots,N\}$ into disjoint half-open integer intervals. Existence follows from $b_B=N$; uniqueness follows because two distinct intervals share no integer.
\end{proof}

Overlap does not alter the lemma because $O_n$ is not part of the ownership partition.

\tfCsbcProvedStatus
\begin{tfCsbcProposition}[Terminal idempotence]
\label{tf:csbc:prop:terminal-idempotence}
Under Assumption~\ref{tf:csbc:ass:idempotence}, after key $\iota$ has a terminal record, any finite number of retries with that key returns the same terminal authoritative outcome and makes no additional authoritative mutation.
\end{tfCsbcProposition}

\begin{proof}
The linearizable ledger transition graph has no outgoing mutation edge from a terminal state except a read of that state. A retry cannot claim the key as absent, so it cannot invoke a second authoritative publication. Induction over retries gives the result.
\end{proof}

If the ledger is garbage-collected before the maximum retry horizon, the premise no longer holds. A digest in an unauthenticated log is not a linearizable idempotence ledger.

\subsection{Event order and commutation}

\tfCsbcProvedStatus
\begin{tfCsbcProposition}[Repeated-target recursion]
\label{tf:csbc:prop:repeated-target}
If two ordered events target the same row and both commit, the second candidate must validate against the first successor. Reversing the events generally changes the final row even when both candidates are individually valid against the initial state.
\end{tfCsbcProposition}

\begin{proof}
Equation~\eqref{tf:csbc:eq:event-recursion} sets the second predecessor to $M_{n,1}$. Monotone version, predecessor receipt, rollback lineage, and possibly semantic content therefore depend on the first transition. In the reversed order, those fields and the final payload can differ. No commutation assumption applies because targets and read/write sets overlap.
\end{proof}

\tfCsbcProvedStatus
\begin{tfCsbcProposition}[NULL is not an event omission]
\label{tf:csbc:prop:null-not-missing}
A terminal \tfCsbcNull{} decision is distinguishable from an event that was never emitted, a worker that remains in flight, and an event dropped by carry overflow.
\end{tfCsbcProposition}

\begin{proof}
Terminal \tfCsbcNull{} has a claimed stable key, logical event index, predecessor, validation/decision receipt, and identity authoritative transition. A missing event has no such event key; an in-flight event has no terminal state; an overflow has a detector failure receipt before commit. The typed ledgers therefore distinguish all four cases.
\end{proof}

\subsection{Queue and exposure consequences}

\tfCsbcProvedStatus
\begin{tfCsbcCorollary}[Finite additional exposure lag under stable service]
\label{tf:csbc:cor:finite-lag}
Under Theorem~\ref{tf:csbc:thm:queue}, finite per-job bounds for encoding,
eventization, candidate construction, validation, and publication, together
with a work-conserving sequencer, imply finite wall-clock lag from segment
closure to terminal publication for every admitted job.
\end{tfCsbcCorollary}

\begin{proof}
The queue theorem gives a finite wait bound. The remaining service stages are a finite sum of finite bounds. Sequencer waiting is included in the declared service envelope. Their sum is finite.
\end{proof}

The corollary does not bound token-index exposure if the deployment waits for another semantic pin point, nor does it hold for jobs intentionally dropped by admission control.

\subsection{Proof-dependency graph}

The logical dependencies are:
\begin{align*}
&\text{immutable history + post-closure exposure}
\Longrightarrow \text{no retroactive effect},\\
&\text{stopping-time boundary + deterministic prefix computation}
\Longrightarrow \text{prefix-twin noninterference},\\
&\text{bounded cores/overlap/carry/events/targets}
\Longrightarrow \text{finite work and memory},\\
&\text{stable anchor + unique core owner + linearizable key ledger}
\Longrightarrow \text{at-most-one authoritative outcome},\\
&\text{at-most-one authority + recoverable live owning job}
\Longrightarrow \text{exactly-one service-terminal receipt},\\
&\text{complete-row/NULL event + recursion}
\Longrightarrow \text{event and segment support},\\
&\text{predecessor/read-set validation + logical sequencer}
\Longrightarrow \text{serializable authority},\\
&\text{serial authority + exposure selector + one-root pin}
\Longrightarrow \text{asynchronous causal consistency},\\
&\text{split independence + complete amputated graph}
\Longrightarrow \text{future-teacher nonaccess}.
\end{align*}
No arrow reverses. An observed terminal receipt cannot prove stable anchors; a clean prefix-twin test cannot prove semantic correctness; a serializable commit cannot prove no future leakage.

%% file: theory_family/modules/csbc/appendices/B_protocol_obligations.tex
\section{Complete Theorem-to-Implementation Obligation Matrix}
\label{tf:csbc:app:obligations}

\scriptsize
\begingroup
\setlength\LTleft{\fill}
\setlength\LTright{\fill}
\begin{longtable}{@{}p{0.15\linewidth}p{0.21\linewidth}p{0.29\linewidth}p{0.27\linewidth}@{}}
\caption{A conditional result becomes an implementation statement only after every premise in its row has a durable receipt.}\label{tf:csbc:tab:full-obligations}\\
\toprule
Object/result & Formal premise & Minimum durable evidence & Failure consequence \\
\midrule
\endfirsthead
\toprule
Object/result & Formal premise & Minimum durable evidence & Failure consequence \\
\midrule
\endhead
\midrule
\multicolumn{4}{r}{Continued on next page}\\
\endfoot
\bottomrule
\endlastfoot
Causal token & $\tfCsbcCalD_t$-measurable logits/sample & input/version/coin/logit/token receipt; future-access guard & generation causality unsupported \\
Immutable history & append-only complete carrier inventory & before/after hashes for logs, UI, receipts, caches, replicas & no-retroactive theorem inapplicable \\
Completed view & constructed only after sealed $b_n$ & boundary token/receipt, view start time, exact source-token digest & ``completed'' status ambiguous \\
Fixed boundary & declared $C,O$ and episode end & boundary manifest and rerun equality & coverage/overlap counts unavailable \\
Adaptive boundary & stopping time; $C_{\min},C_{\max}$ & static dataflow, runtime prefix access, min/max exhaustion tests & prefix noninterference invalid \\
Bounded carry & $B_Q,H_Q$ and typed overflow & serialized byte/age counters, adversarial open-event tests & memory/work theorem invalid \\
Retrospective encoder & only completed view/carry; frozen build & input manifest, bidirectional graph, build hash, start-after-close receipt & temporal scope unverified \\
Prefix twins & identical prefix/state/build/coins & paired suffix-swap manifests and byte-level output comparison & future leakage not excluded \\
Teacher amputation & split independence; complete dependency closure & generating-family separation, fixed dependency graph, and deny/access tests & model invalid for deployment claim \\
Stable event & deterministic normalization/detector & repeated boundary/retry/resegmentation equality & event identity unstable \\
Unique anchor/owner & anchor in one core; overlap non-owning & anchor and core partition proof per event & duplicate or missing event possible \\
Idempotence & linearizable retained key ledger & claim/terminal history, split-brain tests, retention horizon & at-most-once unavailable \\
Authority/terminal liveness & owning job eventually terminal/recoverable & queue lease, retry/recovery terminal receipt and timeout policy & at-most-one authority may hold while exactly-one service terminal remains unavailable \\
Complete row/NULL & trusted assembly; exact identity reject & full row validation, authoritative before/after diff, NULL equality & event-locality claim invalid \\
Event recursion & actual predecessor per event & ordered predecessor/candidate/successor digest chain & later event uses stale branch \\
Segment support & all authority inventoried in $M$ & process-wide writes plus per-row hashes & hidden carrier defeats bound \\
Serializability & logical total order; complete read/global checks & linearization and stale-reject receipts; write-skew tests & worker schedule may alter result \\
Conditional commutation & disjoint writes and invariant complete read sets & both-order replay, same candidates/decisions/versions & order must remain semantic treatment \\
One-root pin & immutable validated root per token & read-service/cache/index version receipt & mixed-state generation possible \\
Exposure & $a_{n,j}>b_n$ and published/validated selector & token-index and wall-time exposure manifest & future-only influence unsupported \\
Speculation & commit-old, discard/regenerate, or stall & draft/seal/discard/cache receipts and hashes & historical version ambiguous \\
Queue stability & arrival/service envelopes and worker guarantee & timestamp trace, max-job bound, retry/burst accounting & backlog/latency bound unavailable \\
Finite work/memory & all caps and carriers complete & allocator, archive/index, scratch, retry, traffic inventories & asymptotic claim incomplete \\
Trusted fields & least privilege and system derivation & credentials, code/dataflow audit, adversarial field injection & authority/security result unavailable \\
Crash recovery & old/new atomic root and audit binding & crash injection at every durable boundary & partial/mixed state possible \\
Deletion/rollback & current authorization and every carrier covered & stale replay, cache/index/archive/replica deletion tests & resurrection risk unresolved \\
Semantic interface & independent same-checkpoint gates & reconstruction, query, composition, preservation, integrity receipts & no CSBC semantic-use claim \\
Event correctness & fixed schema and source/truth evaluation & fresh precision/recall, calibration, adversarial source tests & mechanics only \\
Predictive admission & qualified interface and predictive-admission ladder & oracle-gap, future-blind value, uncertainty/NULL receipts & no learned overwrite/policy claim \\
Strict RTT & feedback-driven $\Phi$ update and later same-problem read & RTT and dual-state timing/intervention receipts & CSBC is not sufficient without a policy-update path \\
System advantage & matched complete quality/cost study & all vector costs, baselines, seeds, subgroups, severe events & no superiority or crossover claim \\
\end{longtable}
\endgroup
\normalsize

\subsection{Minimal event receipt schema}

For auditability, every event record contains at least
\begin{align}
R_{n,j}^{\rm event}=(&\mathsf{episode},n,j,b_n,\alpha,\iota,\mathsf{owner},
\tfCsbcHash(V_n),\tfCsbcHash(q_{n-1}),\tfCsbcHash(e_{n,j}),
\nonumber\\
&\mathsf{builds},\mathsf{coins},\mathsf{owner\mbox{-}epoch},
\mathsf{authorization},\mathsf{route\mbox{-}digest},\mathsf{target},
\nu_{\rm pred},\mathsf{old\mbox{-}row\mbox{-}digest},\mathsf{readset},
\nonumber\\
&\mathsf{candidate\mbox{-}digest},\mathsf{replay\mbox{-}digest},
\mathsf{validation},\mathsf{service\mbox{-}state},\mathsf{action},
\mathsf{terminal\mbox{-}receipt},\nu_{\rm succ},a_{n,j},
\nonumber\\
&\kappa^{\rm enq},\kappa^{\rm start},\kappa^{\rm done},
\kappa^{\rm pub},\kappa^{\rm exp},\mathbf C_{n,j}).
\label{tf:csbc:eq:event-receipt-schema}
\end{align}
A field can be a bounded authenticated digest or typed absence marker. Missing is never silently converted to zero, \tfCsbcNull{}, no event, or no cost.

\subsection{State-machine conformance sketch}

The permitted key states are
\begin{equation}
\begin{aligned}
\textsc{Absent}&\rightarrow\textsc{Detected}\rightarrow\textsc{Owned}
\rightarrow\textsc{Queued}\rightarrow\textsc{Running},\\
\textsc{Running}&\rightarrow\textsc{Retryable}\rightarrow\textsc{Queued},\\
\textsc{Running}&\rightarrow
\{(\textsc{Commit},\nu),\textsc{Null},\textsc{Failed-Terminal}\}.
\end{aligned}
\end{equation}
Transient worker failures advance through \textsc{Retryable} under a renewable bounded lease and an appended attempt receipt. A protocol may choose \textsc{Failed-Terminal} for unrecoverable detector, authorization, integrity, or service errors; that state is not authoritative \tfCsbcNull{}, performs no memory mutation, and cannot be reported as an admission decision. Any later authorized recovery is a new, explicitly linked logical event and key, never reuse of the failed key.

%% file: theory_family/modules/csbc/appendices/C_counterexample_catalogue.tex
\section{Counterexample, Failure, and Audit Catalogue}
\label{tf:csbc:app:counterexamples}

\subsection{Boundary and leakage catalogue}

\begin{table}[H]
\centering
\scriptsize
\begin{tabularx}{\linewidth}{@{}p{0.21\linewidth}p{0.27\linewidth}Y Y@{}}
\toprule
Witness & Violated premise & Observable audit & Required status \\
\midrule
next-token peek & stopping-time boundary & prefix-twin boundary mismatch & \tfCsbcInvalid{} timing result \\
teacher-derived feature cache & graph closure/split independence & provenance ancestor or suffix-swap change & \tfCsbcInvalid{} model \\
late UI rewrite & complete immutable-history inventory & displayed-history hash changes & \tfCsbcFail{} no-retro implementation \\
future-dependent timeout & causal exposure policy & matched queue state, differing suffix changes fallback & \tfCsbcInvalid{} causal decision \\
cross-episode ID collision & episode-bound stable key & same digest under distinct episode authority & \tfCsbcFail{} idempotence scope \\
\bottomrule
\end{tabularx}
\caption{Temporal leakage witnesses and their direct audits.}
\label{tf:csbc:tab:leakage-catalogue}
\end{table}

\subsection{Event and authority catalogue}

\begin{table}[H]
\centering
\scriptsize
\begin{tabularx}{\linewidth}{@{}p{0.21\linewidth}p{0.27\linewidth}Y Y@{}}
\toprule
Witness & Violated premise & Observable audit & Required status \\
\midrule
overlap emits twice & unique owner & same normalized event from two core owners & \tfCsbcFail{} authority/terminal contract \\
wall-time key & stable idempotence identity & retry creates second terminal & \tfCsbcFail{} at-most-once \\
unbounded open quotation & carry cap & serialized carry grows with boundary count & \tfCsbcFail{} boundedness \\
partial row publish & atomic complete-row authority & mixed field/version receipt & severe \tfCsbcFail{} \\
proposal self-assigns ACL & least privilege & adversarial field injection succeeds & severe \tfCsbcFail{} \\
stale worker commit & sequencer/predecessor check & accepted candidate names obsolete root & \tfCsbcFail{} serializability \\
shared archive capacity & commutation independence & both-order decisions differ & order is required treatment \\
semantic poisoning & truth outside mechanics & mechanically valid but false row & semantic/safety \tfCsbcFail{}, mechanics separately reported \\
\bottomrule
\end{tabularx}
\caption{Event, transaction, and authority witnesses.}
\label{tf:csbc:tab:authority-catalogue}
\end{table}

\subsection{Hidden-state and hidden-cost audit}

\tfCsbcPrescriptionStatus\quad A future conformance review inventories all of the following, including zero/absent declarations:
\begin{enumerate}
  \item generator parameters, adapters, optimizer/TTT state, sampler seeds, KV caches, speculative buffers, and visible context;
  \item boundary detector state, overlap buffers, carry, retrospective activations, event queues, candidate buffers, validation caches, and idempotence keys;
  \item authoritative rows, versions, ACL/policy state, tombstones, rollback snapshots, commit roots, and derived indexes;
  \item archives, retrieval indexes, source documents, provenance graphs, audit ledgers, replicas, backups, filesystem/journal overhead, and deletion carriers;
  \item training futures, labels, feature stores, preprocessing caches, split metadata, calibration thresholds, checkpoints, and evaluation state;
  \item CPU/accelerator work, memory/network/storage traffic, barriers, failed/retried work, queue wait, wall time, and energy measurement/proxy.
\end{enumerate}

For each carrier the inventory gives capacity, authority, writer, reader, lifetime, reset, deletion, version, cost unit, and receipt. Calling a carrier ``cache,'' ``metadata,'' or ``temporary'' does not remove its causal or resource role.

\subsection{Worked event trace}

Consider a completed core containing: ``The package is in Rome. Correction: it is in Milan.'' The example is illustrative only.
\begin{enumerate}
  \item Token receipts seal both sentences under prior memory version $\nu_4$.
  \item Boundary $b_n$ closes after the period. The view contains the bounded core and declared overlap.
  \item The eventizer proposes one correction event anchored at the observed token ``Milan'' and key $\iota$.
  \item Trusted assembly checks tenant, object identity, source receipt, ACL, current version, protection, and canonical/neural profile. It may reject with \tfCsbcNull{}.
  \item If accepted in logical event order, the row publishes as $\nu_5$ and receives exposure $a_{n,1}>b_n$.
  \item A later question that pins $\nu_5$ may read Milan. The two earlier sentences and both emission receipts remain unchanged.
\end{enumerate}
The trace demonstrates the contract's chronology, not that an eventizer will recognize the correction or that Milan is true.

%% file: theory_family/modules/csbc/appendices/D_symbols.tex
\section{Symbol, Status, and Evidence Ledger}
\label{tf:csbc:app:symbols}

\subsection{Core symbols}

\small
\begingroup
\setlength\LTleft{\fill}
\setlength\LTright{\fill}
\begin{longtable}{@{}p{0.20\linewidth}p{0.24\linewidth}p{0.48\linewidth}@{}}
\caption{Primary notation.}\label{tf:csbc:tab:symbols}\\
\toprule
Symbol & Type & Meaning \\
\midrule
\endfirsthead
\toprule
Symbol & Type & Meaning \\
\midrule
\endhead
\bottomrule
\endlastfoot
$t,n,(n,j),s,\kappa$ & five clocks & token, boundary, logical event, training-supervision, wall service time \\
$X_t,L_t,\xi_t$ & token/logits/coin & causal sample at token time $t$ \\
$\rho_t^X$ & immutable receipt & binds prefix, logit, coin, token, read version, workspace \\
$b_n,I_n,C_n$ & boundary/core & terminal core index, half-open core, core length \\
$O_n$ & bounded token set & read-only overlap preceding core $I_n$ \\
$q_n\in\tfCsbcCalQ$ & bounded bytes & unfinished cross-boundary detector carry \\
$\beta_n\in\tfCsbcCalB$ & rule artifact & fixed or adaptive causal boundary rule \\
$V_n\in\tfCsbcCalV$ & completed view & overlap, core, and prior carry after $b_n$ \\
$B_\psi$ & frozen map & post-closure bidirectional/retrospective encoder \\
$E_n,e_{n,j},K_n$ & event list/event/count & deterministic ordered events for segment $n$ \\
$\alpha(e)$ & token index & stable event anchor \\
$\tfCsbcOwner(e)$ & boundary index & unique core containing the anchor \\
$\iota(e)$ & authenticated digest & retry/idempotence identity \\
$\mathcal A_{n,j},R_{n,j}$ & routed set/count & authorized candidate targets and their bounded count \\
$\widehat r_{n,j,a}$ & complete row or $\bot$ & untrusted candidate conditioned on routed target $a$ \\
$M_{n,j},\nu$ & bounded state/version & authoritative memory after event $j$ \\
$\tfCsbcNull$ & terminal action & bit-identical authoritative identity transition \\
$a_{n,j}$ & token index & earliest declared exposure of successor event state \\
$\mathcal Q^{\rm svc}$ & bounded queue & asynchronous consolidation jobs \\
$\sigma^{\rm svc}$ & finite service state & detected/owned/queued/running/retryable or terminal receipt \\
$F,Y^{\rm T}$ & training-only & future branches and teacher labels \\
$\theta,\psi$ & frozen parameters & generator and retrospective encoder artifacts \\
$d_h=d/h$ & head dimension & per-head attention width for $h$ heads \\
$P_{\max}$ & positive integer & maximum simultaneously resident candidate/replay/validation jobs \\
$F^{\rm replay}$ & bounded count & reconstruction or replay attempts charged to the resource ledger \\
$\mathbf C$ & nonnegative vector & complete resource ledger \\
\end{longtable}
\endgroup
\normalsize

\subsection{Status ledger}

\begin{table}[H]
\centering
\small
\begin{tabularx}{\linewidth}{@{}p{0.30\linewidth}p{0.24\linewidth}Y@{}}
\toprule
Object & Status in this paper & Boundary \\
\midrule
Typed CSBC timing/system contract & \tfCsbcDefinitionStatus{} & new editorial mathematical construction \\
No-retro, prefix-twin, amputation, ownership, recursion, async, and resource results & \tfCsbcProvedStatus{} & conditional on displayed premises \\
Fixed/adaptive protocols and receipts & \tfCsbcPrescriptionStatus{} & no implementation exists \\
All falsification ladders & \tfCsbcPlannedStatus{} & no stage executed \\
Same-checkpoint semantic interface & \tfCsbcUnvalidatedStatus{} & inherited failed/unqualified gate \\
Event extraction correctness & \tfCsbcUnvalidatedStatus{} & no data or eventizer test \\
Predictive overwrite/admission & \tfCsbcUnvalidatedStatus{} & no oracle gap or learned policy \\
Strict RTT & \tfCsbcUnvalidatedStatus{} & no feedback-driven policy update path \\
CSBC utility, safety, and efficiency & \tfCsbcUnvalidatedStatus{} & no empirical or systems result \\
\bottomrule
\end{tabularx}
\caption{Formal completeness does not promote any empirical status.}
\label{tf:csbc:tab:status-ledger}
\end{table}

\subsection{Frozen numerical boundary}

Only inherited negative/diagnostic counts are admitted:
\begin{itemize}
  \item frozen-checkpoint semantic-interface study: $0/9$ registered interfaces qualified.
  \item fitted-support fault-localization study direct candidate: $40/73{,}728$ exact; direct present: $0$.
  \item Learned route/learned content: $0/73{,}728$.
  \item Oracle route/learned content: $0/73{,}728$.
  \item Learned route/oracle content: $18{,}544/73{,}728$, exactly the routing count.
  \item Oracle route/oracle content: $73{,}728/73{,}728$, a mechanical ceiling only.
  \item Preservation: $0$; missing cases: $43{,}008$.
\end{itemize}
These values delimit what is not qualified. They estimate no CSBC boundary, event, queue, memory transition, semantic accuracy, latency, energy, or cost.

%% file: theory_family/final_synthesis.tex
\clearpage
\section{Dependency and Stop-Rule Synthesis}
\label{sec:integrated-dependency-synthesis}

The five formal parts are complementary contracts, not five independent
sources of empirical support. Their composition is therefore governed by a
conjunctive dependency rule: a downstream system claim is available only when
every upstream scientific gate and every local implementation premise is
satisfied. A theorem proved inside one part cannot replace a failed premise
in another part.

\subsection{Claim--evidence matrix}

The report-wide status ledger is reproduced here so that the detailed basis and
boundary for each headline claim remain available without enlarging the front
matter. Status belongs to the strongest justified statement, not to the
architecture as a whole.

\begin{table}[H]
\centering
\scriptsize
\setlength{\tabcolsep}{3.3pt}
\begin{tabularx}{\linewidth}{lYYl}
\toprule
Claim & Accepted basis & Boundary & Status \\
\midrule
Causal boundary & Causal factorization and completed-segment interface & Formal contract only & \textbf{FORMALIZED} \\
Controlled lifecycle & Version, lock, tombstone, eviction executor & Mechanical ancestor & \textbf{IMPLEMENTED} \\
Lifecycle and typed binding & 300/300 and 240/240 per seed under controls & Controlled grammars & \textbf{VALIDATED} \\
Historical frozen-checkpoint interfaces & External codec, raw representation, matched readers, fitted-support localization & Learned content fails under target-row oracle in those protocols & \textbf{FAILED / UNQUALIFIED} \\
Restricted post-trained readout & Candidate and generated-answer following; update, unchanged-key and restoration controls & Fixed models, prompts, support, and small-event protocols & \textbf{VALIDATED WITHIN SCOPE} \\
Generalized continuous-event readout & Nine completed covered-supervision, bridge, and teacher-aligned trajectories & No latent arm qualifies both families; text and restoration pass & \textbf{NOT MET} \\
Expected-risk overwrite & No registered positive future-risk oracle margin or learned admission comparison & Readout training is not admission learning & \textbf{UNTESTED} \\
Dual-view / AMR formal contract & Complete typed row, trusted assembly, dual-view relation, exact NULL & Formal definitions and transition algebra & \textbf{FORMALIZED} \\
AMR conditional invariants and recovery & Fixed-width, preservation, serializability, tamper, and old/new results & Conditional proofs with explicit assumptions & \textbf{PROVED UNDER STATED ASSUMPTIONS} \\
AMR-1 implementation conformance & Schema, serializer, trusted boundary, crash/security receipts & No implementation qualification bundle & \textbf{UNVALIDATED} \\
Proposed dual-view semantic use & Reconstruction, query, calibration, preservation, and stale-read utility & No qualified same-checkpoint row & \textbf{NOT TESTED / UNVALIDATED} \\
End-to-end cost advantage & No repeated-write or quality--cost crossover & Asymptotic motivation only & \textbf{PLANNED} \\
\bottomrule
\end{tabularx}
\caption{Report-wide claim--evidence status ledger. Formal contracts,
conditional consequences, implementation conformance, semantic usefulness,
validated components, failed interfaces, planned work, and work closed by gate
are kept distinct.}
\label{tab:claim-evidence}
\end{table}

\subsection{Typed dependency chain}

Let $G_{\mathrm{sem}}$ denote qualification of a same-checkpoint semantic row
interface; $G_{\mathrm{row}}$ conformance to the Atomic Memory Row authority
contract; $G_{\mathrm{seg}}$ conformance to the completed-segment timing,
ownership, recursion, and exposure contract; $G_{\mathrm{oracle}}$ a
preregistered, uncertainty-controlled positive future-risk oracle gap;
$G_{\mathrm{policy}}$ a future-blind admission policy passing its registered
fresh-group gates; $G_{\mathrm{rtt}}$ a feedback-driven policy successor read
on a later attempt while the same problem remains unresolved; and
$G_{\mathrm{dual}}$ a receipt-complete policy-by-memory intervention. Then the
claim dependencies used by this report are
\begin{align}
G_{\mathrm{memory\ system}}
  &\Rightarrow G_{\mathrm{sem}}\wedge G_{\mathrm{row}},\\
G_{\mathrm{CSBC\ system}}
  &\Rightarrow G_{\mathrm{sem}}\wedge G_{\mathrm{row}}\wedge G_{\mathrm{seg}},\\
G_{\mathrm{predictive\ overwrite}}
  &\Rightarrow G_{\mathrm{sem}}\wedge G_{\mathrm{row}}\wedge
      G_{\mathrm{{oracle}}}\wedge G_{\mathrm{policy}},\\
G_{\mathrm{dual\ benefit}}
  &\Rightarrow G_{\mathrm{rtt}}\wedge G_{\mathrm{memory\ system}}
      \wedge G_{\mathrm{dual}}.
\label{eq:integrated-dependency-chain}
\end{align}
These implications define necessary gates, not sufficient conditions for
utility, safety, efficiency, or deployment fitness. They preserve the carrier
distinction: $G_{\mathrm{rtt}}$ concerns policy state $\Phi$; the row,
segment, oracle, and admission gates concern authoritative memory $M$ and its
services. A pinned view may coordinate $(\Phi,M)$ without merging ownership,
lifetime, reset, rollback, or ledgers.

\subsection{Current stop frontier}

\begin{table}[H]
\centering
\small
\setlength{\tabcolsep}{4pt}
\begin{tabularx}{\linewidth}{p{0.18\linewidth}p{0.31\linewidth}Yp{0.18\linewidth}}
\toprule
Gate & Required receipt bundle & Stop consequence & Current record \\
\midrule
Semantic interface & Construction, preservation, mapping, query, calibration, stale suppression, and declared generalization under matched models & Restricted success does not imply full authoritative-memory or dual-state benefit & Historical interfaces fail; later restricted readout passes; generalized continuous-event latent readout not met \\
AMR formal contract & Complete typed row, trusted fields, exact NULL, version/lineage, dual-view consistency, recovery, and cost receipts & Formal definitions and transition algebra & FORMALIZED \\
AMR conditional invariants and recovery & Fixed-width, preservation, serializability, tamper, and old/new results & Conditional proofs with explicit assumptions & PROVED UNDER STATED ASSUMPTIONS \\
AMR-1 implementation conformance & Schema, serializer, trusted boundary, crash/security, and measured-cost receipts & No implementation qualification bundle & UNVALIDATED \\
Proposed dual-view semantic use & Reconstruction, query, calibration, preservation, and stale-read qualification & No qualified same-checkpoint row & NOT TESTED / UNVALIDATED \\
Historical latent-row interface & Frozen-checkpoint readers and fitted-support localization & Original negative results remain unchanged by later protocols & FAILED / UNQUALIFIED \\
Completed segment & Prefix-causal boundary, immutable history, unique ownership/key, actual-predecessor recursion, serial exposure, and bounded service & No claim of leakage-free or exactly-once CSBC implementation & Formal contract and planned tests only \\
Expected-risk oracle & Complete supported actions, causal grouped futures, valid evaluator, frozen loss/horizon/margin, simultaneous uncertainty & Do not train or promote a future-blind selector & Unrun; no measured oracle gap \\
Learned admission & Future-blind inputs, calibration, action regret, NULL behavior, preservation, pollution, and fresh-group generalization & Predictive overwrite remains closed & No value model or learned policy \\
Dual-state intervention & Independently pinned carrier variants, compliance, negative controls, endpoint validity, complete cost and severe-event accounting & No complementarity, substitution, neutrality, harm, or benefit claim & Unrun $2\times2$ protocol \\
System comparison & Matched model, context, resident bytes, archive access, attempts, hardware, evaluator, latency, traffic, energy, failures, and recovery & No quality--cost, safety, or superiority claim & Unrun \\
\bottomrule
\end{tabularx}
\caption{Dependency and stop-rule ledger. A failed upstream gate prevents the
corresponding downstream empirical claim.}
\label{tab:integrated-final-stop-ledger}
\end{table}

Under the necessary-condition structure in
Eq.~\eqref{eq:integrated-dependency-chain}, a missing conjunct rules out the
corresponding downstream claim. The cutoff now contains both restricted latent
readout successes and generalized continuous-event failures, not one uniformly
negative semantic record. Neither establishes complete-row conformance,
predictive overwrite, the dual-state intervention, or the completed-segment
system. Those claims remain open empirical work.

%% file: studies/version_and_corrections.tex
\section{Version Identity, Evidence Cutoff, and Explicit Corrections}
\label{app:version-method}

\subsection{Three historical artifacts, not one moving ``latest''}

The document baseline is the complete author-supplied integrated report,
dated September 4, 2026, with 200 letter-size pages and the current
Memory-Mediated Learning Architecture title. Its corresponding LaTeX source
was located, copied into a new document tree, and built before revision.
The rebuilt baseline has 200 pages and byte-identical extracted text to
the supplied PDF. Its historical source Git commit is not established;
content hashes and the complete source/build closure identify it instead.
The source copy, not a short conference draft, is the revision baseline.

Two separately verified public artifacts carry the historical title
\emph{MMLA: How Memory Lets the Past Shape the Future}. The complete
GitHub PDF at the inspected August 17, 2026 repository tree has 196 pages.
That public tree contains PDFs and descriptive files, not corresponding
LaTeX source; a publication-tree commit therefore cannot be represented as
a verified manuscript-source commit. The official arXiv record
\href{https://arxiv.org/abs/2606.28876v3}{arXiv:2606.28876, version 3}
records initial submission on June 27 and revision on August 12, 2026.
Its version-specific PDF has 16 pages. It is a distinct short public
document, not the complete report baseline. A local filename or PDF creation
timestamp does not establish an arXiv publication version or publication date.

The present archival revision is dated September 14, 2026 and is prepared as
the complete source for a replacement of the existing arXiv record. A document
date does not establish submission or announcement: those external states are
recorded separately after they occur. The artifact manifest records exact
document and source-file hashes; the private maintenance map retains historical
paths and evidence bindings without exposing internal task names in the
scientific narrative.

\subsection{A fixed scientific evidence cross-section}

The evidence cutoff is September 13, 2026, 10:11:45 UTC. Completed studies
are included only on the basis of their actual execution and aggregate
records available at that cutoff, cross-checked against source, registered
protocol, and closeout evidence. Later document editing and literature
verification do not move that scientific cutoff. Plans, a clean build,
code inspection, and operational completion are not positive experimental
results. The preparation of this report adds no training, model evaluation,
generated benchmark data, newly opened final holdout, or altered threshold.

Recorded aggregate means and sample standard deviations are checked from
integer counts and stated denominators. Existing interval and significance
outputs retain their original confidence levels, multiplicity treatment,
pairing, and scope; no new resampling replaces a registered decision.
The newly assembled cross-study presentation is retrospective synthesis,
not a prospectively randomized comparison across studies. Its hypotheses
about failure causes are explicitly not unique causal identifications.

\subsection{Preserving the complete scientific record}

The complete theory divisions, expanded proofs, counterexamples, method
sections, positive and negative experiments, figures, protocol details, and
appendices remain in this PDF. The accompanying chapter map and content
inventory audit both the 196-page public report and the 200-page integrated
baseline. Page count is not the sole completeness measure: the audit also
tracks effective text, formal environments, equations, figures, tables,
and stable cross-reference labels. Original body size, paper geometry,
line spacing, and type specification are retained. New space is used for
the actual post-training evidence, original-method comparisons, limitations,
and necessary corrections, not repeated material or blank pages.

Scientific names replace opaque internal task names. Studies are located by
model, intervention, data support, evaluation unit, and date. Original
runtime artifacts are not renamed or rewritten. The accompanying aggregate
table is a neutral presentation of recorded results, not a replacement
for the underlying records. Exact hashes establish identity, not numerical
truth, semantic correctness, or public availability.

\subsection{Correction: adaptedness is not a structural access proof}

The earlier dual-state measurability proof ended by asserting that a version
created later could not be measurable at an earlier event. That assertion
is too strong: a deterministic future value can already be predictable
from the present. Proposition~\ref{tf:dual:prop:measurable} and its expanded
proof now separate two premises. Measurable causal transition maps establish
adaptedness by induction; the independently stated predecessor-access rule
and actual pinned-view trace exclude a later version as a structural parent.
The corrected result does not derive forbidden-access absence from probability
measurability alone, and still does not certify a real implementation.

\subsection{Correction: fixed dimension is not a fixed-bit capacity}

The Gaussian sufficient-statistic example previously called its two real
coordinates a ``bounded pair.'' The dimensionality is fixed, but arbitrary
precision and an unbounded observation count do not satisfy a fixed-bit
contract. Counterexample~\ref{tf:dual:cex:sufficient} preserves that Gaussian
example as an ideal-real-arithmetic comparator and adds an explicit
finite-horizon binary construction. Two bounded counters suffice for its
posterior prediction. This supplies an actual finite-bit example without
claiming exact Gaussian inference at finite precision for arbitrary horizons.

The no-universal-dominance argument also now states a strictly positive
extra resource charge in its redundant-carrier construction. Nonnegative
cost alone may give a tie, not a strict advantage. The converse construction
states the overwritten bit, target, Bayes error, and allowable row cost.
The conclusion is a lack of an architecture-name-only ordering across
unrestricted realizations, not a theorem comparing optimal methods under
one fixed common capacity and cost model.

\subsection{Correction: additive work, peak storage, and elapsed time}

The earlier dual-state cost vector included wall time and retained bytes
inside an unrestricted sum over operations. That is not generally valid
under concurrency or shared storage. Equation~\eqref{tf:dual:eq:ledger-identity}
now applies only to explicitly additive counters with disjoint attribution.
Elapsed makespan and peak simultaneous occupancy are measured separately
from scheduling and allocation traces. For example, two concurrent
one-second operations need not take two seconds, and two reads of one
resident buffer do not create two copies of that buffer. The corrected
accounting result still requires complete costs and does not establish
an efficiency gain. The readout runtime and teacher-cache numbers in this
report are limited measurements, not the unmeasured full PDSA cost vector.

The lifecycle-reduction proof also separates zero changed rows (identity)
from exactly one changed row (complete replacement); an ``at most one''
premise does not itself supply a unique changed row. The coordination digest
is also distinguished from its authenticated envelope: hashing is not signing.

\subsection{Correction: a fixed receipt normal form and reconciled byte ledger}

The earlier AMR-1 table allocated 160 bytes to a receipt while requiring fields
that could not all fit in that allocation. The corrected profile serializes a
96-byte typed predecessor descriptor and a 256-byte receipt. The descriptor
has an explicit normal form for genesis, ordinary replacement, and migration;
unused fields are zero and a kind tag prevents an all-zero string from becoming
a predecessor. The receipt now has explicit allocations for the core digest,
predecessor descriptor, validation-registry digest and bitmap, algorithm and
format identifiers, authorization and event bindings, authentication tag, and
reserved bytes. The unchanged 6,432-byte core therefore yields a 6,688-byte
complete row. All resident-table, working-set, read, write, and publication
totals in the AMR-1 ledger are recomputed from those dimensions. This repairs
an internal serialization contradiction; it is not evidence of an implemented
storage engine or of cryptographic deployment security.

\subsection{Correction: cryptographic tamper evidence does not exclude replay}

The earlier receipt theorem treated every changed accepted row as requiring a
hash collision or authentication forgery. That omitted whole-row substitution:
a previously authenticated row can still verify against its own receipt. The
corrected theorem fixes an expected current validation context and covers a
core-bit change relative to that context. It states separately that replay or
substitution of an intact old row is not a forgery and must be rejected by
namespace, slot, sequence/version, event/authorization, and freshness checks.
Cryptographic integrity and current-context validity are therefore distinct
obligations.

\subsection{Correction: a version pair is weaker than a lineage witness}

The earlier dual-state proposition said that a passing compatibility receipt
need not reveal either lineage, contradicting the definition that the witness
binds exact lineages. The corrected proposition makes the intended claim only
about the nominal pair $(v^{\Phi},v^M)$: that pair alone cannot determine
predecessors, lifetimes, or rollback targets. A verified witness does bind the
exact registered lineages for its view, although it neither merges the two
carrier histories nor authorizes rollback. The revised proof constructs
different histories with the same nominal pair and observes that their valid
witnesses must differ.

\subsection{Corrections to historical reporting, not to experimental records}

The controlled lifecycle split seeds are 202607091--93; its learned writer
seeds are 202607094--96. The earlier reproducibility paragraph did not
distinguish these roles. The secondary generation comparison uses writer
seed 202607094 and must not be counted as three independent generation runs.
For extended-context Llama QA, the recorded dense-baseline F1 is
0.6214590422, which rounds to 0.621, not the previously printed 0.622.
The two extended-context cache-path token counts are now consistently
averaged over all three seeds (1,325 and 1,368), rather than mixing
first-seed token counts with seed-mean F1. These are presentation corrections
from retained aggregates; no question, output, interval, or frozen outcome
was regenerated or reclassified.

\subsection{Availability and reproducibility boundary}

The v4 reconstruction package contains the complete English PDF, corresponding
LaTeX/bibliography/figure assets, a document-only build script, English and
Chinese public descriptions, a neutral aggregate evidence table, and a
file manifest. Document reconstruction is supported from those materials.
Raw model weights, private training and validation payloads, full runtime
logs, credentials, and private review transcripts are not included or
newly published. The upstream model revision names identify public model
origins; they do not license redistribution or prove reproducibility of
every historical scientific execution from this document package alone.

Scientific rerun reproducibility remains distinct from document and
aggregate-arithmetic reproducibility. Where source history or artifact
availability is incomplete, it is reported as not established or not
included, rather than inferred from a filename, hash, or planned release.
No additional experiment is performed by preparing this revision. Submission,
processing, and public announcement are external states and are recorded only
after they occur.